\documentclass{article}

\usepackage{microtype}
\usepackage{graphicx}
\usepackage{subcaption}
\usepackage{tabularx}
\usepackage{booktabs} 
\usepackage{lipsum} 
\usepackage{hyperref}
\usepackage{enumitem}
\usepackage{xurl}
\PassOptionsToPackage{hyphens}{url}\usepackage{hyperref}

\newcommand{\sw}[1]{\textcolor{red}{SW: #1}}

\usepackage[accepted]{icml2026}
\usepackage[T1]{fontenc}
\usepackage{amsmath}
\usepackage{amssymb}
\usepackage{mathtools}
\usepackage{amsthm}
\usepackage[utf8]{inputenc}
\usepackage{inconsolata} 
\usepackage{listings}
\usepackage{xcolor}
\usepackage{twemojis}

\usepackage{pifont}
\definecolor{backcolour}{rgb}{0.96,0.96,0.97}
\definecolor{codegreen}{rgb}{0,0.6,0}
\definecolor{codegray}{rgb}{0.5,0.5,0.5}
\definecolor{codepurple}{rgb}{0.58,0,0.82}
\definecolor{codemagenta}{rgb}{0.8,0.0,0.4}

\lstdefinestyle{mystyle}{
    backgroundcolor=\color{backcolour},   
    commentstyle=\color{codegreen},
    keywordstyle=\color{codemagenta},
    numberstyle=\tiny\color{codegray},
    stringstyle=\color{codepurple},
    basicstyle=\ttfamily\footnotesize, 
    breakatwhitespace=false,         
    breaklines=true,                 
    captionpos=b,                    
    keepspaces=true,                 
    numbers=left,                    
    numbersep=8pt,                  
    showspaces=false,                
    showstringspaces=false,
    showtabs=false,                  
    tabsize=2,
    frame=lines, 
    rulecolor=\color{black!30}, 
    framesep=5pt, 
    xleftmargin=18pt,   
    xrightmargin=2pt,  
    framexleftmargin=18pt,    
}

\lstdefinestyle{promptstyle}{
    basicstyle=\ttfamily\small,
    breaklines=true,
    frame=single,
    backgroundcolor=\color{gray!10},
    rulecolor=\color{gray!60},
    columns=fullflexible,
    keepspaces=true,
    aboveskip=1em,
    belowskip=1em,
    extendedchars=true,
    literate={_}{\_}1
             {❌}{{\textcolor{red}{\ding{55}}}}1
             {✓}{{\textcolor{green}{\ding{51}}}}1
             {±}{{$\pm$}}1
             {≥}{{$\geq$}}1
             {→}{{$\to$}}1
}

\usepackage[capitalize,noabbrev]{cleveref}

\theoremstyle{plain}

\theoremstyle{definition}

\theoremstyle{remark}

\usepackage[textsize=tiny,colorinlistoftodos]{todonotes}
\renewcommand{\sw}[1]{}

\icmltitlerunning{Procedural Generation of Algorithm Discovery Tasks in Machine Learning}

\begin{document}

\twocolumn[
  \icmltitle{Procedural Generation of Algorithm Discovery Tasks in Machine Learning
  }



  \icmlsetsymbol{first}{*}
  \icmlsetsymbol{core}{$\dagger$}
  \icmlsetsymbol{sup}{$\mathsection$}
  \icmlsetsymbol{task}{$\ddagger$}

  \begin{icmlauthorlist}
    \icmlauthor{Alexander D. Goldie}{first,ox}
    \icmlauthor{Zilin Wang}{core,ox}
    \icmlauthor{Adrian Hayler}{core,ox}
    \icmlauthor{Deepak Nathani}{core,ucsb}
    \icmlauthor{Edan Toledo}{core,ucl}
    \icmlauthor{Ken Thampiratwong}{core,ucsb}
    \icmlauthor{Aleksandra Kalisz}{core,ox}
    \icmlauthor{Michael Beukman}{task,ox}
    \icmlauthor{Alistair Letcher}{task,ox}
    \icmlauthor{Hannah Erlebach}{task,ox}
    \icmlauthor{Shashank Reddy Chirra}{task,ox}
    \icmlauthor{Clarisse Wibault}{task,ox}
    \icmlauthor{Theo Wolf}{task,ox}
    \icmlauthor{Charles O'Neill}{task,ox}
    \icmlauthor{Uljad Berdica}{task,ox}
    \icmlauthor{Nicholas Roberts}{task,uwm}
    \icmlauthor{Saeed Rahmani}{task,ox,delft}
    \icmlauthor{Roberta Raileanu}{sup,ucl}
    \icmlauthor{Shimon Whiteson}{sup,ox}
    \icmlauthor{Jakob N. Foerster}{sup,ox}

  \end{icmlauthorlist}

  \icmlaffiliation{ox}{University of Oxford}
  \icmlaffiliation{ucsb}{University of California, Santa Barbara}
  \icmlaffiliation{ucl}{University College London}
  \icmlaffiliation{uwm}{University of Wisconsin--Madison}
  \icmlaffiliation{delft}{Delft University of Technology}

  \icmlcorrespondingauthor{Alexander D. Goldie}{goldie@robots.ox.ac.uk}

  \icmlkeywords{Machine Learning, ICML}

  \vskip 0.3in
]



\printAffiliationsAndNotice{$^*$Lead Author, $^\dagger$Core Contributor, $^\ddagger$Task Contributor, $^\mathsection$Equal Supervision}  


\begin{abstract}
  Automating the development of machine learning algorithms has the potential to unlock new breakthroughs. However, our ability to \textit{improve} and \textit{evaluate} algorithm discovery systems has thus far been limited by existing task suites. They suffer from many issues, such as: poor evaluation methodologies; data contamination; and containing saturated or very similar problems. Here, we introduce \textit{DiscoGen}, a procedural generator of algorithm discovery tasks for machine learning, such as developing optimisers for reinforcement learning or loss functions for image classification. Motivated by the success of procedural generation in reinforcement learning, DiscoGen spans billions of tasks of varying difficulty and complexity from a range of machine learning fields. These tasks are specified by a small number of configuration parameters and can be used to optimise algorithm discovery agents (ADAs). We present \textit{DiscoBench}, a fixed, small subset of DiscoGen tasks for principled evaluation of ADAs. Finally, we propose a number of ambitious, impactful research directions enabled by DiscoGen, and demonstrate its use for ADA optimisation through scaling experiments for automated prompt tuning. DiscoGen is released \href{https://github.com/AlexGoldie/discogen/}{open-source}.
\end{abstract}

\section{Introduction}


Automating the development of machine learning (ML) algorithms with AI offers the potential to unlock new breakthroughs in research. Furthermore, since algorithm discovery agents (ADAs) can alleviate the bottleneck of human ideation, implementation and experimentation, their utility scales directly with computational resources.


However, existing ADA benchmarks (e.g., MLE-Bench \citep{chan_mle-bench_2025} and MLGym-Bench \citep{nathani2025mlgymnewframeworkbenchmark}) suffer from structural problems that inhibit principled evaluation. They generally fail to separate the \textit{discovery} (meta-train) and \textit{evaluation} (meta-test) of algorithms, meaning ADAs discover algorithms for the same problems they are evaluated on. Additionally, they often require agents to write entire codebases, effectively measuring software engineering rather than research skills, or initialise from full file systems, biasing agents away from discovering novelty \citep{nathani2025mlgymnewframeworkbenchmark}. Finally, these benchmarks run the risk of data contamination from pre-training \citep{dong-etal-2024-generalization, liang_swe-bench_2025}; ADAs may have learned from the fixed task sets during pre-training, and thus change their behaviour or use previously seen problem solutions to compensate for poor research skills \citep{liang_swe-bench_2025}.

Furthermore, our ability to develop better ADAs for ML remains limited, principally because there are \textit{too few} different algorithm discovery problems to learn from. These existing suites of tasks for algorithm discovery are constrained due to a reliance on manual creation. Therefore, developing new approaches and architectures for existing suites, or training ADAs on them, risks overfitting.



\begin{figure*}[h]
    \centering
    \includegraphics[width=0.99\linewidth]{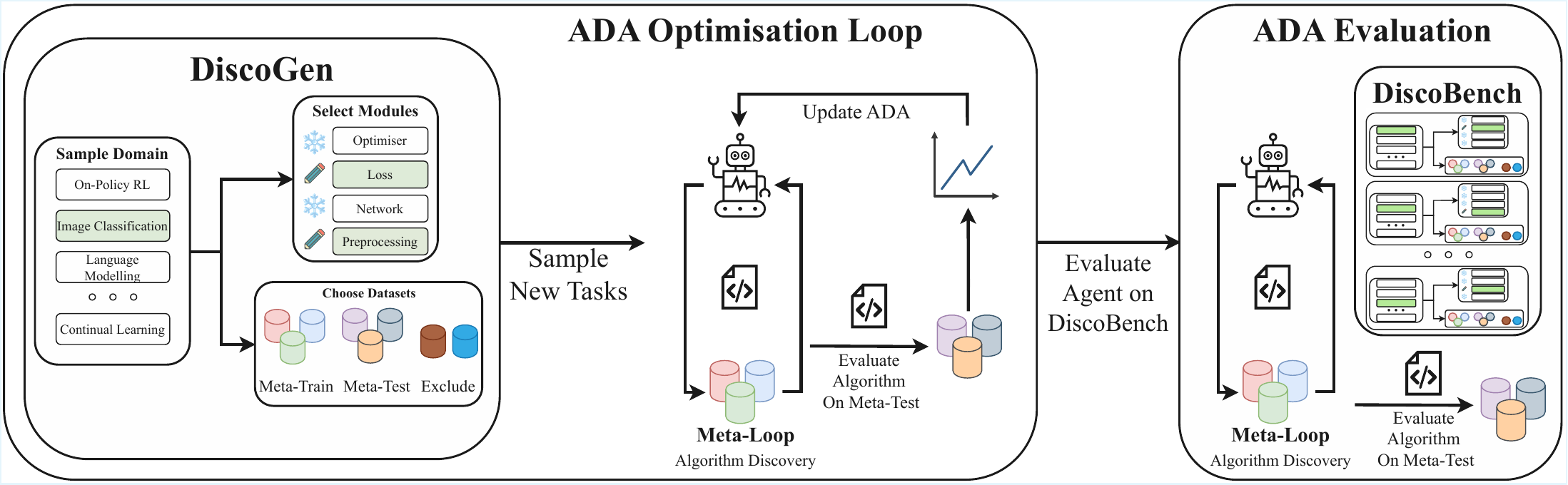}
    \vspace{-5pt}
    \caption{A typical DiscoGen setup. DiscoGen procedurally generates new algorithm discovery tasks. For every generated task, an algorithm discovery agent iteratively develops new algorithms (the meta-loop) for training in the task's meta-train datasets (the inner-loops). The developed algorithm is evaluated on meta-test datasets, with the evaluation score used to optimise the agent (the ADA optimisation-loop). Datasets that are available in a task domain can also be excluded from the task. After each step, DiscoGen can be sampled for a new task. After ADA optimisation has completed, the agent is evaluated on DiscoBench, a set of ADA test tasks.\vspace{-14pt}}
    \label{fig:discogen}
\end{figure*}

To address these issues, we introduce DiscoGen, a procedural generator of algorithm discovery tasks for ML. DiscoGen supports \textit{billions} of different tasks, of varying difficulty. DiscoGen tasks have distinct meta-train/meta-test datasets, where meta-test datasets are hidden from the ADA, ensuring principled evaluation. Furthermore, DiscoGen supports many diverse ML subfields and uses a modular structure that defines \textit{which components} of an algorithm an ADA discovers, meaning tasks vary over a number of axes. 

DiscoGen enables the use of an \textit{ADA optimisation loop}, as shown in Figure \ref{fig:discogen}. Thus, we establish our terminology as:
\begin{itemize}[leftmargin=*] \vspace{-6pt}
\item \textbf{Inner-loop:} An algorithm optimises a specific model on a single dataset's train set. When the inner-loop finishes, the model is evaluated on the dataset's test set. For example, the inner loop could be training an image classifier on the ImageNet train set \citep{russakovsky2015imagenet}, and evaluating it on the ImageNet test set.\vspace{-5pt}
\item \textbf{Meta-loop:} The ADA iteratively improves the algorithm based on inner-loop feedback from its \textit{meta-train} set, which contains many inner-loop datasets. When the meta-loop finishes, the final algorithm is evaluated on a held-out meta-test set. An example meta-loop could involve an ADA developing image classifier loss functions, based on feedback from ImageNet and CIFAR-10 \citep{krizhevsky2009learning} (meta-train), and evaluating the loss by training an image classifier on CIFAR-100 (meta-test).\vspace{-5pt}
\item \textbf{Task:} A single algorithm discovery problem, defining the ADA's objective and the meta-train and meta-test datasets.
\vspace{-16pt}
\item \textbf{ADA Optimisation loop:} The ADA is optimised for meta-loop performance in DiscoGen tasks in a \textit{meta-meta-}loop \citep{Schmidhuber1987EvolutionaryPI}. The ADA optimisation loop produces an ADA for evaluation. ADA optimisation could update ADA weights based on meta-test performance.\vspace{-5pt}
\item \textbf{ADA Evaluation}: The ADA is evaluated on a set of tasks that it has not been directly optimised for (in other words, \textit{meta-meta-}test). This could involve developing a language model optimiser, for instance.\vspace{-9pt}\end{itemize}



Much as procedural environment generation enabled training of general reinforcement learning agents \citep{cobbe_leveraging_2020, team_open-ended_2021, team_human-timescale_2023,matthews_kinetix_2025}, DiscoGen enables new research directions in algorithm discovery. Ideas like autocurricula \citep{leibo_autocurricula_2019,dennis_emergent_2021,parker-holder_evolving_2022} or recursive self-improvement \citep{clune_ai-gas_2019} \textit{rely} on an ability to sample new, interesting tasks of varying difficulty to prevent overfitting. Given the number of possible tasks in DiscoGen, it is a crucial tool for enabling open-ended learning for algorithm discovery \citep{stanley2019open,hughes_open-endedness_2024}.

We create DiscoBench, a set of hand-designed tasks from DiscoGen, for ADA evaluation.
Similar to \citet{matthews_kinetix_2025} or \citet{samvelyan2021minihack}, which build task suites from procedural environment generators, DiscoBench lies in the support of DiscoGen but should \textit{not} be intentionally optimised over. While DiscoBench, like other task sets, is susceptible to data contamination, it benefits from the design of DiscoGen, such as a meta-train/meta-test distinction. Since our ADA evaluation was run prior to publication, it is not subject to this contamination, and we propose mitigations in Section \ref{sec:conclusion} to overcome this in the future.

We provide further research proposals enabled by DiscoGen in Section \ref{sec:new_research}. Finally, as an example of ADA optimisation, we use procedurally generated tasks from DiscoGen for prompt tuning an ADA in \cref{sec:prompts}. We explore how ADA evaluation changes when their prompts are optimised over different numbers of \textit{randomly generated} tasks, finding that prompts developed over a wider range perform better; both in-support, and for completely held out domains. \vspace{-3pt}


\vspace{-6pt}\section{Related Work} \label{sec:related_work}\vspace{-5pt}

Due to the depth of prior work in this field, we abridge our related work discussion here and expand it in Appendix \ref{app:related_work}. 
\vspace{-6pt}
\subsection{Automated Research}\vspace{-4pt}
Automated machine learning \citep[AutoML]{hutter2019automated} focuses on applying machine learning to new problems without expert knowledge. Prior research generally augments machine learning algorithms with methods like hyperparameter tuning (e.g., \citep{li_hyperband_2018, parker-holder_provably_2021}) or data-cleaning (e.g., \citep{activeclean2016krishnan}). However, whereas most AutoML research is limited to fitting human-designed solutions to new data, algorithm discovery has the inverse goal: autonomously developing new algorithms.
\begin{table*}[t]
    \centering
    \caption{Overview of selected desirable properties for a number of existing algorithm discovery tasks. \twemoji{orange circle} denotes partial satisfaction.}
    \label{tab:benchmark_flaws}
    \resizebox{0.83\textwidth}{!}{%
    \begin{tabular}{lrccccc}
        \toprule
        & & \multicolumn{5}{c}{\textbf{Task Property}} \\
        \cmidrule(lr){3-7}
        \textbf{Benchmark} & \textbf{\# Tasks} & \textbf{Meta-Test} & \textbf{Flexible Init.} & \textbf{Low Contam. Risk} & \textbf{Tunable Diff.} & \textbf{Editable Evals}\\
        \midrule
        MLGym-Bench & 13 & \twemoji{cross mark} & \twemoji{cross mark} & \twemoji{cross mark} & \twemoji{cross mark} &\twemoji{cross mark}\\
        MLAgentBench & 13 & \twemoji{cross mark}& \twemoji{cross mark} & \twemoji{cross mark} & \twemoji{cross mark} &\twemoji{cross mark}\\
        MLE-Bench & 75 & \twemoji{cross mark} & \twemoji{cross mark} & \twemoji{cross mark} & \twemoji{orange circle} &\twemoji{cross mark}\\
        AIRSBench & 20 & \twemoji{cross mark}& \twemoji{cross mark} & \twemoji{orange circle} & \twemoji{orange circle} &\twemoji{cross mark}\\
        REBench & 7 & \twemoji{cross mark}& \twemoji{cross mark} & \twemoji{orange circle} & \twemoji{orange circle} &\twemoji{cross mark} \\
        \midrule
        \textbf{DiscoGen (Ours)} & \textbf{$\sim$100B} & \twemoji{check mark button} & \twemoji{check mark button} & \twemoji{orange circle} & \twemoji{check mark button} & \twemoji{check mark button}\\
        \bottomrule
    \end{tabular}%
    }\vspace{-15pt}
\end{table*}

That said, meta-learning is a subset of AutoML which aims to \textit{learn} algorithms from data \citep{Schmidhuber1987EvolutionaryPI, real2020automl, beckSurveyMetaReinforcementLearning2023a}. Often, meta-learned algorithms train a neural network to replace a component of a machine learning algorithm, such as the optimiser \citep{andrychowiczLearningLearnGradient2016, metzVeLOTrainingVersatile2022, goldie_can_2024} or loss function \citep{Kirsch2020Improving, bechtleMetaLearningLearnedLoss2021}. Recently, using large language models (LLMs) to propose new algorithms has proven fruitful \citep{lu_discovering_2024, romera-paredes_mathematical_2024, novikov_alphaevolve_2025}. However, optimising and evaluating these systems is difficult due to a lack of diverse, interesting and well-designed tasks. In this paper, we consider how procedural generation can be used to create new algorithm discovery tasks to this end.

As language models have improved \citep{measuring-ai-ability-to-complete-long-tasks,chollet_arc_2025}, developing more complex research and coding \textit{agents} has emerged as an important pursuit. Agents augment language models with the ability to take actions, use tools and run code \citep{wang_survey_2024,schick_toolformer_2023, yang2024sweagent}. AI research agents \citep{toledo_ai_2025} use these tools to automate parts of the research process in a ReAct loop \citep{yao_react_2023}, where they receive feedback from tools while developing solutions. Rather than focusing on designing new agents, we build a framework for sampling millions of tasks to aid their development.

Agents can be applied throughout the research pipeline. Examples include automating ideation \citep{si_can_2024}, implementing research ideas with a human-in-the-loop \citep{gottweis_towards_2025,weston_ai_2025}, judging research papers \citep{si_can_2024,thakkar_can_2025}, or automating the entire research workflow \citep{lu_ai_2024, yamada_ai_2025, zochi2025, si_towards_2026}. Specifically in algorithm discovery, considerations include how agents should search over new algorithms \citep{jiang_aide_2025,toledo_ai_2025} or when to run experiments \citep{yu_alpharesearch_2025, nathani2025mlgymnewframeworkbenchmark}. In this work, we provide a rigorous and scalable generator of tasks for optimising and evaluating ADAs.

\vspace{-7pt}
\subsection{Optimising Agents}

\vspace{-3pt}




Pretraining language models on more data leads to improved performance \citep{kaplan_scaling_2020}, and large procedurally generated environments have led to generalist reinforcement learning (RL) policies (Section \ref{sec:procgen}). Motivated by these findings, we consider how ADAs can be optimised using procedurally generated algorithm discovery tasks.

Optimising agents specifically for mathematics (e.g., \citep{lewkowycz_solving_2022,trinh_solving_2024, hubert_olympiad-level_2025}) or coding (e.g., \citep{alphacode2022, roziere_code_2024}) has led to significant gains. Underpinning these advances are large, diverse and verifiable problem sets to research for or train models on \citep{shao_deepseekmath_2024, wen_reinforcement_2025}. For example, there are many suites of mathematical problems (e.g., \citep{cobbe2021gsm8k, hendrycksmath2021}) or open-source code repositories \citep{jimenez_swe-bench_2024,chen2021codex}. However, developing similarly large sets of algorithm discovery tasks has proven difficult, as manual curation requires expert knowledge and, often, adaptation to integrate different data. Additionally, developing superhuman algorithms requires measuring `\textit{how good}' algorithms are, rather than correctness. Here, we mitigate these limitations of manual task creation using procedural generation.\vspace{-6pt}
\subsection{Procedural Generation} \label{sec:procgen}\vspace{-3pt}

Procedural content generation (PCG) involves creating levels or environments algorithmically, according to rules, rather than manually \citep{togelius_procedural_2013}. To do so, PCG environments are defined as Contextual Markov Decision Processes \citep[Contextual MDPs]{hallak_contextual_2015} or Underspecified Partially Observable MDPs \citep{dennis_emergent_2021} which define levels by a small number of configuration variables. In deep RL, PCG has proven effective for training agents to generalise over a smooth \textit{distribution} of levels, rather than solving specific levels only. Such approaches have been applied to environments of ranging complexity, from gridworlds \citep{MinigridMiniworld23} to physics engines \citep{matthews_kinetix_2025} or 3-dimensional worlds \citep{team_open-ended_2021}. Procedural generation enables new research directions, such as autocurricula (e.g., \citep{jiang_prioritized_2021, dennis_emergent_2021}) or large scale meta-learning (e.g., \citep{team_human-timescale_2023,nikulin_xland-minigrid_2024}). We explore how to apply similar principles to algorithm discovery.
\vspace{-6pt}

\vspace{-1pt}\section{The Problems With ADA Benchmarks} \label{sec:existing}
\vspace{-3pt}

ADA improvement is bottlenecked by current algorithm discovery benchmarks. 
They are heavily limited in scale due to a reliance on manual task creation. Task suites like MLGym-Bench \citep[13 tasks]{nathani2025mlgymnewframeworkbenchmark}, MLAgentBench \citep[13 tasks]{huang_mlagentbench_2024}, MLE-Bench \citep[75 tasks]{chan_mle-bench_2025}, AIRSBench \citep[20 tasks]{lupidi_airs-bench_2026} and REBench \citep[7 tasks]{wijk_re-bench_2025} assess ADA performance, but none provide enough tasks to optimise ADAs over, nor separate ADA evaluation from optimisation. Ideally, as in PCG for RL, we want to optimise ADAs over a smooth \textit{distribution} of tasks to robustly develop algorithm discovery capabilities. Furthermore, these benchmarks suffer from flaws which limit their value for evaluation.

\vspace{-8pt}
\subsection{Issues With Existing Task Designs} \label{sec:flaws}\vspace{-4pt}

Beyond limited scope, we believe task design in these benchmarks is insufficient. Here, we discuss a number of their structural flaws, which DiscoGen rectifies (Section \ref{sec:advantages}).
\vspace{-8pt}
\paragraph{Poor Evaluation} Proper evaluation in machine learning relies on a distinct train/test split \citep{Goodfellow-et-al-2016} to avoid overestimating performance. Algorithm discovery is no different; despite not fitting a \textit{model} to test data in the \textit{inner-loop}, hill-climbing algorithms on meta-train datasets is susceptible to the same flawed evaluation as humans using validation signals to design methods \citep{langford2005clever,ADPRL11-shimon}. However, existing benchmarks miss the proper train-test boundary. Rather than measuring algorithm transfer from \textit{meta}-train to unseen \textit{meta}-test datasets, they evaluate the performance of inner-loop trained models on the test set of the (known) meta-train datasets (e.g., \citep{nathani2025mlgymnewframeworkbenchmark, chan_mle-bench_2025}). In effect, testing an algorithm on the dataset it was developed on, rather than how well it generalises. Whilst this can be valid evaluation in certain settings, it is not generally the objective in algorithm discovery \citep{goldie_how_2025}. We demonstrate the importance of meta-test evaluation in \cref{sec:discobench_results}, where algorithms often perform worse in meta-test over meta-train.\vspace{-8pt}

\paragraph{Limited Diversity} Due to their limited scale, benchmarks are often restricted to similar classes of problems, such as small Kaggle-style challenges \citep{chan_mle-bench_2025} or quick-to-run problems \citep{nathani2025mlgymnewframeworkbenchmark}. As such, rather than understanding the general performance of ADAs, they measure their ability in specific \textit{types} of problems only.\vspace{-8pt}

\paragraph{Limiting Initialisation} While in-context examples help elicit reasoning in LLMs \citep{wei_chain--thought_2023}, they can limit output diversity \citep{turpin_language_2023}. Many benchmarks only initialise tasks from full implementations, potentially limiting the creativity of agents; in \citet{nathani2025mlgymnewframeworkbenchmark}, agents devolve to hyperparameter tuning. Similarly, starting from empty files can sometimes prove too difficult for agents, limiting the potential optimisation signal.\vspace{-8pt}

\paragraph{Slow Manual Expansion} Adding \textit{every} new task to these suites is manual, meaning scaling their quantity is inherently limited by the number of human-hours available. \vspace{-8pt}

\paragraph{Data Contamination} Data contamination, where evaluation data leaks into training, is an issue in LLM benchmarks due to large-scale pretraining \citep{dong-etal-2024-generalization}. It can be especially problematic in `challenge-based' benchmarks, like MLE-Bench. Since agents can often use the internet, or may have pre-trained on challenges, ADAs can reproduce public solutions to boost their score \citep{cheatingai2025hamin}. Such logic can be extended to other contamination types, like agents seeing open-sourced dataset labels. While limiting internet access is a mitigation (e.g., \cite{yang_programbench_2026}), it is significantly more advantageous to design benchmarks that are as robust as possible instead, such that we evaluate a true agent's capabilities.\vspace{-8pt}

\paragraph{Floor and Ceiling Effects} Many suites include \textit{saturated} or \textit{solved} problems (e.g., \citep{huang_mlagentbench_2024}) with hard-to-change difficulties, limiting their signal for optimisation.\vspace{-8pt}

\vspace{-3pt}\section{DiscoGen} \label{sec:discogen}


DiscoGen generates \textbf{tasks}: modular algorithm discovery problems consisting of an objective and meta-train and meta-test datasets. Each task is defined by seven components.\vspace{-3pt}

\paragraph{1. Task Domain} The task domain is the area of machine learning a task pertains to (e.g., \textit{Image Classification}). It defines the initial codebase, performance metrics, and which datasets and modules are available.\vspace{-8pt}

\paragraph{2. Editable Modules} For each domain, we identify important building blocks, or \textit{modules}, of an algorithm that can be set as \textit{editable} or \textit{fixed}. Only \textit{editable} modules are to be discovered by the agent; \textit{fixed} modules use standard implementations (e.g., a baseline optimiser or loss). There can be many modules per domain, combinatorially expanding the number of possible tasks that can be generated.\vspace{-8pt}

\paragraph{3. Meta-Training Datasets} Meta-training datasets are the problems which an agent can run experiments on during algorithm discovery. They are known to the agent.\vspace{-8pt}

\paragraph{4. Meta-Test Datasets} Meta-test datasets are used to evaluate an algorithm \textit{after} the meta-loop completes. They measure algorithm-transfer to held-out problems, and are unknown to the ADA.\vspace{-8pt}

\paragraph{5. Backend} Some task domains include extra \textit{backends} that expand the task space. For example, in On-Policy RL, tasks can require either a feed-forward or recurrent policy.\vspace{-8pt}

\paragraph{6. Evaluation Type} DiscoGen supports different meta-objectives: final score (maximise performance); energy, in kWh, to reach a proportion of the baseline's score (maximise efficiency); or time taken, in seconds, to reach a proportion of the baseline's score (maximise speed, like \citet{zhao2025the,kellerjordan2024moddednanogpt}). In this paper we focus on performance only, due to resource constraints.\vspace{-8pt}

\paragraph{7. Initialisation} Tasks are set to either `\textit{empty}', where only the interface of each editable module is provided (to ensure implementations fit the codebase), or `\textit{baseline}', where editable modules start from baseline algorithms. We compare these settings in Section \ref{sec:baselines}.\vspace{-8pt}

\subsection{An Example Task}

We use Example \ref{fig:example_config} to demonstrate the task interface. For this on-policy RL task, the ADA will work in JAX \citep{bradbury_jax_2018} to maximise the performance of an RL agent. The majority of the codebase is \textit{fixed}: domain-specific code like wrapper functions, environment creation, and the optimiser, training loop, target and activation modules. While ADAs \textit{can} interact with these files, changes are overwritten prior to meta-testing to prevent evaluation hacking. 

The ADA must work with two \textit{editable} files: the loss, which starts as an empty function mapping inputs (e.g., data batches) to a scalar loss; and the network architecture, which maps environment observations to an RL policy, value function and recurrent state, and also contains no logic. The ADA can make arbitrary code edits, such as writing additional functions, importing extra packages, and filling templates.  The agent can \textit{run} inner-loop training and evaluation for both meta-train environments. To ensure security during agent execution, the ADA operates in a containerised environment. If one editable module is \textit{only} called by another, the ADA can edit its interface, expanding the search-space of algorithms in DiscoGen tasks further.

\begin{lstlisting}[language=Python, caption={An example task configuration. A task is defined by its task domain, meta-train/meta-test datasets, backend and modules.}, label={fig:example_config}]
task_domain = "OnPolicyRL"
meta_train = ["Ant", "Freeway"]
meta_test = ["Pusher", "Craftax"]
backend = "recurrent"
change_optim = False
change_loss = True
change_networks = True
change_train = False
change_target = False
change_activation = False
eval_type = "performance"
initialisation = "empty"
\end{lstlisting}\vspace{-12pt}
\subsection{Procedural Task Generation} \label{sec:domains}
\vspace{-5pt}
DiscoGen is a procedurally generated benchmark of ML tasks. As a generator, it is designed for \textit{improving} ADAs, manually or automatically in an ADA optimisation loop (\cref{fig:discogen}). A task is specified by a small configuration file, like Example \ref{fig:example_config}, which takes the same role as the parameters in other PCG environments \citep{cobbe_leveraging_2020, team_open-ended_2021}. This can be randomly generated, user-specified, or sampled using autocurricula. The technical details of sampling tasks in DiscoGen are described in Appendix \ref{app:technical_implementation}.\vspace{-1pt}

DiscoGen creates tasks in a two-stage process to reduce the risk of meta-test leakage. The meta-train portion of the task is generated first; only \textit{after} the meta-loop is complete does DiscoGen create the meta-test codebase. As such, the agent is never given any details of the meta-test datasets.\vspace{-1pt}

PCG involves creating levels, or tasks, programmatically. DiscoGen is no different; the number of tasks for a domain is combinatorial with respect to how many modules and datasets it supports. Specifically, for a task domain with $m$ modules, $d$ datasets, and $b$ backends, and our currently supported $3$ evaluation types and $2$ task initialisations, \vspace{-5pt}
\begin{equation}
    N_{tasks} = 2 \cdot 3 \cdot b \cdot (2^m-1)\cdot\left(3^d-2^{(d+1)}+1\right).
\end{equation}\vspace{-19pt}

Derived in Appendix \ref{app:derivation}, this assumes at least one editable module and assigns each dataset to either meta-train, meta-test or unused (with at least one meta-train and meta-test).

\vspace{-5pt}
\subsection{Available Task Domains}\vspace{-3pt}

Table \ref{tab:task_domains} shows the domains in DiscoGen, which we expect to grow from open-source contributions;
adding new domains needs only mild adaptation of existing codebases. Due to imbalanced task counts, we recommend stratified sampling over domains to reduce bias, which is supported by the DiscoGen library. We describe each domain, its modules and datasets in Appendix \ref{app:domains}.
\begin{table}[h!]
\centering
\caption{Overview of domains and their number of supported tasks.\vspace{-5pt}}
\label{tab:task_domains}
\begin{small}
\setlength{\tabcolsep}{4pt}
\begin{tabularx}{1\columnwidth}{@{}Xcccr@{}}
\toprule
\textbf{Task Domain} & \textbf{$m$} & \textbf{$d$} & \textbf{$b$} & \textbf{$N_{tasks}$} \\
\midrule
Bayesian Optimisation          & 6 & 11 & 1 & 65,413,656 \\
Brain Speech Detection          & 3 & 7  & 1 & 81,144 \\
Computer Vision Classification  & 4 & 9  & 1 & 1,679,400 \\
Continual Learning             & 5 & 3  & 3 & 6,696 \\
Greenhouse Gas Prediction       & 2 & 4  & 1 & 900 \\
Language Modelling             & 3 & 4  & 2 & 4,200 \\
Model Unlearning\footnotemark  & 1 & 3  & 1 &  85,176 \\
Neural Cellular Automata       & 5 & 5  & 1 & 33,480 \\
Off-Policy RL                   & 7 & 4  & 1 & 38,100 \\
Offline RL                   & 5 & 10  & 1 & 10,602,372 \\
On-Policy MARL               & 6 & 17  & 2 & 97,431,783,120 \\
On-Policy RL               & 6 & 13 & 3 & 1,789,383,960 \\
Trajectory Prediction          & 4 & 3 & 3 & 1,080 \\
Unsupervised Environment Design & 3 & 4  & 1 & 2,100 \\
\bottomrule
Total & & & & 99,299,115,384 \\
Median & & & & 59,622
\end{tabularx}
\end{small}\vspace{-6pt}

\end{table}
\footnotetext{Model unlearning entails finetuning pretrained models. For $n$ models, $N_{tasks} = 2\cdot3\cdot b \cdot (2^m-1)\cdot\left((2n+1)^d-2(n+1)^{d}+1\right)$} 
\vspace{-12pt}

DiscoGen exhibits useful diversity across these millions of tasks; in \cref{sec:prompts}, we show that ADA optimisation improves as more DiscoGen tasks are experienced. In \cref{sec:onpolicy} we demonstrate how performance varies across different editable module combinations; as the number of editable modules lowers the ADA's success rate. The domains supported in DiscoGen span a range of machine learning fields, incorporate datasets of varying complexity and difficulty, and are built upon codebases of differing scales. Most task domains include unique modules, and when there is overlap, implementations are domain-specific. 

We further validate this diversity through \textit{rank correlation analysis} over the performance of different ADAs in DiscoBench (\cref{sec:discobench_results}), a subset of DiscoGen tasks, in Figures \ref{fig:all_data_corr} \& \ref{fig:split_corrs} (Appendix \ref{app:correlation}). Hierarchical clustering of the correlation matrix reveals distinct patterns; while correlation is often high between similar modules in different domains or different modules in the same domain, there are also anti-correlations where strong performance in one task implies poor performance in another, including within the same domain. The Fisher-Z transformed mean Spearman correlation is $\sim0.4$; high enough to indicate non-random signal, while sufficiently small to show low redundancy in DiscoGen. Notably, we also find that clustering patterns are \textit{distinct} between meta-train and meta-test, demonstrating how the \textit{same algorithm} can rank differently across datasets. 

Per-dataset analysis in each task reinforces the meaningfulness of DiscoGen's large task space. Considering Appendix \ref{app:discobench_results}, the ranking of the discovered algorithms changes across datasets within the \textit{same task}. This is intuitive, given prior literature suggests optimal algorithms differ between datasets or RL environments (e.g., in reinforcement learning \citep{goldie_can_2024, jackson2025clean}, computer vision \citep{rodrigo_comprehensive_2024, takahashi_comparison_2024} or language modelling \citep{transformersssms2024dao, repeatafter2024jelassi}).
\vspace{-8pt}
\subsection{Advantages of DiscoGen}\label{sec:advantages}
\vspace{-3pt}
Individual task design in DiscoGen also overcomes the many flaws of previous task definitions raised in \cref{sec:flaws}.\vspace{-9pt}

\paragraph{Principled Evaluation \& Contamination Resistance} DiscoGen tasks clearly distinguish between meta-train and meta-test. Since DiscoGen is procedural, and there is no knowledge of the meta-test datasets in meta-training, the potential for test leakage is limited. Even as DiscoGen enters pre-training datasets, this is a step towards fairer evaluation, ensuring DiscoGen remains pertinent for a long time. Furthermore, DiscoGen supports different evaluation \textit{types}, enabling discovery for factors other than performance.\vspace{-8pt}

\paragraph{High Diversity} DiscoGen generates highly diverse tasks. As demonstrated in \cref{sec:domains}, DiscoGen supports a range of domains with different data structures, filesystem complexities and modules. Since DiscoGen tasks are parameterised combinatorially, its tasks represent a smooth range of difficulties between ``\textit{implement a single module for one easy dataset}'' to ``\textit{implement many modules for many hard datasets}'', rather than simply \textit{easy} or \textit{hard} subsets.\vspace{-8pt}

\paragraph{Different Initialisations} Agents need not implement full codebases, and the initialisation of editable modules can be set to just the inputs and outputs of the module (empty) or full  baseline implementations. In essence, tasks are either `\textit{improve a baseline}' or `\textit{de novo discovery}'. This enables better analysis of the biases elicited by ADAs, can make tasks easier or harder, and may increase agent creativity.\vspace{-9pt}

\paragraph{Ease of Adding Tasks} Beyond our currently implemented domains, adding many tasks to DiscoGen is significantly easier than for other suites. For similar effort to adding one task to, say, MLE-Bench \citep{chan_mle-bench_2025} or MLGymBench \citep{nathani2025mlgymnewframeworkbenchmark}, DiscoGen can gain potentially millions of new tasks in its support. When the base code for a task domain is complete, adding more tasks is even easier; isolating a new module effectively doubles the number of possible tasks, and adding a new dataset near-triples it.

\paragraph{Unsaturated Problems} Since tasks can be made more or less difficult by changing the module and dataset configurations (e.g., Section \ref{sec:onpolicy}), DiscoGen tasks span a wide range of difficulties. We find that the more modules there are to implement, the harder the problem but the higher the potential ceiling. Additionally, almost all datasets currently in DiscoGen have yet to be solved by humans, let alone agents, and adding more, harder datasets is straightforward.\vspace{-7pt}

\vspace{-2pt}\section{ADA Evaluation: DiscoBench} \label{sec:benchmark}\vspace{-5pt}

Despite the contamination risk that arises from releasing a fixed public task suite, there is still merit to evaluating ADA performance over a set of DiscoGen tasks that resolve the other flaws from prior benchmarks (\cref{sec:flaws}).
\begin{table*}[t]
\centering
\caption{ADA evaluation performance in DiscoBench (Elo Scores with 95\% CIs). Bold indicates best mean performance.\vspace{-7pt}}
\label{tab:elos}
\resizebox{0.91\textwidth}{!}{%
\begin{tabular}{@{}lccccccccc@{}}
\toprule
& \multicolumn{3}{c}{\textit{DiscoBench (Single Edit)}} & \multicolumn{3}{c}{\textit{DiscoBench (All Edit)}} & \multicolumn{3}{c}{\textbf{\textit{DiscoBench}}} \\
\cmidrule(lr){2-4} \cmidrule(lr){5-7} \cmidrule(lr){8-10}
\textbf{Model} & \textbf{Succ.} & \textbf{Meta-Train} & \textbf{Meta-Test} & \textbf{Succ.} & \textbf{Meta-Train} & \textbf{Meta-Test} & \textbf{Succ.} & \textbf{Meta-Train} & \textbf{Meta-Test} \\
\midrule
Baseline (All Fixed)      & --- & \textbf{1111} \tiny{[1088, 1133]} & \textbf{1133} \tiny{[1114, 1153]} & --- & \textbf{1443} \tiny{[1340, 1621]} & \textbf{1388} \tiny{[1264, 1575]} & --- & \textbf{1129} \tiny{[1109, 1150]} & \textbf{1144} \tiny{[1120, 1166]} \\
GPT-OSS 120B  & 59.5\% & 907 \tiny{[885, 931]}             & 944 \tiny{[924, 966]}             & 17.8\% & 596 \tiny{[366, 712]}             & 595 \tiny{[142, 758]}             & 52.0\% & 892 \tiny{[871, 914]}             & 931 \tiny{[911, 955]} \\
Devstral2     & 55.2\% & 963 \tiny{[932, 986]}             & 924 \tiny{[900, 945]}             & 39.3\% & 790 \tiny{[634, 890]}             & 1010 \tiny{[909, 1127]}        & 52.4\% & 948 \tiny{[928, 969]}             & 928 \tiny{[909, 951]} \\
Deepseek-v3.2 & 71.3\% & 1019 \tiny{[997, 1042]}        & 999 \tiny{[978, 1020]}          & 40.5\% & 1171 \tiny{[1090, 1320]}        & 1007 \tiny{[926, 1174]}           & 65.7\% & 1031 \tiny{[1013, 1049]}        & 998 \tiny{[974, 1016]} \\
\bottomrule
\end{tabular}%
}\vspace{-13pt}
\end{table*}

DiscoBench is a manually curated ADA evaluation set akin to hand-designed levels built in PCG environments (e.g., \citep{matthews_kinetix_2025, nikulin_xland-minigrid_2024}). For each domain in DiscoGen, we create $m+1$ tasks; $m$ tasks where each of the $m$ modules is active (\textit{Single Edit}), and $1$ where all modules are active simultaneously (\textit{All Edit}). DiscoBench is the union of these sets; we separate them here for analysis only. We do not include other module combinations to ensure DiscoBench stays manageable, and to enable principled expansion with new domains. Meta-train and meta-test sets are fixed between tasks to enable comparison, and are selected pseudo-randomly; long-to-run datasets are reserved for meta-testing, for computational reasons. These splits are included in Appendix \ref{app:task_list}. ADAs should not be optimised on DiscoBench to ensure it is an appropriate test suite (though the probability of sampling tasks from DiscoBench is non-zero, as in other PCG environments). Since DiscoBench is not yet public, our evaluation is not subject to contamination; however, this work entering the public domain exposes it to pretraining or internet search agents. As such, we plan to release a private DiscoBench `API' using datasets not mentioned publicly.
\vspace{-9pt}
\subsection{Experimental Setup}\vspace{-5pt}

We explore the performance of different LLMs with the MLGym ADA \citep{nathani2025mlgymnewframeworkbenchmark}, a ReAct agent \citep{yao_react_2023} which can run code, read files and choose when to submit, within a fixed action budget. Due to resource constraints, and for reproducibility purposes, we evaluate three open-source language models: Deepseek-v3.2 \citep{deepseek-ai_deepseek-v32_2025}, Devstral2 \citep{mistralIntroducingDevstral} and GPT-OSS 120B \citep{openai_gpt-oss-120b_2025}. We include an `\textit{all fixed}' baseline (i.e., the code with no editable modules) for comparison; it is \textit{always} possible for the ADA to implement this. We provide experimental details and hyperparameters in Appendix \ref{app:experimental_details}, include our generic ADA system prompt in Appendix \ref{app:systemprompt}, and detail cost and compute usage in Appendix \ref{compute}.

In both \textit{DiscoBench Single} and \textit{DiscoBench All}, we aggregate scores over three seeds per task and model. We report two per-model success-rates and Elo ratings \citep{elo1978rating} based on same-dataset comparisons; one for meta-train, and one for meta-test. We report 95\% confidence intervals for Elo, estimated using 100 bootstrap samples as in Appendix \ref{app:experimental_details}. Since agents frequently fail to consistently produce valid solutions for many tasks\footnote{Since at least one model produced a valid solution for every task, we have an existence proof that all tasks are solvable.}, we penalise failure such that a model with more successful runs dominates one with fewer; this penalty does not apply in baseline comparisons. We also report the total success rate for each model.

To understand how models could perform without failure, we run extra experiments on \textit{DiscoBench (Single Edit)} until each model has three successful attempts; we include results in Appendix \ref{app:until_success}. Due to low success rates, this was unaffordable for \textit{DiscoBench (All Edit)} and a small number of tasks in \textit{DiscoBench (Single Edit)}. This failure is expected; in \citet{nathani2025mlgymnewframeworkbenchmark}, many frontier closed source models failed over their four attempts in MLGym-Bench tasks. These tasks are omitted from the \textit{Until Success} analysis.

\vspace{-9pt}
\subsection{Results} \label{sec:discobench_results}\vspace{-5pt}

We report results in \cref{tab:elos}, and include a per-task score breakdown in Appendix \ref{app:discobench_results}. To demonstrate that agents can discover interesting and performant algorithms, we discuss two hand-selected algorithms in Appendix \ref{app:example_algos}.

Firstly, success rates in \textit{All Edit} are significantly lower than for \textit{Single Edit}, confirming the hypothesis that including more editable modules increases task difficulty. This is explored further in Section \ref{sec:onpolicy}, where we sweep over all module combinations in On-Policy RL and find that the success rates of ADAs \textit{consistently} fall as more editable modules are added. In fact, average success rates for all three models are low. In contrast, we examine \textit{Success@3} rates (i.e., what proportion of tasks had \textit{at least one} successful solution from 3 attempts) in Appendix \ref{app:pass_k_results}, and find that they are 10-30 percentage points higher than the aggregated rates. Considering this performance gap, and that \textit{all} baselines follow the same interface as the editable modules, it is clear that agents struggle to \textbf{robustly} produce even simple, well-known algorithms. We find that failures are broadly driven by syntax errors or, often, code overfitting to the meta-train datasets (e.g., hardcoded shapes).

Elo shows a similar pattern; the baseline has a much higher score in \textit{All Edit} than in \textit{Single Edit}. Even when agents have three successful solutions, no agent consistently outperforms the baseline to the point of having a higher Elo (Appendix \ref{app:until_success}). Since ADAs do not yet match well-known human algorithms, even though they \textit{could} be implemented and are often \textit{suboptimal} algorithms for the datasets, there is clearly a significant margin for ADAs to improve.

Comparing ADAs, Deepseek-v3.2 performs well in \textit{DiscoBench (Single Edit)}, both in meta-train and meta-test, as well as on aggregate in the full \textit{DiscoBench} set. However, due to currently low success rates, drawing conclusions for \textit{DiscoBench (All Edit)} is difficult. Overall, the relative baseline performance increases between meta-train and meta-test, suggesting there are signs of meta-overfitting for ADAs; there is clearly space for ADA improvement.

It is important to ensure DiscoBench is diverse; a single model being uniformly dominant would suggest DiscoBench only measures general ability. We explore this using rank-correlation analysis in Appendix \ref{app:correlation} and find high variation in rankings between tasks. In fact, the per-task results (Appendix \ref{app:discobench_results}) show that even the ranking over datasets within the \textit{same} task varies, demonstrating how different algorithms are better for different datasets and justifying claims of diversity in DiscoGen. Furthermore, this per-task breakdown confirms the range of task difficulties; sometimes agents outperform the baseline, usually they produce weak-but-valid solutions, and often they fail completely.

\vspace{-9pt}
\section{Enabling New Research} \label{sec:new_research}\vspace{-3pt}

In addition to enabling discovery of better algorithms, by extracting discovered artifacts, DiscoGen is a platform for a plethora of new research directions. To serve as inspiration to the wider research community, we propose some ideas here. In \cref{sec:prompts}, we show how DiscoGen can be used for prompt optimisation, demonstrating one such use-case.

\vspace{-6pt}
\subsection{Understanding The Pathologies of ADAs}\vspace{-4pt}

Our ability to analyse the pathologies of algorithm discovery systems is hindered by the design and limited controllability of existing benchmarks. DiscoGen provides an expansive space in which to run such analysis. For example, research could seek to understand creativity differences \citep{haase_has_2025, franceschelli_creativity_2025} as task initialisation changes, the limits of instruction-following in complex file-systems \citep{zeng2024evaluating, ouyang_training_2022}, or whether ADAs are biased towards certain domains or modules (e.g., they might be better at designing optimisers than losses). This enables more scientific development of ADAs.\vspace{-8pt}

\subsection{Learning to Discover Algorithms}\vspace{-5pt}

LLM reasoning has significantly improved since the introduction of RL \citep{shao_deepseekmath_2024, ouyang_training_2022, gehring_rlef_2025, kazemnejad_vineppo_2025} or evolution \citep{sarkar_evolution_2025, qiu_evolution_2025} post-training. Prior work generally focuses on mathematics or programming, where there are many verifiable problems to learn from. As DiscoGen enables sampling of \textit{billions} of unique algorithm discovery tasks, findings from these other task-rich domains can be transferred towards training better ADAs. Such \textit{meta-meta-learning} could optimise for efficient, quick, or performant algorithms, or all three.\vspace{-8pt}

\subsection{Sampling Hard-Yet-Learnable Discovery Tasks}\vspace{-5pt}

Prior work has shown how autocurricula \citep{dennis_emergent_2021, parker-holder_evolving_2022, foster2025lilo} can sample hard-yet-learnable tasks to improve minimum expected performance bounds \citep{beukman2024Refining} and improve training efficiency \citep{foster2025lilo}. Since tasks in DiscoGen are of varying difficulty, it is naturally suited to curriculum methods, and their application could improve the performance and efficiency over random task sampling. This is especially important in ADA optimisation, where task completion times can vary by orders of magnitude.\vspace{-8pt}

\subsection{Algorithm World Models}\vspace{-5pt}

\citet{team_cwm_2025} train a ``Code World Model'' to replicate a code interpreter's state as it ran, to improve an LLM's programming abilities. However their work necessitates vast amounts of data. Combining DiscoGen with similar computational resources could enable collection of a similarly large-scale, structured dataset of algorithm-performance pairs. This could be used to train an `\textit{algorithm world model}' (AWM) that predicts an algorithm's performance. Such a model could be used directly in an ADA, fine-tuned as above, or integrated into tree-search agents as below.\vspace{-9pt}

\subsection{Training LLM-As-A-Judge in Tree-Search ADAs}\vspace{-5pt}

``\textit{How to explore?}'' is an open question in algorithm discovery and AI research \citep{toledo_ai_2025}. Some agents designed for automated research and algorithm discovery use tree-search \citep{jiang_aide_2025, toledo_ai_2025}, but evaluating a tree's leaves requires running expensive inner-loop trainings. Instead, using an LLM or otherwise trained model to evaluate algorithms could act as a filter, selecting promising leaves to run \citep{yu_alpharesearch_2025, herr_llm-first_2025} or acting as a value function \citep{wang_mcts-judge_2025} or reward model \citep{zhang_lessons_2025} to skip inner-loop training and reduce the cost of search.\vspace{-2pt}

However, off-the-shelf judge performance would likely be poor, since we ideally want to evaluate super-human (and thus, out-of-distribution) algorithms. Using data generated using DiscoGen, it would be possible to train either a full model or a value prediction head that could be integrated into Monte-Carlo Tree Search \citep{banditbases_2006_levente}.\vspace{-9pt}

\subsection{Symbolic Evolution of Algorithms}\vspace{-6pt}

Since DiscoGen provides templates for module (i.e., defining inputs and outputs), and the rest of an algorithm's code is pre-implemented, it is well-placed for working with non LLM-based algorithm discovery methods; for example, symbolic search or black-box meta-learning. This could include developing methods using genetic programming, as in \citep{ramachandran_searching_2017, zheng2022symbolic, chenSymbolicDiscoveryOptimization2023a,goldie_how_2025}, or black-box evolution (e.g., \citep{lu_discovered_2022, metzVeLOTrainingVersatile2022, goldie_can_2024}).\vspace{-9pt}

\subsection{Self-Improving ADAs}\vspace{-6pt}\label{sec:self_improve}

An alternative to \textit{designing} ADAs is letting them build their \textit{own} scaffolds (e.g., \citep{hu_automated_2024, zhang_darwin_2025, wang_huxley-gomachine_2025, zhang_hyperagents_2026}), for data-driven open-ended self-improvement. However, optimising an agent's scaffold on existing algorithm discovery suites could lead to overfitting. With DiscoGen, such systems could be run near-indefinitely with low risk of overfitting to individual tasks, enabling the discovery of super-human ADAs.\vspace{-8pt}

\vspace{-3pt}

\section{ADA Optimisation Using DiscoGen} \label{sec:prompts}
\vspace{-5pt}
\begin{table*}[t]
\centering
\caption{ADA evaluation performance after prompt optimisation (Elo Scores with 95\% CIs). Bold indicates the best (point-estimate) Elo.\vspace{-2pt}}
\label{tab:results_prompting}
\resizebox{0.69\textwidth}{!}{
\begin{tabular}{@{}lcccccc@{}}
\toprule
& \multicolumn{3}{c}{\textit{In-Distribution Domains}} & \multicolumn{3}{c}{\textit{Held-Out Domains}} \\
\cmidrule(lr){2-4} \cmidrule(lr){5-7}
\textbf{$K_{tasks}$} & \textbf{Succ.} & \textbf{Meta-Train} & \textbf{Meta-Test} & \textbf{Succ.} & \textbf{Meta-Train} & \textbf{Meta-Test} \\
\midrule
1   & 69.9\% & 955 \tiny{[931, 980]}   & 965 \tiny{[940, 990]}   & 65.0\% & 979 \tiny{[947, 1012]}   & 983 \tiny{[954, 1011]} \\
5   & 74.2\% & \textbf{1063} \tiny{[1038, 1087]} & 978 \tiny{[955, 1002]}  &   68.5\%     & 985 \tiny{[952, 1018]}   & 994 \tiny{[966, 1022]} \\
10  & 71.6\% & 935 \tiny{[911, 961]}   & 991 \tiny{[967, 1014]}  & 72.4\% & \textbf{1031} \tiny{[998, 1065]} & 1000 \tiny{[973, 1028]} \\
30  & 75.2\% & 1047 \tiny{[1023, 1071]} & \textbf{1067} \tiny{[1042, 1093]} & 72.3\% & 1005 \tiny{[971, 1037]} & \textbf{1024} \tiny{[995, 1052]} \\
\bottomrule
\end{tabular}
}\vspace{-15pt}
\end{table*}

Given the impact of prompting on LLM performance \citep{lester-etal-2021-power, promptbreeder2024}, we optimise an ADA's prompt for DiscoGen tasks. We query an `ADA-Optimisation' LLM, distinct from the ADA, to iteratively update the prompt in the \textit{ADA optimisation loop}, based on past meta-train and meta-test performances in sampled DiscoGen tasks. Its objective is to optimise \textbf{meta-test performance}.
Since the `prompting' LLM is queried infrequently, it uses a more expensive closed-source model; Claude Sonnet 4.5 \citep{anthropic_claude_sonnet45_2025}. As the best performing LLM tested in DiscoBench, the ADA uses DeepSeek-V3.2.\vspace{-2pt}

We explore how task quantity, $K_{tasks}$, correlates to ADA optimisation performance over $30$ updates. We sample from DiscoGen at different frequencies; when $K_{tasks}=1$, we tune the prompt on the same task $30$ times, and when $K_{tasks}=30$, we use a different task each iteration.
To prevent bias, we uniformly sample task domains \textit{before} task configurations. To test if ADA optimisation in DiscoGen improves ADA generalisation, we hold out a small number of domains during ADA optimisation (Appendix \ref{app:init_hparams}); in practice, all domains should be used to maximise the breadth of experience.
Results are presented in \cref{tab:results_prompting}, using DiscoBench for ADA evaluation and the Elo methodology from \cref{sec:discobench_results}, with additional analysis in Appendix \ref{app:prompt_tuning}. We include the prompt-tuner prompt in Appendix \ref{app:sysprompts}.

Our results suggest a positive relationship between increasing $K_{tasks}$ and meta-test performance (i.e., the ADA optimisation objective): as task count increases, ADA performance improves. While this pattern is strongest for the in-distribution domains (but still unseen tasks), the monotonic trend holds in the held-out set, which are out-of-support for the optimised ADAs. Such results demonstrate the value of procedural generation for ADA optimisation; given $K_{tasks}$ can still be dramatically scaled, with additional resources, ADA performance can likely be improved much further. Interestingly, this pattern does \textit{not} exist for meta-train (which we do not optimise for), meaning conventional evaluations would have misrepresented the ADA performance.

Considering the prompts themselves (Appendix \ref{app:discovered_prompts}), there is a noticeable progression from $K_{tasks}=1$ to $K_{tasks}=30$. Whereas lower task counts over-index on specific tasks, prompts developed over more tasks emphasise broader discovery and machine learning principles.

\vspace{-8pt}\section{Analysing DiscoGen}\vspace{-4pt}

Here, we analyse the design of DiscoGen for the three ADAs from Section \ref{sec:discobench_results}. Task redundancy and the meta-train/meta-test performance gap are analysed in Appendix \ref{app:correlation}.\vspace{-6pt}

\subsection{Changing Modules}\label{sec:onpolicy}\vspace{-4pt}

\begin{figure}[h]
\begin{center}
    \includegraphics[width=0.77\linewidth]{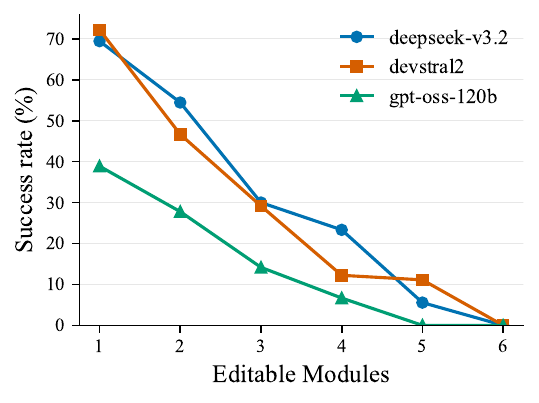}
    \end{center}
    \vspace{-12pt}
    \caption{Success rate vs. editable module count in On-Policy RL.}\label{fig:change_correlation}\vspace{-10pt}
\end{figure}

Using PCG in DiscoGen provides a smooth distribution of tasks to optimise over, potentially enabling autocurricula methods. To verify this smoothness, we evaluate ADAs for a sweep over the \textit{editable} modules in On-Policy RL domain. We measure the success rate for each LLM over every possible combination of 6 modules (i.e., $2^6-1=63$ tasks), for a fixed meta-train/meta-test split, in Figure \ref{fig:change_correlation}. \vspace{-7pt}

For every model, as the number of editable modules \textit{increases}, the success rate \textit{decreases}. As such, tasks in DiscoGen can smoothly be made more or less difficult simply by increasing or decreasing its number of editable modules.\vspace{-8pt}

\subsection{Using a Different Initialisation}\label{sec:baselines}\vspace{-4pt}

DiscoGen enables different task initialisations; either \textit{empty} files, which only provide a ``function interface'' to the agent, or starting from \textit{baseline} implementations. While we experiment with \textit{empty} initialisations through most of this paper, we compare the two settings here to verify that \textit{empty} initialisations are more difficult than \textit{baseline} ones. Due to computational limitations, we restrict these experiments to one LLM (Deepseek-v3.2) and a set of quick-to-run DiscoGen domains, as set out in Appendix \ref{app:init_hparams}.

We find that success rates for starting from \textit{baseline} ($92.4\%$) are significantly higher than for \textit{empty} initialisations ($75.4\%$), and Elo scores are higher ($1101$ vs $899$, respectively). Given all else is kept equal, baseline initialisation clearly makes tasks easier. Furthermore, a non-exhaustive, qualitative analysis of the discovered algorithms from the \textit{baseline} experiments shows that they often do remain close to the original implementation.

\vspace{-8pt}\section{Conclusion}\label{sec:conclusion}\vspace{-5pt}

In this paper, we introduced DiscoGen, a procedural generator of algorithm discovery tasks. We motivated the design of DiscoGen by the shortcomings of existing algorithm discovery suites, such as poor evaluation methodology and limited scale. We demonstrated that DiscoGen overcame many of these flaws, and established DiscoBench, an ADA evaluation set. We subsequently introduced a number of possible research avenues enabled by DiscoGen. As demonstration of the potential of DiscoGen, we used it for ADA prompt optimisation and showed that prompt performance improved with the number of tasks it was optimised for; both within and beyond the support of the optimisation distribution. Finally, we analysed the design of DiscoGen, showing how different axes of task variation affect difficulty.\vspace{-6pt} 
\paragraph{Future Work} Beyond the research \textit{enabled} by DiscoGen (\cref{sec:new_research}), we highlight a number of avenues for future work. Firstly, DiscoGen is non-exhaustive; expanding its support of domains, modules and datasets, via open-source contributions, would increase its utility. Additionally, ADA evaluation is still affected by contamination (though less than other benchmarks). Implementing an API-only DiscoBench suite, with datasets not available in DiscoGen, could solve this issue. Testing better models and scaffolds in DiscoBench would provide a greater understanding of the current ADA ceiling. Finally, our ADA optimisation experiments only focus on prompt tuning, a relatively limited form of training, due to cost constraints; exploring scaffold- or weight-based improvement is an obvious next step.


\section*{Acknowledgements}

\textbf{AG} is funded by the EPSRC Centre for Doctoral Training in Autonomous Intelligent Machines and Systems EP/S024050/1. \textbf{ZW} is funded by a generous grant from Waymo. \textbf{AK} is supported by Exscientia and the SABS CDT. \textbf{CW} is funded by the EPSRC DTP Research Studentship. \textbf{TW} is funded by the EPSRC Centre for Doctoral Training in Autonomous Intelligent Machines and Systems EP/Y035070/1. \textbf{UB} is supported by the EPSRC Centre for Doctoral Training in Autonomous Intelligent Machines and Systems and the Rhodes Scholarship. \textbf{NR} is supported by the Defense Advanced Research Projects Agency (DARPA). Our experiments were made possible by an equipment grant from NVIDIA. \textbf{SR} is funded by the Transport and Mobility Institute at Delft University of Technology. \textbf{JF} is partially funded by the UKRI grant EP/Y028481/1, which was originally selected for funding by the ERC. \textbf{JF} is also supported by the JPMC Research Award and the Amazon Research Award. This project received compute resources from a generous grant provided by the Isambard-AI National AI Research Resource, for the projects "Robustness via Self-Play RL" and "FLAIR Summer Moonshots".


The authors thank \textbf{Jonathan Cook} and \textbf{Edward Hughes} for their comments and advice in this project.

\section*{Contributions}

\textbf{Alexander D. Goldie} led the project, created and designed DiscoGen, ran all experiments, wrote the paper and implemented the On-Policy RL task domain.

\textbf{Zilin Wang} contributed to the design of DiscoGen and implemented the Computer Vision Classification and Brain Speech Detection task domains.

\textbf{Adrian Hayler} contributed to the design of DiscoGen and implemented the Language Modelling task domain.

\textbf{Deepak Nathani} contributed to the design of DiscoGen and the structure of the DiscoGen repository.

\textbf{Edan Toledo} contributed to the design of DiscoGen and helped edit the paper.

\textbf{Ken Thampiratwong} contributed to the design of DiscoGen.

\textbf{Aleksandra Kalisz} contributed to the design of DiscoGen.

\textbf{Michael Beukman} implemented the Unsupervised Environment Design task domain and helped edit the paper.

\textbf{Alistair Letcher} implemented the Model Unlearning task domain.

\textbf{Hannah Erlebach} implemented the Neural Cellular Automata task domain.

\textbf{Shashank Reddy Chirra} implemented the Off-Policy RL and Multi-Agent RL task domains, and contributed additional modules for the On-Policy RL task domain.

\textbf{Clarisse Wibault} implemented the Bayesian Optimisation task domain.

\textbf{Theo Wolf} implemented the Greenhouse Gas Prediction task domain.

\textbf{Charles O'Neill} implemented the Continual Learning task domain.

\textbf{Uljad Berdica} implemented the Offline RL task domain.

\textbf{Nicholas Roberts} implemented the state-space model backend for the Language Modelling task domain.

\textbf{Saeed Rahmani} implemented the Trajectory Prediction task domain.


\textbf{Roberta Raileanu}, \textbf{Shimon Whiteson} and \textbf{Jakob N. Foerster} provided equal supervision over the course of the project.

\section*{Impact Statement}\label{app:safety}\vspace{-6pt}

Considering the social, economic and safety effects of automated algorithm discovery is crucial due to the potential magnitude of its impacts. In this work, we are considering not only how a specific algorithm can be meta-learned (which has more limited impact), as in much prior work, but how to \textit{optimise} agents towards the development of new machine learning algorithms. While we believe there are a range of significant benefits that such work can lead to, we are also wary and considerate of the risks.\vspace{-2pt}

It is often preferable to develop models with specialist capabilities; consistently focusing on generalist improvements can introduce safety concerns \citep{bommasani_opportunities_2022,weidinger_ethical_2021}, and may not even be the most effective way to enhance desired capabilities \citep{belcak_small_2025}.  We believe algorithm discovery provides a surgical tool with which to automate research, offering a much more specific objective than `\textit{automating science}', as in a lot of automated AI scientist literature \citep{lu_ai_2024, zochi2025, yamada_ai_2025}. Rather than giving black-box agents the ability to decide what problems they work on freely, algorithm discovery (and specifically, DiscoGen) is designed to involve a human-in-the-loop; discovery agents are only allowed to make changes to certain files, with \textit{all} other changes being overwritten at test time. Furthermore, despite the massive task space available in DiscoGen, all tasks are well-defined, scoped and based on human-selected codebases, meaning they are constrained. Given the analysed suite of 14 domains, we do not believe any tasks deemed unsafe \textit{could} be sampled from DiscoGen. However, since we hope for community contributions in the future, it is important to ensure that this remains the case moving forward.\vspace{-2pt}

As we emphasise optimisation for algorithm discovery \textit{only}, DiscoGen provides a platform for AI-Human Co-Improvement \citep{weston_ai_2025}. Crucially, rather than training towards more generally-able agents (where algorithm discovery capabilities are improved simultaneously with adverse capabilities), optimising on DiscoGen improves performance \textit{directly} for human-designed algorithm discovery tasks. We believe this type of scoped optimisation is a necessary path towards safe super-human agents; it is safer to develop specifically super-human agents than developing poorly understood general superintelligence.\vspace{-2pt}

It is important to recognise that development of automated algorithm discovery systems can pose ethical and economic issues. While we focus on general and beneficial task domains, harmful actors could instead look to optimise agents in unethical domains. Fully mitigating this misuse is beyond the scope of our paper, but it is important to ensure DiscoGen is comprised only of broadly beneficial domains to limit any negative behaviours. Similarly, discovering better machine learning algorithms automatically enables them to be used for harmful datasets; while this is a risk of all machine learning research, we must acknowledge that DiscoGen accelerates their development. At the least, to reduce the risk of directly meta-learning on these datasets, it is important to ensure there is broad human-alignment in foundation models, making it hard to optimise for misaligned behaviour \citep{bai_constitutional_2022}. A counter-measure to these risks is to ensure DiscoGen \textit{supports} AI safety tasks. For example, discovering algorithms for the \textit{Model Unlearning} task in DiscoGen \citep{yuanshun_unlearning_2024} is one avenue for reducing unwanted behaviours in foundation models.\vspace{-2pt}

Ensuring people from affected fields are kept in-the-loop as these systems develop is necessary for ensuring automated algorithm systems \textit{complement}, rather than \textit{replace}, humans. The potential impact of AI on the labour market is significant \citep{doi:10.1126/science.adj0998}, and thus developing systems which protect the role of humans is important. We believe that automated algorithm discovery provides a strong balance of developing research breakthroughs using AI, while maintaining the agency and empowerment of humans who design the problem settings for it to operate in. In general, this fulfils the idea that research agents are most effective when provided as tools to humans \citep{gottweis_towards_2025, shneiderman_human-centered_2022}, rather than acting to substitute them. \vspace{-2pt}

Moreover, the requirements for operating in this field are high; running or querying large language models is expensive, and optimising research agents is more so. This introduces large financial \citep{chen_frugalgpt_2023} and environmental \citep{strubell_energy_2019,faiz_llmcarbon_2024} costs, which should be considered when experimenting with DiscoGen. One high-impact area for future research would involve developing energy-efficient automated algorithm discovery systems; in fact, these could be developed automatically, using a well-defined objective in combination with the research proposal of Section \ref{sec:self_improve}. Furthermore, using \textit{energy} as the evaluation criterion to discover new, hyper-efficient algorithms could offset the costs of running ADAs, given the potential downstream savings enabled by such algorithms.\vspace{-2pt}

Democratisation of AI is one necessary tool for ensuring it can act in the benefit of all, rather than a few major players. We believe DiscoGen helps with this in two ways. Firstly, in making a large-scale research environment for algorithm discovery available, we believe that such research is made more feasible outside of industrial frontier labs. Secondly, by emphasising the development of \textit{specialist}, rather than \textit{generalist}, agents, we hope to enable a landscape of research using smaller models which require less pre-training data. However, it would be na\"{i}ve to suggest that research in automated science and algorithm discovery is not expensive, or doesn't require significant hardware resources. Moving forward, it is necessary to ensure the development of these tools is accessible to a wide range of researchers, from academia through to small and large parts of industry.

\bibliography{main}
\bibliographystyle{icml2026}
\newpage
\appendix

\onecolumn

\section*{Appendix}

Our appendix is structured as follows:
\begin{itemize}
    \item Appendix \ref{app:domains} provides an overview of each task domain included in DiscoGen. We provide a high-level overview of the goal of each domain, its implementation details, and what editable modules and datasets it supports.
    \item Appendix \ref{app:related_work} extends the abridged related work from the main text (Section \ref{sec:related_work}) to include discussion of the wider field.
    \item Appendix \ref{app:experimental_details} provides hyperparameters and experimental details for our paper. It also includes a description of our Elo calculation and discussion of the experimental compute and cost for the paper.
    \item Appendix \ref{app:implementation} introduces implementation details of DiscoGen. We discuss how to derive the expression for its task count, and how DiscoGen creates new tasks when sampled from.
    \item Appendix \ref{app:task_list} provides a breakdown of what the meta-train and meta-test datasets are for each domain in DiscoBench.
    \item Appendix \ref{app:correlation} examines the redundancy of tasks in DiscoBench, using average-rank correlation, to demonstrate the semantic difference between different tasks.
    \item Appendix \ref{app:prompt_tuning} expands the prompt optimisation results from Section \ref{sec:prompts} and provides further experimental details and analysis of the system.
    \item Appendix \ref{app:example_algos} introduces two example discovered algorithms, demonstrating that the most performant meta-train/meta-test runs are making novel discoveries rather than just rehashing baselines.
    \item Appendix \ref{app:discobench_extra} reports and discusses success@3 metrics (i.e., the rate of \textit{at least} one of the three seeds producing a successful solution), and \textit{until success} scores (i.e., the Elo scores when ADAs are run until they have three successful independent seeded runs).
    \item Appendix \ref{app:discobench_results} provides per-task results for all experiments in the paper.
    \item Appendix \ref{app:sysprompts} includes all prompts used in this paper. This includes prompts developed by the prompt optimisation loop.
\end{itemize}
\newpage

\section{Task Domain Overview}\label{app:domains}

In this section, we provide a brief overview of the implementations of each task domain included in DiscoGen, as well as all references covering their original implementations and the origin of their datasets.

\subsection{Bayesian Optimisation}

\paragraph{Task Domain} 
Bayesian Optimisation \cite{Jones1998, garnett_bayesoptbook_2023} addresses the problem of optimising an expensive, black-box objective function under a limited evaluation budget. A probabilistic surrogate model is fit to observed function evaluations, and an acquisition function, balancing exploration of uncertain regions with exploitation of promising candidates, is used to select the next evaluation point. The objective is to identify the global minimum or maximum of the function within a fixed number of queries.

\paragraph{Implementation} 
Our Bayesian Optimisation implementation is based on Boax \cite{boax2023github}.

\paragraph{Editable modules}
We include six editable modules for Bayesian Optimisation: the surrogate model; the optimiser used to fit the surrogate model; the acquisition function; the optimiser used to maximise the acquisition function; the initial domain sampling strategy; and the query selection policy.

\paragraph{Datasets} We include 11 synthetic optimisation functions that are standard in Bayesian Optimisation literature \cite{simulationlib}: Ackley 1D, Ackley 2D, Branin 2D, Bukin 2D, Cosine 8D, Drop-Wave 2D, Egg-Holder 2D, Griewank 5D, Hartmann 6D, Holder-Table 2D and Levy 6D.

\subsection{Brain Speech Detection}

\paragraph{Task Domain} 
Brain Speech Detection is a neural signal processing task in which a model predicts the presence of speech from non-invasive or invasive brain recordings.
The goal is to learn a two-class classifier conditioned on neural activity (e.g., MEG signals).

\paragraph{Implementation}
The code for Brain Speech Detection is adapted from the official LibriBrain competition \cite{landau20252025} codebase, which provides a standardised pipeline for neural signal preprocessing, model training, and evaluation.

\paragraph{Editable modules} 
We include three editable modules in Brain Speech Detection: the loss function; the optimiser; and the network architecture.

\paragraph{Datasets} 
We split the original LibriBrain dataset \cite{ozdogan2025libribrain} into seven parts, constituting seven datasets, each of which contains MEG data and labels collected during the process of 
the same participant listening to one chapter of
Sherlock Holmes in spoken English.

\subsection{Computer Vision Classification}

\paragraph{Task Domain} 
Computer Vision Classification is a supervised learning task in which a model assigns a discrete semantic label to an input image.
The objective is to learn robust visual representations that generalise across variations in appearance, scale, illumination, and data distribution.
This task is a foundational benchmark in computer vision and is widely used to evaluate model architectures, optimisation strategies, and robustness to dataset shift. 
In DiscoGen, the Computer Vision Classification task spans standard, corrupted, long-tailed, and fine-grained classification settings.

\paragraph{Implementation} 
The code for Computer Vision Classification is adapted from the MLGym \cite{nathani2025mlgymnewframeworkbenchmark} benchmarking infrastructure.
The implementation provides a unified training and evaluation pipeline for image classification models, including dataset loading via HuggingFace Datasets, model initialisation, optimisation, and metric computation.

\paragraph{Editable modules} 
We include four editable modules in Computer Vision Classification: the loss function; the optimiser; the network
architecture; and the image preprocessing.

\paragraph{Datasets} 
We support nine widely used image classification datasets, covering a range of difficulty levels and distributional properties. 
These include MNIST \cite{lecun2010mnist} and Fashion-MNIST \cite{xiao2017/online} for grayscale digit and apparel classification; CIFAR-10 and CIFAR-100 \cite{krizhevsky2009learning} for small-scale natural image classification; Tiny ImageNet \cite{russakovsky2015imagenet, wu2017tiny} for large-class-count evaluation; CIFAR-10-C \cite{hendrycks2019benchmarking} for corruption robustness; CIFAR-10-LT \cite{krizhevsky2009learning} for long-tailed class imbalance; Oxford Flowers-102 \cite{Nilsback08} and Stanford Cars \cite{Krause_2013_ICCV_Workshops} for fine-grained object classification. 
All datasets are accessed through HuggingFace Datasets and follow standardised train, validation, and test splits.

\subsection{Continual Learning}

\paragraph{Task Domain} Continual learning is a broadly defined field in which a model must learn from non-stationary data sources without forgetting its previous learnings \citep{liyuan2024continual}. Non-stationarity can arise through a number of means; in our tasks, which are focused on image classification under nonstationarity, these include randomly permuting the labels attached to images or randomly subsampling classes shown throughout training.

\paragraph{Implementation} Our default implementation is based on elastic weight consolidation \citep{kirkpatrick2017overcomingcatastropic}. 

\paragraph{Editable modules} We support five different modules in continual learning: the optimisation algorithm; the regulariser for mitigating catastrophic forgetting; the replay buffer for storing past experience; the sampler for mixing replay data with new data; and the learning rate scheduler.

\paragraph{Additional Backends} Our default backend uses a ResNet-18 for its base network \citep{he_deep_2015}. However, we also offer backends supporting vision transformers \citep{dosovitskiy2021an, rw2019timm} and parameter isolation models \citep{rusu_progressive_2022}, a common architecture for preventing catastrophic forgetting \citep{kirkpatrick2017overcomingcatastropic} in continual learning.

\paragraph{Datasets} We support three datasets for continual learning: PermutedMNIST \citep{lecun2010mnist}, SplitCIFAR100 \citep{krizhevsky2009learning} and TinyImageNetSplit \citep{wu2017tiny}.

\subsection{Greenhouse Gas Prediction}

\paragraph{Task Domain} Forecasting the concentration of greenhouse gases in the atmosphere is an important tool for predicting and mitigating the effects of climate change.

\paragraph{Implementation} Our base implementation is based on a standard scikit-learn \citep{scikit-learn} training loop. 

\paragraph{Editable modules} We support two modules in Greenhouse Gas Prediction: the model architecture and the way the data is processed before modelling.

\paragraph{Datasets} We support four datasets from \citet{noaa_gml_co2} that are used for predicting concentrations of CO$_2$, CH$_4$, N$_2$O and SF$_6$ in the atmosphere. Each dataset is split into a training dataset (pre-2014) and a validation dataset (2015-2025).

\subsection{Language Modelling}

\paragraph{Task Domain} Language Modelling is the task of learning the underlying distribution of text data via next-token prediction over vast bodies of text, mostly scraped from the internet. We evaluate the quality of a trained language model by computing the exponential of the average negative log-likelihood across next-token predictions on a validation set, also called perplexity.

\paragraph{Implementation} Our default implementation builds on a modified codebase based on the \texttt{modded-nanogpt} repository~\cite{kellerjordan2024moddednanogpt}.

\paragraph{Editable modules} We support three editable modules: the network architecture, the loss function, and the optimiser.

\paragraph{Additional Backends} In addition to a base nanogpt implementation, we support a state-space model backend based on Mamba \citep{mamba, transformersssms2024dao}.

\paragraph{Datasets} We support the following four datasets: LMFineWeb 10B~\cite{penedo2024finewebdatasetsdecantingweb}, TinyStories~\cite{eldan2023tinystoriessmalllanguagemodels}, OPC-FineWeb Math and OPC-FineWeb Code~\cite{huang2025opencoderopencookbooktoptier}.

\subsection{Model Unlearning}

\paragraph{Task Domain} Model Unlearning, also called Machine Unlearning \cite{yinzhi_unlearning_2015}, is the task of modifying (e.g. fine-tuning) a model to ``forget'' targeted information such as sensitive personal data, copyrighted content, or harmful knowledge, all while preserving the model's overall capabilities on unrelated tasks. We specifically focus on LLM unlearning \cite{yuanshun_unlearning_2024} across 3 datasets and a variety of open-weight models (see below).

\paragraph{Implementation} The code for all tasks is adapted from  OpenUnlearning \cite{openunlearning_2025}. Preservation of general capability is evaluated using LMEvalHarness \cite{lmevalharness_2024}.

\paragraph{Editable modules} We include a single editable module in Model Unlearning: the loss function, which should typically balance two objectives: unlearning specific knowledge from the forget set while preserving performance on the retain set.

\paragraph{Datasets} We support 3 different datasets for Model Unlearning: TOFU \cite{tofu_2024}, MUSE \cite{muse_2024}, and WMDP-Cyber \cite{wmdp_2024}.

\paragraph{Models} We support 13 different open-weight LLMs from 5 providers: Llama-2-7b-chat-hf, Llama-2-7b-hf, Llama-2-13b-hf, Llama-3.2-1B-Instruct, Llama-3.2-3B-Instruct, Llama-3.1-8B-Instruct \cite{llama_2023, grattafiori_llama_2024}, Gemma-7b-it \cite{gemma_2024}, Phi-1.5, Phi-3.5-mini-instruct \cite{abdin_phi-3_2024}, Qwen2.5-1.5B-Instruct, Qwen2.5-3B-Instruct, Qwen2.5-7B-Instruct \cite{qwen_2024} and Zephyr-7b-beta \cite{zephyr_2023}.

\subsection{Neural Cellular Automata}

\paragraph{Task Domain} Neural Cellular Automata (NCAs) are a form of cellular automata in which updates to cells (a function of their neighbours' states) are parameterised and learned by a neural network \cite{mordvintsev2020growing}. This task is about designing the components that the NCA uses to perceive its neighbours and learn its update rule.

\paragraph{Implementation} The code for Neural Cellular Automata is adapted from CAX \cite{faldor2025cax}, a library which provides a unified JAX framework for implementing cellular automata.

\paragraph{Editable modules} We include five editable modules: (i) the perceive module, (ii) the update module, and (iii) the loss, (iv) optimiser and (v) train modules for training the update module.

\paragraph{Datasets} We support five datasets: two growing tasks \cite{mordvintsev2020growing}, a self-classifying MNIST task \cite{randazzo2020self}, matrix operations \cite{bena2025path} and MNIST inpainting \cite{tesfaldet2022attention}.


\subsection{Off-Policy RL}

\paragraph{Task Domain} 
Off-policy RL refers to a class of reinforcement learning approaches in which an agent learns from experience generated by a behaviour policy that may differ from the policy being optimised, including experience collected in the past or by other agents. In this task, we focus on value-based methods that learn value functions by minimising the temporal-difference (TD) error \cite{sutton_reinforcement_2020}.

\paragraph{Implementation} 
The code for Off-Policy RL is adapted from the Deep Q-learning \citep[DQN]{dqn} implementation from PureJaxRL \citep{lu_discovered_2022}, which is itself based on CleanRL \citep{shengyi2022the37implementation, huang2022cleanrl}. 

\paragraph{Editable modules} 
We consider six editable modules: (i) the loss function, which determines the prediction targets for the value network; (ii) the optimiser; (iii) the network architecture; (iv) the replay mechanism, which specifies how experience is stored and sampled during training; (v) the policy, which governs the trade-off between exploration and exploitation; and (vi) the training loop.

\paragraph{Datasets} 
We support Off-Policy RL on four environments from the MinAtar suite \citep{young_minatar_2019}—Asterix, Breakout, Freeway, and Space Invaders—as reimplemented in Gymnax \citep{gymnax2022github} using JAX. These environments are simplified versions of their corresponding Atari games \citep{bellemare13arcade}.

\subsection{Offline RL}

\paragraph{Task Domain} Offline Reinforcement Learning (offline RL) is a class of algorithms that trains a policy from previously collected environment transitions, without any other additional environment interactions~\citep{levine2020offline}. Unlike in online RL, the agent cannot explore and collect new trajectories from the \textit{online} environment and must extract the policy from pre-existing datasets. Such datasets may contain suboptimal trajectories or insufficient state coverage~\citep{kumar2022should}. In practice, the real environment---the \textit{online} domain inaccessible during training---can be used to tune the algorithm's hyperparameters~\cite{jackson2025clean}.

\paragraph{Implementation} We base our implementation on Revisited Behaviour Regularized Actor-Critic (ReBRAC)~\citep{tarasov2023revisiting}, a high-performing and hyperparameter-robust offline RL algorithm. Our JAX implementation is based on those by \citet{jackson2025clean} and \citet{park2025flowqlearning}.

\paragraph{Editable Modules} We include five editable modules for the offline RL task domain: (i) the actor loss, (ii) the critic loss, (iii) the network architectures, (iv) the optimizer, and (v) the training loop.

\paragraph{Datasets} We support all single-task, reward-labeled datasets from OGBench~\citep{ogbench_park2025}, a currently unsaturated offline RL benchmark. This encompasses several robot morphologies across various locomotion and manipulation tasks.


\subsection{On-Policy Multi-Agent RL}

\paragraph{Task Domain} 
On-policy multi-agent reinforcement learning (MARL) refers to a class of multi-agent methods in which each agent updates its policy using an on-policy RL algorithm, thereby relying solely on trajectories generated by their current policy.

\paragraph{Implementation} 
The implementation of our on-policy MARL task is adapted from the Independent PPO (IPPO) algorithm \cite{ippo}, as provided in the JaxMARL library \cite{flair2024jaxmarl}.

\paragraph{Editable modules} 
We include six editable modules in On-Policy MARL: the advantage estimation and critic target computation, the loss function, the optimiser, the network architecture, the activation function used by the network, and the training loop.

\paragraph{Additional Backends}
In addition to the default, we support a recurrent architecture. In this case, the training loop must handle the recurrent state produced by the agent’s network.

\paragraph{Datasets}
We support 17 environments from the JaxMARL library \cite{flair2024jaxmarl}, comprising 5 multi-agent Brax tasks, 11 tasks from SMACv2 \cite{ellis2023smacv2}, and the MPE Spread environment \cite{mpe}.


\subsection{On-Policy RL} 

\paragraph{Task Domain} On-Policy RL is a subset of reinforcement learning in which an agent learns from experience collected by its own policy \citep{sutton_reinforcement_2020}, rather than from data collected by a different behaviour policy.

\paragraph{Implementation} The code for On-Policy RL is adapted from the Proximal Policy Optimisation \citep[PPO]{schulman_proximal_2017} implementation from PureJaxRL \citep{lu_discovered_2022}, which is itself based on CleanRL \citep{shengyi2022the37implementation, huang2022cleanrl}.

\paragraph{Editable modules} We include four editable modules in On-Policy RL: the loss function; the optimiser; the network architecture; and the training loop.

\paragraph{Additional Backends} In addition to the default, we support two backends that can be used to augment tasks in On-Policy RL. These are: a recurrent agent, in which the train loop must support a recurrent variable produced by the agent's networks; and a transformer agent \citep{vaswani_attention_2017}, in which the agent architecture uses attention.

\paragraph{Datasets} We support 13 different environments for On-Policy RL, from three different environment suites. These include: Ant, HalfCheetah, Hopper, Humanoid, Pusher, Reacher and Walker2D from Brax \citep{freeman_brax_2021}, a JAX-based \citep{bradbury_jax_2018} implementation of MuJoCo \citep{todorov_mujoco_2012}; Asterix, Breakout, Freeway and SpaceInvaders from Gymnax \citep{gymnax2022github}, a reimplementation of four MinAtar environments \citep{young_minatar_2019} which are simplifications of some Atari games \citep{bellemare13arcade}; and Craftax and Craftax-Classic \citep{matthews2024craftax}, which are based on Crafter \citep{hafner2021crafter}.

\subsection{Trajectory Prediction}

\paragraph{Task Domain} 
Trajectory prediction is the task of forecasting the future positions of traffic participants, such as vehicles, pedestrians, and cyclists, given their observed past motion and surrounding road context \cite{huang2022survey}.
Effective trajectory prediction must account for the multi-modal nature of human behaviour; at any moment, an agent may accelerate, brake, turn, or change lanes, which can lead to multiple plausible futures.
The objective is to produce a set of $K$ diverse future trajectories for a target agent, each with an associated probability, that collectively cover the space of likely outcomes.
Models are evaluated on how closely their best predictions match the ground truth using minimum Average Displacement Error (minADE), minimum Final Displacement Error (minFDE), miss rate, and Brier-minFDE. 
The ability to anticipate these possibilities is critical for safe motion planning in self-driving systems.

\paragraph{Implementation} 
The base code for Trajectory Prediction is adapted from UniTraj \citep{feng2024unitraj}, a unified framework for scalable vehicle trajectory prediction, and AutoBot \citep{girgis2022autobot}, a latent-variable sequential set transformer for joint multi-agent motion prediction.
The implementation provides a standardised pipeline for encoding past agent trajectories and road geometry via attention-based modules, decoding multi-modal future trajectories with uncertainty estimates, and evaluating predictions against ground truth.

\paragraph{Editable modules} 
We include four editable modules in Trajectory Prediction: (i) the loss function, which defines the multi-modal training objective balancing trajectory likelihood with mode diversity; (ii) the optimiser, which controls weight updates and learning rate scheduling; (iii) the network architecture, which encodes agent dynamics, social interactions, and map context before decoding future trajectories; and (iv) the training loop, which orchestrates data loading, model training, validation, and model selection.

\paragraph{Datasets} 
We support three large-scale autonomous driving datasets: Argoverse~2 \citep{wilson2023argoverse}, collected across six US cities (Austin, Detroit, Miami, Pittsburgh, Palo Alto, and Washington D.C.); nuScenes \citep{caesar2020nuscenes}, recorded in Boston and Singapore and covers diverse urban conditions; and the Waymo Open Motion Dataset \citep{ettinger2021large}, spanning six US cities (San Francisco, Phoenix, Mountain View, Los Angeles, Detroit, and Seattle).
Each dataset is preprocessed into a standardised format with 21 observed timesteps (2.1\,s at 10\,Hz), up to 32 surrounding agents, 128 map polylines, and a 60-timestep (6\,s) prediction horizon.
All three datasets contain 850 preprocessed scenarios, hosted on HuggingFace.

\subsection{Unsupervised Environment Design}
\paragraph{Task Domain} 
 Unsupervised Environment Design (UED) is a field focused on training agents that are robust, and able to generalise to a wide distribution of tasks~\citep{dennis_emergent_2021,jiang_prioritized_2021,parker-holder_evolving_2022,samvelyan2023Maestro}. In particular, this tends to be posed as a two-player game between a task-proposing adversary and a task-solving student~\citep{dennis_emergent_2021}. 
There are several objective functions that the adversary can use, ranging from theoretically principled ones such as regret~\citep{dennis_emergent_2021} to more empirically motivated ones like positive value loss~\citep{jiang2021Replayguided} and learnability~\citep{rutherford2024no}.

\paragraph{Implementation} 
We use the base scaffold from Sampling for Learnability~\citep{rutherford2024no}, which uses PPO as the underlying learning algorithm, and training comprises two stages. The first is sampling a large batch of random environments, then rolling out the agent on these, and using these trajectories to score each level. The second phase then trains the agent on a mixture of the high-scoring levels, and uniform random environments.

\paragraph{Editable modules} 
The first editable module is responsible for rolling the agent out on candidate environments, and then providing a score for each of them. The second is the train step, which controls the actual RL agent update, as well as how the filtered levels are used during training. Finally, the last module consists of the hyperparameters for the sampling and training process.

\paragraph{Datasets} 
We consider two distinct domains; the first is Minigrid~\citep{MinigridMiniworld23}, which is a partially-observable 2D navigation task. The second is Kinetix~\citep{matthews_kinetix_2025}, an open-ended domain of 2D physics tasks. The task distribution consists of a broad distribution of randomly-generated physics puzzles, and the goal is to generalise to interesting, human-designed problems.
There are three levels of difficulty in the Kinetix benchmark, corresponding to how many entities there are in a scene.
We use Minigrid and the \texttt{Small} Kinetix setting as training tasks, and \texttt{Medium} and \texttt{Large} as the testing tasks.


\newpage
\section{Additional Related Work} \label{app:related_work}

\subsection{Automated Research}
AutoML involves trying to apply machine learning to new data without needing human experts \citep{hutter2019automated, he_automl_2021, parker-holder_automated_2022}. Historically, AutoML has focused on areas like hyperparameter tuning (e.g., \citep{smac3, li_hyperband_2018, parker-holder_provably_2021, eimer_hyperparameters_2023}), selecting algorithms from a possible set \citep{feurer_efficient_2015,lindauer_open_2017,feurer_auto-sklearn_2022}, neural architecture search \citep{neat, hyperneat, zoph_neural_2017,zoph_learning_2018,white_neural_2023} and data cleaning and augmentation \citep{timeseries_data_2017_zhang,neutatz_data_2022,activeclean2016krishnan}. While many of these techniques involve applying \textit{existing} machine learning algorithms, the objective of automated algorithm discovery is to develop \textit{new} machine learning algorithms without relying on manual design.

Meta-learning involves learning algorithms from data in a `meta-loop' \citep{Schmidhuber1987EvolutionaryPI, real2020automl, beckSurveyMetaReinforcementLearning2023a}. Often, meta-learned algorithms are black-box, taking the form of a neural network. For example, prior work has considered meta-learning optimisation algorithms (e.g., \citep{andrychowiczLearningLearnGradient2016,metz_meta-learning_2019,oh_discovering_2020,metzPracticalTradeoffsMemory2022,metzVeLOTrainingVersatile2022, goldie_can_2024, lan_learning_2024,  oh_discovering_2025}) or loss functions (e.g., \citep{Kirsch2020Improving,bechtleMetaLearningLearnedLoss2021,lu_discovered_2022, alfano_novel_2023}). Optimising these black-box algorithms frequently relies on evolution \citep{metzVeLOTrainingVersatile2022, lu_discovered_2022, jackson2023discovering,goldie_can_2024, goldie_how_2025} or meta-gradients \citep{oh_discovering_2020, oh_discovering_2025}. However, more interpretable algorithms can also be discovered using symbolic evolution or search \citep{ ramachandran_searching_2017, cranmer_discovering_2020, zheng2022symbolic, chenSymbolicDiscoveryOptimization2023a}, or, driven by the rise of highly capable language models (e.g., \citep{openai_gpt5_2025,google_gemini3_2025, meta_llama4_2025, anthropic_claude_sonnet45_2025}), repeatedly prompting language models for new algorithm suggestions \citep{lu_discovering_2024, romera-paredes_mathematical_2024, novikov_alphaevolve_2025, nathani2025mlgymnewframeworkbenchmark, toledo_ai_2025, gideoni2025random}. In this paper, we frame each algorithm discovery problem as its own \textit{task}, and show how the DiscoGen procedural task generator can help optimise these algorithm discovery systems.

Given a rise in language model capabilities \citep{measuring-ai-ability-to-complete-long-tasks,chollet_arc_2025,phan2025humanityslastexam, paglieri2025balrog}, creating better coding \citep{yang2024sweagent, jiang_aide_2025,anthropic_claude_code, openai_codex_2025} and research agents \citep{lu_ai_2024,nathani2025mlgymnewframeworkbenchmark, toledo_ai_2025} has developed into an important sub-field. Such systems have been used for increasingly complex mathematical \citep{hubert_olympiad-level_2025, luong_towards_2025}, software engineering \citep{yang2024sweagent,liu_large_2025} and research \citep{taylor_galactica_2022,lu_ai_2024, yamada_ai_2025, Karpathy2026AutoResearch} tasks. LLM agents augment base language models with the ability to take actions or use tools \citep{wang_survey_2024,schick_toolformer_2023}, enabling them to run code \citep{nathani2025mlgymnewframeworkbenchmark, yang2024sweagent, anthropic_claude_code}, search the internet \citep{nakano_webgpt_2021,thoppilan_lamda_2022, komeili_internet-augmented_2021, gao_pal_2022,OpenAI_2025_deepresearch,Google_2025_deepresearch}, or access applications like calculators for verifiable tasks \citep{cobbe2021gsm8k}. These tools are used by AI research agents to automate all or part of the research process in a ReAct loop \citep{yao_react_2023}, in which tools are used intermittently throughout a task. In this work, we propose a new procedural task generator for algorithm discovery tasks to help drive forward development of new agents and models for algorithm discovery.

The development and analysis of such agents includes developing agents purely for ideation \citep{si_can_2024}, implementing research ideas proposed by a human-in-the-loop \citep{gottweis_towards_2025,weston_ai_2025}, judging the quality of research papers \citep{lu_ai_2024, si_can_2024,thakkar_can_2025}, or even automating the entire workflow to write scientific articles \citep{lu_ai_2024, yamada_ai_2025, zochi2025, si_towards_2026}. Methods for integrating LLMs into algorithm discovery systems have taken a range of forms. For example, LLMs are often used as mutation and crossover operators in evolutionary algorithms \citep{ma_eureka_2024,klissarov2024maestromotifskilldesignartificial,romera-paredes_mathematical_2024}. However, such systems vary widely in how much agency they give to the LLM. Design decisions include how the agent should search over new algorithms using, for example, tree-search algorithms \citep{jiang_aide_2025,toledo_ai_2025}, whether the LLM should decide which experiments to run \citep{yu_alpharesearch_2025}, or which algorithm to submit as the `final' version \citep{nathani2025mlgymnewframeworkbenchmark}.



\subsection{Optimising Agents}






To optimise a model, architecture, or system for generalising over a problem space, we need to be able to evaluate it over a large amount of data \citep{kaplan_scaling_2020,hoffmann_training_2022, chung_scaling_2024}. For example, language models improve as they are pretrained on more data \citep{kaplan_scaling_2020}, and generalisation in deep reinforcement learning (RL) has benefitted from procedurally generated and larger environments (\citet{cobbe_leveraging_2020}, Section \ref{sec:procgen}). However, it is often preferable to develop models with specialist capabilities; consistently focusing on generalist improvements can introduce safety concerns \citep{bommasani_opportunities_2022,weidinger_ethical_2021}, and may not be the most effective way to enhance desired capabilities \citep{belcak_small_2025}.

Instead, prior work has often focused on building agentic architectures or training models for specific domains. For example, significant effort has gone into developing better agents for mathematics (e.g., \citep{lewkowycz_solving_2022,trinh_solving_2024, hubert_olympiad-level_2025}) or programming (e.g., \citep{alphacode2022, jimenez_swe-bench_2024, roziere_code_2024}). Underpinning many of these advances is access to an evaluation signal from a large, diverse and verifiable problem set which can be optimised by both models, via training  \citep{shao_deepseekmath_2024, wen_reinforcement_2025}, and researchers, through agent and prompt design. For example, there are large suites of mathematical problems (e.g., \citep{cobbe2021gsm8k, hendrycksmath2021}), factual questions \citep{joshi-etal-2017-triviaqa,hendrycks_measuring_2021}, and programming suites can leverage open-source code repositories to verify that inputs lead to expected outputs \citep{jimenez_swe-bench_2024,chen2021codex}. Developing these large suites for algorithm discovery has proven more difficult; manual curation of codebases requires expert knowledge, and often needs adaptation to be applied to different data. Additionally, copying outputs from existing code is insufficient for algorithm discovery, wherein the objective is develop \textit{superhuman} algorithms that are better than those already in existence \citep{romera-paredes_mathematical_2024, novikov_alphaevolve_2025}. In this work, we propose a procedural algorithm discovery task generator that spans millions of possible algorithm discovery tasks, and does not require domain-specific expert knowledge for generating new tasks.

\subsection{Procedural Generation} \label{app:procgen}

Procedural generation involves automatically creating levels or environments, according to rules, rather than manually designing every instance individually \citep{togelius_procedural_2013}. To do so, environments in procedural generation are defined as Contextual Markov Decision Processes \citep[Contextual MDPs]{hallak_contextual_2015} or Underspecified Partially Observable MDPs \citep{dennis_emergent_2021}, where each level or task is defined by a small number of configuration variables. In deep RL, procedural environment generation has proven effective for training agents on a \textit{distribution} of levels, such that they generalise over a distribution rather than just fitting the specific environment levels seen \citep{cobbe_leveraging_2020,frans_powderworld_2023}. Such approaches have been applied to generate levels in environments of ranging complexity, from gridworlds \citep{MinigridMiniworld23} to physics engines \citep{frans_powderworld_2023,matthews_kinetix_2025}, 3-dimensional worlds \citep{team_open-ended_2021}, or games \citep{proceduralcontent2016games, kuttler_nethack_2020,cobbe_leveraging_2020, hafner2021crafter, wang_alchemy_2021,matthews2024craftax}. Procedural generation opens up new avenues of research too, by enabling the development of autocurricula (e.g., \citep{jiang_prioritized_2021, dennis_emergent_2021, parker-holder_evolving_2022}) or large scale meta-learning (e.g., \citep{nikulin_xland-minigrid_2024}).

\newpage

\section{Experimental Details} \label{app:experimental_details}

\subsection{Hyperparameters}

All DiscoBench experiments are run over three meta-seeds; that is, three independent algorithm discovery runs. In all experiments, agents are given a \textbf{24 hour time limit}, and either an 80 (for \textit{DiscoBench Single}) or 100 (for \textit{DiscoBench All}, since tasks are more complicated) action budget, in which to complete a task. Within said limit, the agent is allowed to run many experiments for different implementations, each of which returning performance metrics for all meta-train datasets. Furthermore, for \textit{DiscoBench Single}, we run additional experiments until each agent has 3 valid attempts. For this setting, we remove a small number of tasks (On-Policy RL/Off-Policy RL train\_loop and Model Unlearning) since they proved too difficult to produce three successful attempts.

In all experiments, we use the MLGym agent with different open-source LLMs. We do not add or change the implementations of tools from MLGym. We use default hyperparameters from MLGym for all experiments \citep{nathani2025mlgymnewframeworkbenchmark}, besides those specified below:
\begin{table}[H]
    \centering
    \caption{Non-default MLGym hyperparameters.}
    \label{tab:mlgym_hparams}
    \begin{tabular}{l c c}
        \toprule
        \textbf{Setting} 
        & \texttt{gpus\_per\_agent} 
        & \texttt{max\_steps} \\
        \midrule
        \textit{DiscoBench (Single Edit)} 
        &1 & 80 \\
        \textit{DiscoBench (All Edit)} 
        & 1 & 100 \\
        \textit{Prompt Tuning (Meta-Training)} 
         &4 & 50 \\
        \textit{Prompt Tuning Single} 
         &1 & 50 \\
        \textit{Prompt Tuning All} 
         &1 & 60 \\
        \bottomrule
    \end{tabular}
\end{table}

\subsection{Elo Calculation}

Our results are comparative between models, and based on an Elo score. In Elo, each model's score changes based on pairwise win-loss comparisons of their performance in individual tasks, and the Elo disparity between the model. Before calculating the Elo, we aggregate the per-dataset scores for each model over its meta-seeds; we also include a penalty to scores which ensures that a model with more successful attempts out of its three meta-seeds dominates a model with less successful attempts.

For model $1$ with rating $R_1$ and model $2$ with rating $R_2$, the expected outcome score of model $1$ is
\begin{equation*}
    E_1 = \frac{1}{1+10^{(R_2-R_1)/400}},
\end{equation*}
where a $400$ point difference corresponds to a $\approx91\%$ chance of winning.

After calculating $E_1$, the rating of model $1$ is updated to reflect the difference between the expected and true score as
\begin{equation*}
    R_1' = R_1 + K\cdot(S_1-E_1).
\end{equation*}
In this expression, $S_1$ is the score for model $1$, and is set as $1.0$ for a win, $0.5$ for a draw and $0.0$ for a loss. Scores for all models are initialised at 1000, and $K=32$ to balance volatility of score calculations. We loop through the data for $1000$ epochs, shuffling each epoch and annealing $K$ to $1$ for stability. We use 100 bootstrapping rounds to estimate $95\%$ confidence intervals.

\subsection{Prompt Optimisation Experimental Details}

For our prompt optimisation experiment, we use Claude-4.5 Sonnet \citep{anthropic_claude_sonnet45_2025} to automatically refine prompts for an ADA; here the ADA is the MLGym agent with Deepseek-V3.2 \citep{deepseek-ai_deepseek-v32_2025,nathani2025mlgymnewframeworkbenchmark}. We sweep over different numbers of tasks for prompt optimisation by changing the \textit{frequency} at which a task is sampled. As such, a new task is sampled every $30/N_{tasks}$ steps. 

We explored the generalisation of each prompt to a set of four held-out domains (which is a larger distribution shift than for the 10 in-distribution domains, despite all task specifics being held out from the agent). The four held-out domains we select are: \textit{On-Policy MARL, Offline RL, Neural Cellular Automata, and Trajectory Prediction}.

The task sampling process involved first sampling a task domain from the ten possible in-distribution set from DiscoGen. Each dataset from the task domain had a $40\%$ chance of inclusion in meta-train or meta-test, and a $20\%$ chance of being excluded from the task. Every module has an independent $30\%$ chance of being marked as editable. Any invalid task (i.e., no editable modules or an empty meta-train/meta-test set) is discarded, and we resample from DiscoGen. We cap the maximum amount that each domain can be sampled to 10, to prevent domain bias, but this limit is not reached in practice over $30$ ADA optimisation iterations.

To prevent unbounded context growth for the ADA Optimisation LLM (Claude-4.5 Sonnet), we do not provide full conversation history. Instead, at every ADA optimisation step, we provide: a system prompt (Appendix \ref{app:sysprompts}); the current and three preceding prompts with their performance; and up to 15 previous task domain-performance pairs without their corresponding prompt, to help ground scores. It is not told the specific task configuration, to aid development of a general prompt. Claude is given per-dataset breakdowns, as well as whether each dataset is from meta-train or meta-test in a given task. We also tell Claude if there were any failures or error messages, but do not provide full error messages; again to limit the context size. If Claude does not produce a prompt fitting the prescribed template within three attempts, the task is discarded and resampled without increasing the ADA optimisation iteration (i.e., the task is not counted towards the total).

\subsection{Different Initialisation Domains}\label{app:init_hparams}

Due to computational constraints, we are unable to run an exhaustive comparison between initialisation strategies for \textit{all} tasks in the DiscoGen ADA evaluation set. Instead, we select a number of the quicker-to-run domains to run our ablation on. In particular, we use tasks for the following domains: \textit{Bayesian Optimisation; Brain Speech Detection; Greenhouse Gas Prediction; Computer Vision Classification; Neural Cellular Automata; On-Policy RL; Off-Policy RL; and Trajectory Prediction}.

\subsection{Experimental Compute}\label{compute}
All experiments are run using Nvidia H200 GPUs. We run the majority of experiments on a single GPU, besides prompt optimisation experiments which are run on 4 GPUs to accelerate experimentation, and On-Policy MARL experiments which use 4 GPUs due to environment memory requirements. For all experimentation (including most development and experiments which are not directly included in this paper), we use an estimated 35,000 GPU-hours of compute. However, such high-end compute is not generally a requirement for usage. To ensure DiscoGen is accessible to researchers under different constraints, we verified that many DiscoGen task domains can run on consumer-grade (e.g., Nvidia 2080Ti/3080) and mid-range (Nvidia L40s) hardware.

All models were run through an API. The total cost of API credits used in development and experiments is estimated to be about \$1600. A task generally costs on the order of \$0.10-\$0.50 to run with open-source models.

\newpage

\section{DiscoGen Details} \label{app:implementation}

\subsection{Technical DiscoGen Implementation}\label{app:technical_implementation}

Here, we clarify the implementation details of DiscoGen; in particular, how tasks are generated.

DiscoGen operates by creating \textit{file systems}. To sample a random task from DiscoGen, a random DiscoGen configuration must be created. Doing so involves selecting a task domains, which defines the availability of modules, datasets and backends, and randomising each of these categories. An example task configuration is shown in Example \ref{fig:example_config}.

After a configuration has been created, DiscoGen is queried to build the meta-train portion of the task. To do so, DiscoGen creates: (1) all necessary files for \textit{each} meta-training dataset (in per-dataset folders), including downloading and caching any data; (2) a script for running an inner-loop on all meta-training datasets; (3) a directory which includes all discovered algorithms; and (4) a procedurally generated \texttt{description} file. The task description covers \textit{four} things: a high-level overview of algorithm discovery; a description of the task domain; information about each of the editable modules, including what purpose it serves and what interface it must have; and descriptions of each dataset in meta-training. We provide all of these descriptions in our \href{https://github.com/AlexGoldie/discogen/}{open-source repository}, which also includes annotated task-generation code (\texttt{make\_files.py}).

Any ADA operates in a meta-loop over the meta-train files to develop new algorithms. When the meta-loop is complete, DiscoGen is queried again to build the meta-\textit{test} task. As this is created, \textbf{all} files besides the editable modules in \texttt{discovered/} are overwritten, and rebuilt from scratch for different datasets. This is done to lower the risk of evaluation hacking. Since the meta-test datasets are not known to the agent, we also dramatically reduce the risk of train-test leakage.

\subsection{Deriving Task Counts} \label{app:derivation}

For $m$ different modules, $d$ different datasets, and $b$ different backends in a task domain, we derive the number of valid tasks below.

Each dataset can be marked as part of the \textit{meta-train} set, \textit{meta-test} set, or \textit{excluded} set, given 3 possible options per dataset. The same dataset can not be included more than once, to prevent leakage between the meta-train and meta-test sets (i.e., $\mathcal{D}_{train} \cap \mathcal{D}_{test} = \varnothing$. Therefore, the number of possible combinations of datasets is $3^d$. However, we require \textit{at least} one dataset to be in the meta-train, and \textit{at least} one dataset in meta-test. We remove the two $2^d$ cases where this is not true (i.e., where either meta-train or meta-test are excluded as options), but add back the double-counted case of \texttt{all exclude}. This produces
$(3^d-2\times2^{d}+1) = (3^d-2^{(d+1)}+1)$ valid dataset configurations.

For $m$ modules, each modules can be marked as \textit{editable} or \textit{fixed}, meaning each module has $2$ possible states. This means there are $2^m$ possible module combinations. By removing the case where all modules are \textit{fixed}, this gives $2^m-1$ valid modules configurations.

We currently support $2$ types of initialisation (start-from-interface and start-from-baseline), and $3$ types of evaluation (performance, energy and time).

By combining the number of configurations for datasets and modules, and multiplying by the number of backends, initialisations and evaluation types, this gives
\begin{equation*}
    N_{tasks} = 2 \cdot 3 \cdot b \cdot (2^m-1)\cdot\left(3^d-2^{(d+1)}+1\right).
\end{equation*}

For Model Unlearning, where we can provide one of $n$ different base models for each dataset, the computation changes slightly. As opposed to counting $3$ options for every dataset there are $2n+1$ possible choices; any combination of meta-train/model ($n$), meta-test/model ($n$), or exclude ($1$). This provides $(2n+1)^d$ combinations. The degenerate cases now count as removing $(n+1)^d$ combinations each, meaning we remove $2(n+1)^d$ in place of the previous $2\times2^d$. This gives 
\begin{equation*}    
N_{tasks} = 2 \cdot 3 \cdot b \cdot (2^m-1)\cdot\left((2n+1)^d-2(n+1)^{d}+1\right).
\end{equation*}

\newpage


\section{DiscoBench Task List} \label{app:task_list}

Here, we provide a list of the meta-train/meta-test dataset splits for each DiscoBench task in this paper. All datasets are described and referenced in Appendix \ref{app:domains}.

\subsection{Bayesian Optimisation}

\paragraph{Meta-Train Datasets:}Ackley1D, Branin2D, Cosine8D, Eggholder2D, Hartmann6D, Levy6D,

\paragraph{Meta-Test Datasets:}Ackley2D, Bukin2D, DropWave2D, Griewank5D, HolderTable2D

\subsection{Brain Speech Detection}

\paragraph{Meta-Train Datasets:} LibriBrain Sherlock Holmes 1-3

\paragraph{Meta-Test Datasets:} LibriBrain Sherlock Holmes 4-7

\subsection{Computer Vision Classification}

\paragraph{Meta-Train Datasets:} CIFAR10, FashionMNIST, MNIST, OxfordFlowers

\paragraph{Meta-Test Datasets:} CIFAR100, CIFAR10C, CIFAR10LT, StanfordCars, TinyImageNet

\subsection{Continual Learning}

\paragraph{Meta-Train Datasets:} SplitCIFAR100

\paragraph{Meta-Test Datasets:} PermutedMNIST, TinyImageNetSplit

\subsection{Greenhouse Gas Prediction}

\paragraph{Meta-Train Datasets:} CH4, SF6

\paragraph{Meta-Test Datasets:} CO2, N2O

\subsection{Language Modelling}

\paragraph{Meta-Train Datasets:} LMFineWeb, OPCFineWebMath

\paragraph{Meta-Test Datasets:} OPCFineWebCode, TinyStories

\subsection{Model Unlearning}

\paragraph{Meta-Train Datasets:} MUSE

\paragraph{Meta-Test Datasets:} TOFU, WMDP-Cyber

\paragraph{Base Model:} Qwen2.5-1.5B-Instruct

\subsection{Neural Cellular Automata}

\paragraph{Meta-Train Datasets:} GrowingLizard, SelfClassifyingMNIST

\paragraph{Meta-Test Datasets:} GrowingButterfly, MatrixOperations, MNISTInpainting

\subsection{Off-Policy RL}

\paragraph{Meta-Train Datasets:} MinAtar/Breakout, MinAtar/Freeway

\paragraph{Meta-Test Datasets:} MinAtar/Asterix, MinAtar/SpaceInvaders

\subsection{Offline RL}

\paragraph{Meta-Train Datasets:} OGBench/antmaze-large-navigate, OGBench/cube-single-play, OGBench/humanoidmaze-medium-navigate, OGBench/puzzle-3x3-play, OGBench/scene-play

\paragraph{Meta-Test Datasets:} OGBench/antmaze-giant-navigate, OGBench/antsoccer-arena-navigate,OGBench/cube-double-play, OGBench/humanoidmaze-large-navigate, OGBench/puzzle-4x4-play

\subsection{On-Policy MARL}

\paragraph{Meta-Train Datasets:} MABrax/Ant, MABrax/Walker, SMAX/2s3z, SMAX/6h\_vs\_8z, SMAX/27m\_vs\_30m, SMAX/smacv2\_10\_units

\paragraph{Meta-Test Datasets:}  MABrax/HalfCheetah, MABrax/Hopper, MABrax/Humanoid, MPE/Spread, SMAX/3s\_vs\_5z, SMAX/3s5z, SMAX/3s5z\_vs\_3s6z, SMAX/5m\_vs\_6m, SMAX/10m\_vs\_11m, SMAX/smacv2\_5\_units, SMAX/smacv2\_20\_units

\subsection{On-Policy RL}

\paragraph{Meta-Train Datasets:} MinAtar/Breakout, MinAtar/Freeway

\paragraph{Meta-Test Datasets:}  MinAtar/Asterix, MinAtar/SpaceInvaders

\subsection{Trajectory Prediction}

\paragraph{Meta-Train Datasets:} nuScenes, Argoverse2

\paragraph{Meta-Test Datasets:} Waymo

\subsection{Unsupervised Environment Design}

\paragraph{Meta-Train Datasets:} Kinetix/Small, Minigrid

\paragraph{Meta-Test Datasets:} Kinetix/Medium, Kinetix/Large

\newpage

\section{Redundancy Analysis} \label{app:correlation}

To ensure DiscoGen tasks exhibit semantic diversity, rather than redundancy, in this section we produce and analyse three Spearman Rank Correlation plots based on \textit{DiscoBench Single (Until Success)} results from Section \ref{sec:discobench_results}. 

\begin{figure}[htbp]

    \centering
    \includegraphics[width=\linewidth]{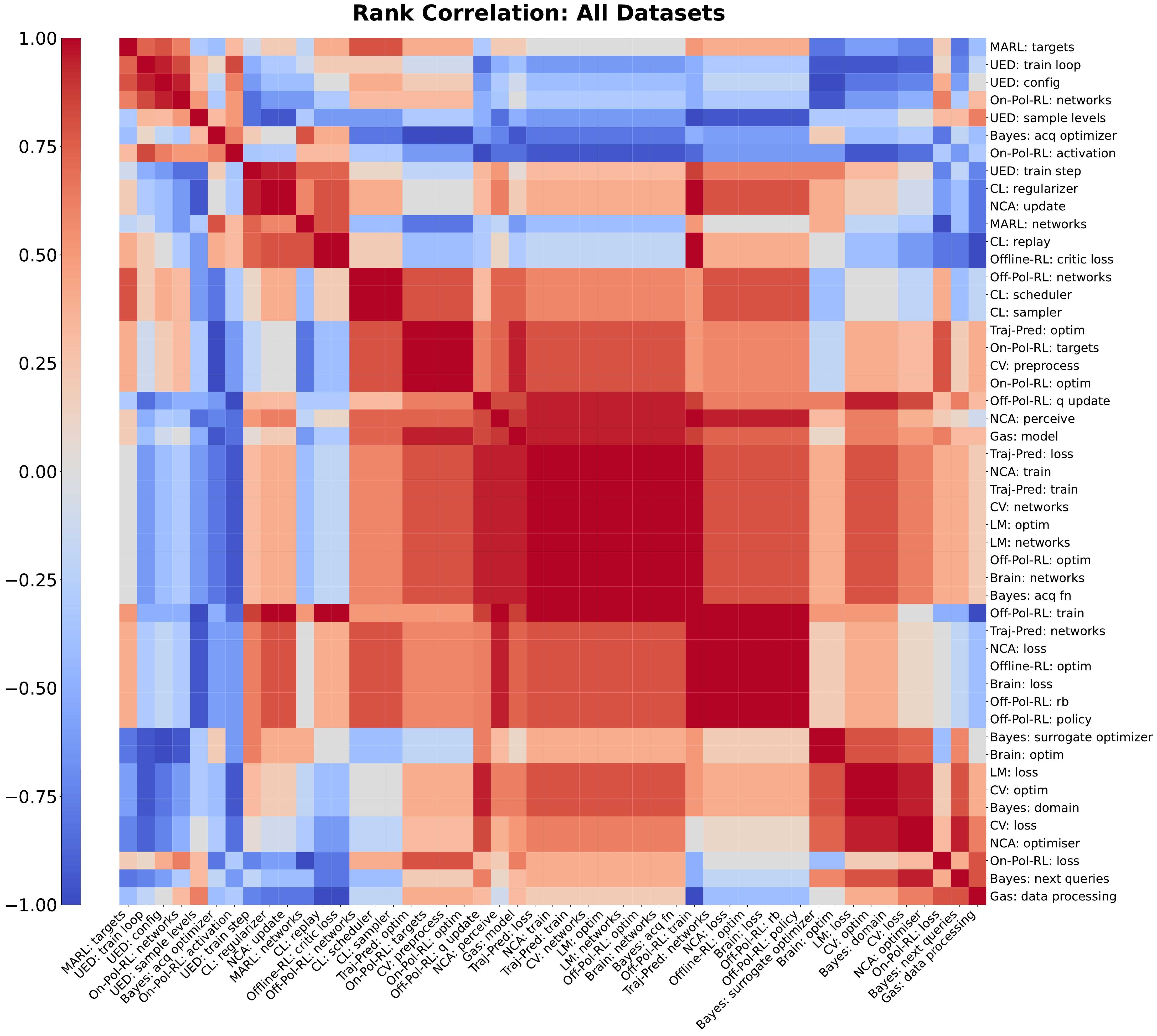}
    
    \caption{Clustered average rank correlation over all datasets (meta-train and meta-test) for each task.}
    \label{fig:all_data_corr}
\end{figure}

\begin{figure}[htbp]
    \centering
    \begin{subfigure}[t]{0.48\linewidth}
        \centering
        \includegraphics[width=\linewidth]{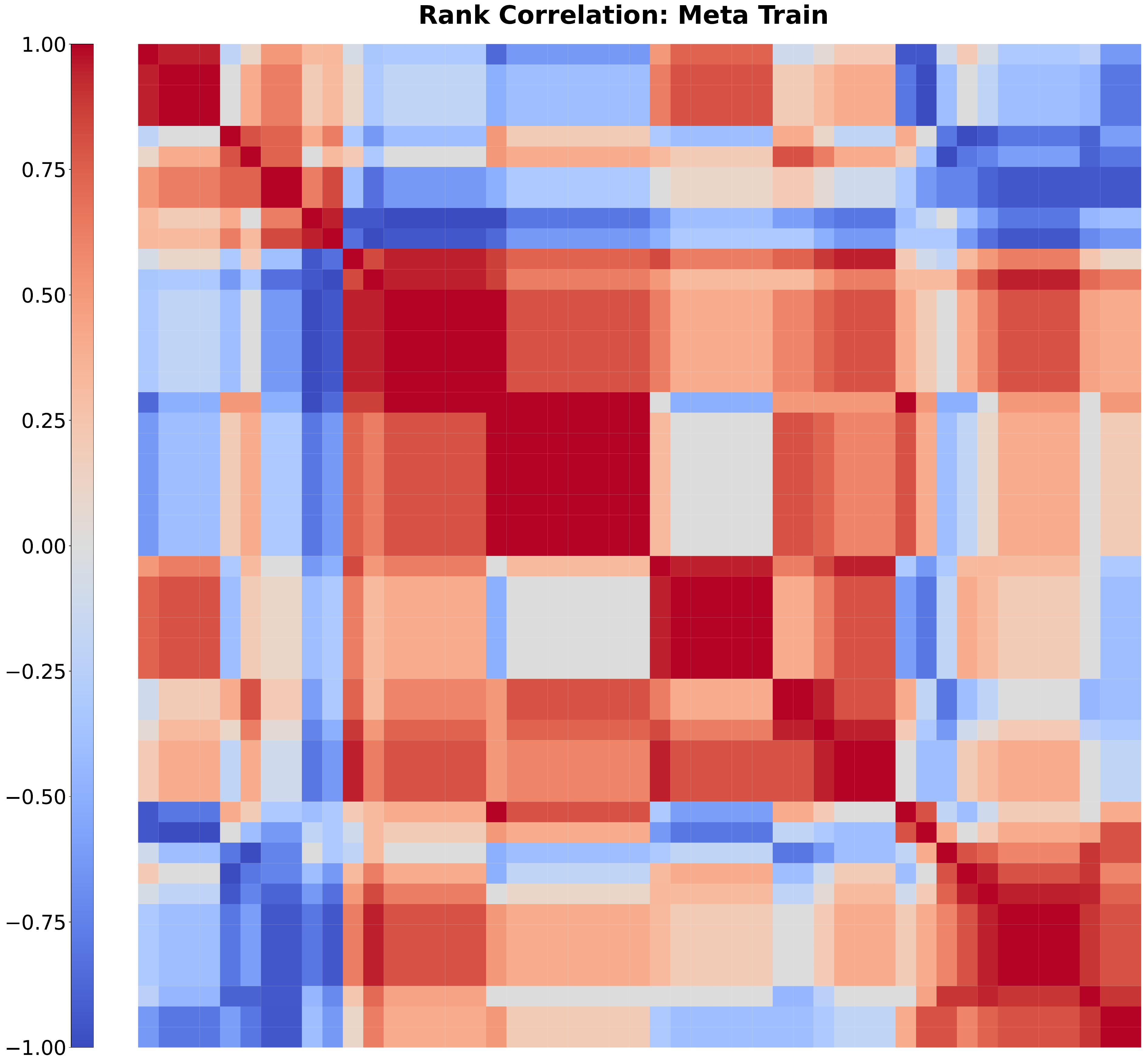}
        \caption{Clustered average rank correlation over meta-train datasets.}
        \label{fig:meta_train_corr}
    \end{subfigure}
    \hfill 
    \begin{subfigure}[t]{0.48\linewidth}
        \centering
        \includegraphics[width=\linewidth]{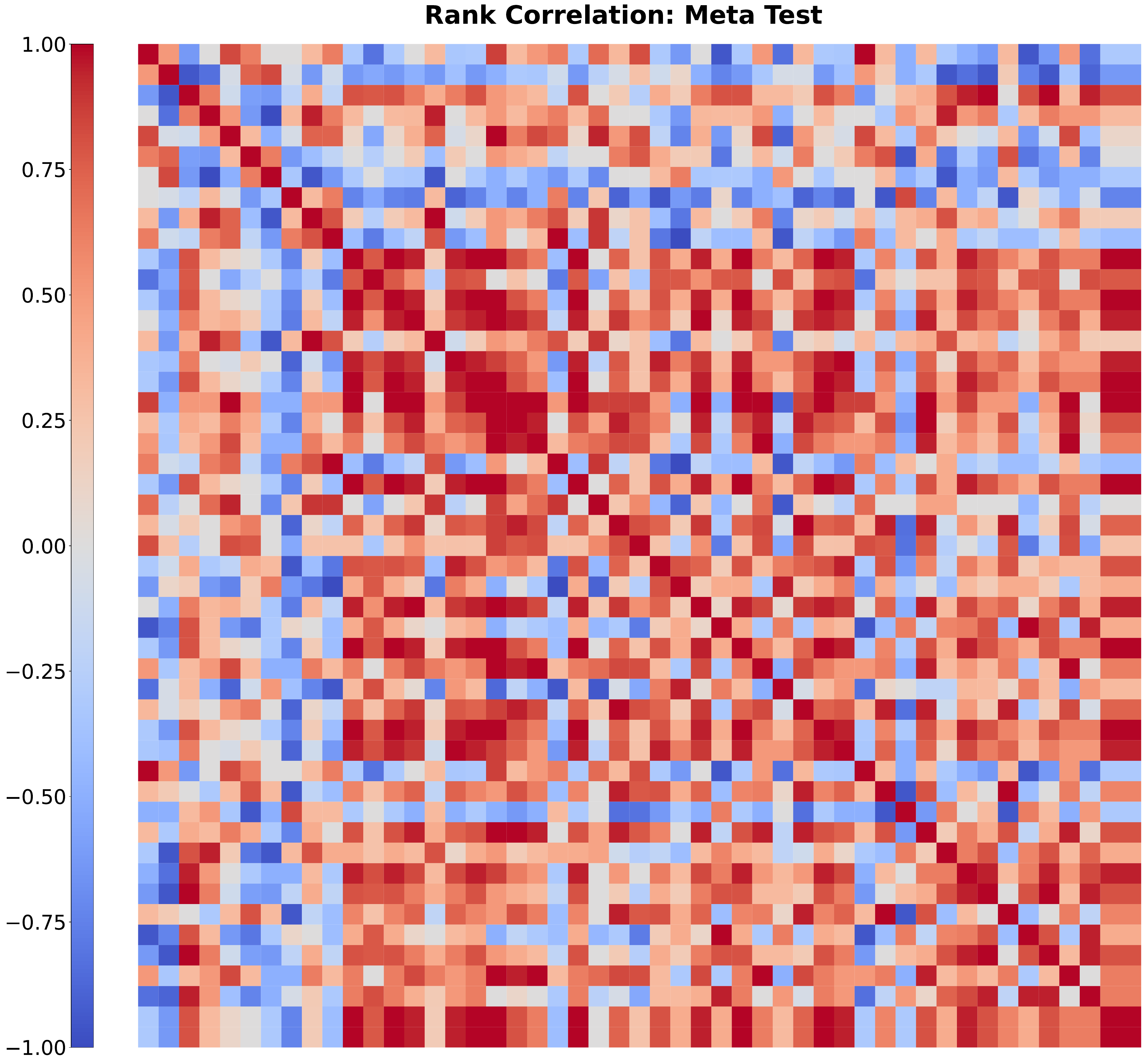}
        \caption{Average rank correlation over meta-test datasets, ordering the tasks based on meta-train clustering.}
        \label{fig:meta_test_corr}
    \end{subfigure}
    
    \caption{Rank correlations for Meta-Train (left) and Meta-Test (right). The Meta-Test plot inherits the clustering order from Meta-Train to visualize consistency across splits.}
    \label{fig:split_corrs}
\end{figure}

For each task, we compute the \textit{average} rank of the models and baseline over datasets in the task. In Figure \ref{fig:all_data_corr}, we combine meta-train and meta-test results into one set and compute average rank over all datasets. We compute the Spearman Rank correlation between all tasks, which is shown through a heatmap. In Figures \ref{fig:meta_train_corr} and \ref{fig:meta_test_corr}, we show the rank correlations over only the meta-train and meta-test sets for each task respectively.

We use hierarchical clustering in Figures \ref{fig:all_data_corr} and \ref{fig:meta_train_corr} to group together tasks with similar average rankings. For Figure \ref{fig:meta_test_corr}, we \textit{keep} the task-ordering from the meta-train clustering; this helps visualise the consistency in rank ordering between meta-train and meta-test for the same tasks.

\vspace{-8pt}
In \cref{fig:all_data_corr}, we find clusters of high-correlation tasks that are occasionally, though not always, intuitive. There is often high correlation between tasks in which the network architecture is the editable module. Alternatively, there is notably \textit{less} correlation when the editable module is the optimiser. We also find frequently mixed correlation between tasks from the same domain; for example, tasks within on-policy RL or Bayesian Optimisation are frequently anti-correlated.

In Figure \ref{fig:split_corrs}, we consider the consistency of clusters between meta-train and meta-test; essentially, exploring whether high and low correlation patterns are reflected between the two regimes, to explore alignment when \textit{only} the datasets are changed. To do so, we first compute the order of labels by clustering the \textit{meta-train} heatmap, and keep this order fixed in the \textit{meta-test} plot. If patterns persist, it would suggest that the rank correlations are similar between the two regimes. 

Compared to the clustered plot for meta-train (\cref{fig:meta_train_corr}), the correlation structure in meta-test (\cref{fig:meta_test_corr}) effectively dissolves to noise. Despite the \textit{only} difference between these plots being the datasets (i.e., \cref{fig:meta_train_corr} plotting over meta-train datasets that the agent develops algorithms for, and \cref{fig:meta_test_corr} plotting over the meta-test datasets that the agent doesn't see during its meta-loop), the average algorithm rank changes \textbf{dramatically} to the extent that old patterns become broadly unrecognisable. This provides strong justification to the semantic diversity, and lack of redundancy, in the combinatorial task space of DiscoGen; simply changing datasets completely changes the ranking of ADAs. It also justifies the fact that the meta-train evaluation of existing benchmarks is insufficient.

\newpage

\section{Prompt Tuning Analysis} \label{app:prompt_tuning}

We do not visualise ADA optimisation curves for $K_{tasks}\neq1$, since tasks are frequently changing and performance is incomparable between different datasets, domains, and module combinations. However, to demonstrate the validity of our prompt improvement ADA optimisation loop, we visualise the ADA optimisation curves over the meta-train and meta-test environments (both of which are known during ADA optimisation, since meta-test is only held out from the ADA itself).

\vspace{-8pt}
\begin{figure}[htbp]
    \centering
    \begin{subfigure}{0.43\textwidth}
        \centering
        \includegraphics[width=\linewidth]{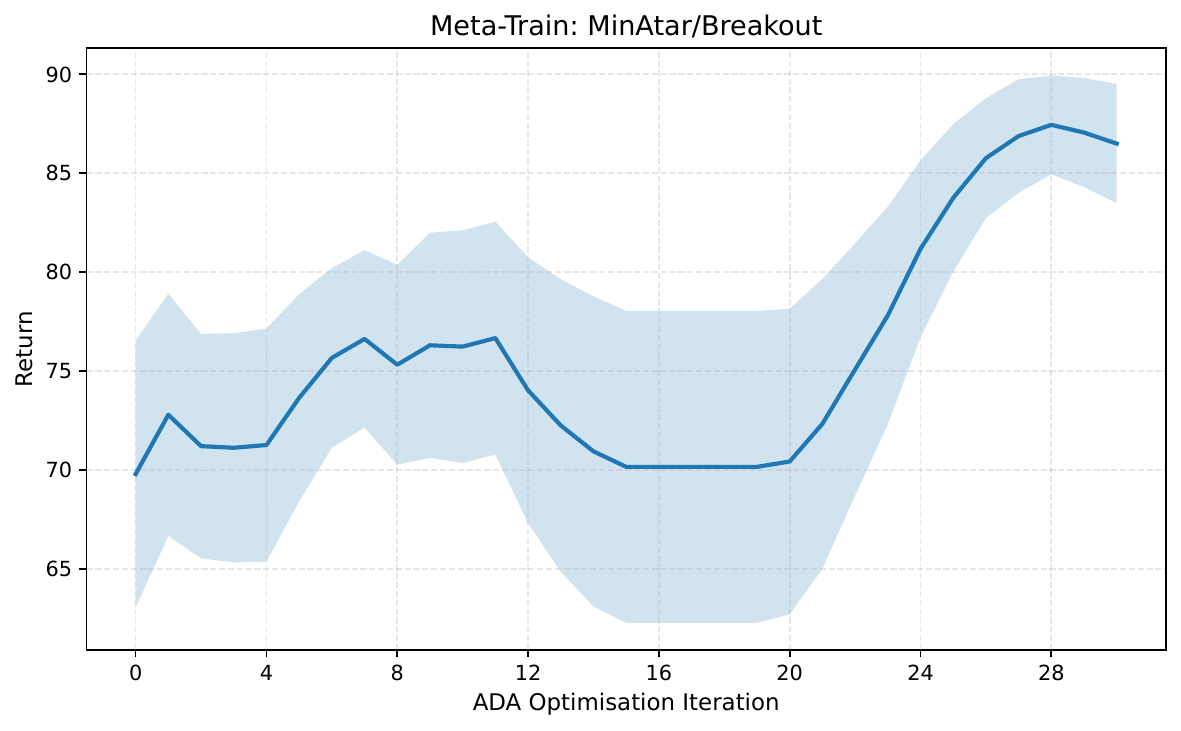}
        \caption{Performance in Breakout (a meta-train environment) over the course of ADA optimisation.}
        \label{fig:top-left}
    \end{subfigure}
    \hfill 
    \begin{subfigure}{0.43\textwidth}
        \centering
        \includegraphics[width=\linewidth]{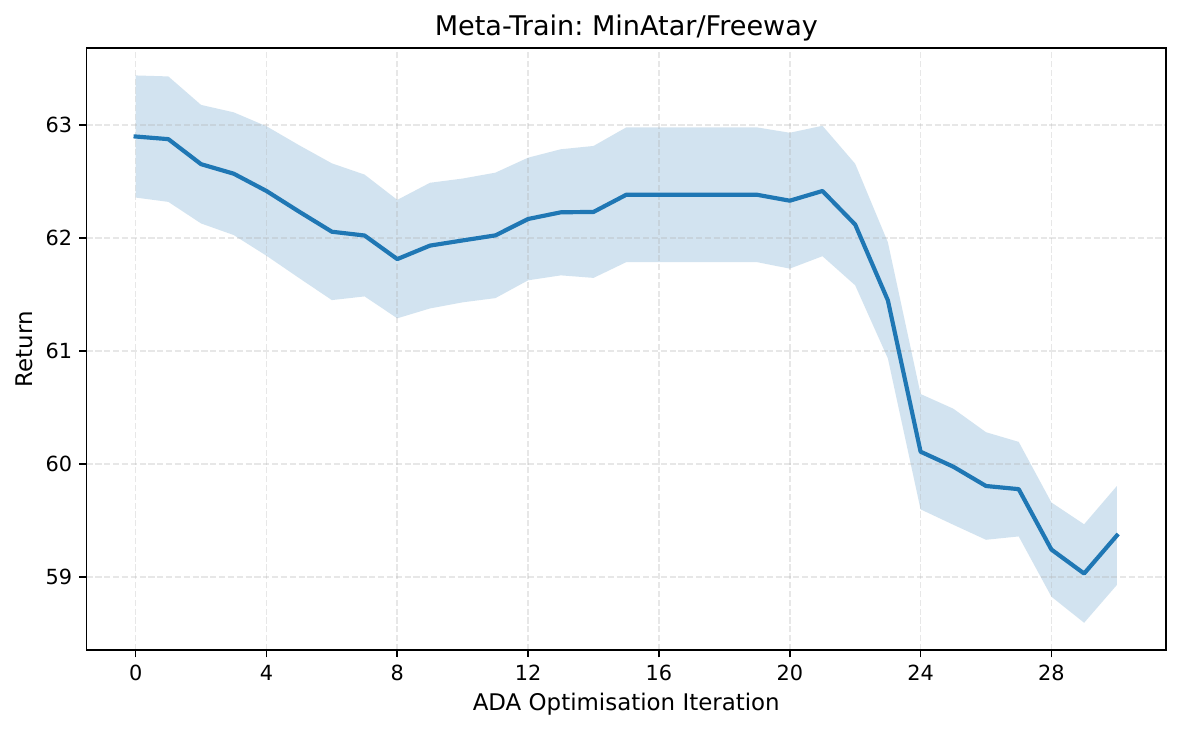}
        \caption{Performance in Freeway (a meta-train environment) over the course of  ADA optimisation.}
        \label{fig:top-right}
    \end{subfigure}


    \begin{subfigure}{0.43\textwidth}
        \centering
        \includegraphics[width=\linewidth]{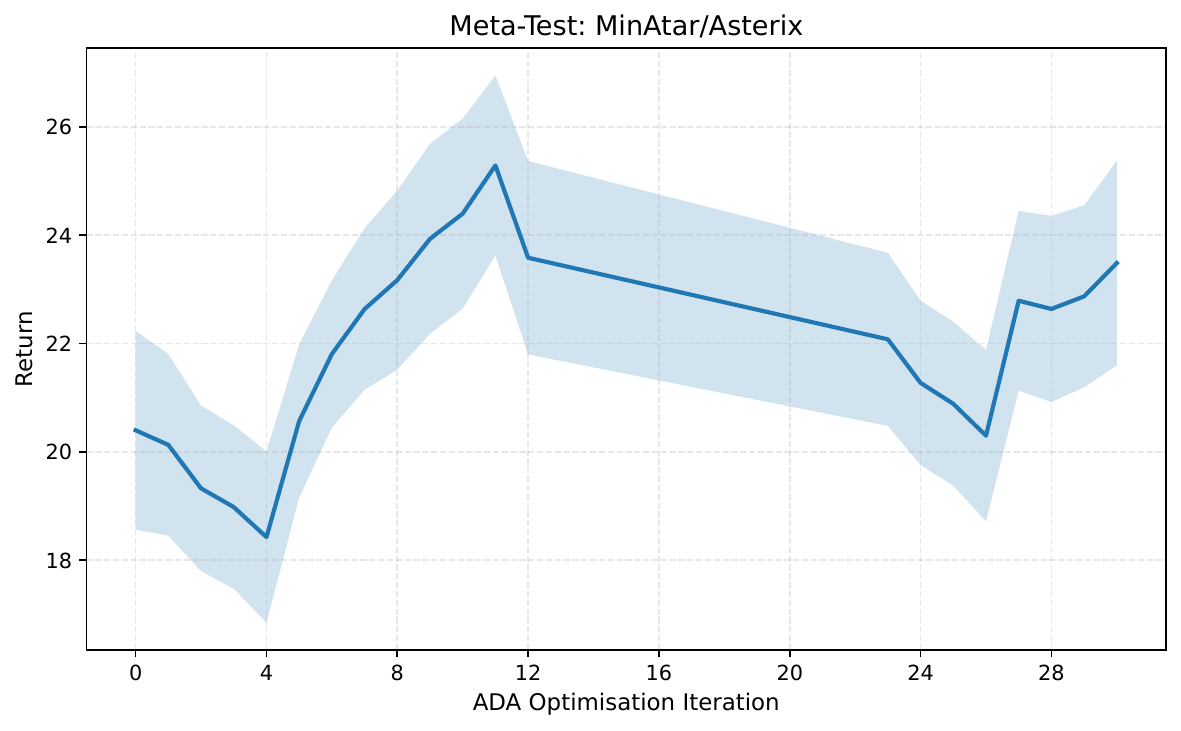}
        \caption{Performance in Asterix (a meta-test environment) over the course of  ADA optimisation.}
        \label{fig:bottom-left}
    \end{subfigure}
    \hfill
    \begin{subfigure}{0.43\textwidth}
        \centering
        \includegraphics[width=\linewidth]{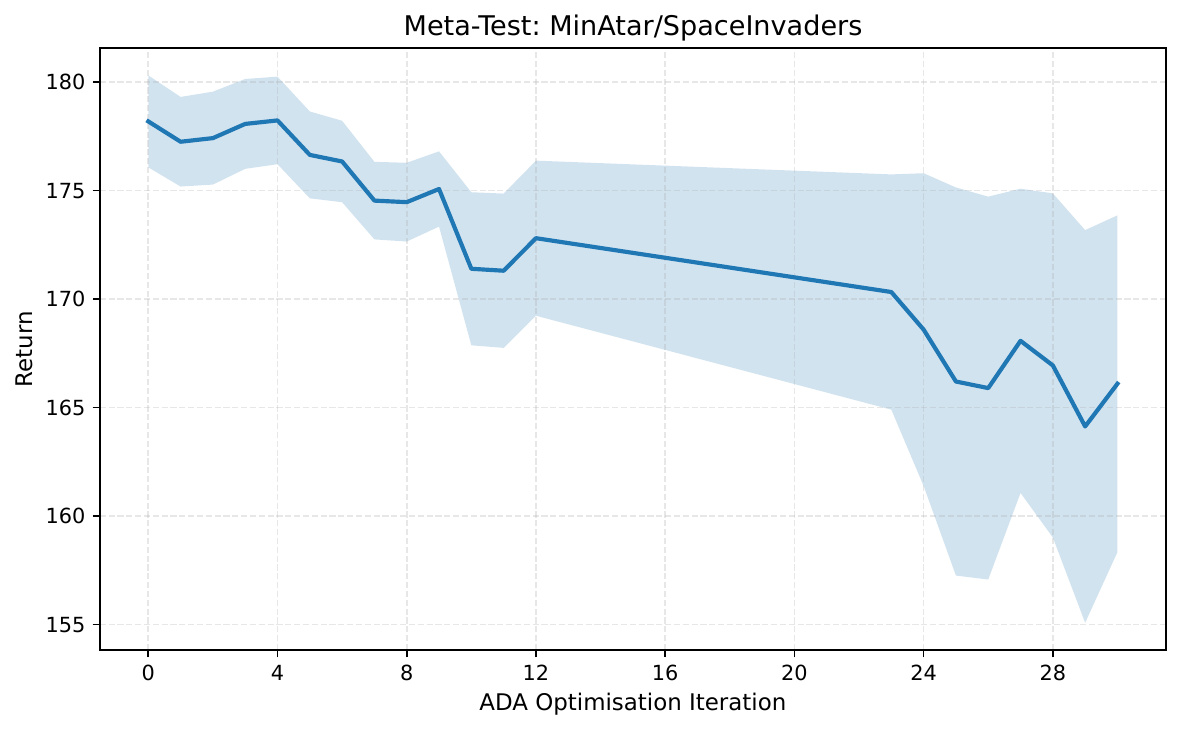}
        \caption{Performance in Space Invaders (a meta-test environment) over the course of  ADA optimisation.}
        \label{fig:bottom-right}
    \end{subfigure}
    
    \caption{Visualisation of performance for the four environments that are used in the $K_{tasks}=1$ prompt optimisation experiment. We plot mean and standard error of the evaluation return achieved by the final policy for each environment over 8 inner-loop seeds, as defined by the On-Policy RL task domain implementation.}
    \label{fig:four-plots}
\end{figure}

Given prompt optimisation using a language model is effectively a search problem, the curves are (expectedly) noisy. However, there is a general trend of improvement. Considering these curves in parallel with the final prompt (Section \ref{app:task1prompt}) helps elucidate these curves. In particular, the prompt emphasises \textit{improvement} in Breakout and \textit{maintaining} in Freeway. Furthermore, the vast majority of the prompt is dedicated to scores in Breakout; it suggests making changes based on the variance of Breakout, for instance. 

\newpage

\section{Example Discovered Algorithms} \label{app:example_algos}

One question to explore is whether ADAs in DiscoGen \textit{actually} discover new algorithms, or whether successful solutions end up reproducing baselines and subsequently tweaking hyperparameters. Searching through all discovered algorithms would be difficult, given the number of independent ADA runs in our work, for different domains and module combinations. As such, we select two interesting discovered algorithms to briefly discuss here. These are chosen as two algorithms which generalised especially well from meta-train to meta-test. We note that these algorithms were \textit{not} the best performers in meta-train (though neither performed badly), which generally did not transfer well to meta-test.

\subsection{Language Modelling:}

We firstly introduce a discovered loss function for language modelling. We include the loss function code in Example \ref{code:loss_func}.

\begin{lstlisting}[language=Python, caption={A discovered loss function for language modelling.}, label={code:loss_func}]
from typing import Sequence

import torch.nn.functional as F
import torch
import torch.nn as nn
from typing import Dict, Optional

def compute_loss(outputs: Dict[str, torch.Tensor], labels: torch.Tensor, num_items_in_batch: Optional[int] = None) -> torch.FloatTensor:

    """Calculate the loss for the model's outputs against the true labels.
    Args:
        outputs (dict): The model's outputs, typically containing logits {"logits": logits}.
        labels (torch.Tensor): The true labels for the batch.
        num_items_in_batch (int, optional): The number of items in the batch. Defaults to None.

    Returns:
        torch.FloatTensor: The computed loss value.
    """
    logits = outputs["logits"]

    # Additive margin softmax (AM-Softmax) with margin=0.1 and scale=10
    margin = 0.1
    scale = 10.0

    one_hot = torch.zeros_like(logits).scatter_(1, labels.unsqueeze(1), 1.0)
    logits_margin = logits - margin * one_hot
    logits_margin = scale * logits_margin

    # Cross-entropy with label smoothing
    loss = F.cross_entropy(logits_margin, labels, label_smoothing=0.05)

    # Entropy regularization
    probs = F.softmax(logits, dim=1)
    entropy = -torch.sum(probs * torch.log(probs + 1e-8), dim=1).mean()
    loss = loss + 0.01 * entropy
    return loss 
\end{lstlisting}

The ADA makes a number of unintuitive design decisions that, when combined, performed close to the maximum in both meta-train and meta-test. Firstly, the agent uses additive margin softmax \citep{additivemargin}, a technique used in face verification to encourage large margins between network outputs, which is generally undesirable in language modelling where synonyms should be treated similarly. This is used in the cross-entropy calculation with a scale of 10; effectively making the target distribution significantly more sharp. Secondly, the agent introduces conflicting objectives: label-smoothing, which optimises towards  \textit{reduced} certainty in outputs \citep{inceptionv3}, and an entropy penalty which \textit{encourages} certainty. This all combines to give an interesting loss function; one which encourages and penalises certainty simultaneously.

\subsection{On-Policy RL: A CNN-MLP Dual-Path Architecture For MinAtar}

We also explore a dual-path network architecture \citep{dualpath} for on-policy reinforcement learning in MinAtar \citep{young_minatar_2019}. We provide discovered python code in Example \ref{code:hybrid_arch}, removing any redundant code for clarity. In the DiscoGen setup, the activation is imported from a separate file.

\begin{lstlisting}[language=Python, caption={A discovered neural network architecture for on-policy reinforcement learning.}, label={code:hybrid_arch}]
from typing import Sequence

import distrax
import flax.linen as nn
import jax
import jax.numpy as jnp
import numpy as np
from flax.linen.initializers import constant, orthogonal


class ActorCritic(nn.Module):
    action_dim: Sequence[int]
    config: dict

    @nn.compact
    def __call__(self, x):
        """Insert your network logic here."""
        # Input = x. x is the environment observation.
        # x shape: (batch, 400) after flatten.
        hsize = self.config.get("HSIZE", 64)
        
        # Map activation string to function
        activation = nn.elu
    
        # Reshape to (..., 10, 10, 4)
        x_img = jnp.reshape(x, (*x.shape[:-1], 10, 10, 4))
        # Apply CNN
        cnn = nn.Conv(features=32, kernel_size=(5,5), strides=(1,1), padding='SAME',
                    kernel_init=orthogonal(np.sqrt(2)), bias_init=constant(0.0))(x_img)
        cnn = activation(cnn)
        cnn = nn.LayerNorm()(cnn)
        cnn = nn.Conv(features=64, kernel_size=(3,3), strides=(1,1), padding='SAME',
                    kernel_init=orthogonal(np.sqrt(2)), bias_init=constant(0.0))(cnn)
        cnn = activation(cnn)
        cnn = nn.LayerNorm()(cnn)
        cnn = nn.Conv(features=64, kernel_size=(3,3), strides=(1,1), padding='SAME',
                    kernel_init=orthogonal(np.sqrt(2)), bias_init=constant(0.0))(cnn)
        cnn = activation(cnn)
        cnn = nn.LayerNorm()(cnn)
        cnn = cnn.reshape((*cnn.shape[:-3], -1))  # flatten
        # Reduce dimension
        cnn = nn.Dense(256, kernel_init=orthogonal(np.sqrt(2)), bias_init=constant(0.0))(cnn)
        cnn = activation(cnn)
        cnn = nn.LayerNorm()(cnn)
        # MLP on flattened input
        mlp = nn.Dense(hsize, kernel_init=orthogonal(np.sqrt(2)), bias_init=constant(0.0))(x)
        mlp = activation(mlp)
        mlp = nn.LayerNorm()(mlp)
        mlp = nn.Dense(hsize, kernel_init=orthogonal(np.sqrt(2)), bias_init=constant(0.0))(mlp)
        mlp = activation(mlp)
        mlp = nn.LayerNorm()(mlp)
        # Concatenate
        x = jnp.concatenate([cnn, mlp], axis=-1)
        # Dense layer to combine
        x = nn.Dense(hsize, kernel_init=orthogonal(np.sqrt(2)), bias_init=constant(0.0))(x)
        x = activation(x)
        x = nn.LayerNorm()(x)

        # Policy head
        logits = nn.Dense(self.action_dim, kernel_init=orthogonal(0.01), bias_init=constant(0.0))(x)
        # Value head
        v = nn.Dense(1, kernel_init=orthogonal(1.0), bias_init=constant(0.0))(x)
        
        pi = distrax.Categorical(logits=logits)

        return pi, jnp.squeeze(v, axis=-1)
\end{lstlisting}

The agent makes two particularly interesting and uncommon design decisions in its discovered architecture. Firstly, rather than using a ReLU activation \citep{relu_activation}, as is common when optimising the PPO objective \citep{schulman_proximal_2017,shengyi2022the37implementation}, the agent uses the \textit{exponential linear unit} \citep[ELU]{clevert_fast_2016}. Secondly, while using convolutional neural networks \citep[CNN]{lecun_deep_2015} is common in Atari games \citep{mnih_human-level_2015, shengyi2022the37implementation}, and MLP policies are often used for MinAtar \citep{lu_discovered_2022}, here the ADA combines the two. Specifically, the agent builds an architecture which learns features with both local inductive bias (from the CNN) and global features (from the MLP), effectively overcoming the shortcomings of each approach, and concatenates them before the policy- and value-head. This network architecture performed exceptionally well, on average, in all four MinAtar environments.

\newpage

\section{Additional DiscoBench Analysis}\label{app:discobench_extra}

\subsection{Success@3 Results}\label{app:pass_k_results}

We believe that the principal objective of ADA research should be to develop robust and performant algorithm discovery agents. As such, in the main body of our text we report aggregated success metrics over independent seeds of algorithm discovery; an ADA which only produces a valid algorithm in one out of its three seeds would receive a success rate of $33\%$. However, such reporting introduces the question of whether tasks are too difficult to provide signal for either evaluation or optimisation. As such, in Table \ref{tab:success3} we report \textit{success@3}. This acts as a complement to \cref{tab:elos} and verifies that, for most tasks, agents are able to create \textit{at least one} valid solution.

\begin{table*}[h]
\centering
\caption{\textit{Success@3} rates in DiscoBench. Bold indicates best performance.}
\label{tab:success3}
\begin{tabular}{@{}lcc@{}}
\toprule
& \multicolumn{1}{c}{\textit{DiscoBench (Single Edit)}} & \multicolumn{1}{c}{\textit{DiscoBench (All Edit)}} \\
\cmidrule(lr){2-2} \cmidrule(lr){3-3}
\textbf{Model} & \textbf{Success@3} & \textbf{Success@3} \\
\midrule
GPT-OSS 120B & 82.1\% & 26.5\% \\
Devstral2 & 79.7\% & 46.2\% \\
Deepseek-v3.2 & \textbf{89.0}\textbf{\%} & \textbf{63.1\%} \\
\bottomrule
\end{tabular}%
\end{table*}
\textit{Success@3} illuminates that ADAs are \textit{able} to produce valid solutions to most tasks, albeit not consistently. As such, the large gap between average success over 3 seeds, compared to \textit{success@3} broadly stems from low ADA robustness as opposed to an insurmountable set of tasks in DiscoBench. Furthermore, by extrapolating these results from DiscoBench to DiscoGen, they confirm that there is \textit{signal} for ADA optimisation; as long as models can sometimes produce valid solutions, there is sufficient evidence to differentiate between meta-loop trajectories, and thus to optimise ADAs.

\subsection{Successful Seeds Analysis} \label{app:until_success}

We include additional results over a subset of domains, where affordable and feasible, until each ADA has 3 \textbf{successful} seeds (i.e., 3 independent runs with syntactically valid submissions). Such results allow us to disentangle Elo penalties from \textit{task failure} and Elo comparisons from \textit{better or worse solutions}. These results are included in Table \ref{tab:elos_until_success}.
\begin{table*}[h]
\centering
\caption{ADA evaluation performance in DiscoBench (Elo Scores with 95\% CIs). Bold indicates best mean performance.\vspace{-2pt}}
\label{tab:elos_until_success}
\begin{tabular}{@{}lcc@{}}
\toprule
& \multicolumn{2}{c}{\textit{DiscoBench (Until Success)}} \\
\cmidrule(lr){2-3} 
\textbf{Model} & \textbf{Meta-Train} & \textbf{Meta-Test} \\
\midrule
Baseline (All Fixed)  & \textbf{1080} \tiny{[1050, 1112]} & \textbf{1153} \tiny{[1122, 1187]} \\
GPT-OSS 120B  & 885 \tiny{[850, 917]}             & 894 \tiny{[867, 924]}             \\
Devstral2     & 986 \tiny{[960, 1014]}            & 954 \tiny{[928, 982]}             \\
Deepseek-v3.2 & 1049 \tiny{[1026, 1074]}          & 998 \tiny{[976, 1020]}            \\
\bottomrule
\end{tabular}%
\vspace{-13pt}
\end{table*}

We find that, even after accounting for failed runs (as in our DiscoBench results from Section \ref{sec:discobench_results}), models are unable to consistently outperform the all fixed baseline; in fact, the baseline is arguably \textit{more} dominant in meta-test performance. While improving robustness of ADAs is necessary as above, it is also clearly essential to develop ADAs which produce \textit{better} solutions when they do succeed.

\newpage

\section{Per-Task DiscoBench Results} \label{app:discobench_results}

Here, we present per-task DiscoBench results, which are aggregated into a per-model ELO score over meta-train and meta-test sets. These are separated into \textit{DiscoBench Single}, \textit{DiscoBench All}, and \textit{DiscoBench Single (Until Success)}. The metric is provided on the x-axis; in some cases, said metric should be \textit{maximised}, and in some it should be \textit{minimised}.

Results are reported over 3 or less meta-seeds, depending on how many runs were successful for each model. We highlight the baseline scores in red; these are the scores obtained by the default implementation for each task, when all files are set to fixed. For all results, we calculate bootstrap stratified confidence intervals (95\%) and interquartile means, following \citet{agarwal2021deep}.

In our \textit{Until Success} experiments (Appendices \ref{sec:until_success_id} \& \ref{sec:until_success_mt}), we omit a small number of tasks in which we were unable to reliably produce three seeds for every model without incurring excessive cost.


\subsection{DiscoBench (Single Edit) -- Meta-Train}
\label{sec:one_change_id}

\begin{figure}[htbp]
\centering
\setlength{\lineskip}{0pt}
\includegraphics[width=0.48\textwidth]{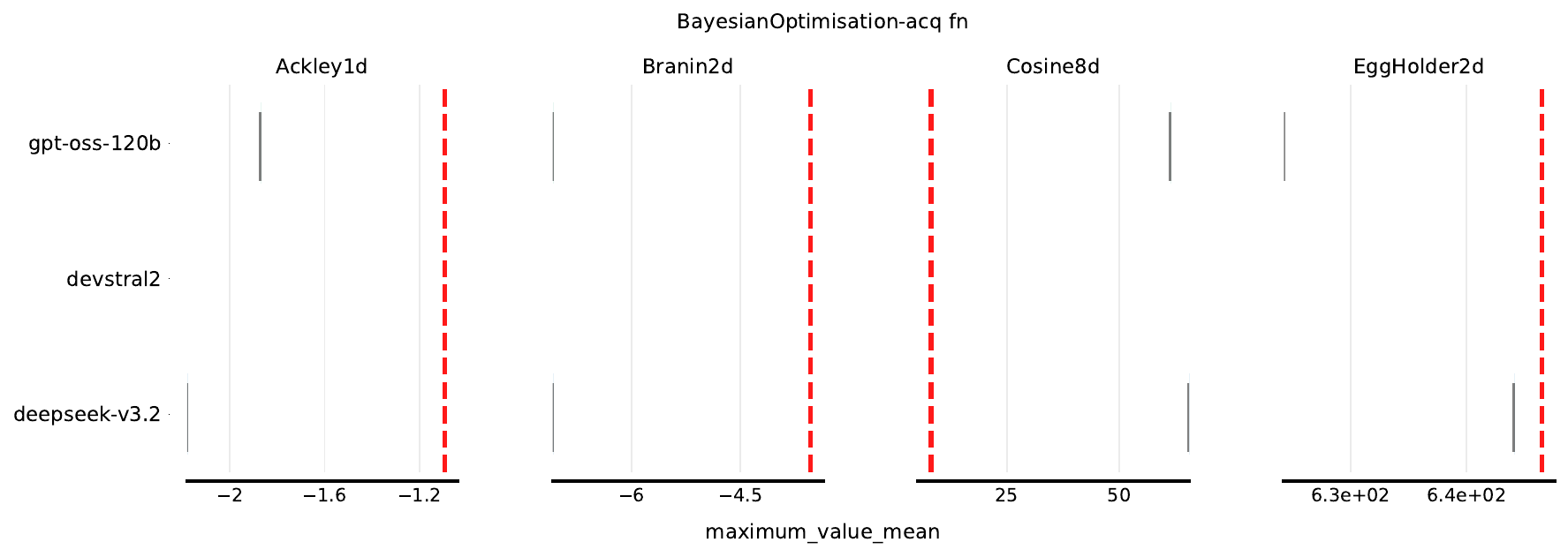}%
\hfill%
\includegraphics[width=0.48\textwidth]{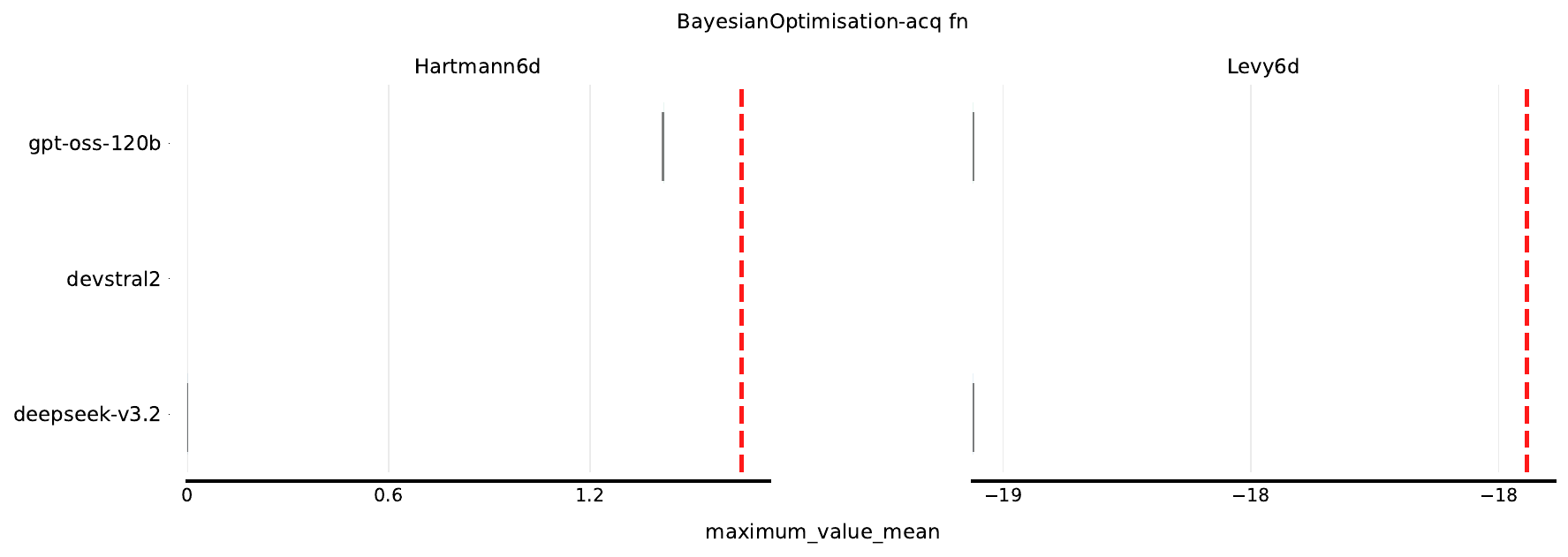}%
\\[0.5em]
\includegraphics[width=0.48\textwidth]{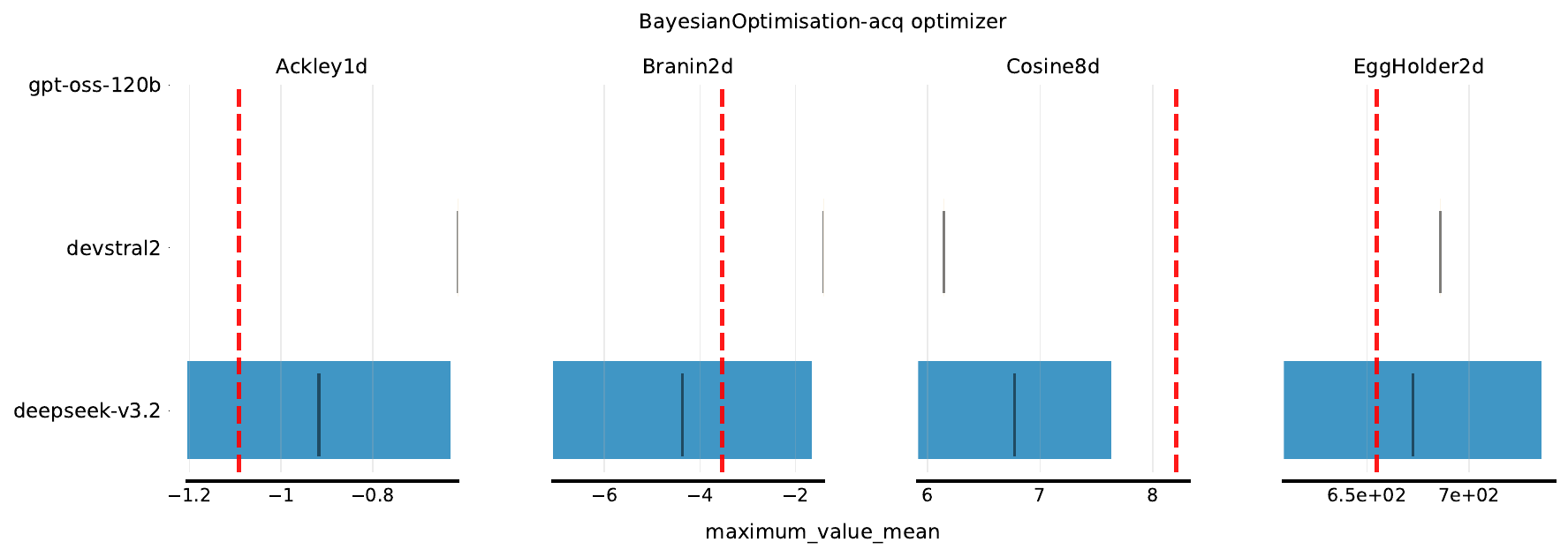}%
\hfill%
\includegraphics[width=0.48\textwidth]{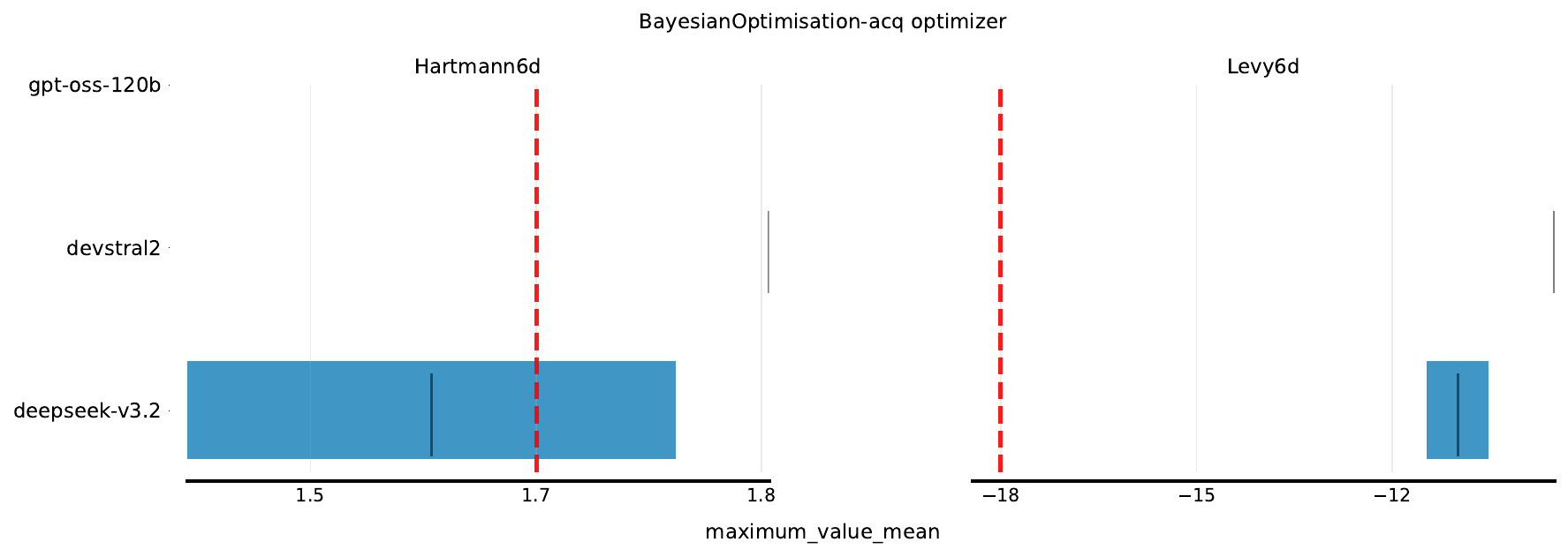}%
\\[0.5em]
\includegraphics[width=0.48\textwidth]{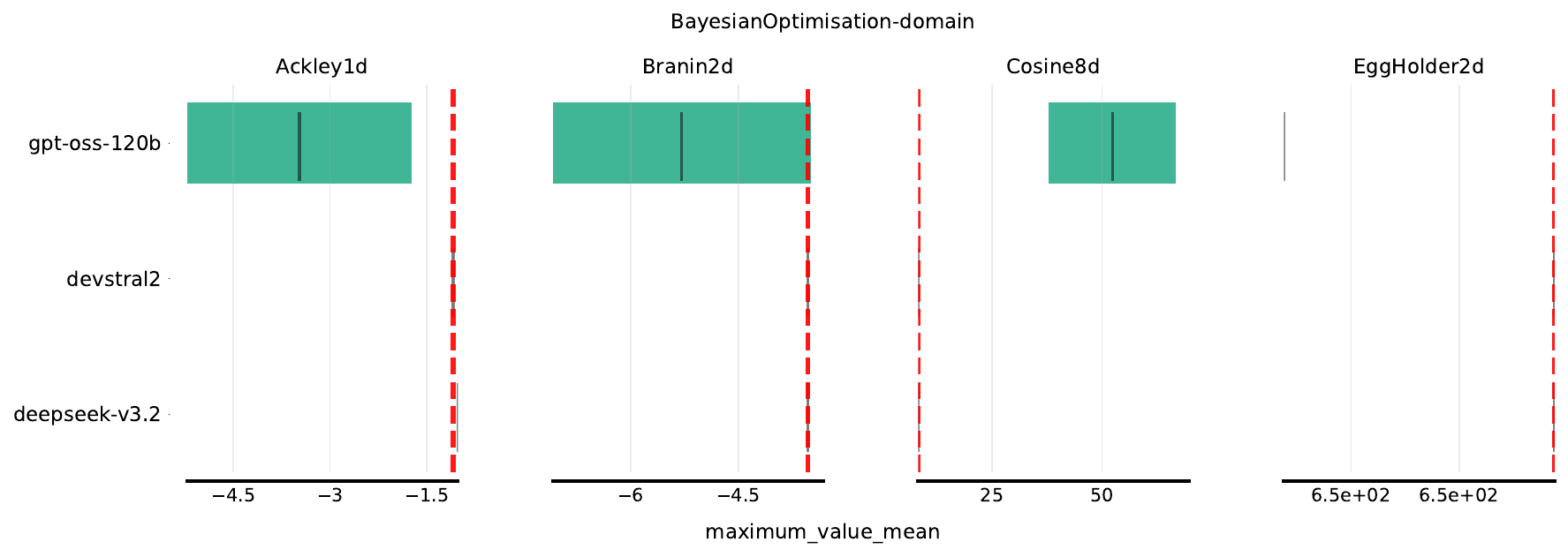}%
\hfill%
\includegraphics[width=0.48\textwidth]{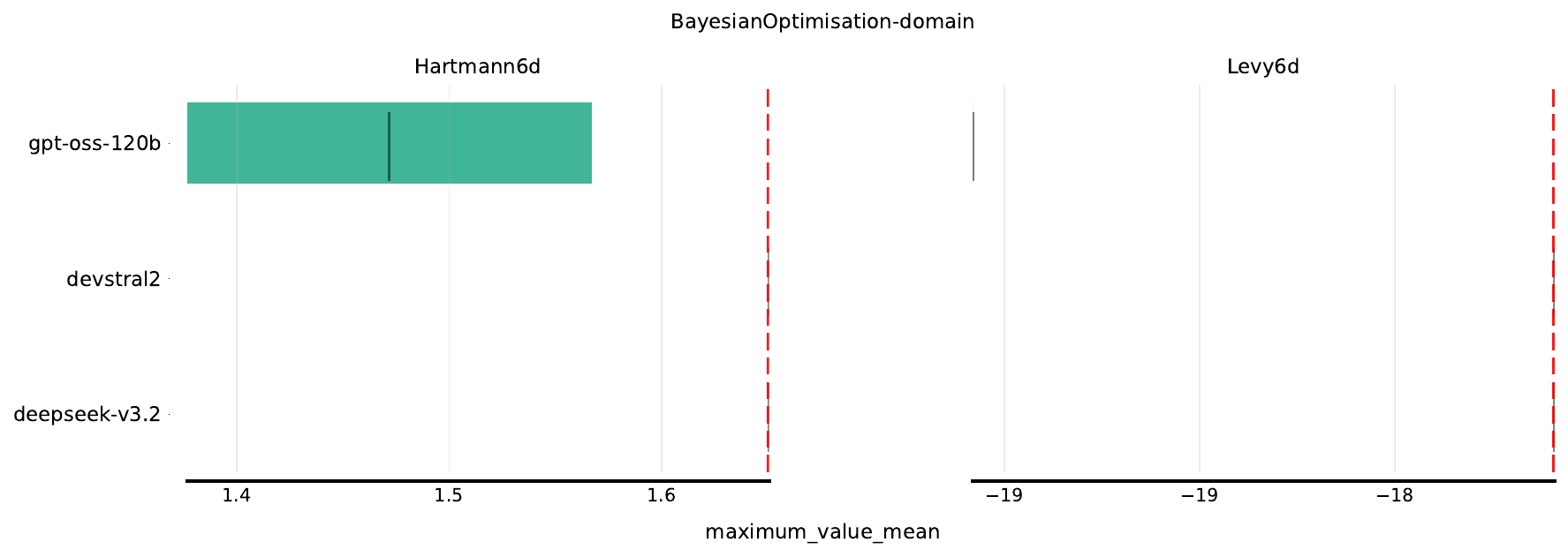}%
\\[0.5em]
\includegraphics[width=0.48\textwidth]{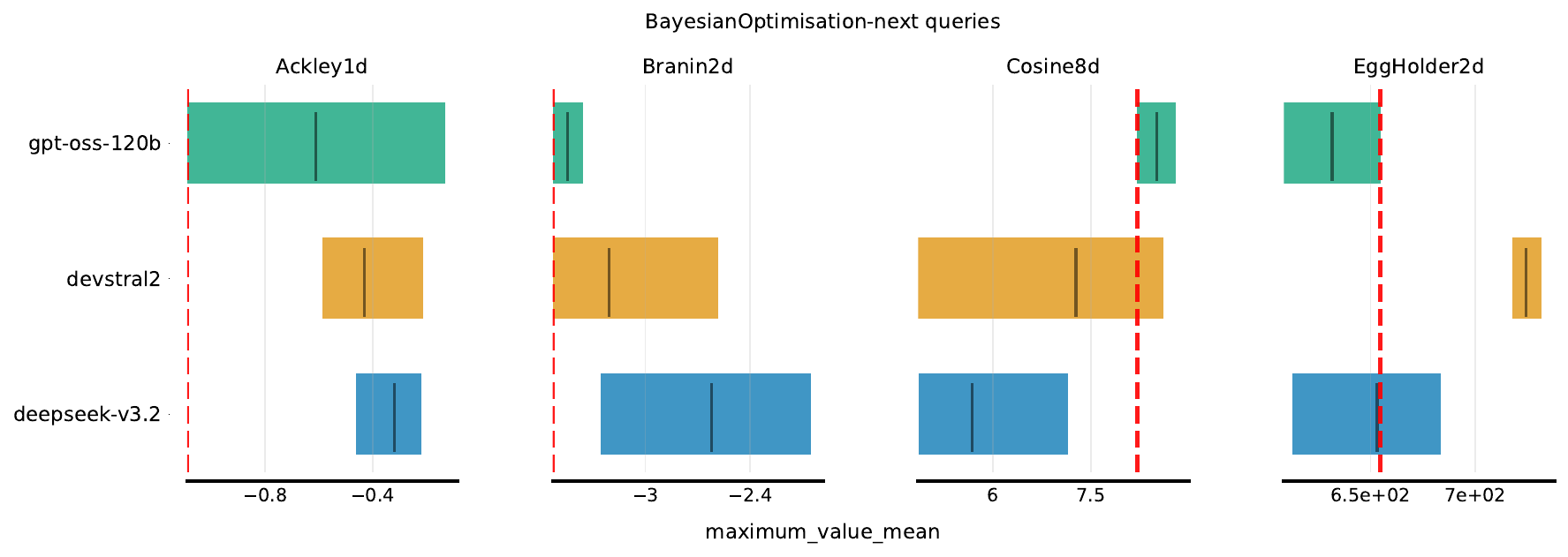}%
\hfill%
\includegraphics[width=0.48\textwidth]{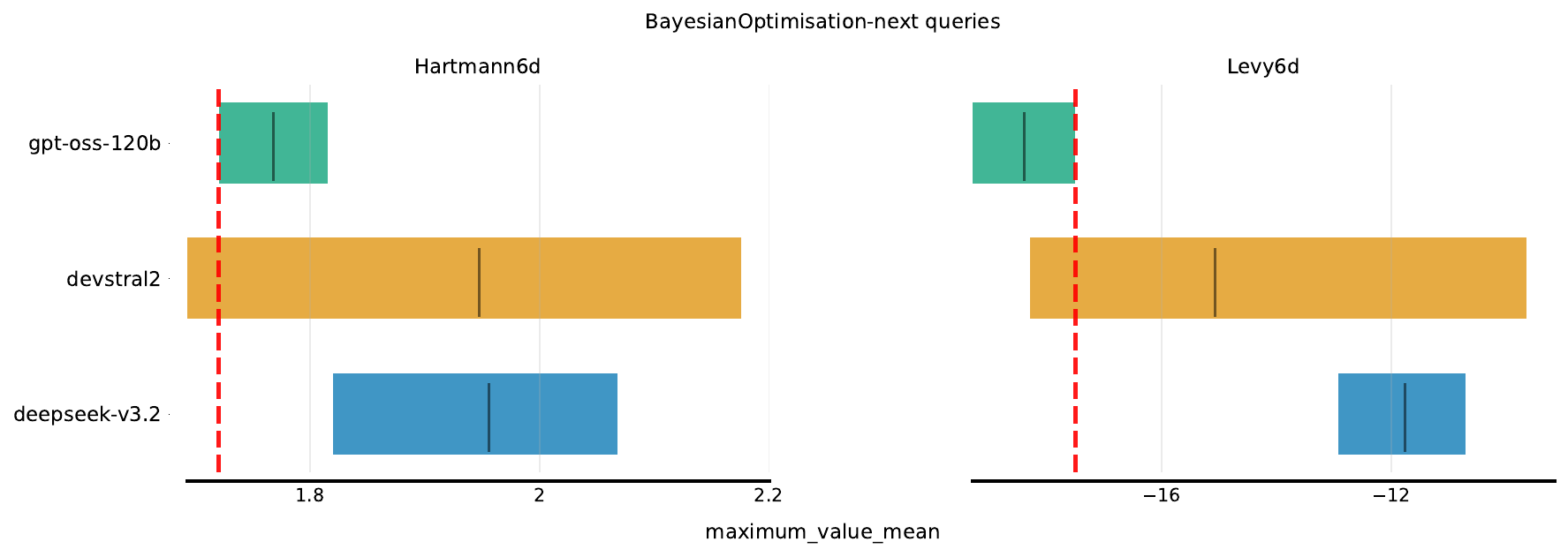}%
\\[0.5em]
\includegraphics[width=0.48\textwidth]{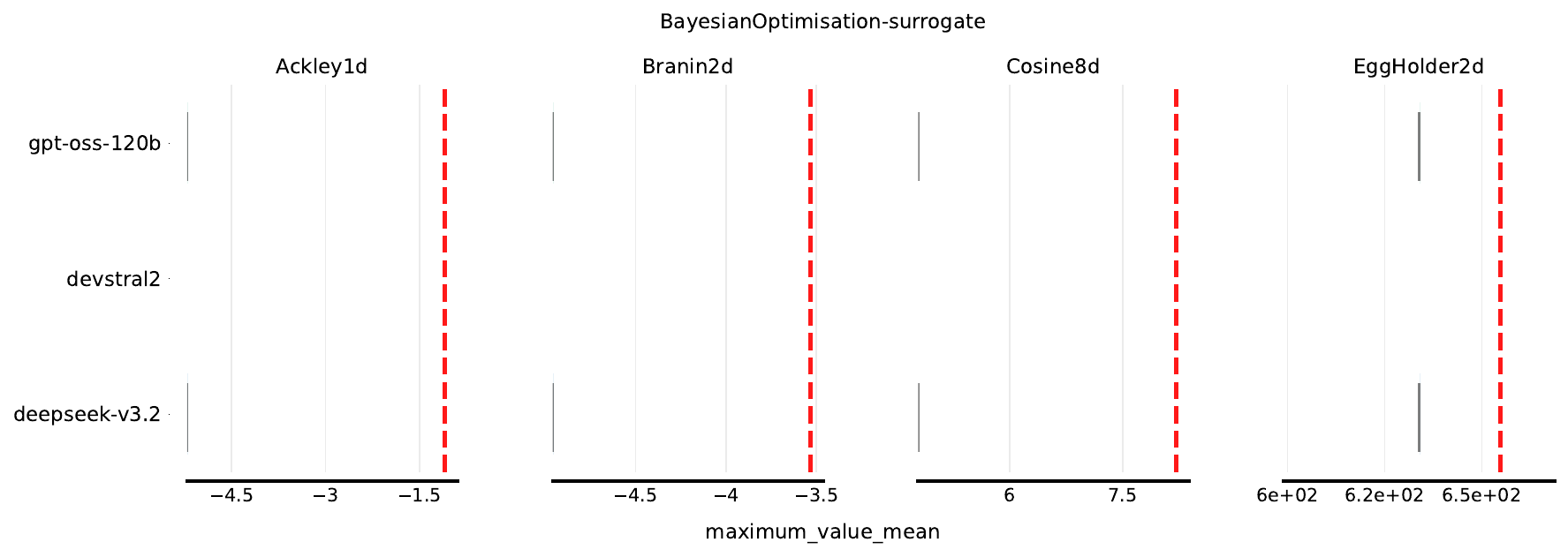}%
\hfill%
\includegraphics[width=0.48\textwidth]{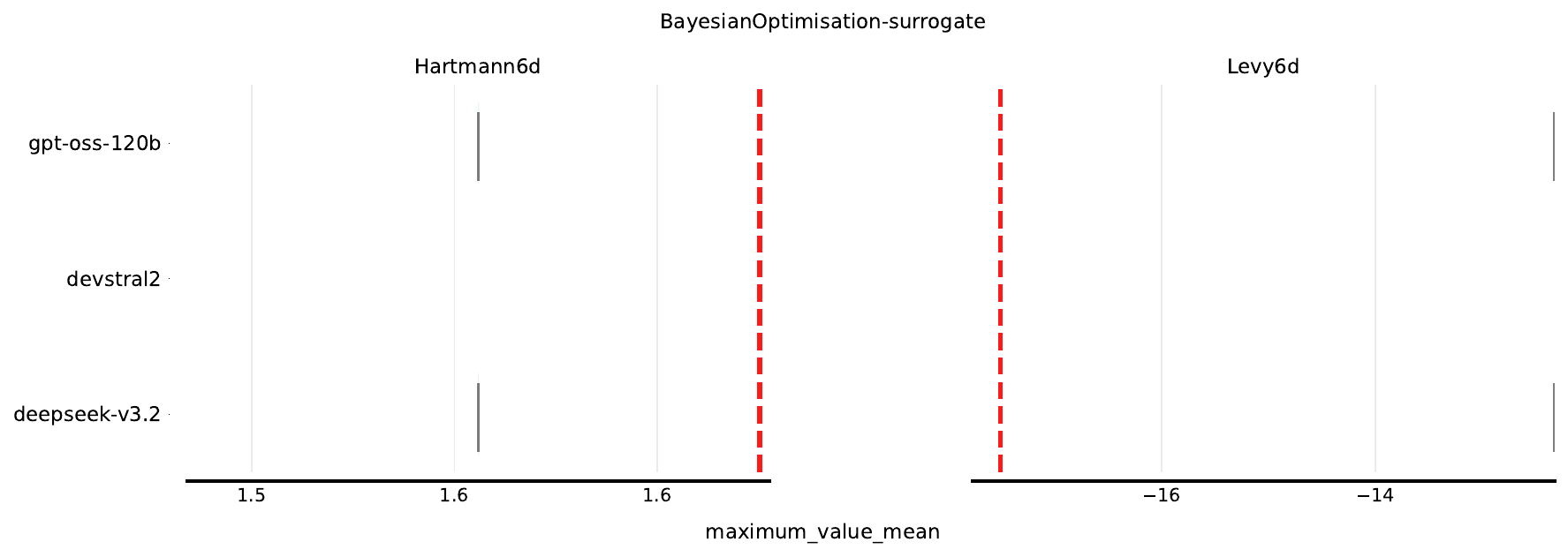}%
\caption{DiscoBench (Single Edit) results on Meta-Train tasks. (Part 1/7)}
\label{fig:one_change_id_1}
\end{figure}
\clearpage

\begin{figure}[htbp]
\centering
\setlength{\lineskip}{0pt}
\includegraphics[width=0.48\textwidth]{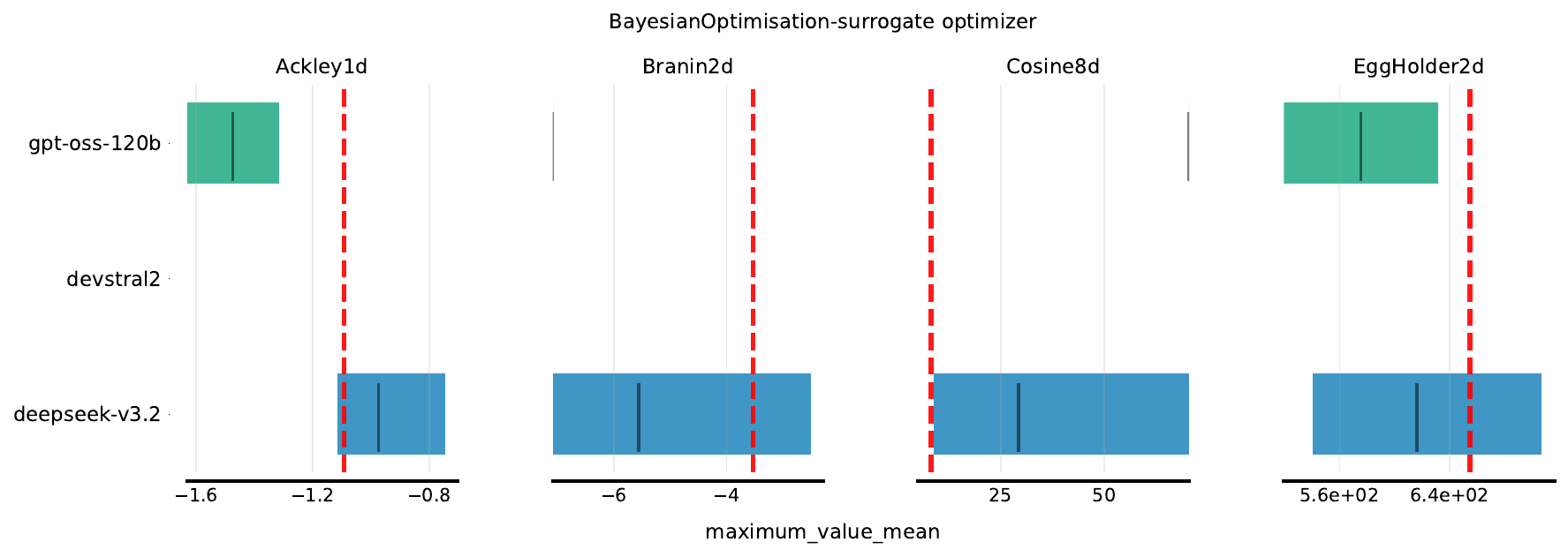}%
\hfill%
\includegraphics[width=0.48\textwidth]{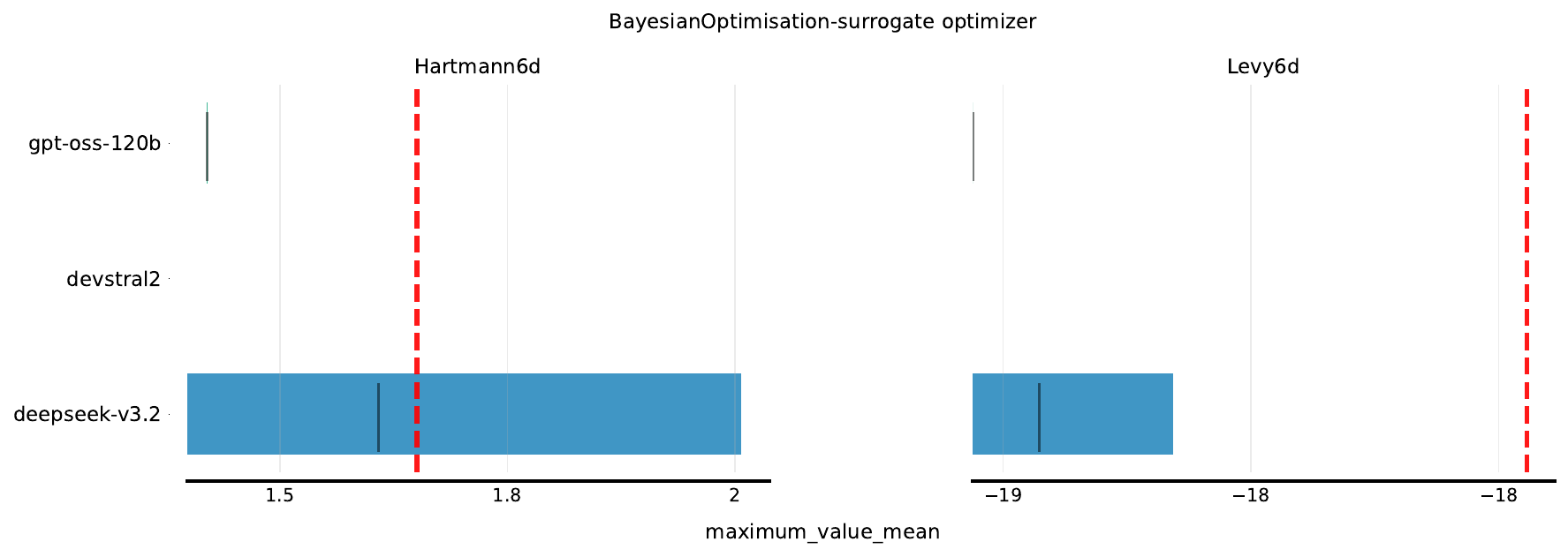}%
\\[0.5em]
\includegraphics[width=0.48\textwidth]{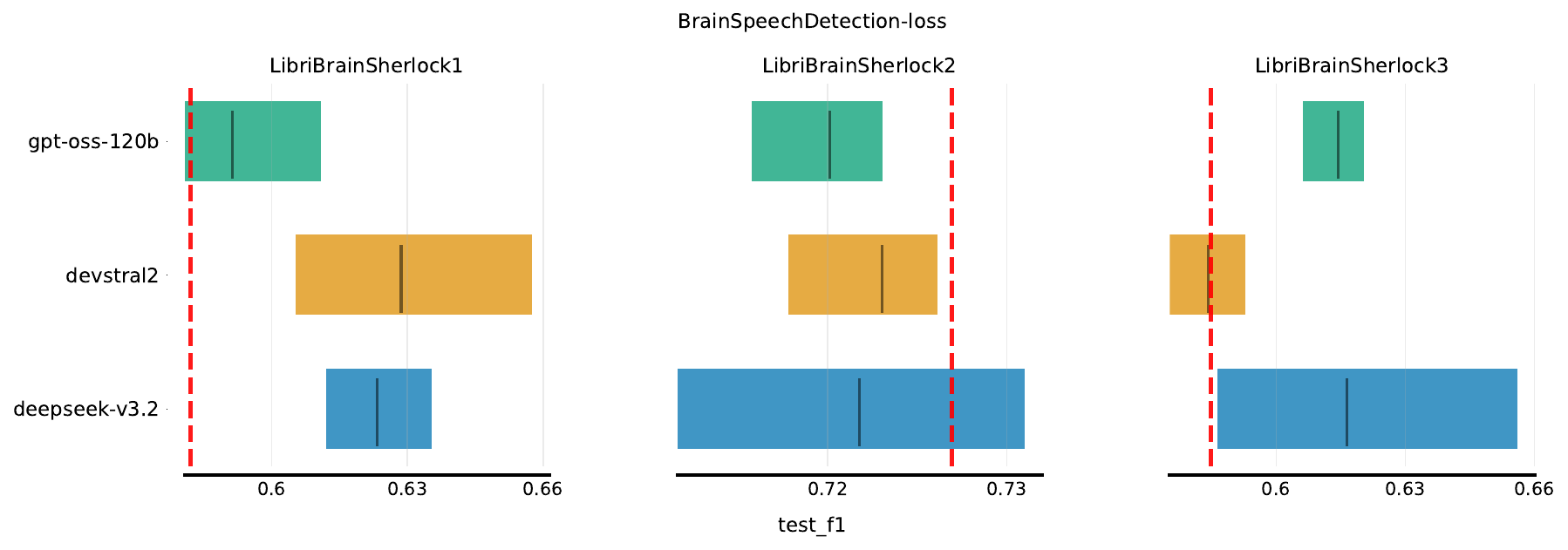}%
\hfill%
\includegraphics[width=0.48\textwidth]{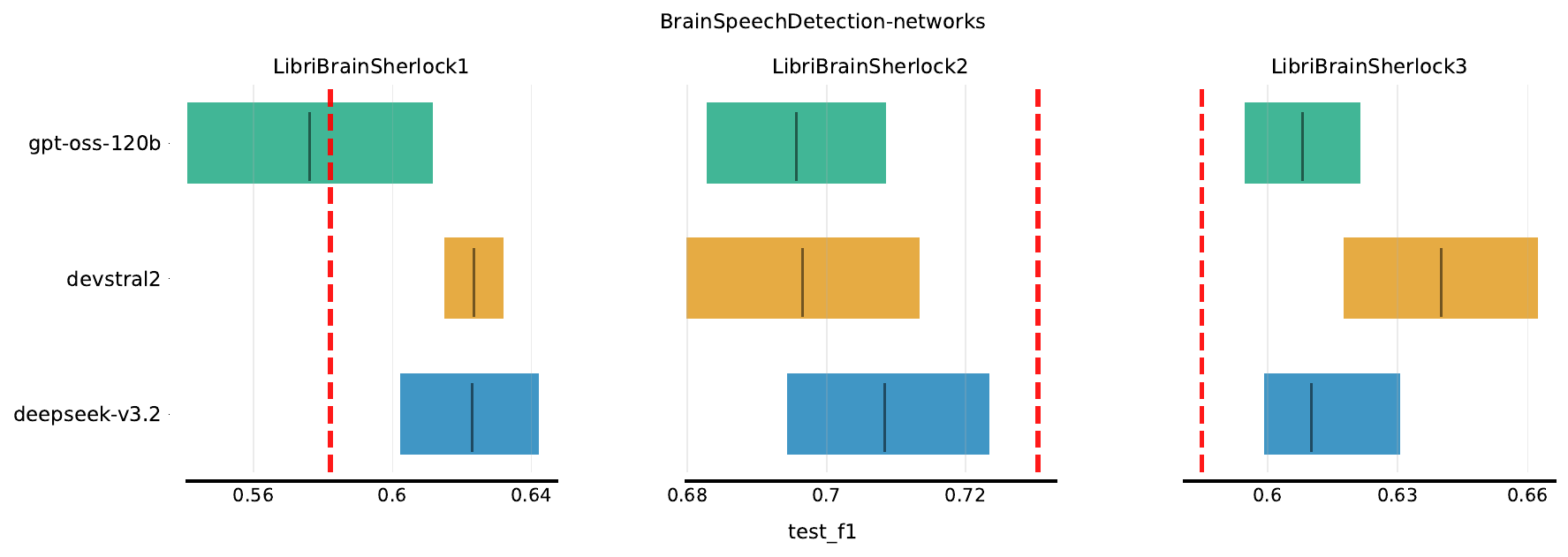}%
\\[0.5em]
\includegraphics[width=0.48\textwidth]{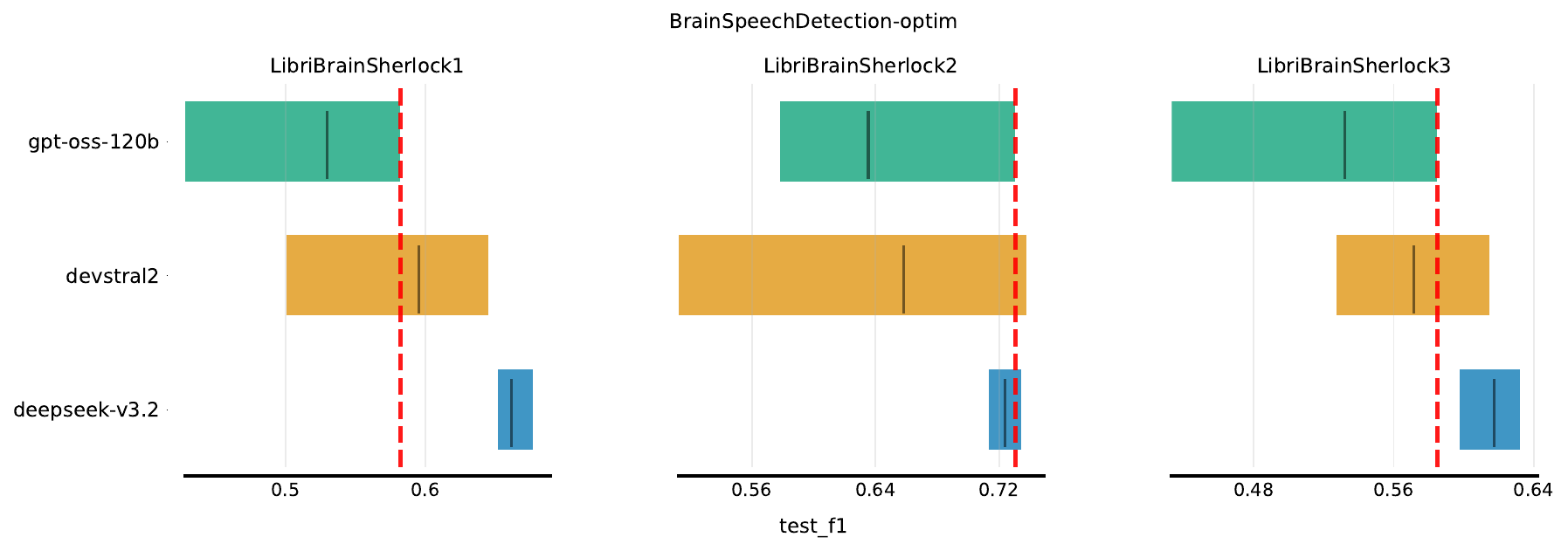}%
\hfill%
\includegraphics[width=0.48\textwidth]{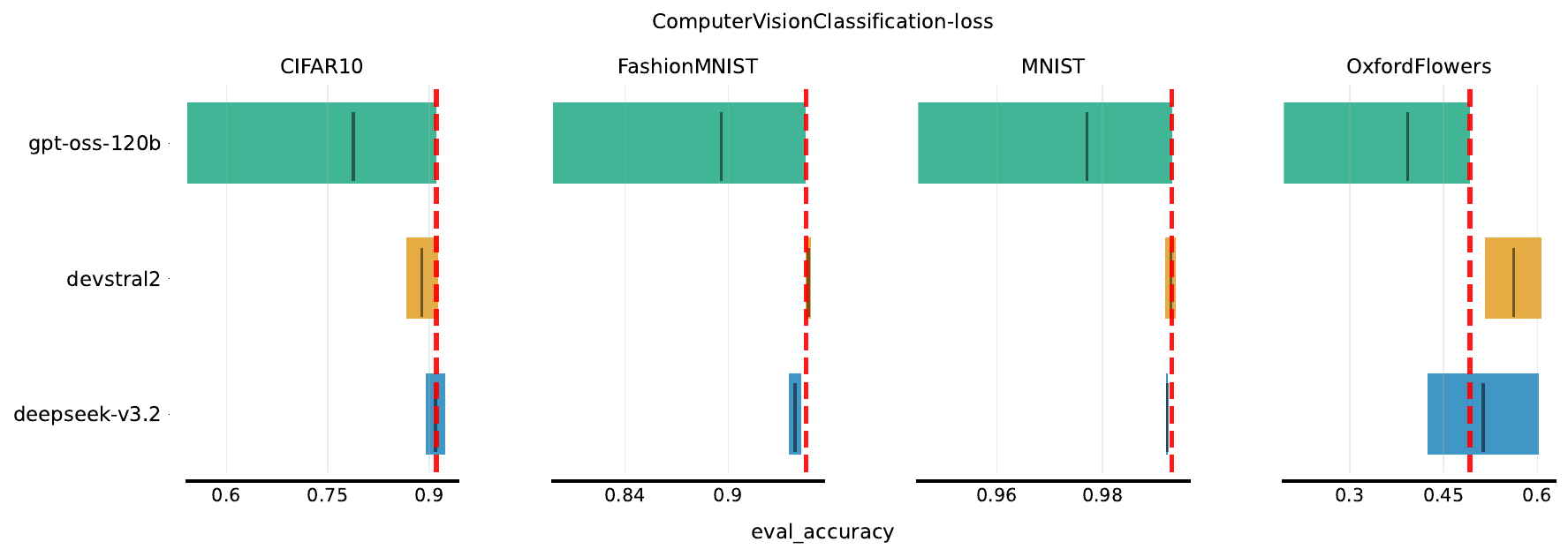}%
\\[0.5em]
\includegraphics[width=0.48\textwidth]{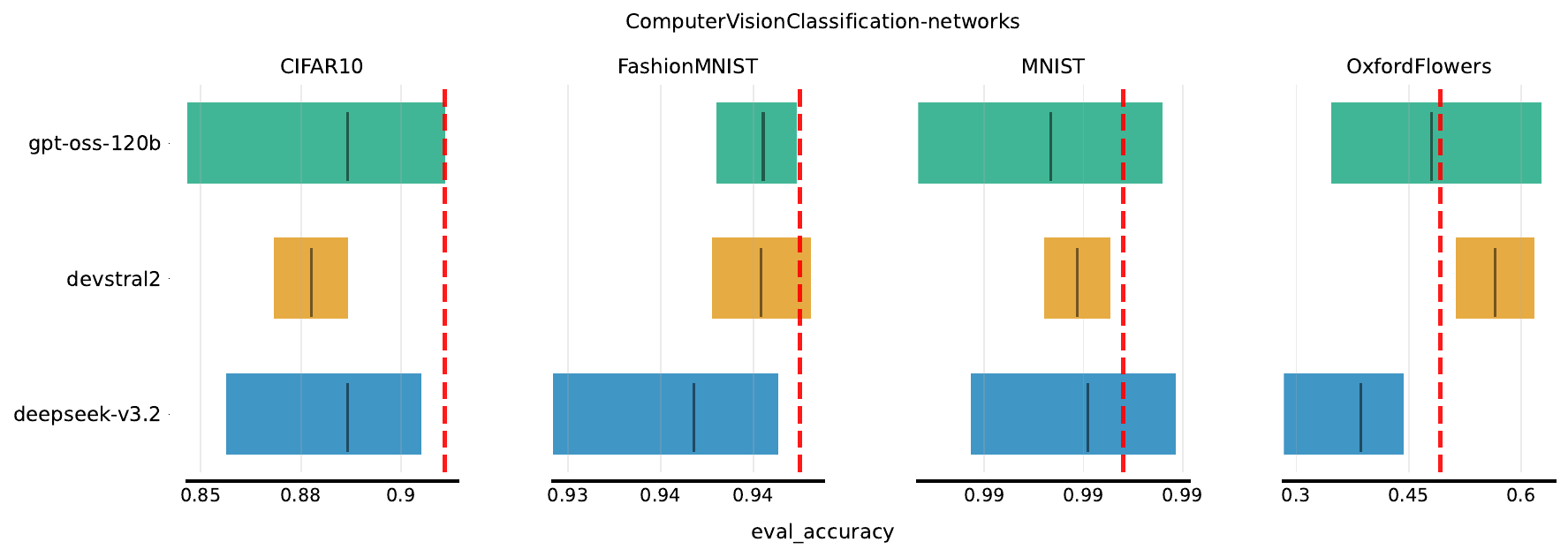}%
\hfill%
\includegraphics[width=0.48\textwidth]{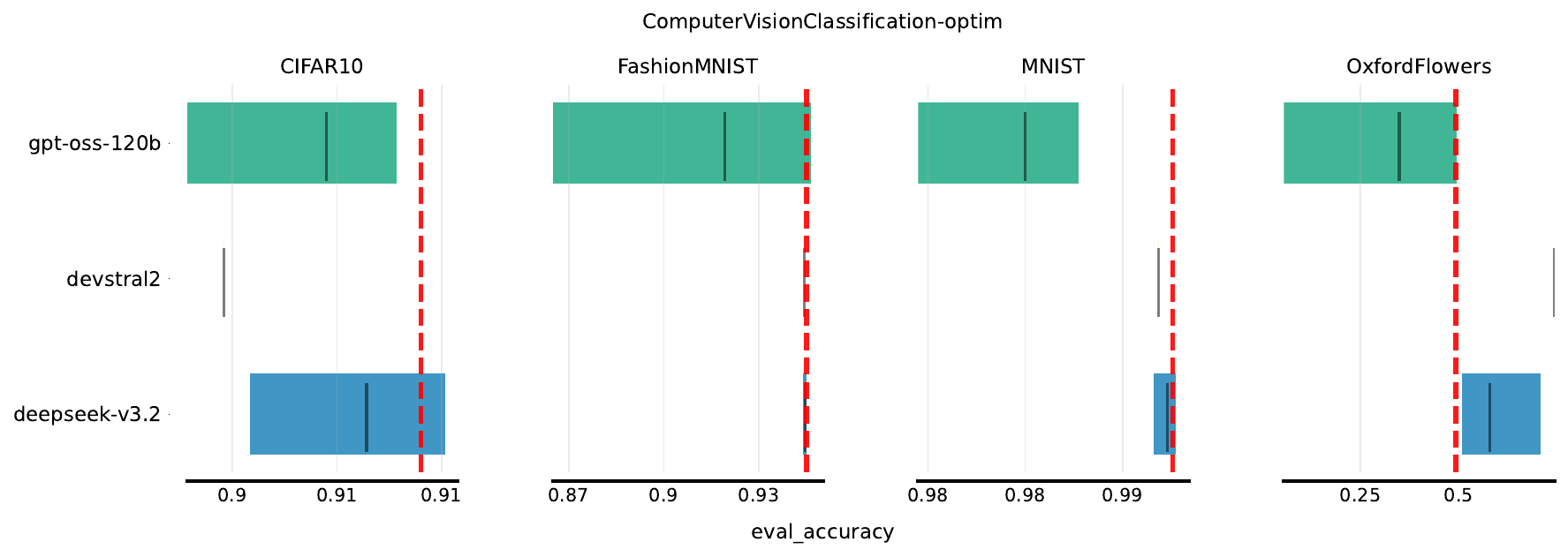}%
\\[0.5em]
\includegraphics[width=0.48\textwidth]{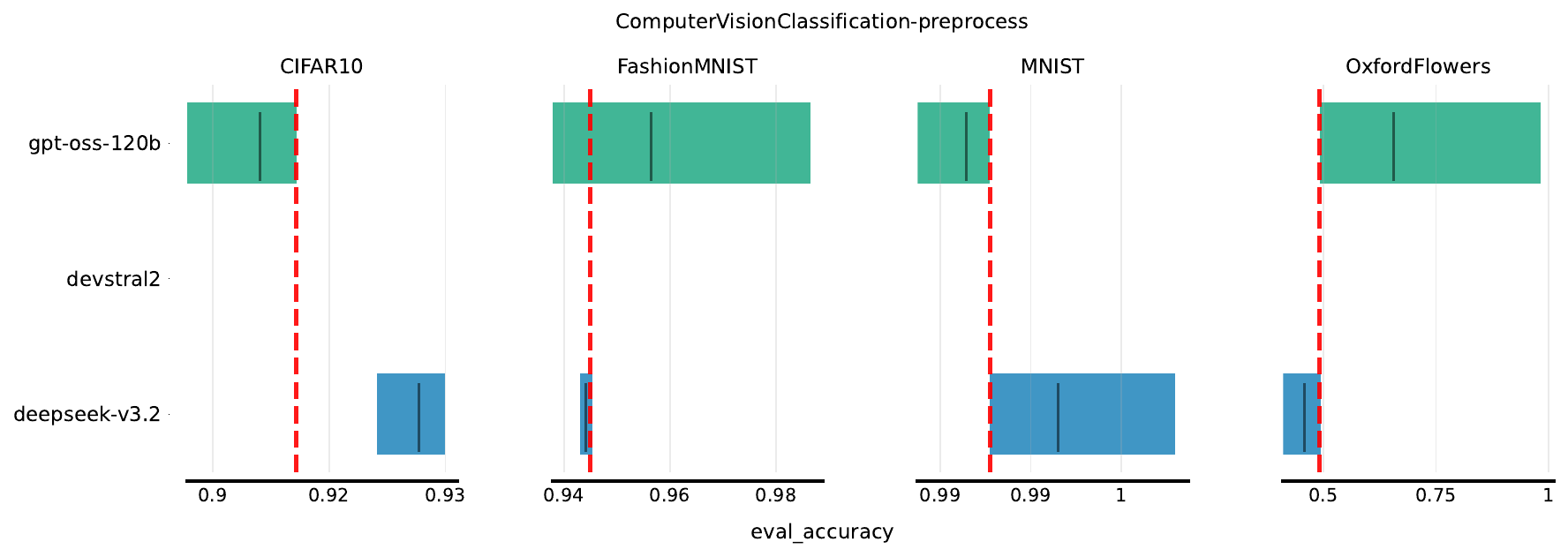}%
\hfill%
\includegraphics[width=0.48\textwidth]{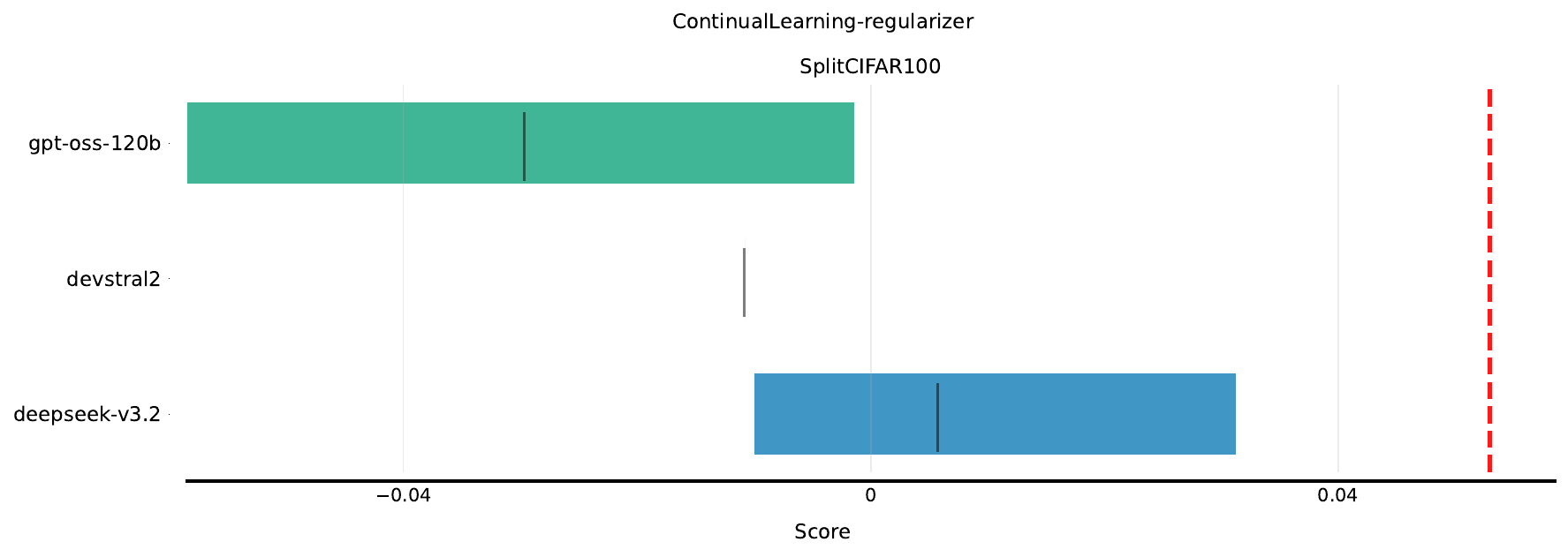}%
\\[0.5em]
\includegraphics[width=0.48\textwidth]{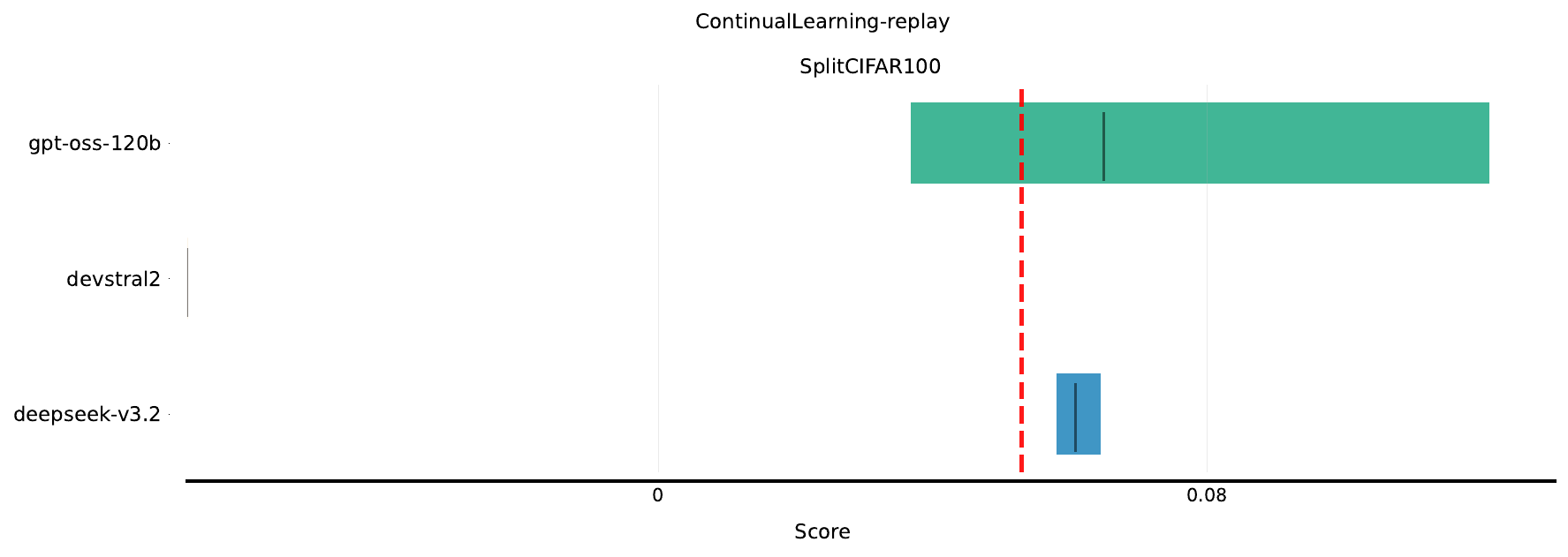}%
\hfill%
\includegraphics[width=0.48\textwidth]{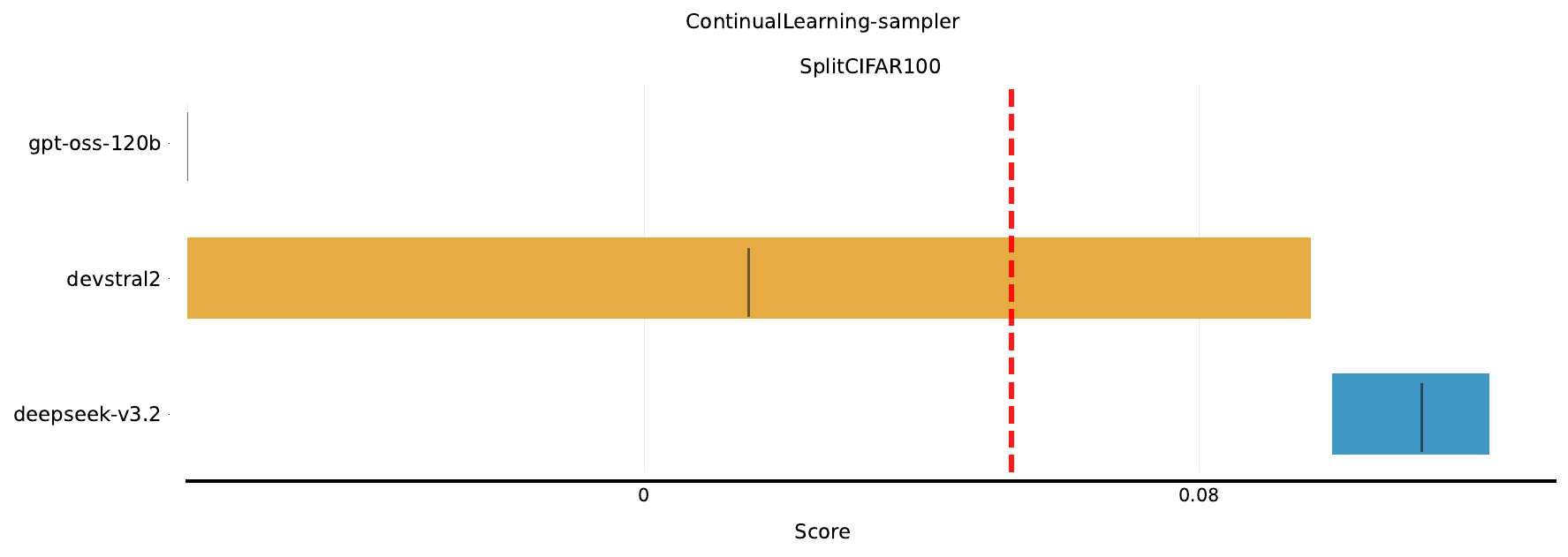}%
\caption{DiscoBench (Single Edit) results on Meta-Train tasks. (Part 2/7)}
\label{fig:one_change_id_2}
\end{figure}
\clearpage

\begin{figure}[htbp]
\centering
\setlength{\lineskip}{0pt}
\includegraphics[width=0.48\textwidth]{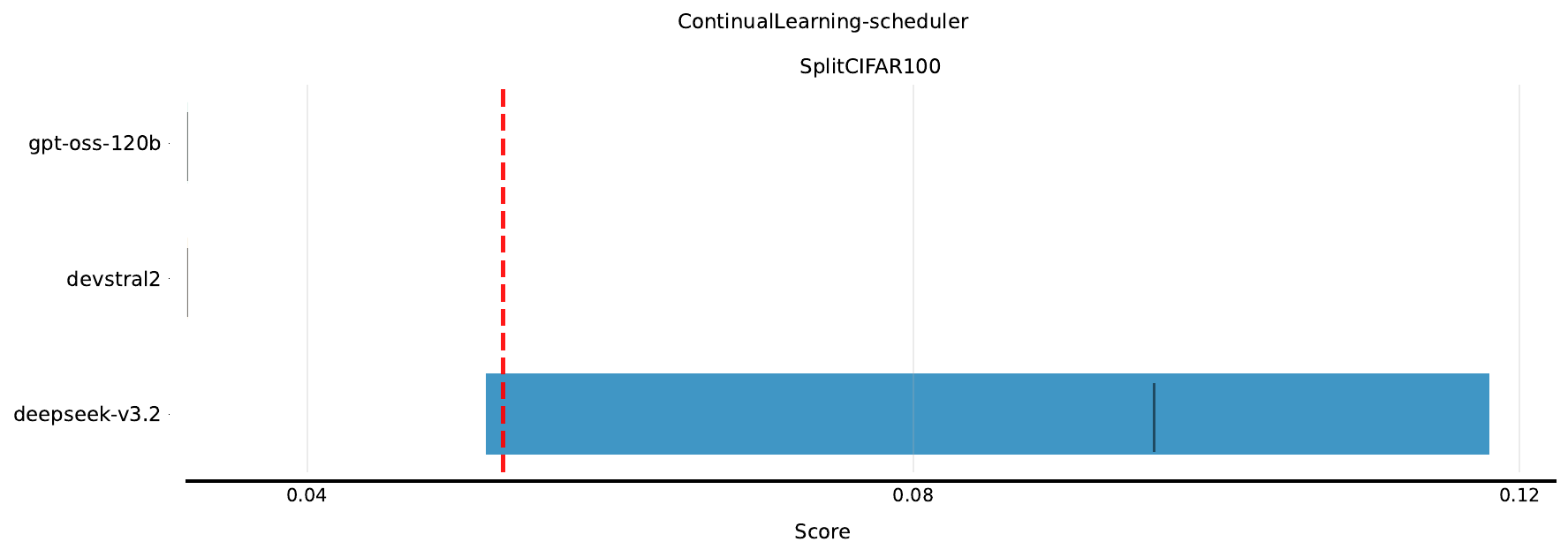}%
\hfill%
\includegraphics[width=0.48\textwidth]{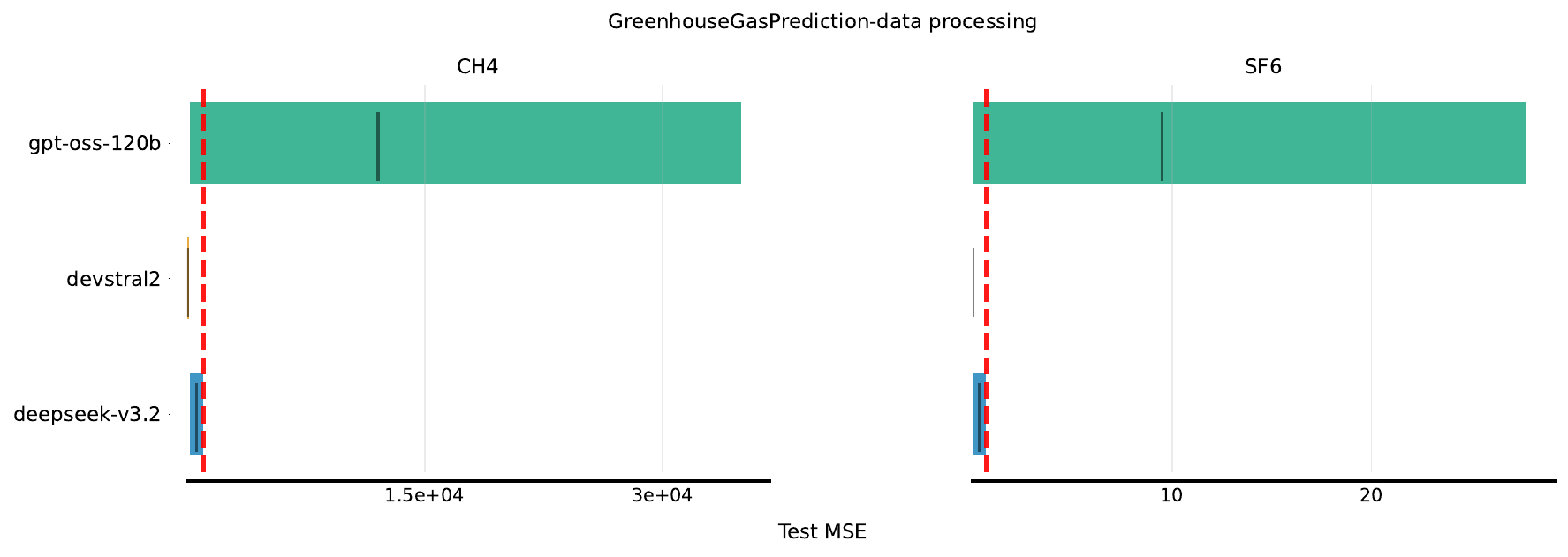}%
\\[0.5em]
\includegraphics[width=0.48\textwidth]{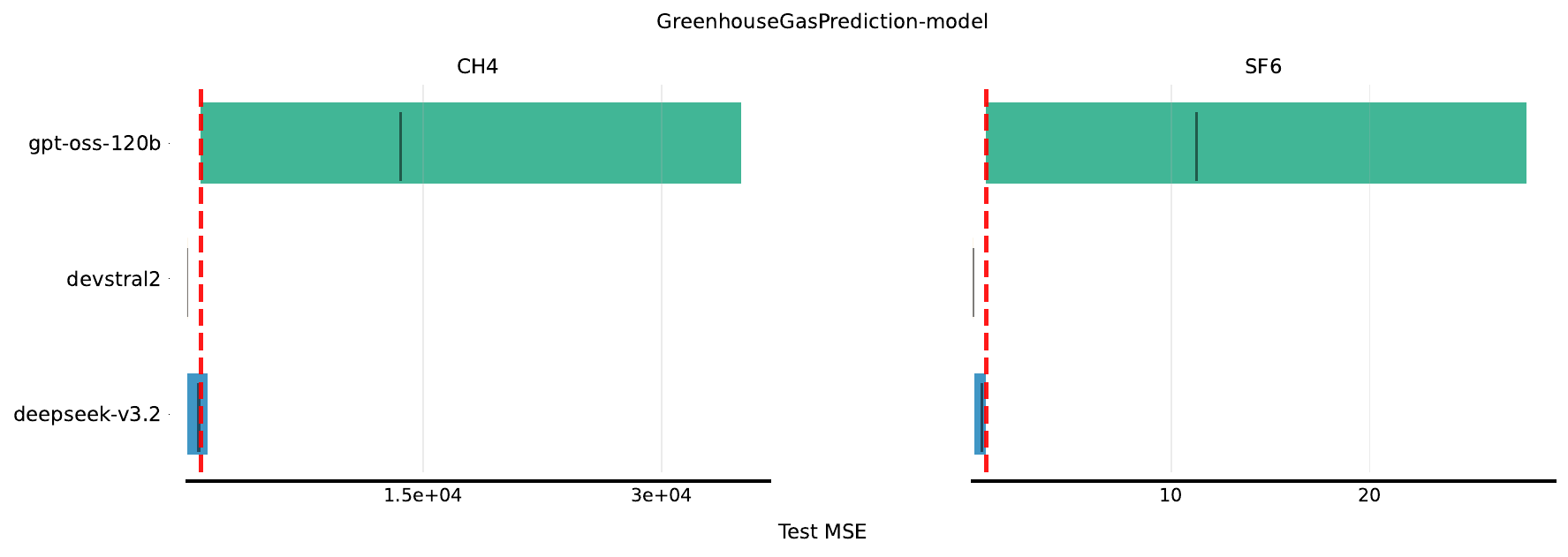}%
\hfill%
\includegraphics[width=0.48\textwidth]{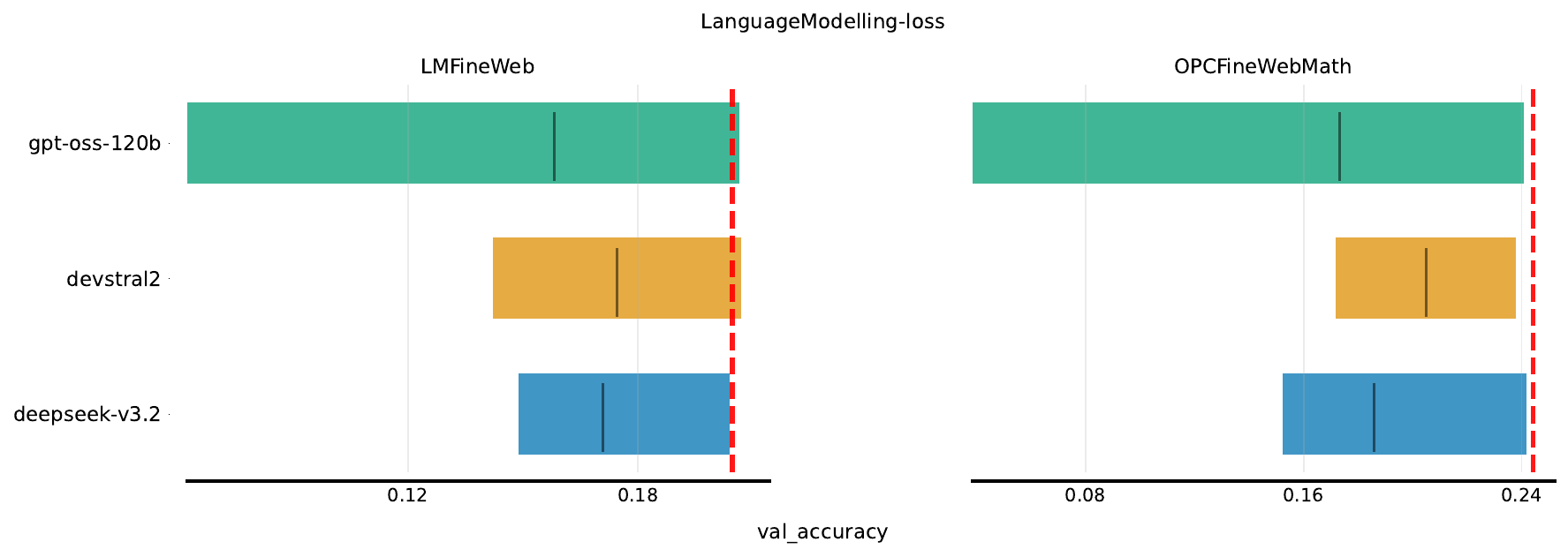}%
\\[0.5em]
\includegraphics[width=0.48\textwidth]{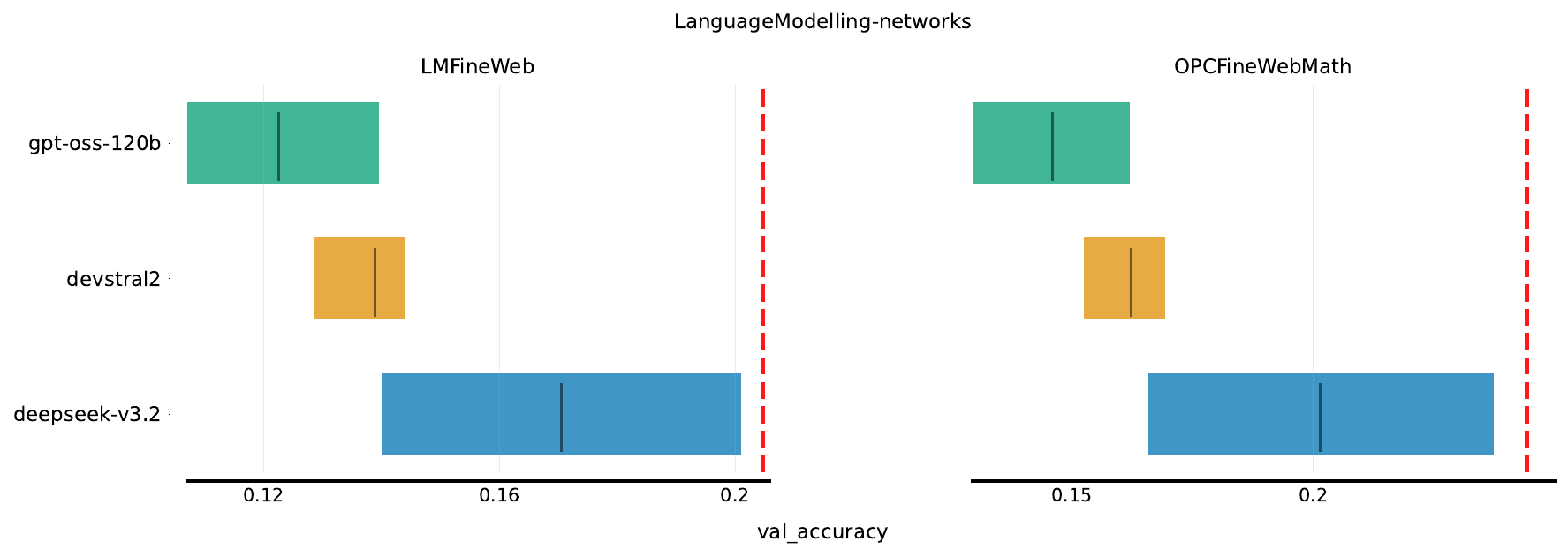}%
\hfill%
\includegraphics[width=0.48\textwidth]{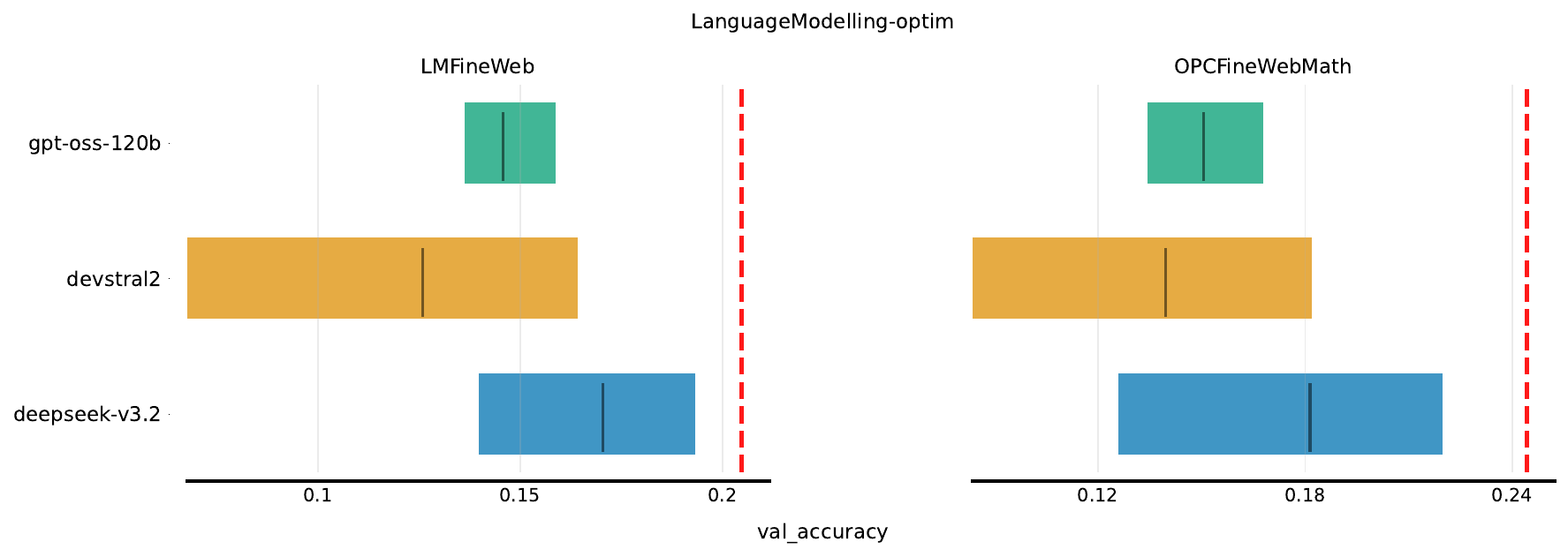}%
\\[0.5em]
\includegraphics[width=0.48\textwidth]{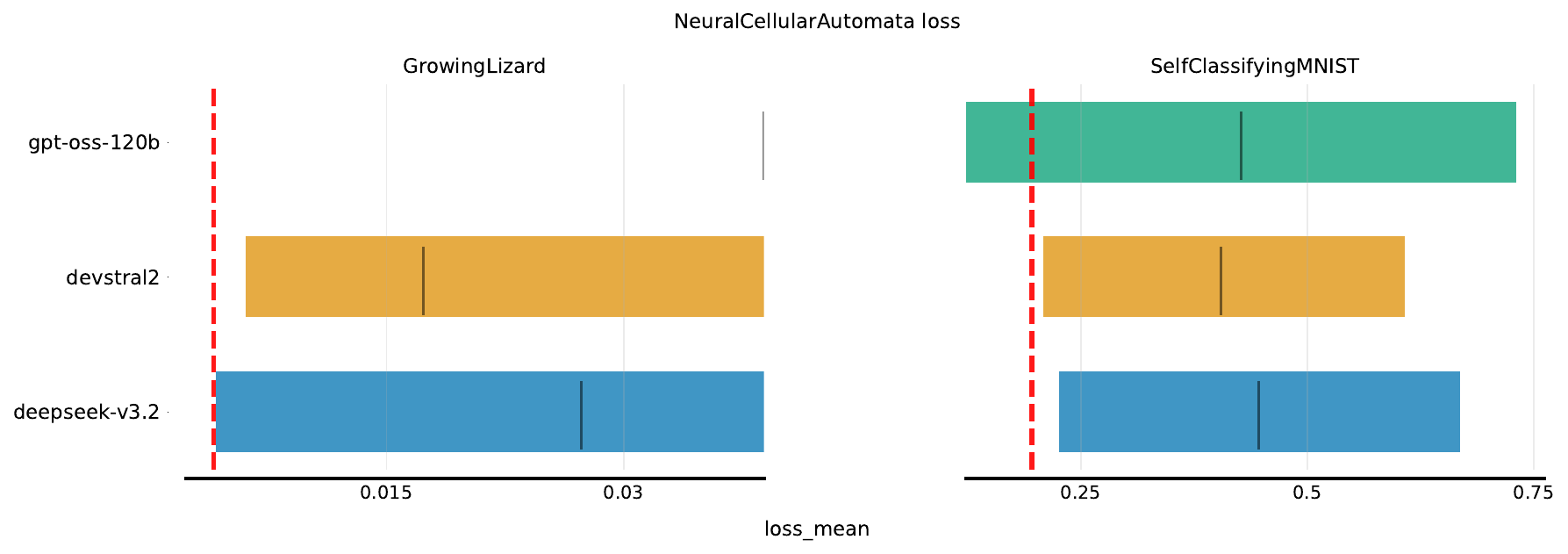}%
\hfill%
\includegraphics[width=0.48\textwidth]{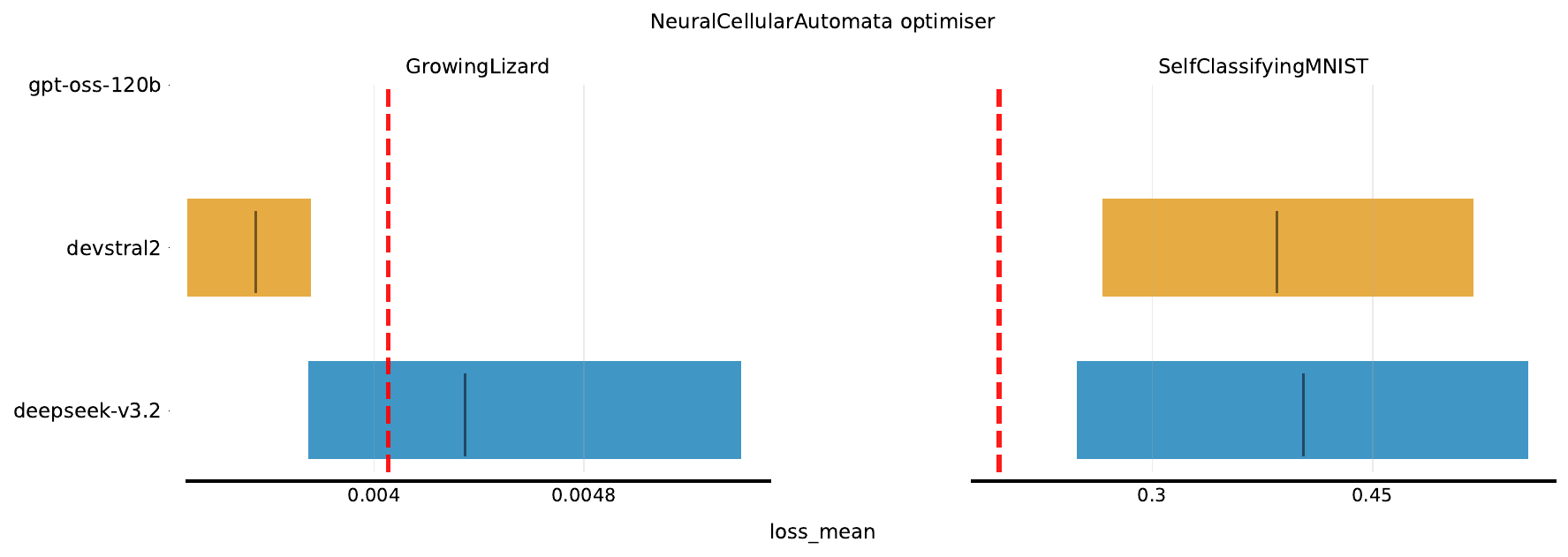}%
\\[0.5em]
\includegraphics[width=0.48\textwidth]{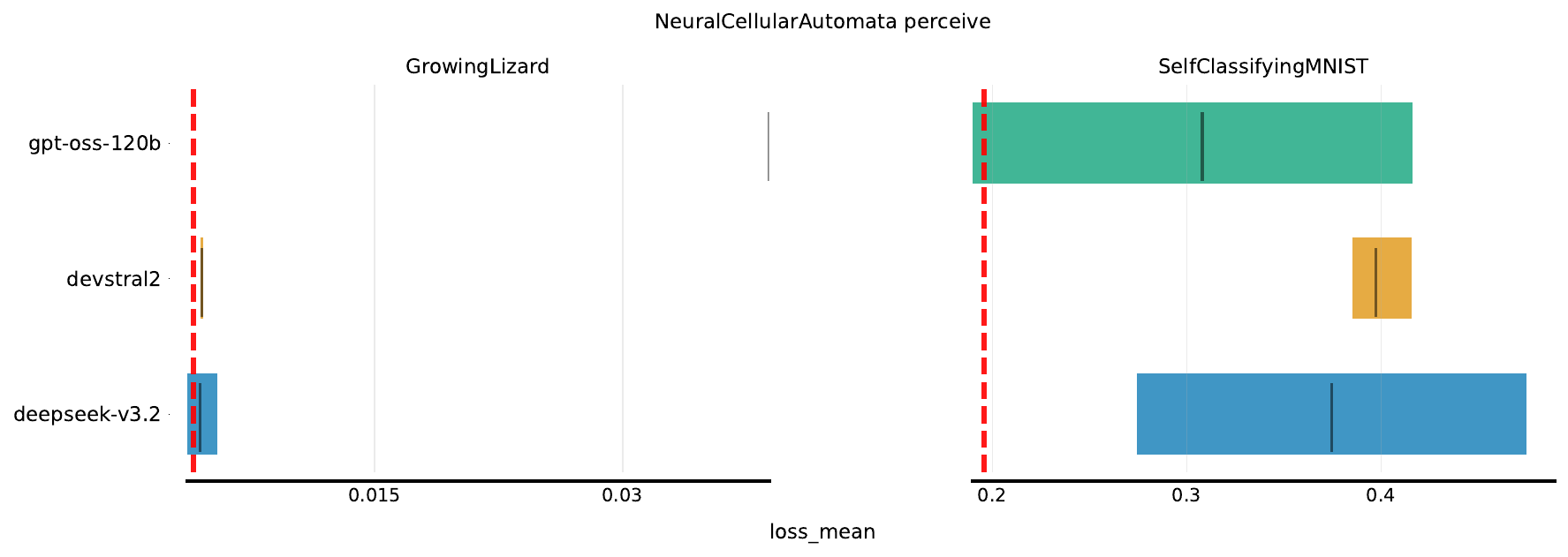}%
\hfill%
\includegraphics[width=0.48\textwidth]{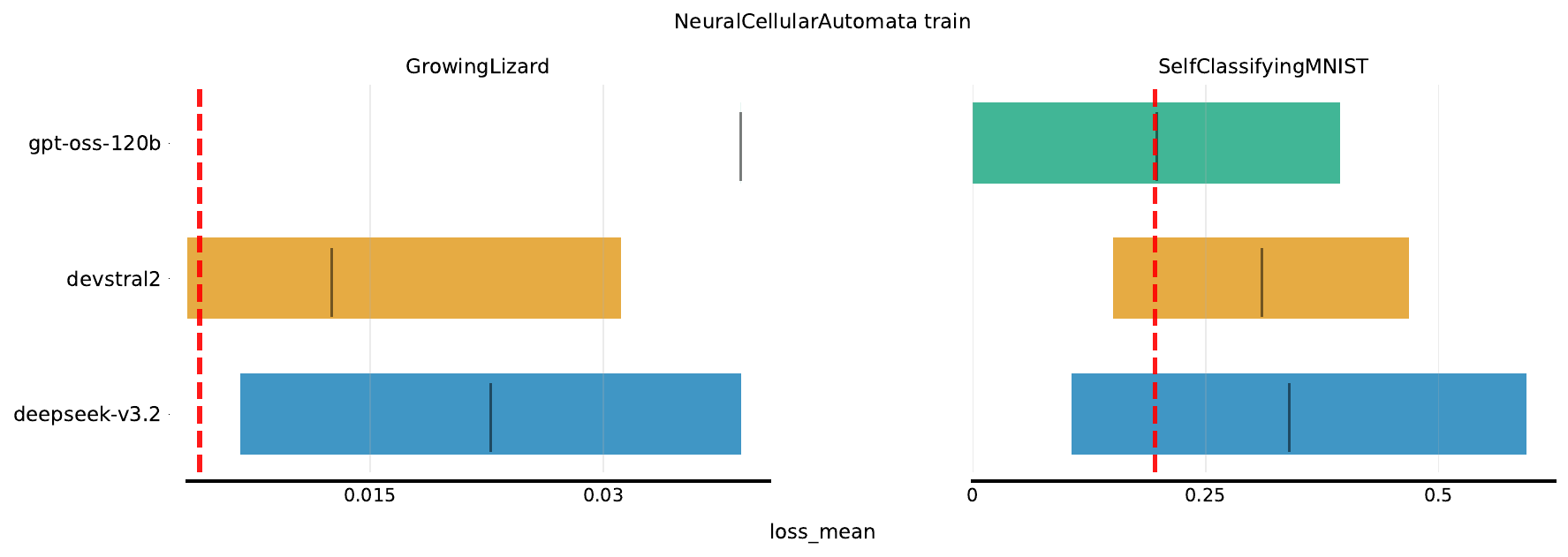}%
\\[0.5em]
\includegraphics[width=0.48\textwidth]{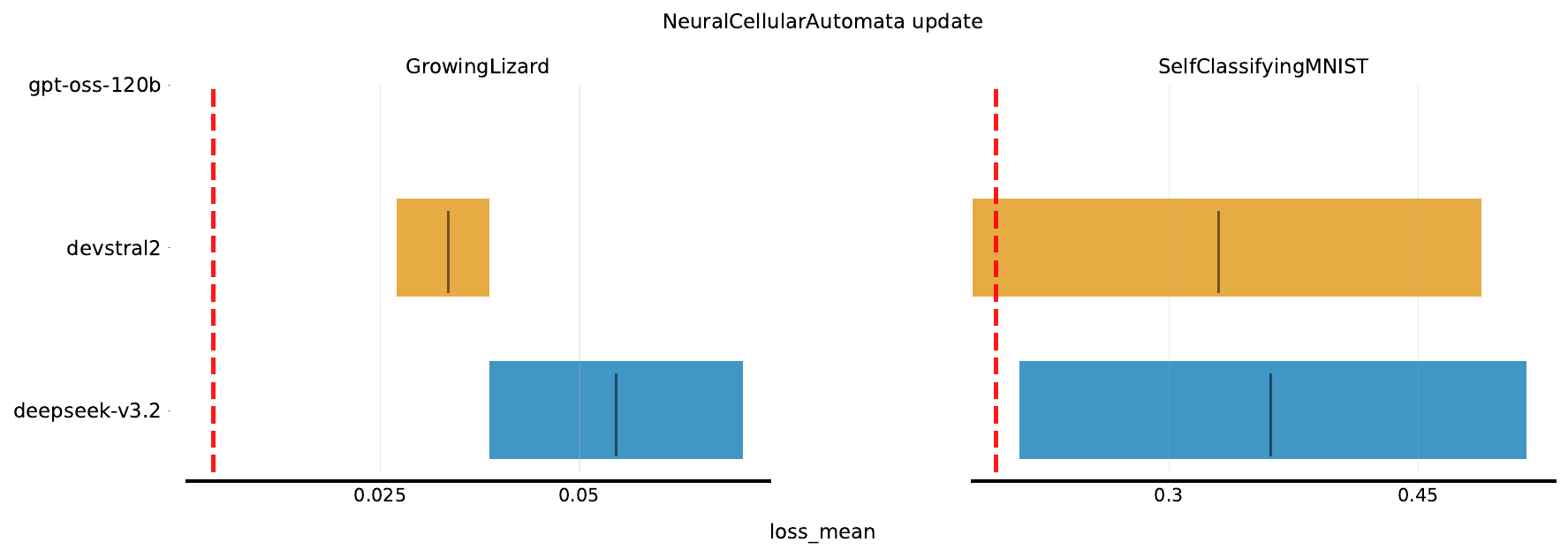}%
\hfill%
\includegraphics[width=0.48\textwidth]{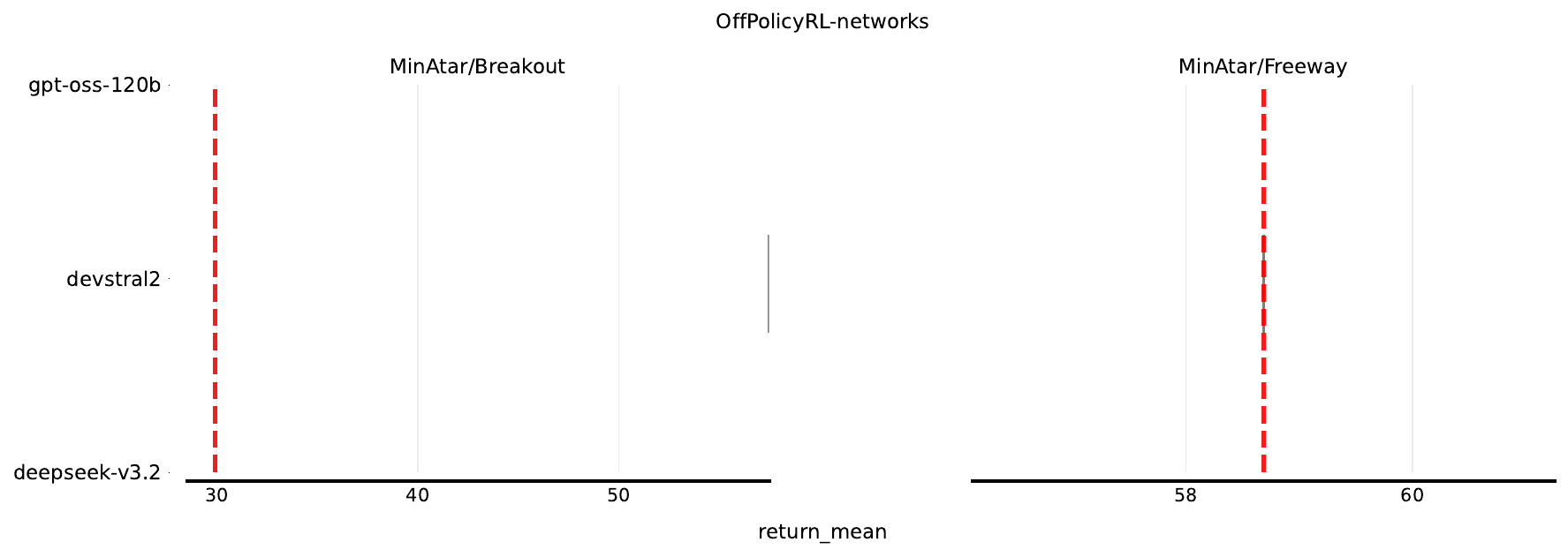}%
\caption{DiscoBench (Single Edit) results on Meta-Train tasks. (Part 3/7)}
\label{fig:one_change_id_3}
\end{figure}
\clearpage

\begin{figure}[htbp]
\centering
\setlength{\lineskip}{0pt}
\includegraphics[width=0.48\textwidth]{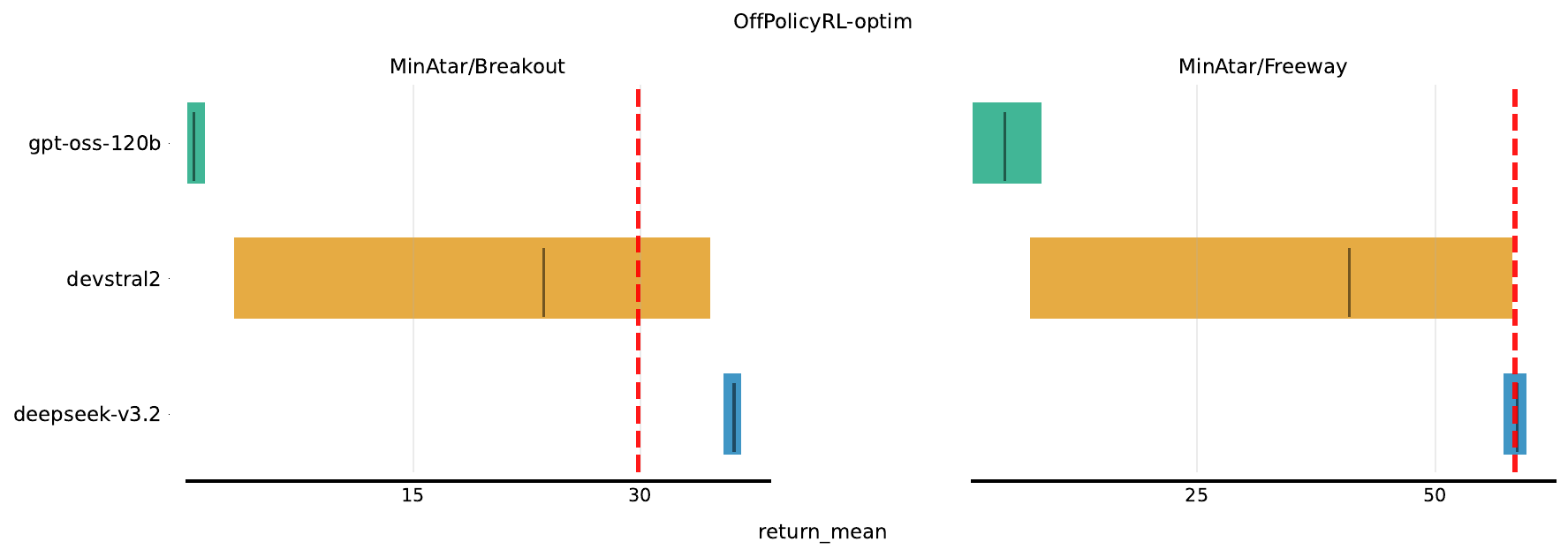}%
\hfill%
\includegraphics[width=0.48\textwidth]{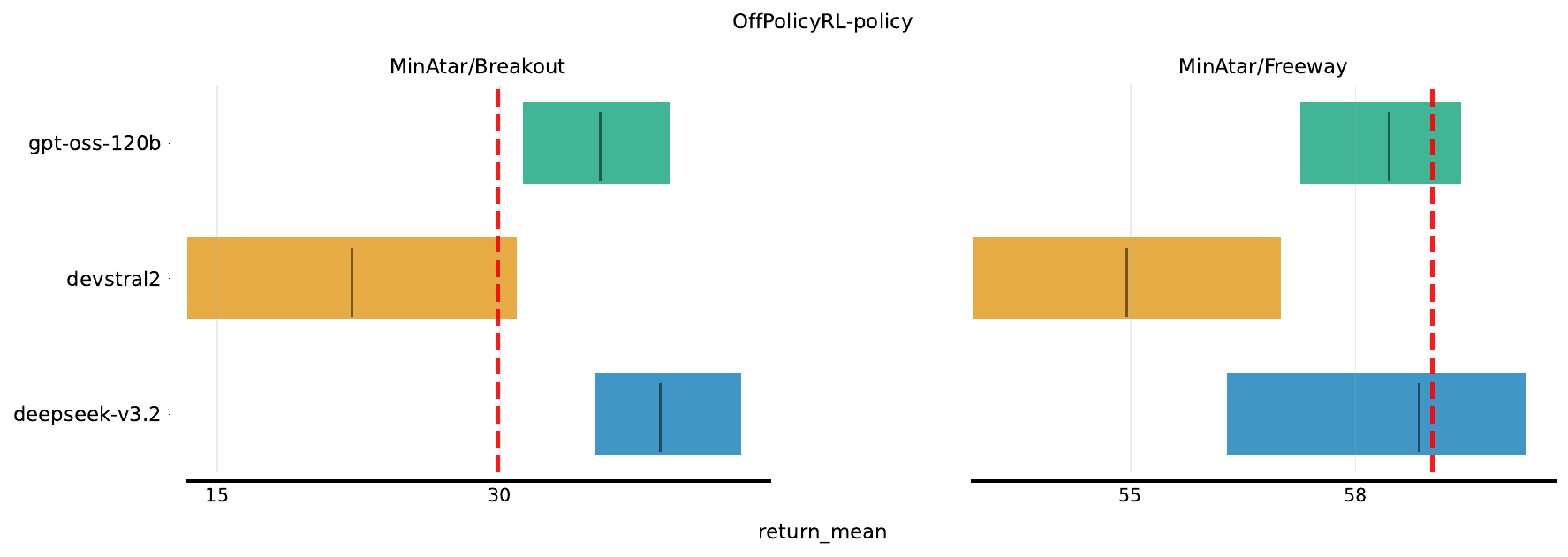}%
\\[0.5em]
\includegraphics[width=0.48\textwidth]{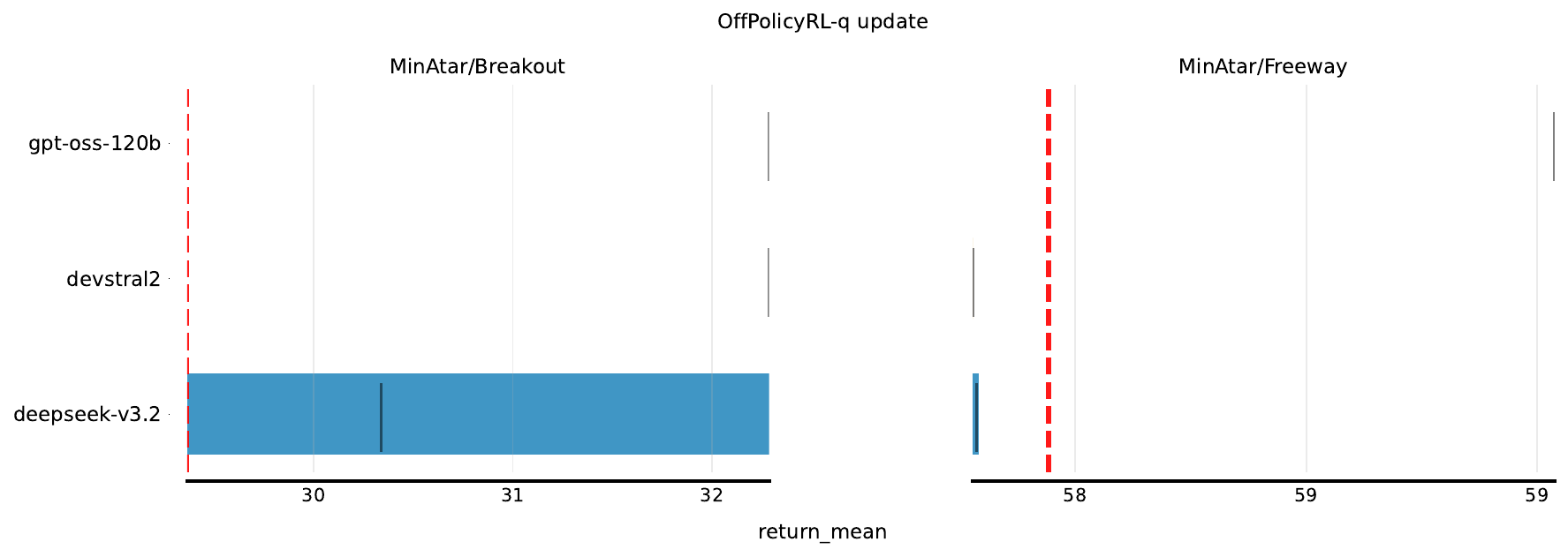}%
\hfill%
\includegraphics[width=0.48\textwidth]{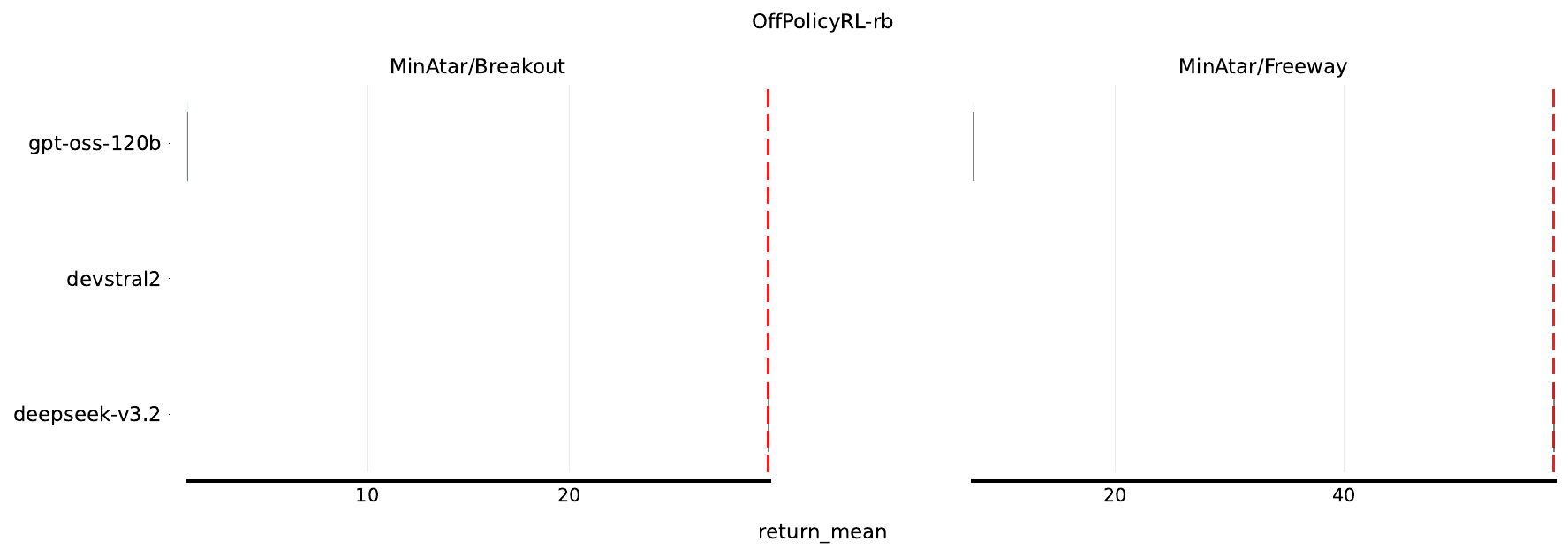}%
\\[0.5em]
\includegraphics[width=0.48\textwidth]{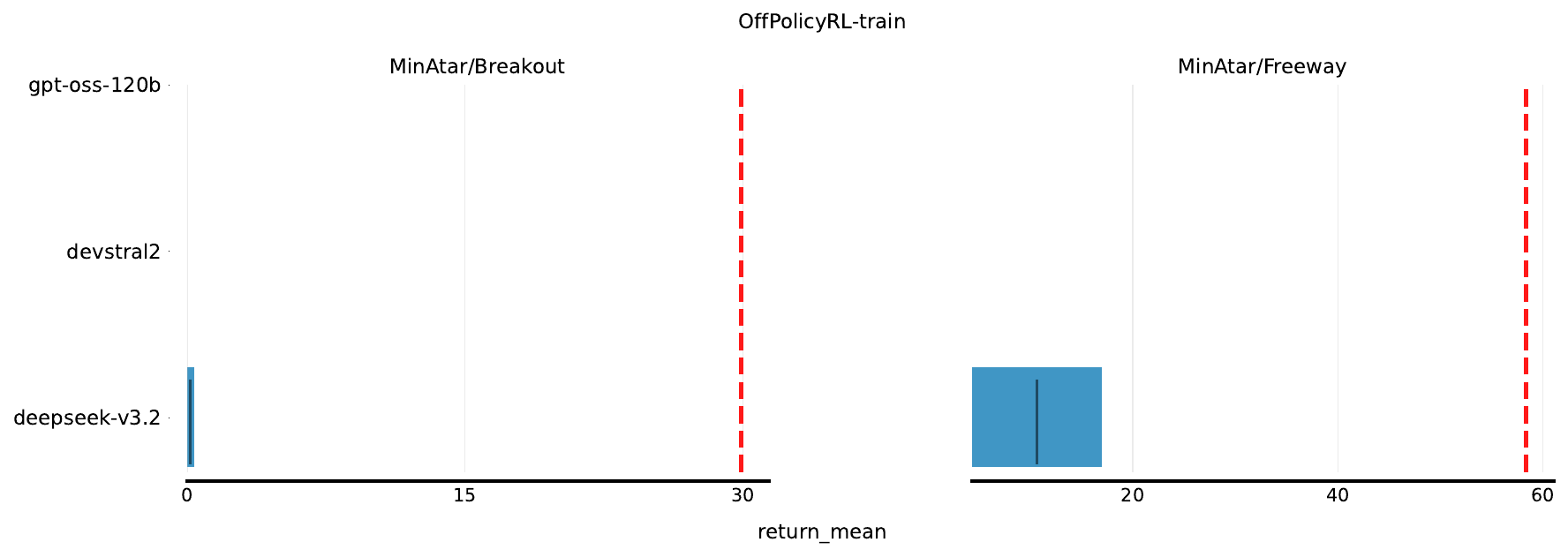}%
\hfill%
\includegraphics[width=0.48\textwidth]{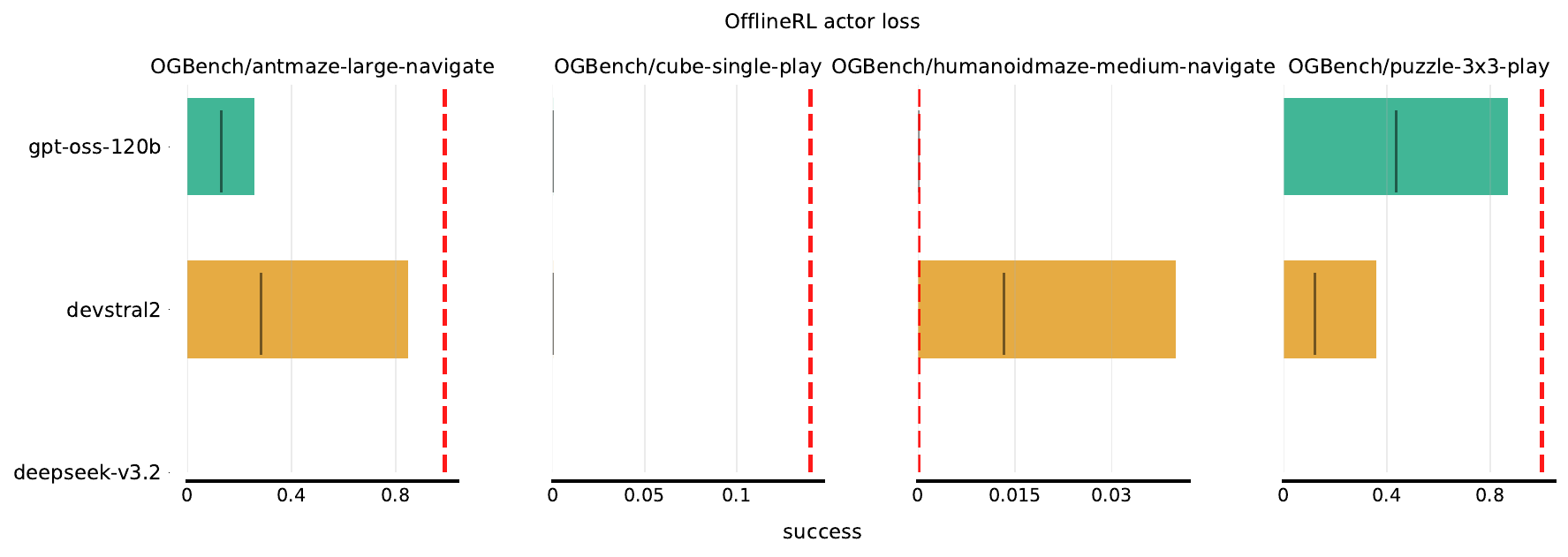}%
\\[0.5em]
\includegraphics[width=0.48\textwidth]{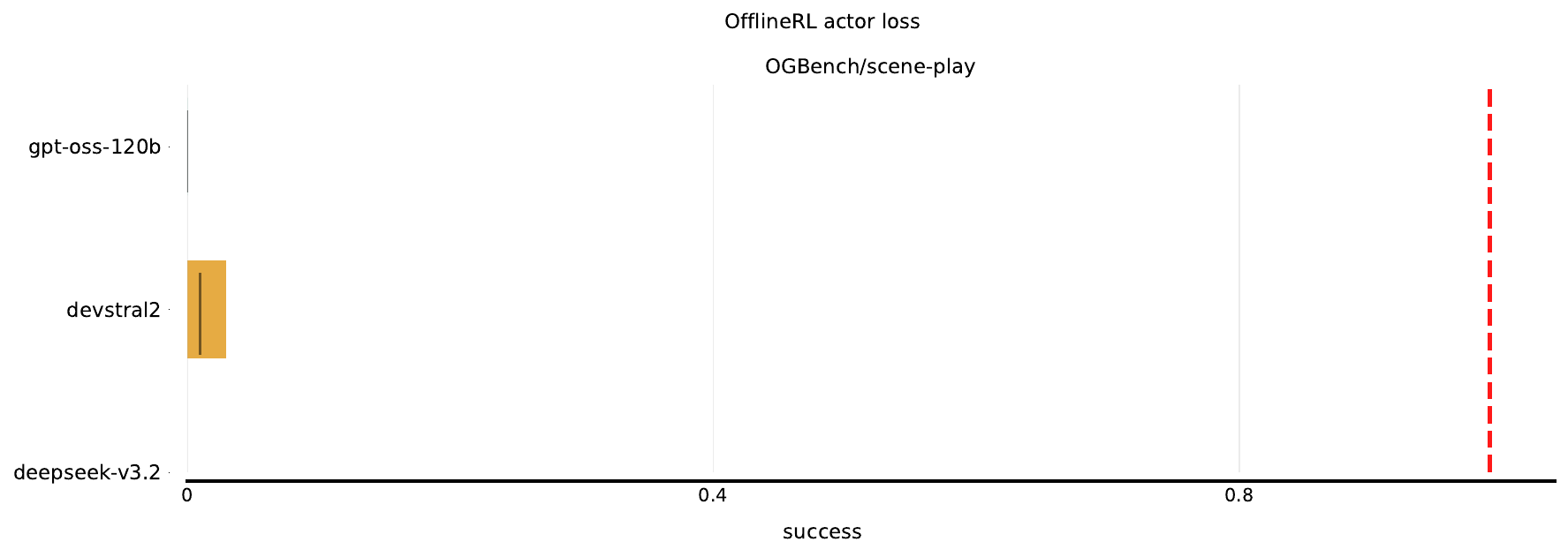}%
\hfill%
\includegraphics[width=0.48\textwidth]{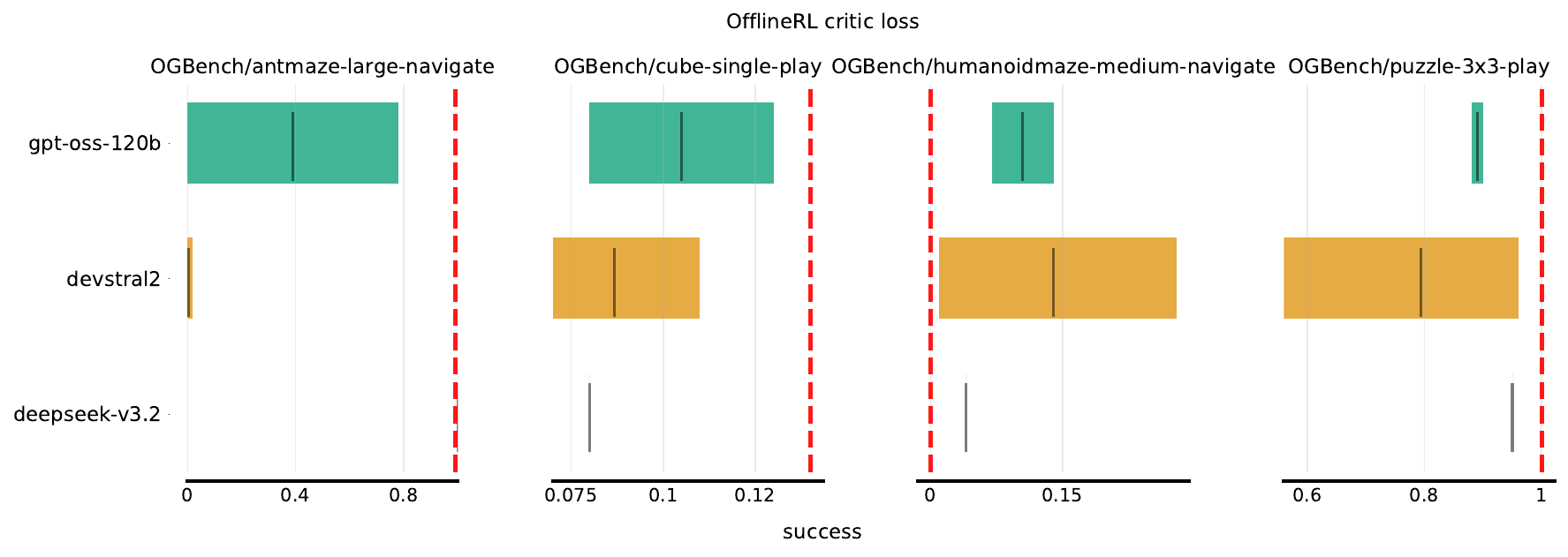}%
\\[0.5em]
\includegraphics[width=0.48\textwidth]{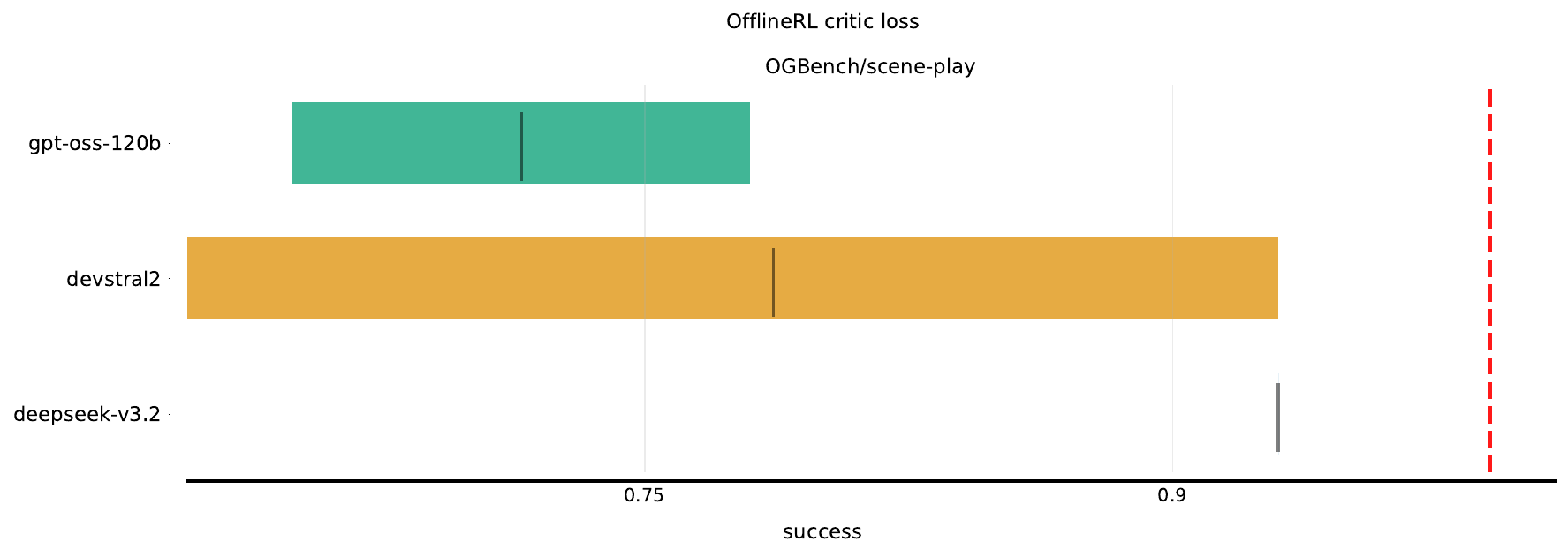}%
\hfill%
\includegraphics[width=0.48\textwidth]{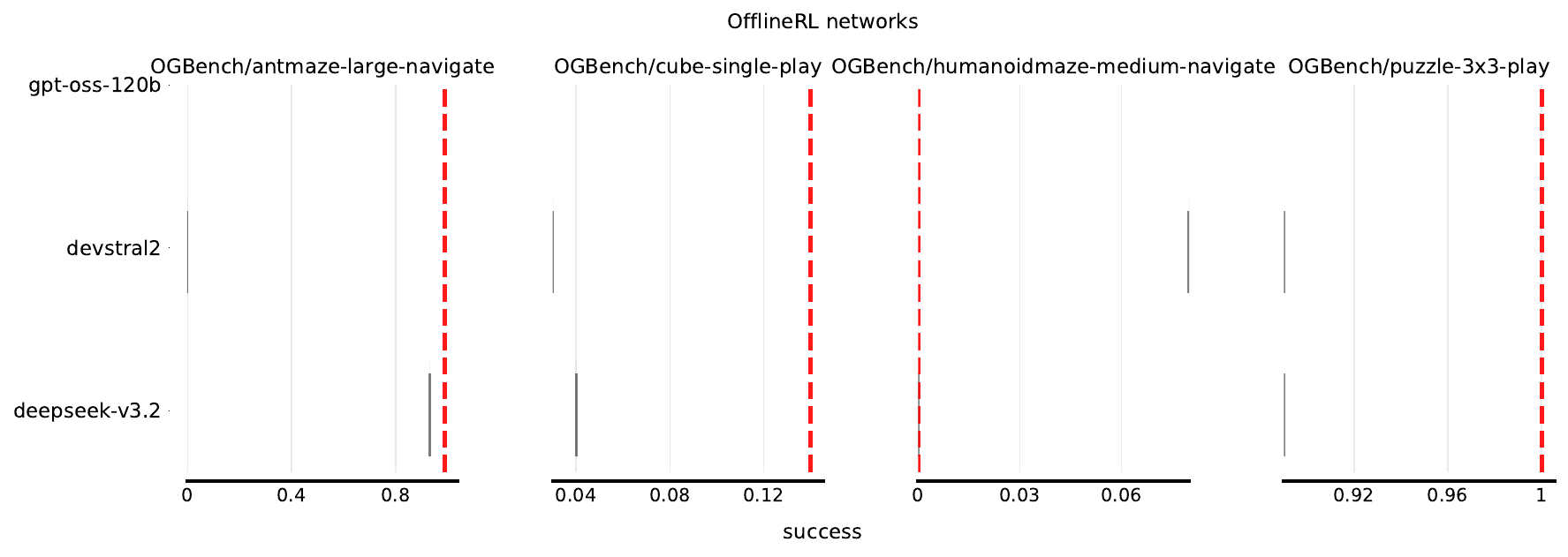}%
\\[0.5em]
\includegraphics[width=0.48\textwidth]{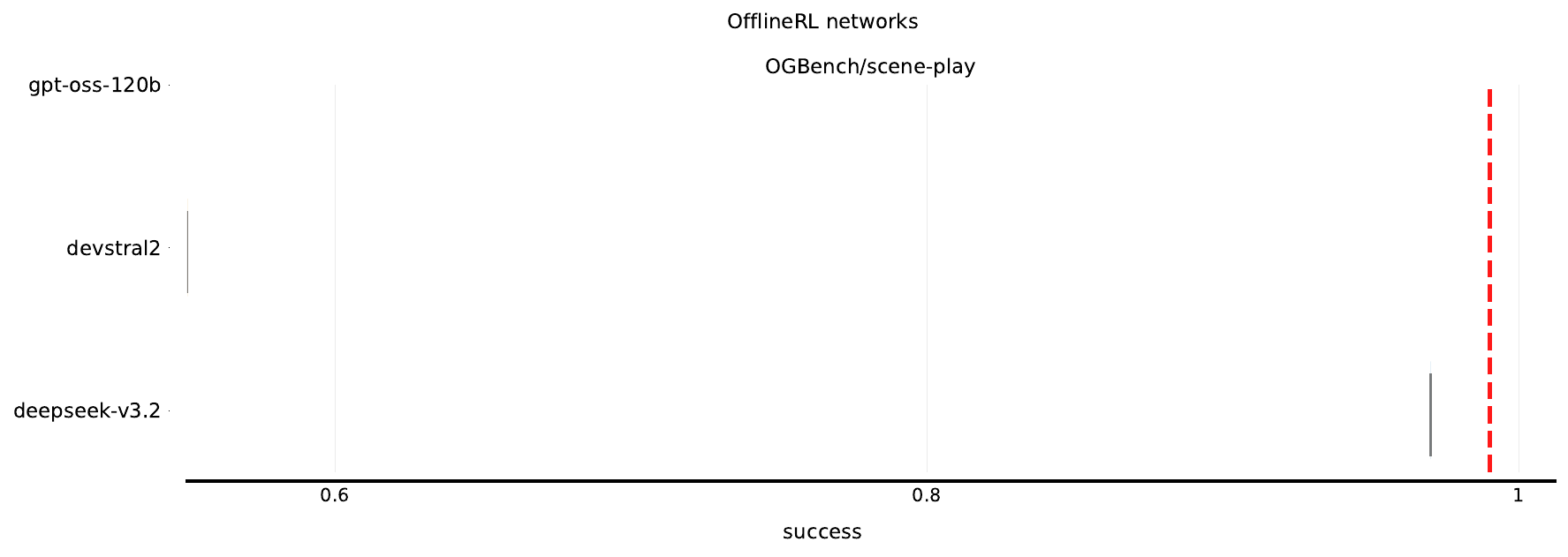}%
\hfill%
\includegraphics[width=0.48\textwidth]{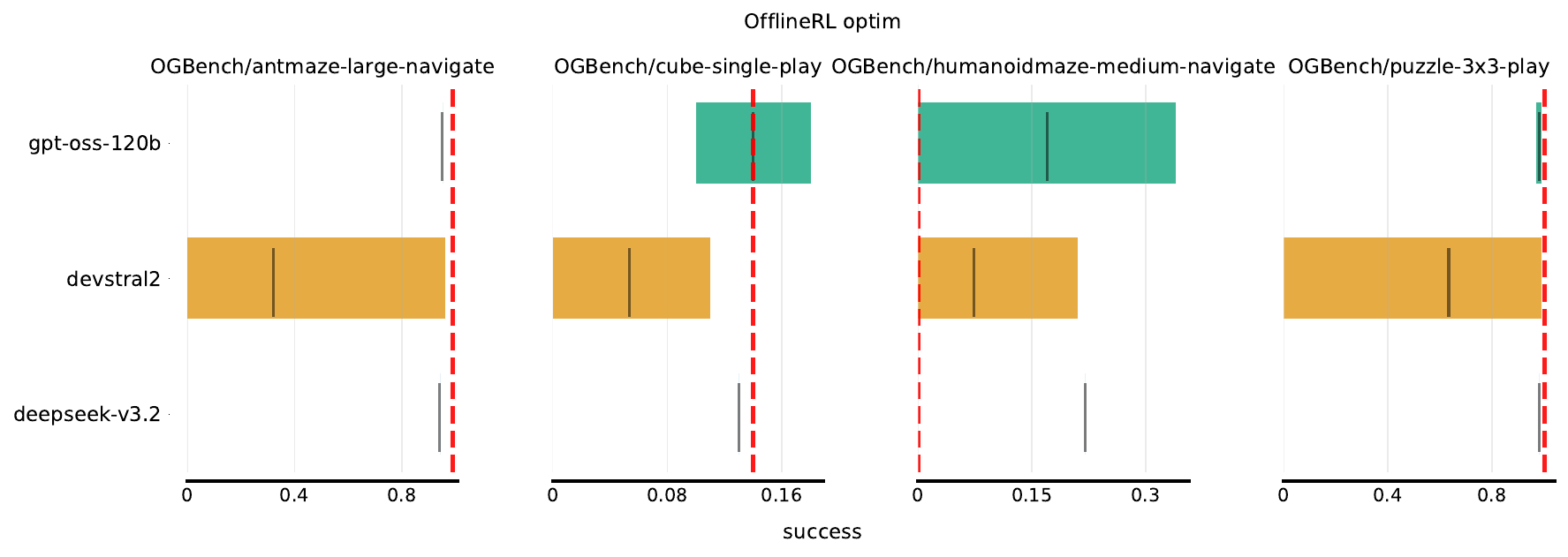}%
\caption{DiscoBench (Single Edit) results on Meta-Train tasks. (Part 4/7)}
\label{fig:one_change_id_4}
\end{figure}
\clearpage

\begin{figure}[htbp]
\centering
\setlength{\lineskip}{0pt}
\includegraphics[width=0.48\textwidth]{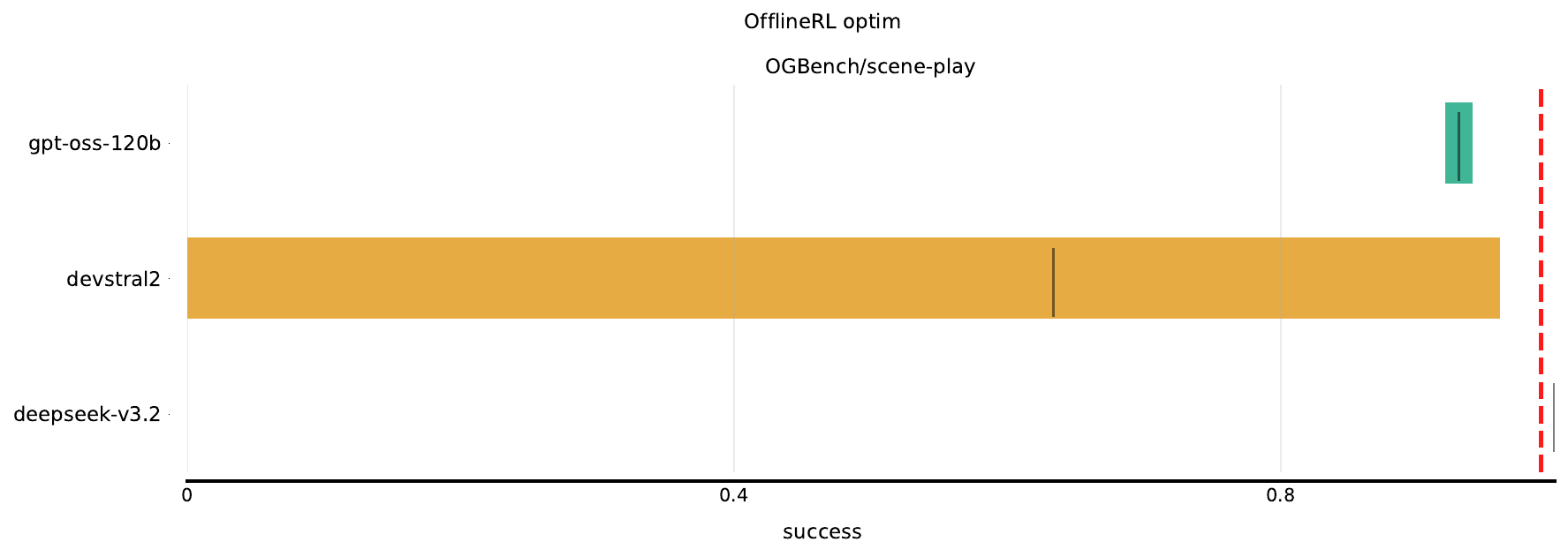}%
\hfill%
\includegraphics[width=0.48\textwidth]{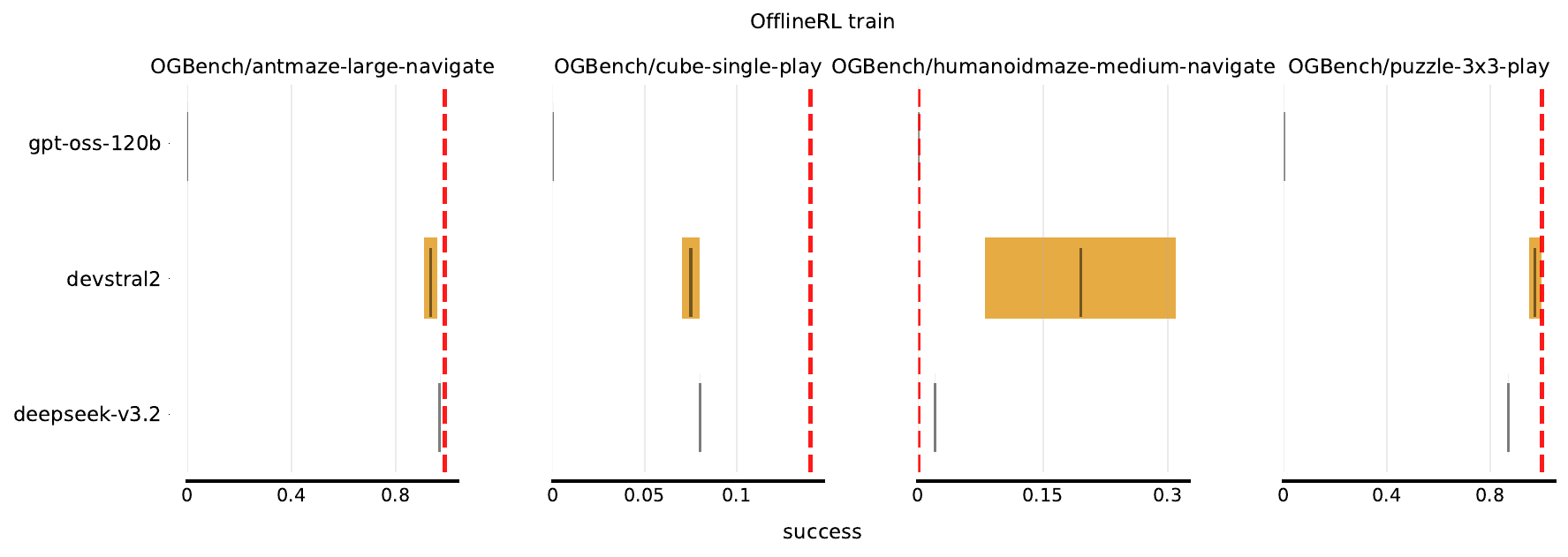}%
\\[0.5em]
\includegraphics[width=0.48\textwidth]{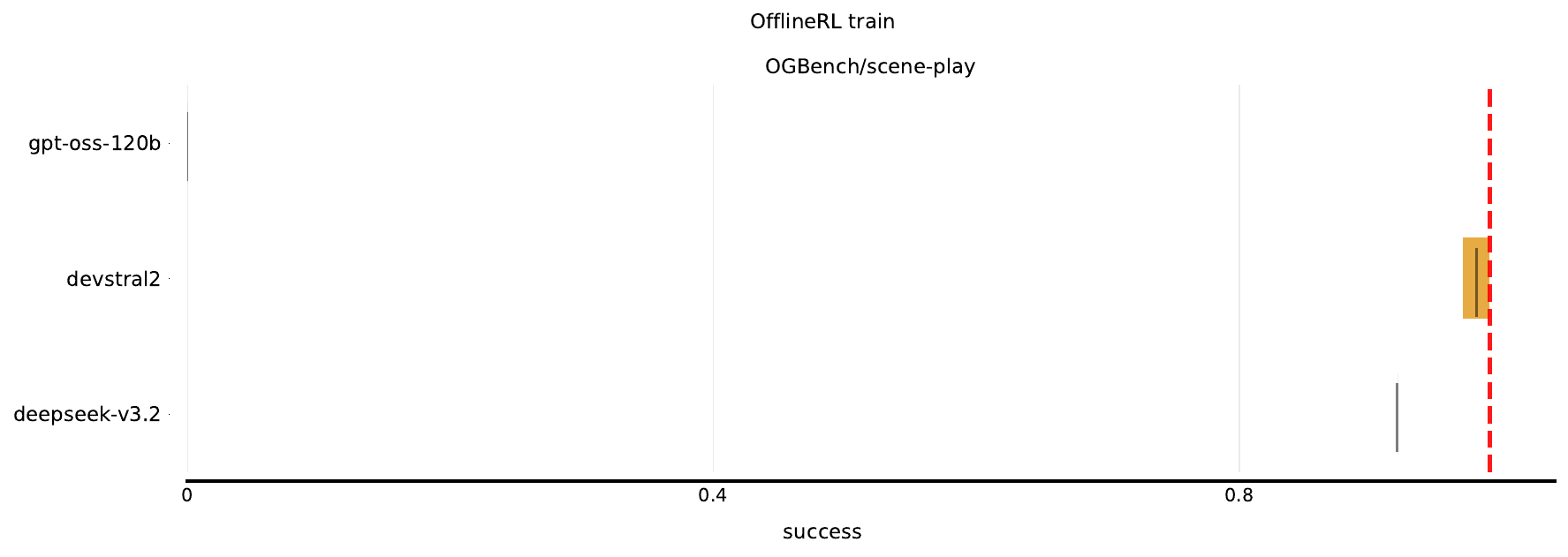}%
\hfill%
\includegraphics[width=0.48\textwidth]{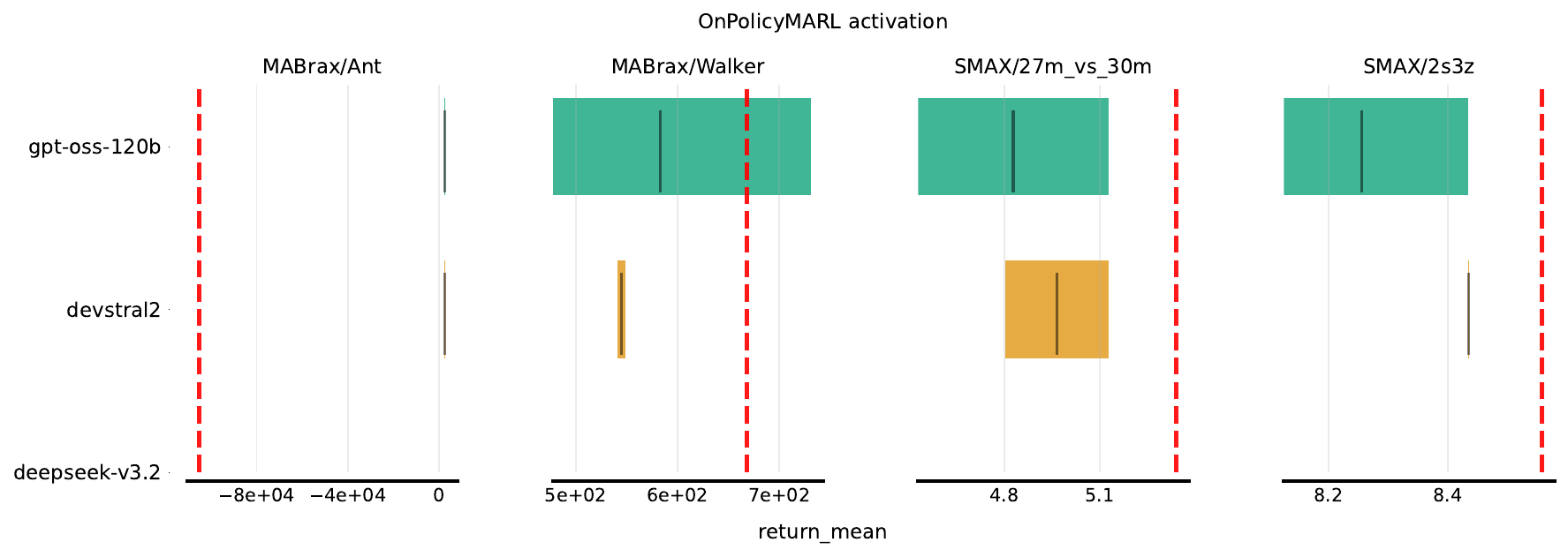}%
\\[0.5em]
\includegraphics[width=0.48\textwidth]{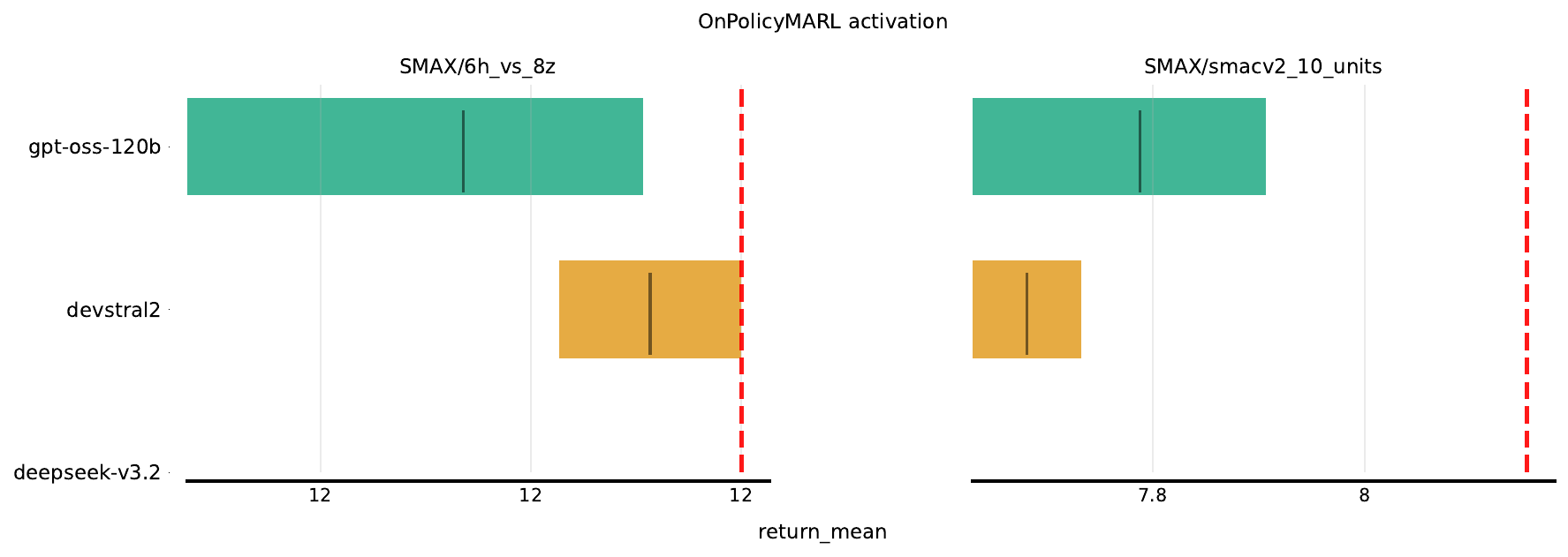}%
\hfill%
\includegraphics[width=0.48\textwidth]{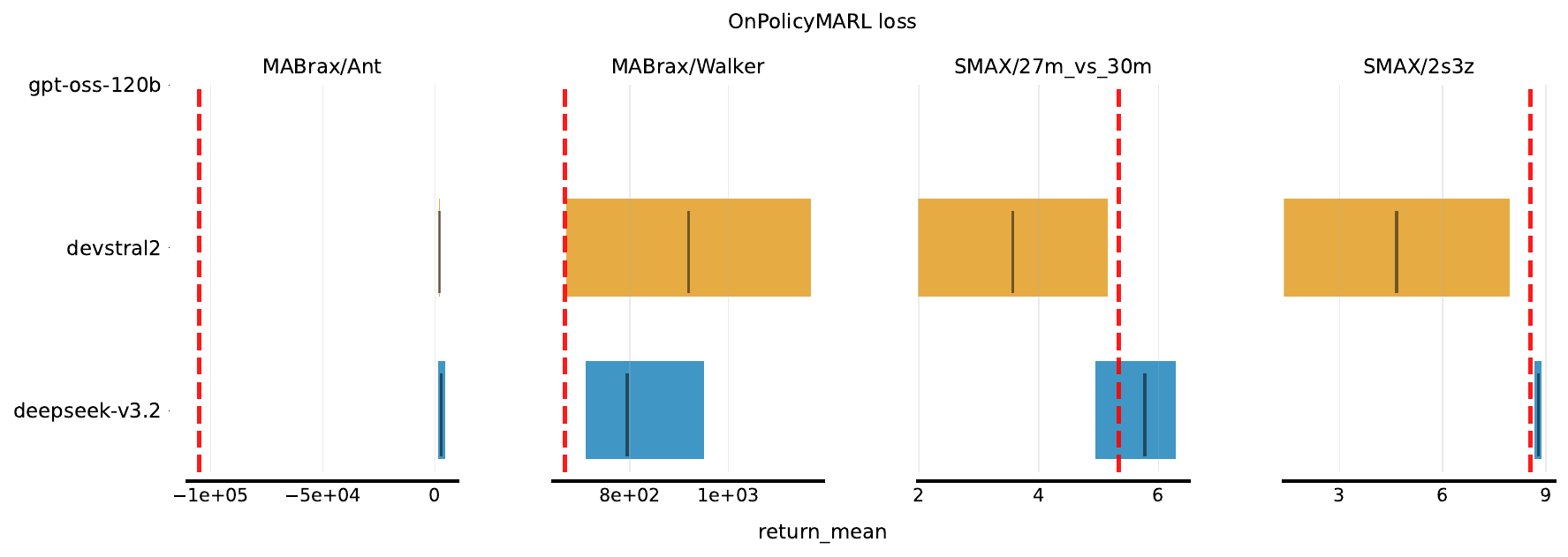}%
\\[0.5em]
\includegraphics[width=0.48\textwidth]{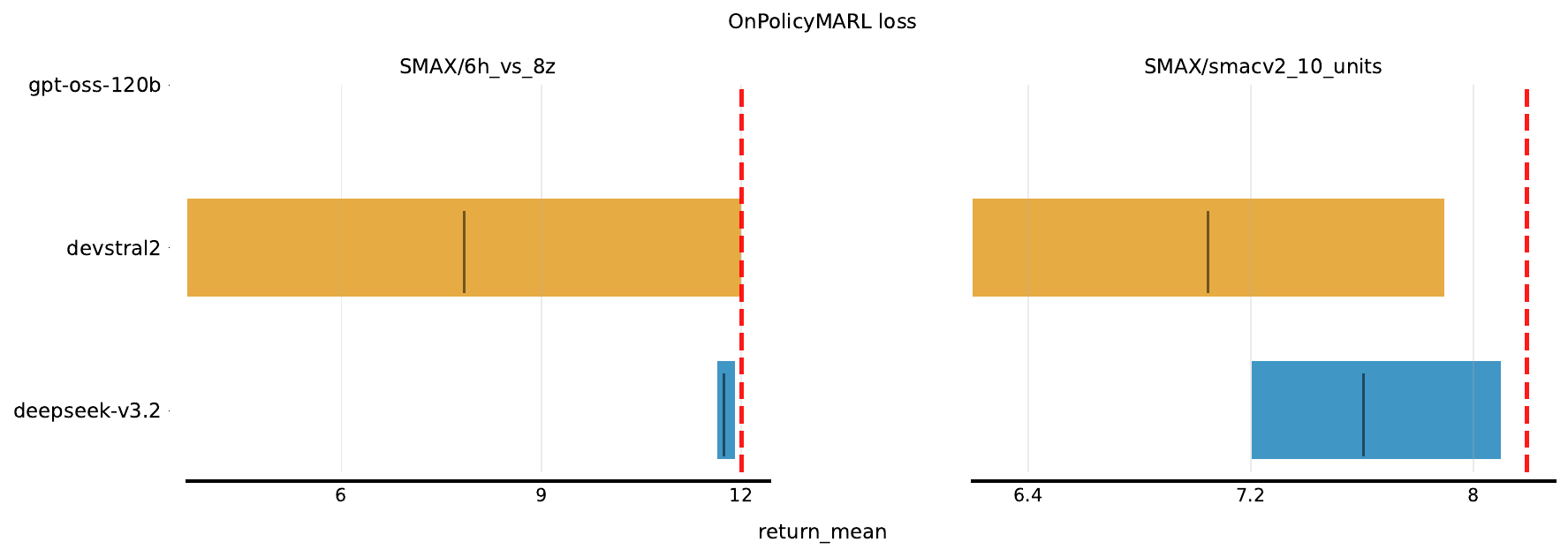}%
\hfill%
\includegraphics[width=0.48\textwidth]{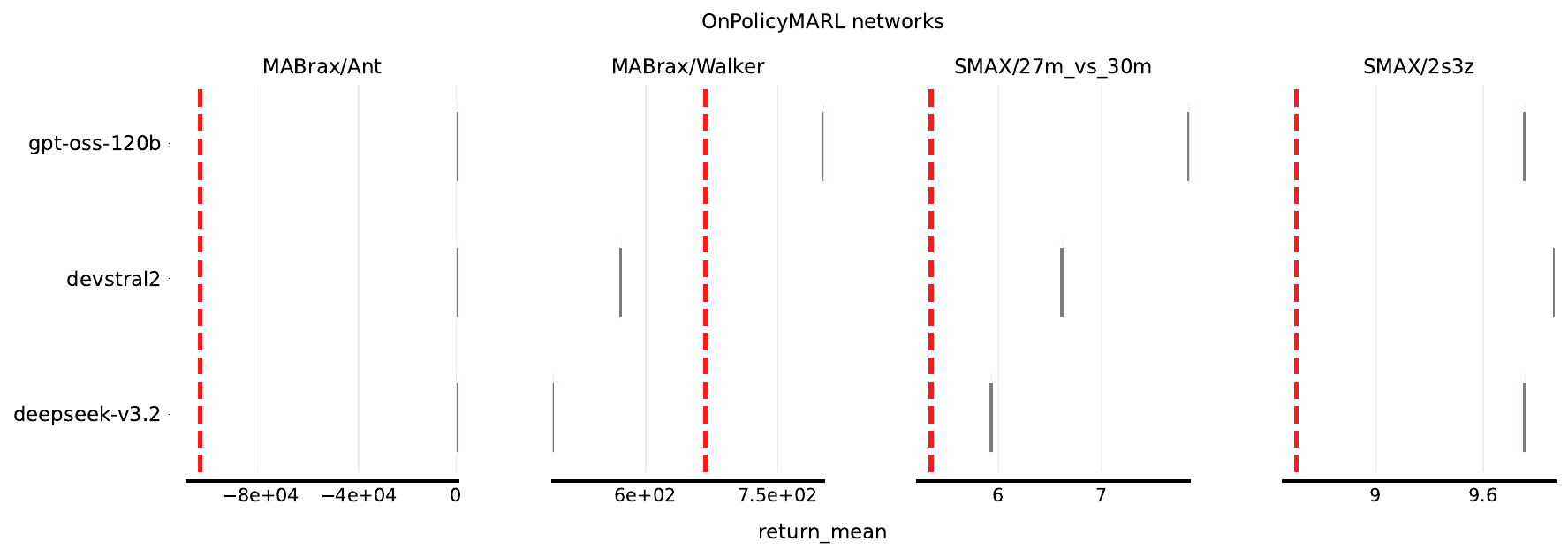}%
\\[0.5em]
\includegraphics[width=0.48\textwidth]{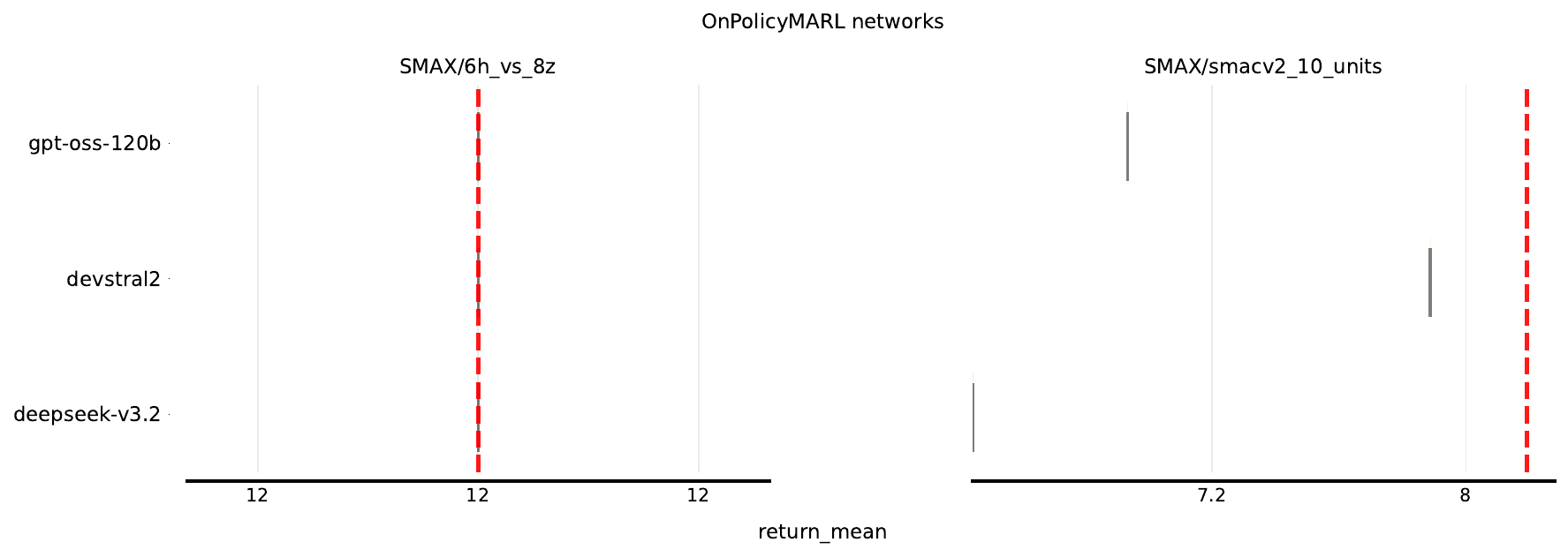}%
\hfill%
\includegraphics[width=0.48\textwidth]{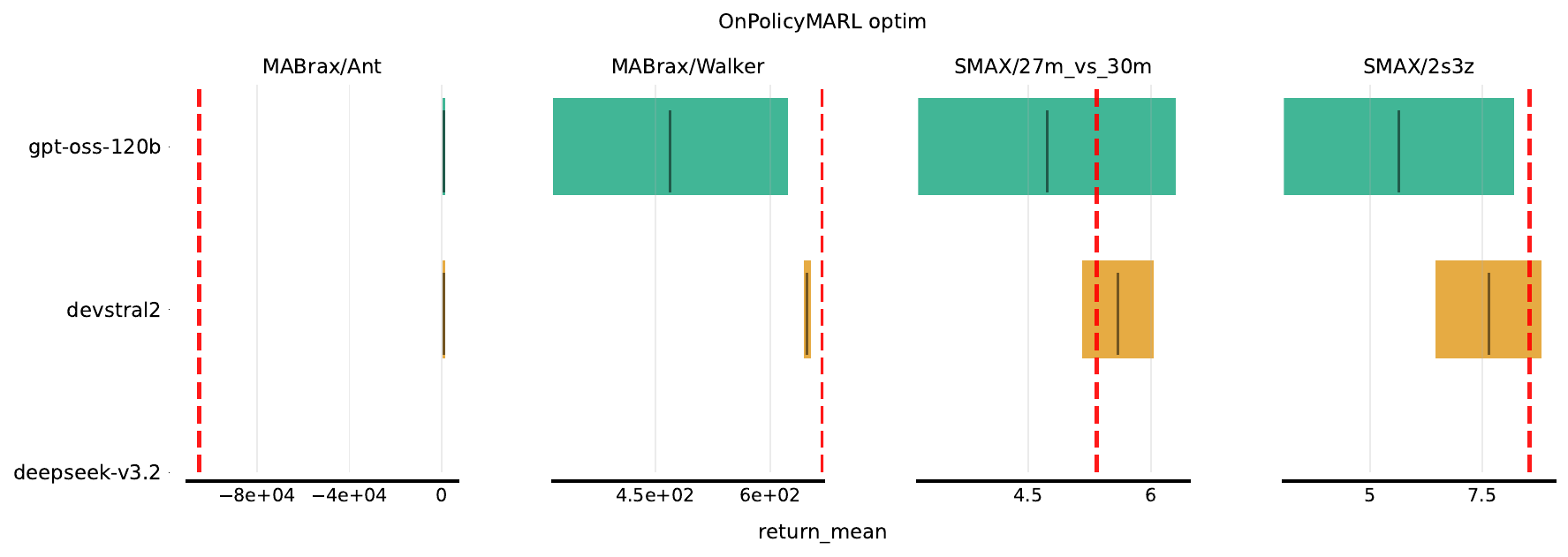}%
\\[0.5em]
\includegraphics[width=0.48\textwidth]{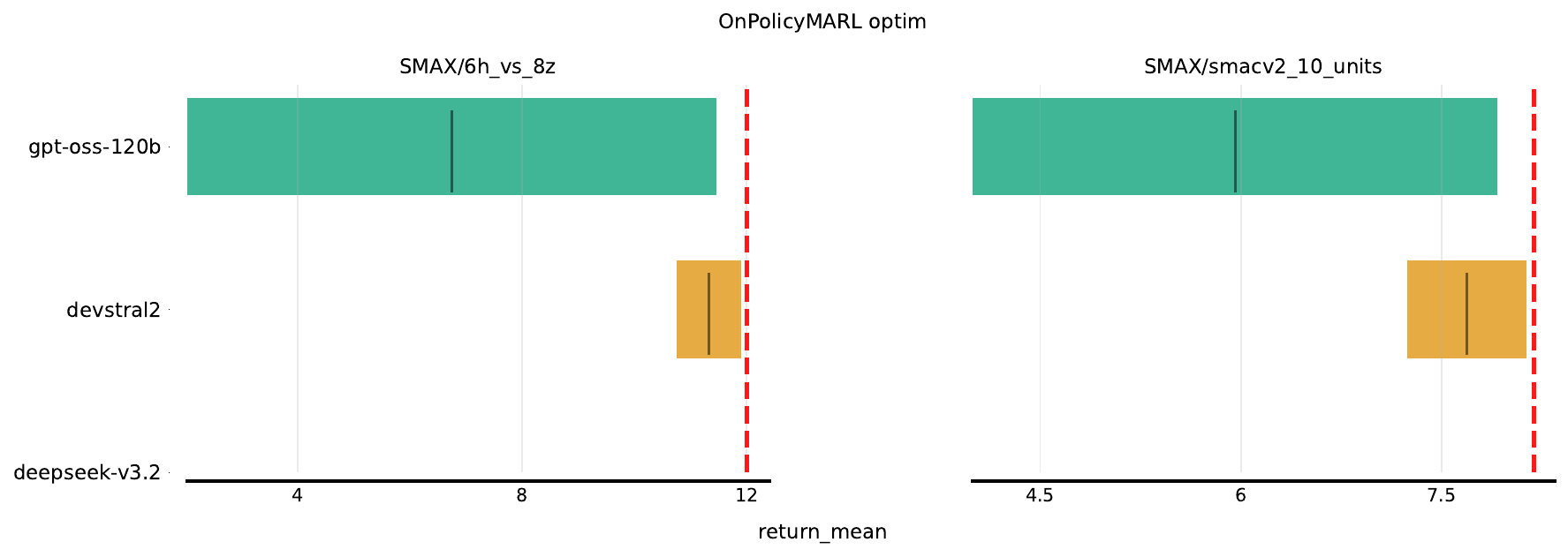}%
\hfill%
\includegraphics[width=0.48\textwidth]{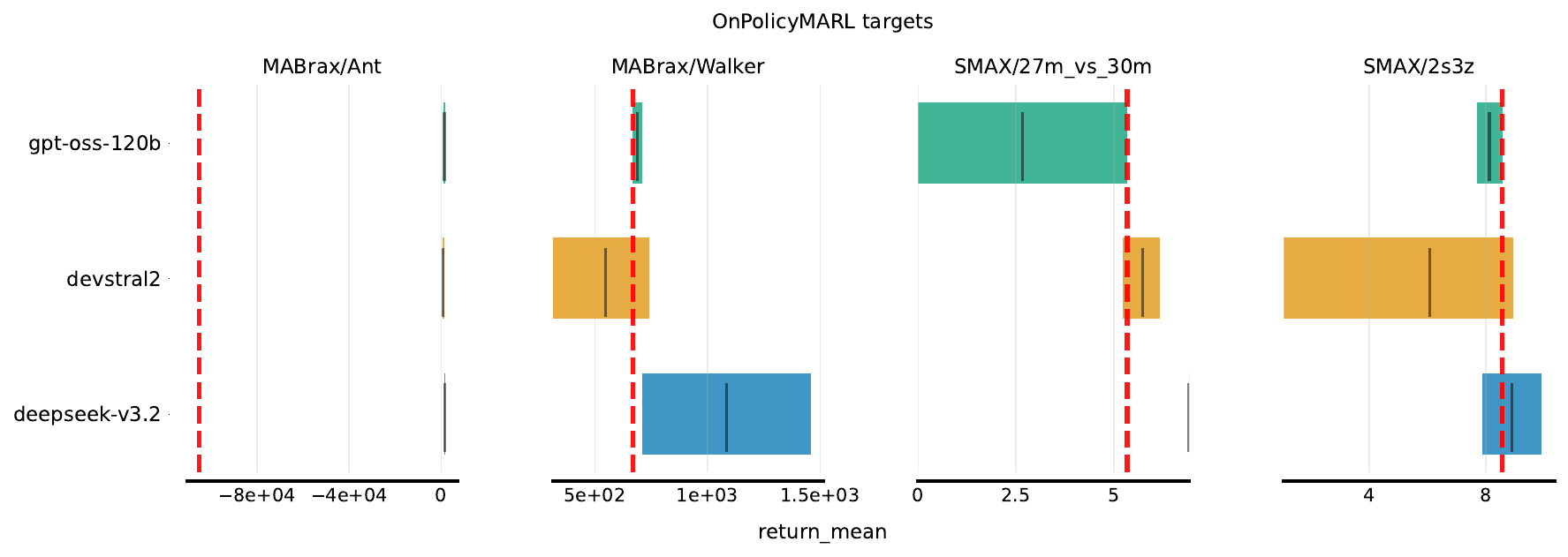}%
\caption{DiscoBench (Single Edit) results on Meta-Train tasks. (Part 5/7)}
\label{fig:one_change_id_5}
\end{figure}
\clearpage

\begin{figure}[htbp]
\centering
\setlength{\lineskip}{0pt}
\includegraphics[width=0.48\textwidth]{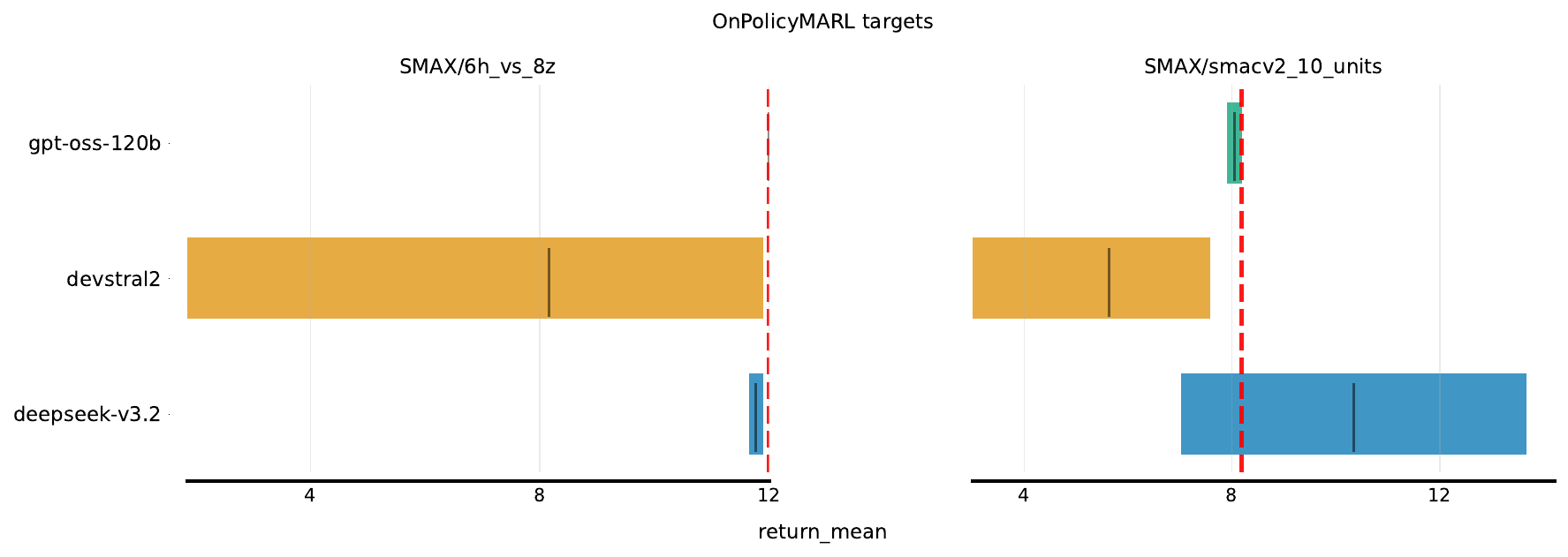}%
\hfill%
\includegraphics[width=0.48\textwidth]{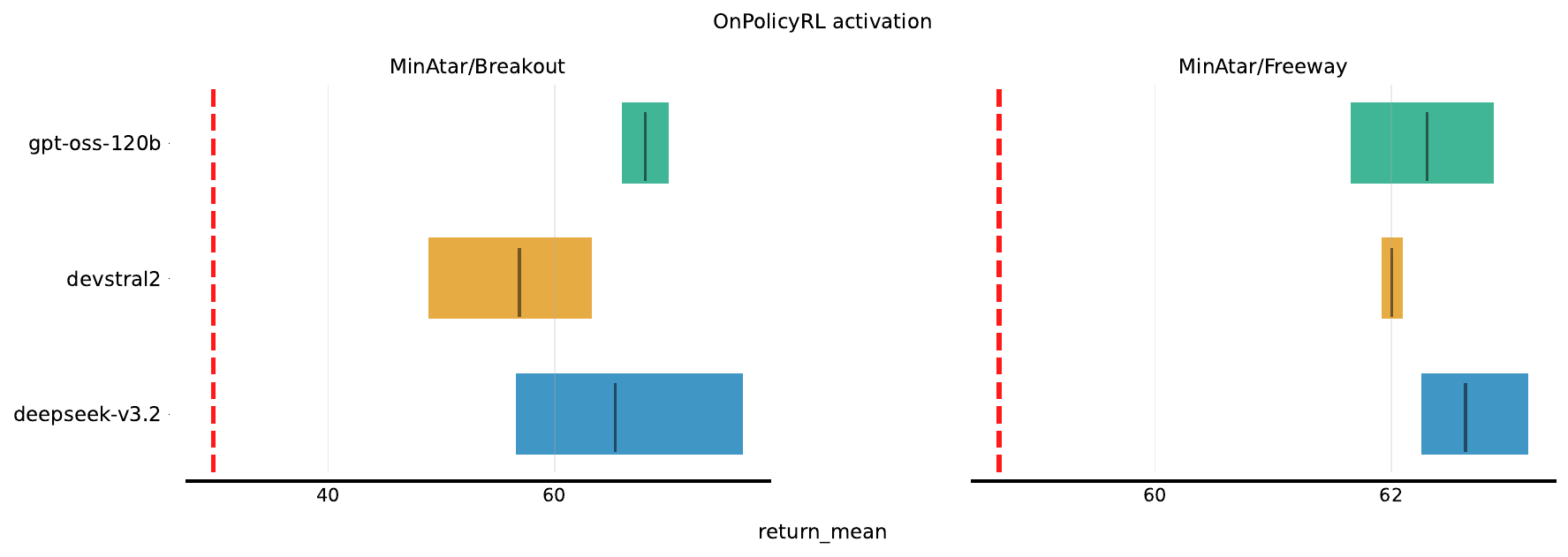}%
\\[0.5em]
\includegraphics[width=0.48\textwidth]{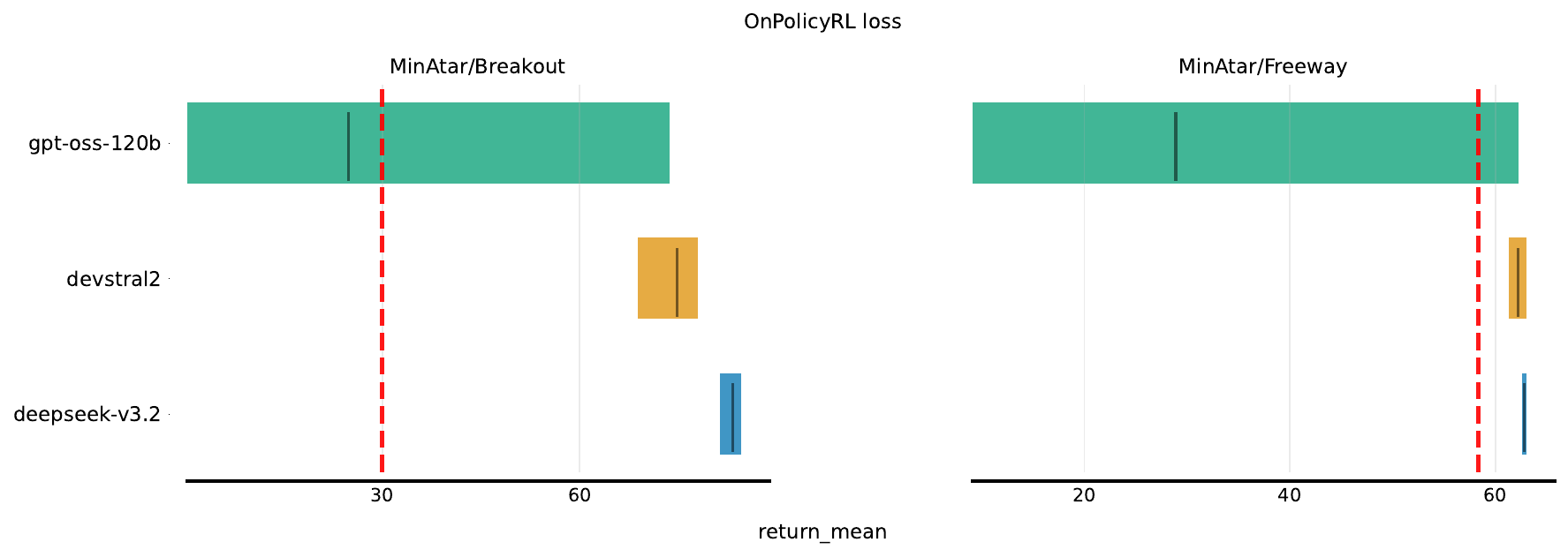}%
\hfill%
\includegraphics[width=0.48\textwidth]{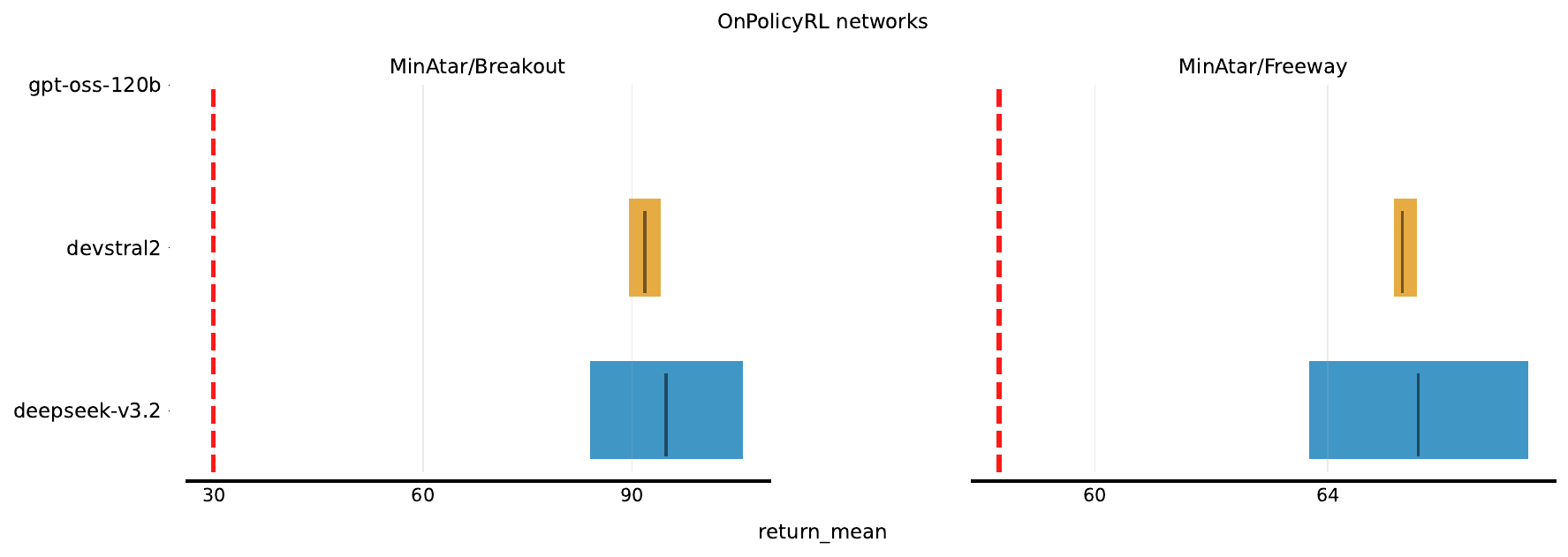}%
\\[0.5em]
\includegraphics[width=0.48\textwidth]{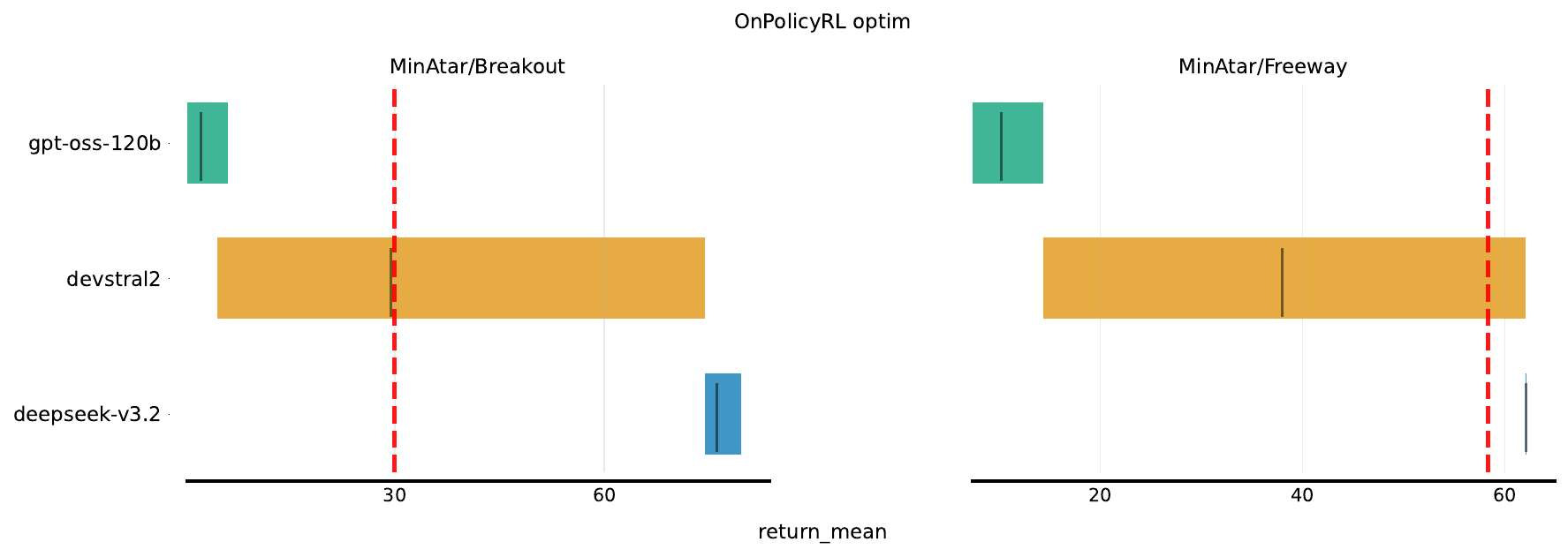}%
\hfill%
\includegraphics[width=0.48\textwidth]{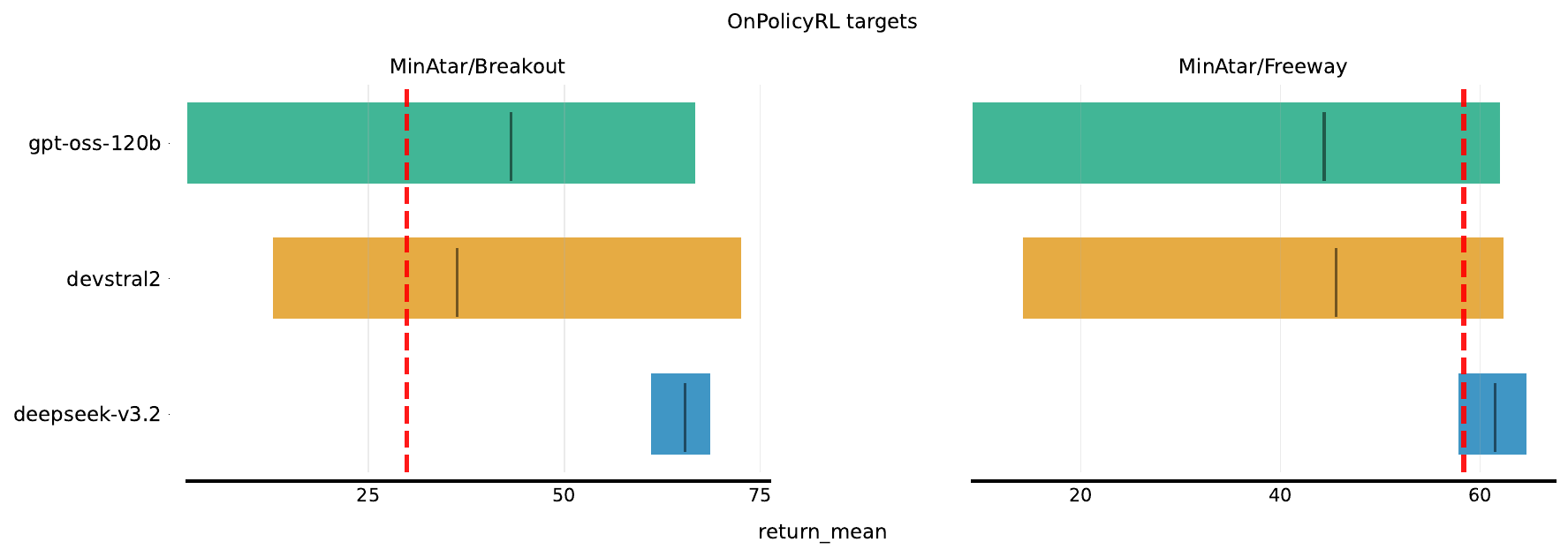}%
\\[0.5em]
\includegraphics[width=0.48\textwidth]{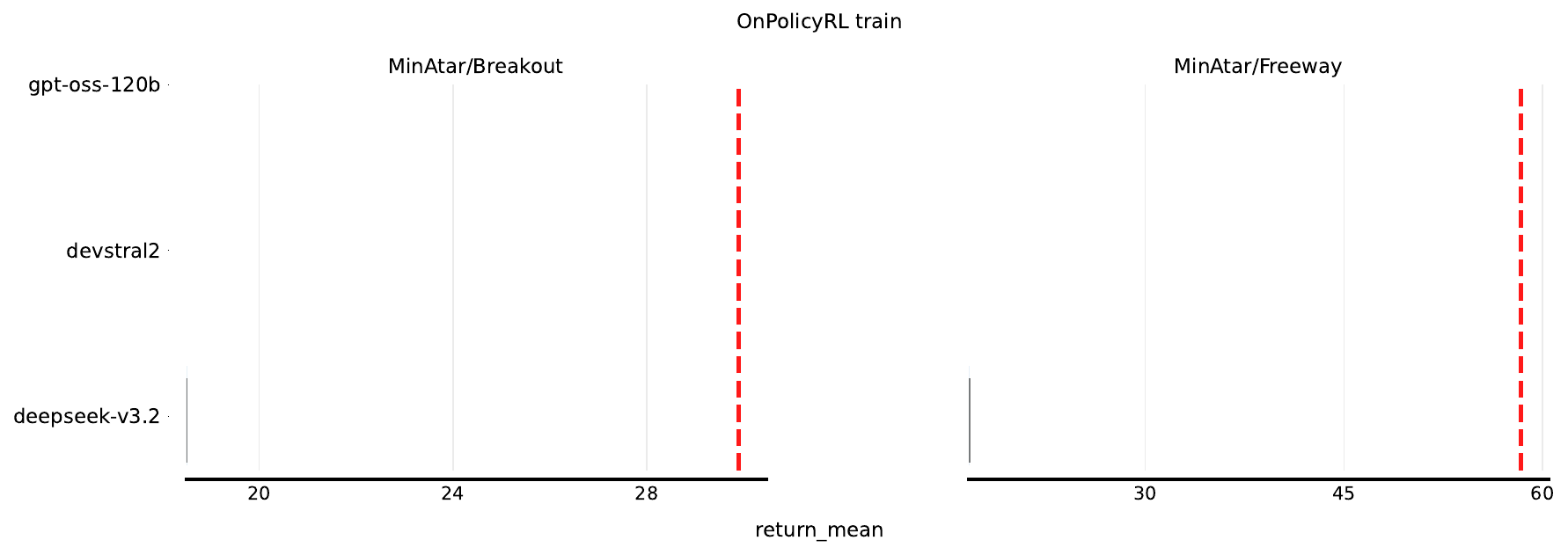}%
\hfill%
\includegraphics[width=0.48\textwidth]{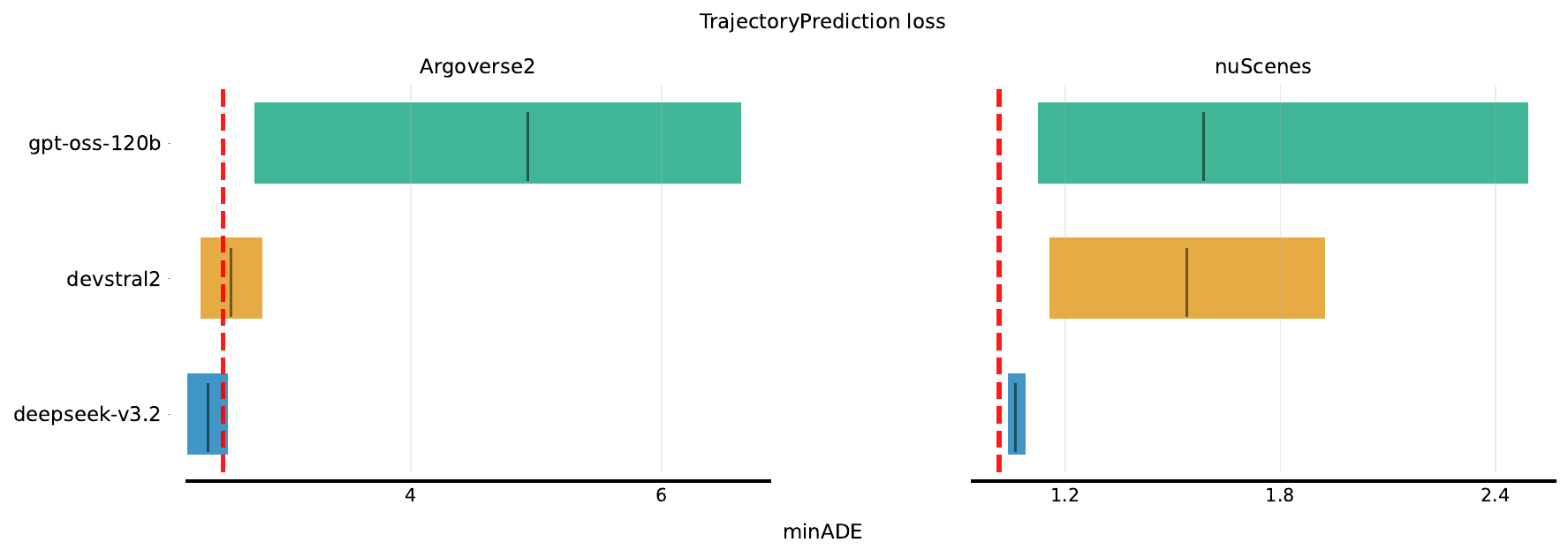}%
\\[0.5em]
\includegraphics[width=0.48\textwidth]{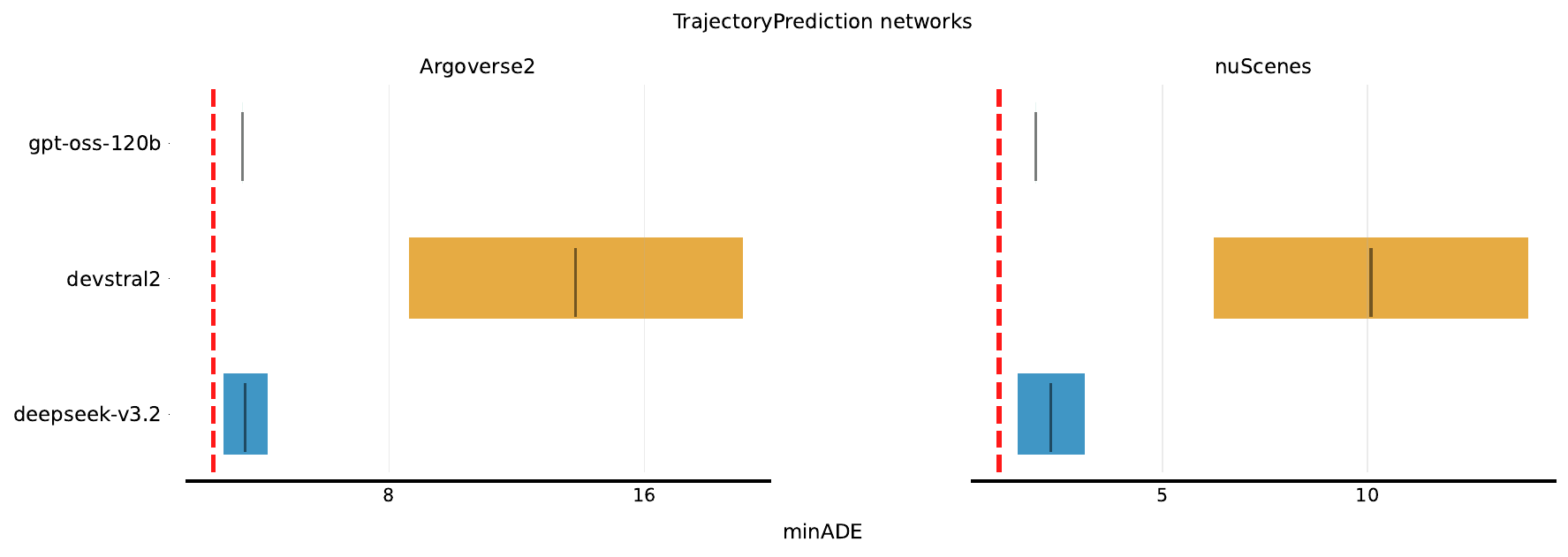}%
\hfill%
\includegraphics[width=0.48\textwidth]{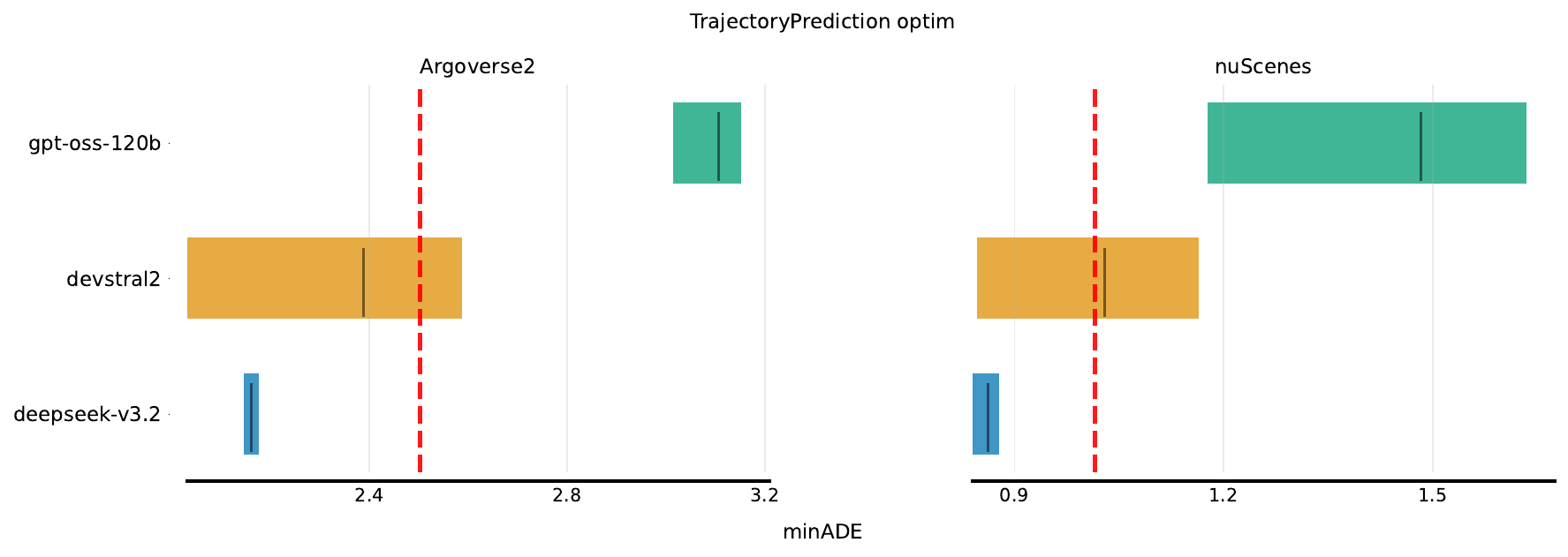}%
\\[0.5em]
\includegraphics[width=0.48\textwidth]{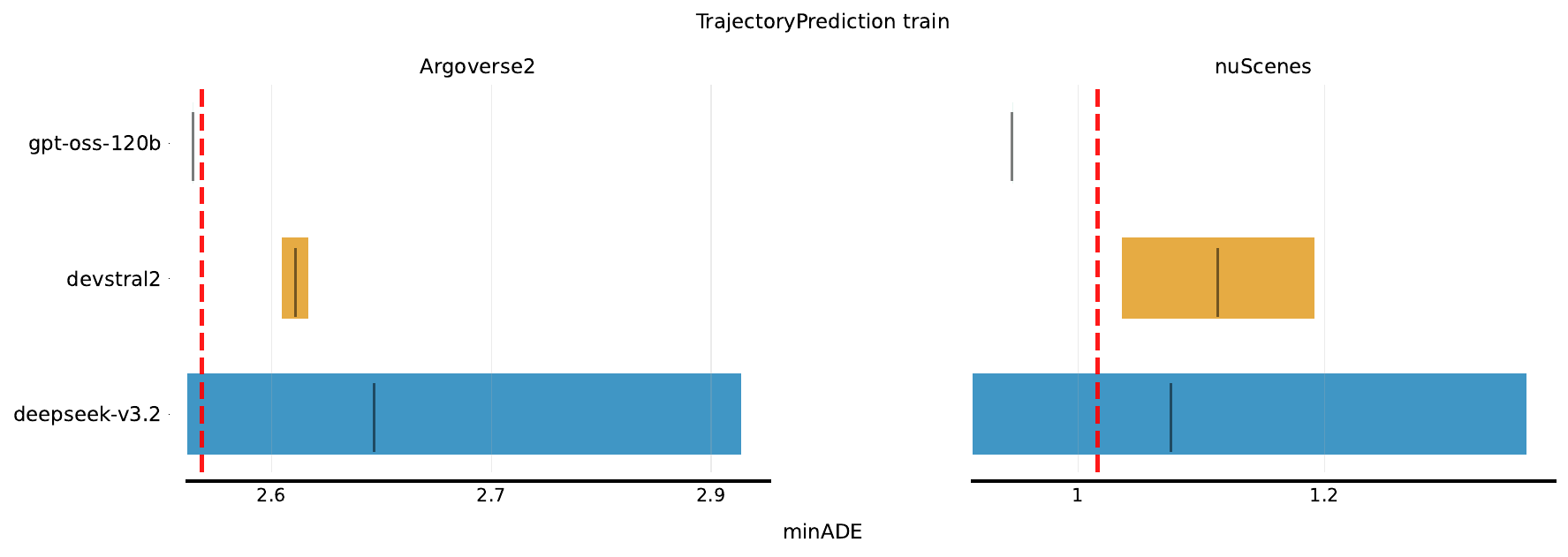}%
\hfill%
\includegraphics[width=0.48\textwidth]{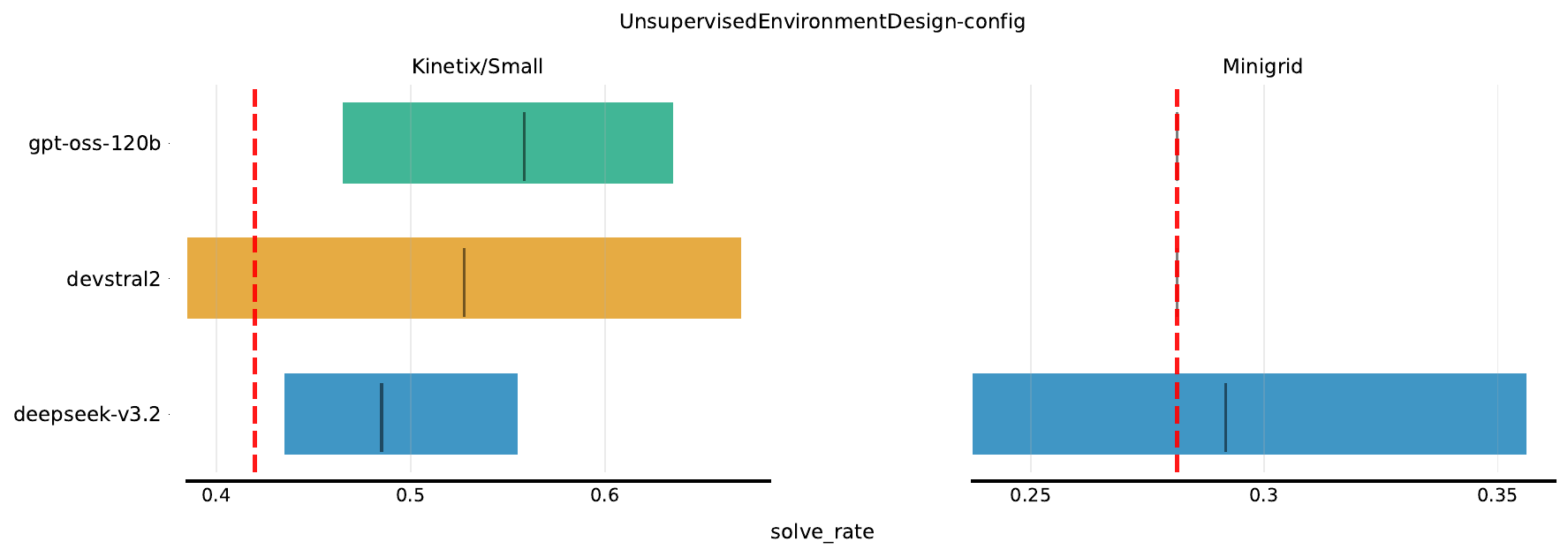}%
\caption{DiscoBench (Single Edit) results on Meta-Train tasks. (Part 6/7)}
\label{fig:one_change_id_6}
\end{figure}
\clearpage

\begin{figure}[htbp]
\centering
\setlength{\lineskip}{0pt}
\includegraphics[width=0.48\textwidth]{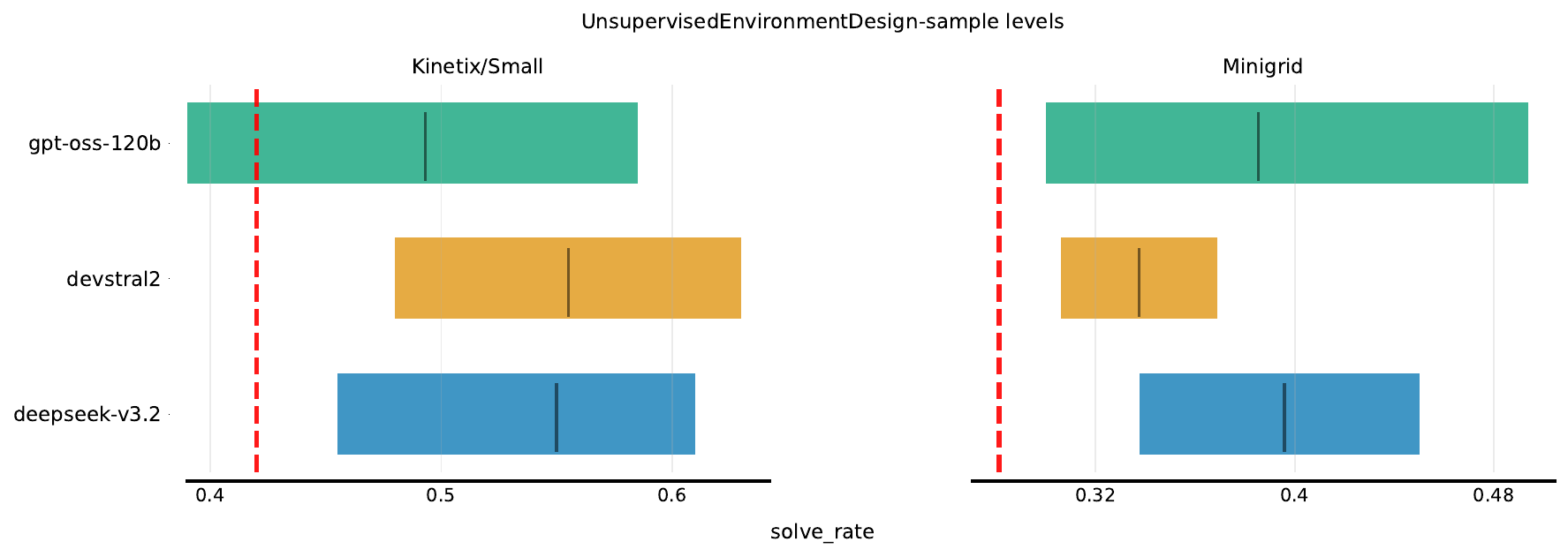}%
\hfill%
\includegraphics[width=0.48\textwidth]{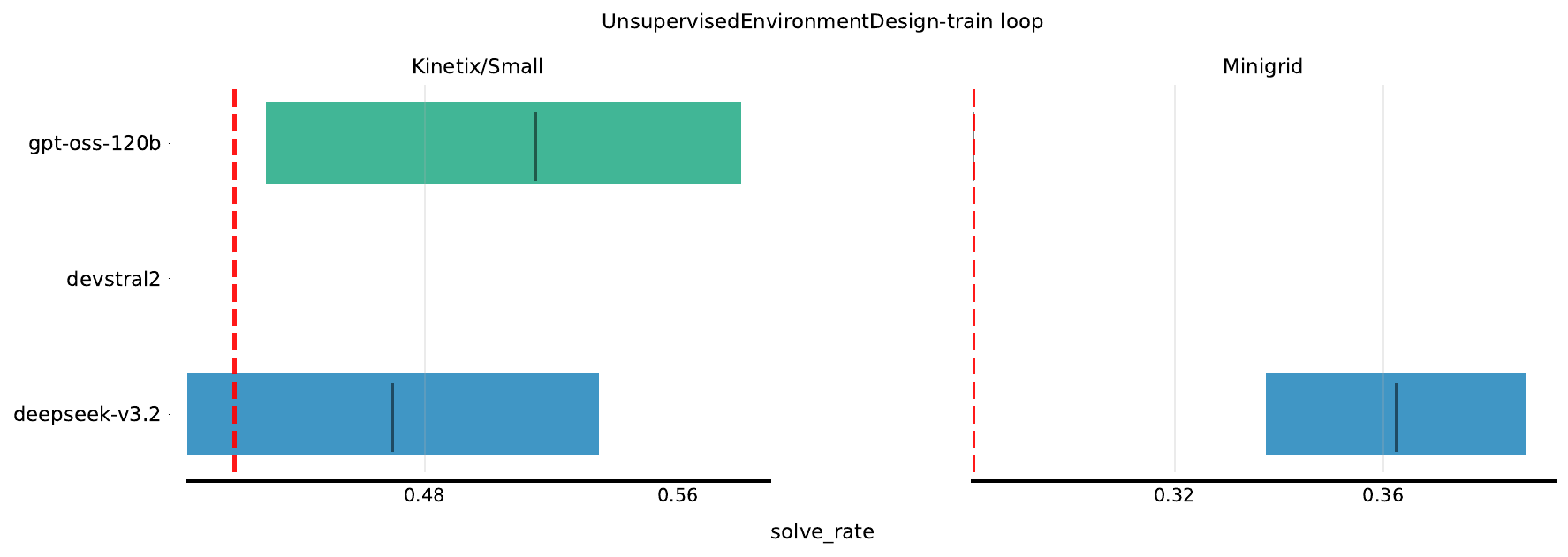}%
\\[0.5em]
\includegraphics[width=0.48\textwidth]{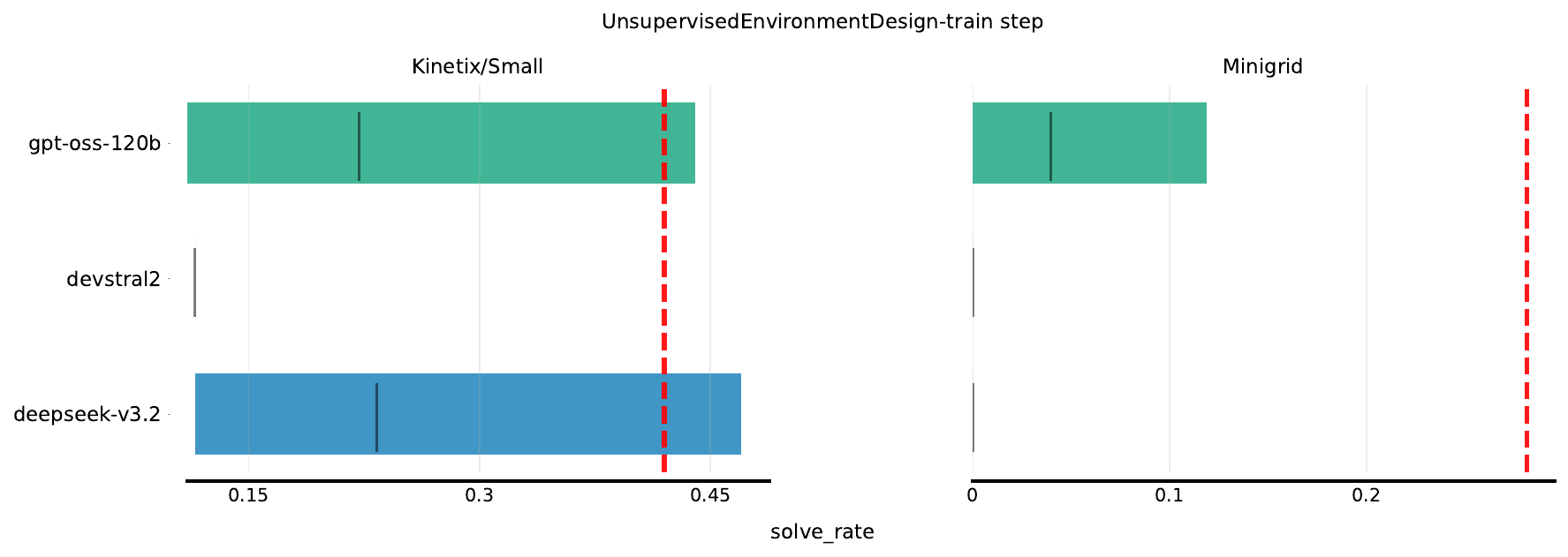}%
\hfill%
\caption{DiscoBench (Single Edit) results on Meta-Train tasks. (Part 7/7)}
\label{fig:one_change_id_7}
\end{figure}
\clearpage

\subsection{DiscoBench (Single Edit) -- Meta-Test}
\label{sec:one_change_mt}

\begin{figure}[htbp]
\centering
\setlength{\lineskip}{0pt}
\includegraphics[width=0.48\textwidth]{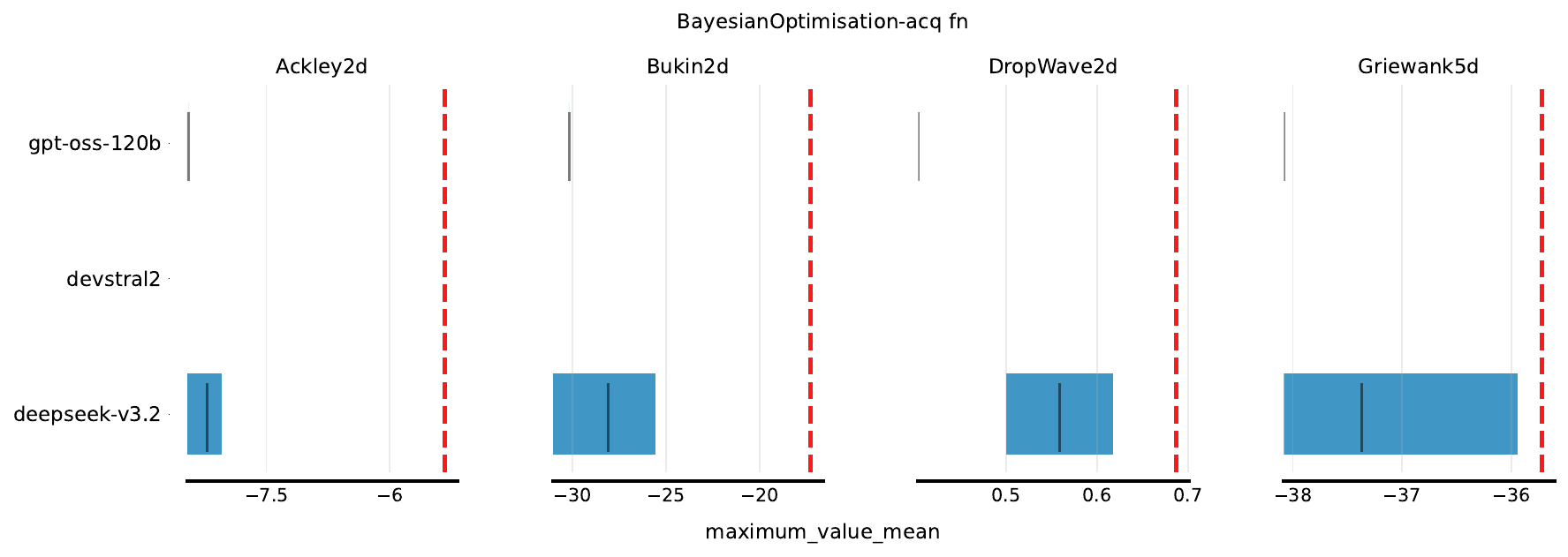}%
\hfill%
\includegraphics[width=0.48\textwidth]{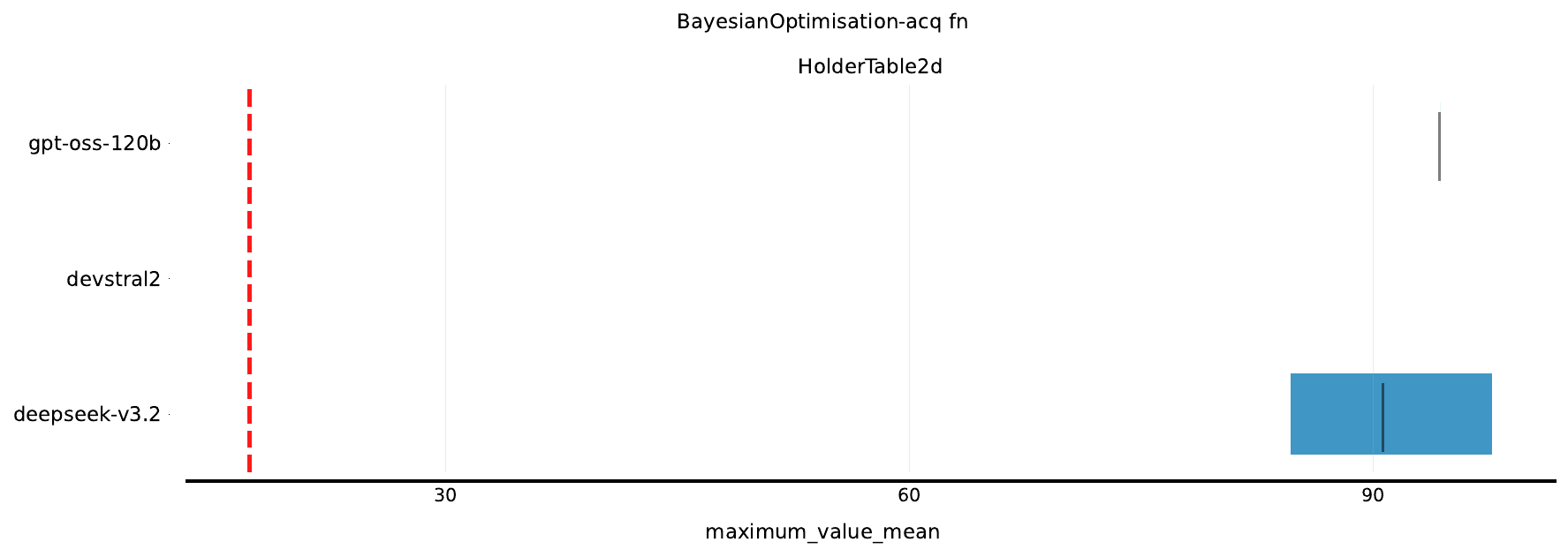}%
\\[0.5em]
\includegraphics[width=0.48\textwidth]{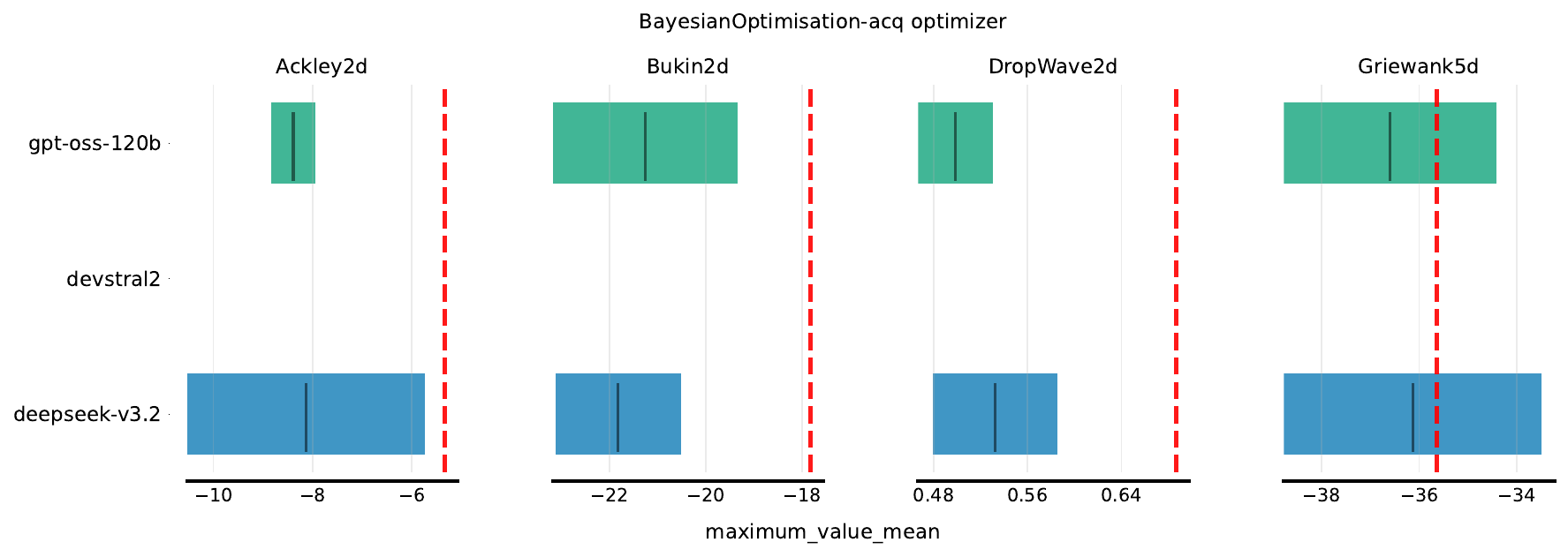}%
\hfill%
\includegraphics[width=0.48\textwidth]{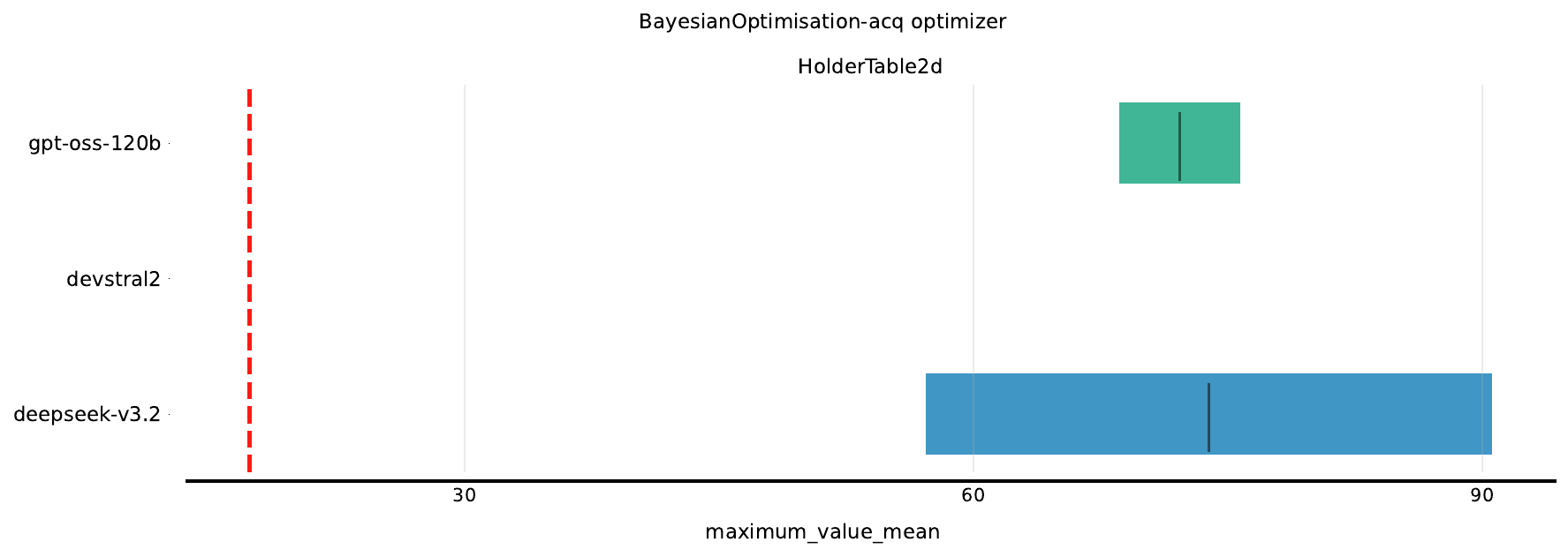}%
\\[0.5em]
\includegraphics[width=0.48\textwidth]{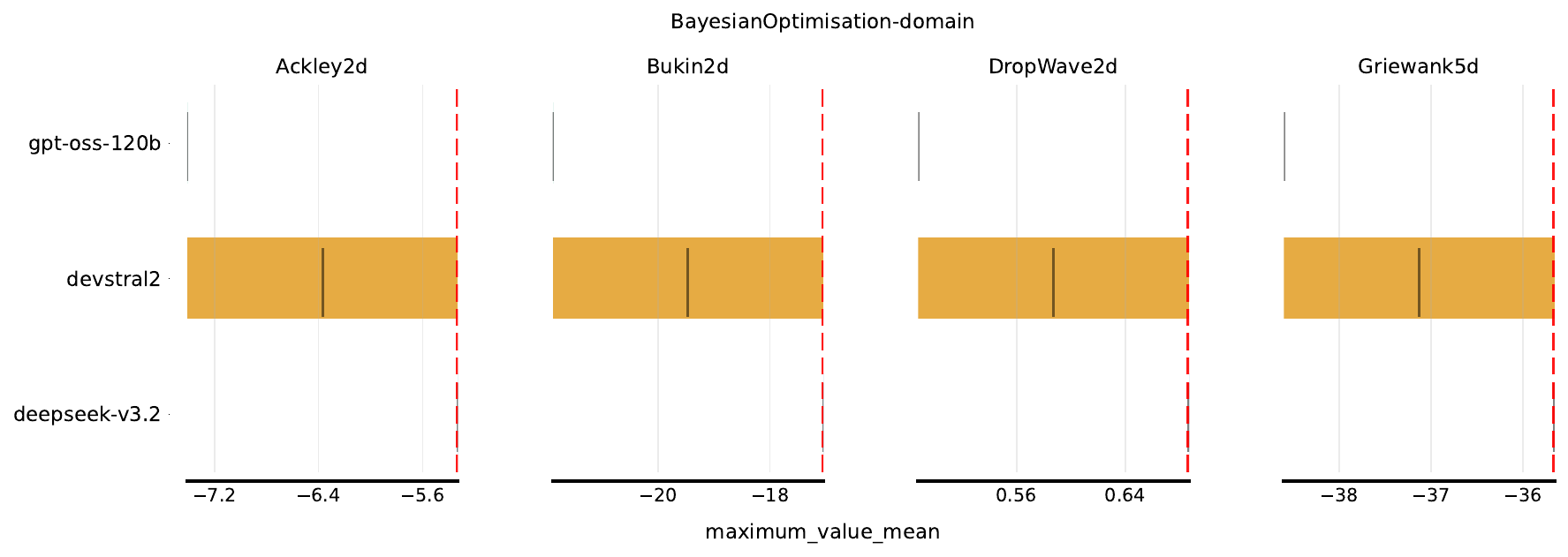}%
\hfill%
\includegraphics[width=0.48\textwidth]{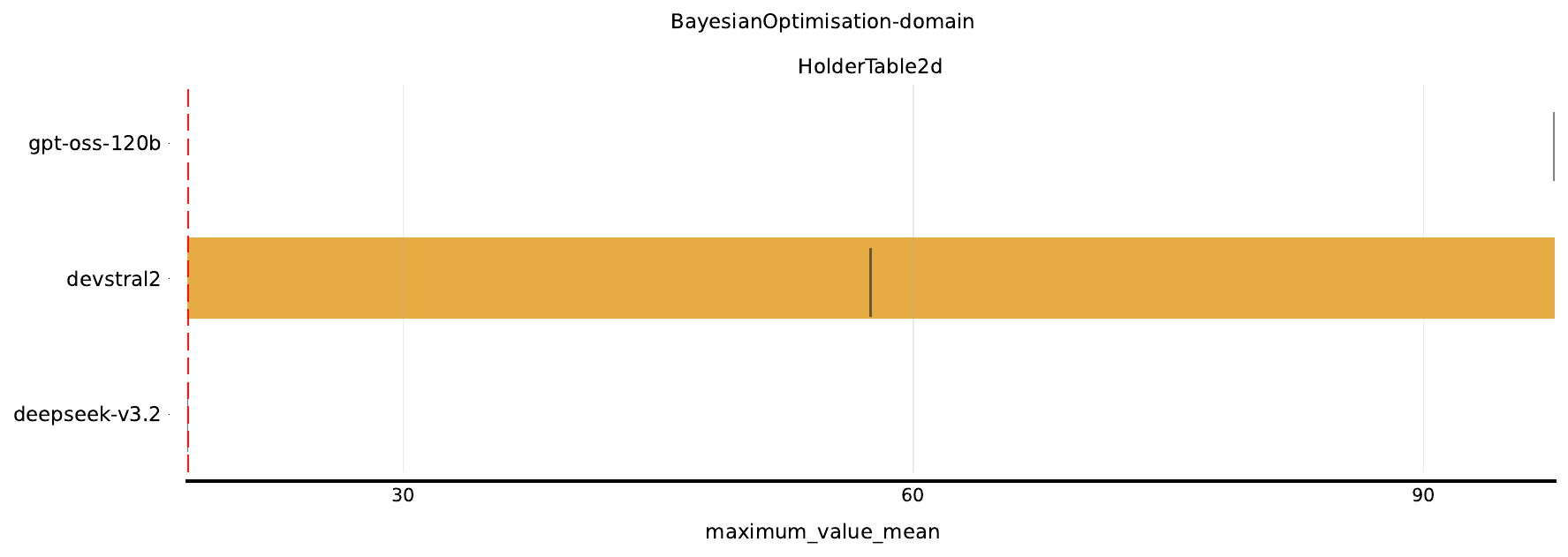}%
\\[0.5em]
\includegraphics[width=0.48\textwidth]{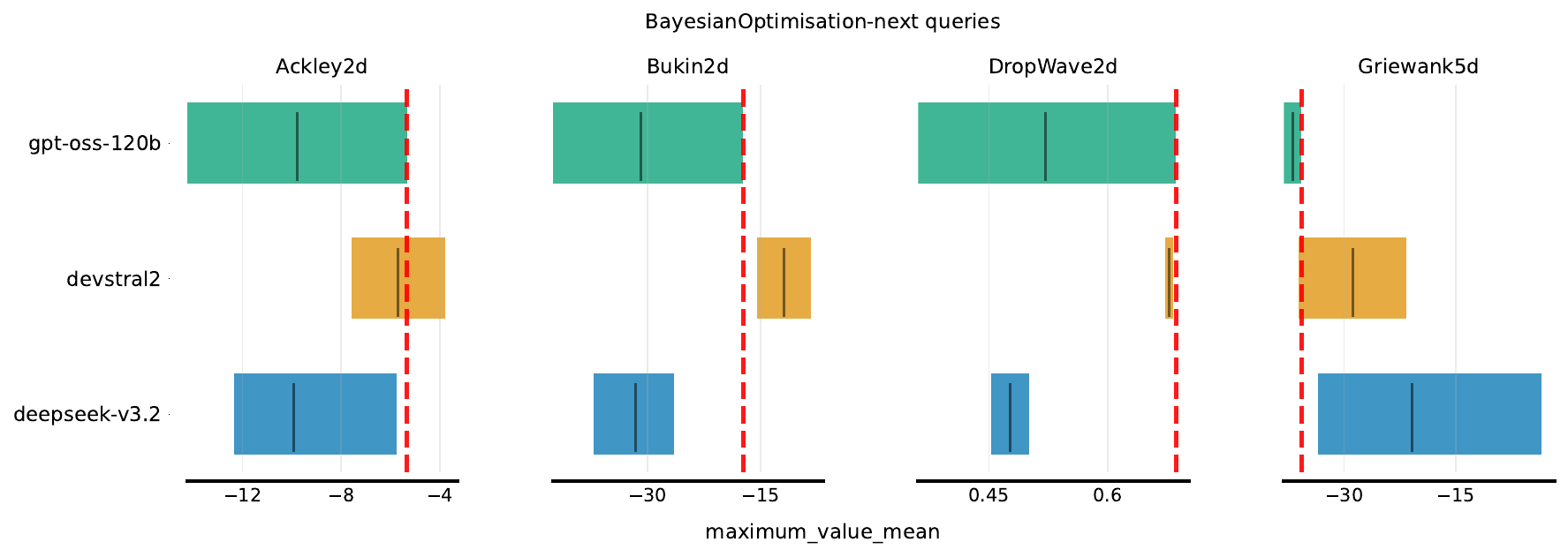}%
\hfill%
\includegraphics[width=0.48\textwidth]{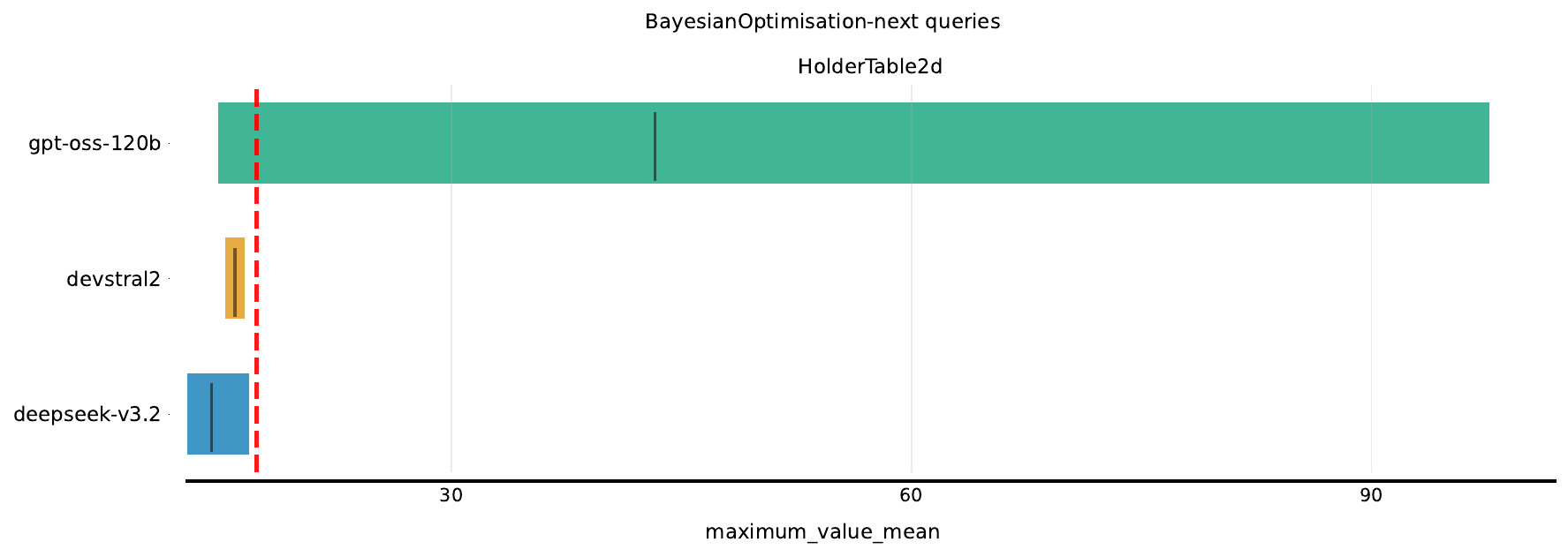}%
\\[0.5em]
\includegraphics[width=0.48\textwidth]{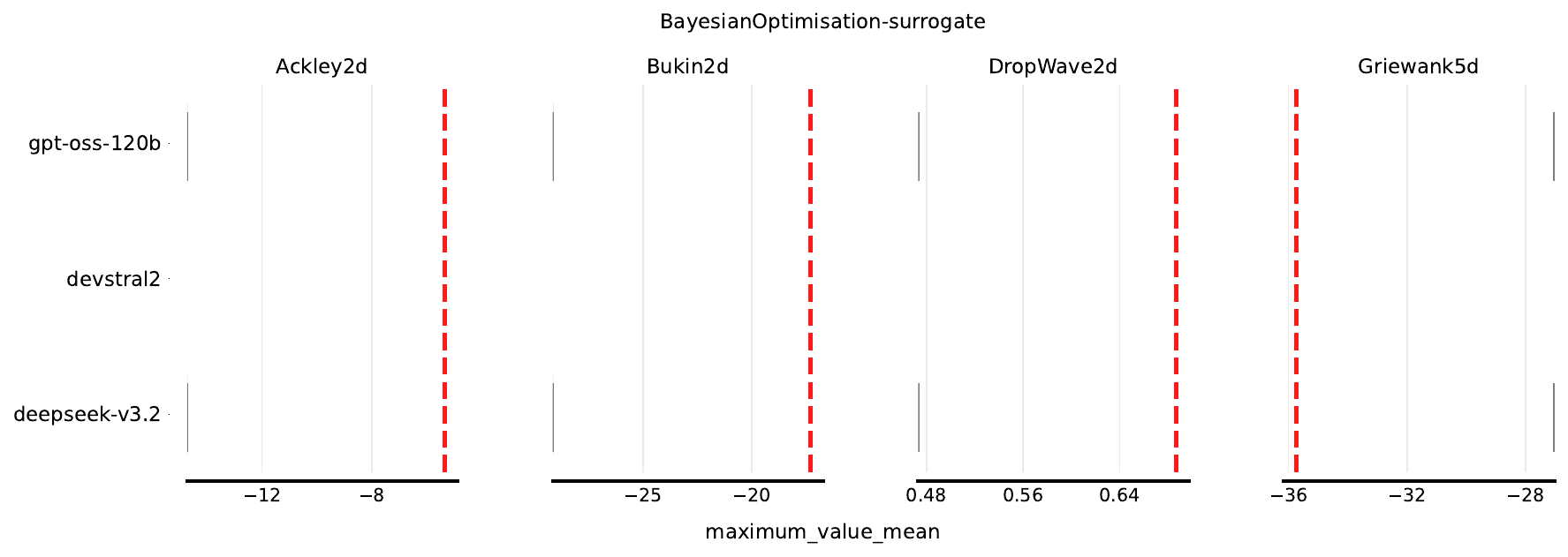}%
\hfill%
\includegraphics[width=0.48\textwidth]{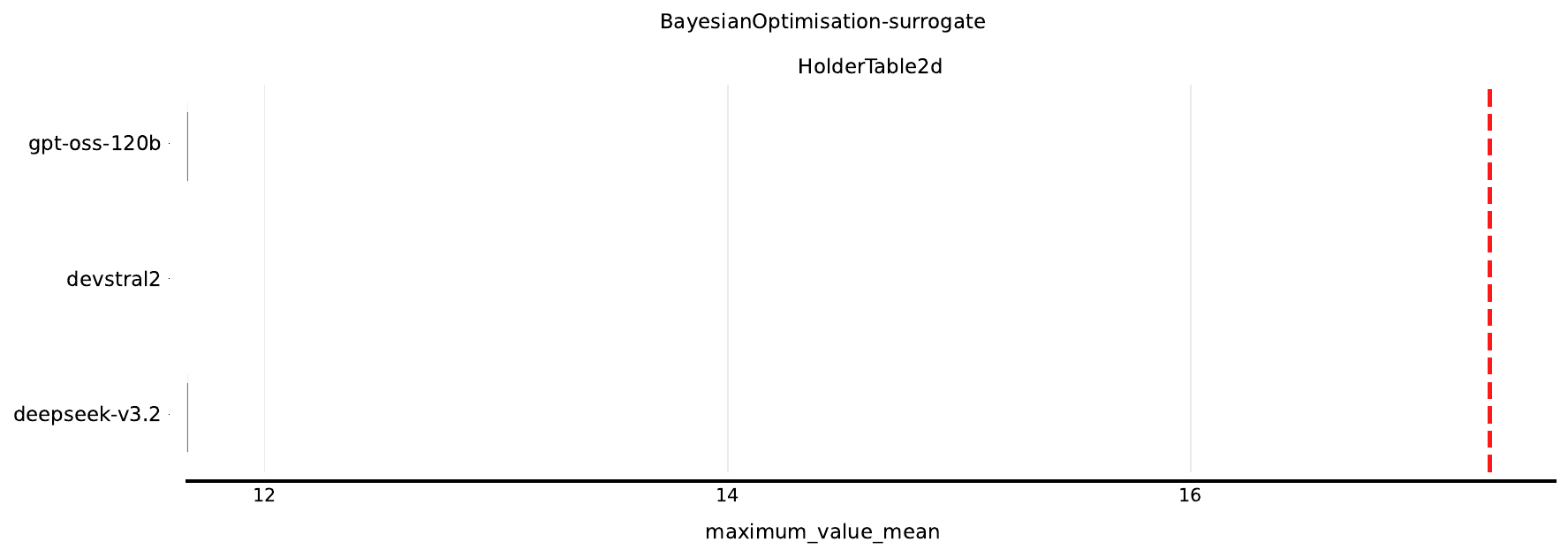}%
\caption{DiscoBench (Single Edit) results on Meta-Test tasks. (Part 1/7)}
\label{fig:one_change_mt_1}
\end{figure}
\clearpage

\begin{figure}[htbp]
\centering
\setlength{\lineskip}{0pt}
\includegraphics[width=0.48\textwidth]{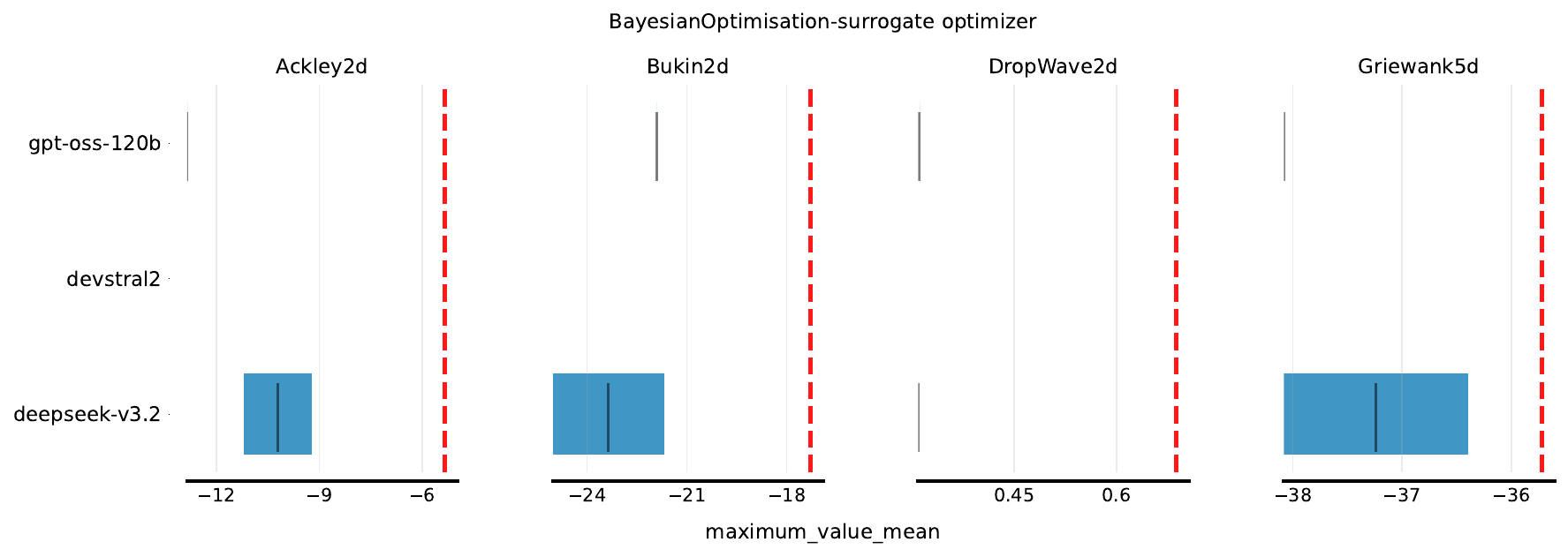}%
\hfill%
\includegraphics[width=0.48\textwidth]{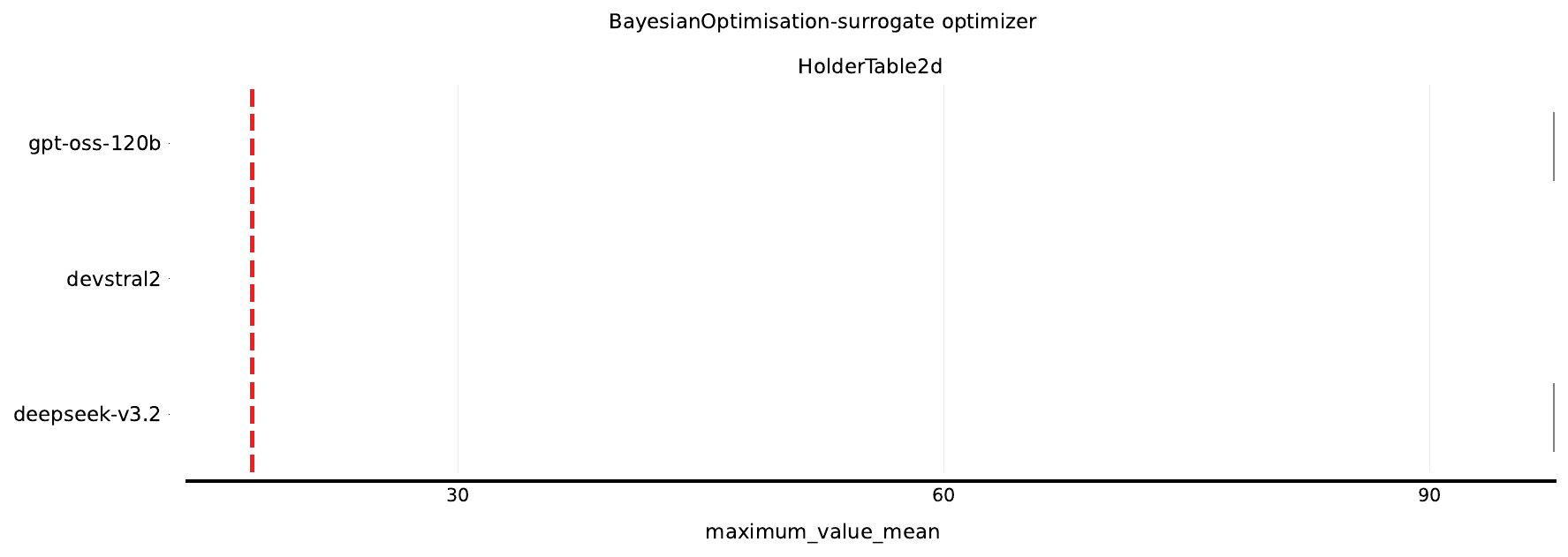}%
\\[0.5em]
\includegraphics[width=0.48\textwidth]{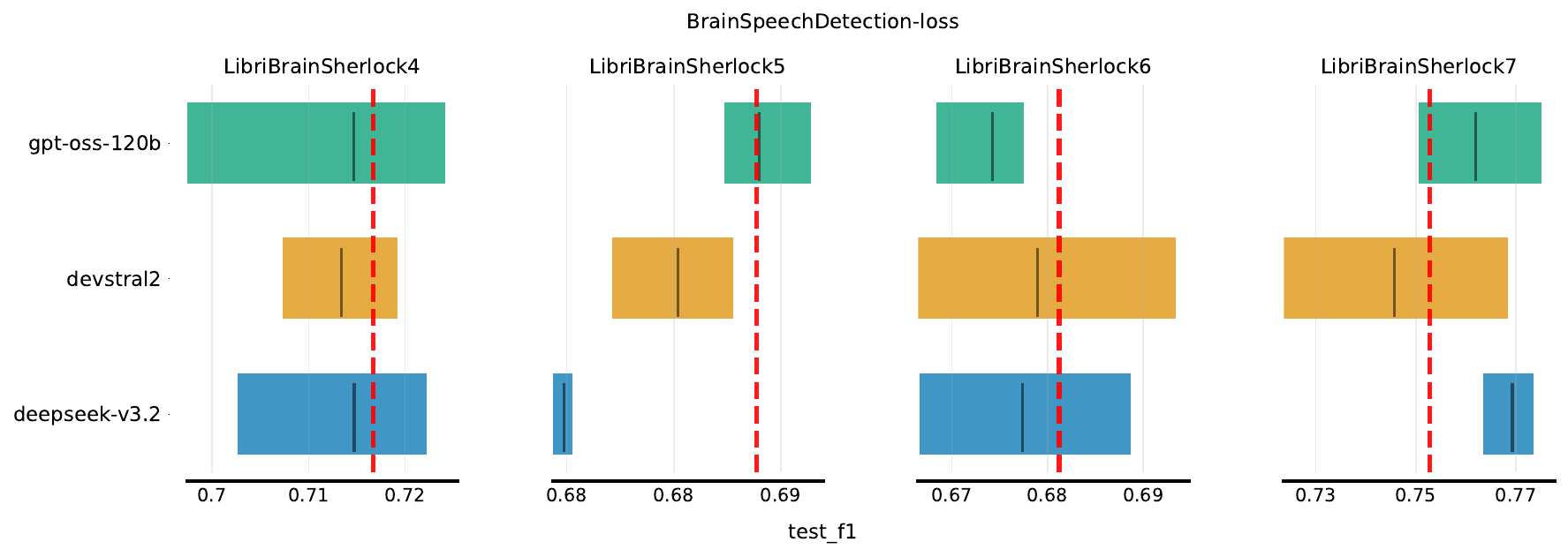}%
\hfill%
\includegraphics[width=0.48\textwidth]{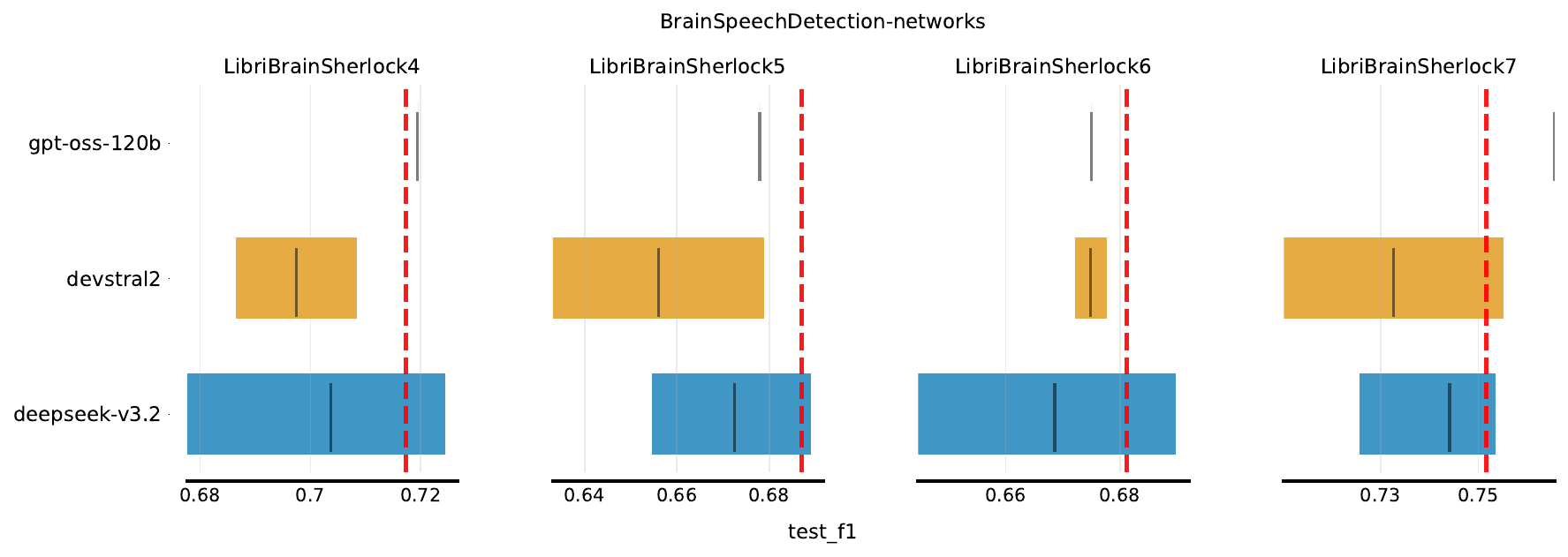}%
\\[0.5em]
\includegraphics[width=0.48\textwidth]{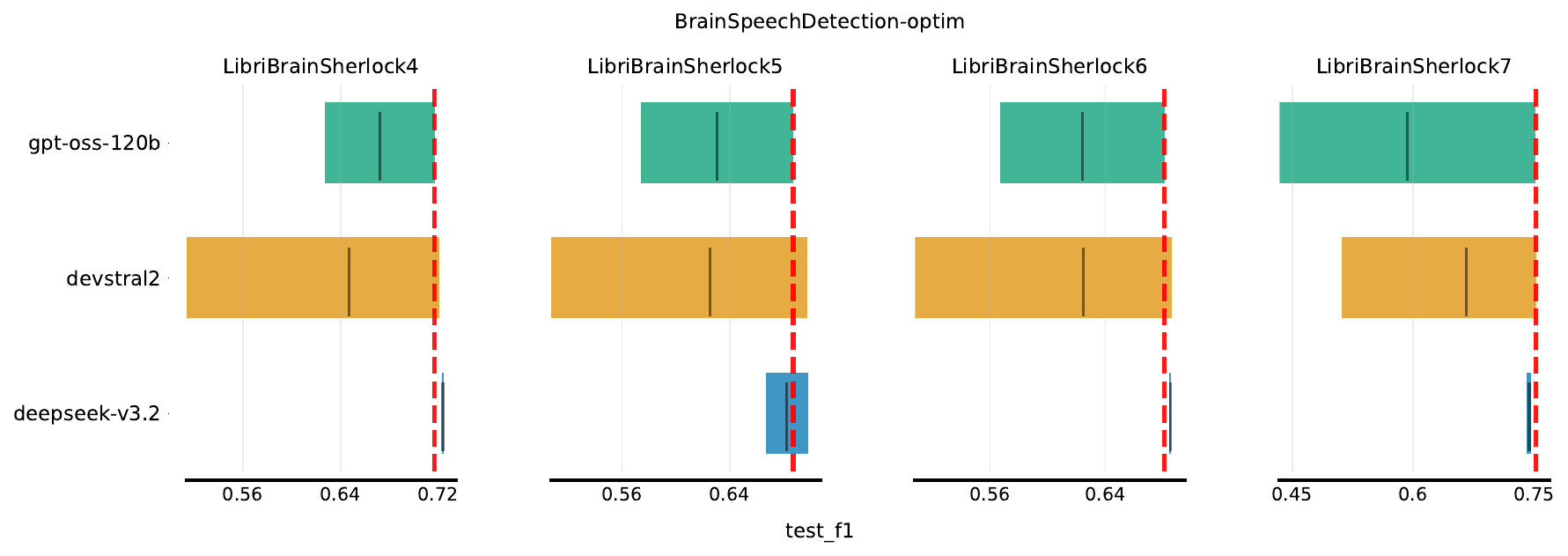}%
\hfill%
\includegraphics[width=0.48\textwidth]{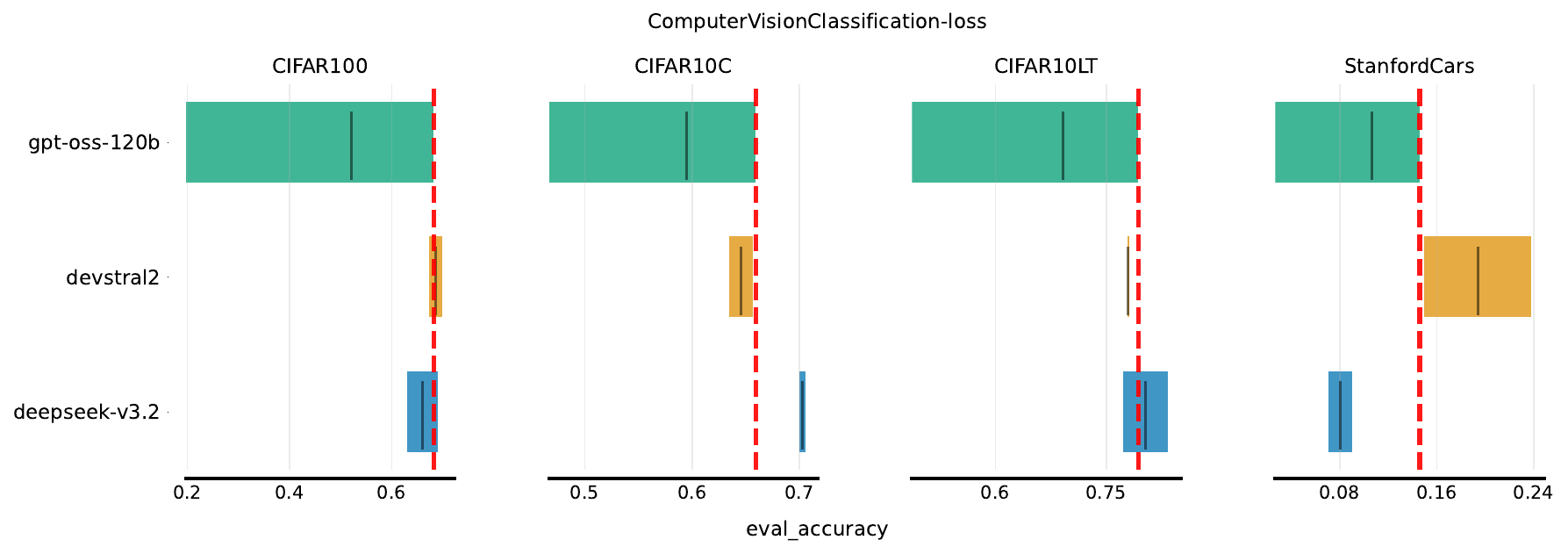}%
\\[0.5em]
\includegraphics[width=0.48\textwidth]{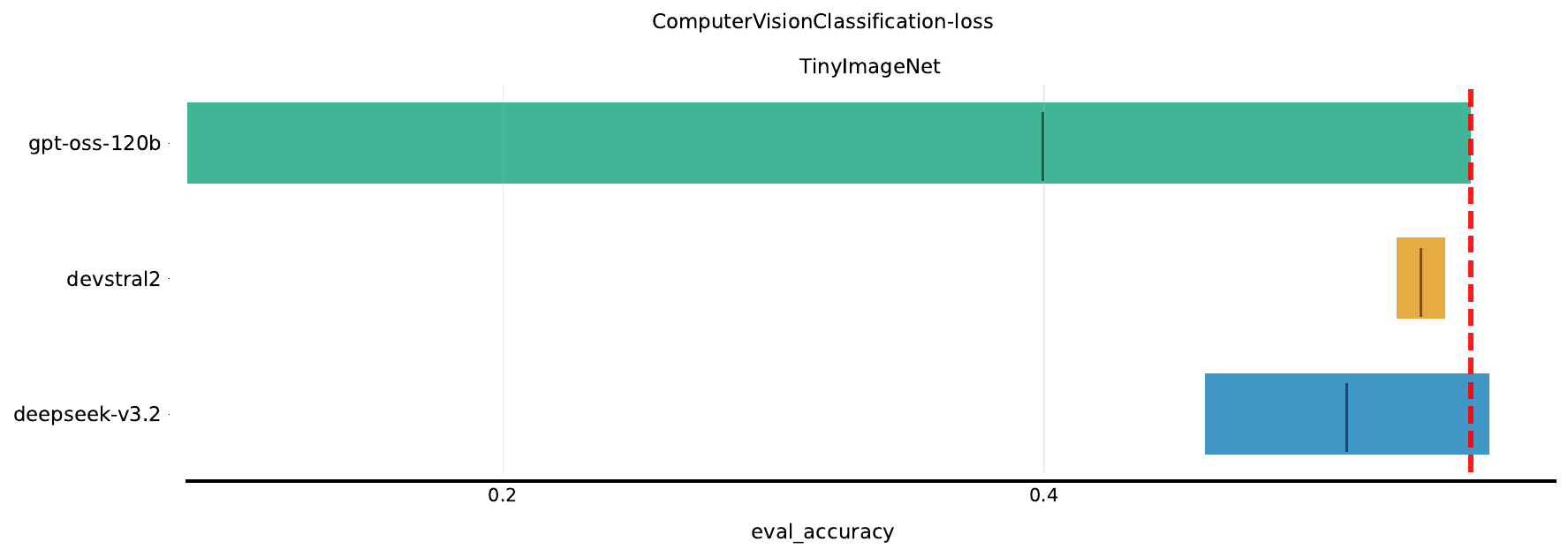}%
\hfill%
\includegraphics[width=0.48\textwidth]{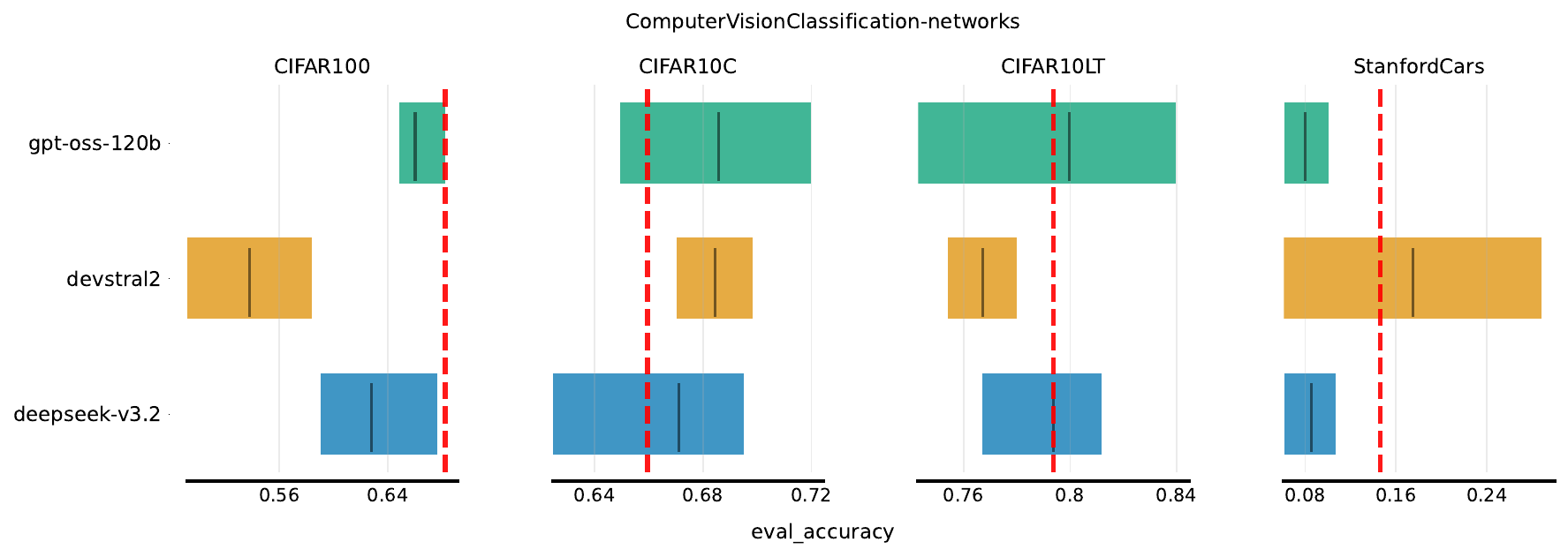}%
\\[0.5em]
\includegraphics[width=0.48\textwidth]{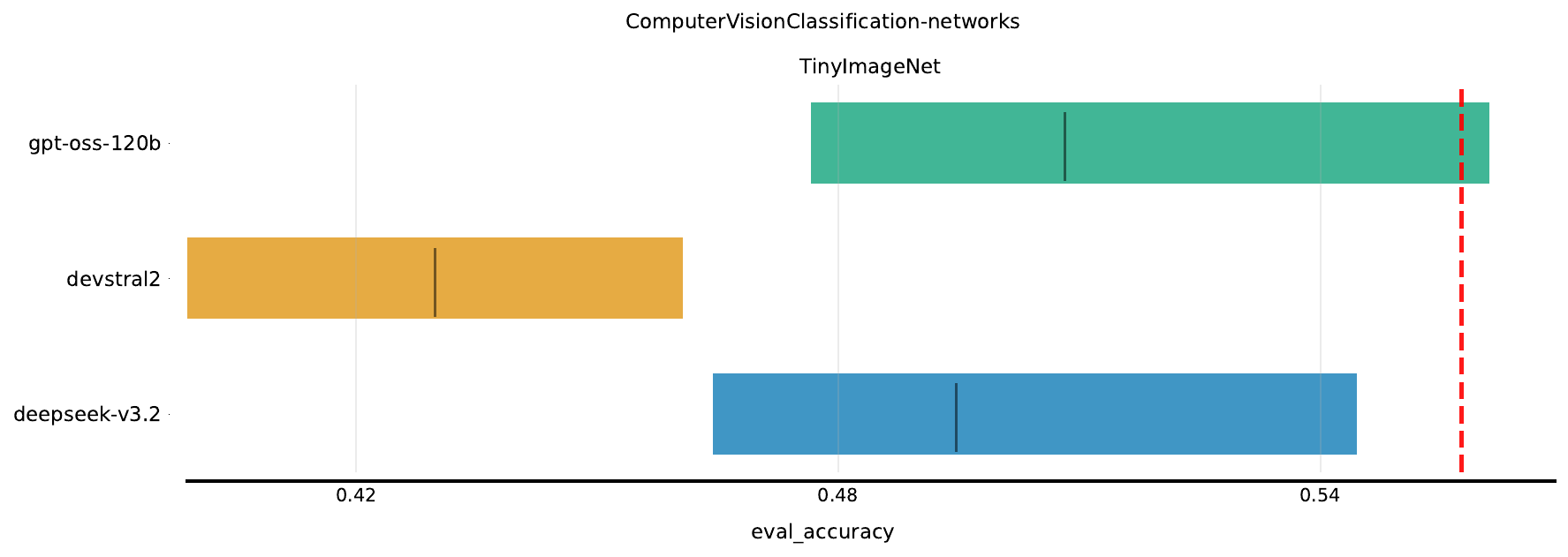}%
\hfill%
\includegraphics[width=0.48\textwidth]{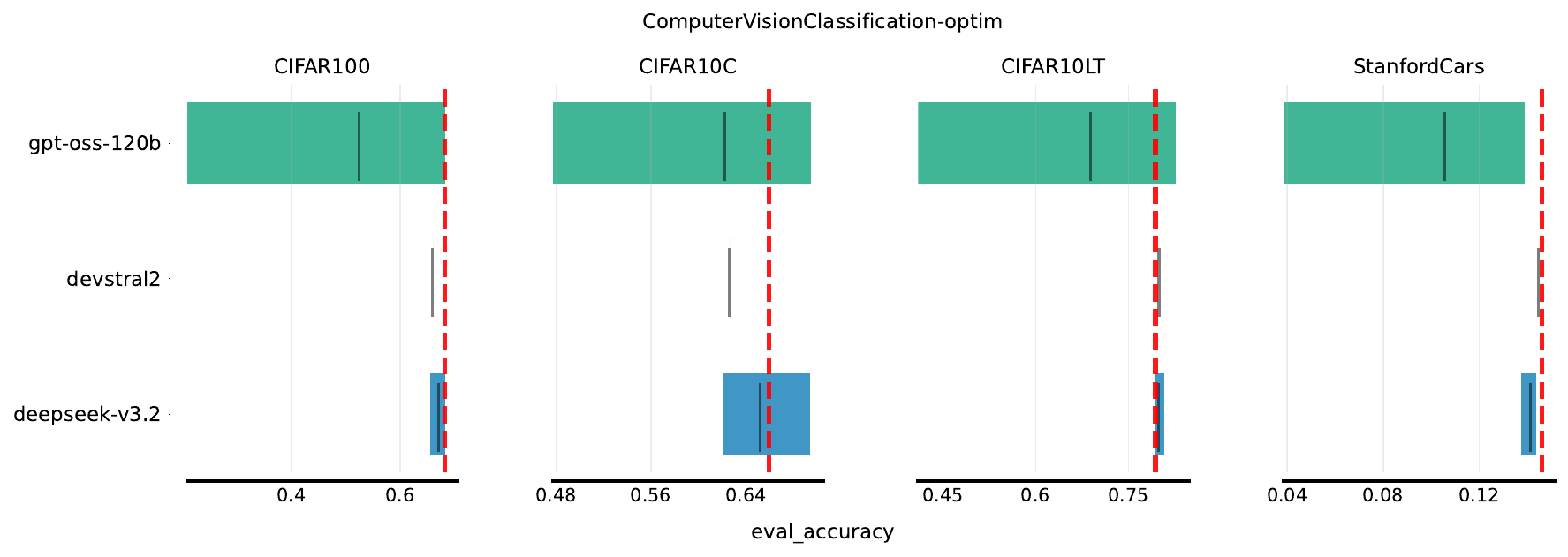}%
\\[0.5em]
\includegraphics[width=0.48\textwidth]{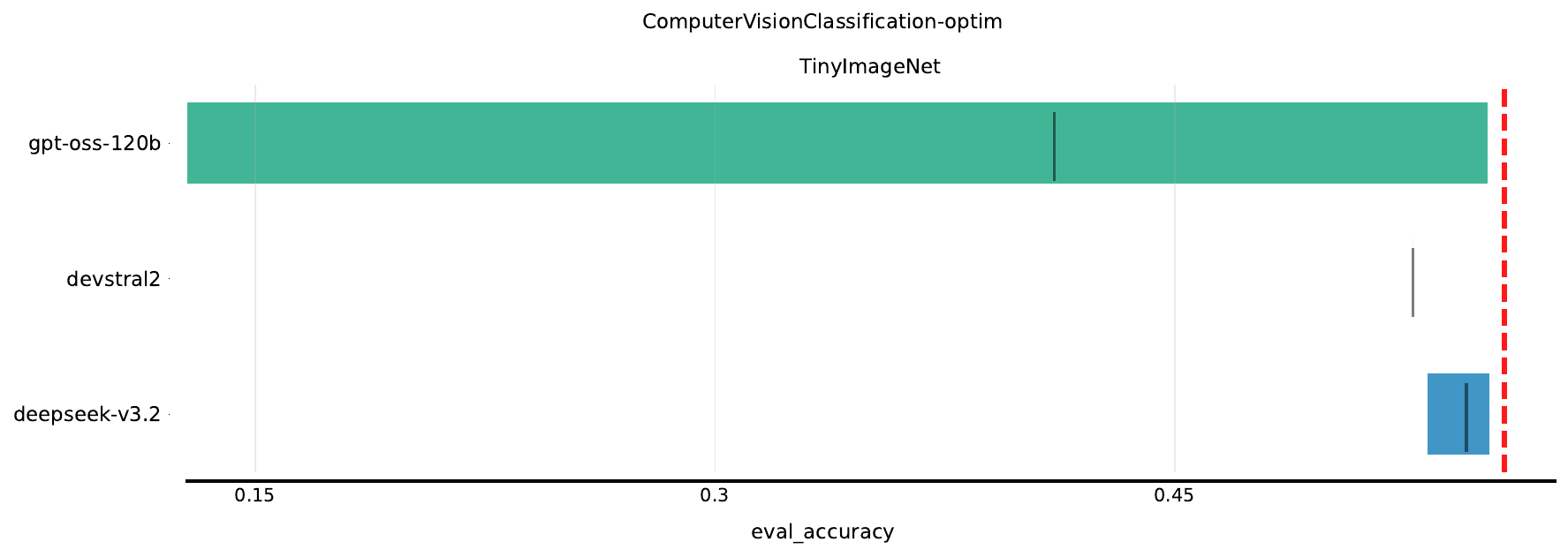}%
\hfill%
\includegraphics[width=0.48\textwidth]{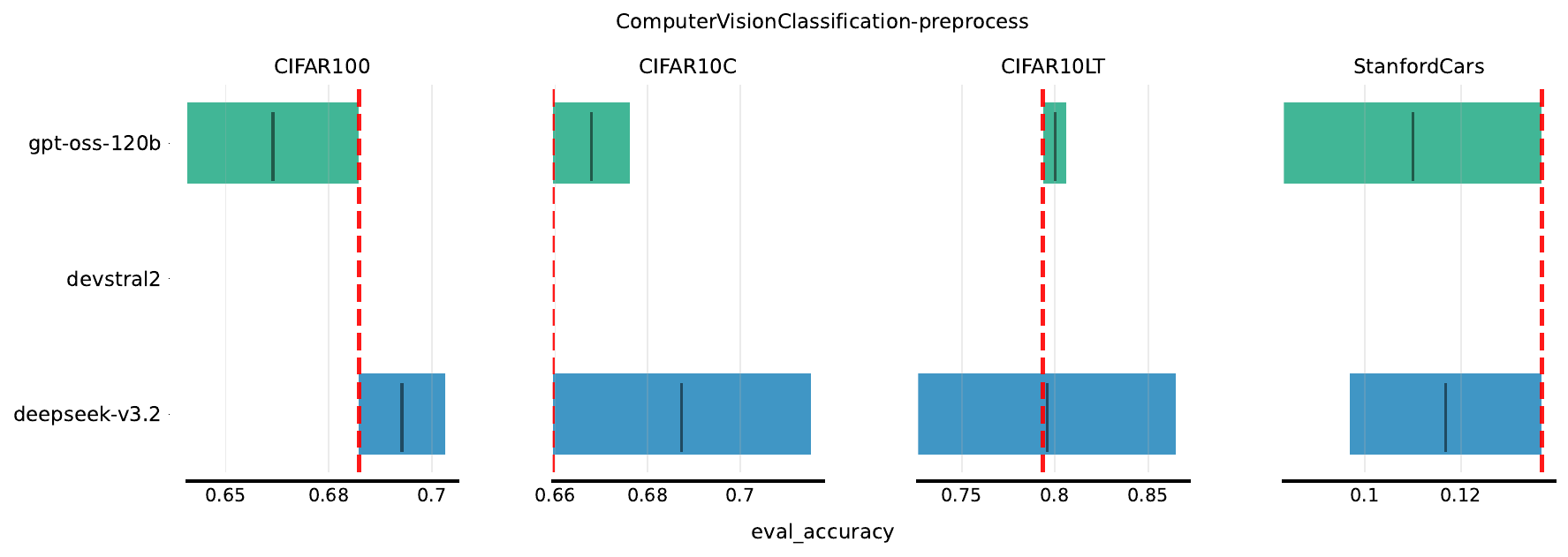}%
\caption{DiscoBench (Single Edit) results on Meta-Test tasks. (Part 2/7)}
\label{fig:one_change_mt_2}
\end{figure}
\clearpage

\begin{figure}[htbp]
\centering
\setlength{\lineskip}{0pt}
\includegraphics[width=0.48\textwidth]{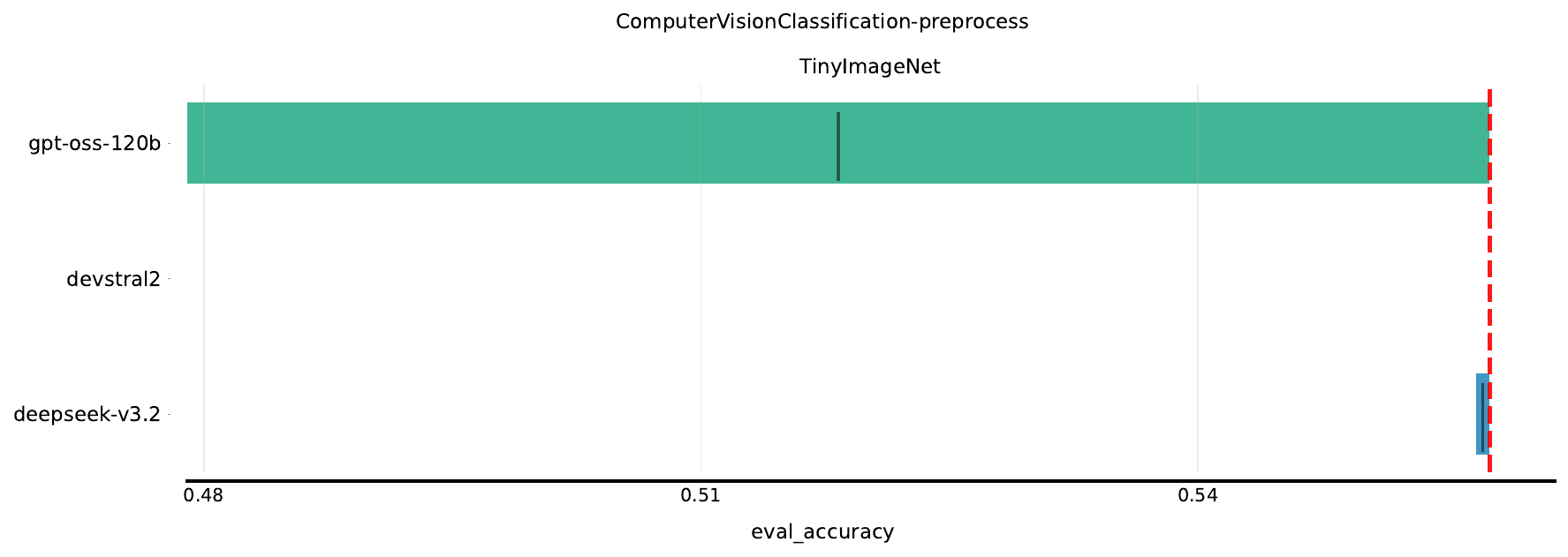}%
\hfill%
\includegraphics[width=0.48\textwidth]{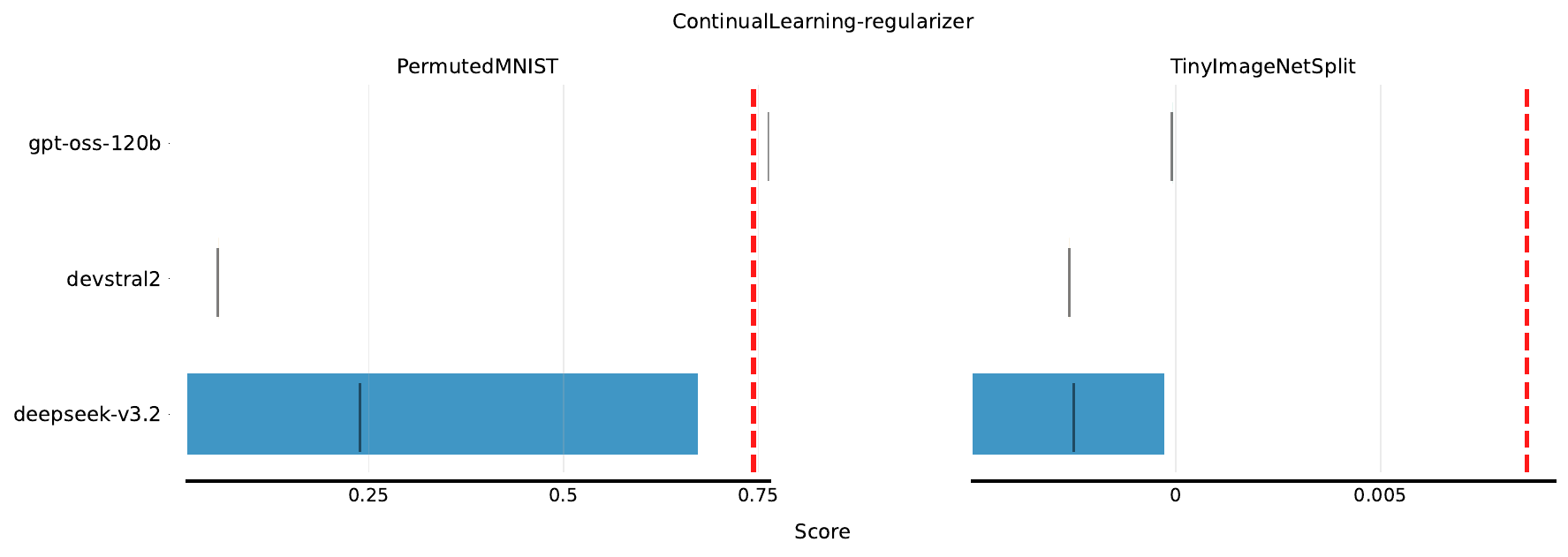}%
\\[0.5em]
\includegraphics[width=0.48\textwidth]{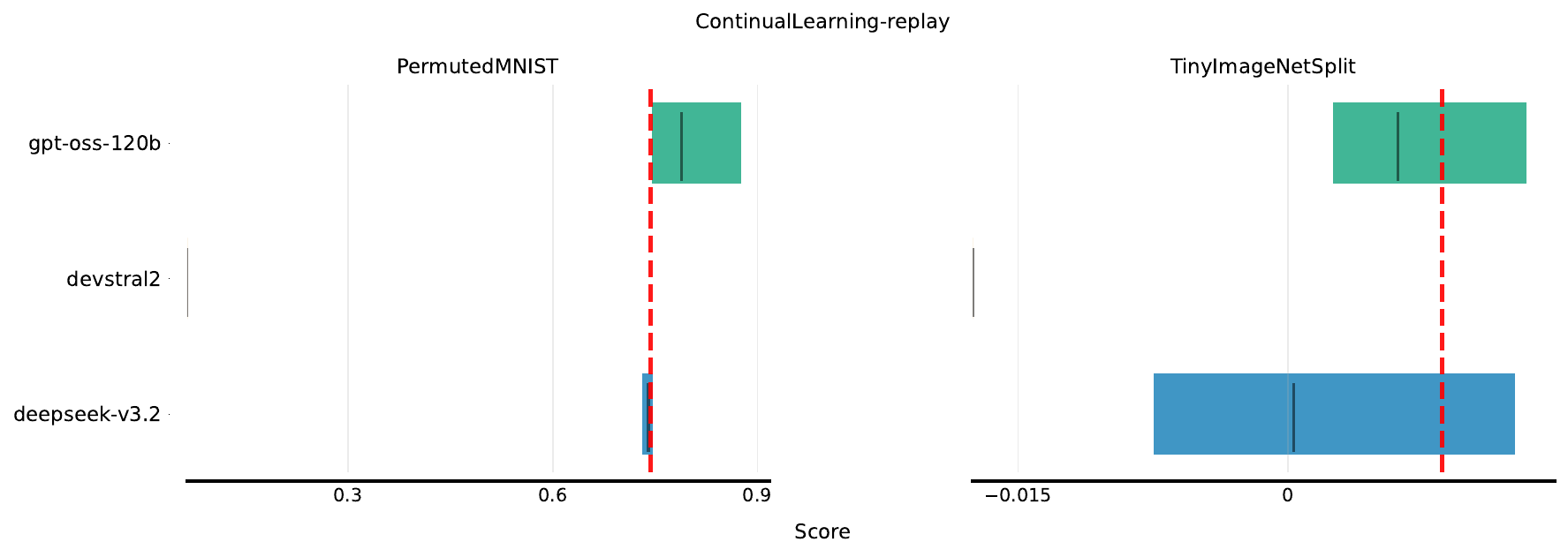}%
\hfill%
\includegraphics[width=0.48\textwidth]{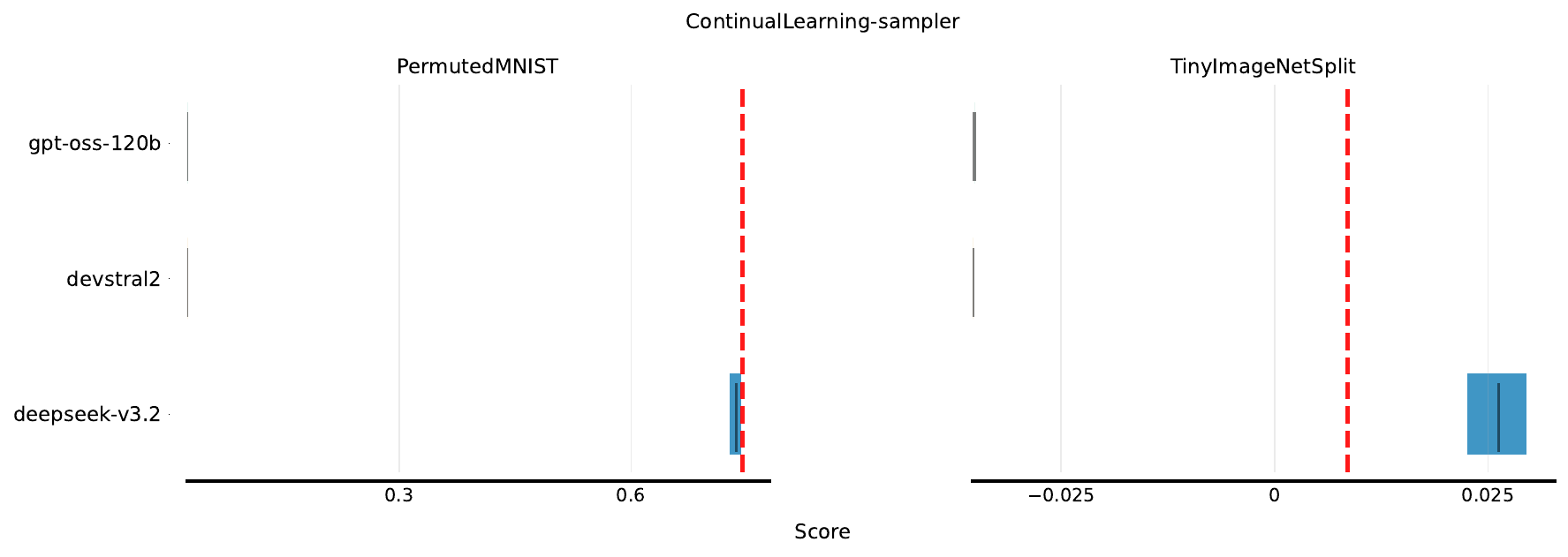}%
\\[0.5em]
\includegraphics[width=0.48\textwidth]{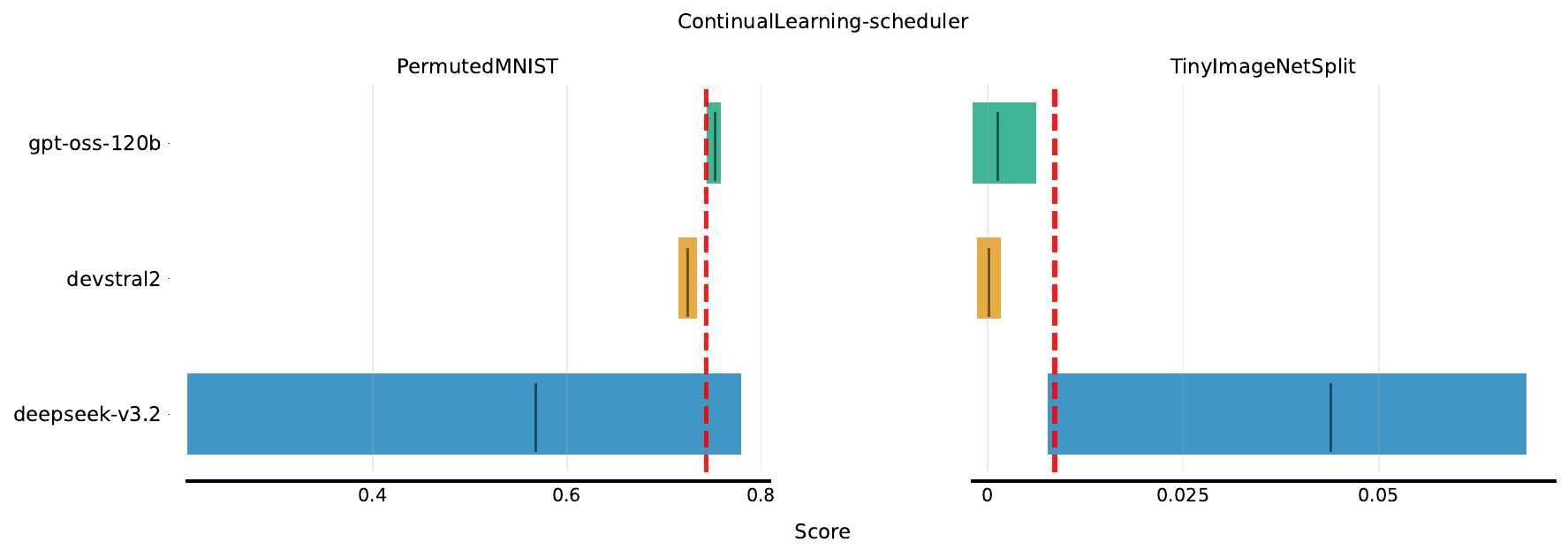}%
\hfill%
\includegraphics[width=0.48\textwidth]{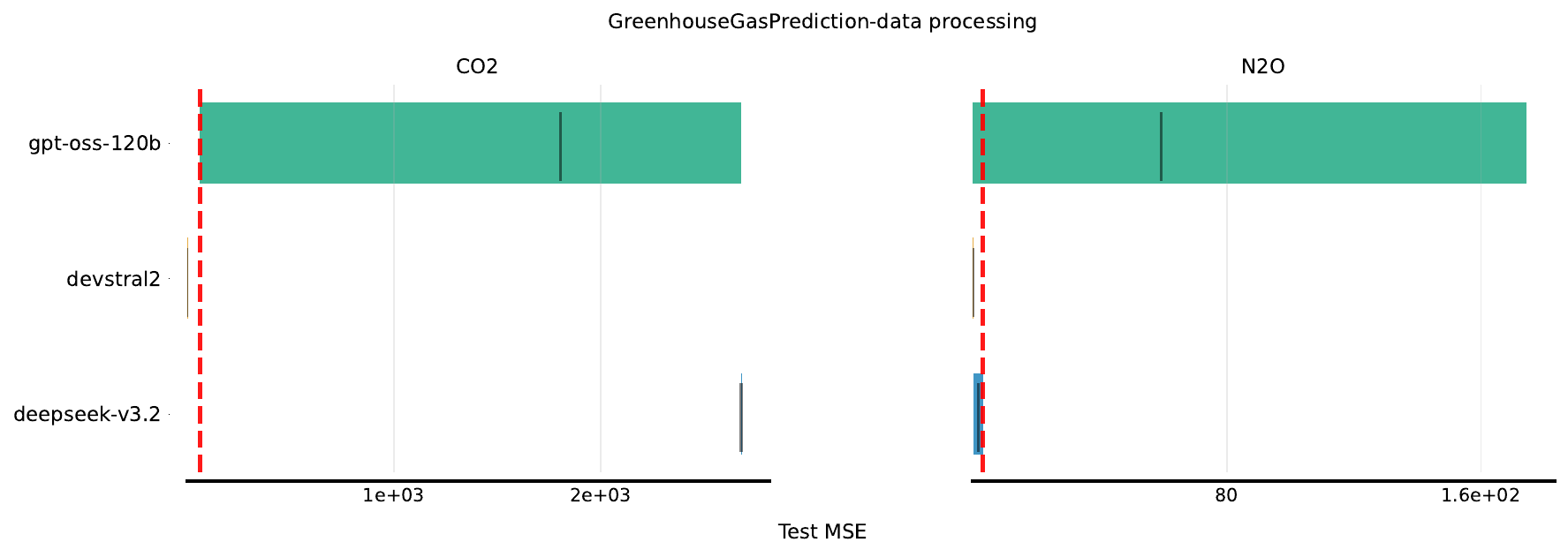}%
\\[0.5em]
\includegraphics[width=0.48\textwidth]{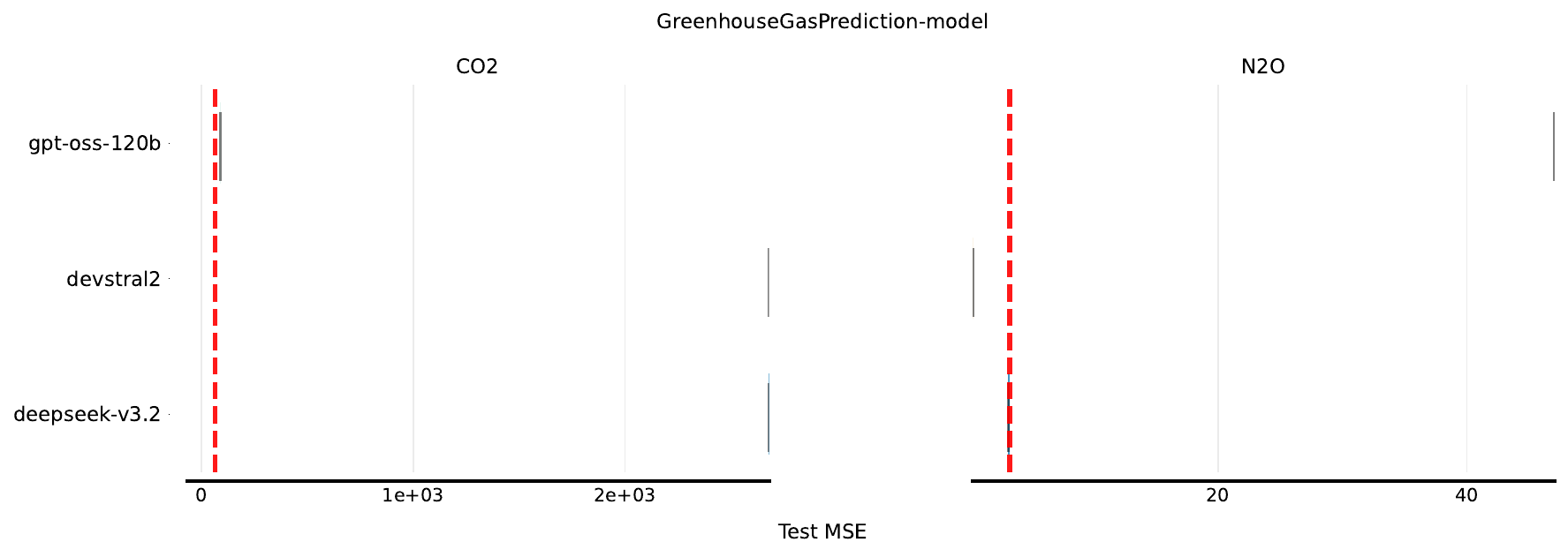}%
\hfill%
\includegraphics[width=0.48\textwidth]{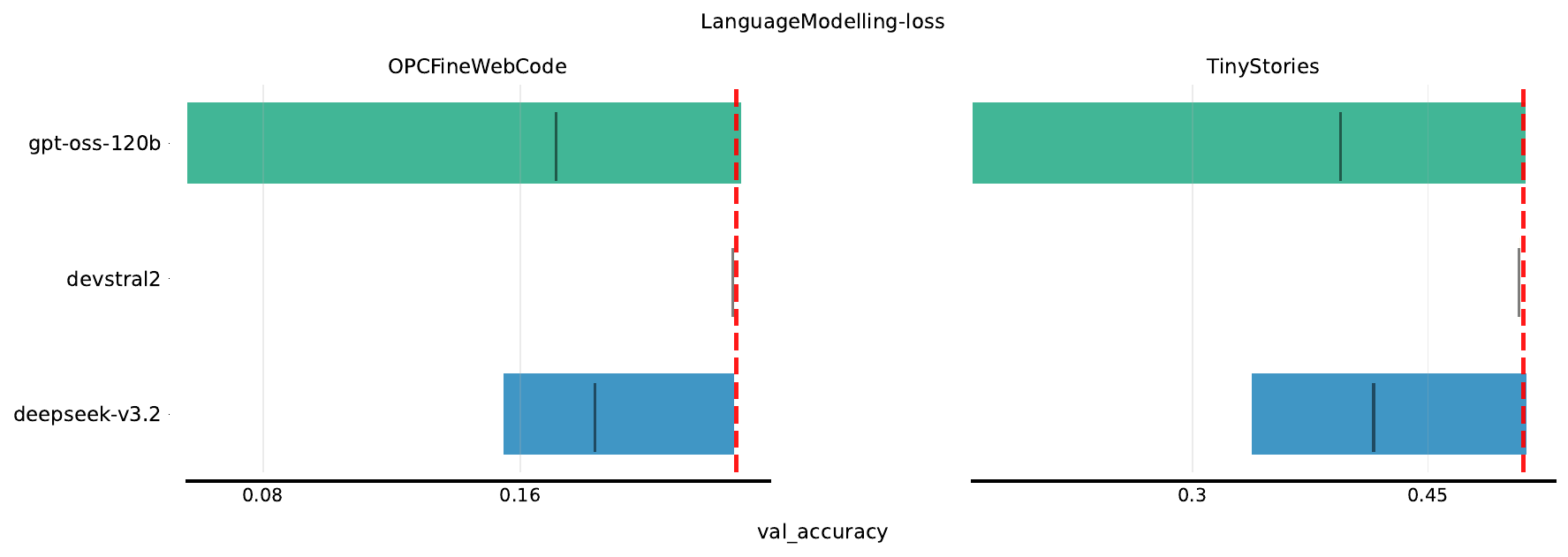}%
\\[0.5em]
\includegraphics[width=0.48\textwidth]{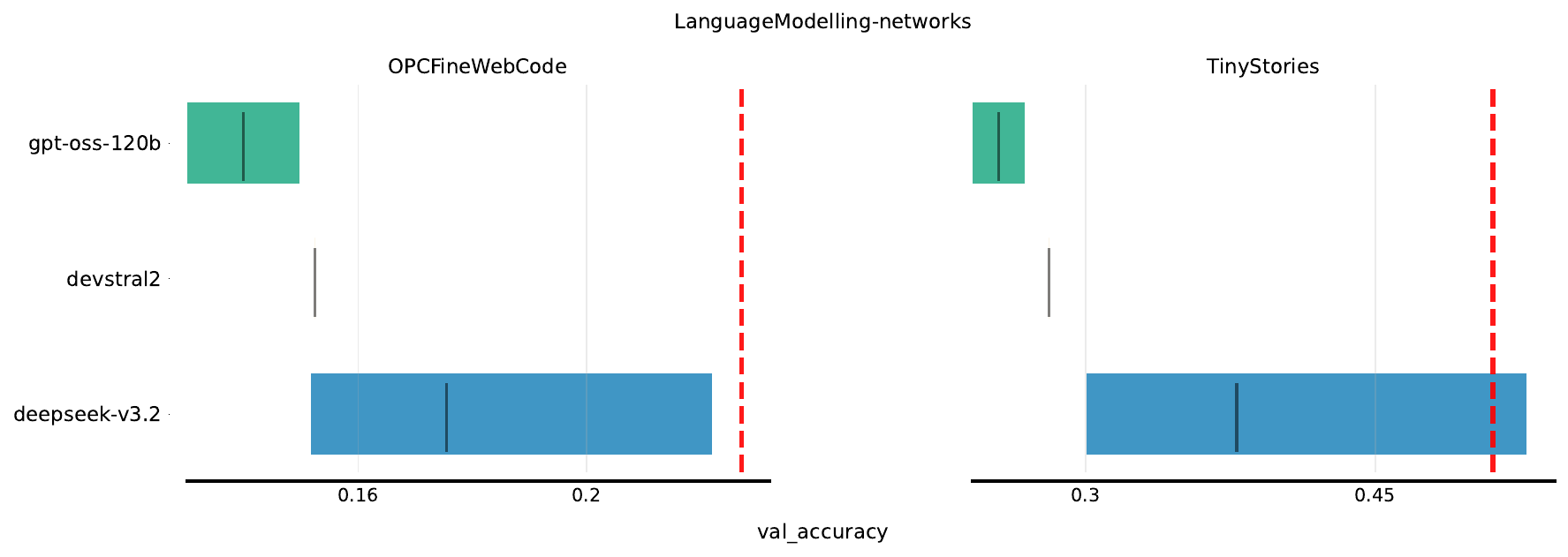}%
\hfill%
\includegraphics[width=0.48\textwidth]{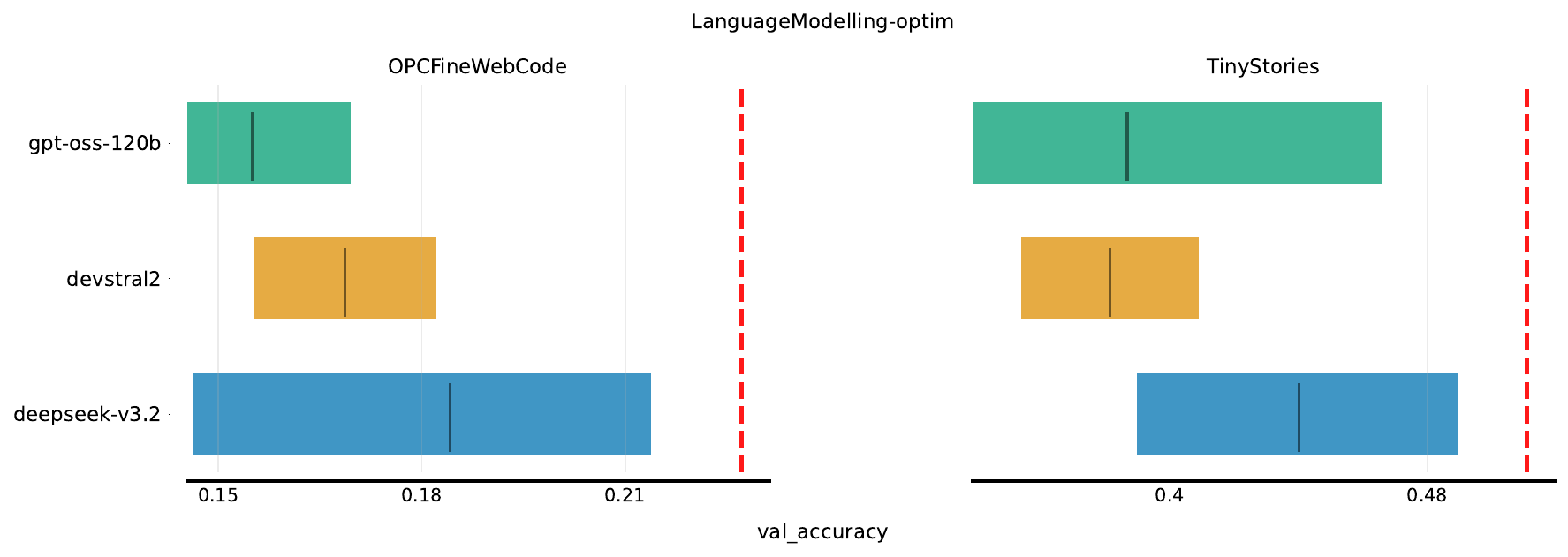}%
\\[0.5em]
\includegraphics[width=0.48\textwidth]{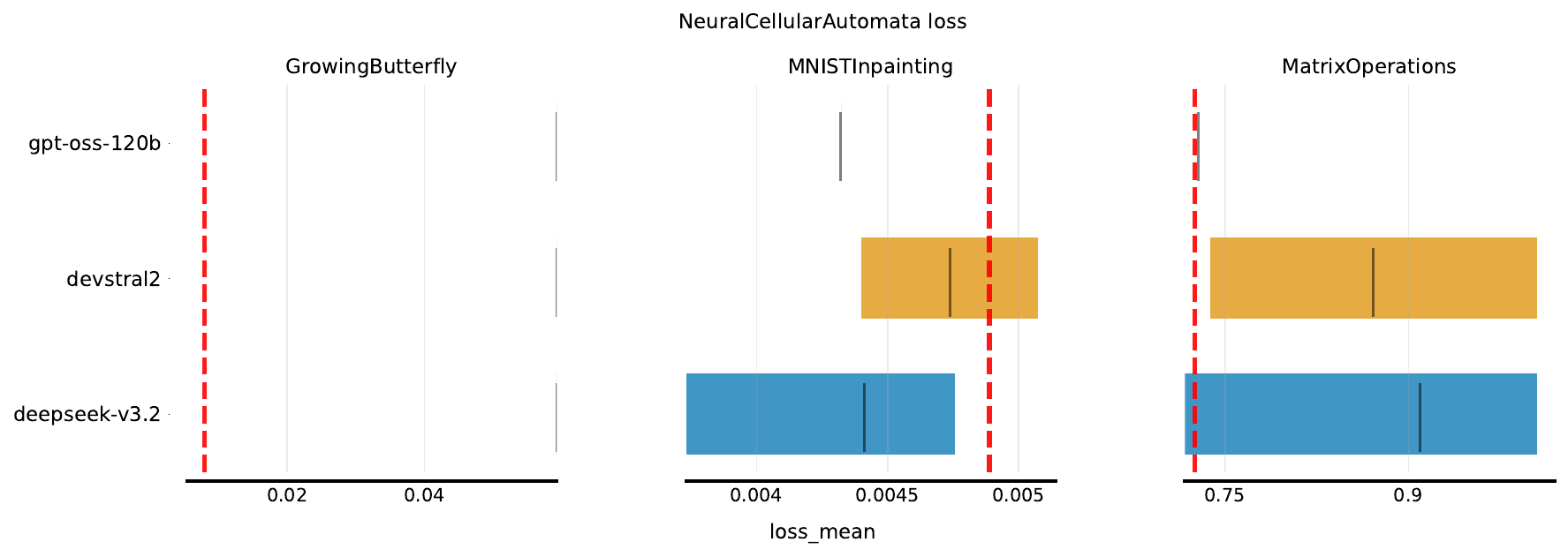}%
\hfill%
\includegraphics[width=0.48\textwidth]{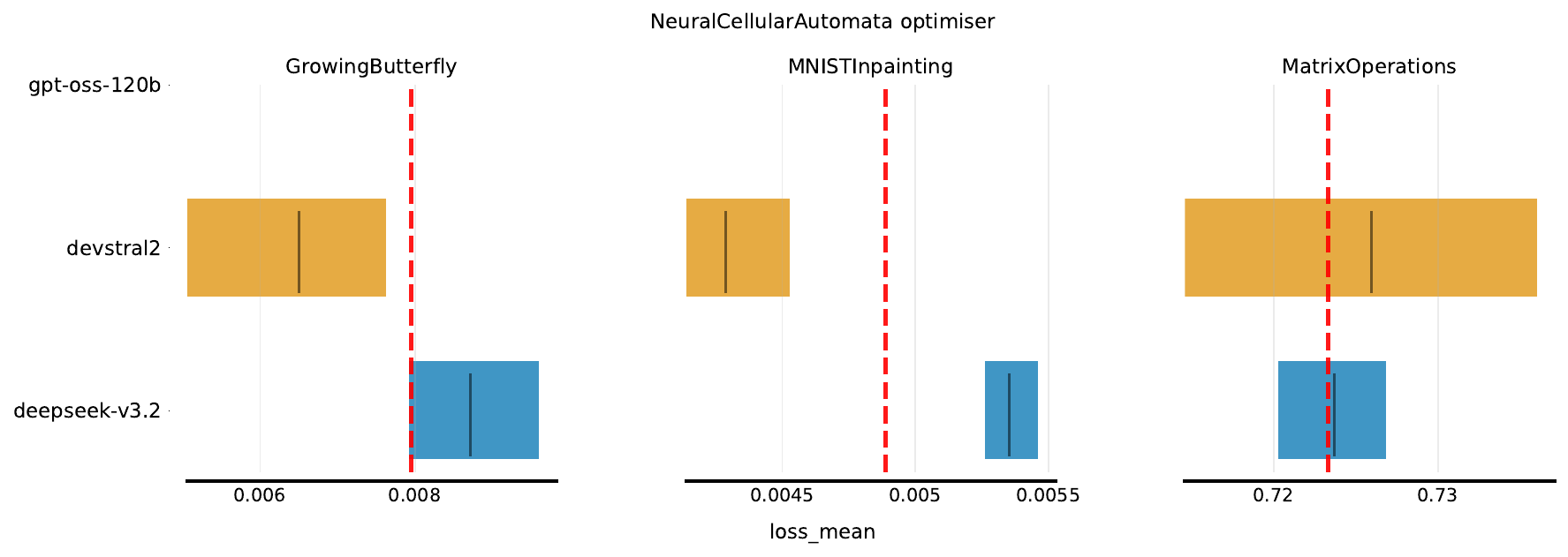}%
\caption{DiscoBench (Single Edit) results on Meta-Test tasks. (Part 3/7)}
\label{fig:one_change_mt_3}
\end{figure}
\clearpage

\begin{figure}[htbp]
\centering
\setlength{\lineskip}{0pt}
\includegraphics[width=0.48\textwidth]{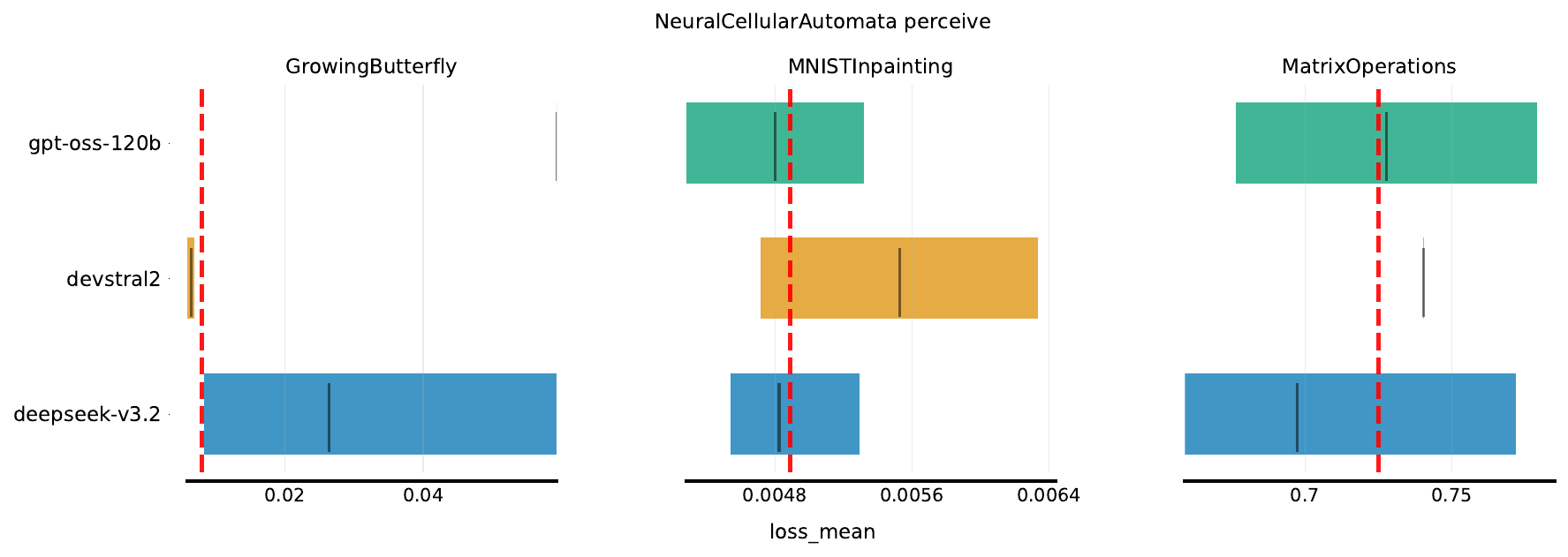}%
\hfill%
\includegraphics[width=0.48\textwidth]{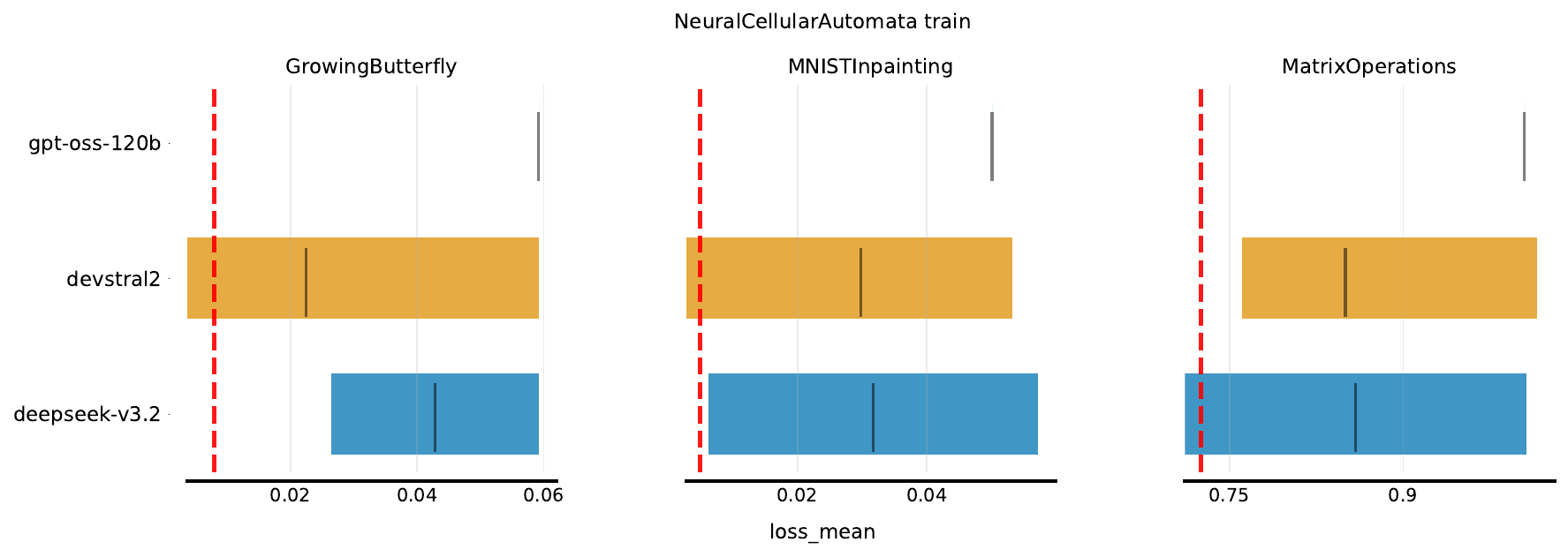}%
\\[0.5em]
\includegraphics[width=0.48\textwidth]{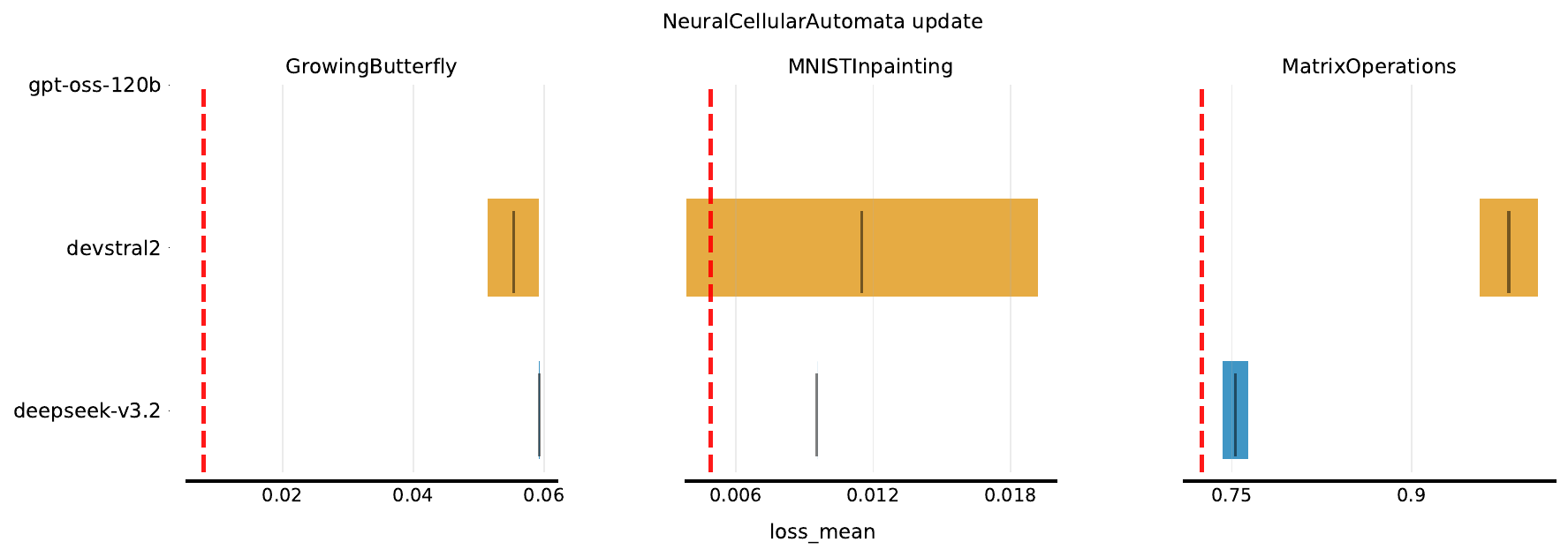}%
\hfill%
\includegraphics[width=0.48\textwidth]{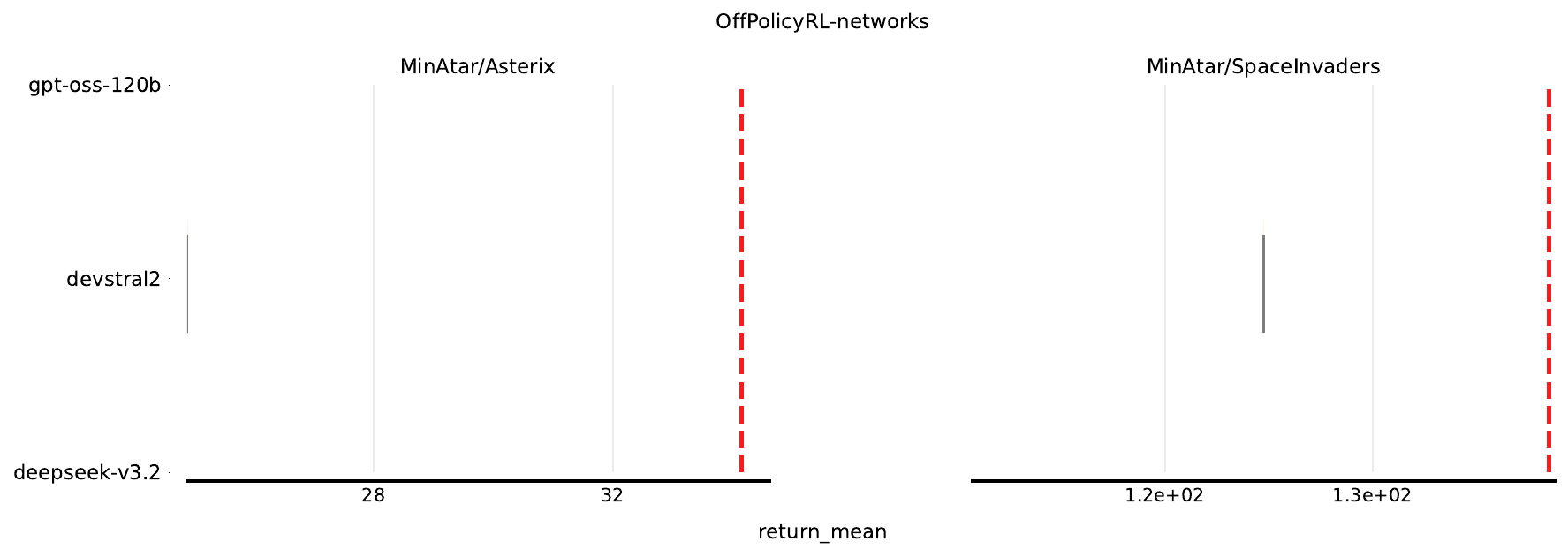}%
\\[0.5em]
\includegraphics[width=0.48\textwidth]{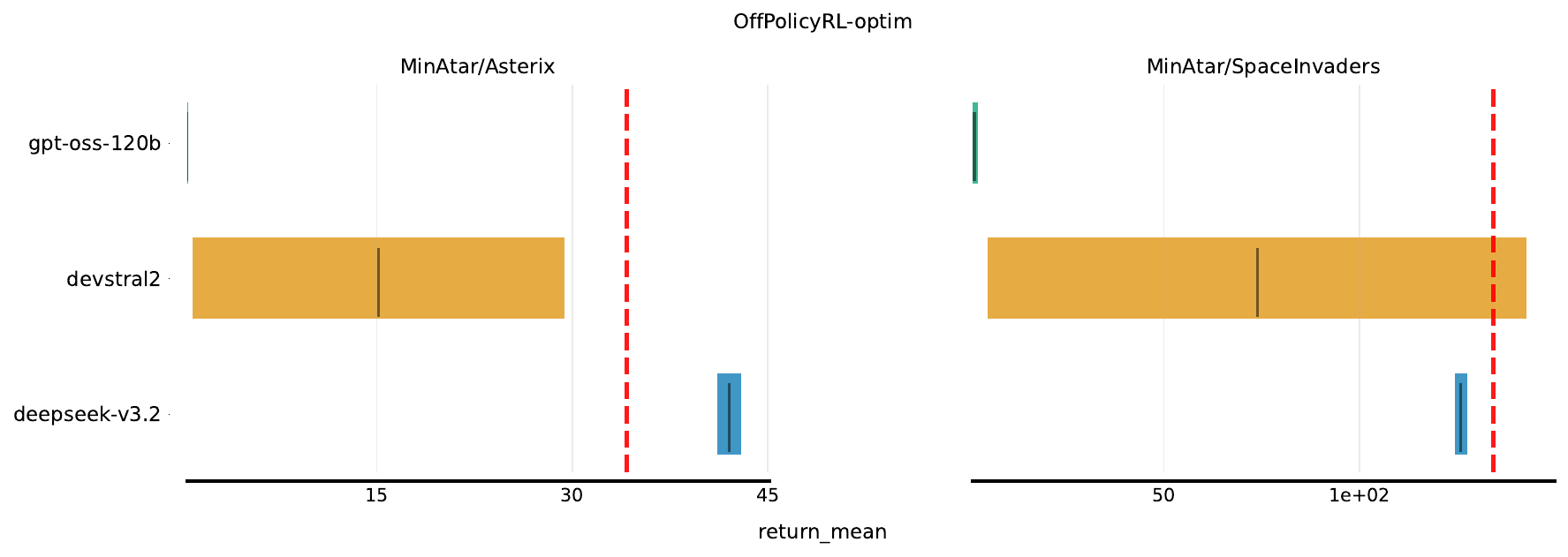}%
\hfill%
\includegraphics[width=0.48\textwidth]{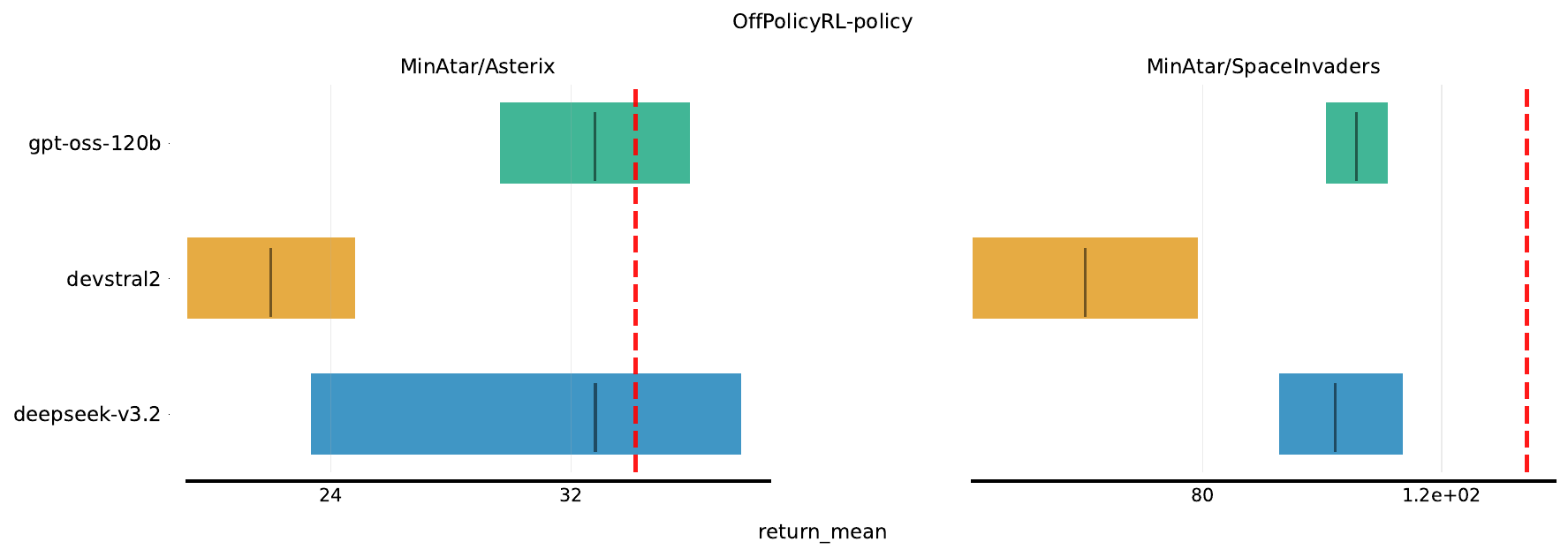}%
\\[0.5em]
\includegraphics[width=0.48\textwidth]{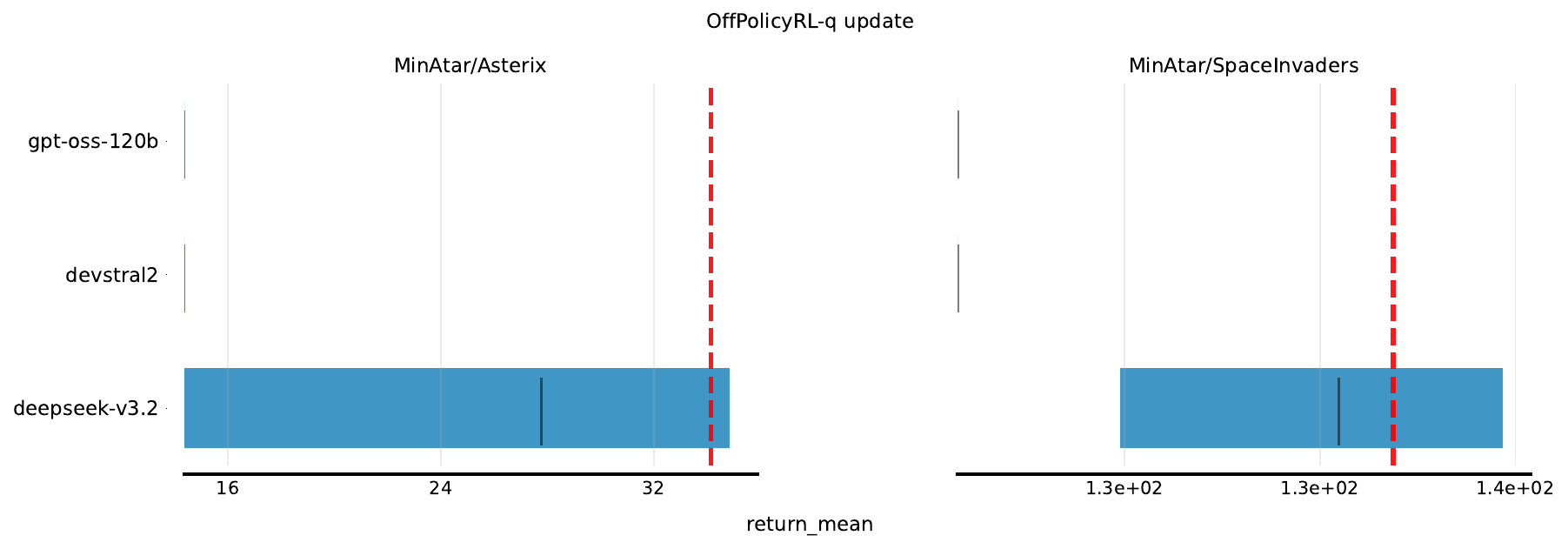}%
\hfill%
\includegraphics[width=0.48\textwidth]{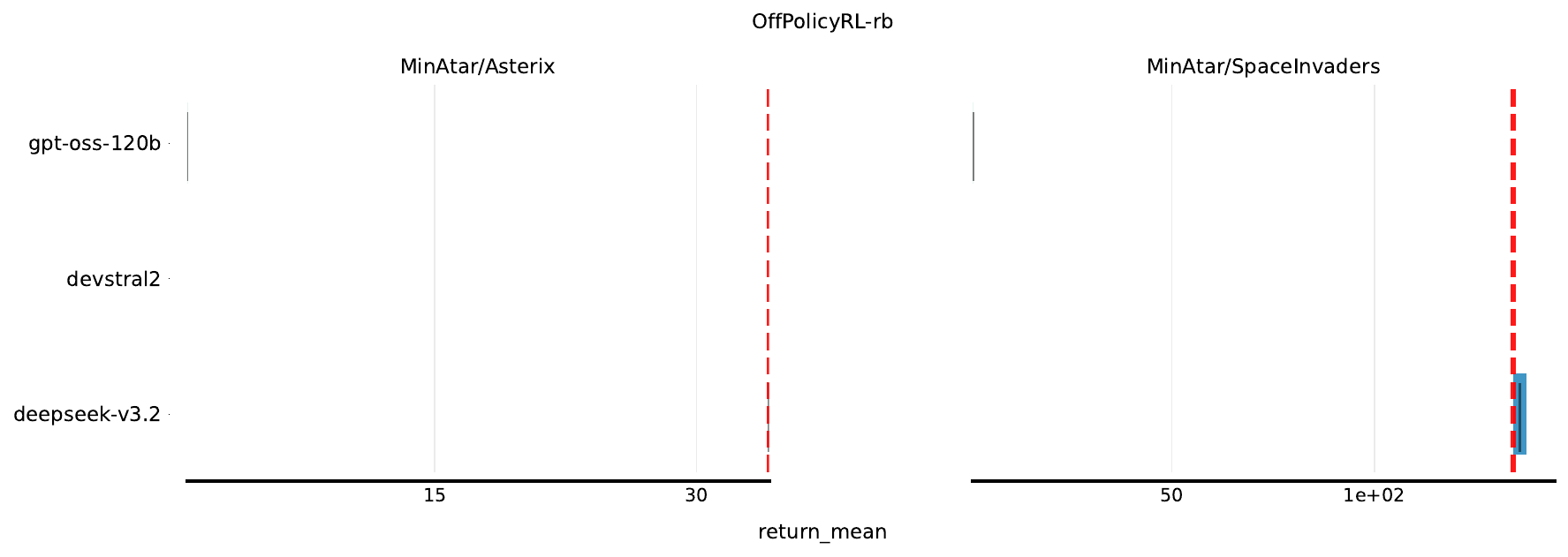}%
\\[0.5em]
\includegraphics[width=0.48\textwidth]{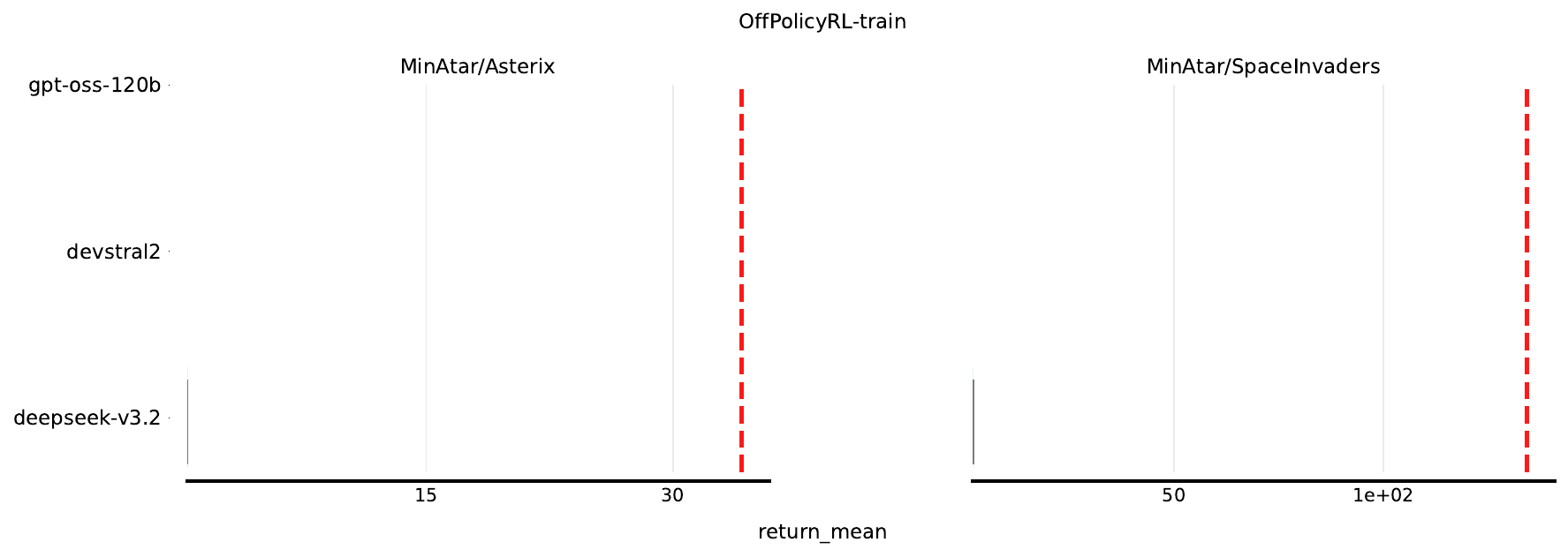}%
\hfill%
\includegraphics[width=0.48\textwidth]{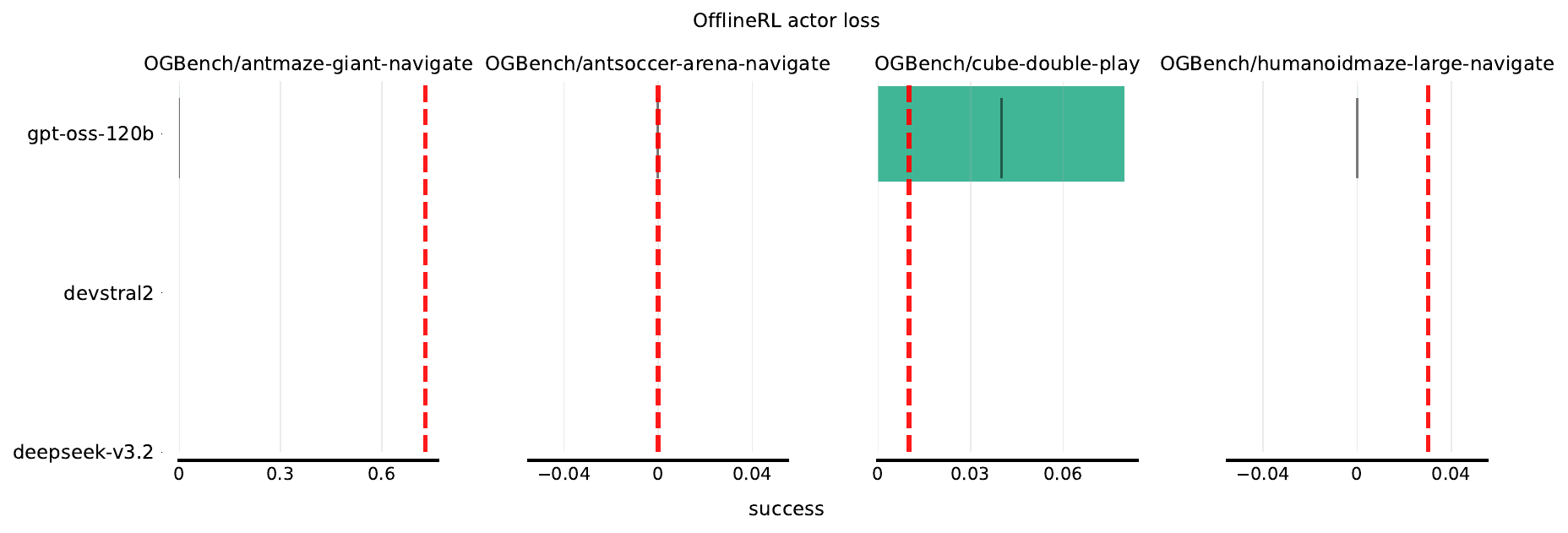}%
\\[0.5em]
\includegraphics[width=0.48\textwidth]{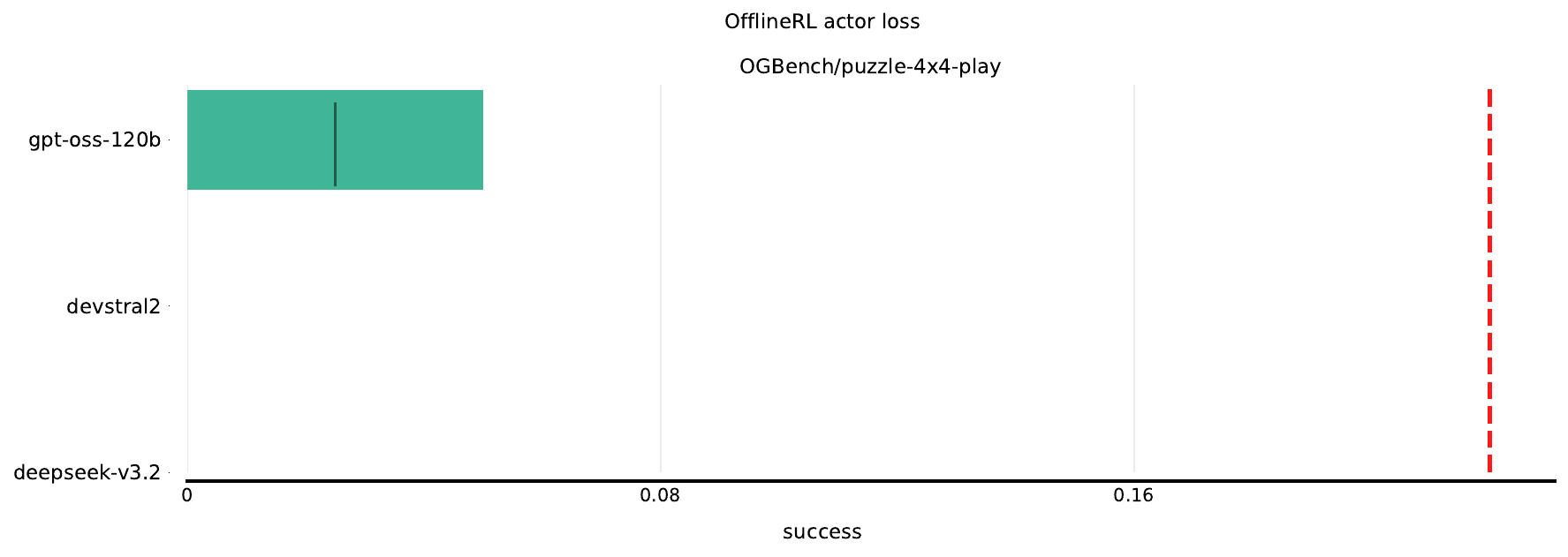}%
\hfill%
\includegraphics[width=0.48\textwidth]{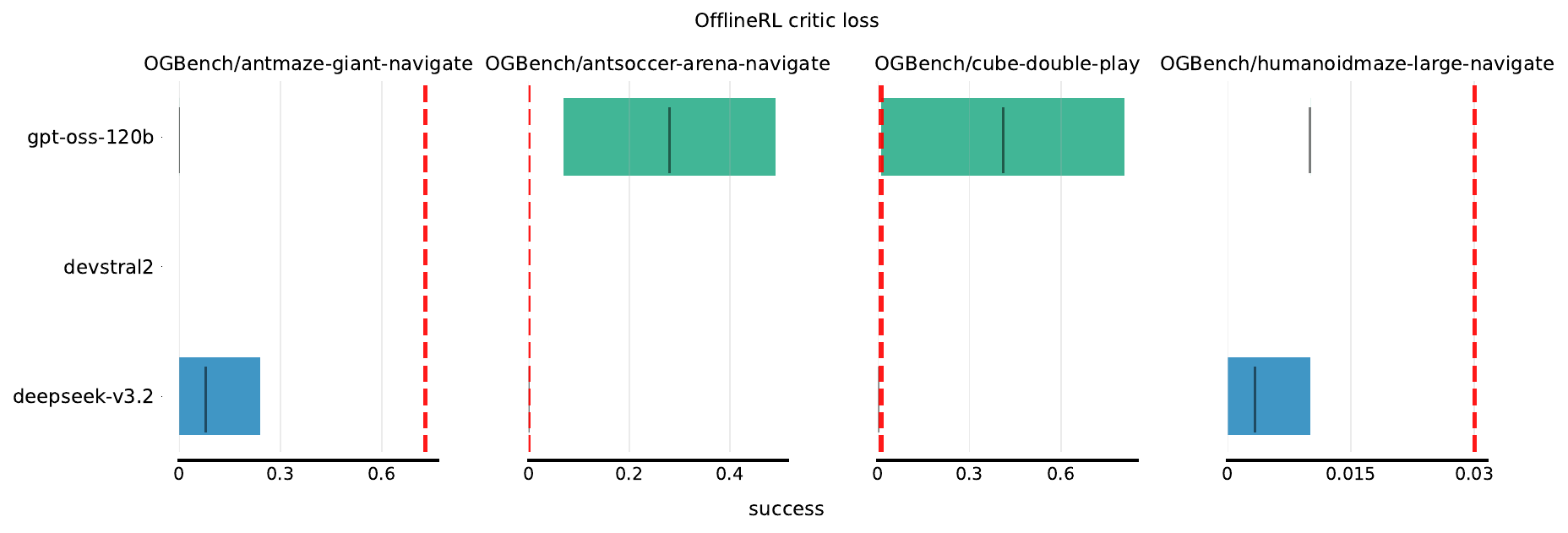}%
\caption{DiscoBench (Single Edit) results on Meta-Test tasks. (Part 4/7)}
\label{fig:one_change_mt_4}
\end{figure}
\clearpage

\begin{figure}[htbp]
\centering
\setlength{\lineskip}{0pt}
\includegraphics[width=0.48\textwidth]{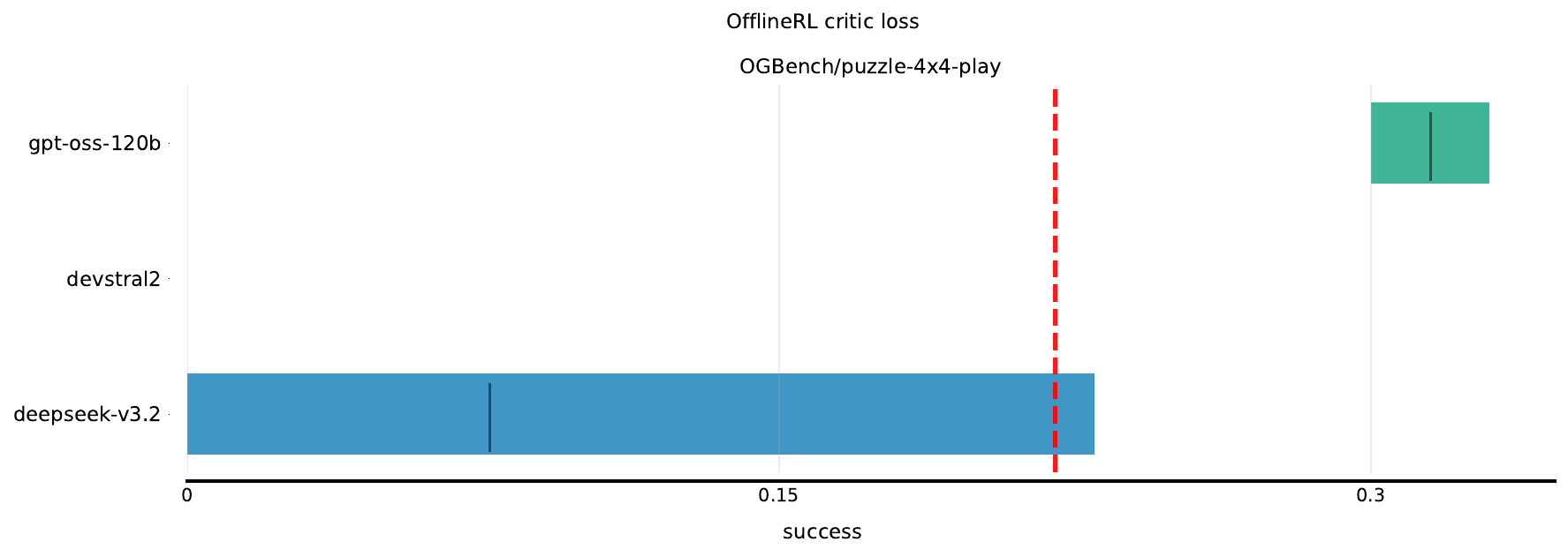}%
\hfill%
\includegraphics[width=0.48\textwidth]{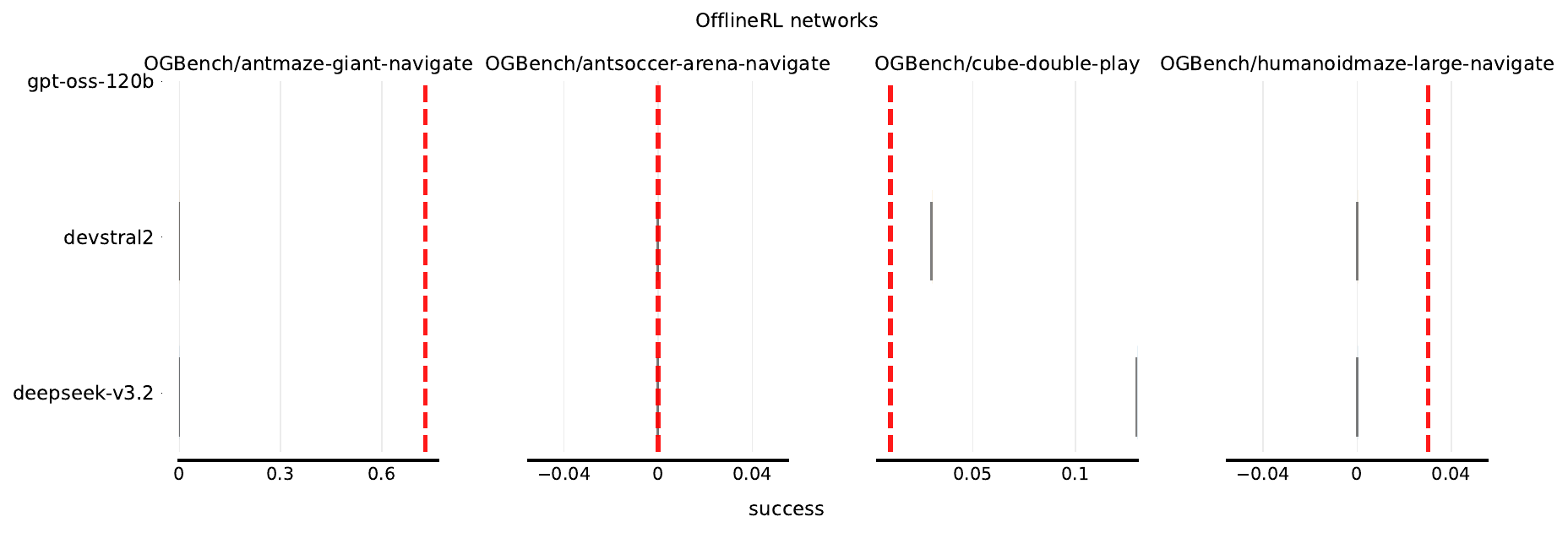}%
\\[0.5em]
\includegraphics[width=0.48\textwidth]{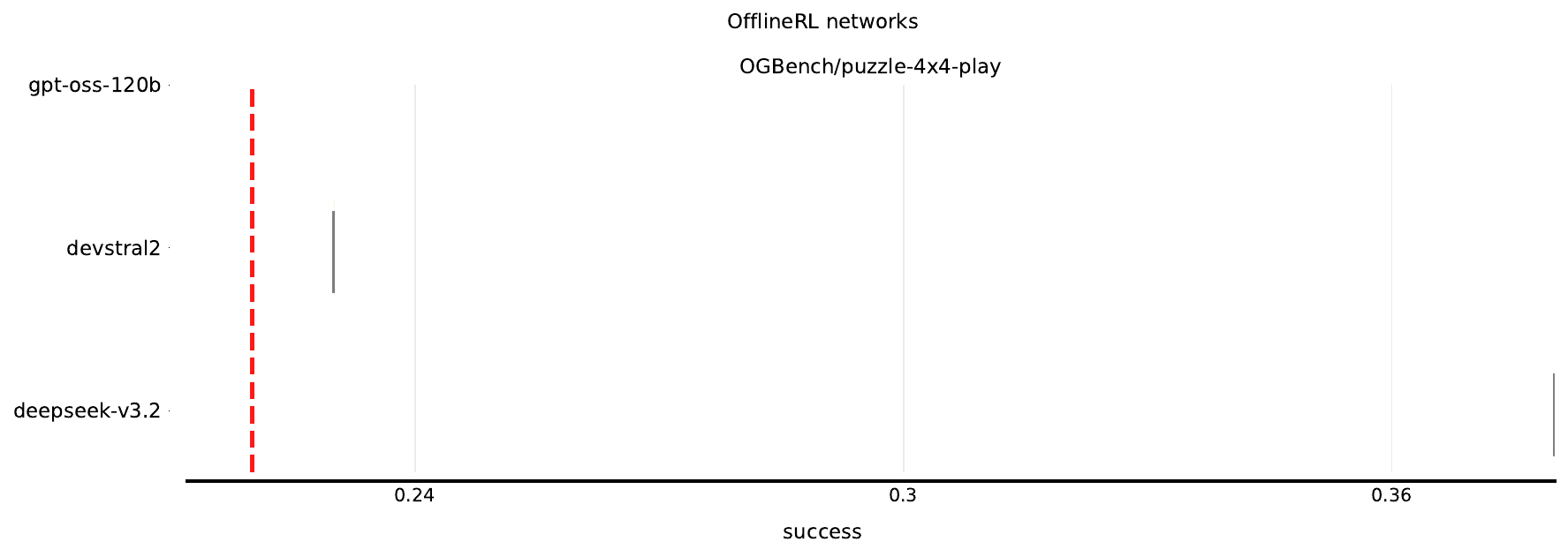}%
\hfill%
\includegraphics[width=0.48\textwidth]{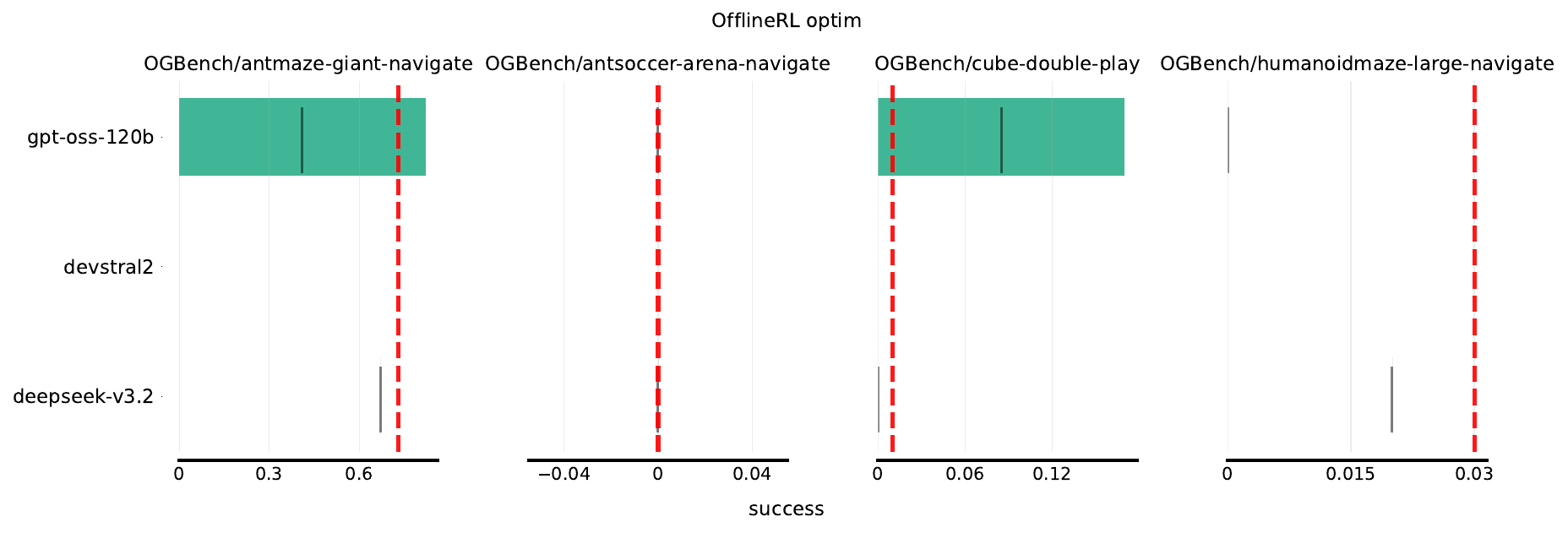}%
\\[0.5em]
\includegraphics[width=0.48\textwidth]{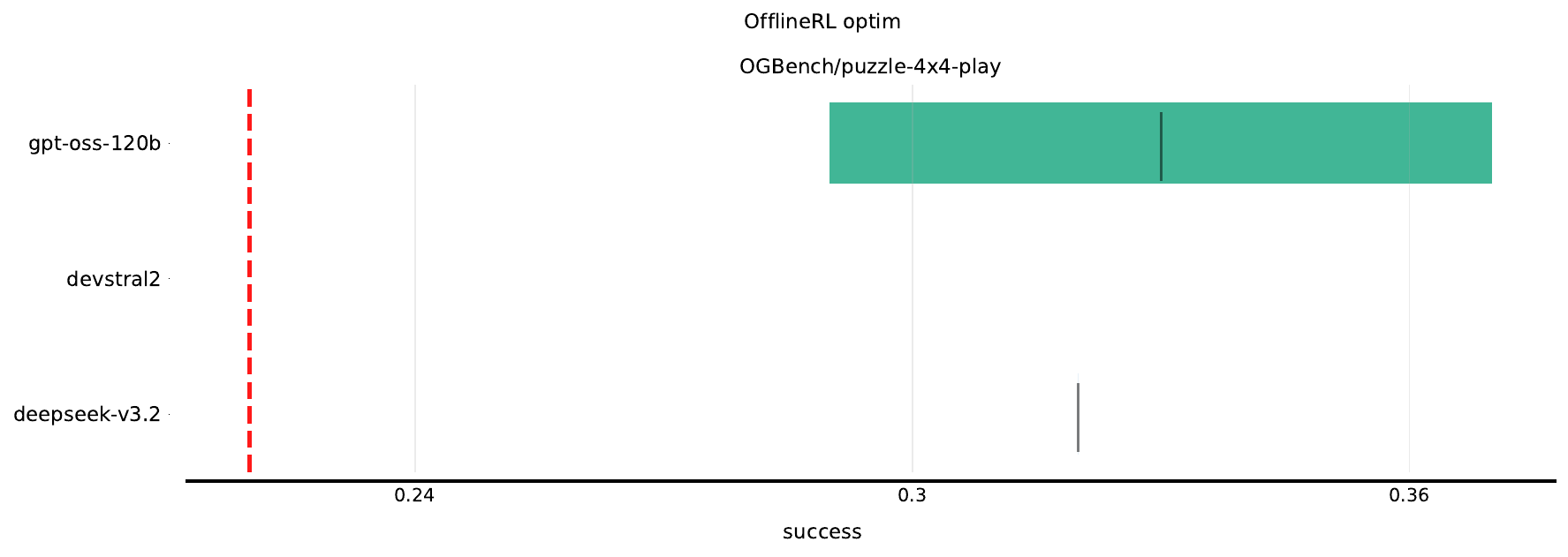}%
\hfill%
\includegraphics[width=0.48\textwidth]{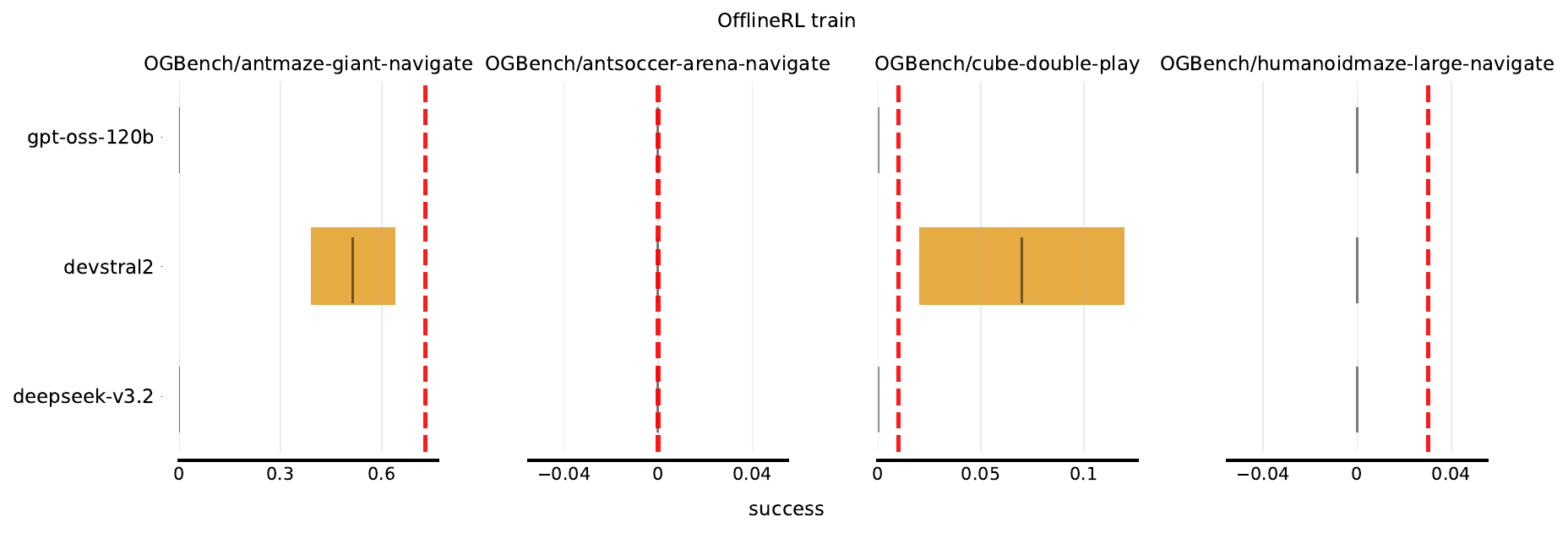}%
\\[0.5em]
\includegraphics[width=0.48\textwidth]{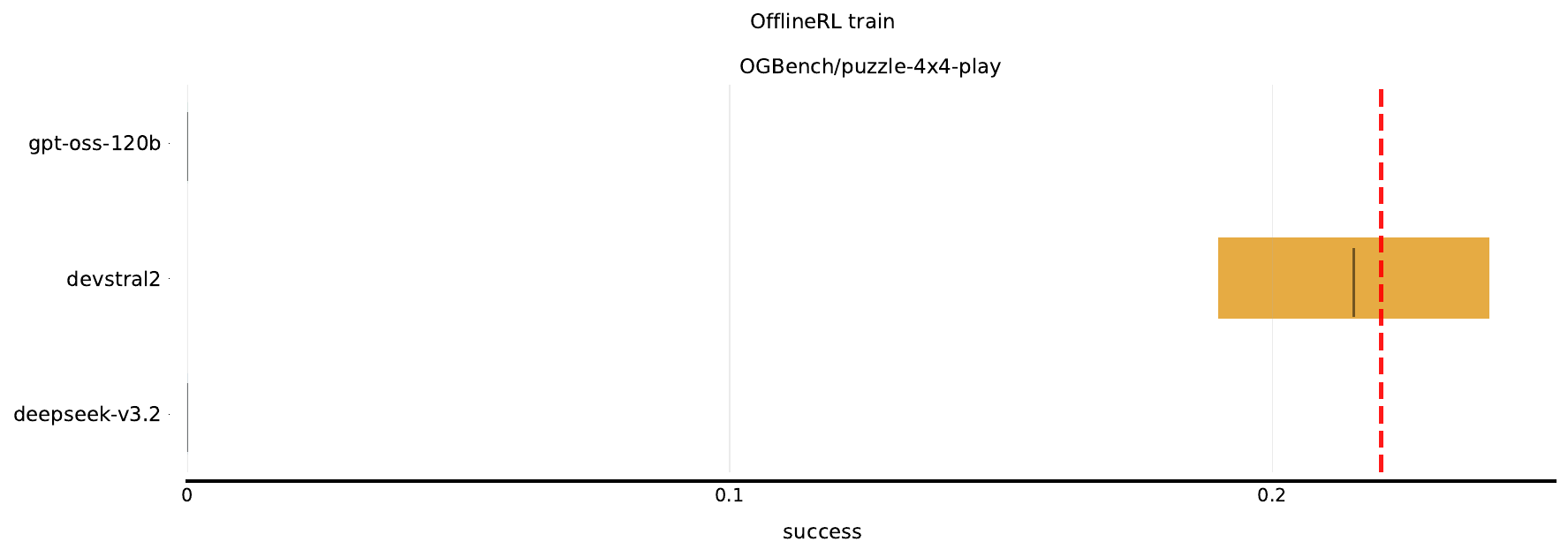}%
\hfill%
\includegraphics[width=0.48\textwidth]{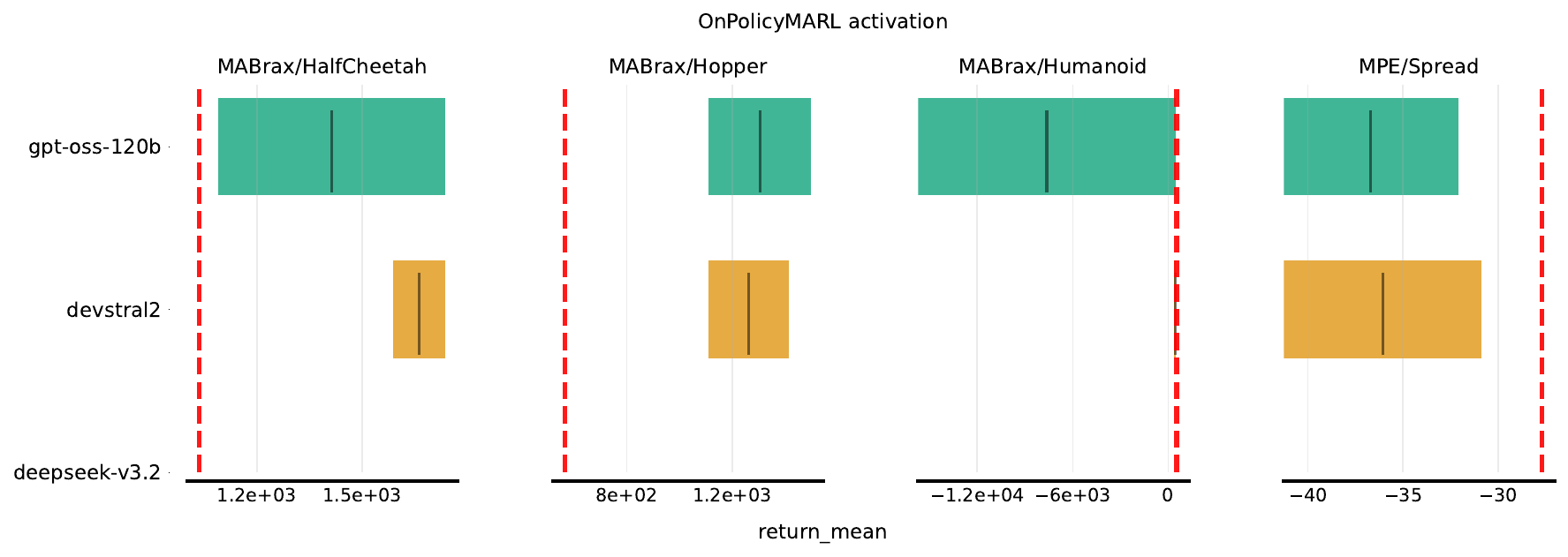}%
\\[0.5em]
\includegraphics[width=0.48\textwidth]{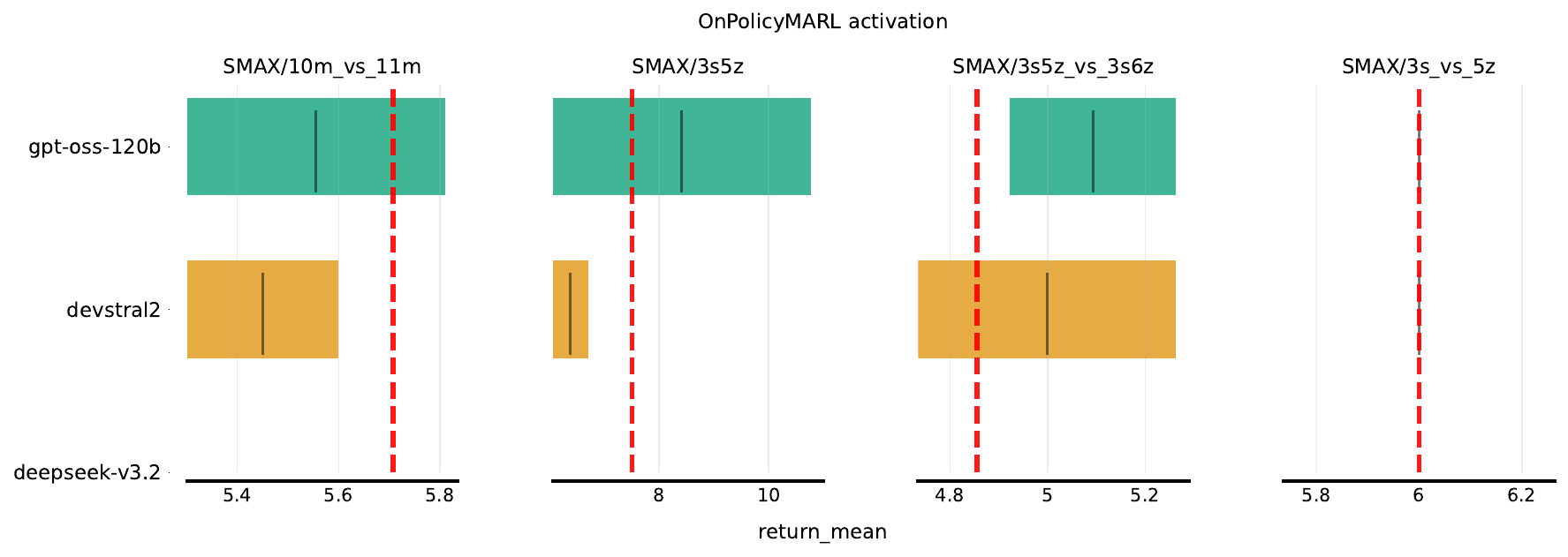}%
\hfill%
\includegraphics[width=0.48\textwidth]{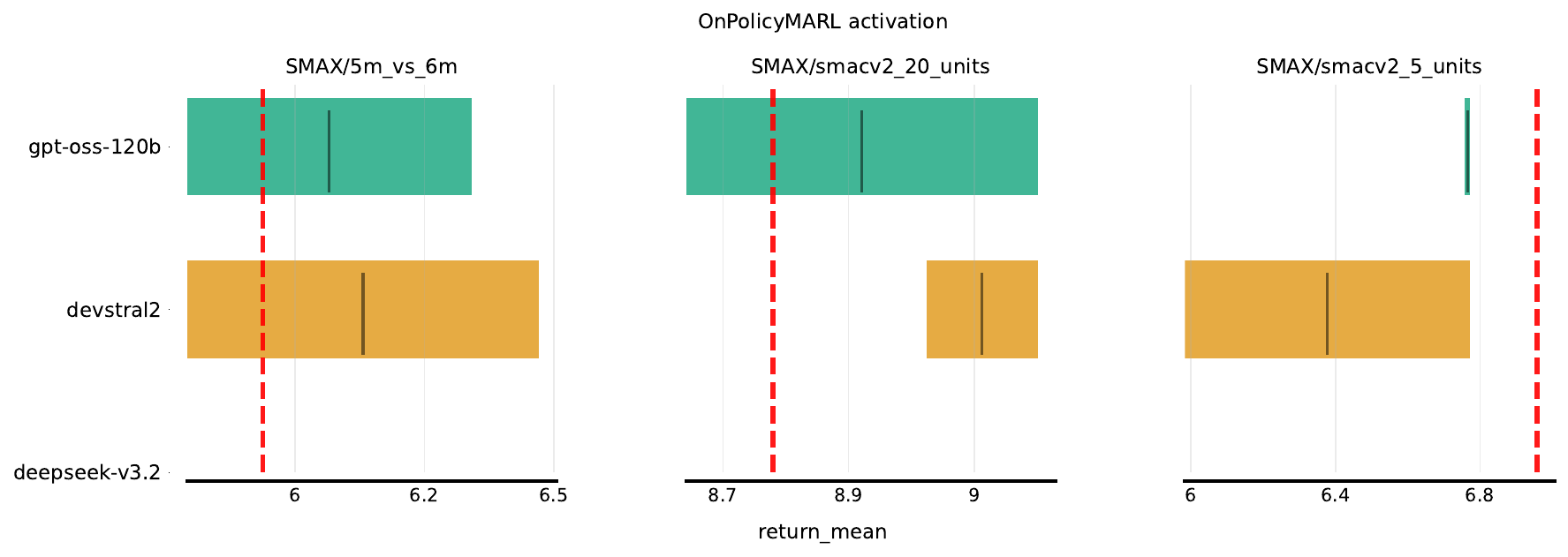}%
\\[0.5em]
\includegraphics[width=0.48\textwidth]{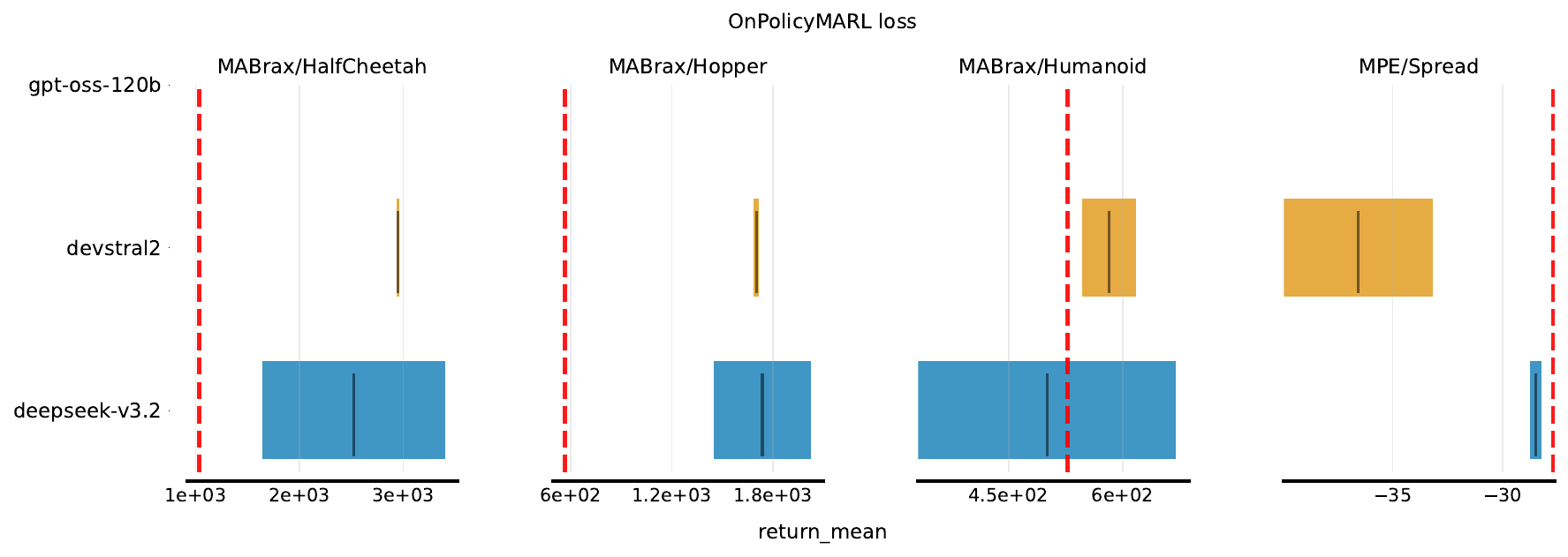}%
\hfill%
\includegraphics[width=0.48\textwidth]{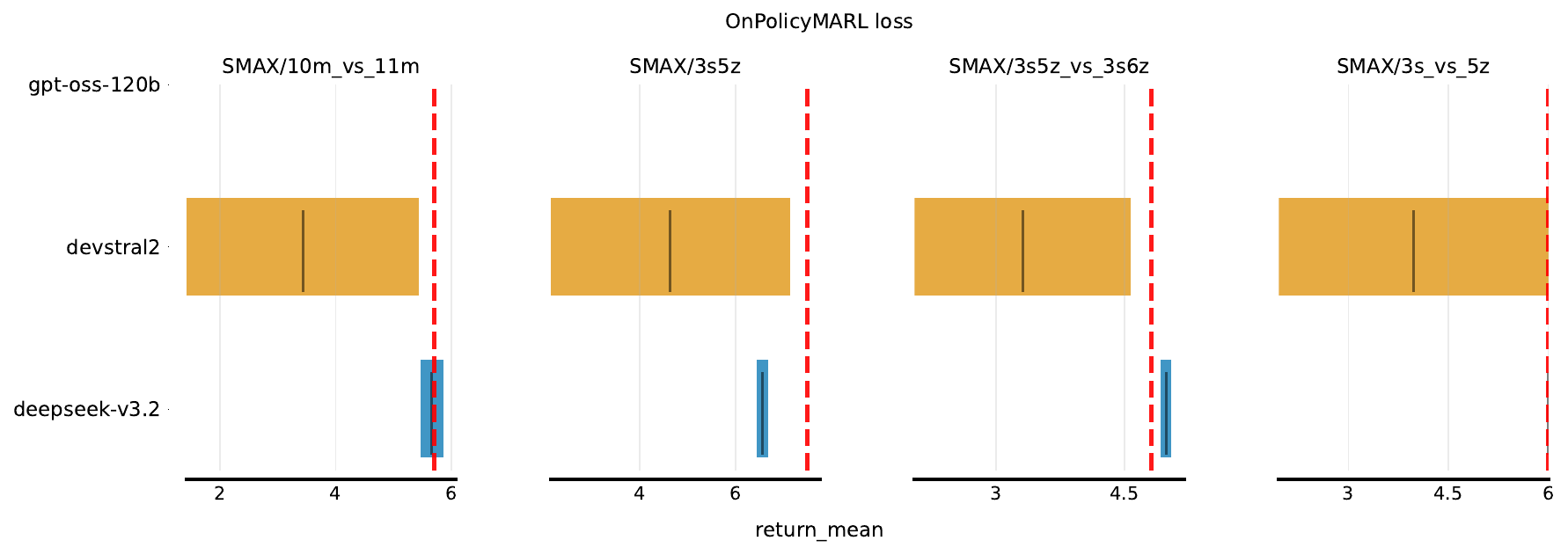}%
\caption{DiscoBench (Single Edit) results on Meta-Test tasks. (Part 5/7)}
\label{fig:one_change_mt_5}
\end{figure}
\clearpage

\begin{figure}[htbp]
\centering
\setlength{\lineskip}{0pt}
\includegraphics[width=0.48\textwidth]{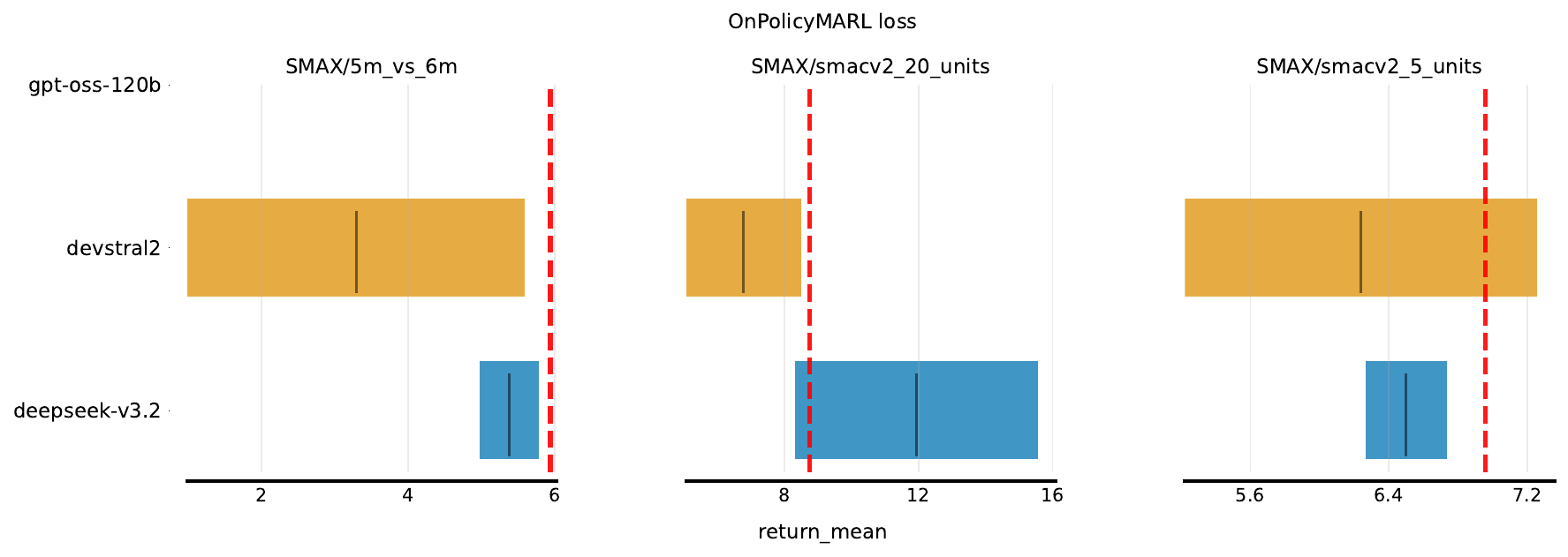}%
\hfill%
\includegraphics[width=0.48\textwidth]{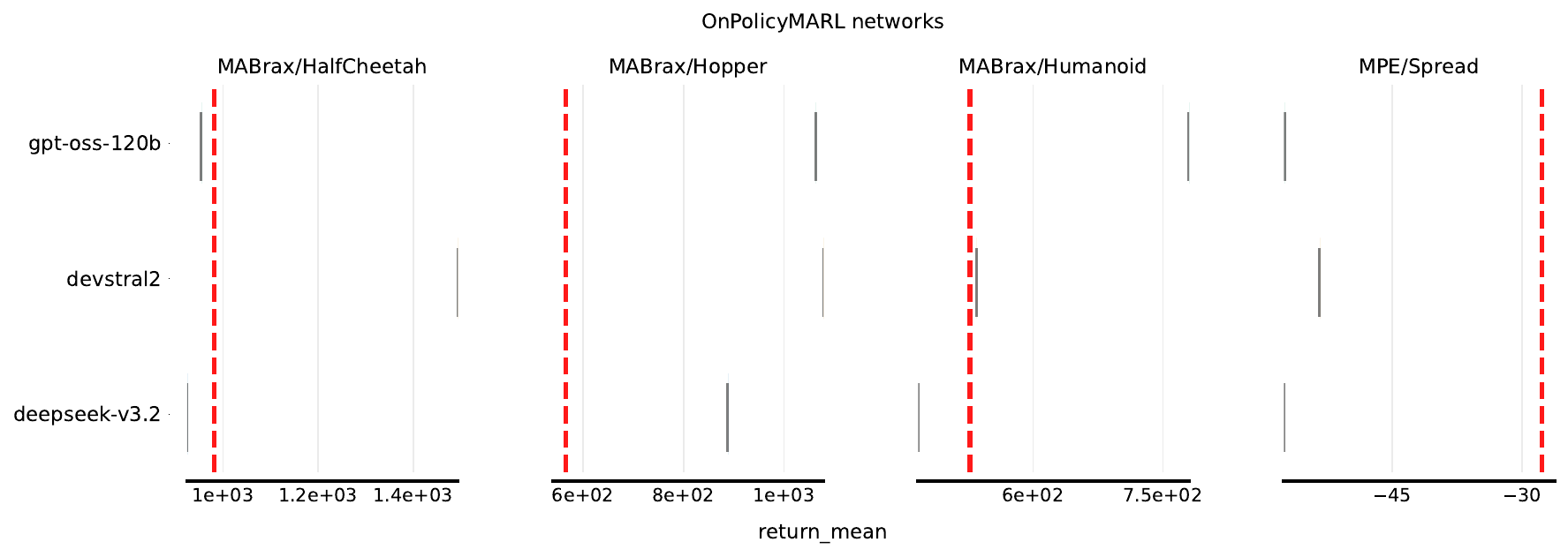}%
\\[0.5em]
\includegraphics[width=0.48\textwidth]{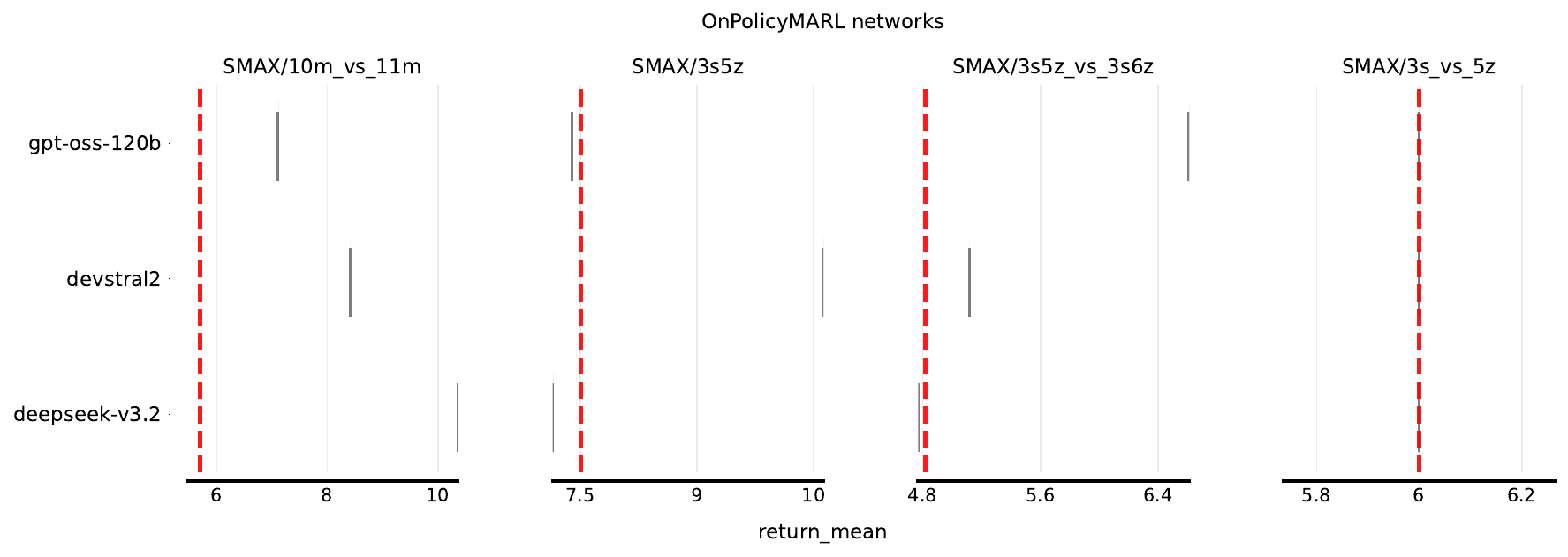}%
\hfill%
\includegraphics[width=0.48\textwidth]{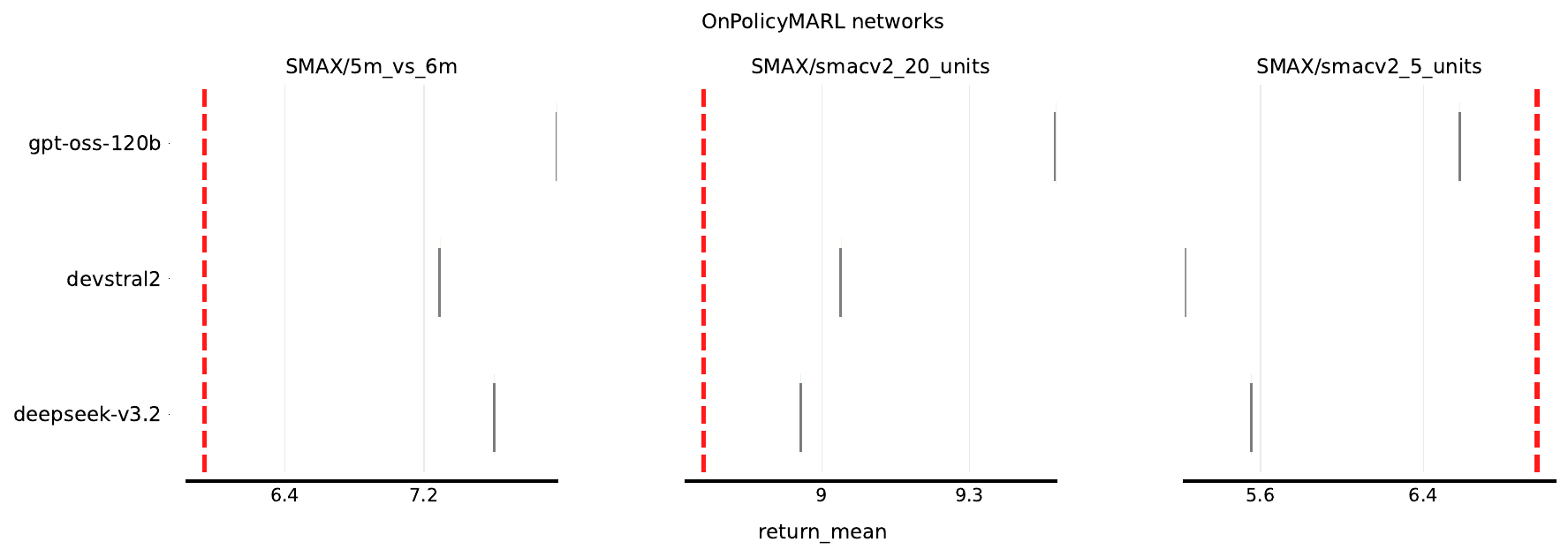}%
\\[0.5em]
\includegraphics[width=0.48\textwidth]{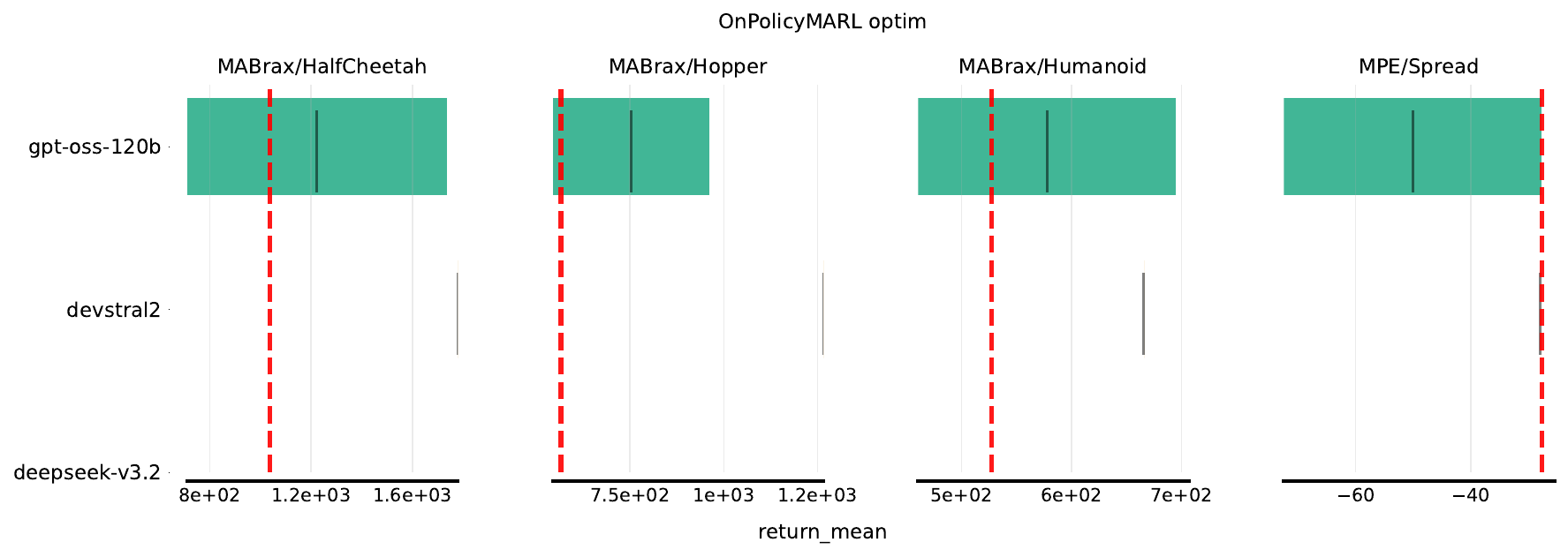}%
\hfill%
\includegraphics[width=0.48\textwidth]{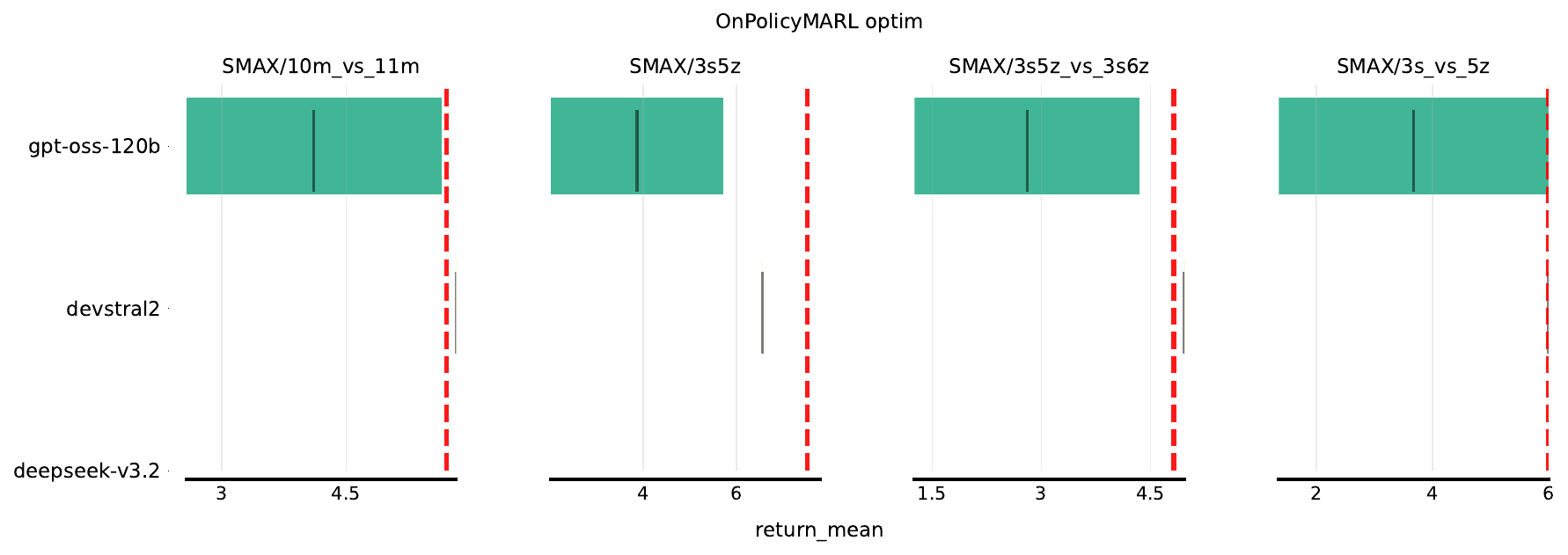}%
\\[0.5em]
\includegraphics[width=0.48\textwidth]{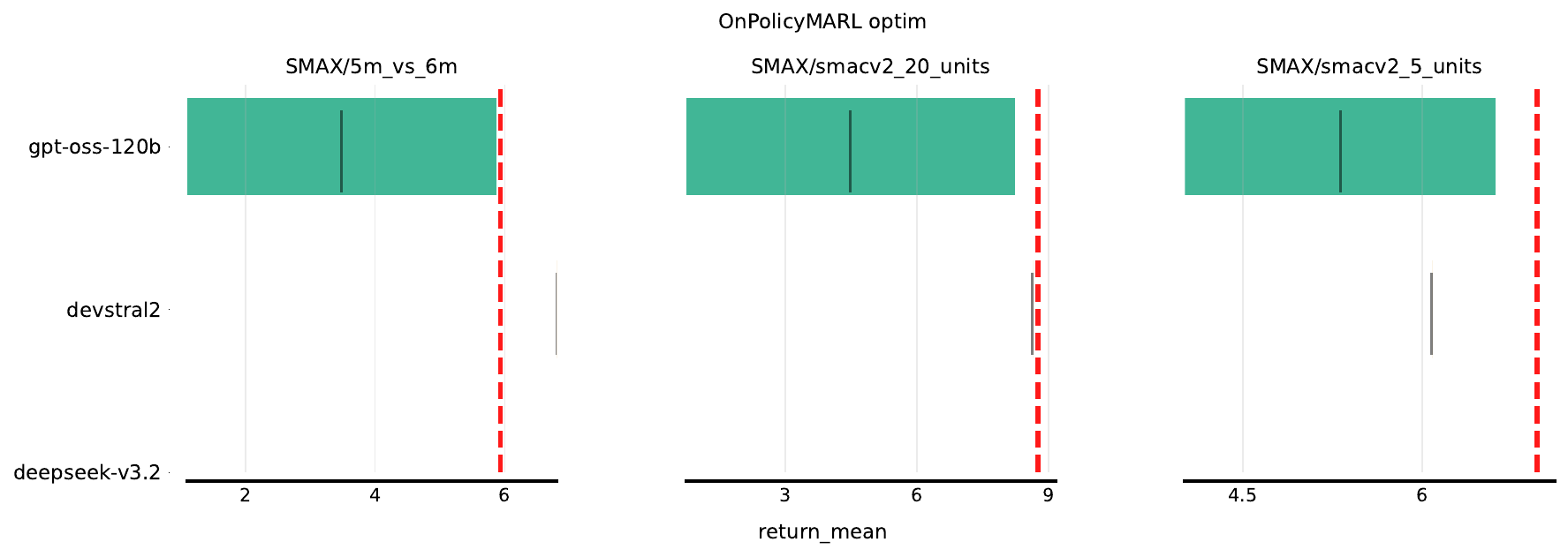}%
\hfill%
\includegraphics[width=0.48\textwidth]{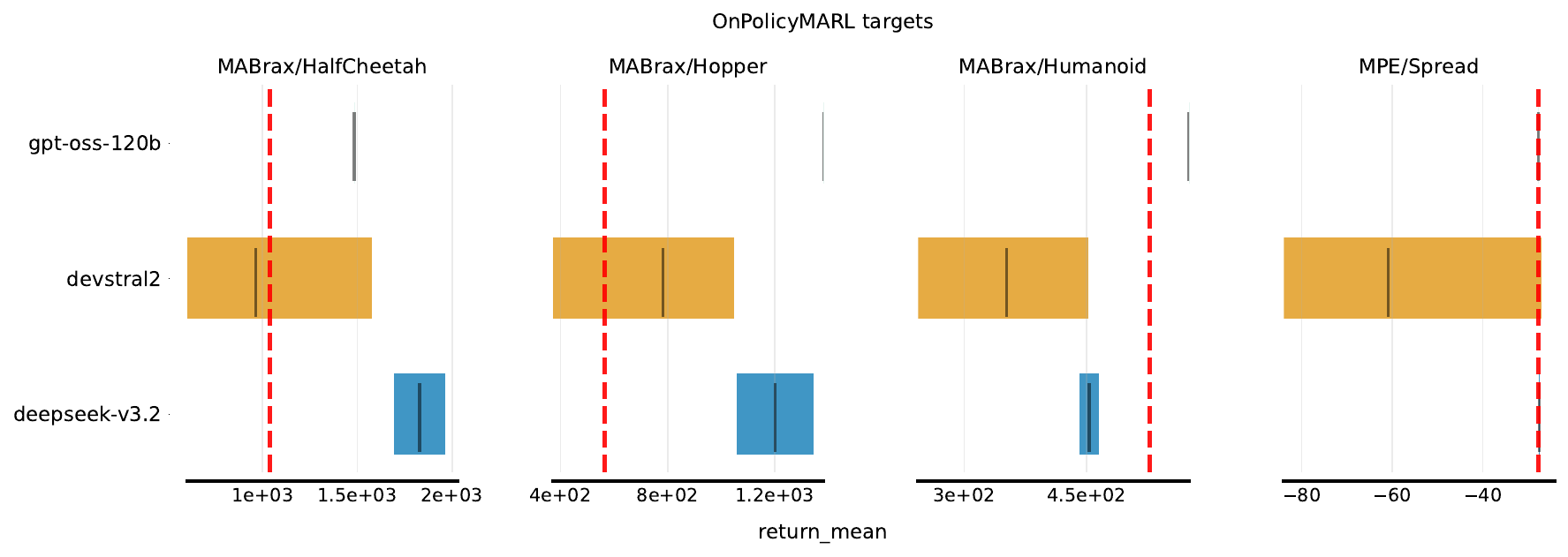}%
\\[0.5em]
\includegraphics[width=0.48\textwidth]{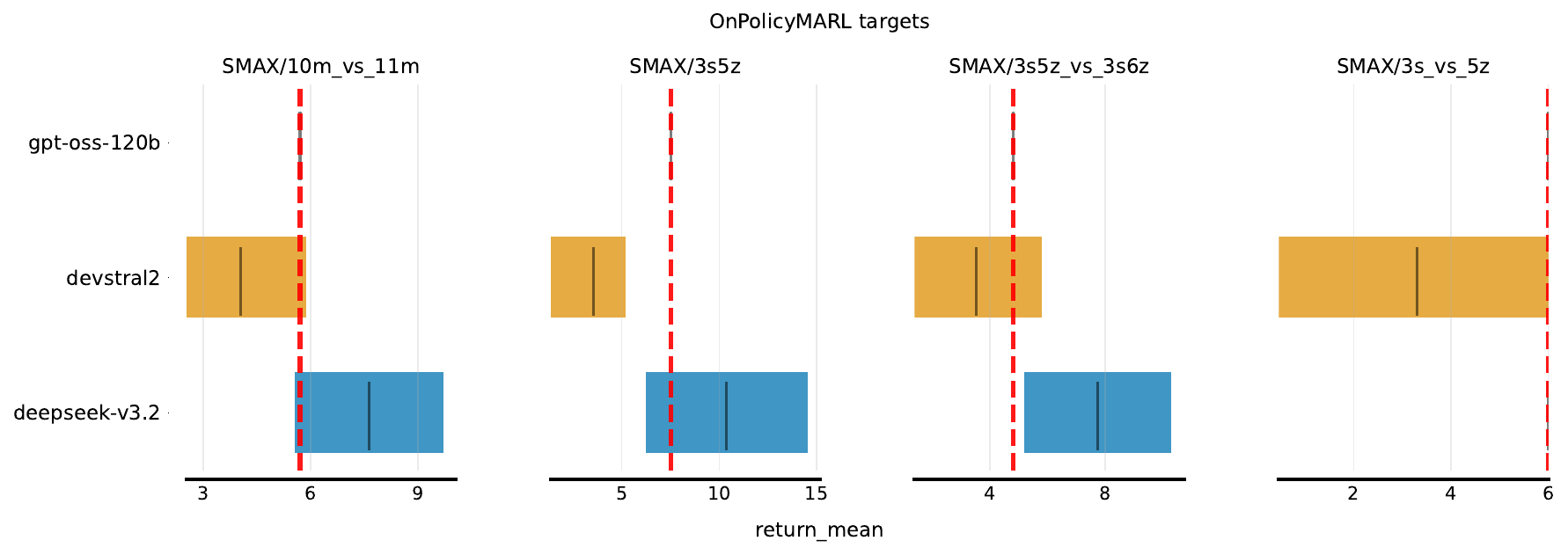}%
\hfill%
\includegraphics[width=0.48\textwidth]{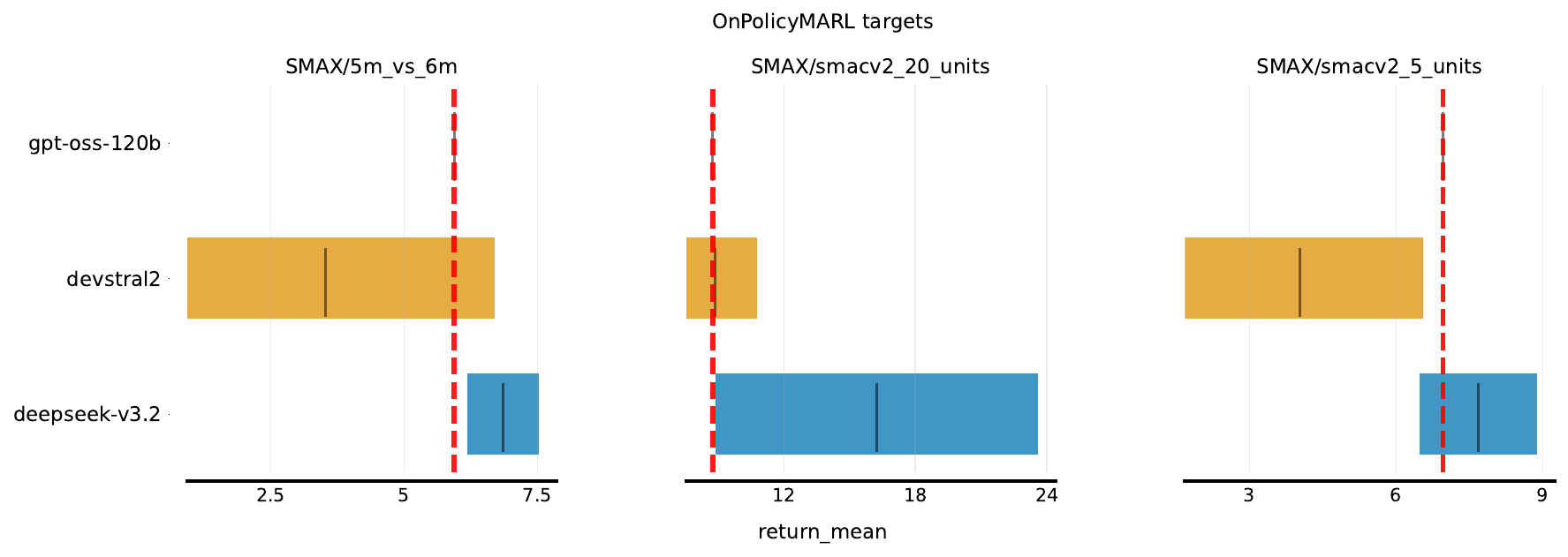}%
\\[0.5em]
\includegraphics[width=0.48\textwidth]{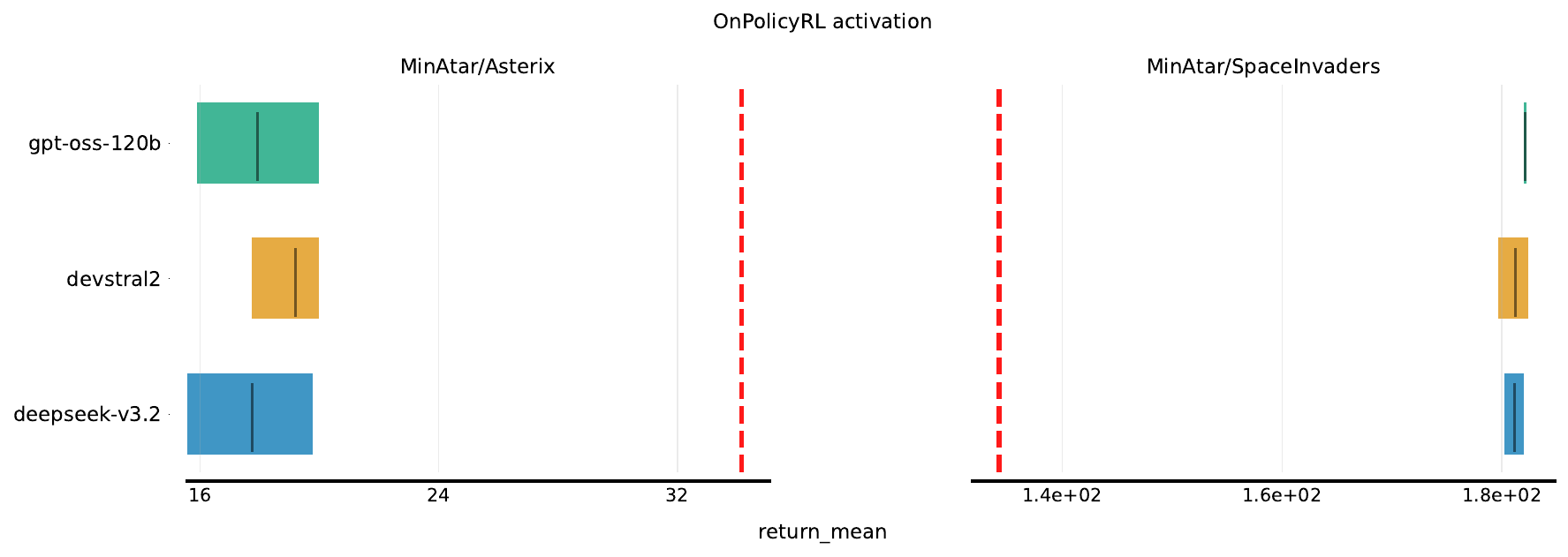}%
\hfill%
\includegraphics[width=0.48\textwidth]{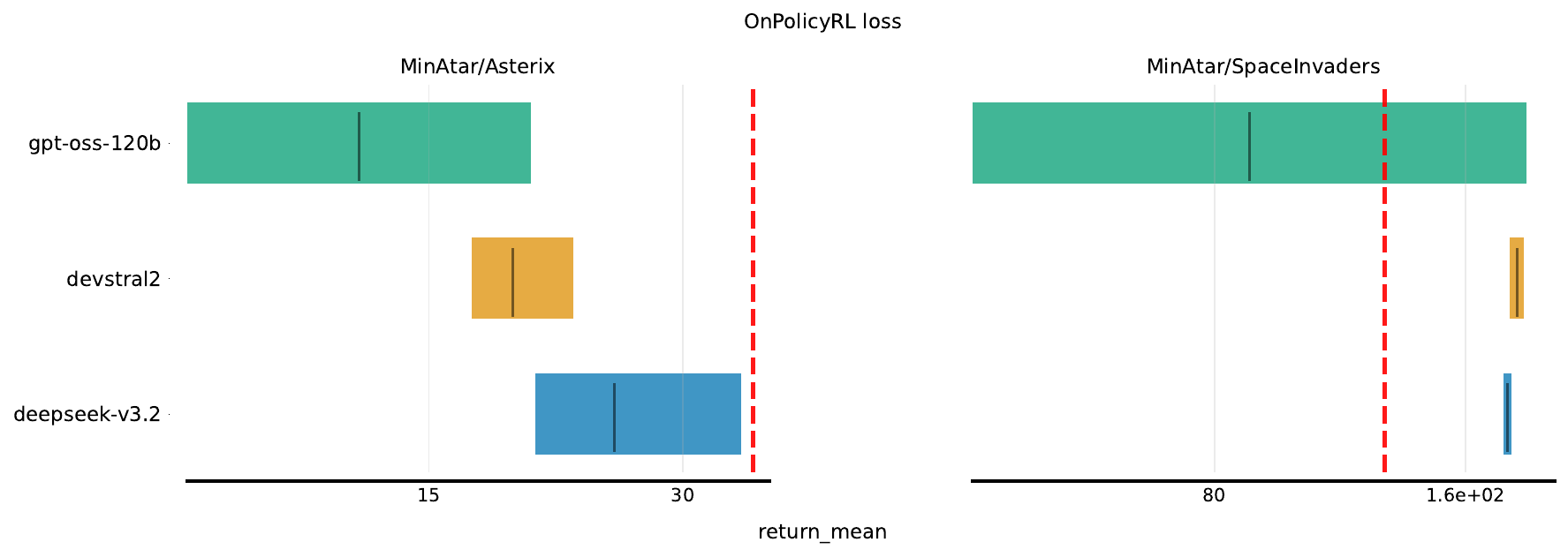}%
\caption{DiscoBench (Single Edit) results on Meta-Test tasks. (Part 6/7)}
\label{fig:one_change_mt_6}
\end{figure}
\clearpage

\begin{figure}[htbp]
\centering
\setlength{\lineskip}{0pt}
\includegraphics[width=0.48\textwidth]{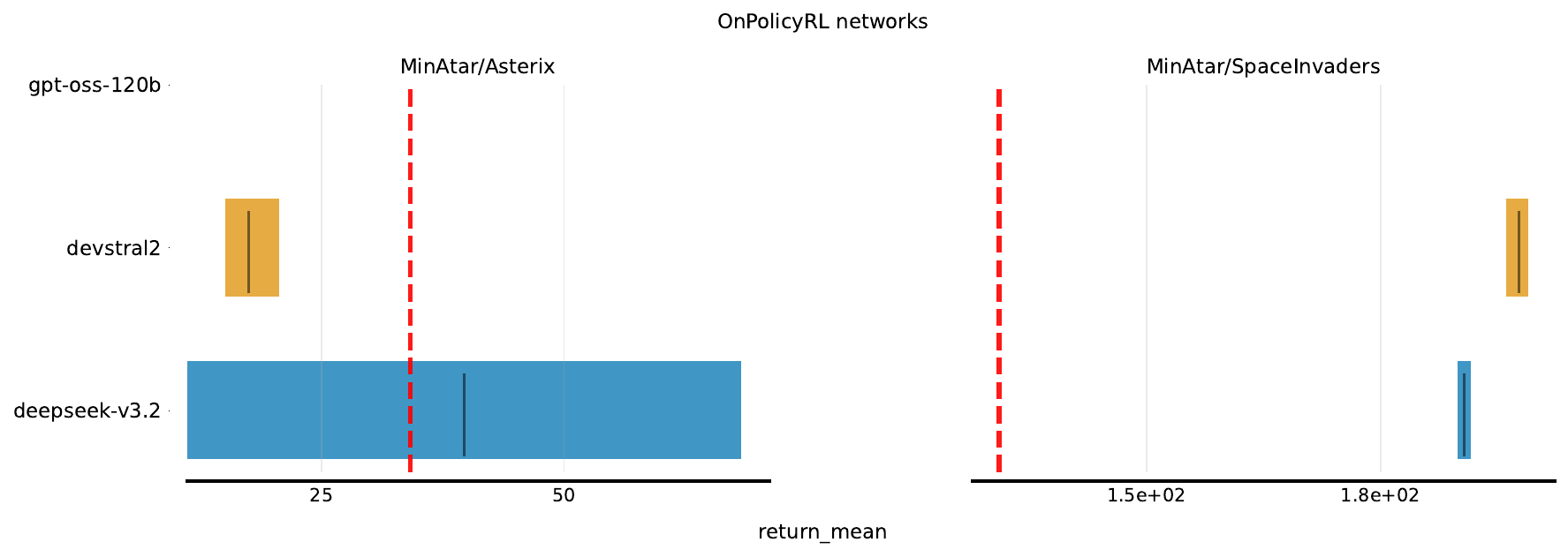}%
\hfill%
\includegraphics[width=0.48\textwidth]{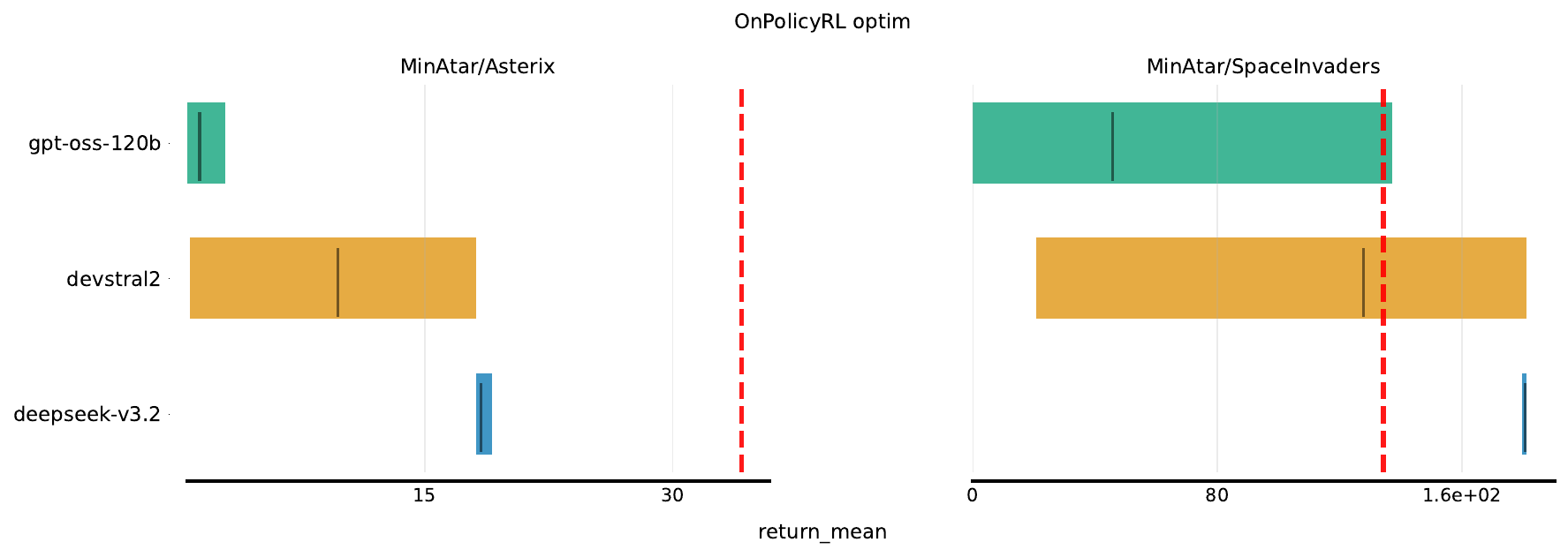}%
\\[0.5em]
\includegraphics[width=0.48\textwidth]{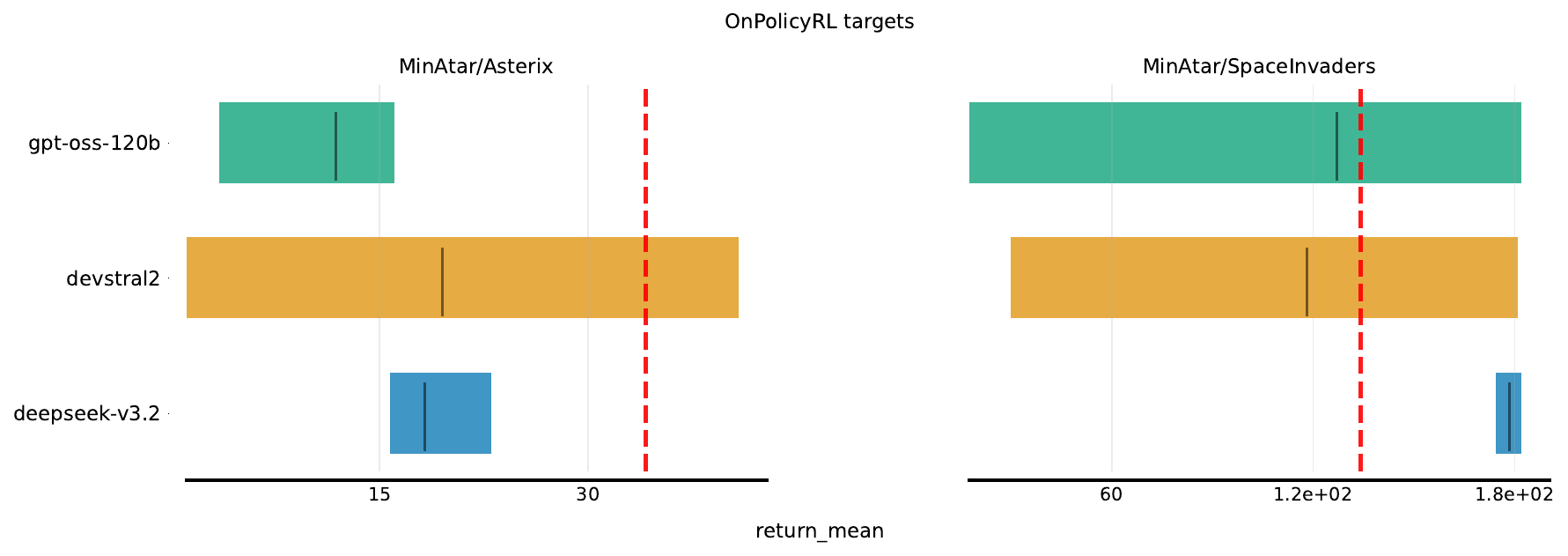}%
\hfill%
\includegraphics[width=0.48\textwidth]{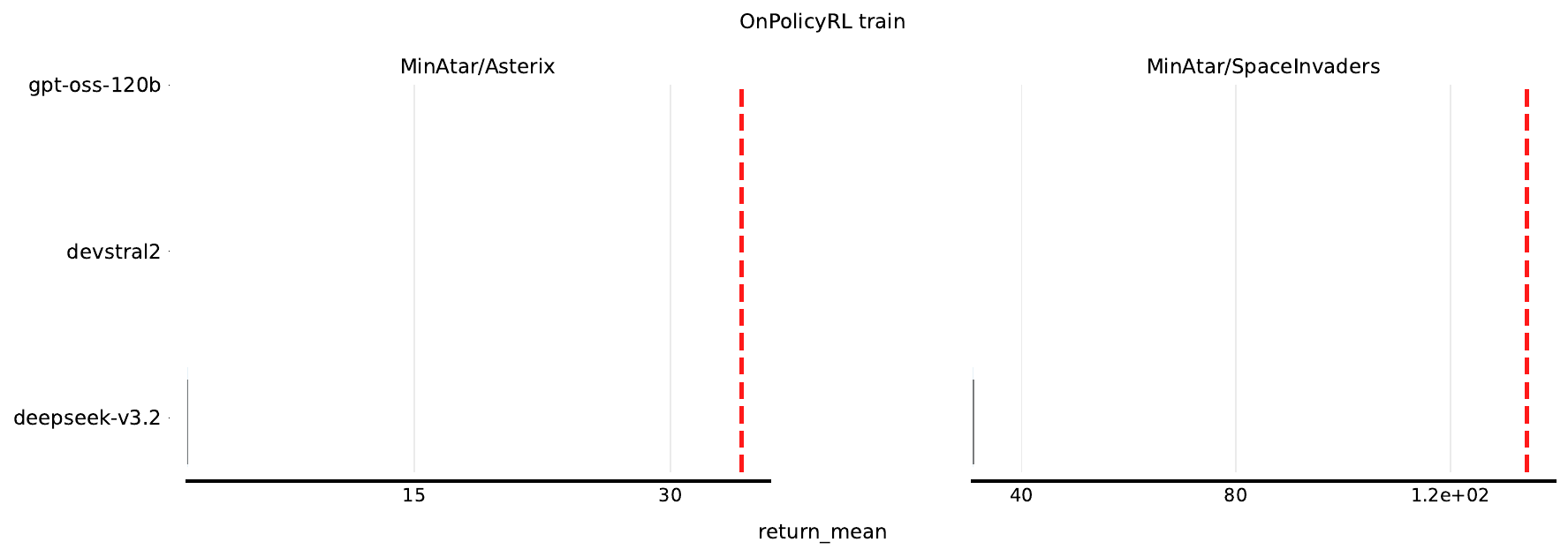}%
\\[0.5em]
\includegraphics[width=0.48\textwidth]{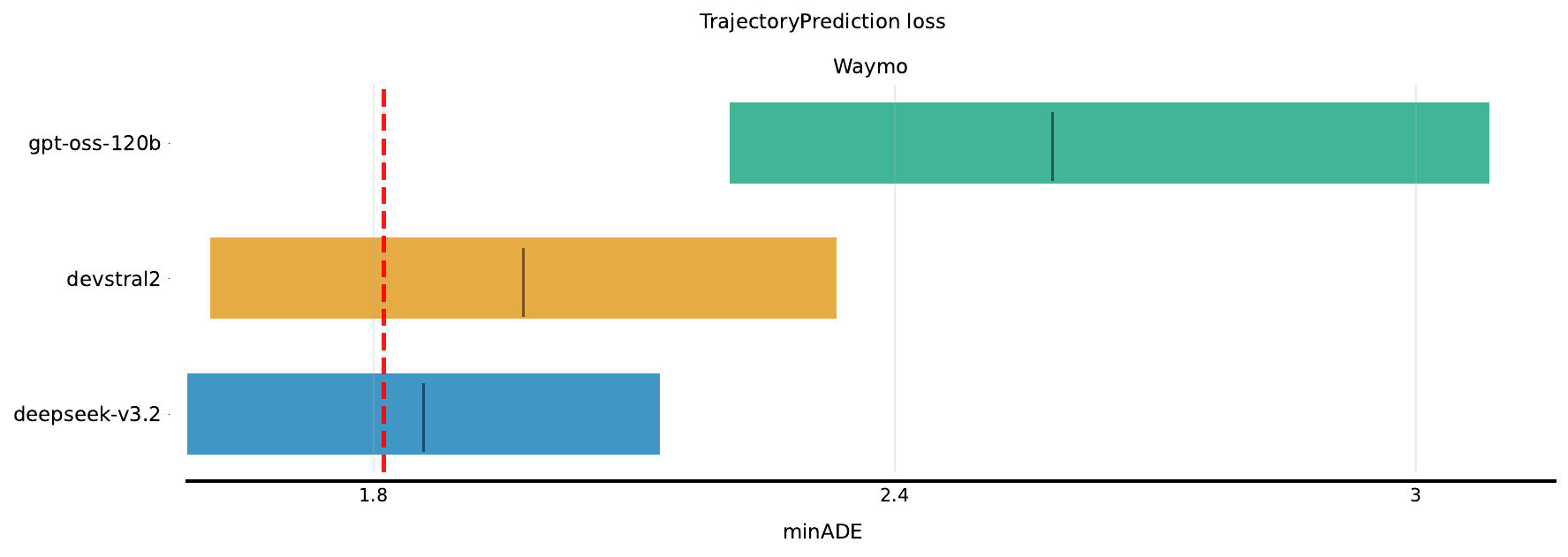}%
\hfill%
\includegraphics[width=0.48\textwidth]{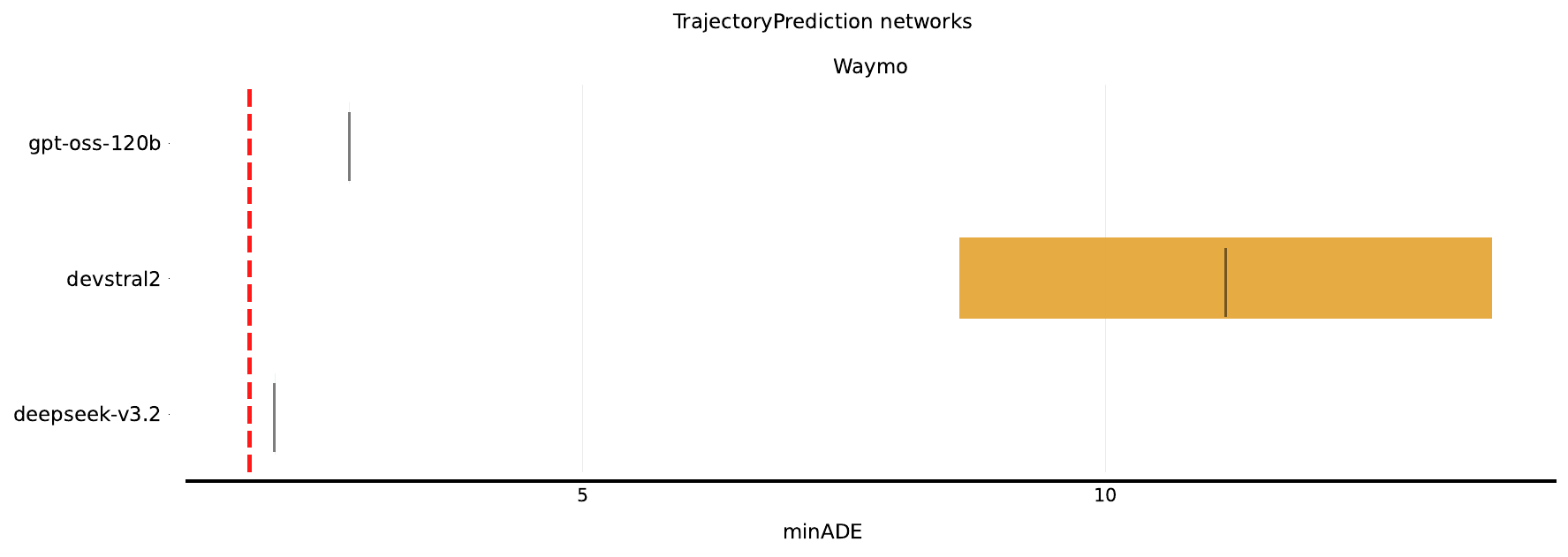}%
\\[0.5em]
\includegraphics[width=0.48\textwidth]{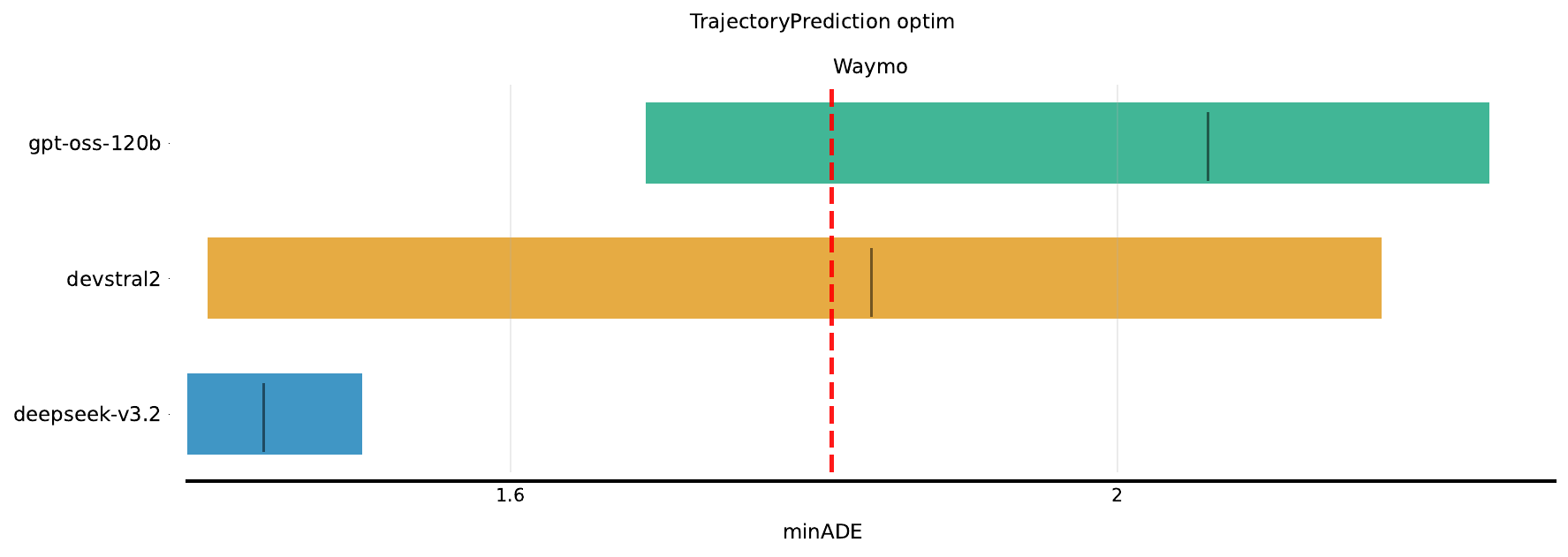}%
\hfill%
\includegraphics[width=0.48\textwidth]{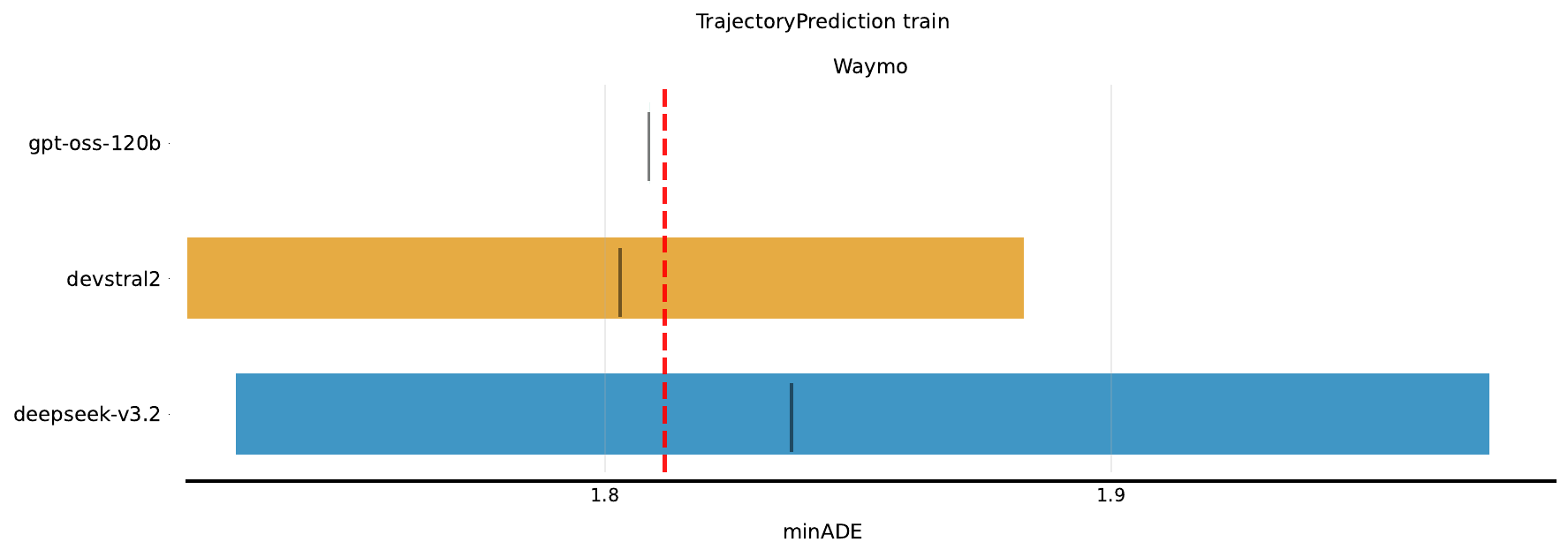}%
\\[0.5em]
\includegraphics[width=0.48\textwidth]{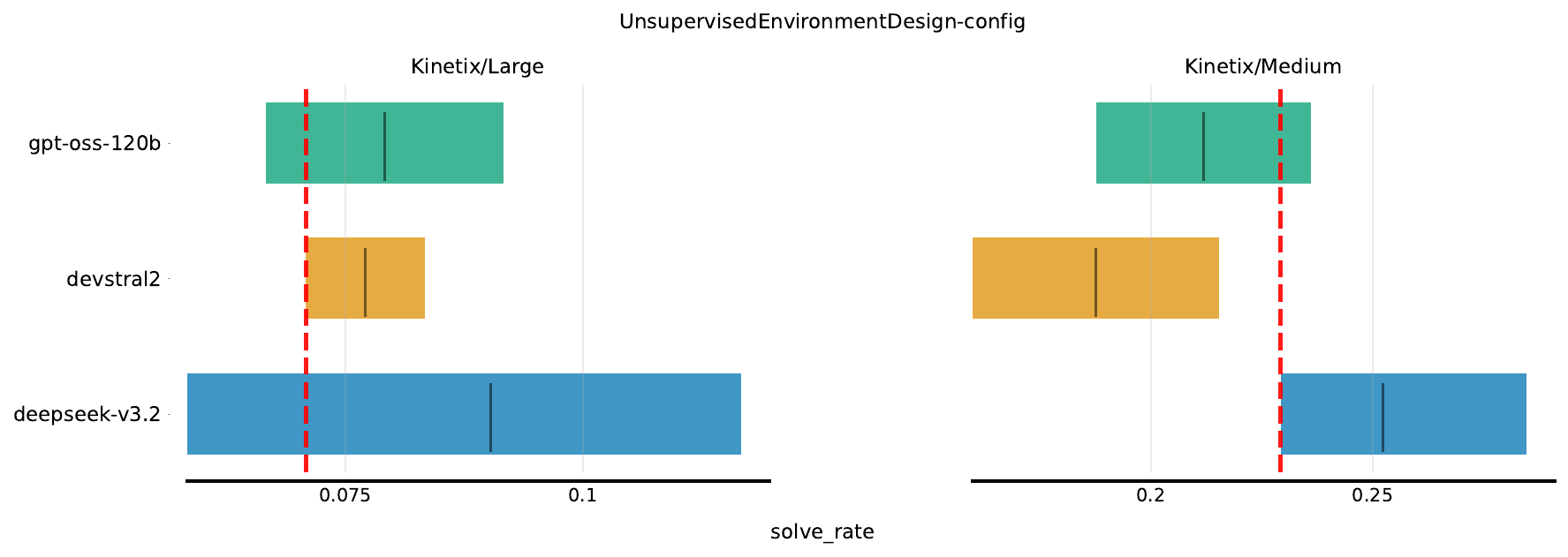}%
\hfill%
\includegraphics[width=0.48\textwidth]{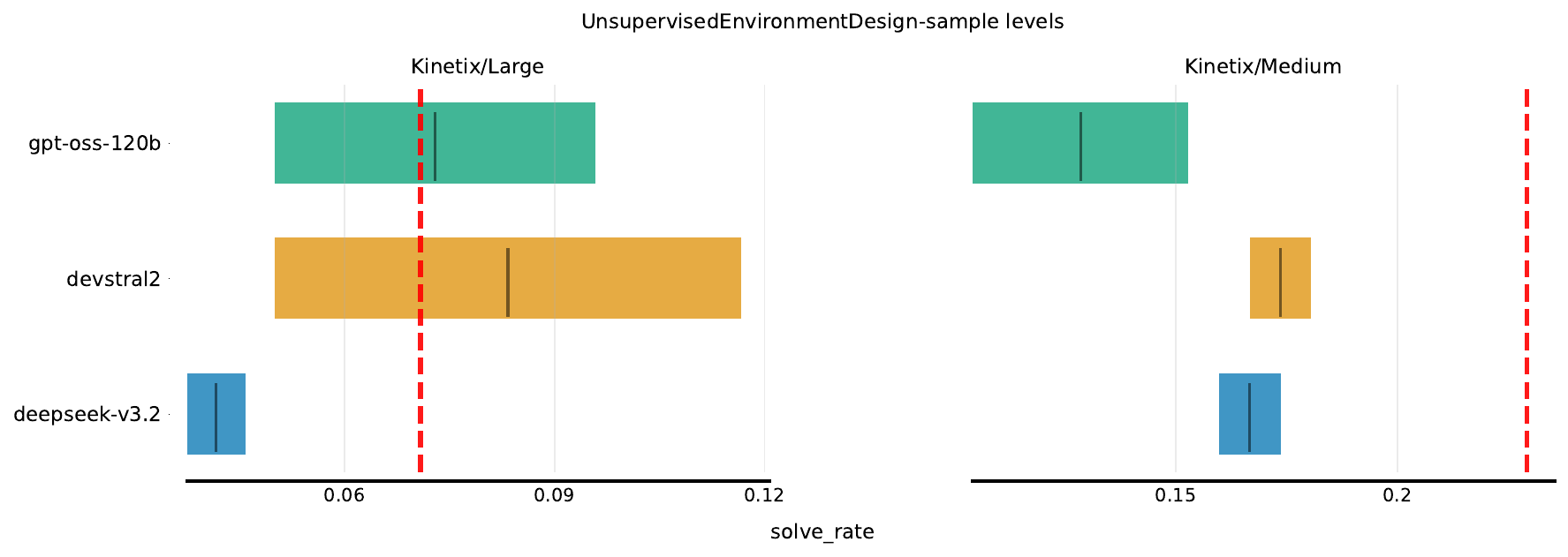}%
\\[0.5em]
\includegraphics[width=0.48\textwidth]{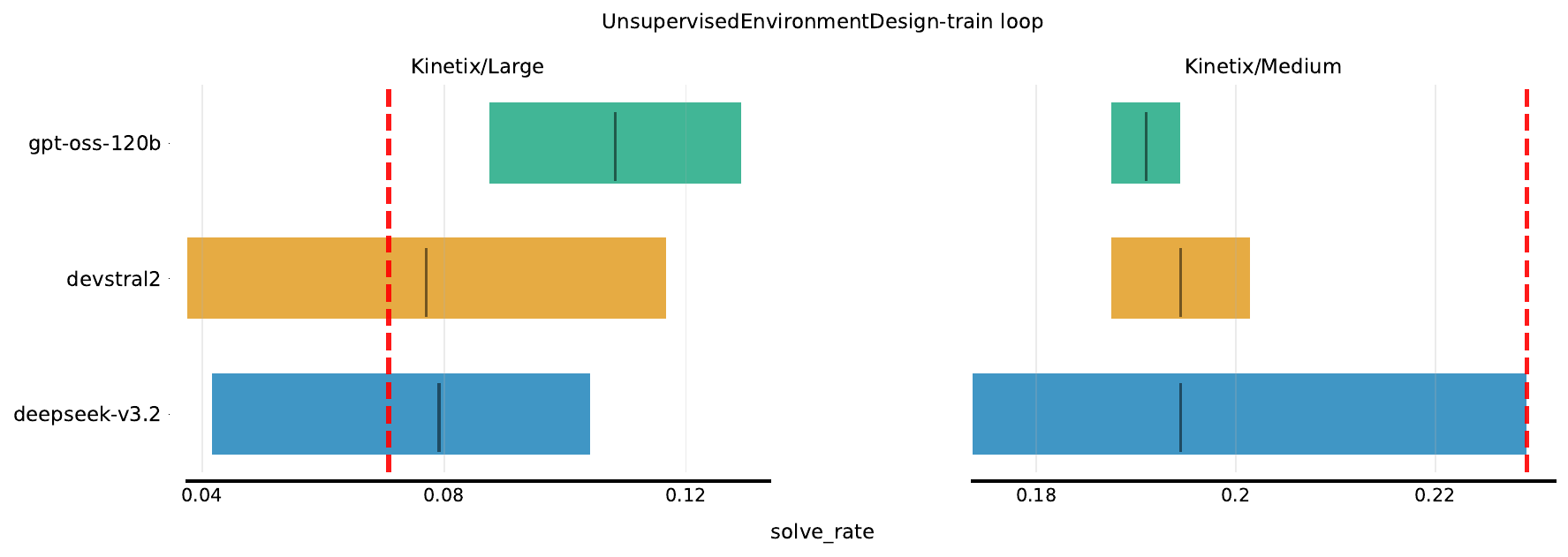}%
\hfill%
\includegraphics[width=0.48\textwidth]{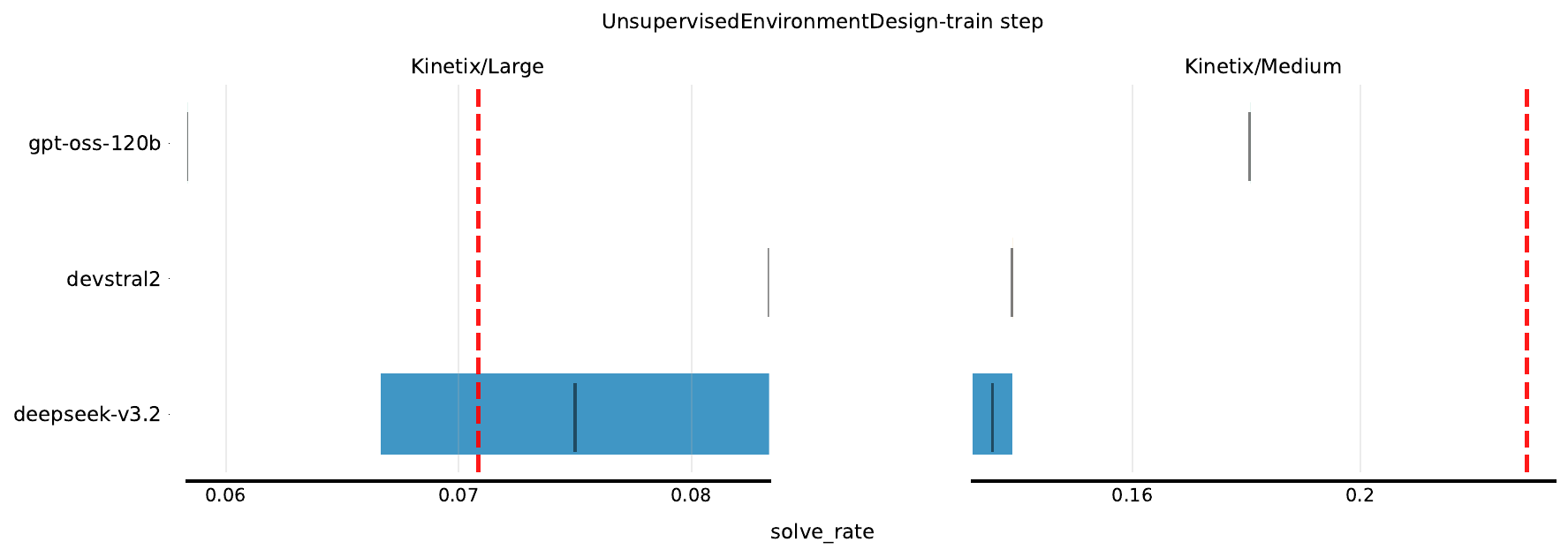}%
\caption{DiscoBench (Single Edit) results on Meta-Test tasks. (Part 7/7)}
\label{fig:one_change_mt_7}
\end{figure}
\clearpage

\subsection{DiscoBench (All Edit) -- Meta-Train}
\label{sec:all_change_id}

\begin{figure}[htbp]
\centering
\setlength{\lineskip}{0pt}
\includegraphics[width=0.48\textwidth]{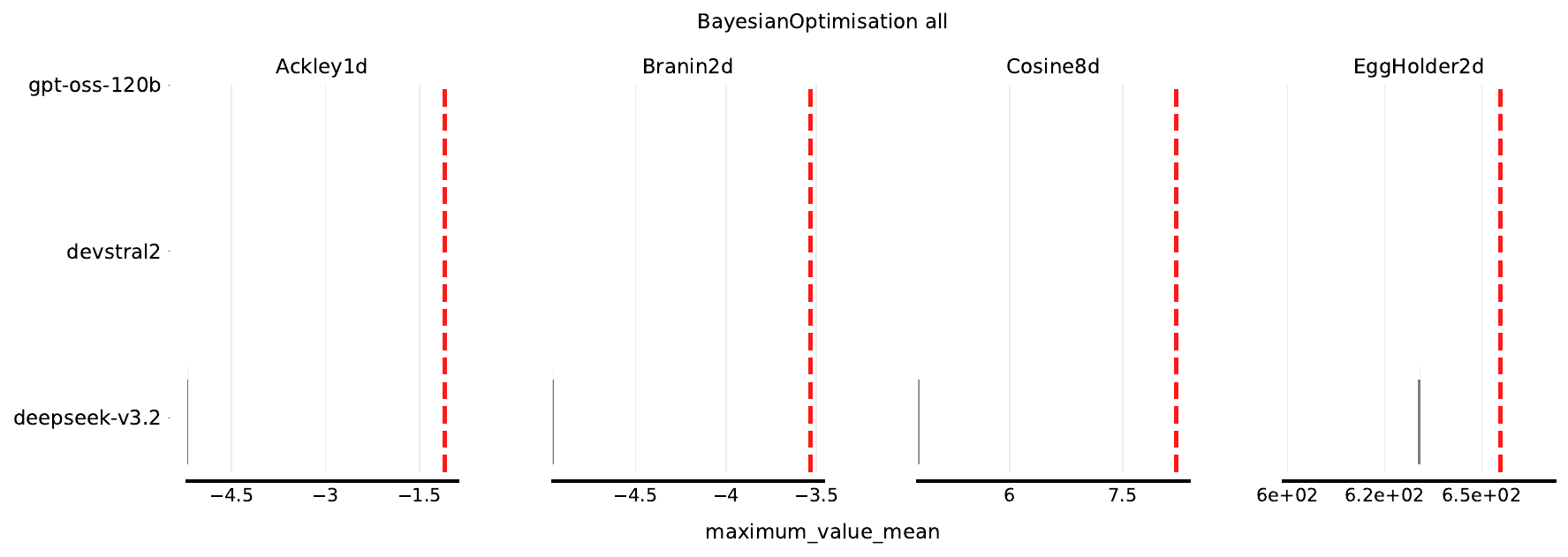}%
\hfill%
\includegraphics[width=0.48\textwidth]{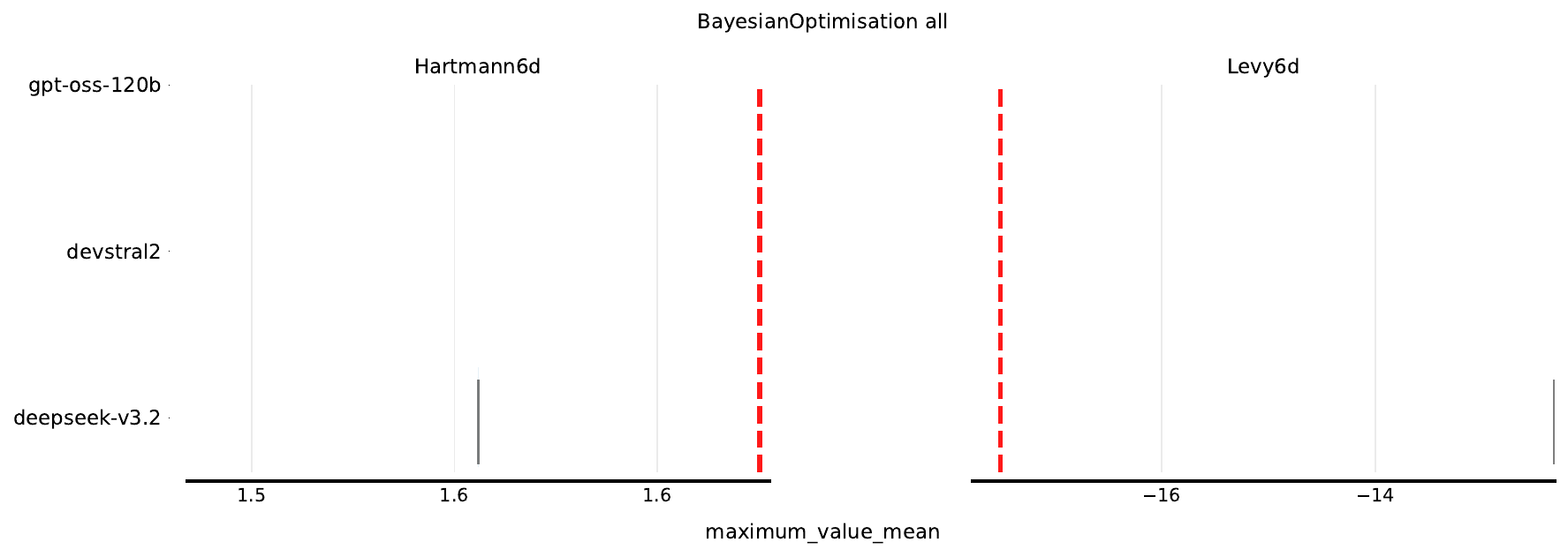}%
\\[0.5em]
\includegraphics[width=0.48\textwidth]{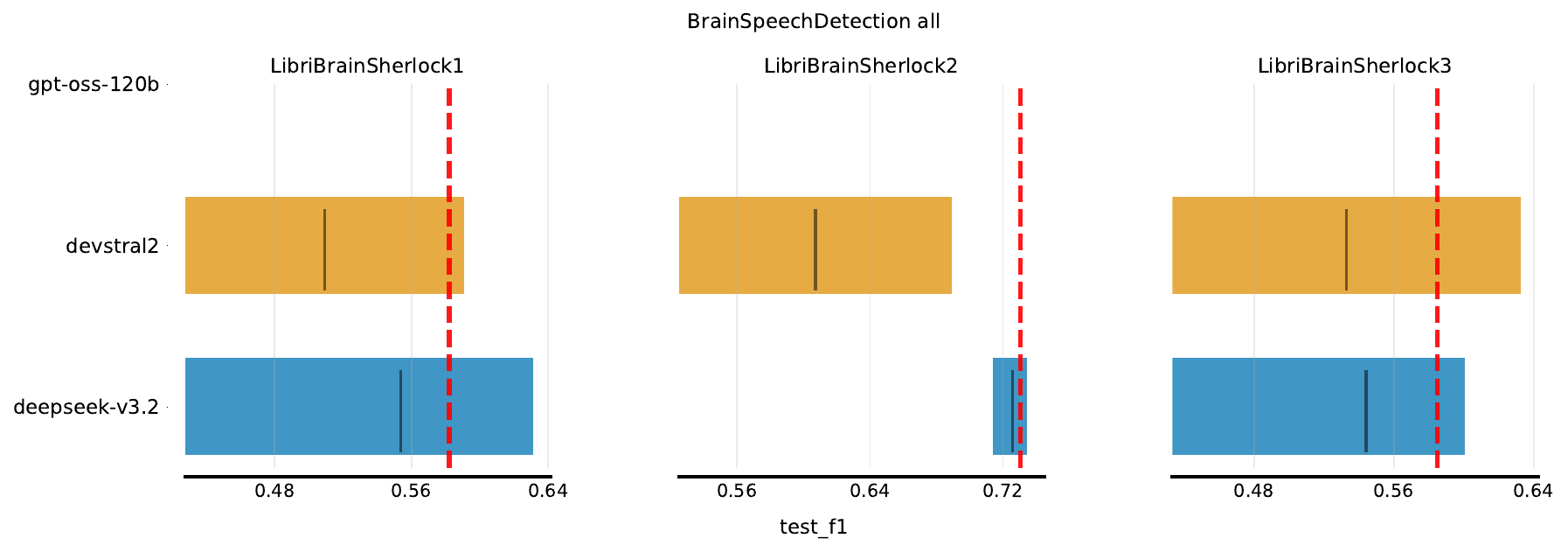}%
\hfill%
\includegraphics[width=0.48\textwidth]{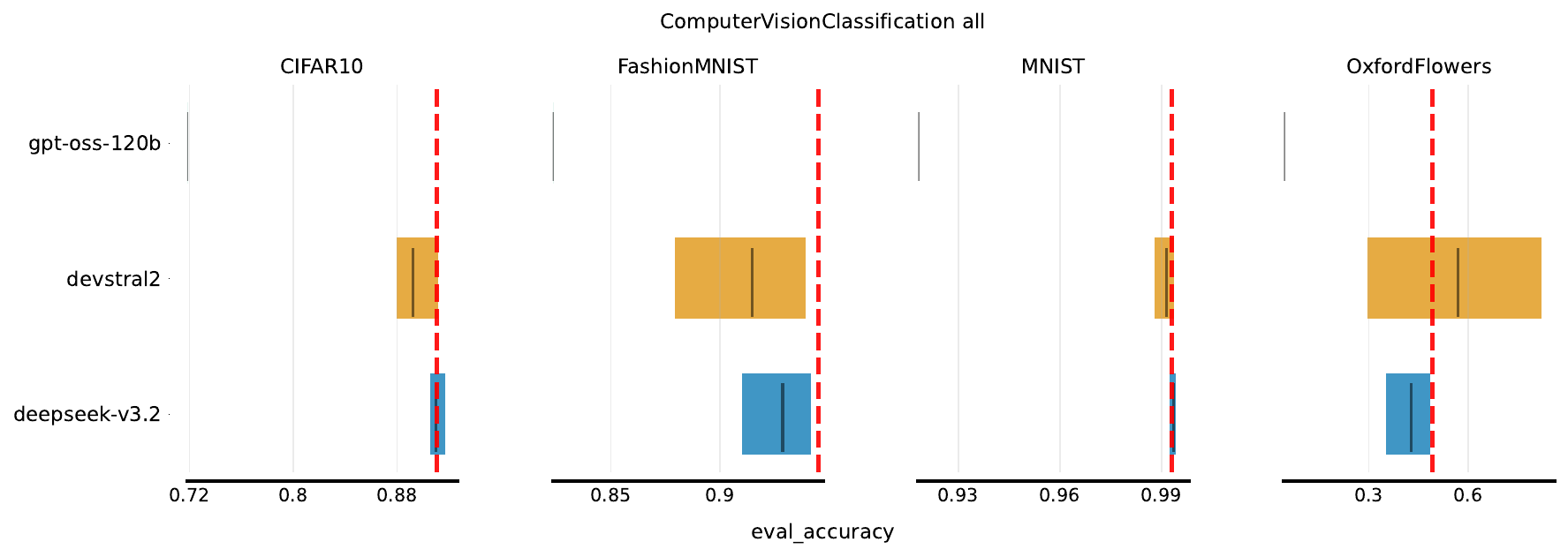}%
\\[0.5em]
\includegraphics[width=0.48\textwidth]{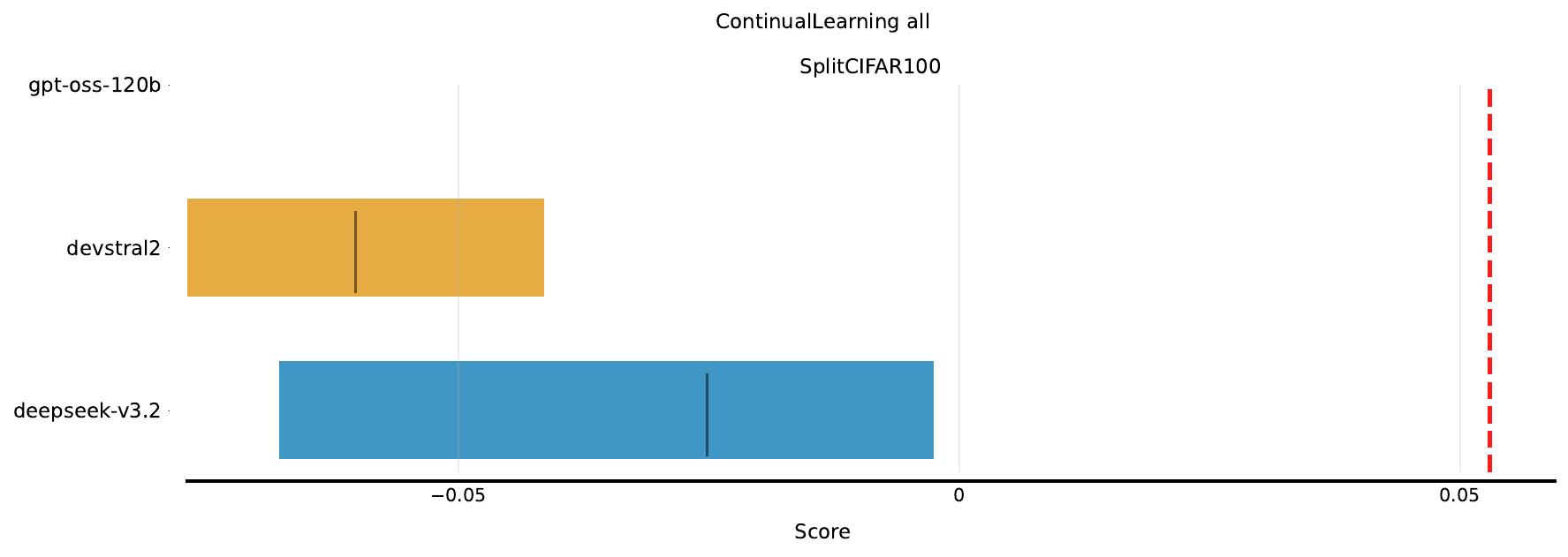}%
\hfill%
\includegraphics[width=0.48\textwidth]{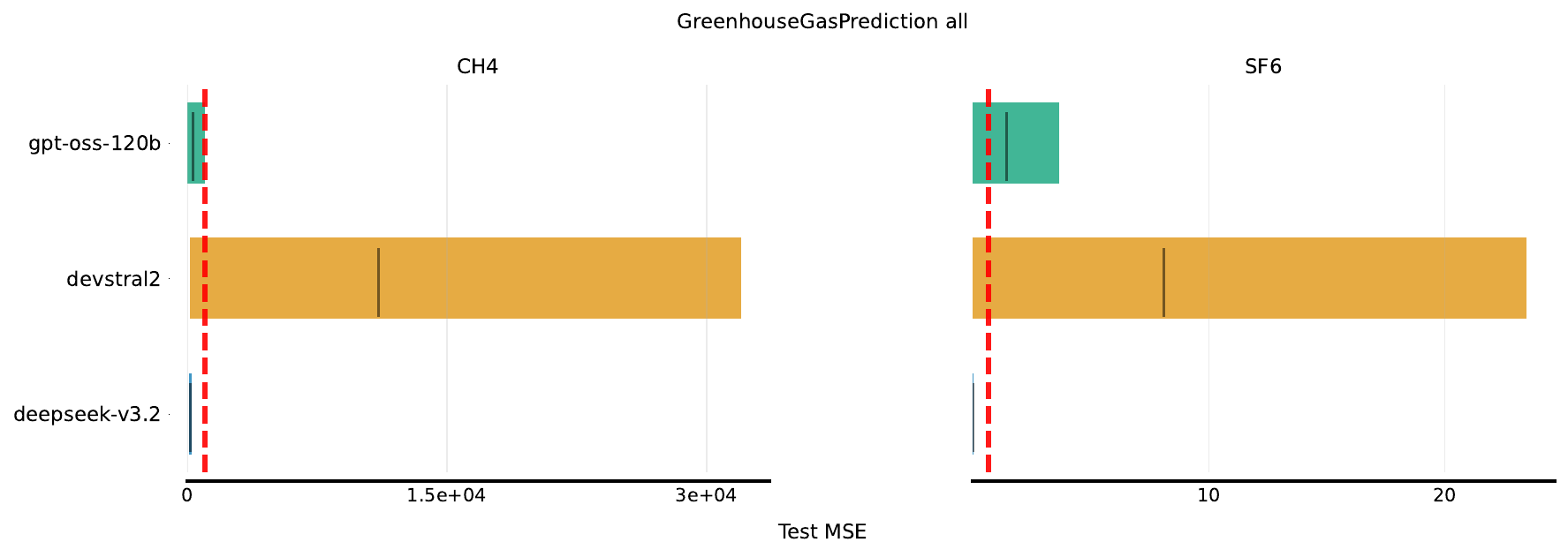}%
\\[0.5em]
\includegraphics[width=0.48\textwidth]{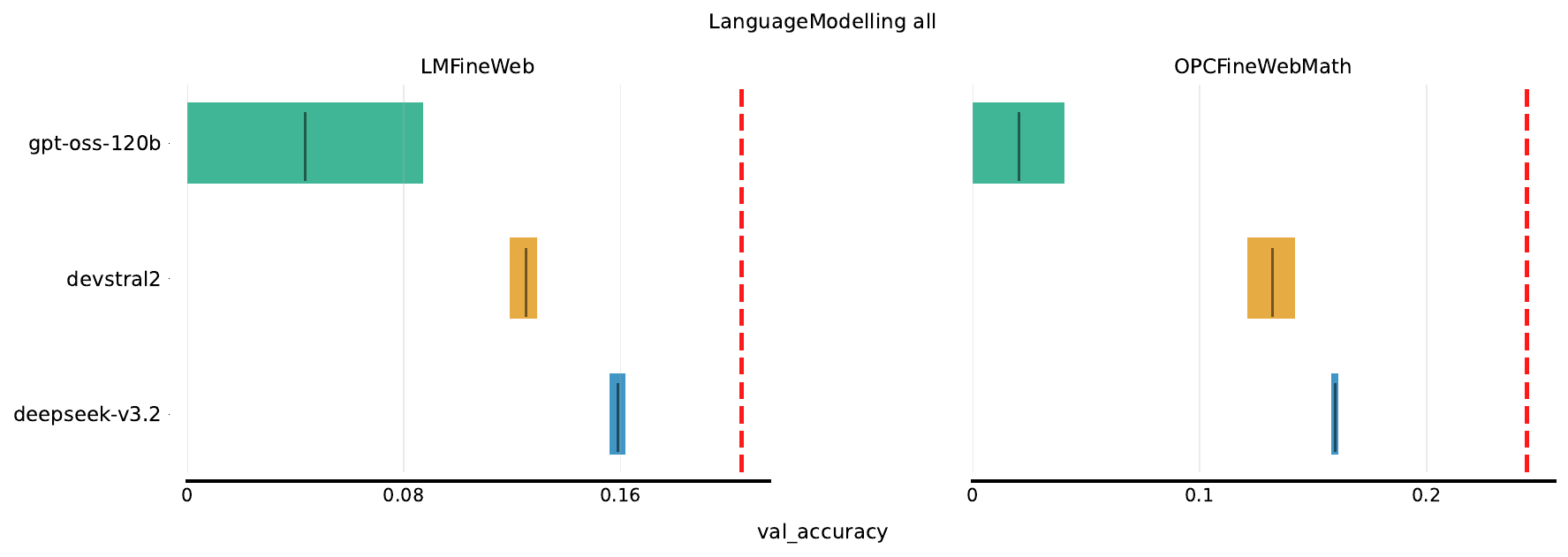}%
\hfill%
\includegraphics[width=0.48\textwidth]{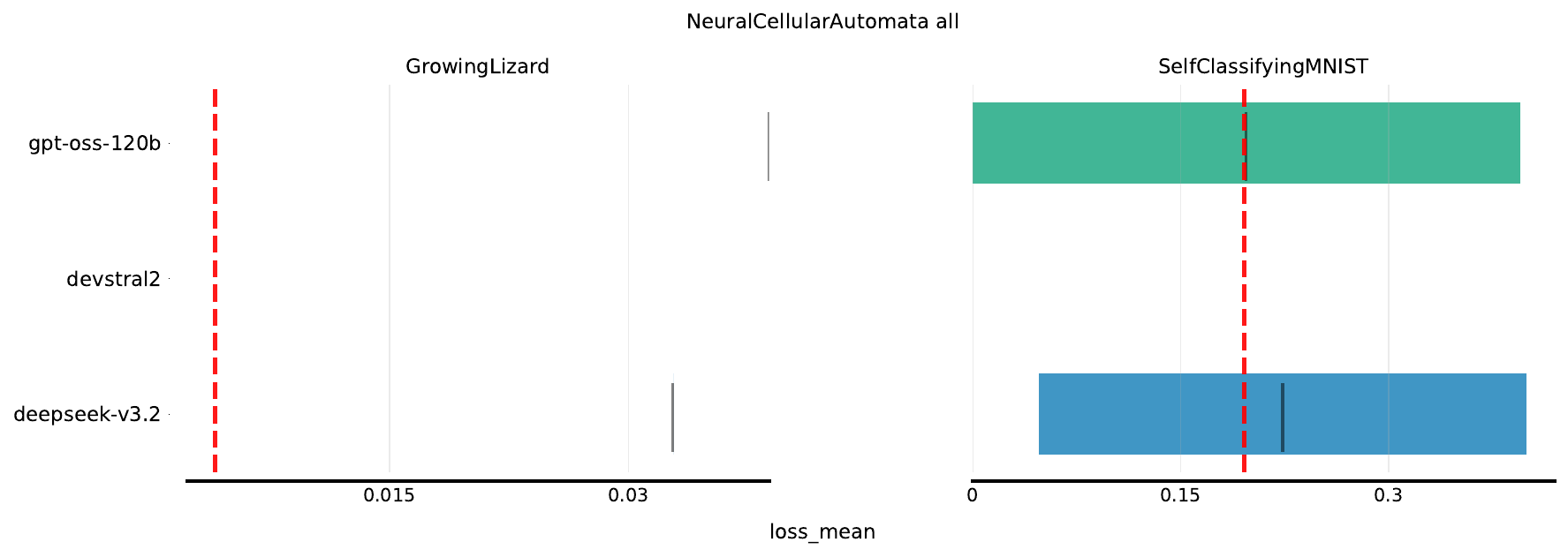}%
\\[0.5em]
\includegraphics[width=0.48\textwidth]{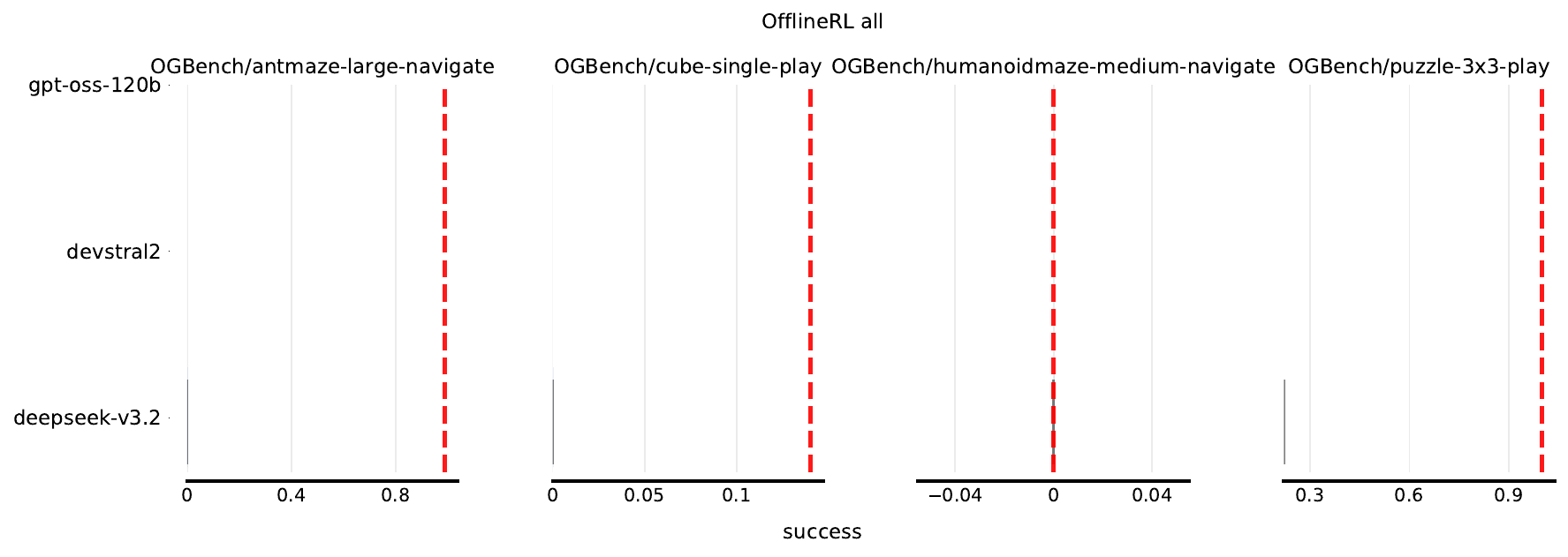}%
\hfill%
\includegraphics[width=0.48\textwidth]{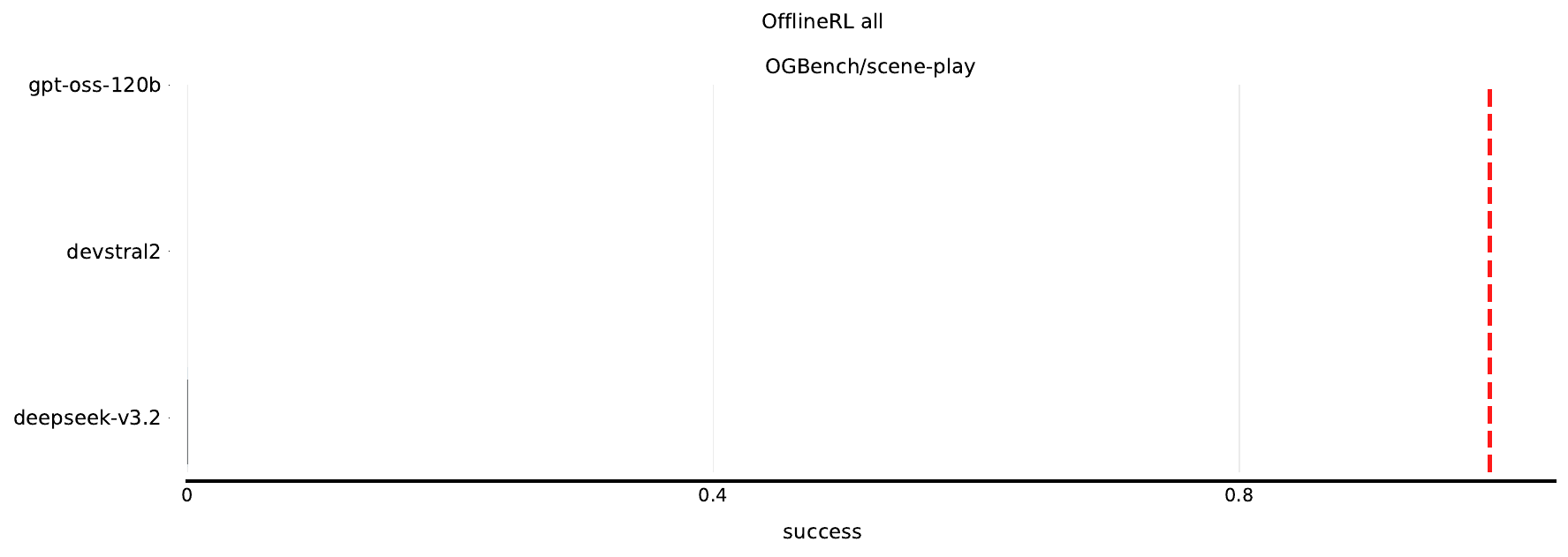}%
\caption{DiscoBench (All Edit) results on Meta-Train tasks. (Part 1/2)}
\label{fig:all_change_id_1}
\end{figure}
\clearpage

\begin{figure}[htbp]
\centering
\setlength{\lineskip}{0pt}
\includegraphics[width=0.48\textwidth]{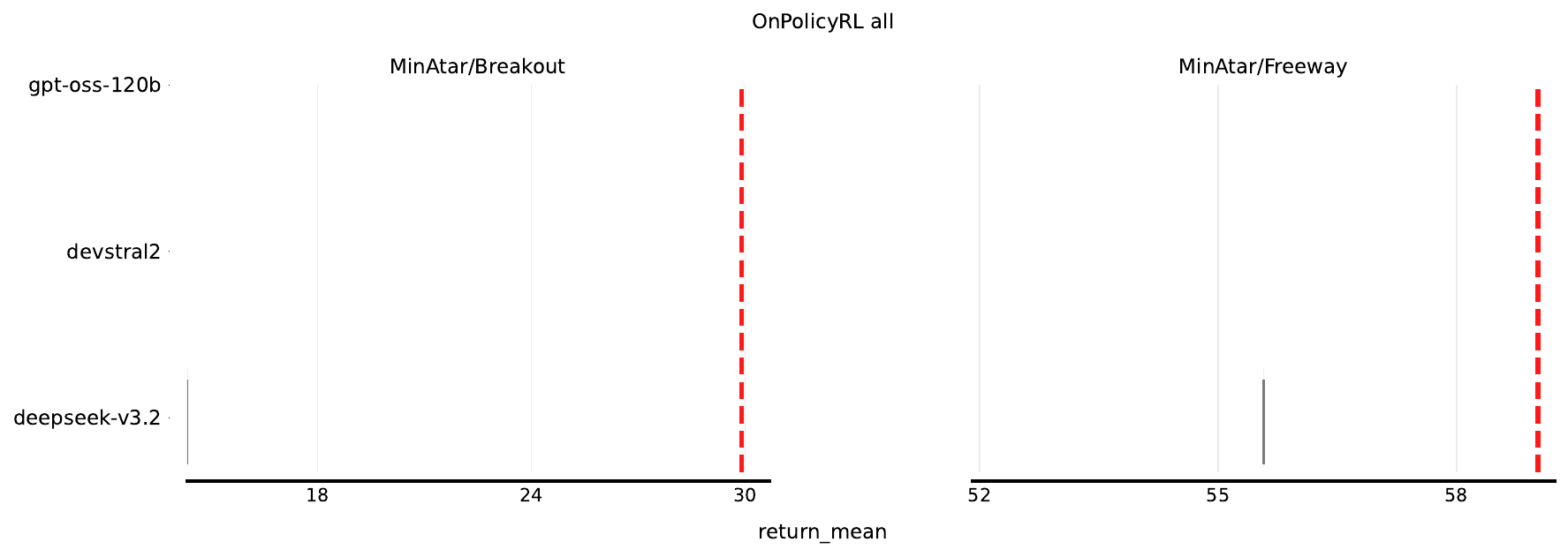}%
\hfill%
\includegraphics[width=0.48\textwidth]{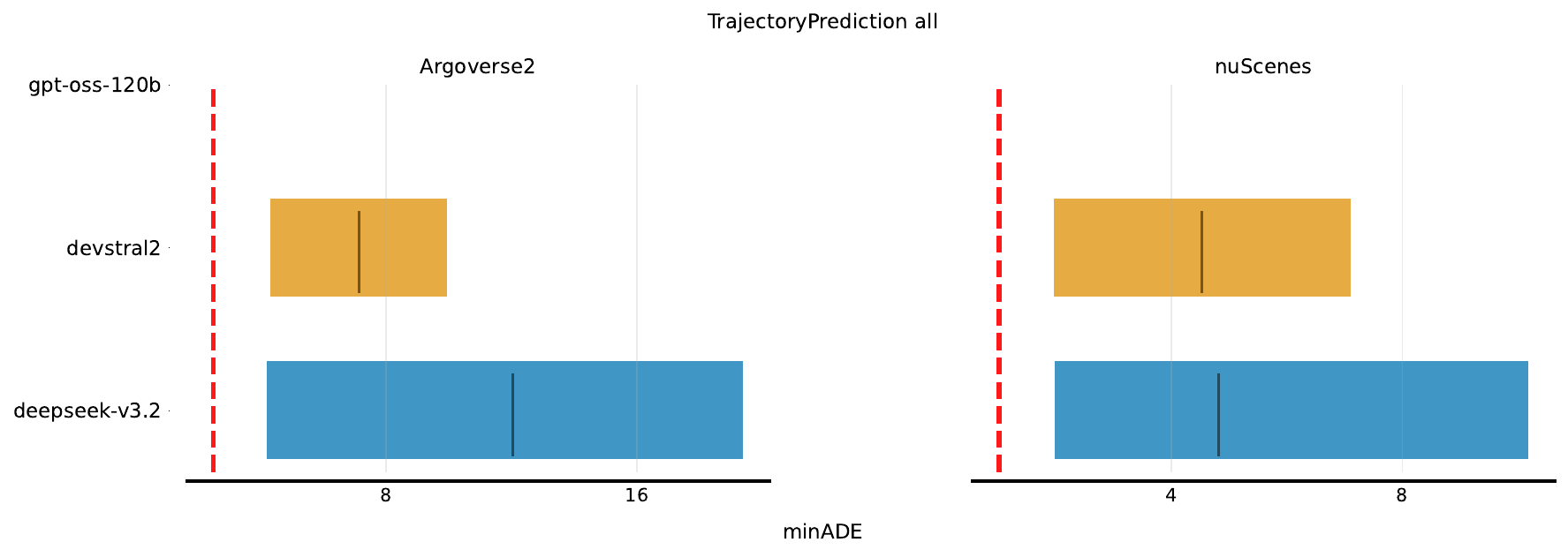}%
\caption{DiscoBench (All Edit) results on Meta-Train tasks. (Part 2/2)}
\label{fig:all_change_id_2}
\end{figure}
\clearpage

\subsection{DiscoBench (All Edit) -- Meta-Test}
\label{sec:all_change_mt}

\begin{figure}[htbp]
\centering
\setlength{\lineskip}{0pt}
\includegraphics[width=0.48\textwidth]{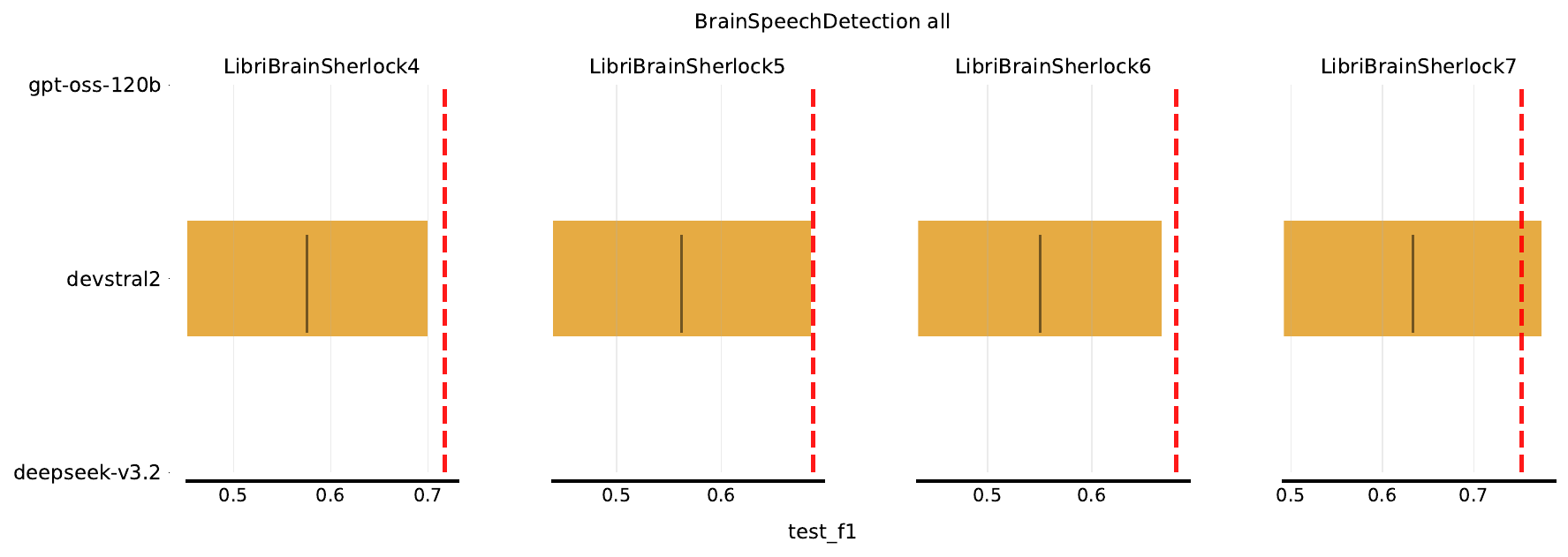}%
\hfill%
\includegraphics[width=0.48\textwidth]{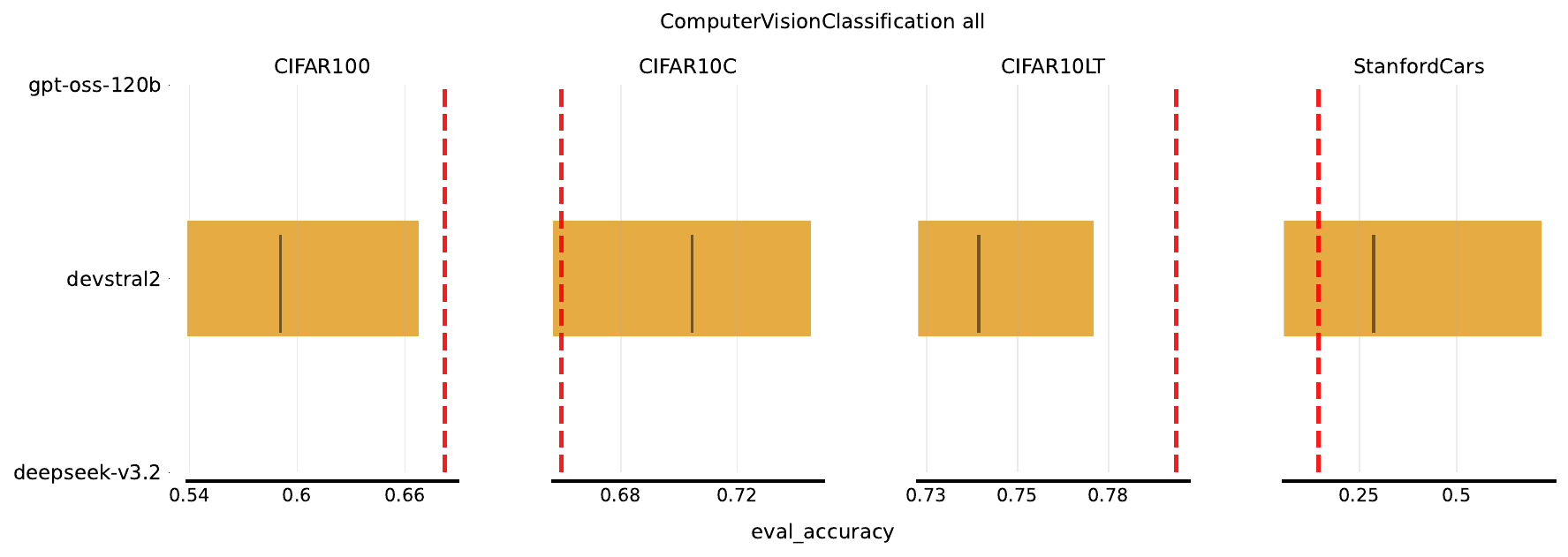}%
\\[0.5em]
\includegraphics[width=0.48\textwidth]{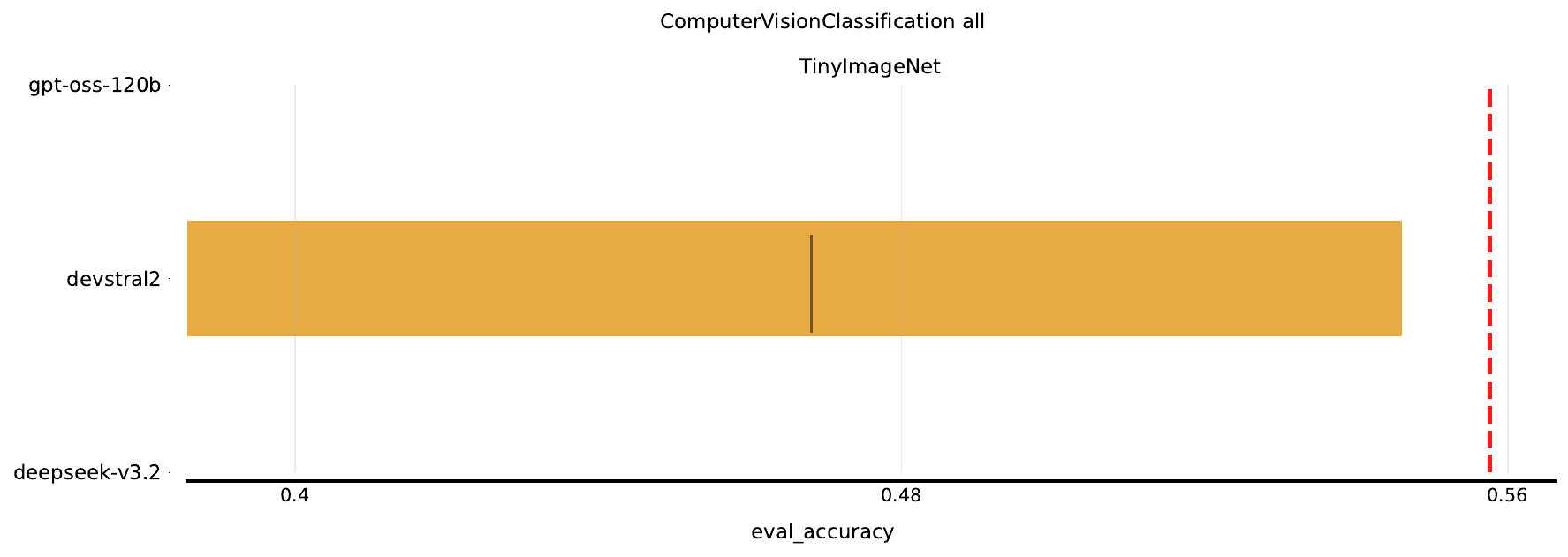}%
\hfill%
\includegraphics[width=0.48\textwidth]{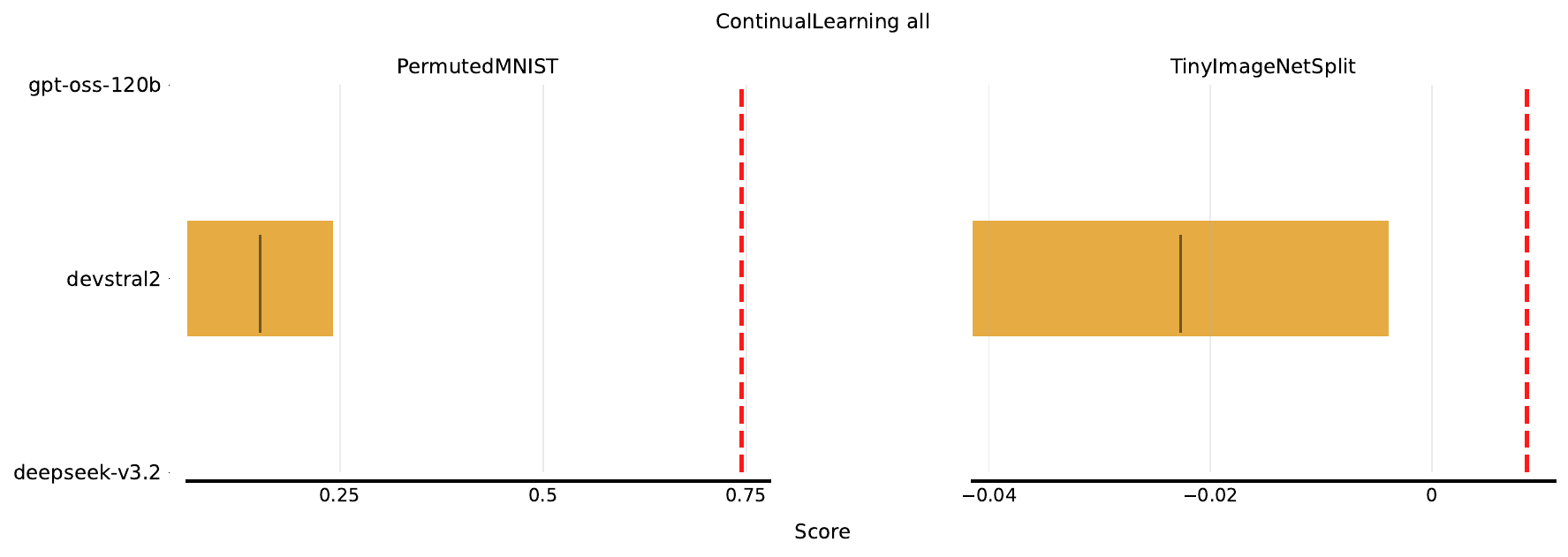}%
\\[0.5em]
\includegraphics[width=0.48\textwidth]{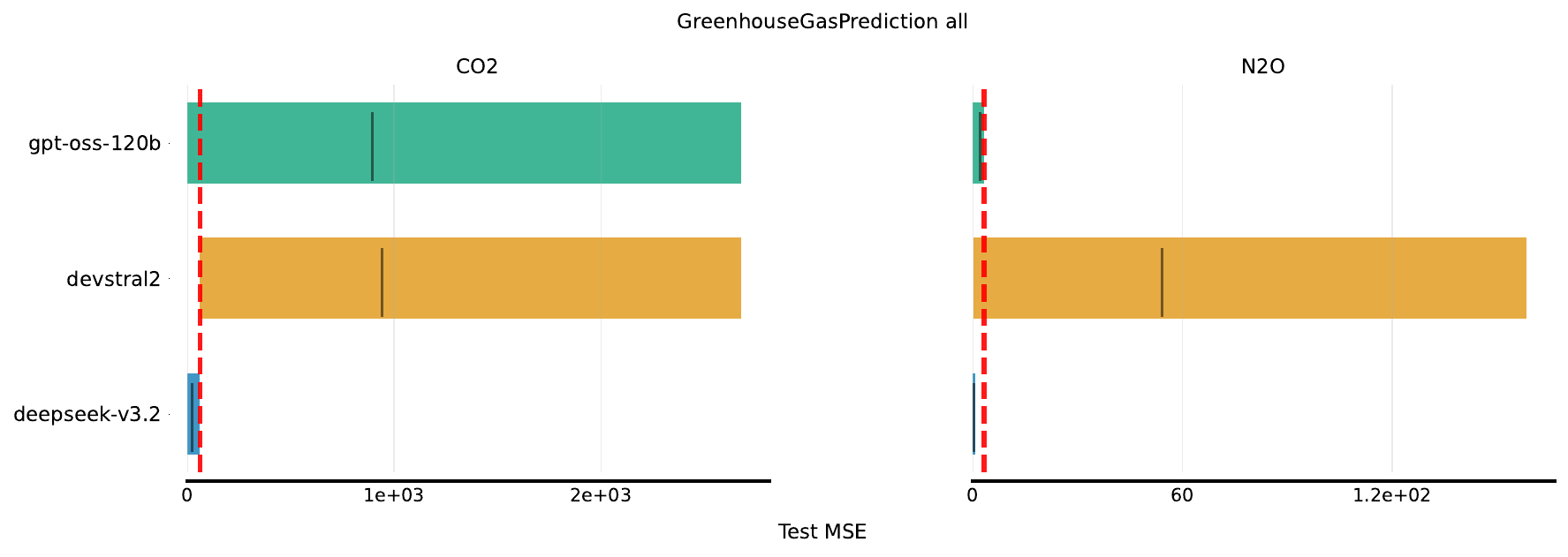}%
\hfill%
\includegraphics[width=0.48\textwidth]{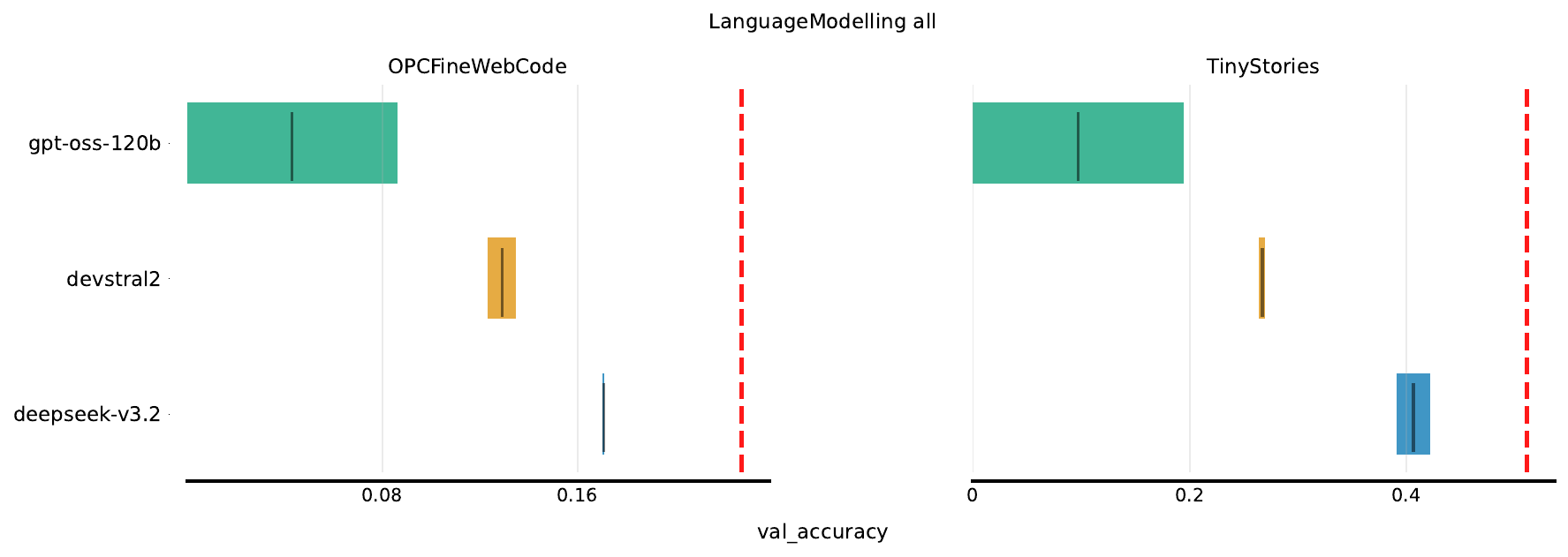}%
\\[0.5em]
\includegraphics[width=0.48\textwidth]{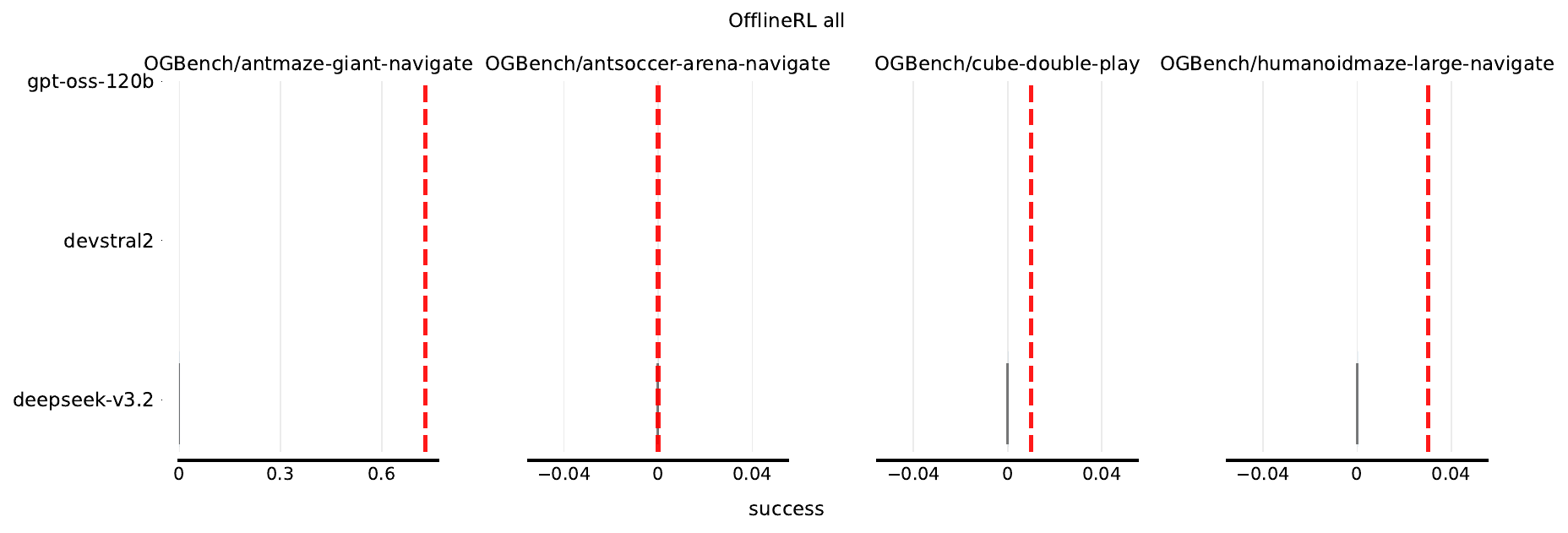}%
\hfill%
\includegraphics[width=0.48\textwidth]{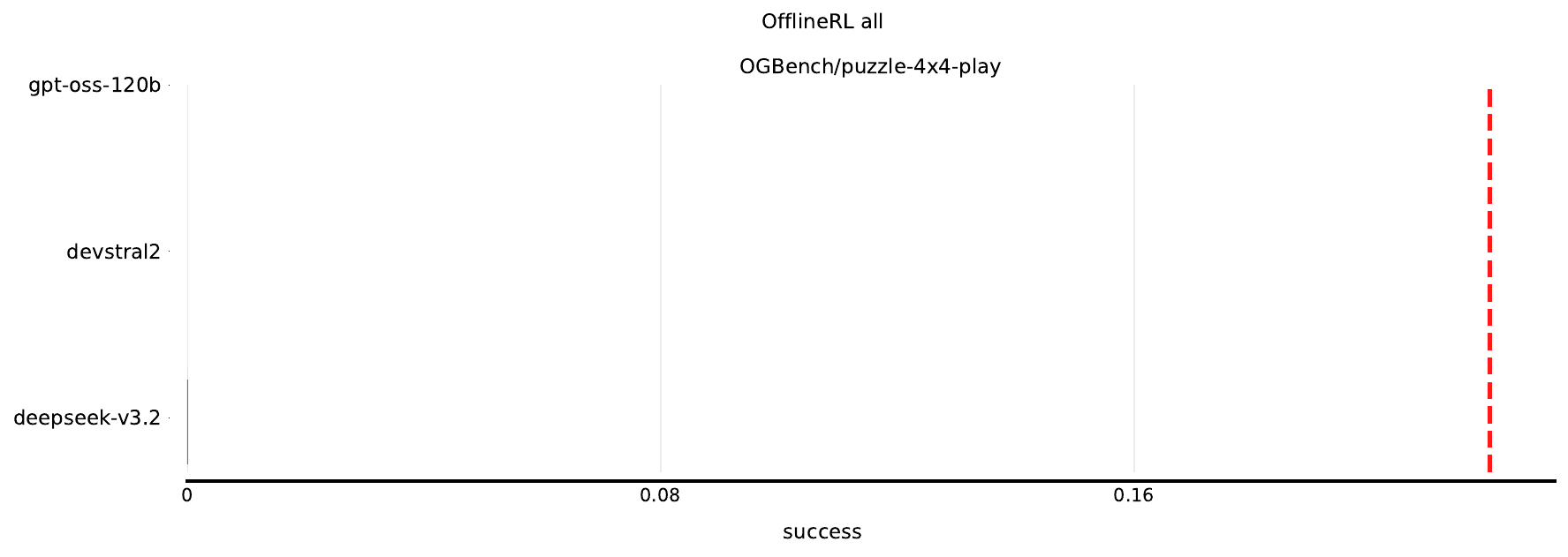}%
\\[0.5em]
\includegraphics[width=0.48\textwidth]{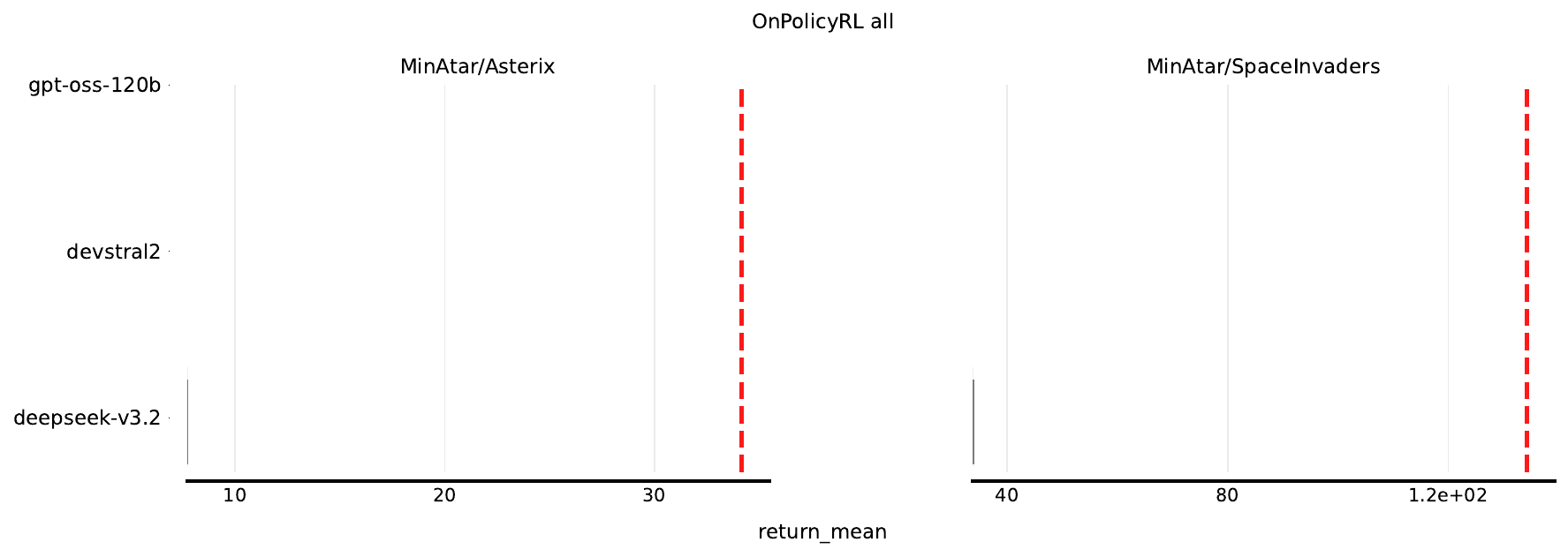}%
\hfill%
\includegraphics[width=0.48\textwidth]{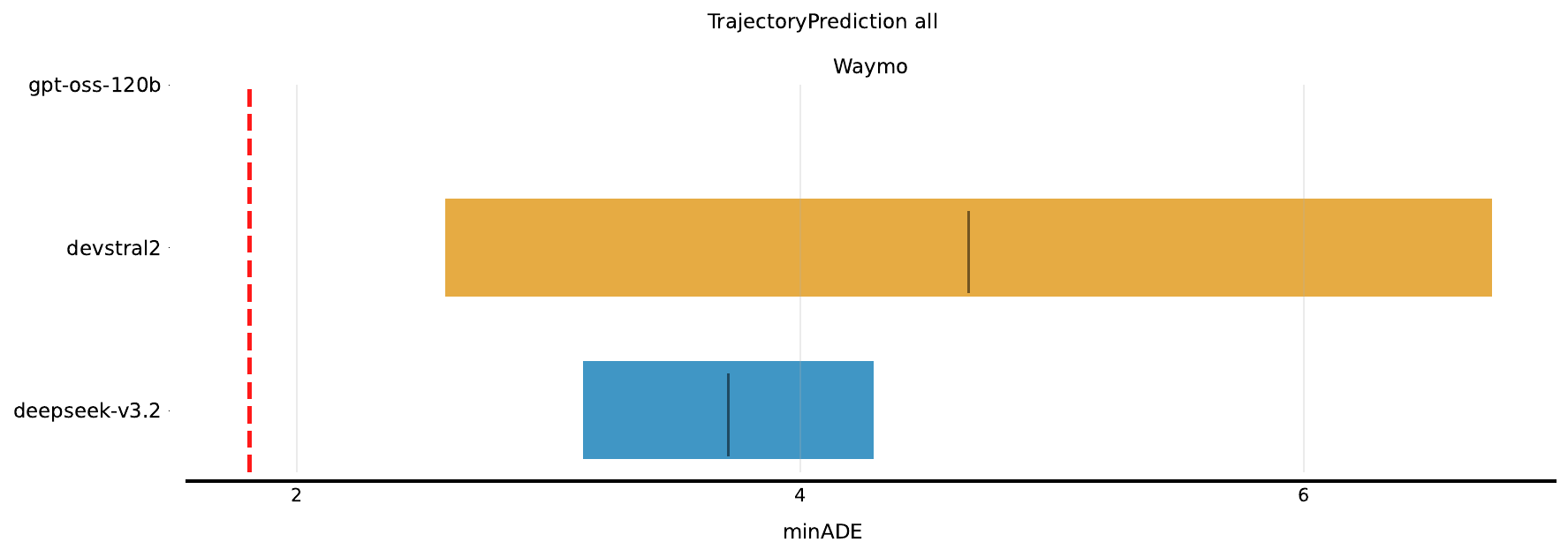}%
\caption{DiscoBench (All Edit) results on Meta-Test tasks.}
\label{fig:all_change_mt_1}
\end{figure}
\clearpage

\subsection{DiscoBench (3 Successful Seeds) -- Meta-Train}
\label{sec:until_success_id}

\begin{figure}[htbp]
\centering
\setlength{\lineskip}{0pt}
\includegraphics[width=0.48\textwidth]{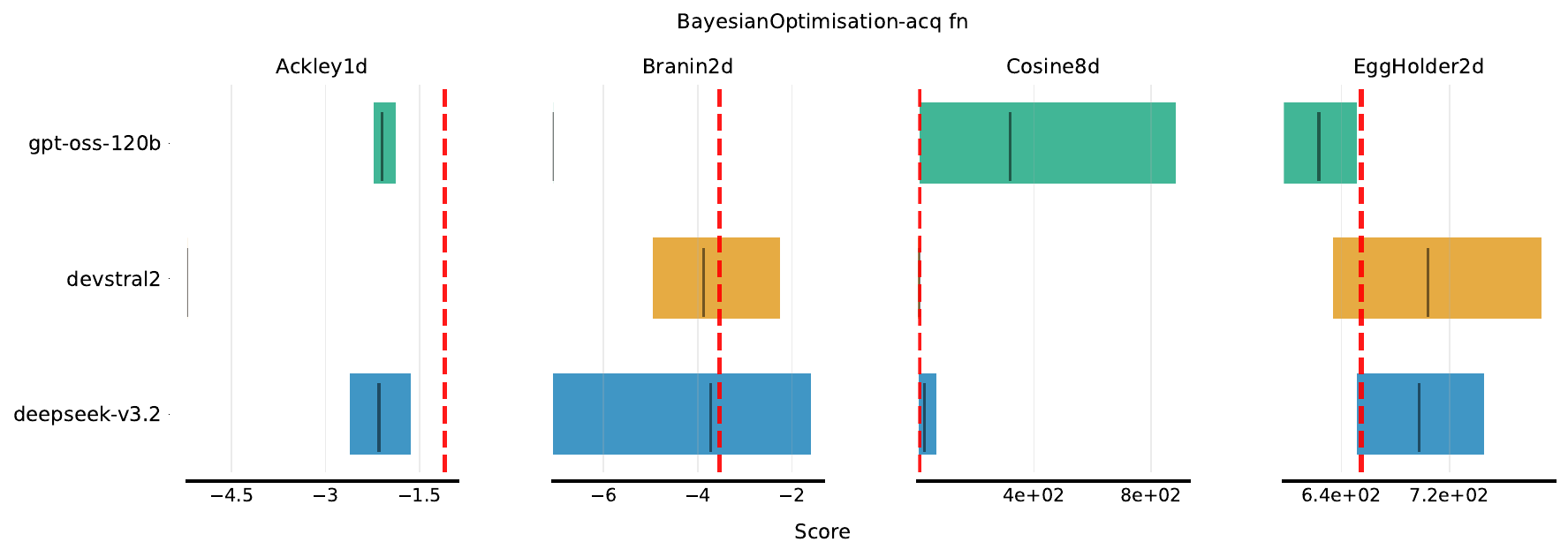}%
\hfill%
\includegraphics[width=0.48\textwidth]{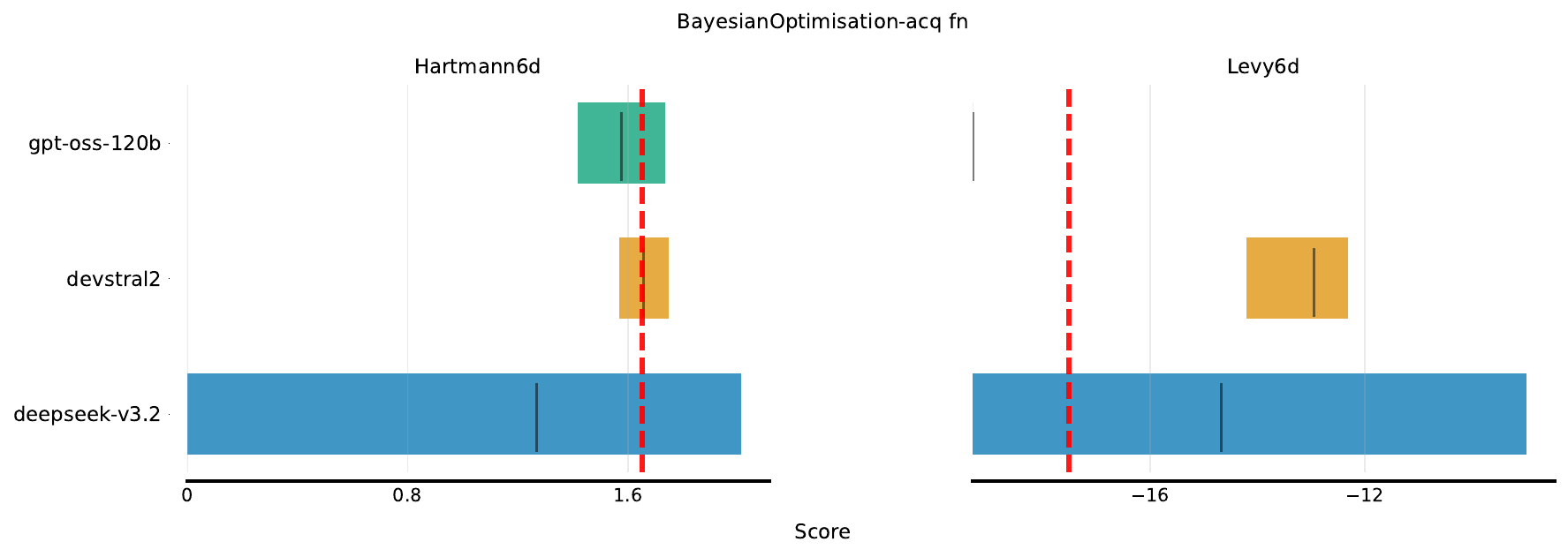}%
\\[0.5em]
\includegraphics[width=0.48\textwidth]{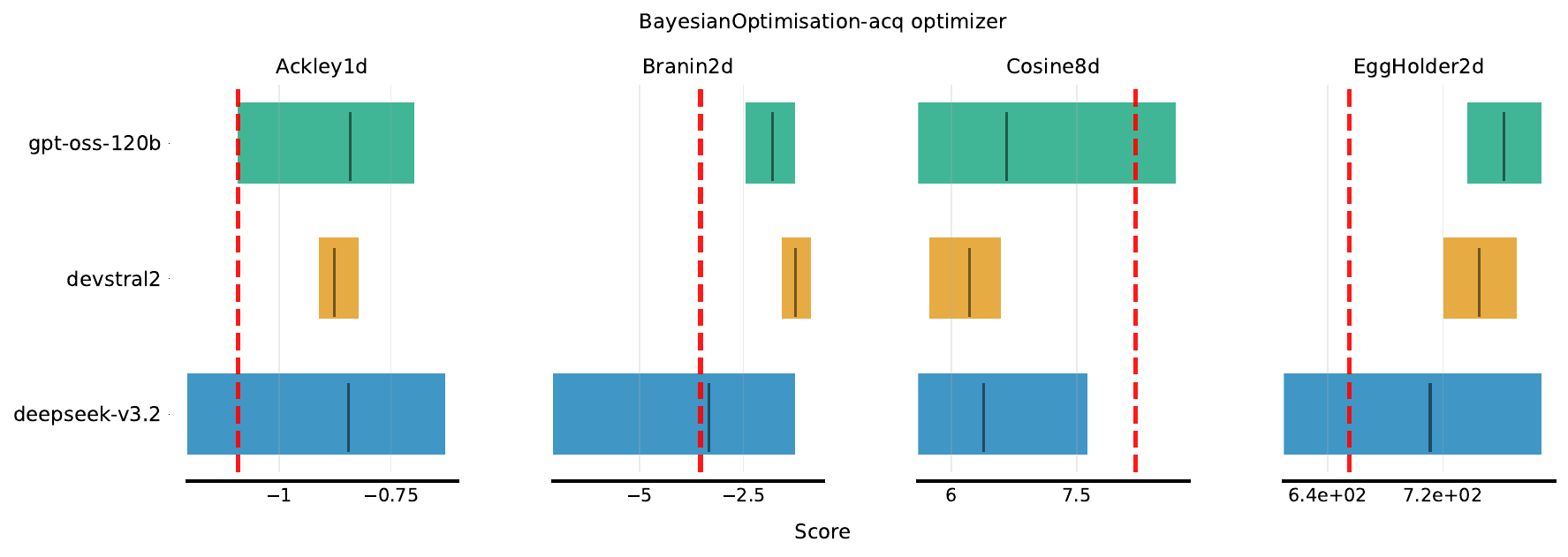}%
\hfill%
\includegraphics[width=0.48\textwidth]{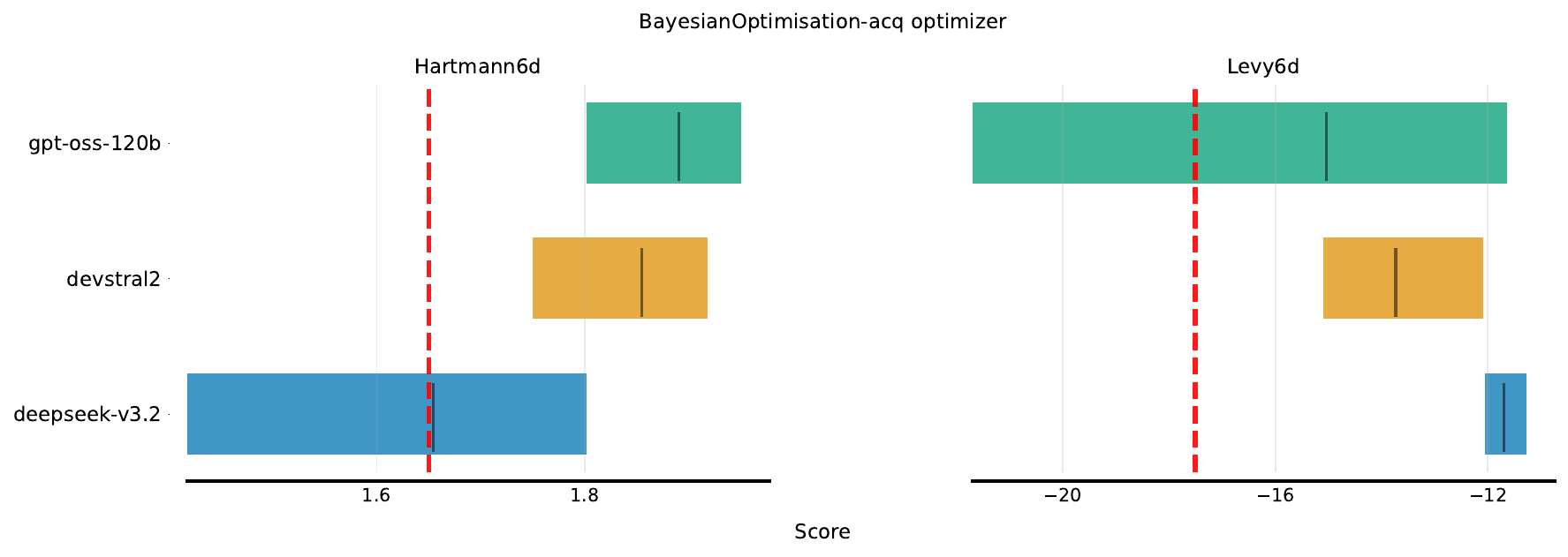}%
\\[0.5em]
\includegraphics[width=0.48\textwidth]{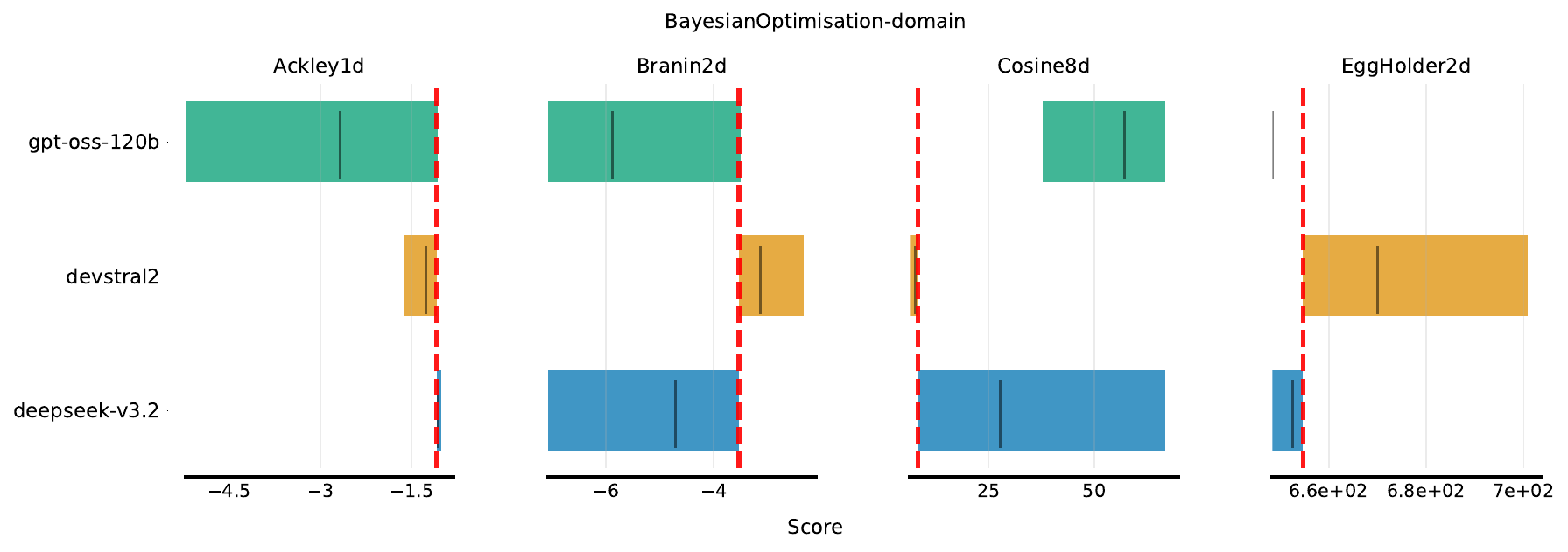}%
\hfill%
\includegraphics[width=0.48\textwidth]{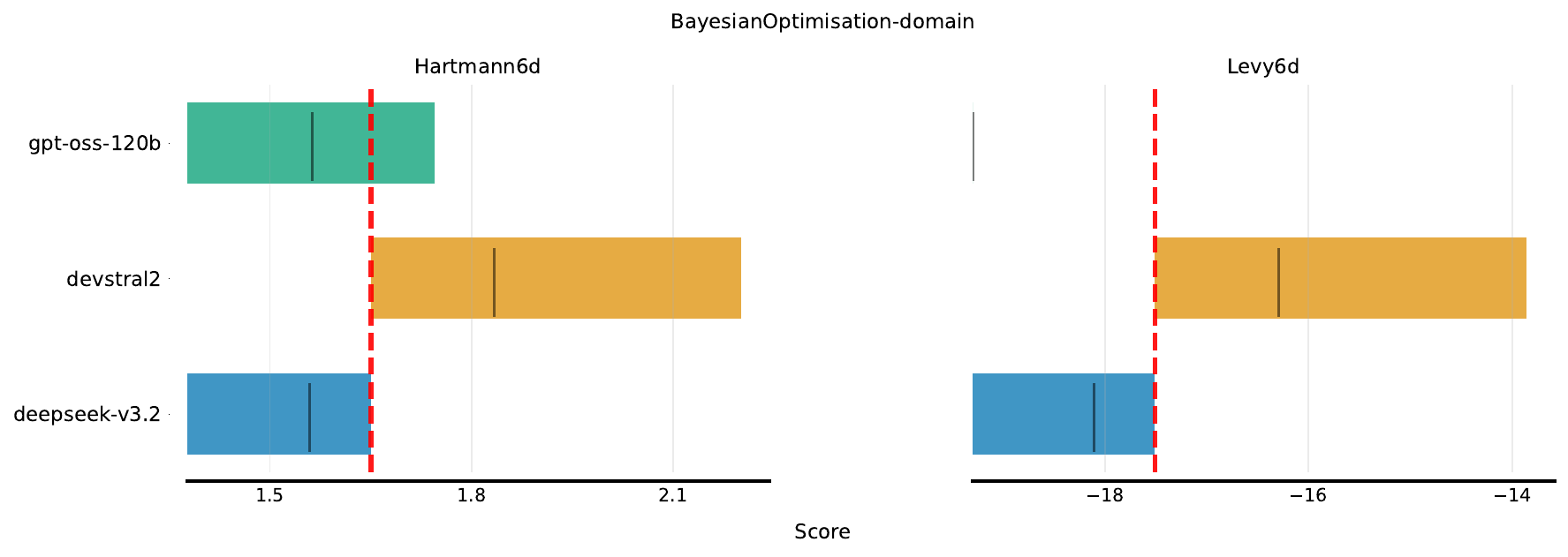}%
\\[0.5em]
\includegraphics[width=0.48\textwidth]{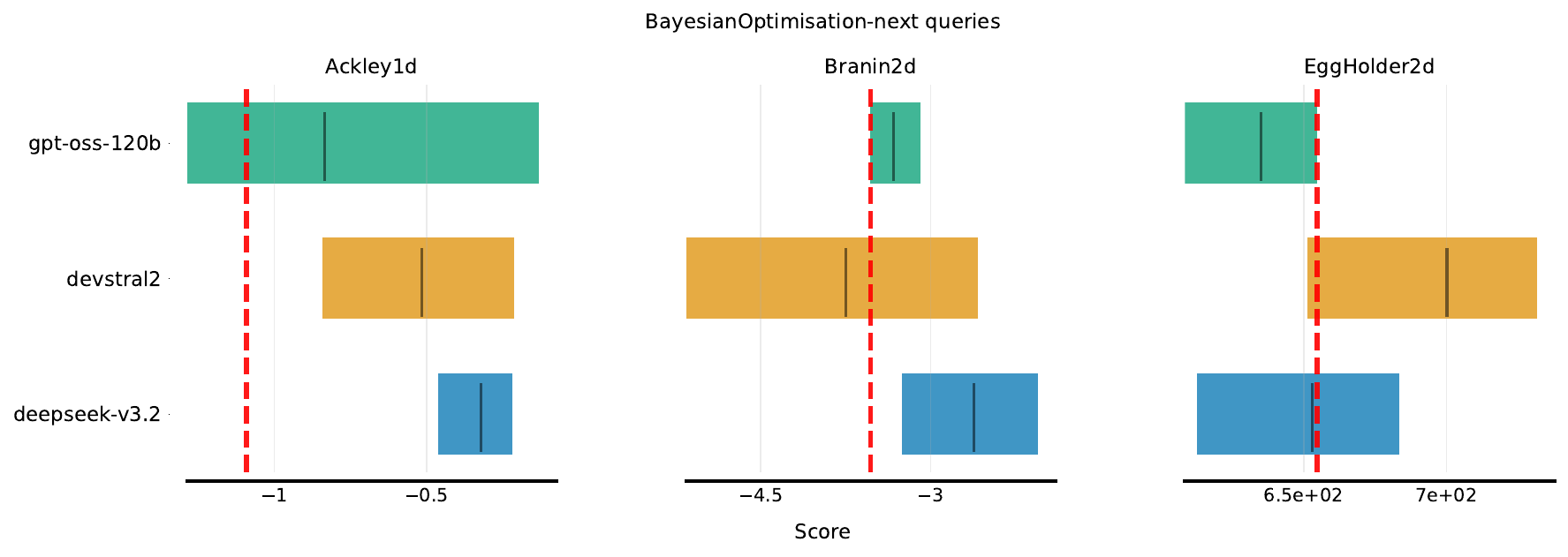}%
\hfill%
\includegraphics[width=0.48\textwidth]{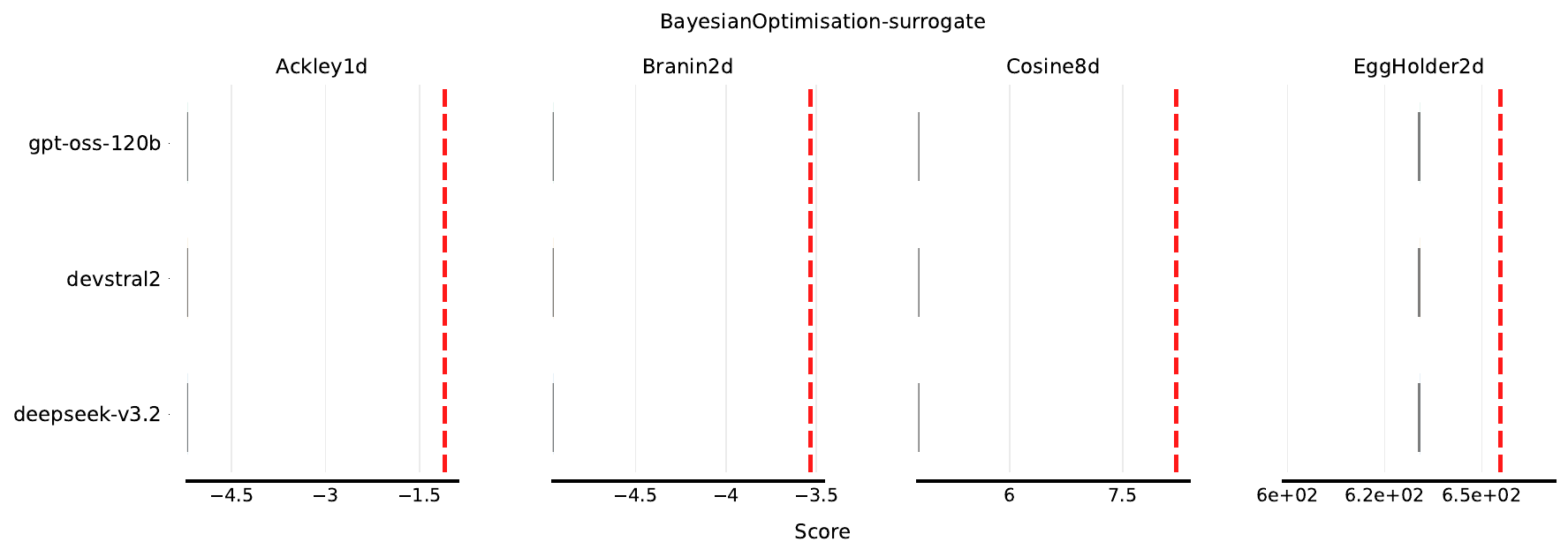}%
\\[0.5em]
\includegraphics[width=0.48\textwidth]{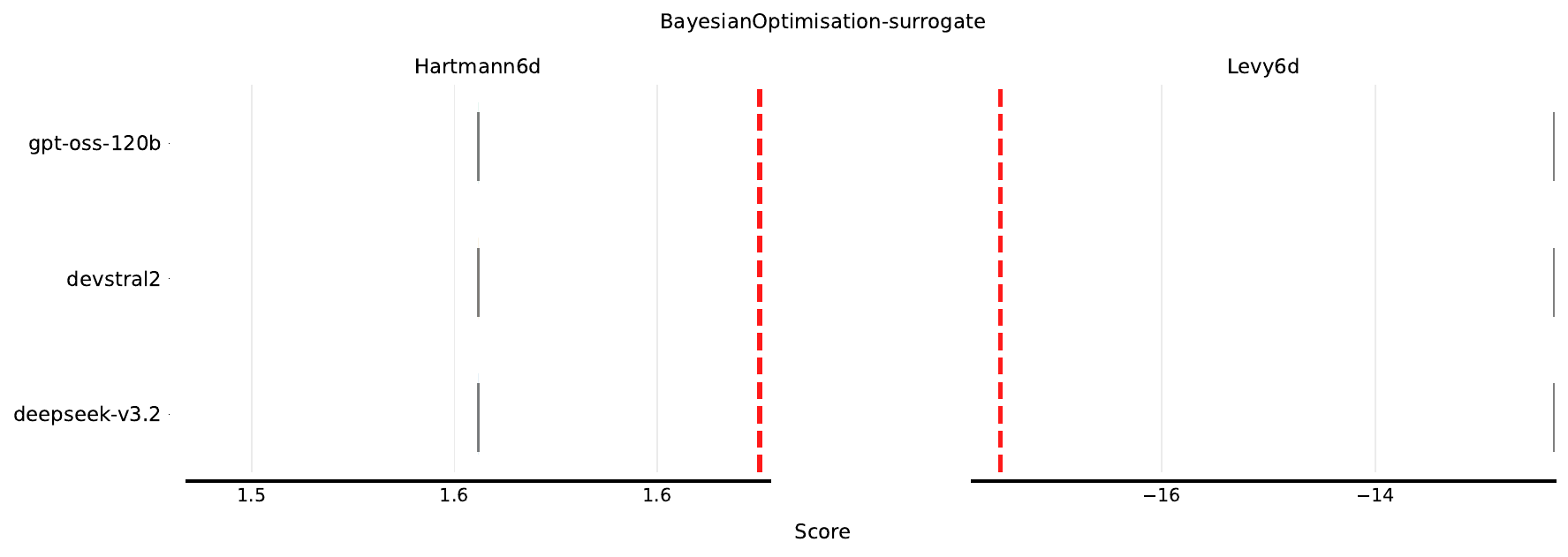}%
\hfill%
\includegraphics[width=0.48\textwidth]{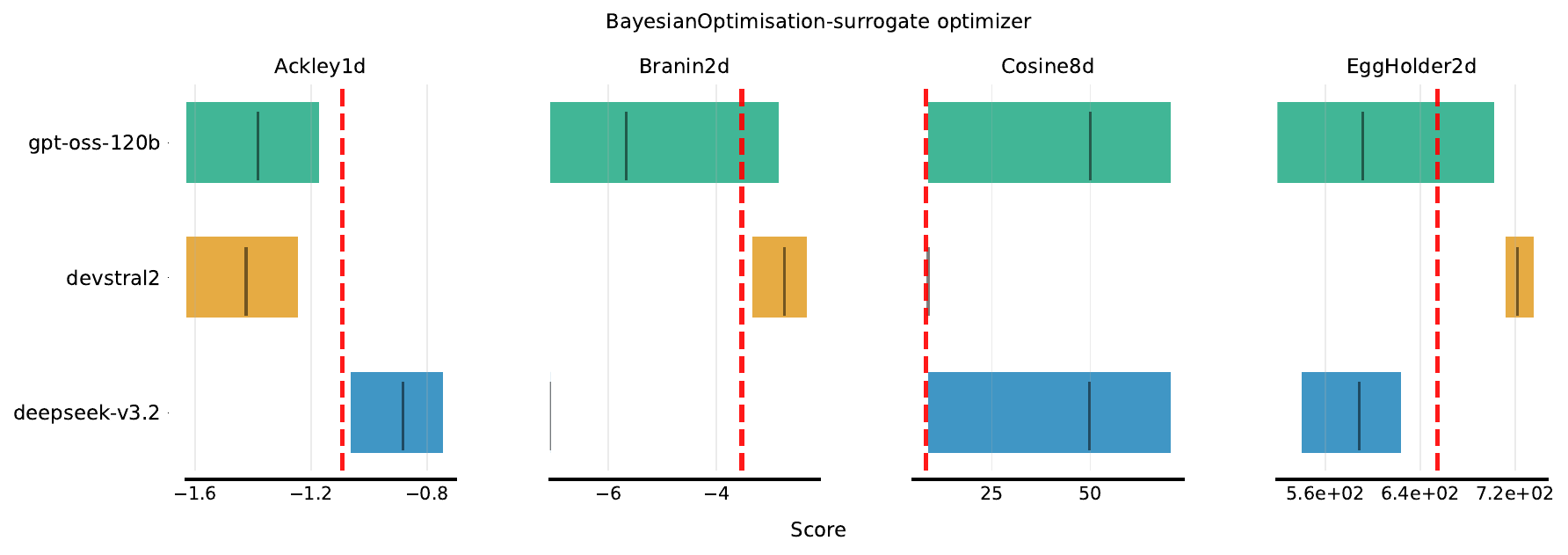}%
\caption{DiscoBench (3 Successful Seeds) results on Meta-Train tasks. (Part 1/4)}
\label{fig:until_success_id_1}
\end{figure}
\clearpage

\begin{figure}[htbp]
\centering
\setlength{\lineskip}{0pt}
\includegraphics[width=0.48\textwidth]{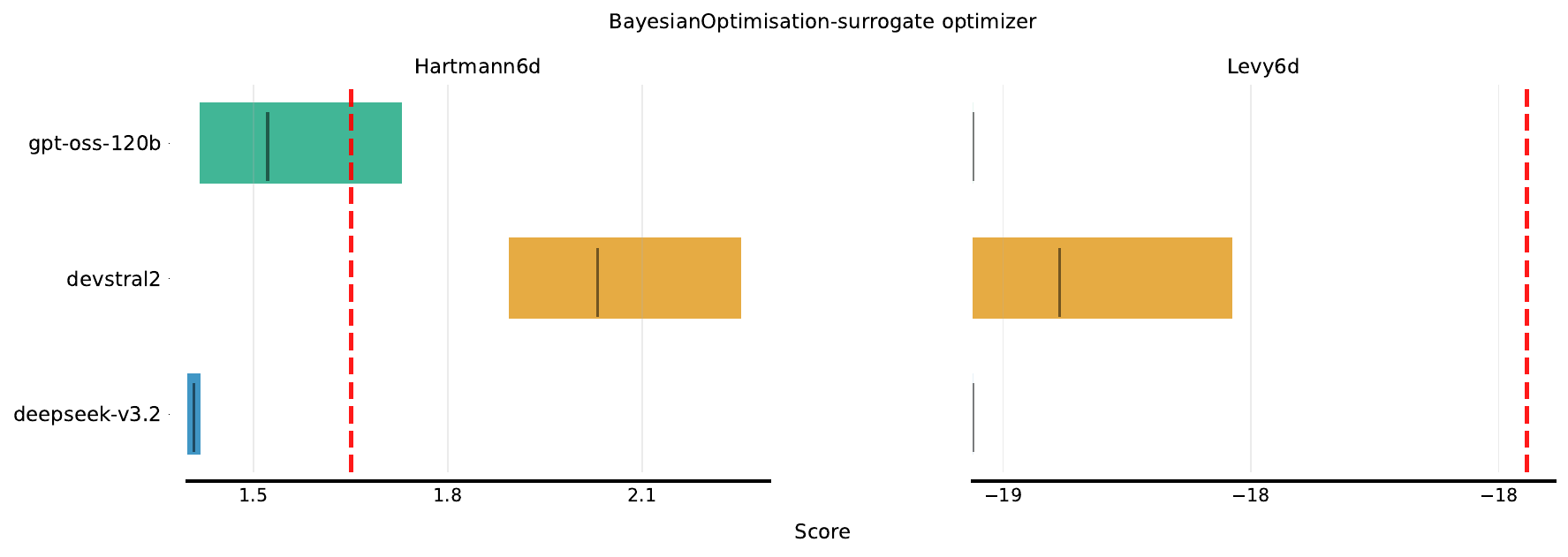}%
\hfill%
\includegraphics[width=0.48\textwidth]{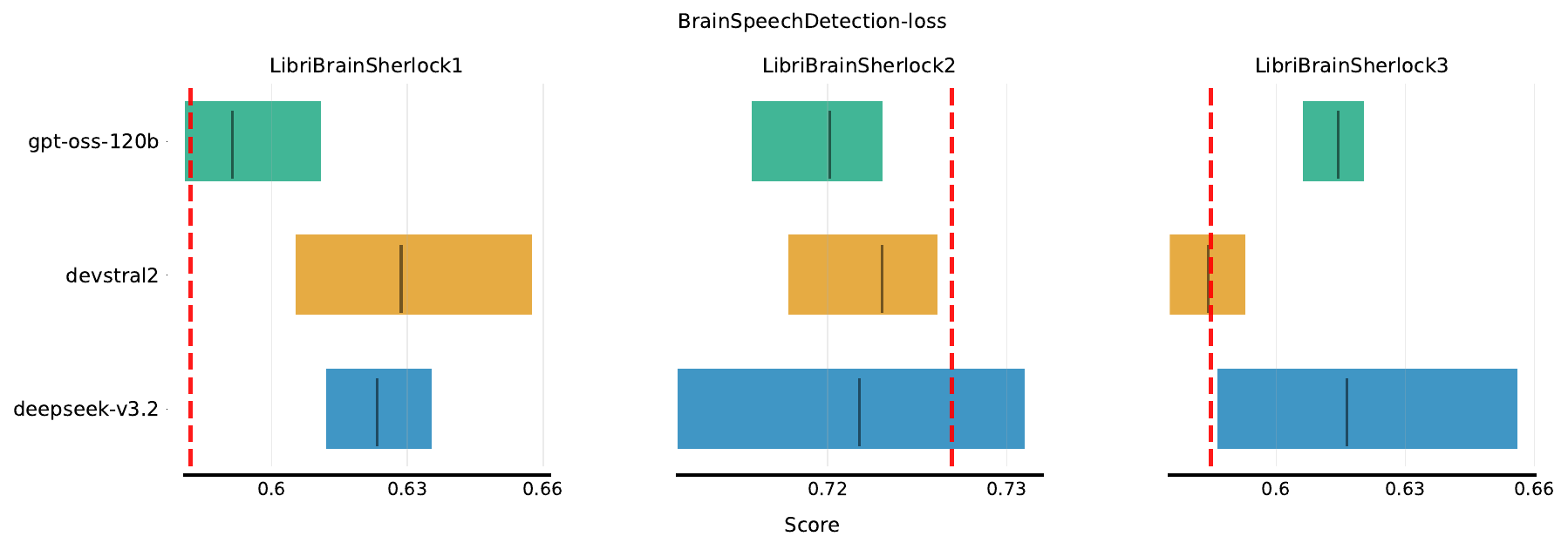}%
\\[0.5em]
\includegraphics[width=0.48\textwidth]{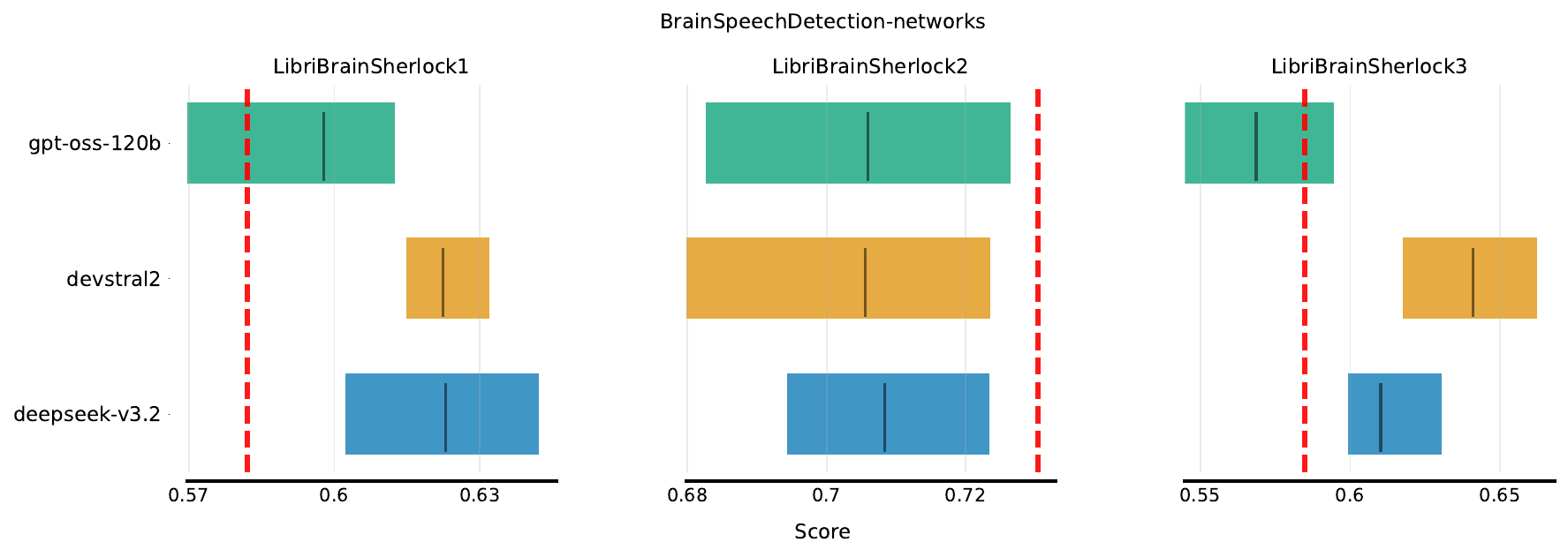}%
\hfill%
\includegraphics[width=0.48\textwidth]{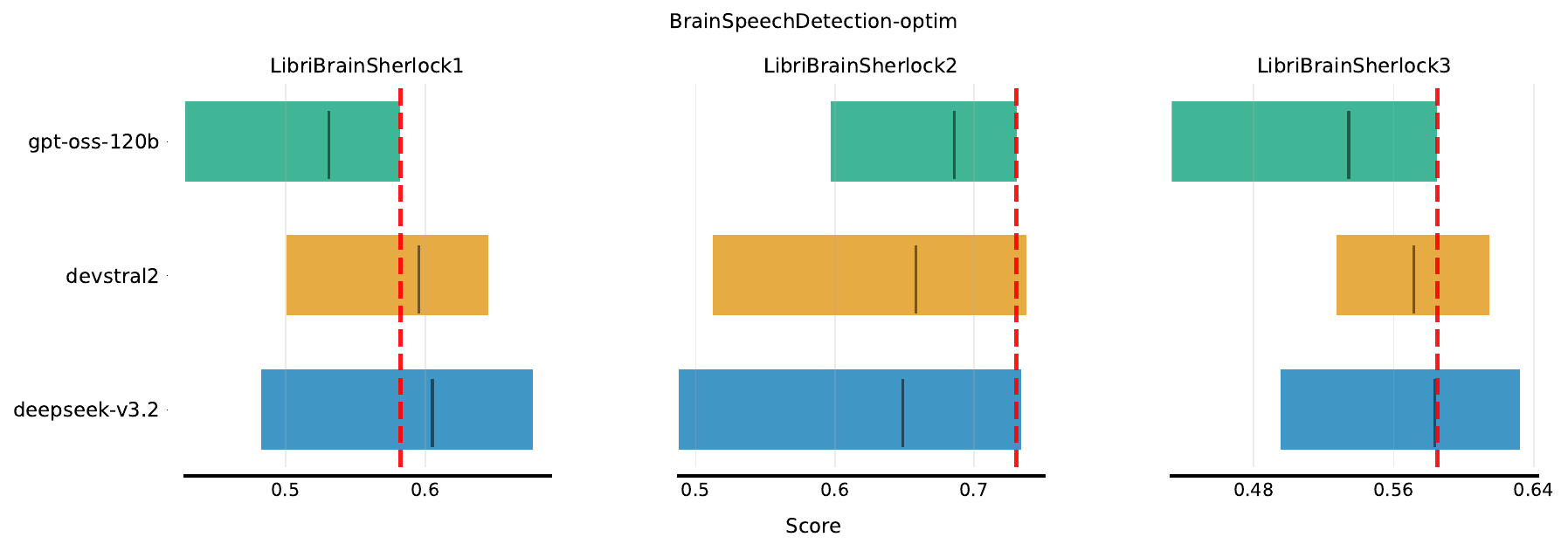}%
\\[0.5em]
\includegraphics[width=0.48\textwidth]{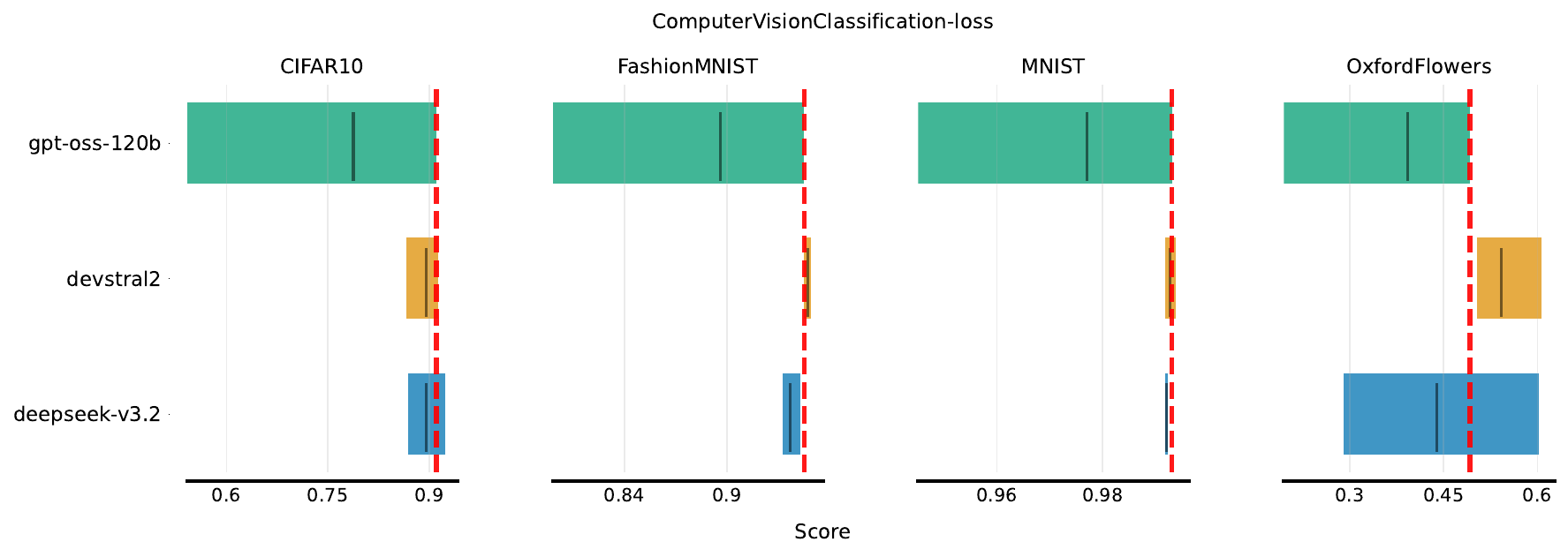}%
\hfill%
\includegraphics[width=0.48\textwidth]{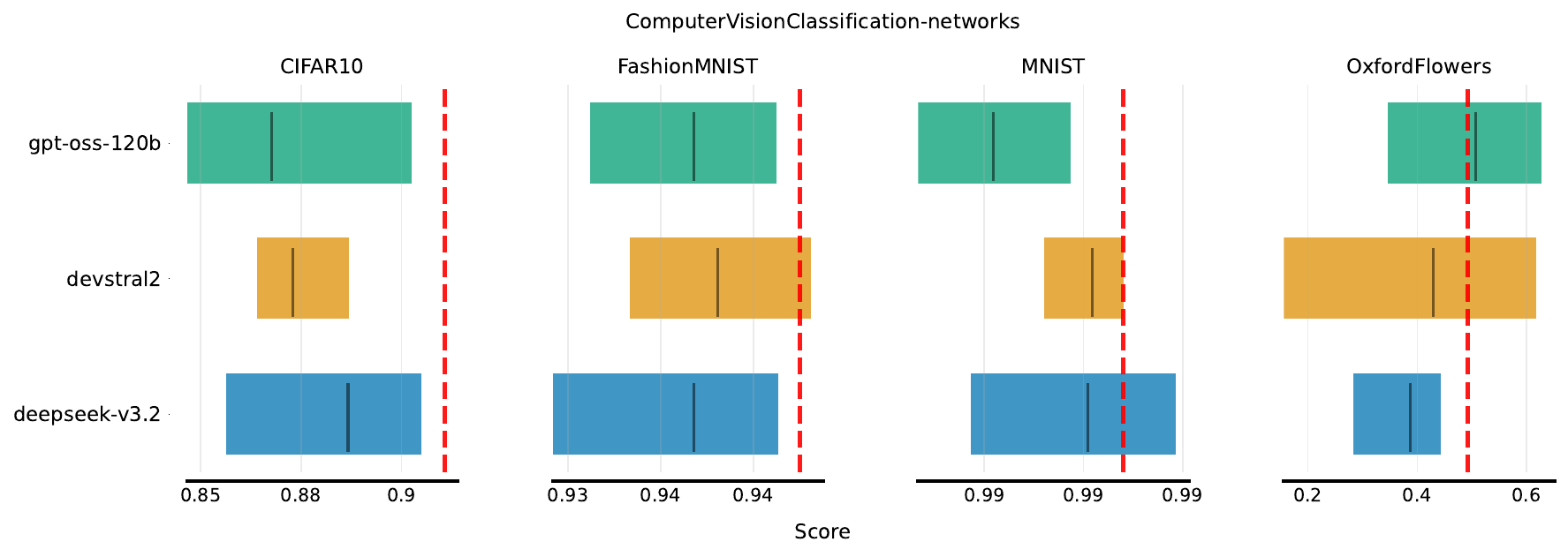}%
\\[0.5em]
\includegraphics[width=0.48\textwidth]{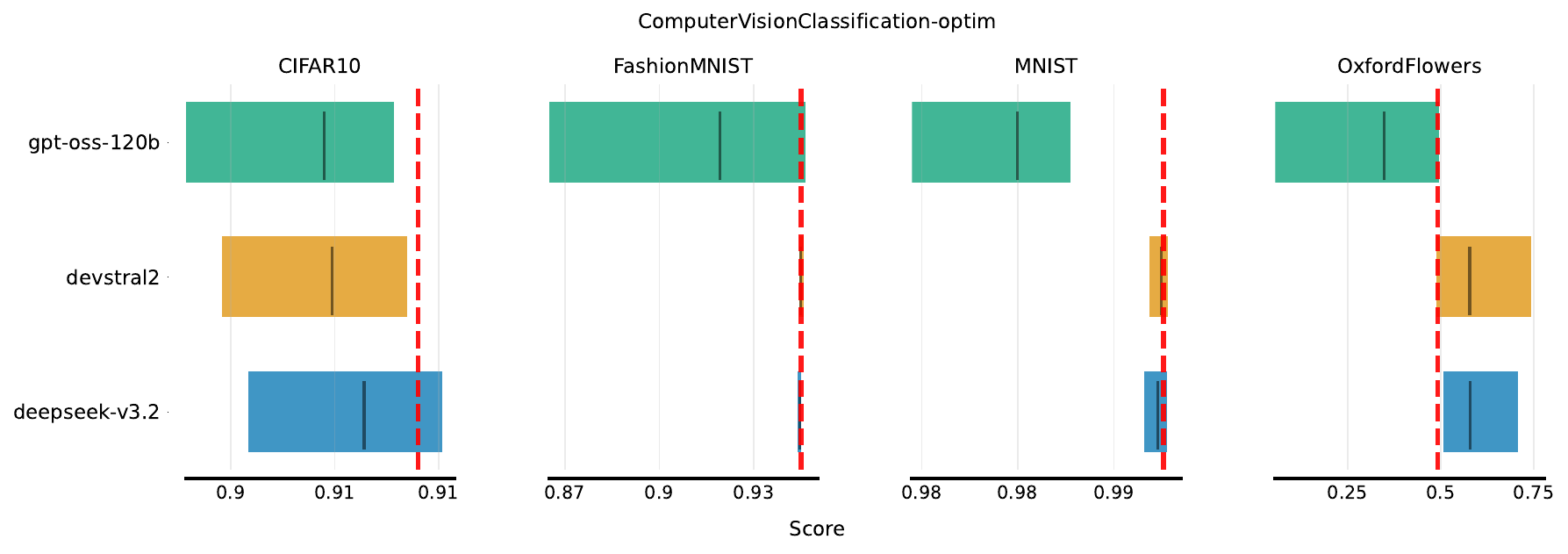}%
\hfill%
\includegraphics[width=0.48\textwidth]{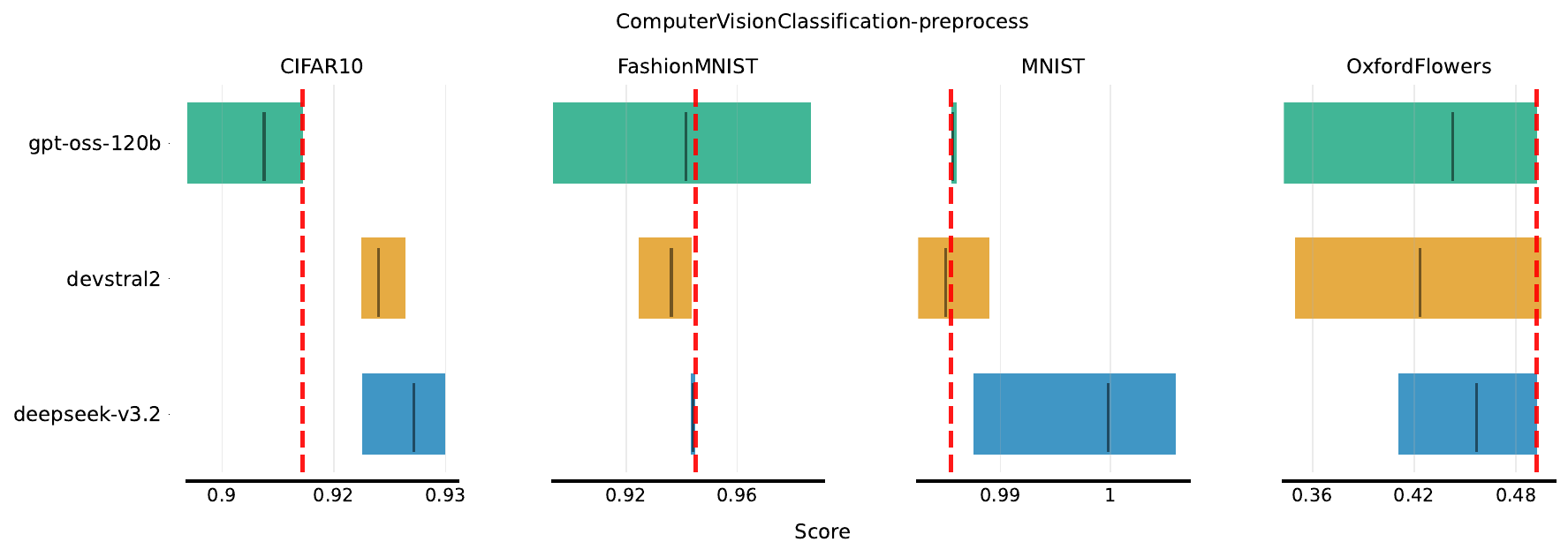}%
\\[0.5em]
\includegraphics[width=0.48\textwidth]{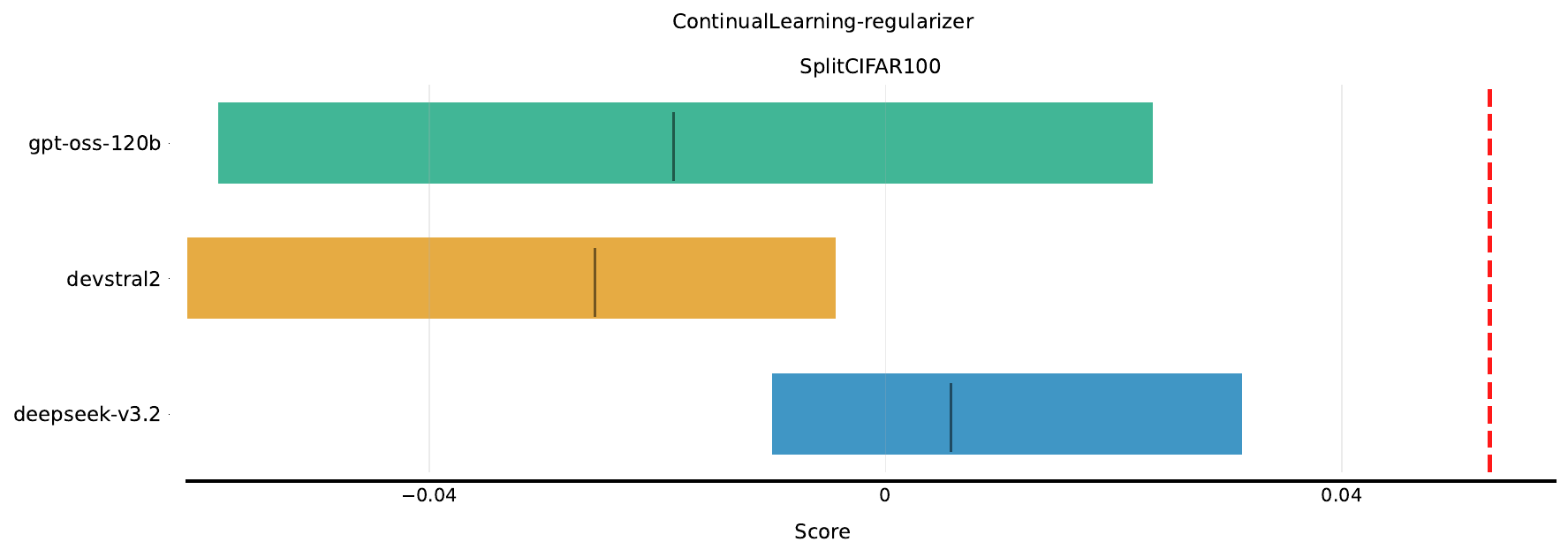}%
\hfill%
\includegraphics[width=0.48\textwidth]{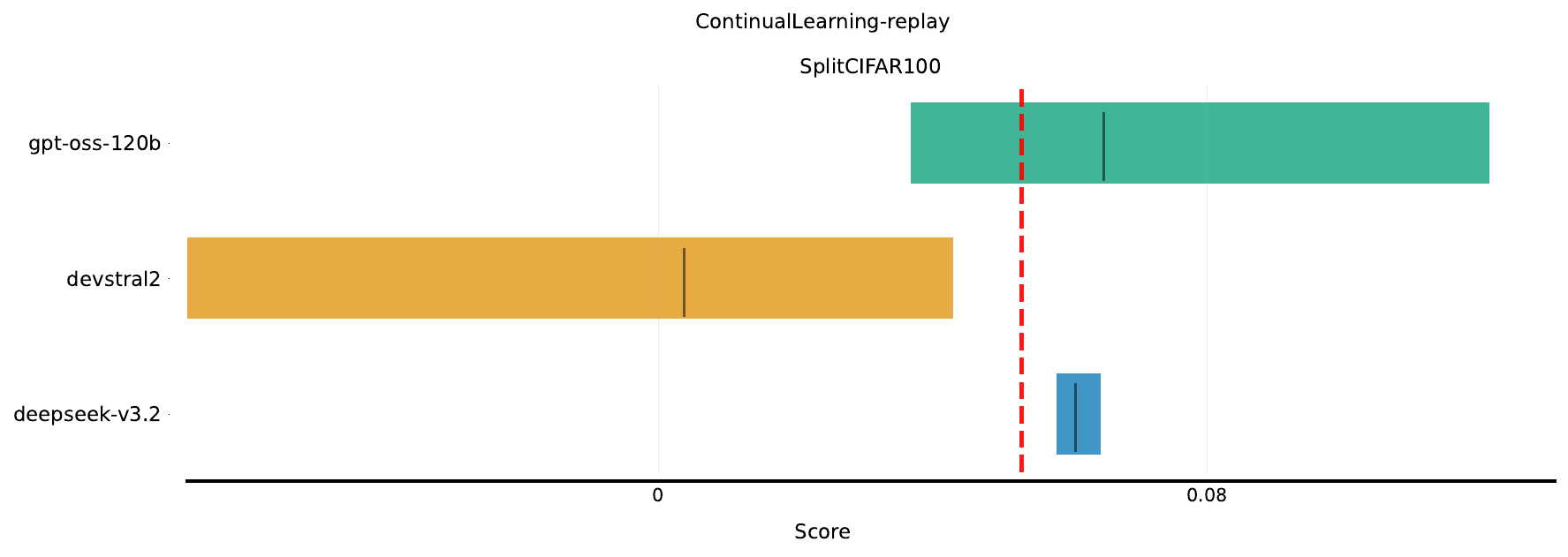}%
\\[0.5em]
\includegraphics[width=0.48\textwidth]{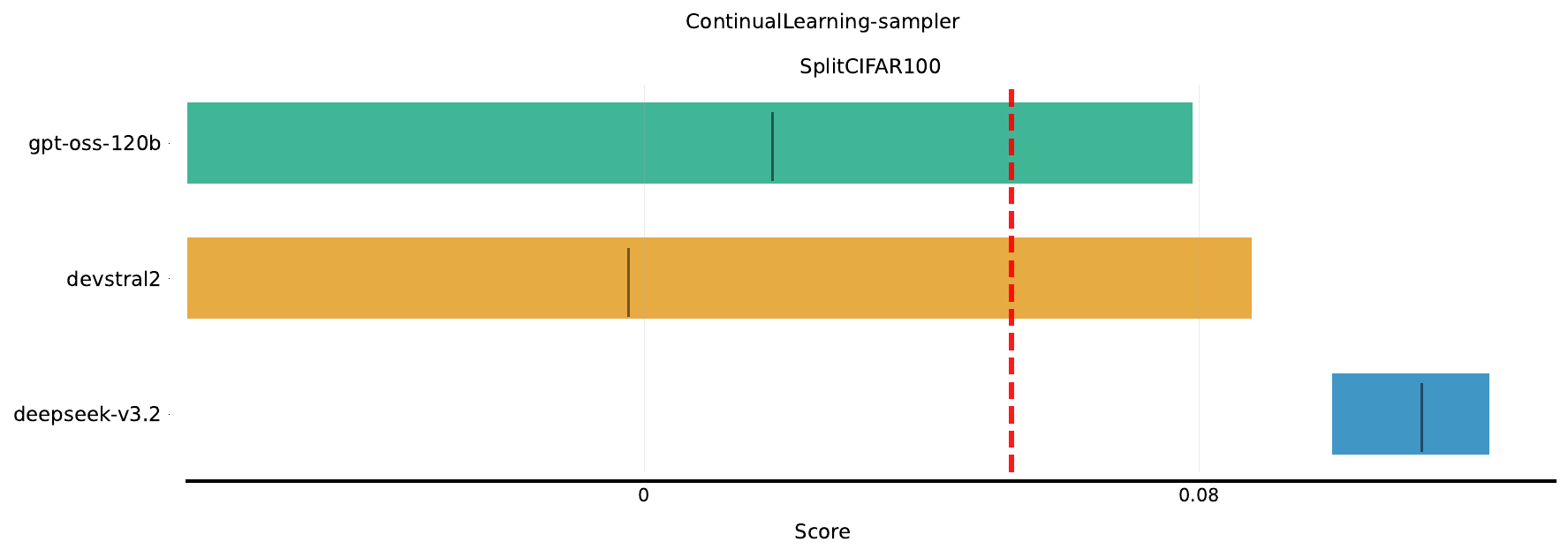}%
\hfill%
\includegraphics[width=0.48\textwidth]{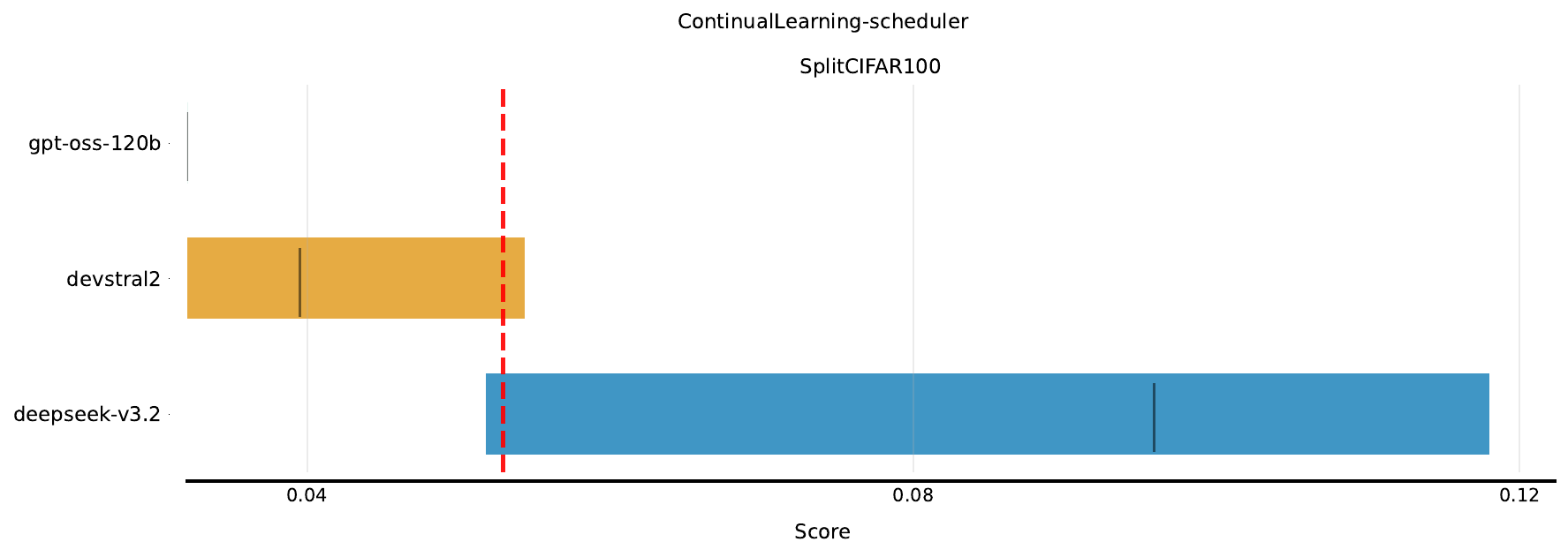}%
\caption{DiscoBench (3 Successful Seeds) results on Meta-Train tasks. (Part 2/4)}
\label{fig:until_success_id_2}
\end{figure}
\clearpage

\begin{figure}[htbp]
\centering
\setlength{\lineskip}{0pt}
\includegraphics[width=0.48\textwidth]{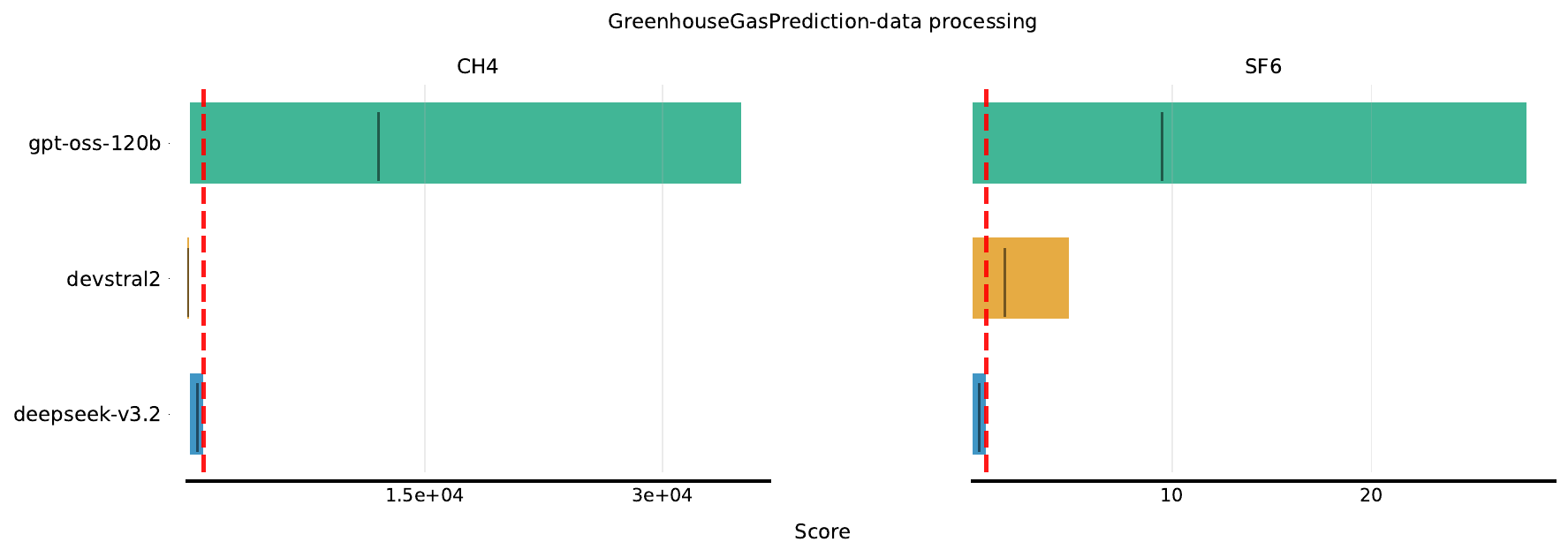}%
\hfill%
\includegraphics[width=0.48\textwidth]{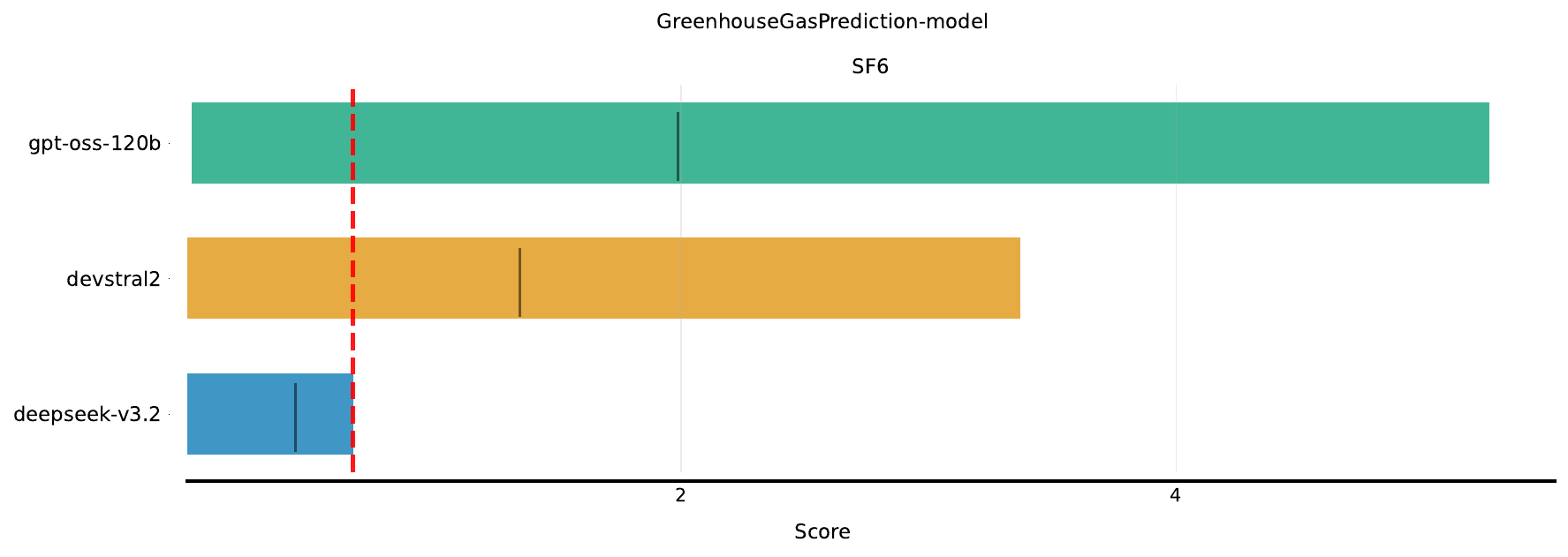}%
\\[0.5em]
\includegraphics[width=0.48\textwidth]{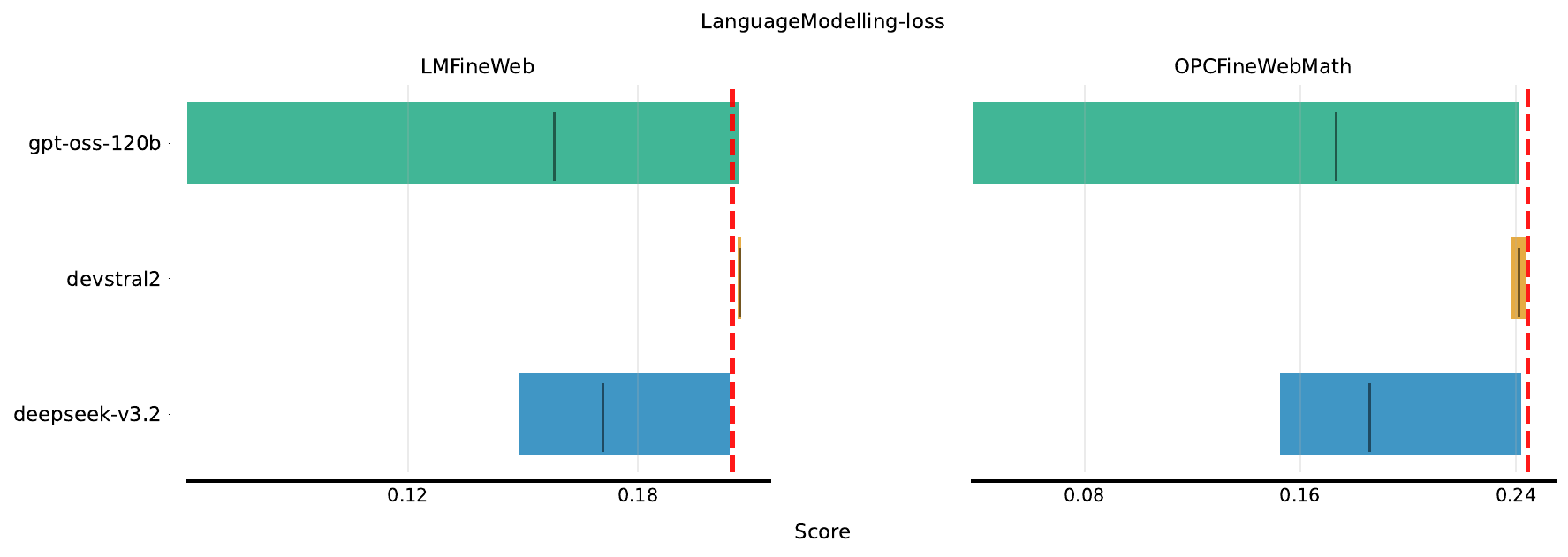}%
\hfill%
\includegraphics[width=0.48\textwidth]{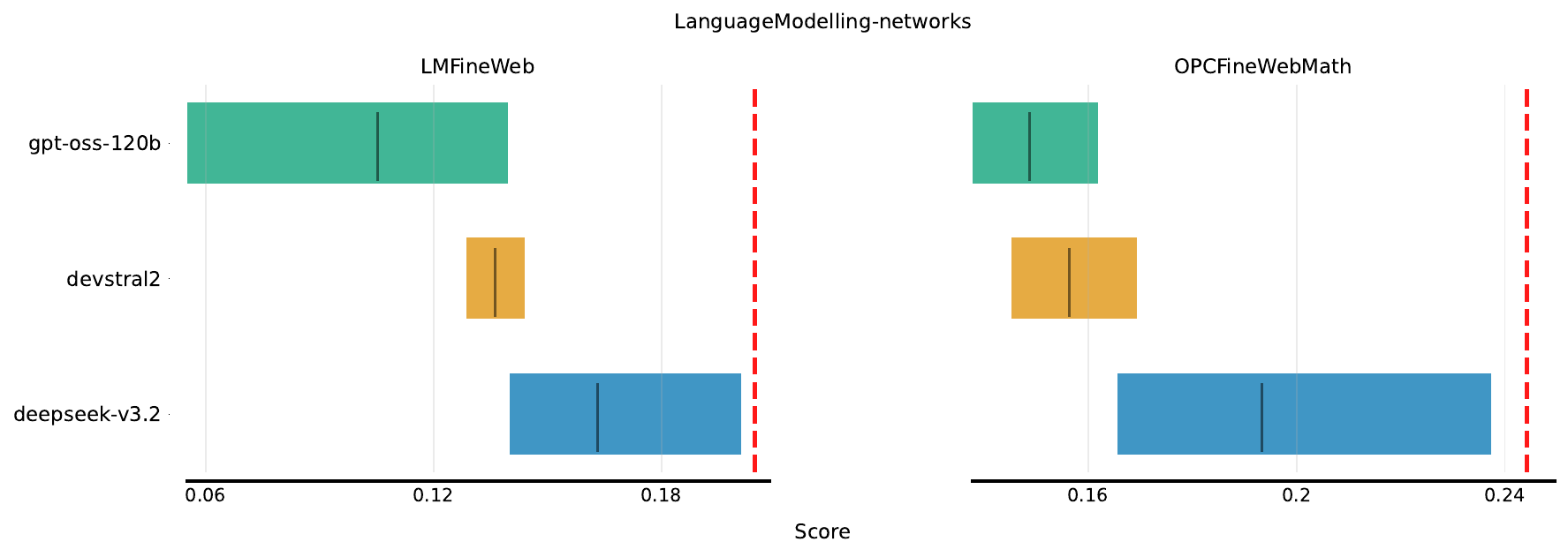}%
\\[0.5em]
\includegraphics[width=0.48\textwidth]{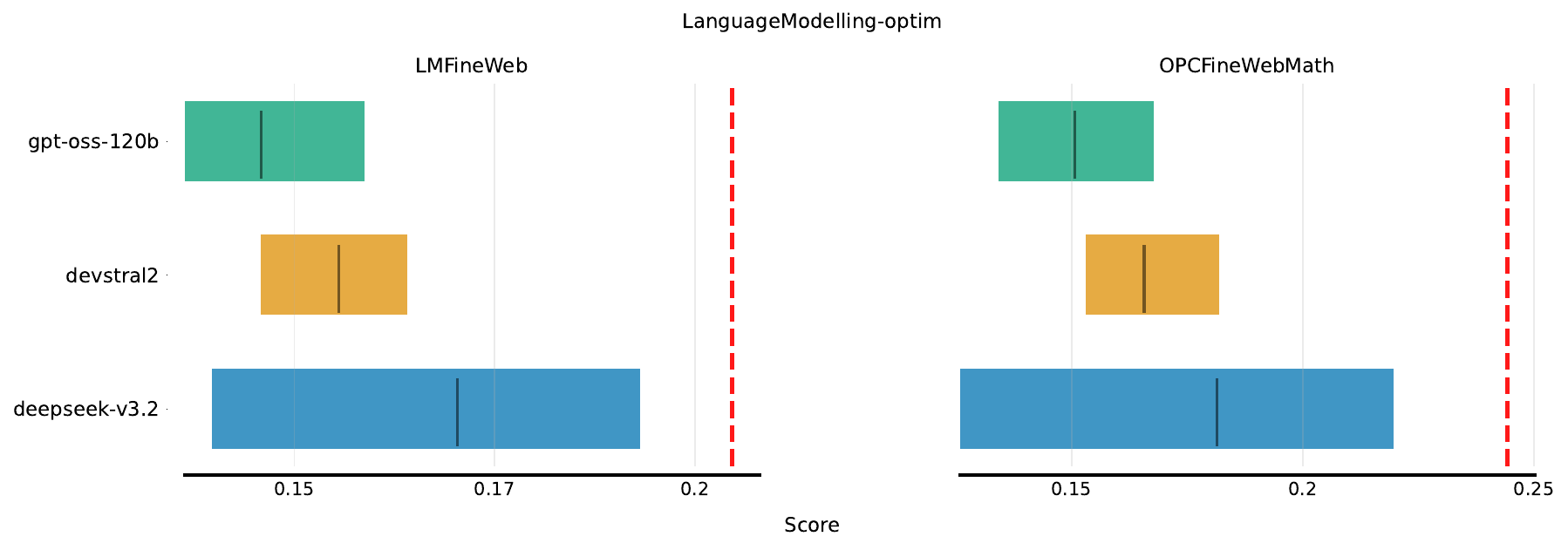}%
\hfill%
\includegraphics[width=0.48\textwidth]{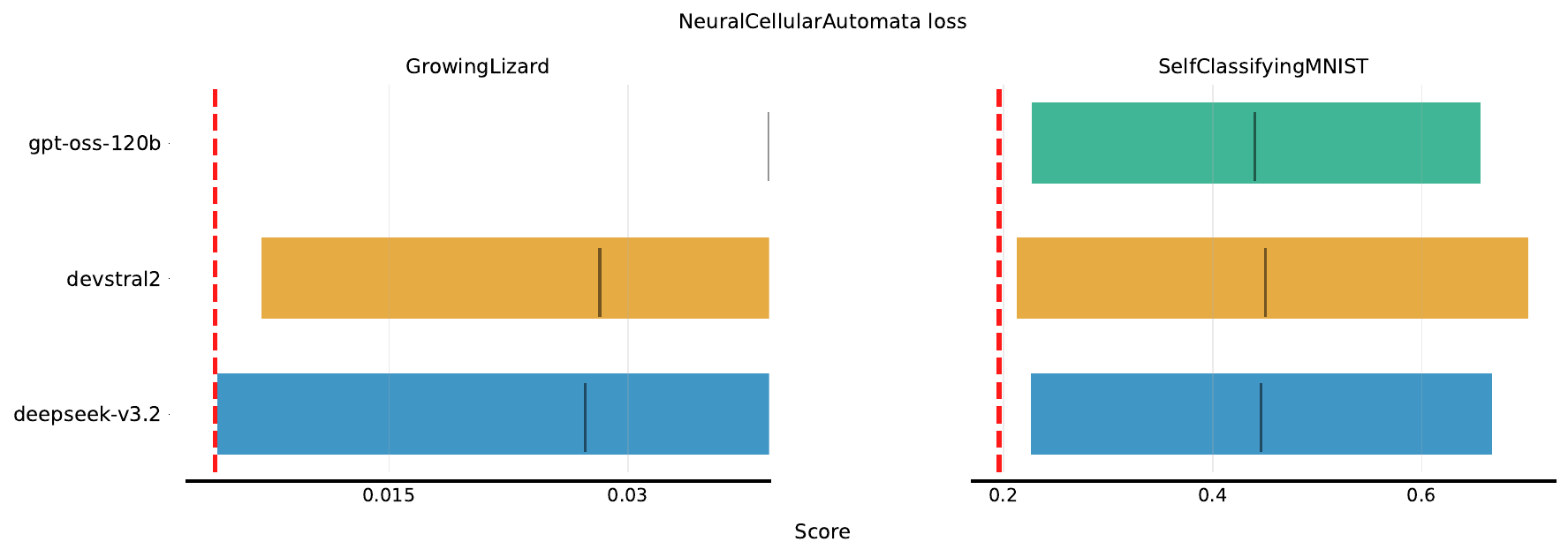}%
\\[0.5em]
\includegraphics[width=0.48\textwidth]{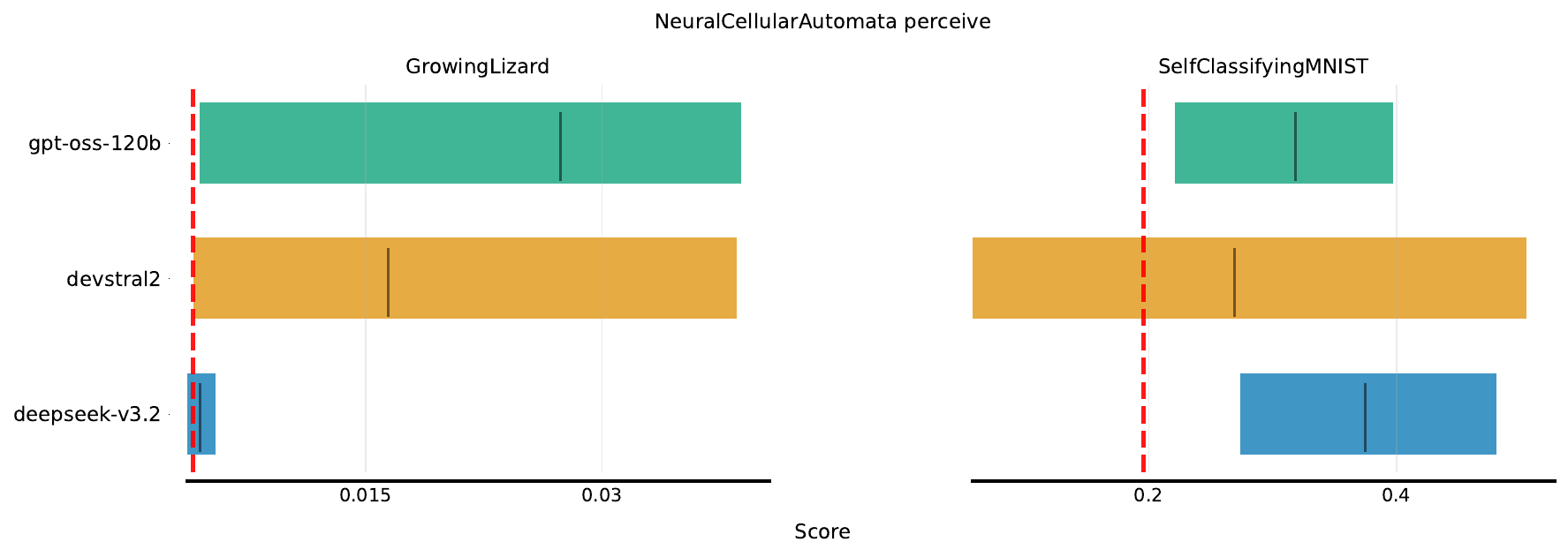}%
\hfill%
\includegraphics[width=0.48\textwidth]{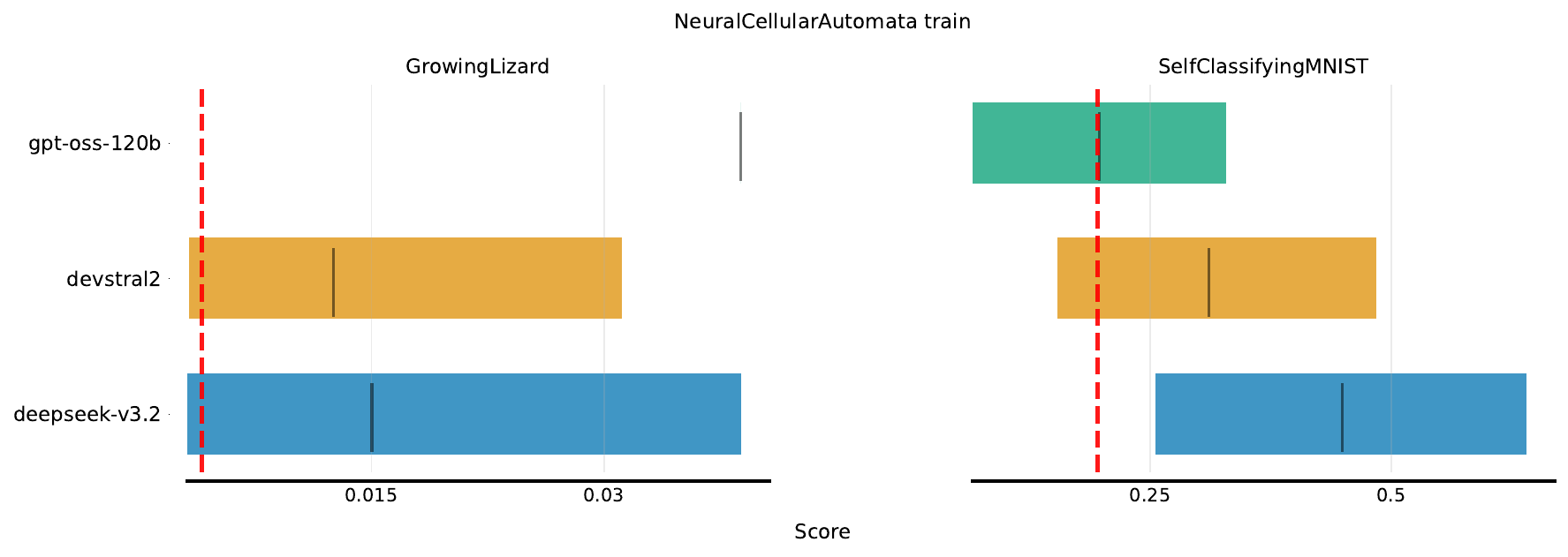}%
\\[0.5em]
\includegraphics[width=0.48\textwidth]{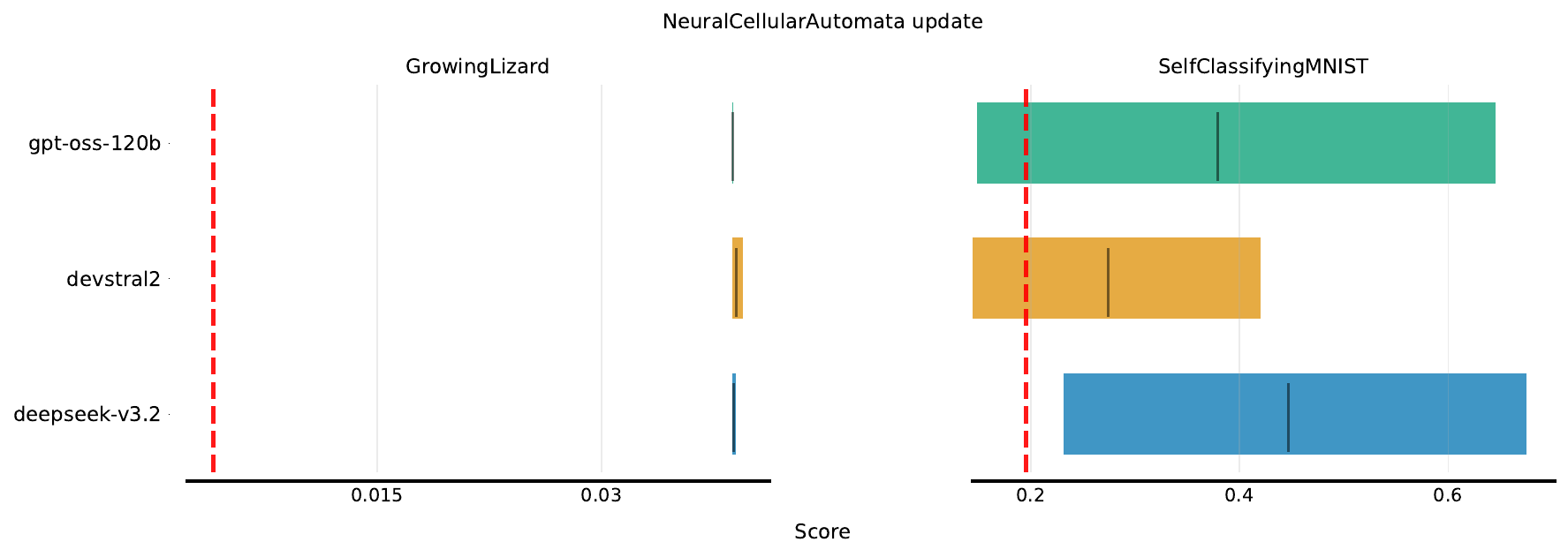}%
\hfill%
\includegraphics[width=0.48\textwidth]{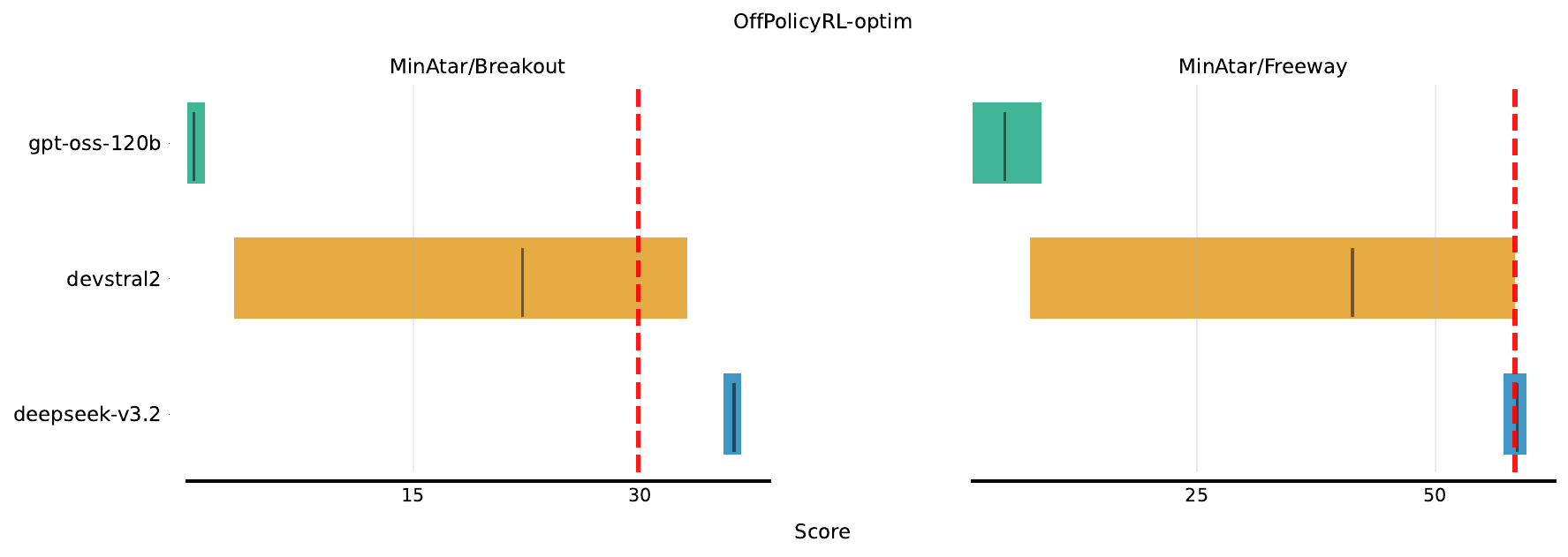}%
\\[0.5em]
\includegraphics[width=0.48\textwidth]{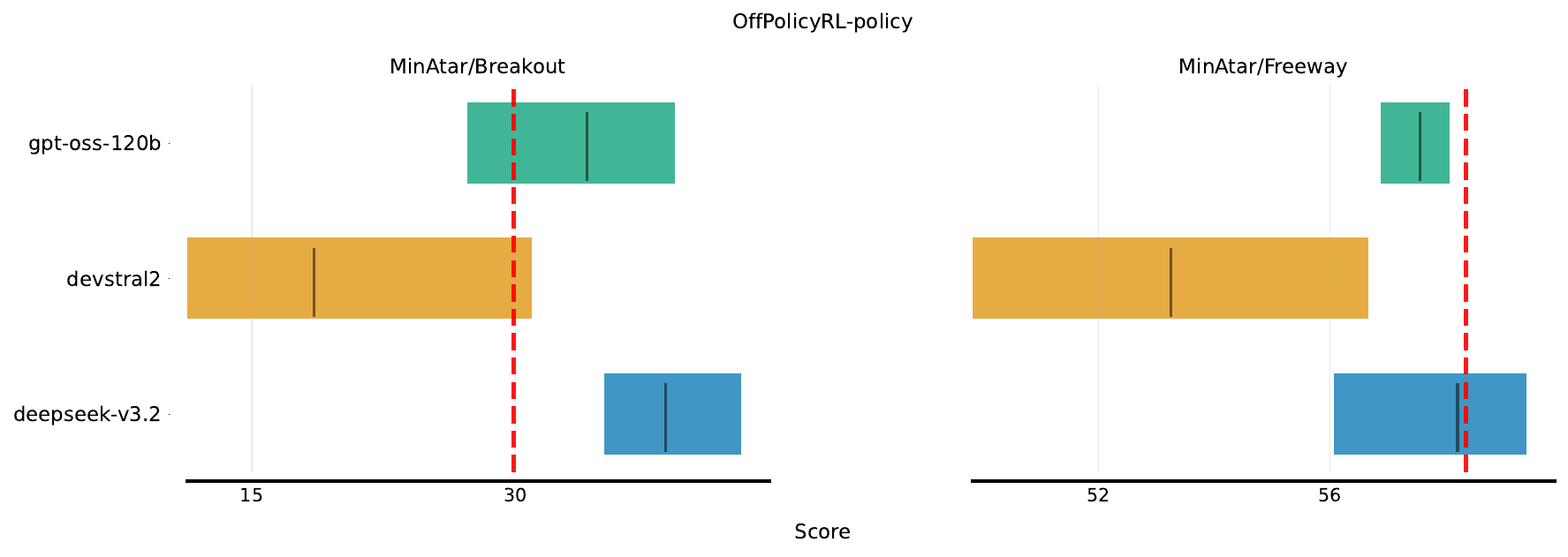}%
\hfill%
\includegraphics[width=0.48\textwidth]{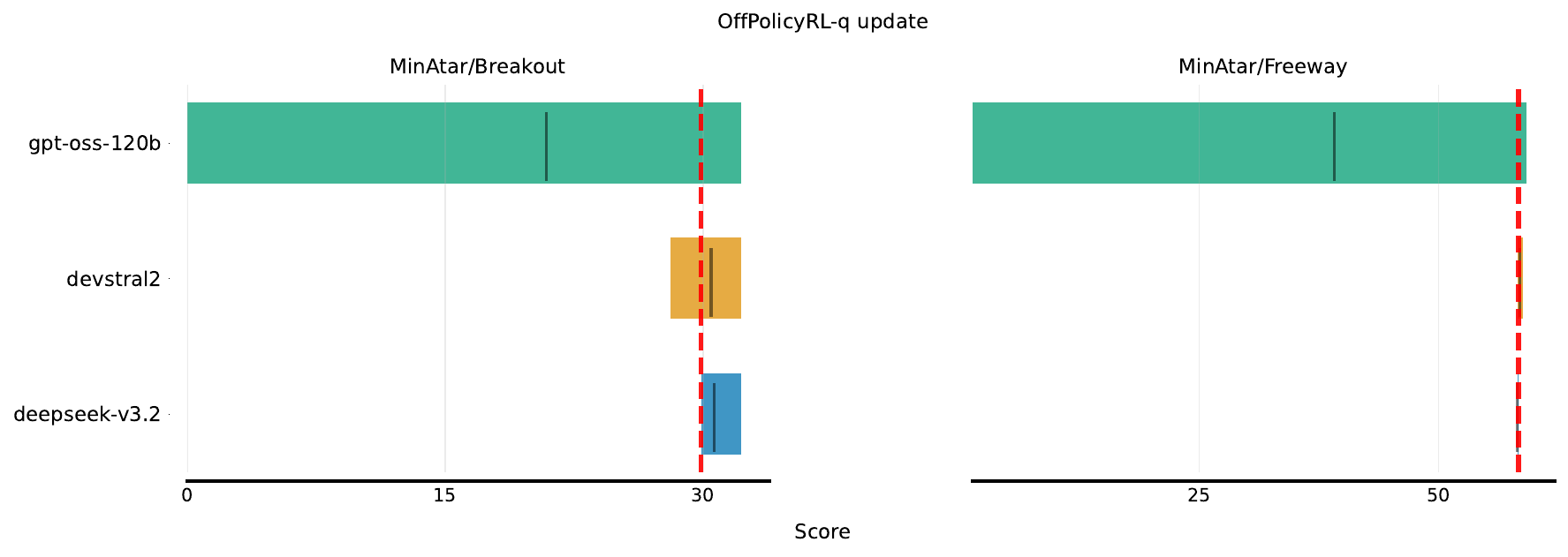}%
\caption{DiscoBench (3 Successful Seeds) results on Meta-Train tasks. (Part 3/4)}
\label{fig:until_success_id_3}
\end{figure}
\clearpage

\begin{figure}[htbp]
\centering
\setlength{\lineskip}{0pt}
\includegraphics[width=0.48\textwidth]{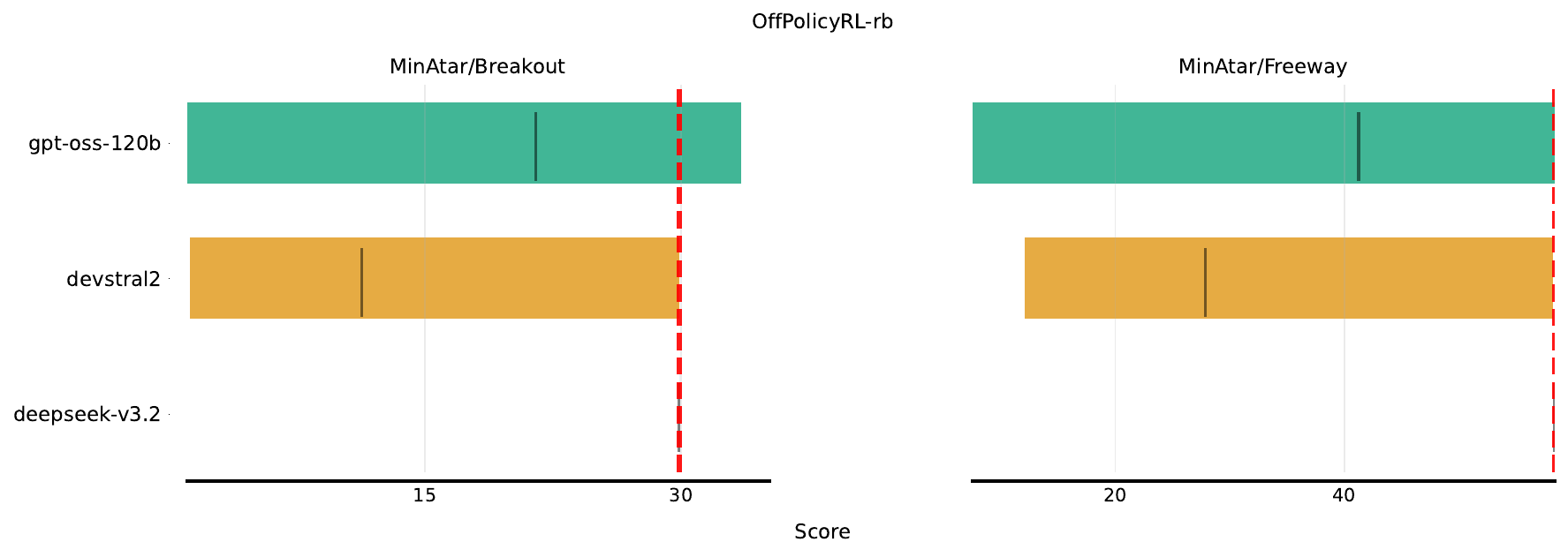}%
\hfill%
\includegraphics[width=0.48\textwidth]{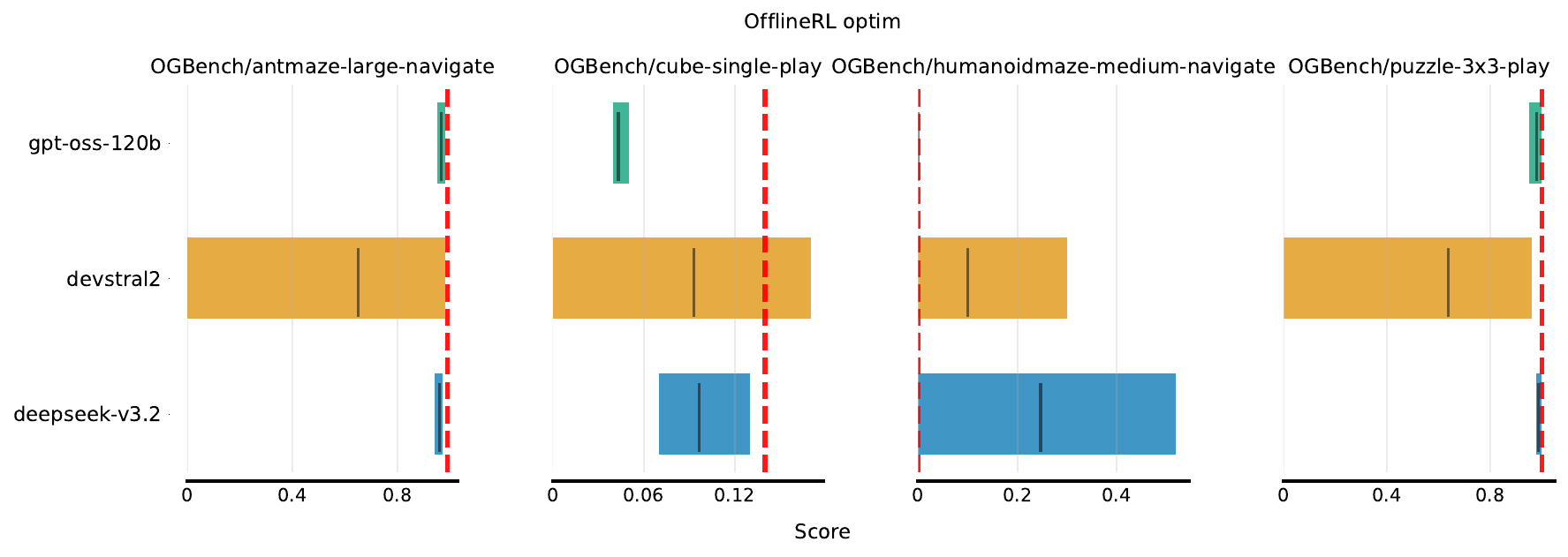}%
\\[0.5em]
\includegraphics[width=0.48\textwidth]{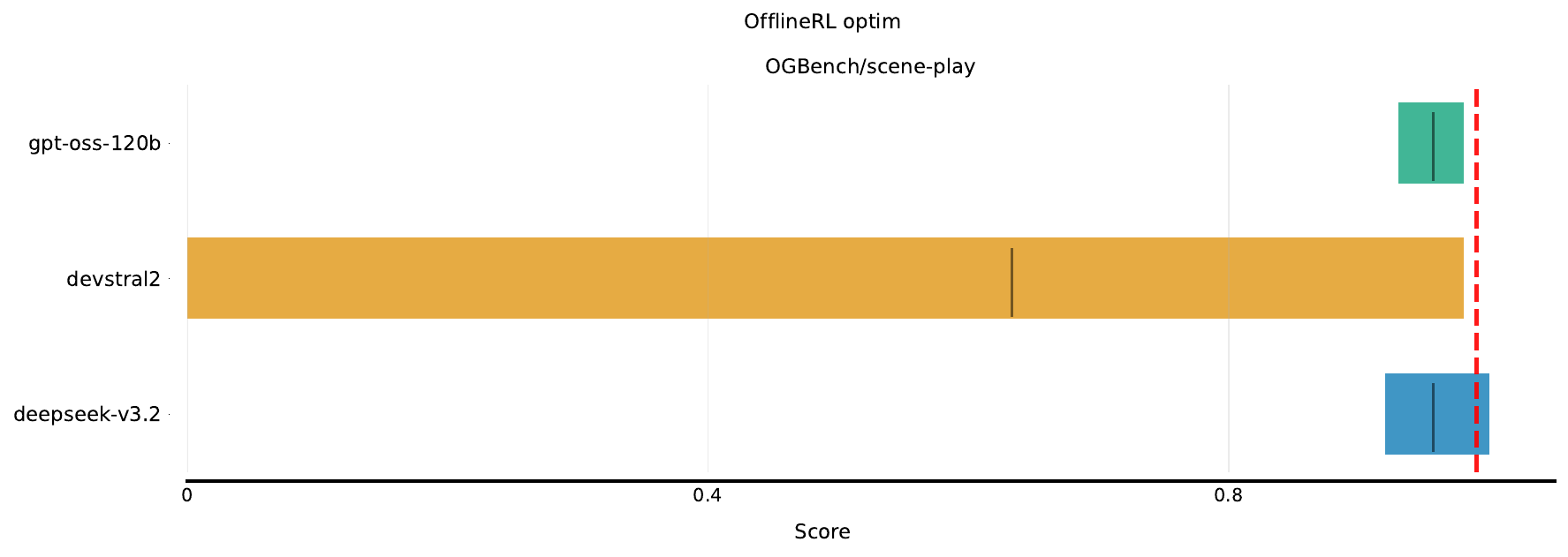}%
\hfill%
\includegraphics[width=0.48\textwidth]{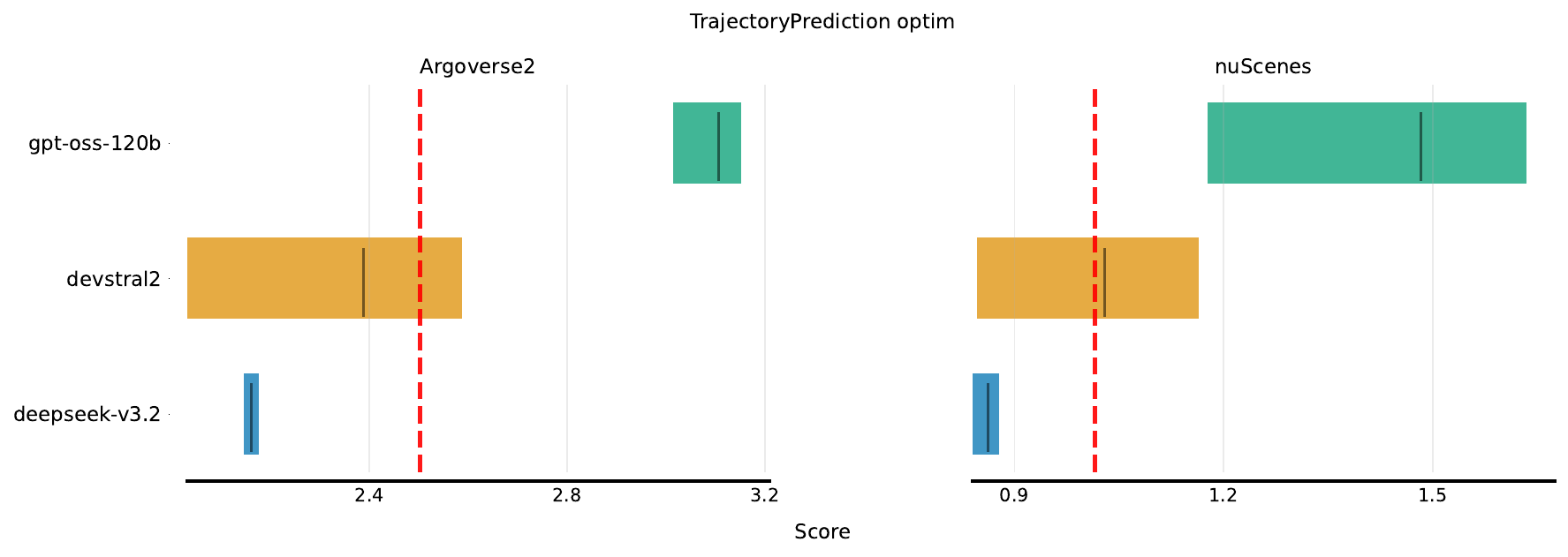}%
\\[0.5em]
\includegraphics[width=0.48\textwidth]{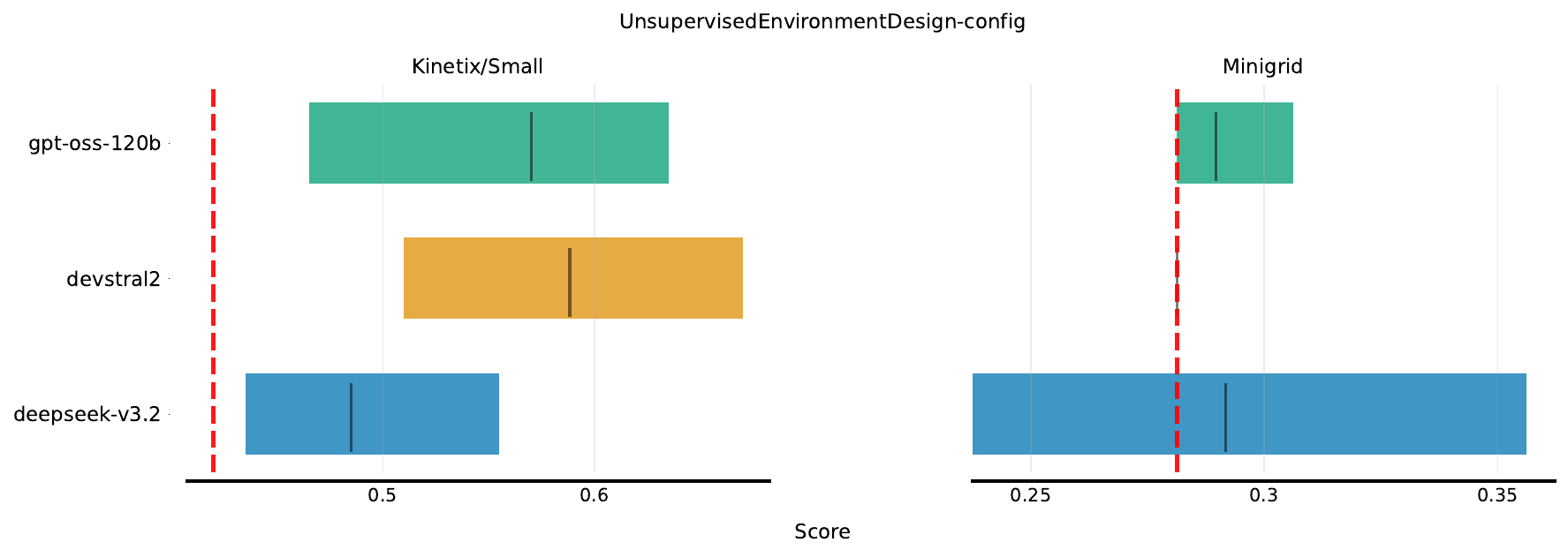}%
\hfill%
\includegraphics[width=0.48\textwidth]{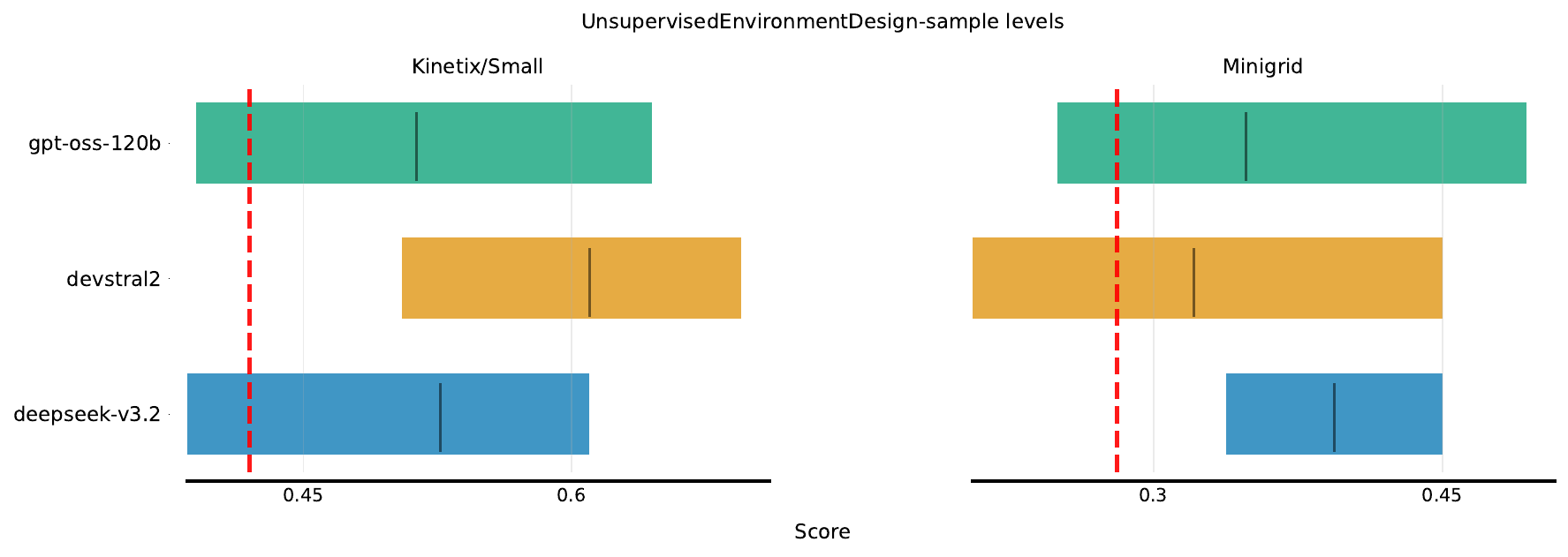}%
\\[0.5em]
\includegraphics[width=0.48\textwidth]{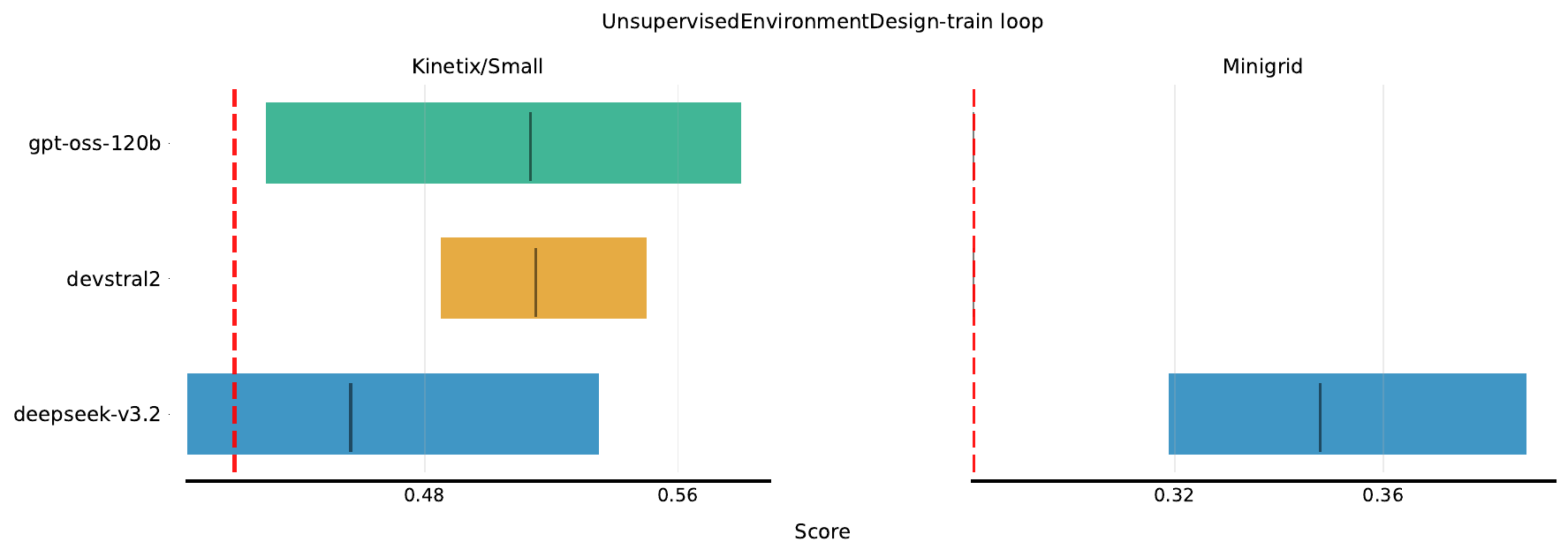}%
\hfill%
\includegraphics[width=0.48\textwidth]{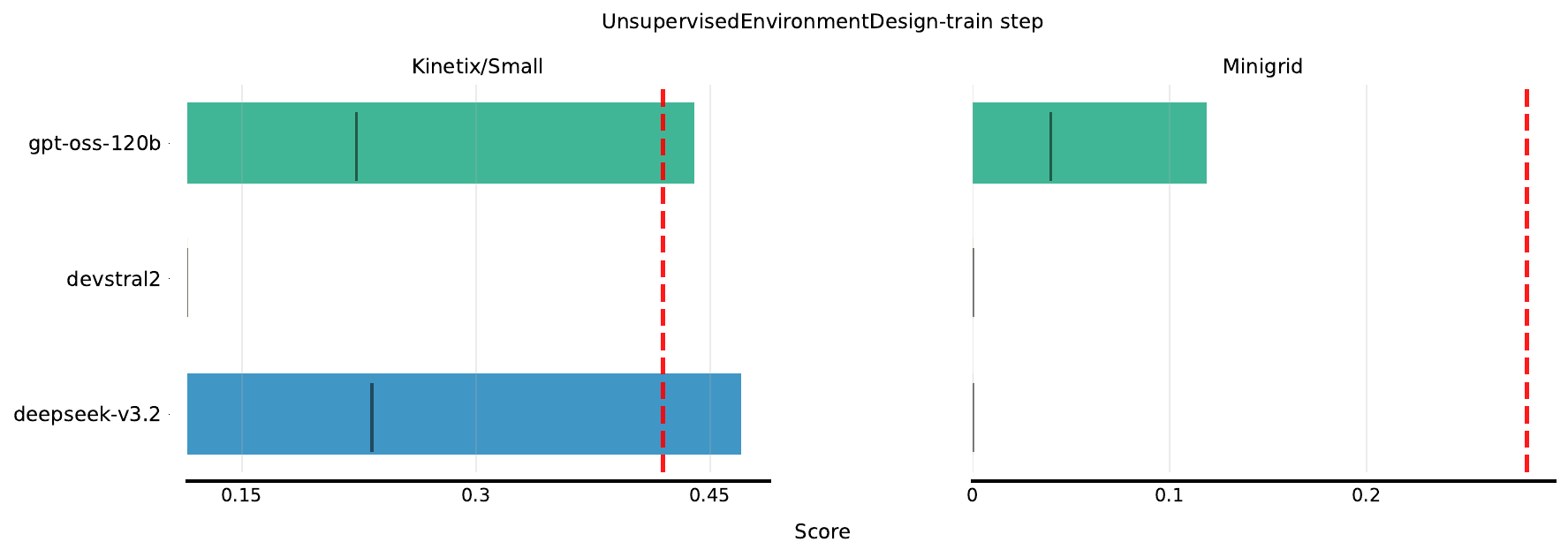}%
\caption{DiscoBench (3 Successful Seeds) results on Meta-Train tasks. (Part 4/4)}
\label{fig:until_success_id_4}
\end{figure}
\clearpage

\subsection{DiscoBench (3 Successful Seeds) -- Meta-Test}
\label{sec:until_success_mt}

\begin{figure}[htbp]
\centering
\setlength{\lineskip}{0pt}
\includegraphics[width=0.48\textwidth]{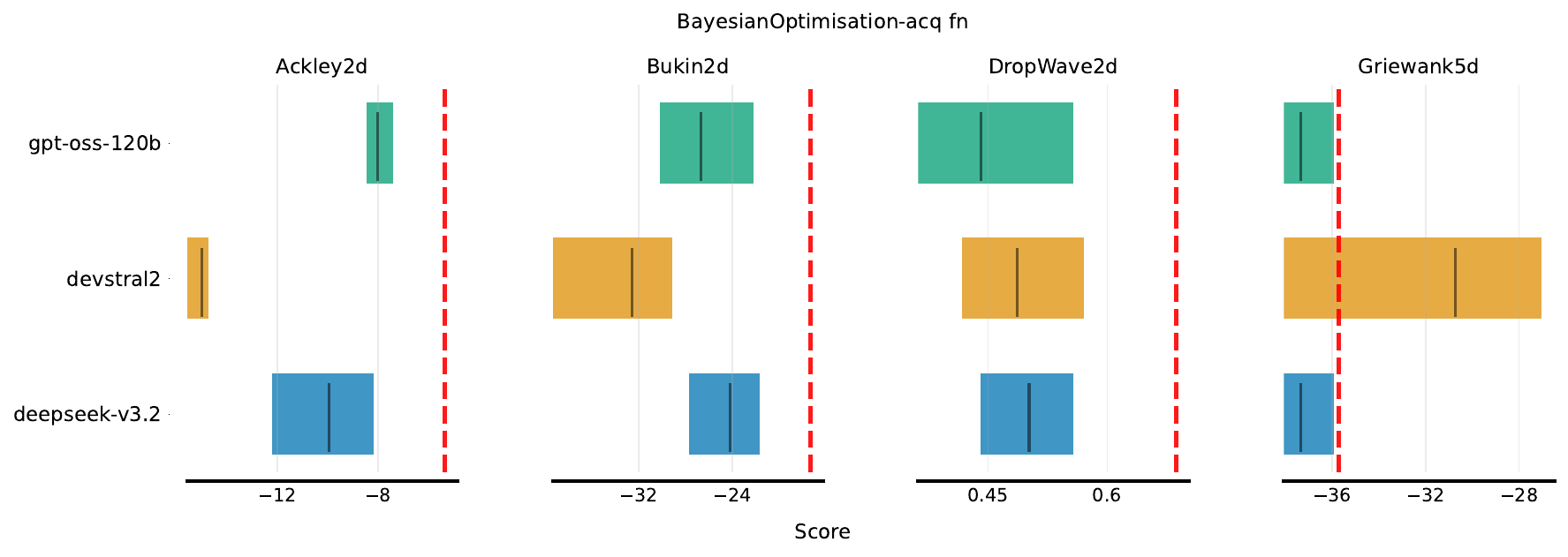}%
\hfill%
\includegraphics[width=0.48\textwidth]{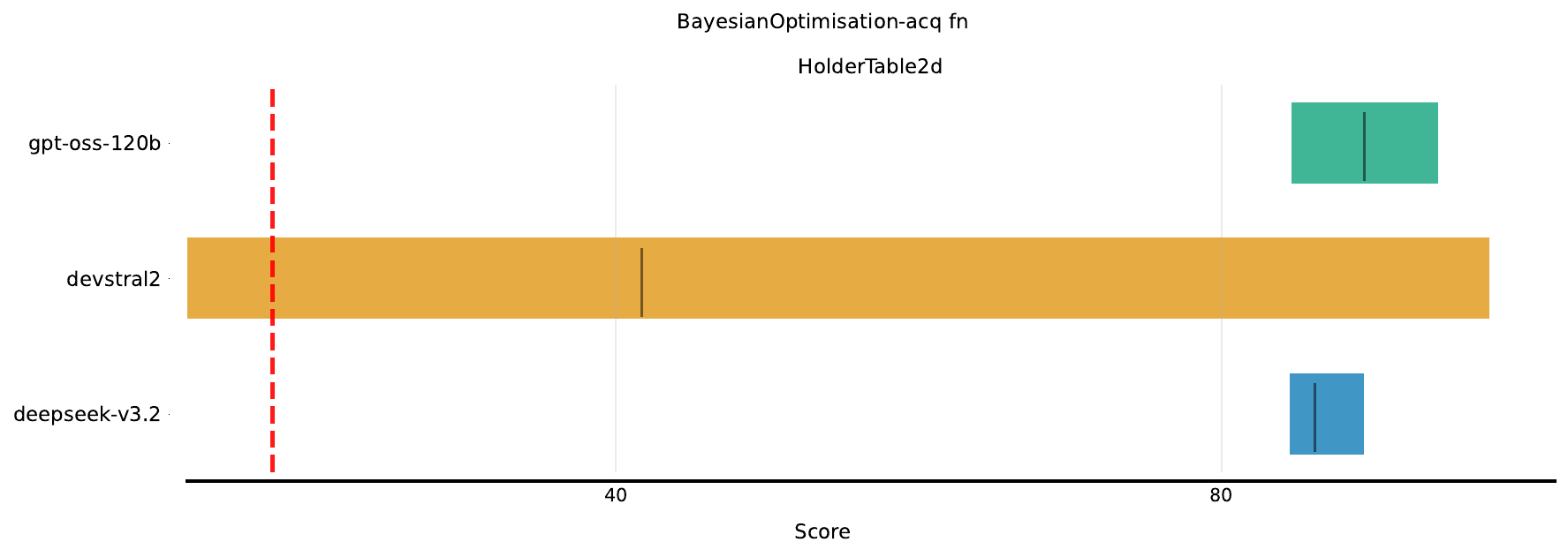}%
\\[0.5em]
\includegraphics[width=0.48\textwidth]{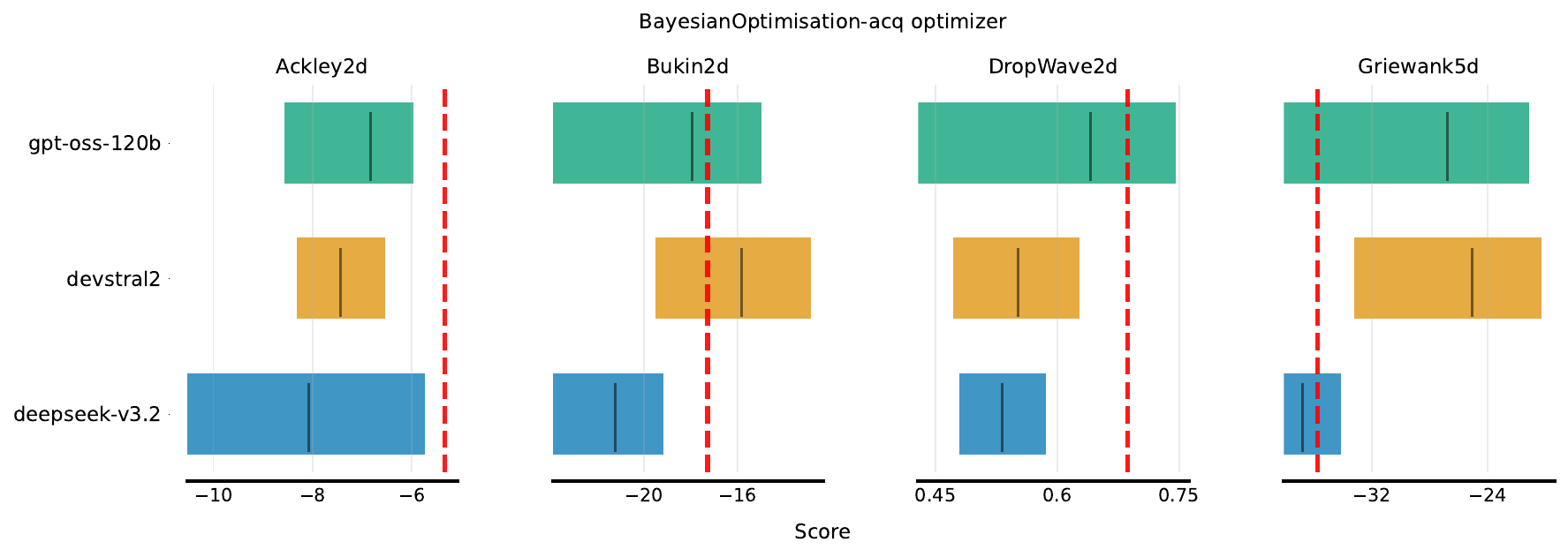}%
\hfill%
\includegraphics[width=0.48\textwidth]{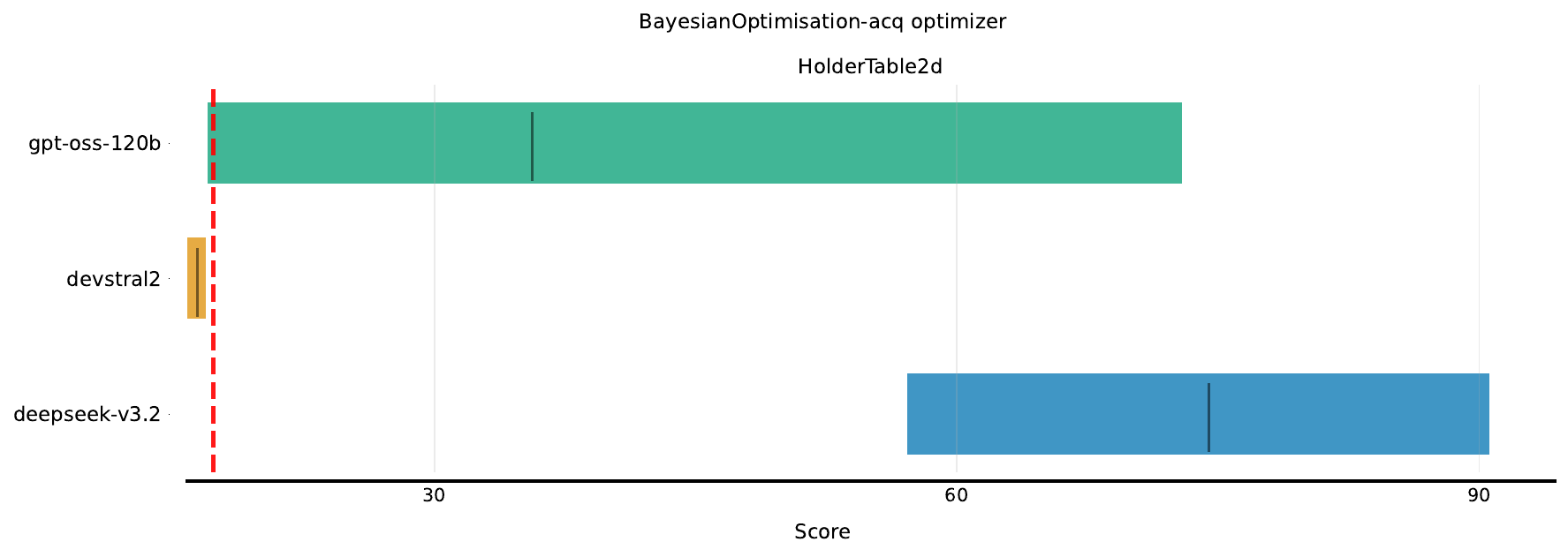}%
\\[0.5em]
\includegraphics[width=0.48\textwidth]{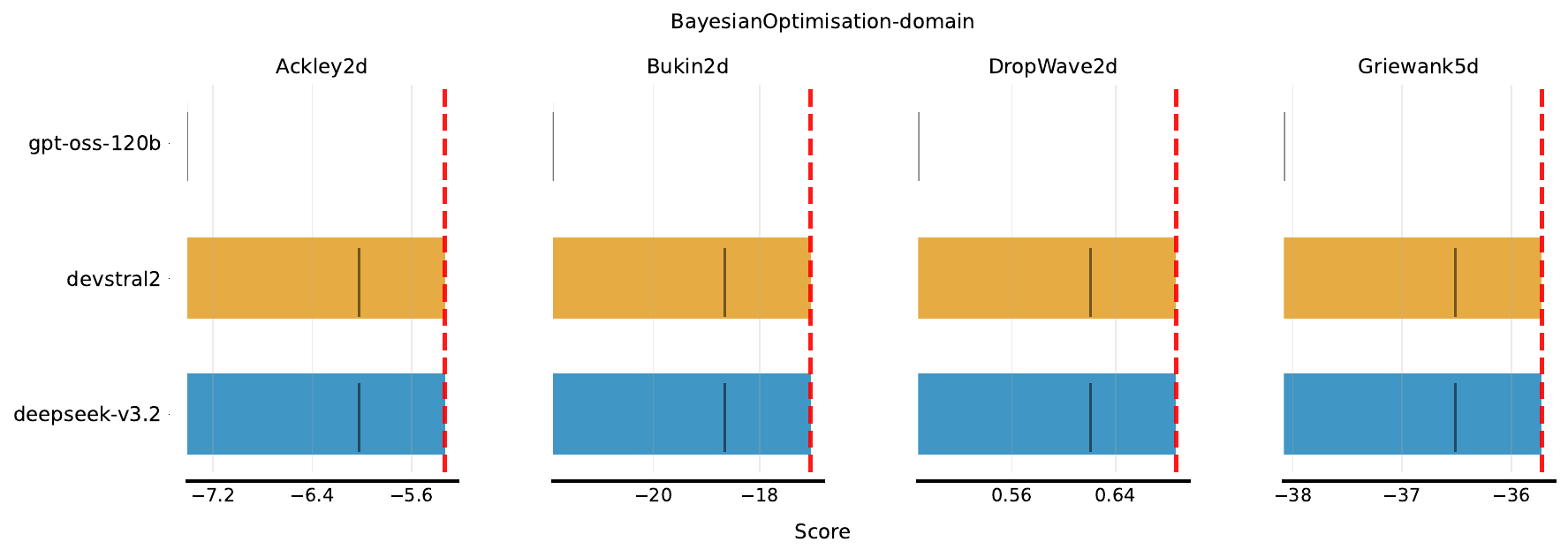}%
\hfill%
\includegraphics[width=0.48\textwidth]{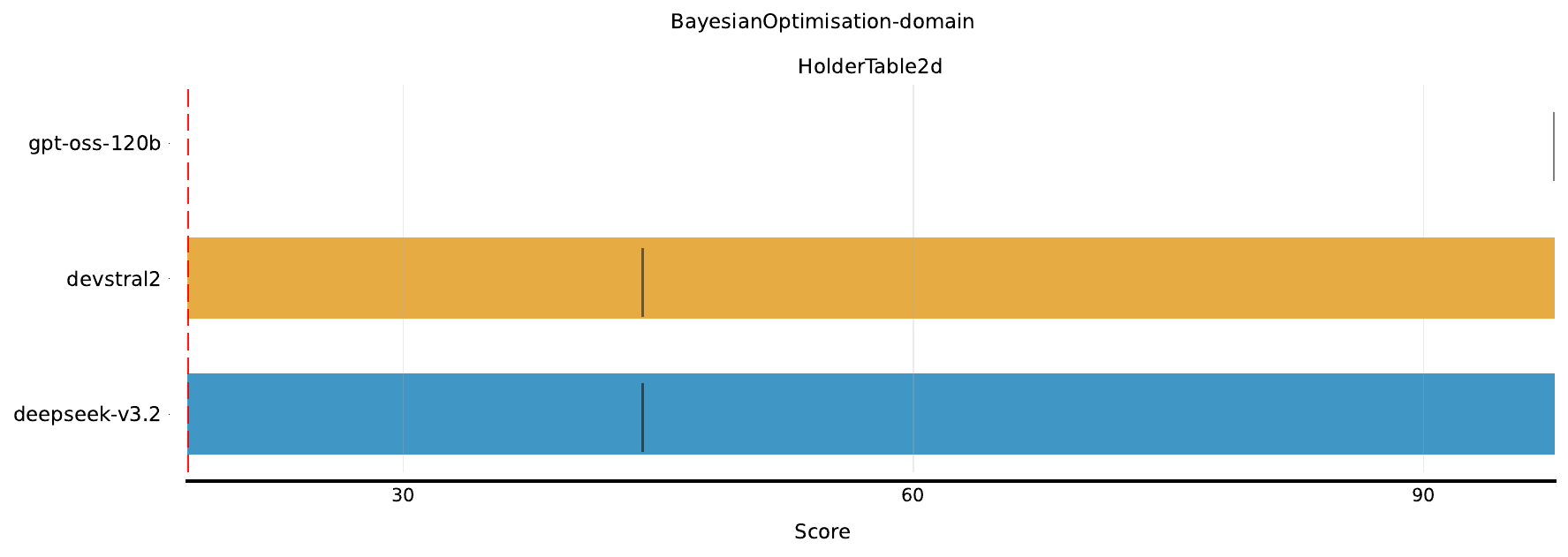}%
\\[0.5em]
\includegraphics[width=0.48\textwidth]{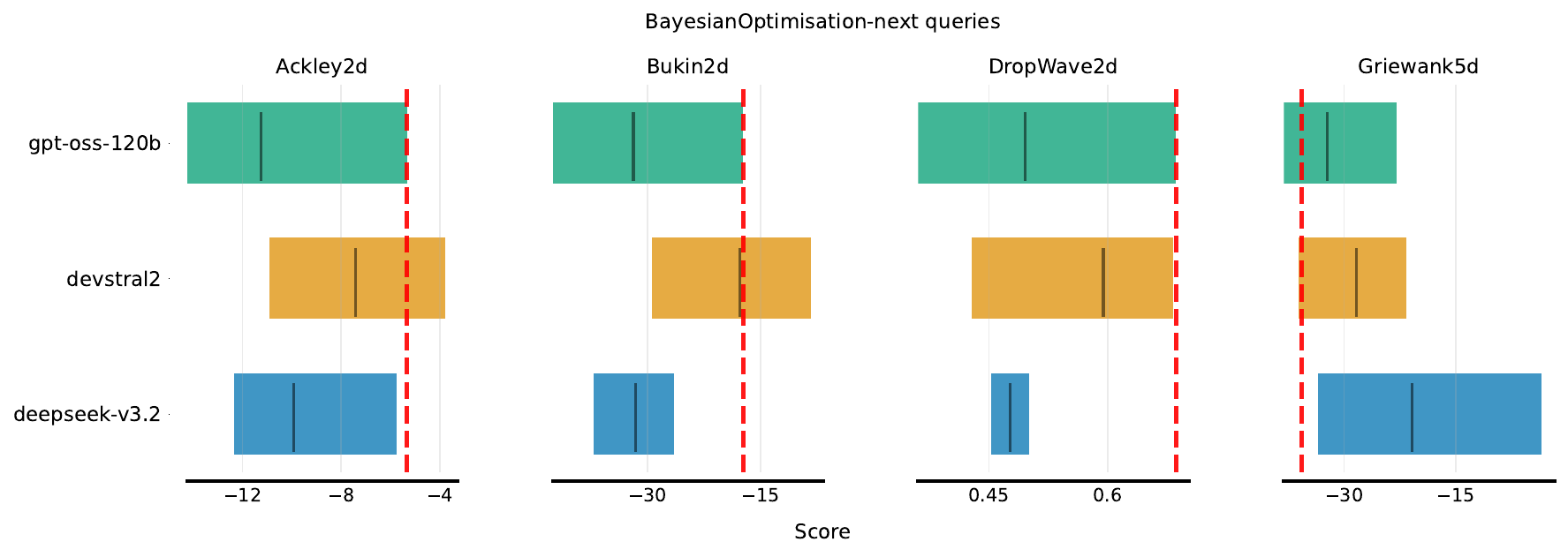}%
\hfill%
\includegraphics[width=0.48\textwidth]{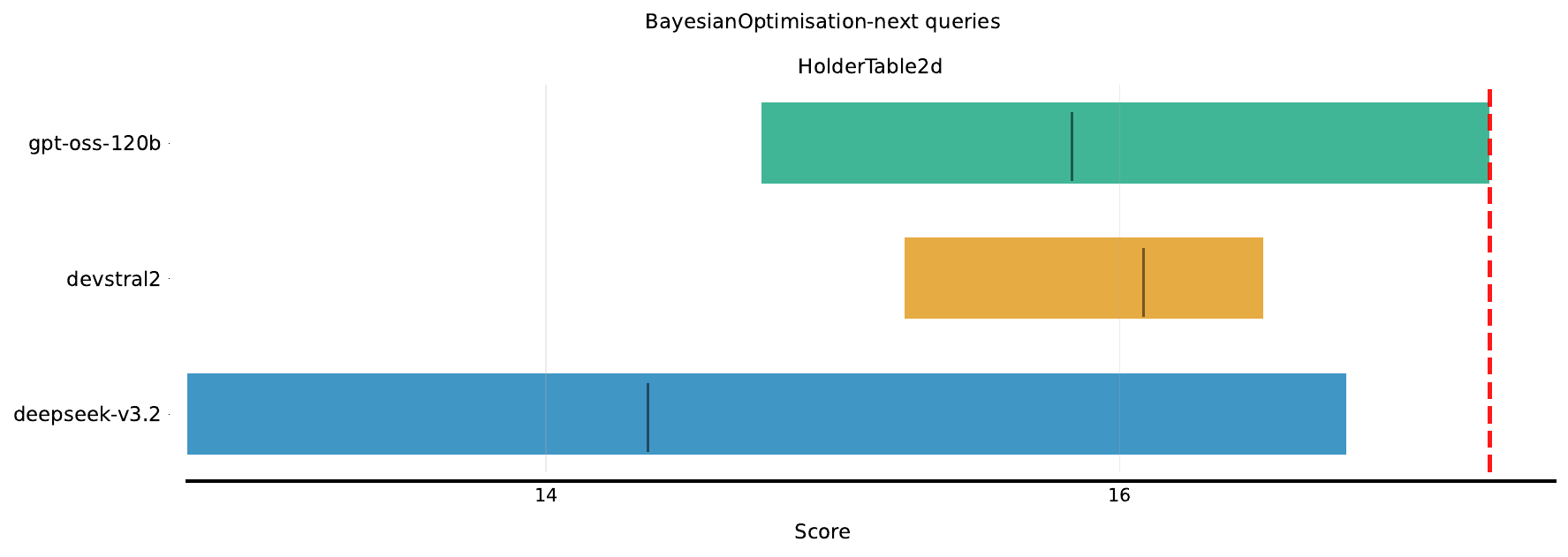}%
\\[0.5em]
\includegraphics[width=0.48\textwidth]{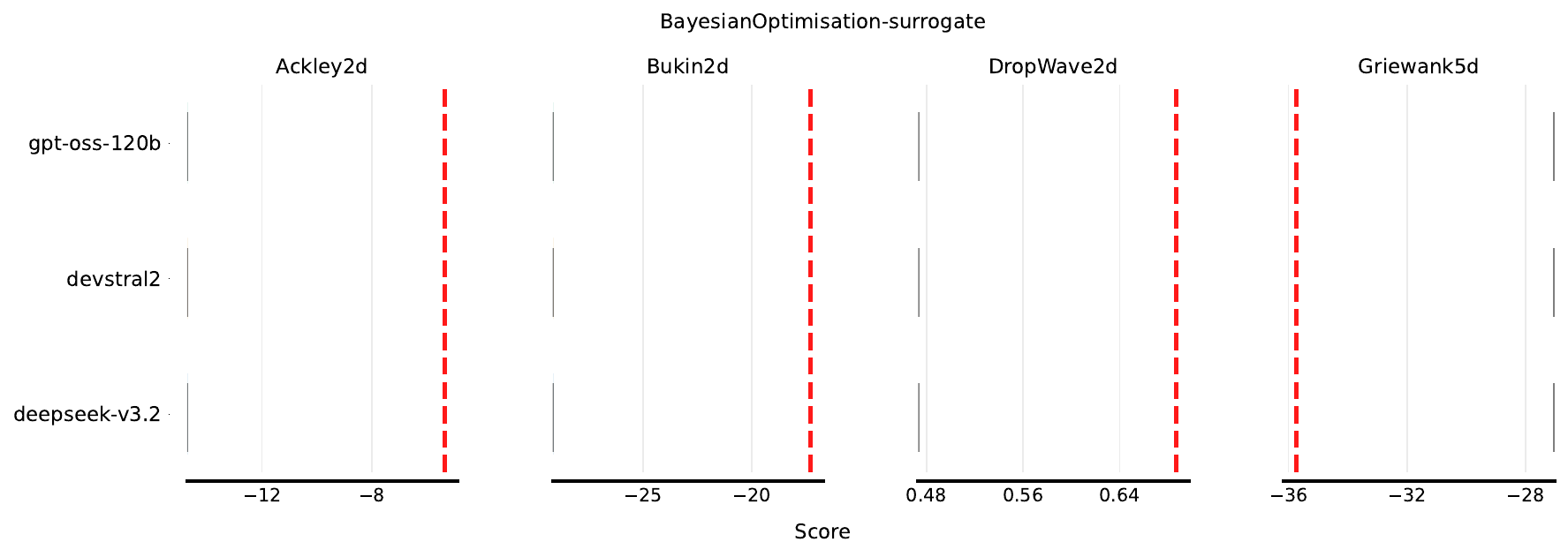}%
\hfill%
\includegraphics[width=0.48\textwidth]{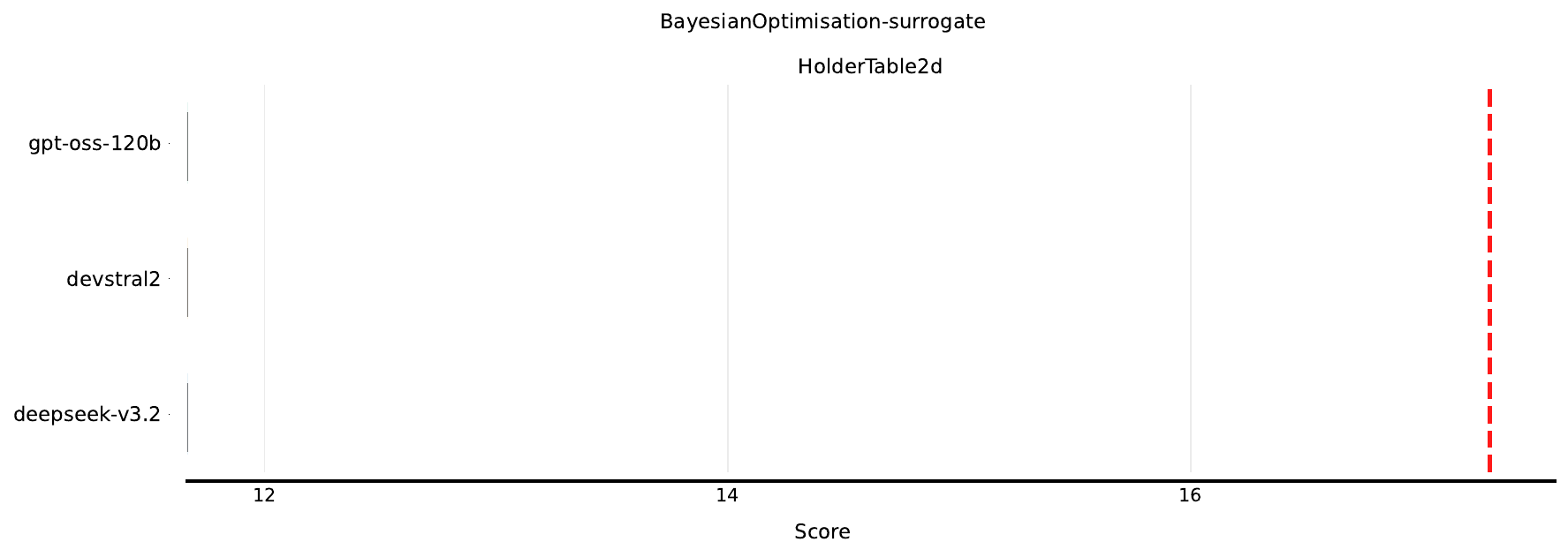}%
\caption{DiscoBench (3 Successful Seeds) results on Meta-Test tasks. (Part 1/5)}
\label{fig:until_success_mt_1}
\end{figure}
\clearpage

\begin{figure}[htbp]
\centering
\setlength{\lineskip}{0pt}
\includegraphics[width=0.48\textwidth]{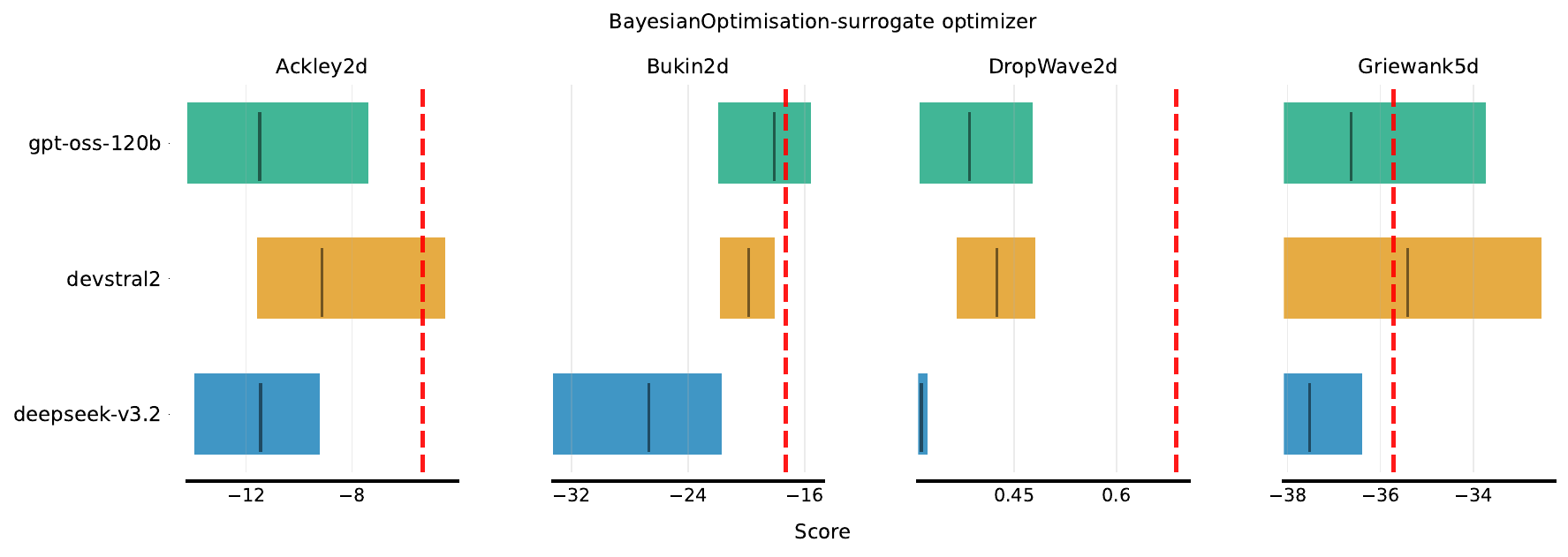}%
\hfill%
\includegraphics[width=0.48\textwidth]{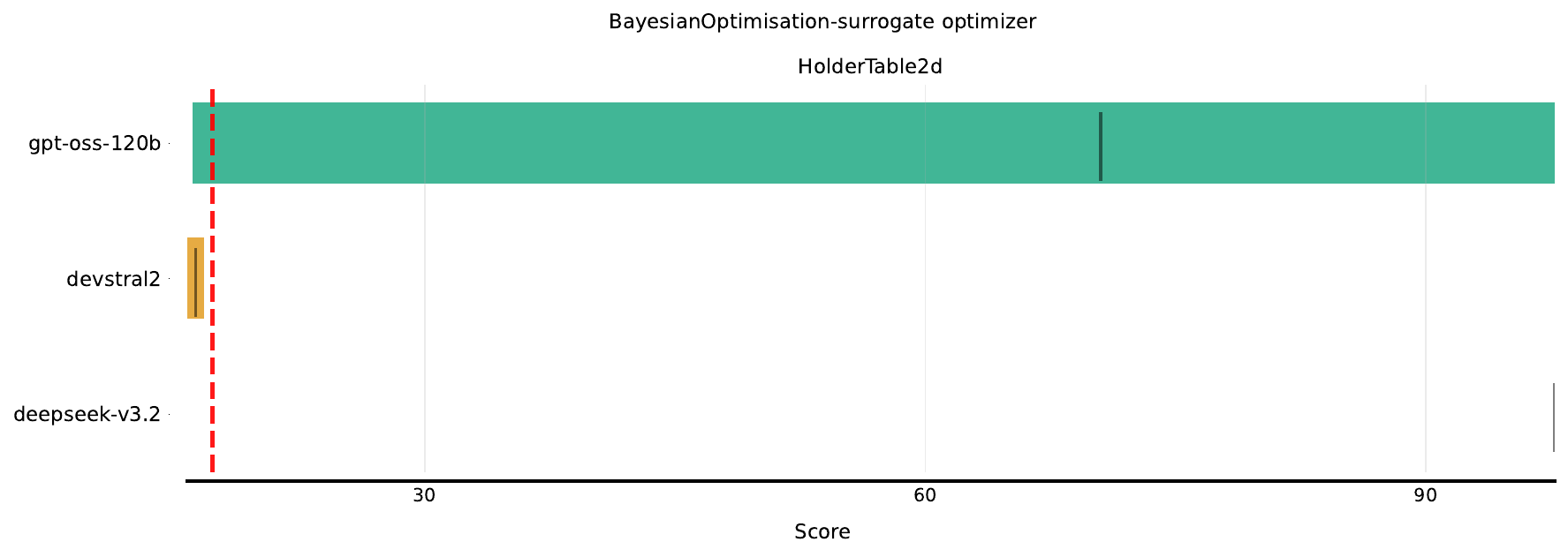}%
\\[0.5em]
\includegraphics[width=0.48\textwidth]{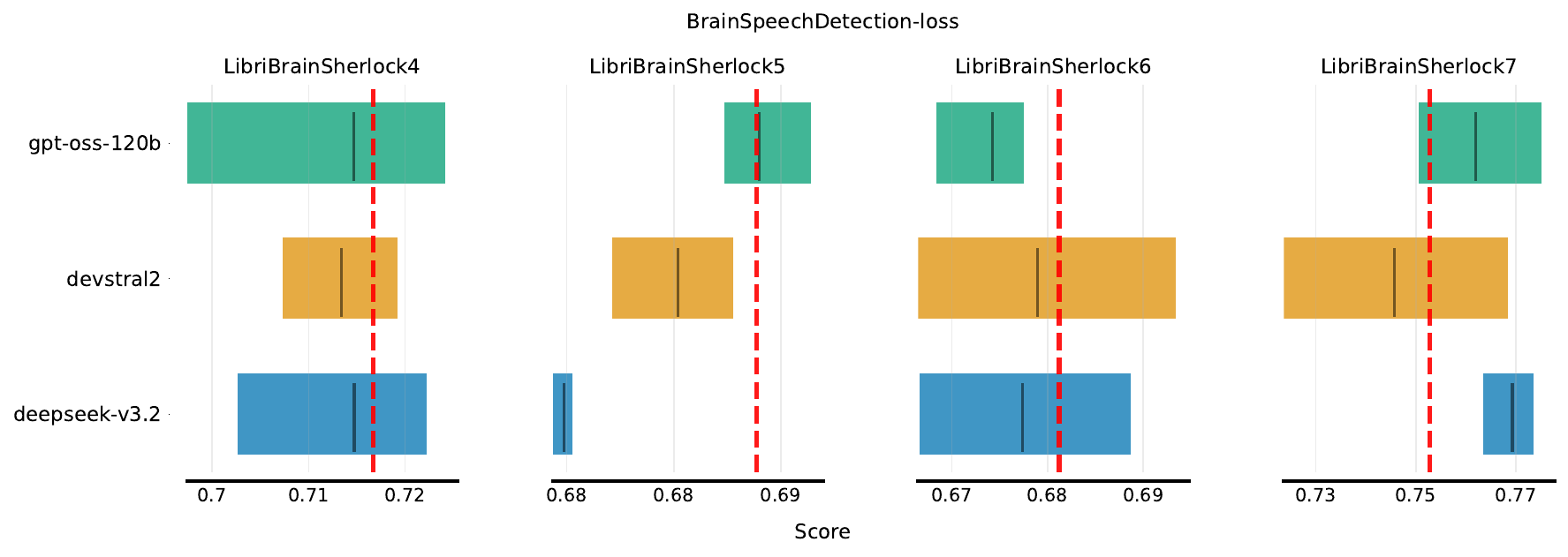}%
\hfill%
\includegraphics[width=0.48\textwidth]{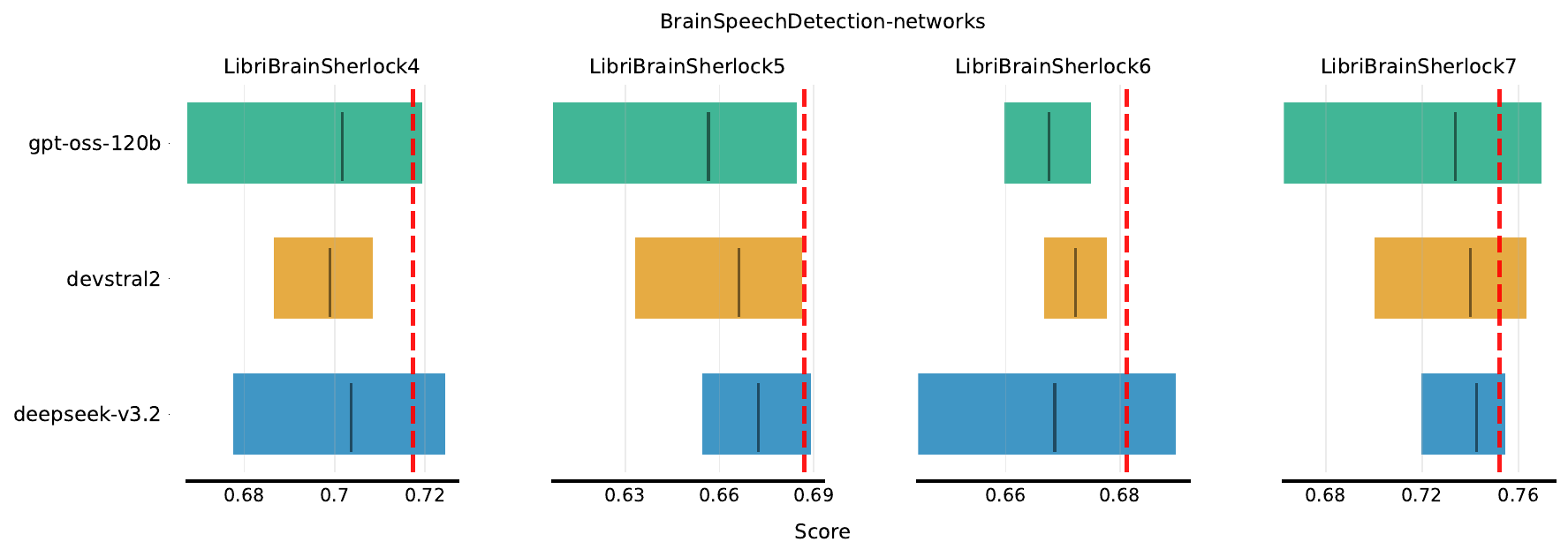}%
\\[0.5em]
\includegraphics[width=0.48\textwidth]{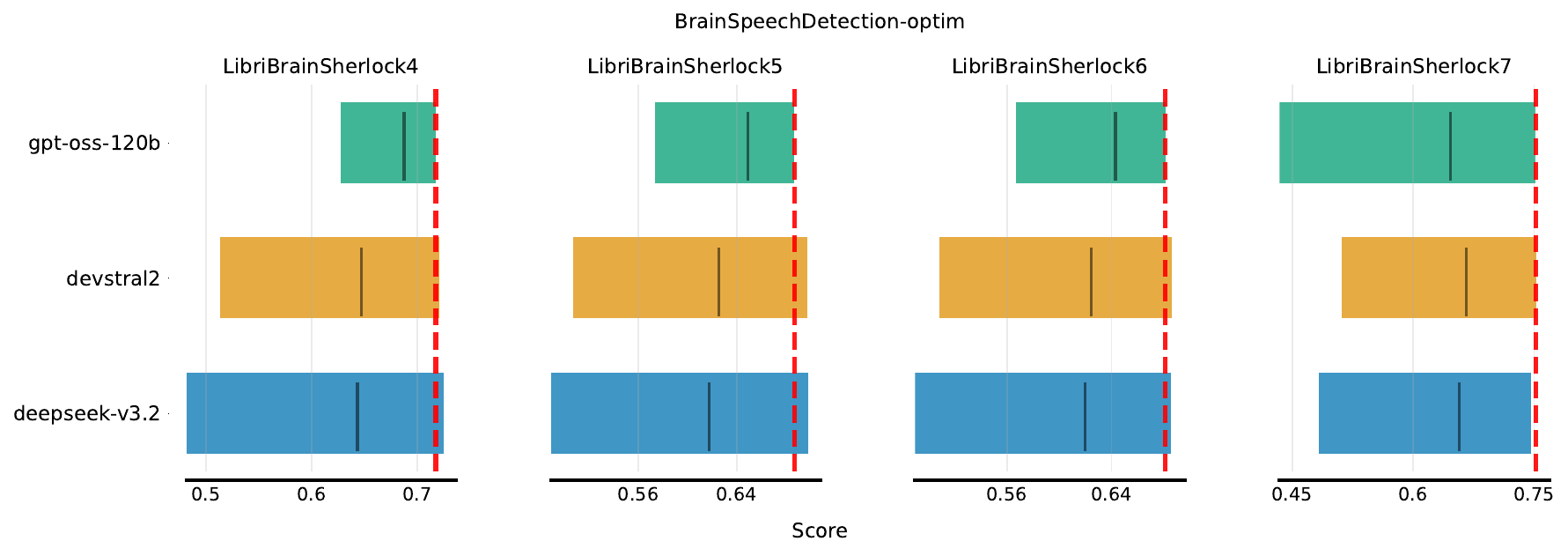}%
\hfill%
\includegraphics[width=0.48\textwidth]{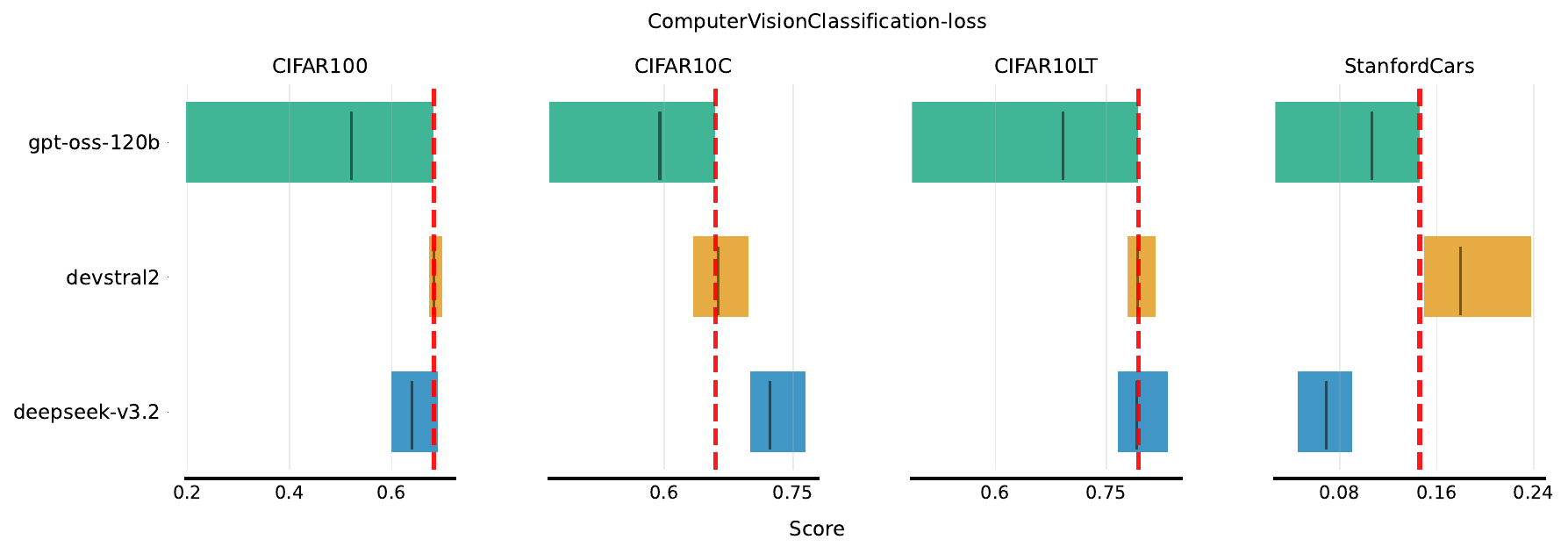}%
\\[0.5em]
\includegraphics[width=0.48\textwidth]{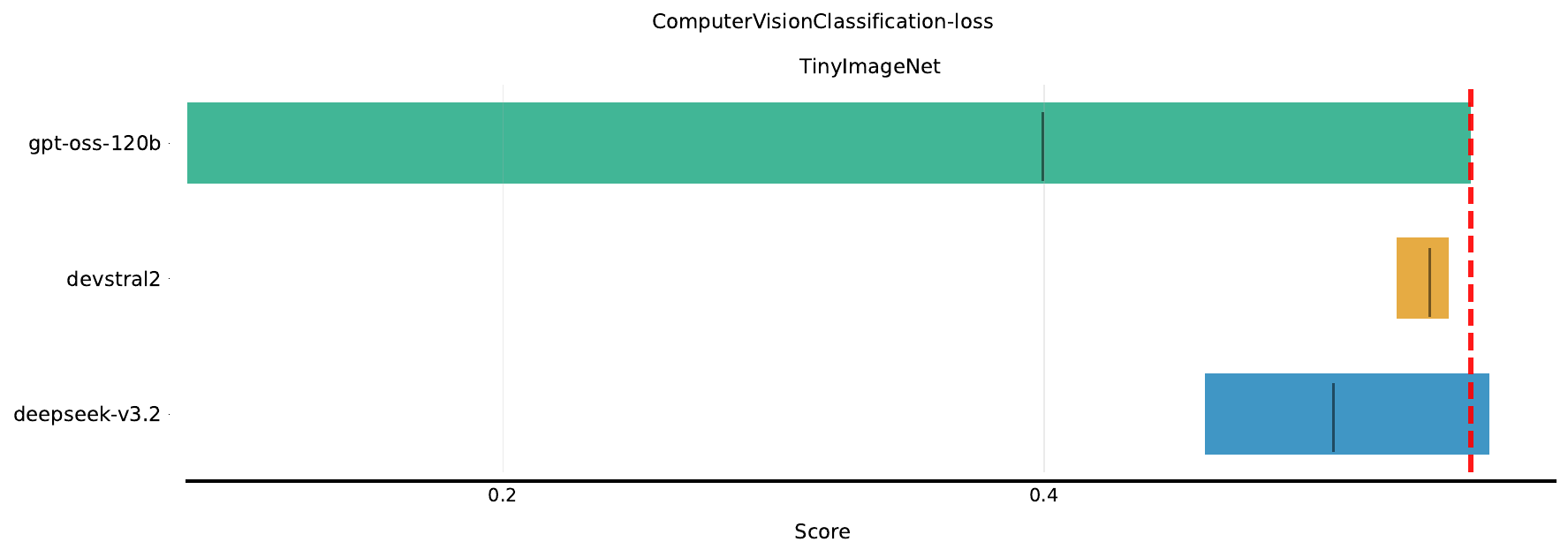}%
\hfill%
\includegraphics[width=0.48\textwidth]{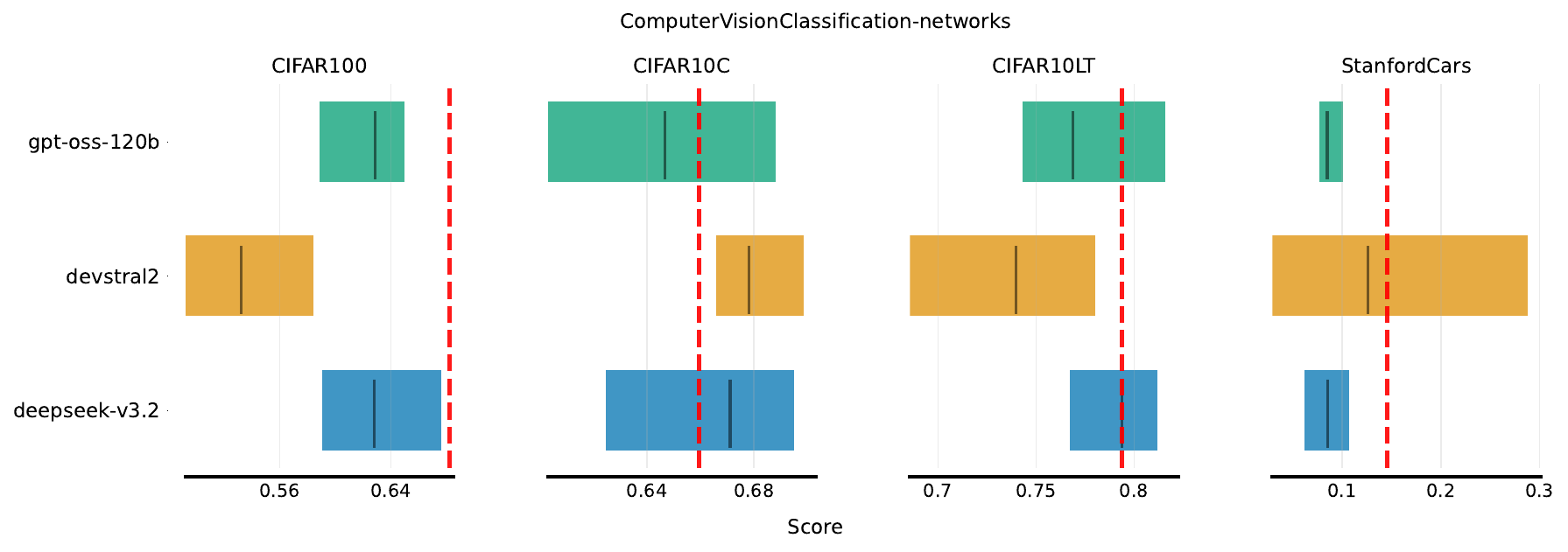}%
\\[0.5em]
\includegraphics[width=0.48\textwidth]{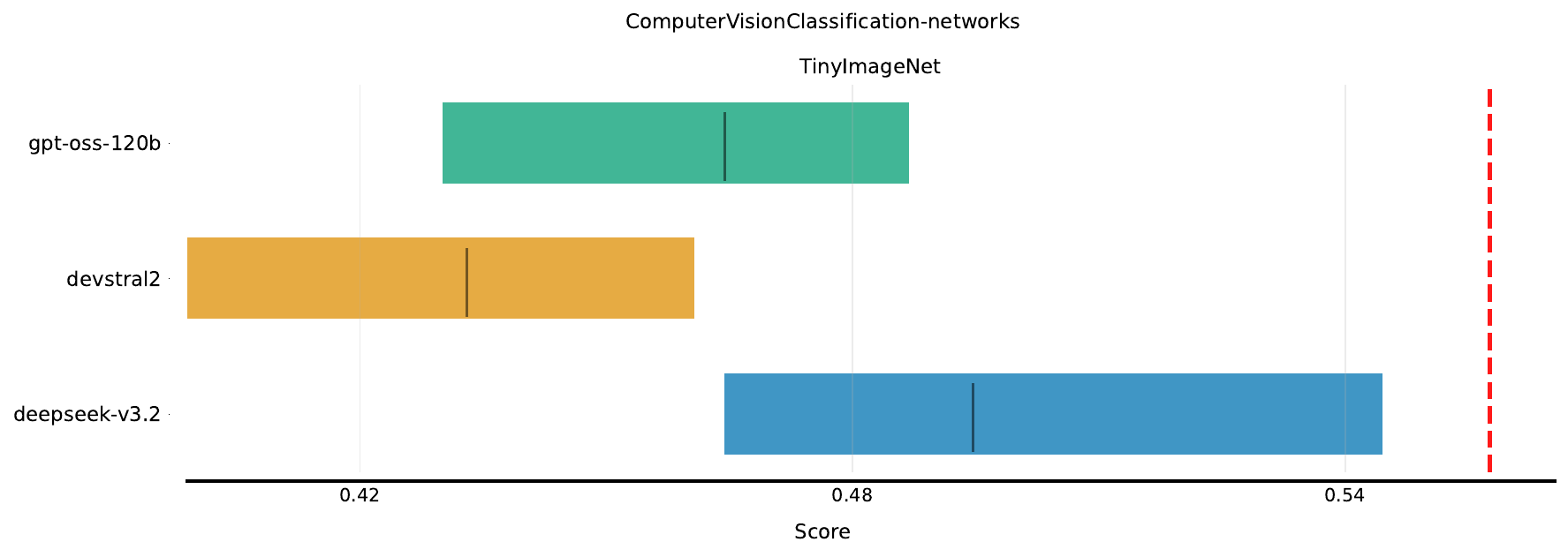}%
\hfill%
\includegraphics[width=0.48\textwidth]{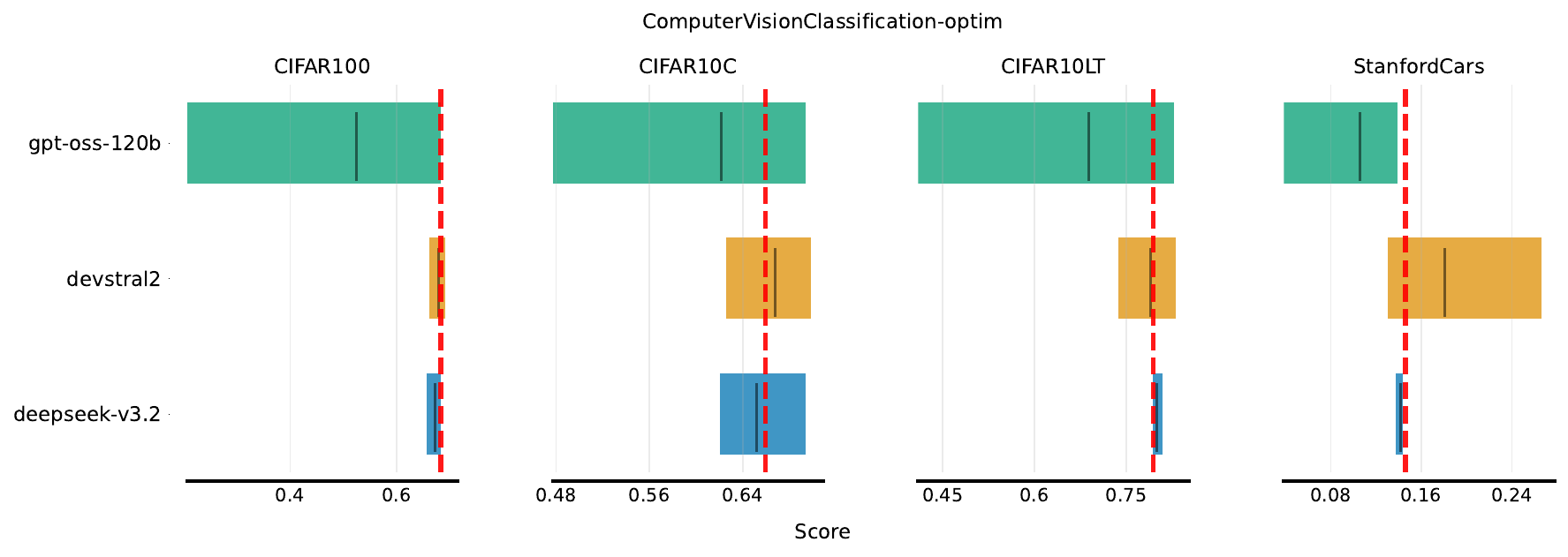}%
\\[0.5em]
\includegraphics[width=0.48\textwidth]{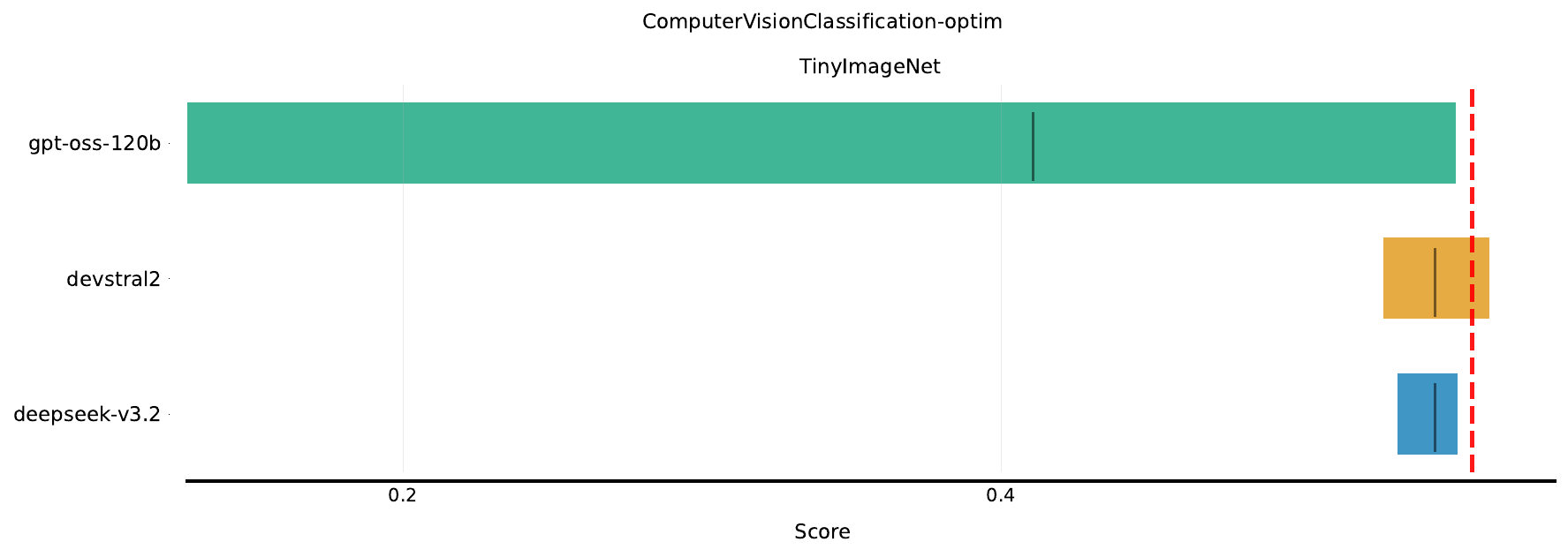}%
\hfill%
\includegraphics[width=0.48\textwidth]{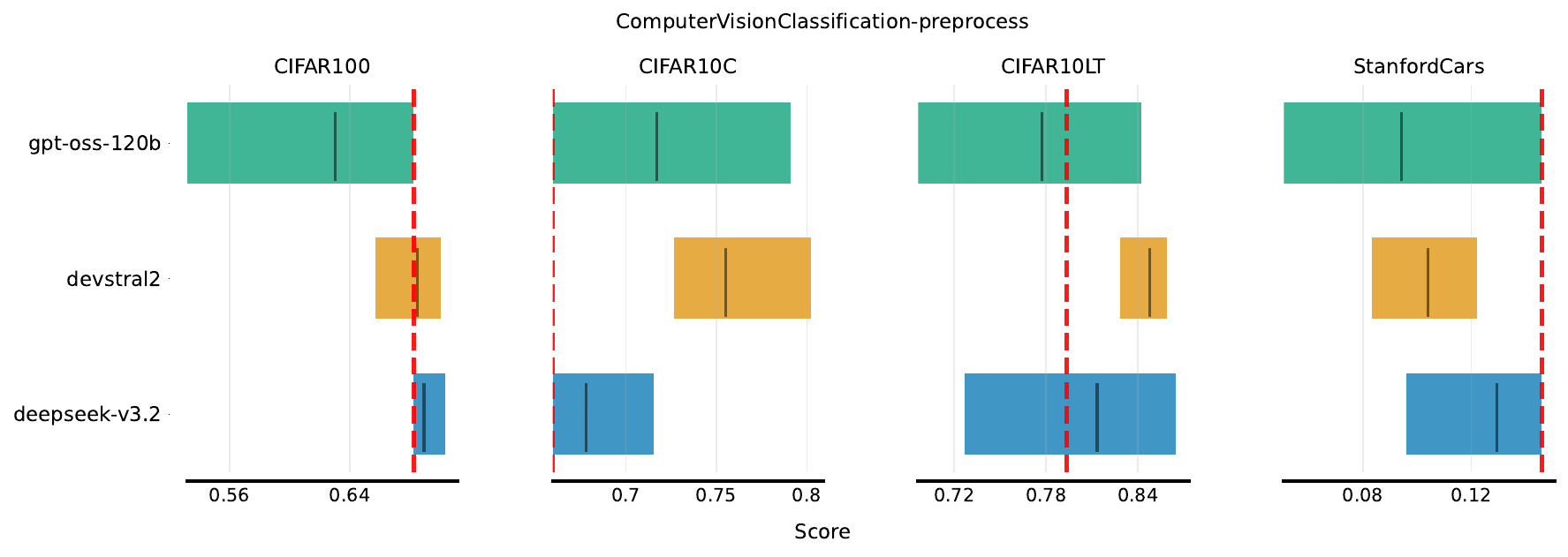}%
\caption{DiscoBench (3 Successful Seeds) results on Meta-Test tasks. (Part 2/5)}
\label{fig:until_success_mt_2}
\end{figure}
\clearpage

\begin{figure}[htbp]
\centering
\setlength{\lineskip}{0pt}
\includegraphics[width=0.48\textwidth]{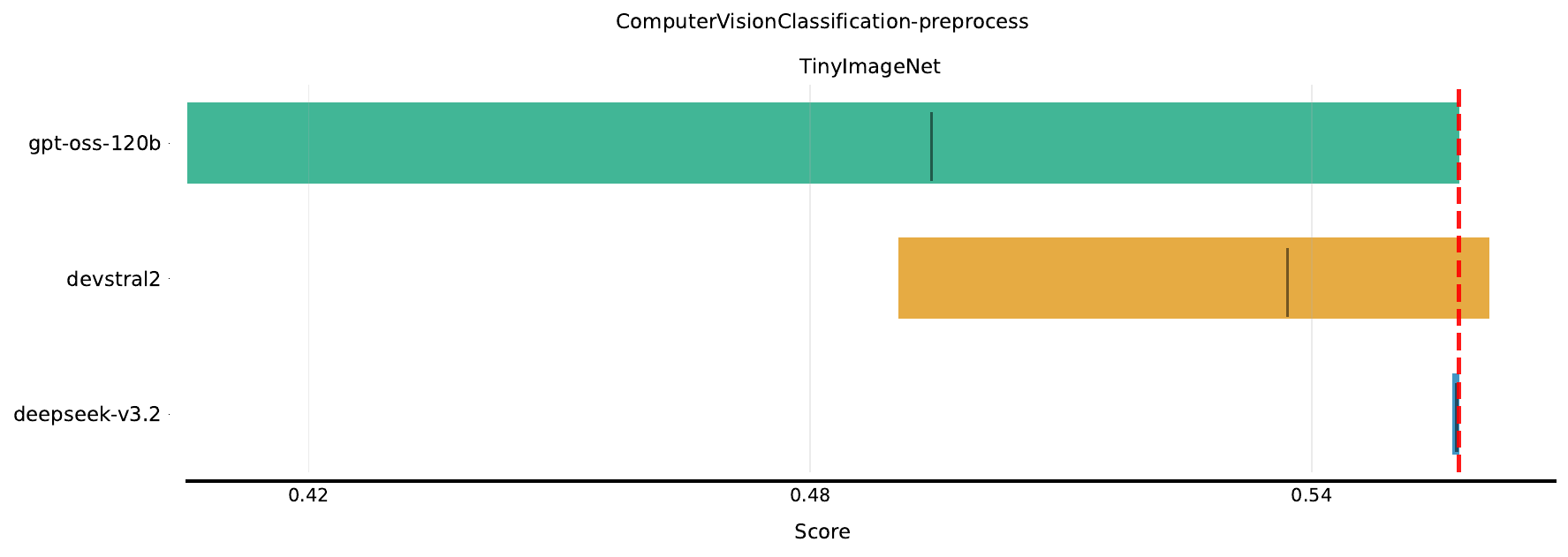}%
\hfill%
\includegraphics[width=0.48\textwidth]{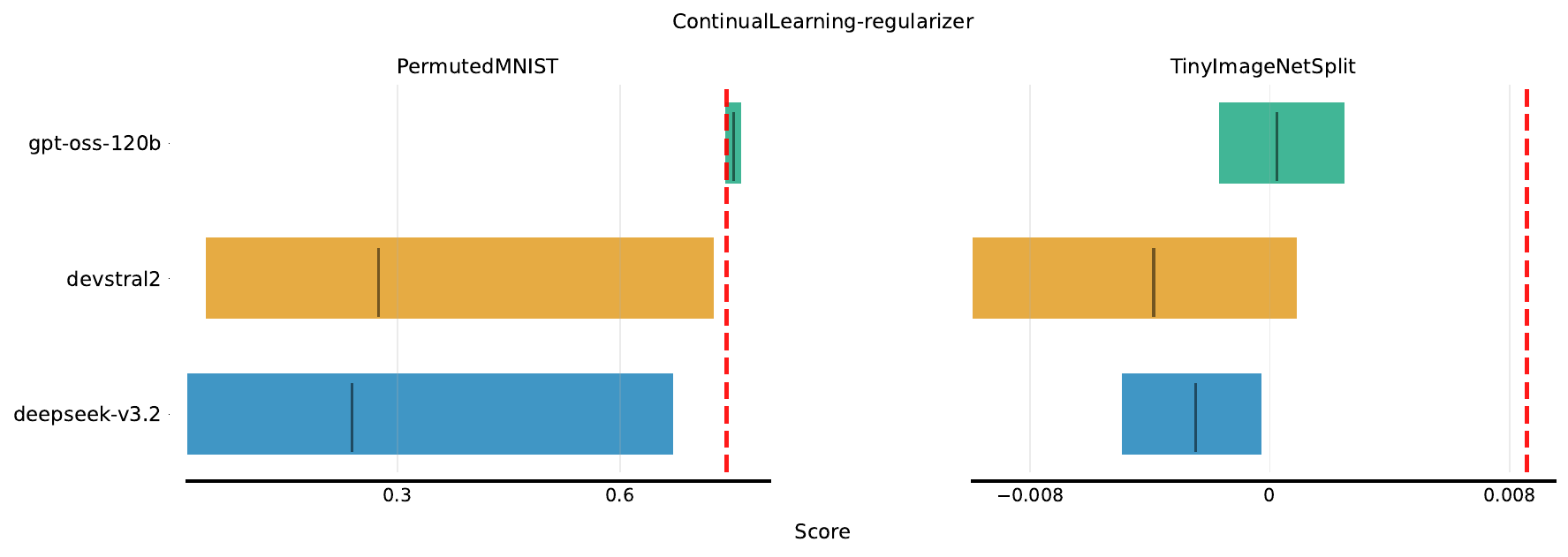}%
\\[0.5em]
\includegraphics[width=0.48\textwidth]{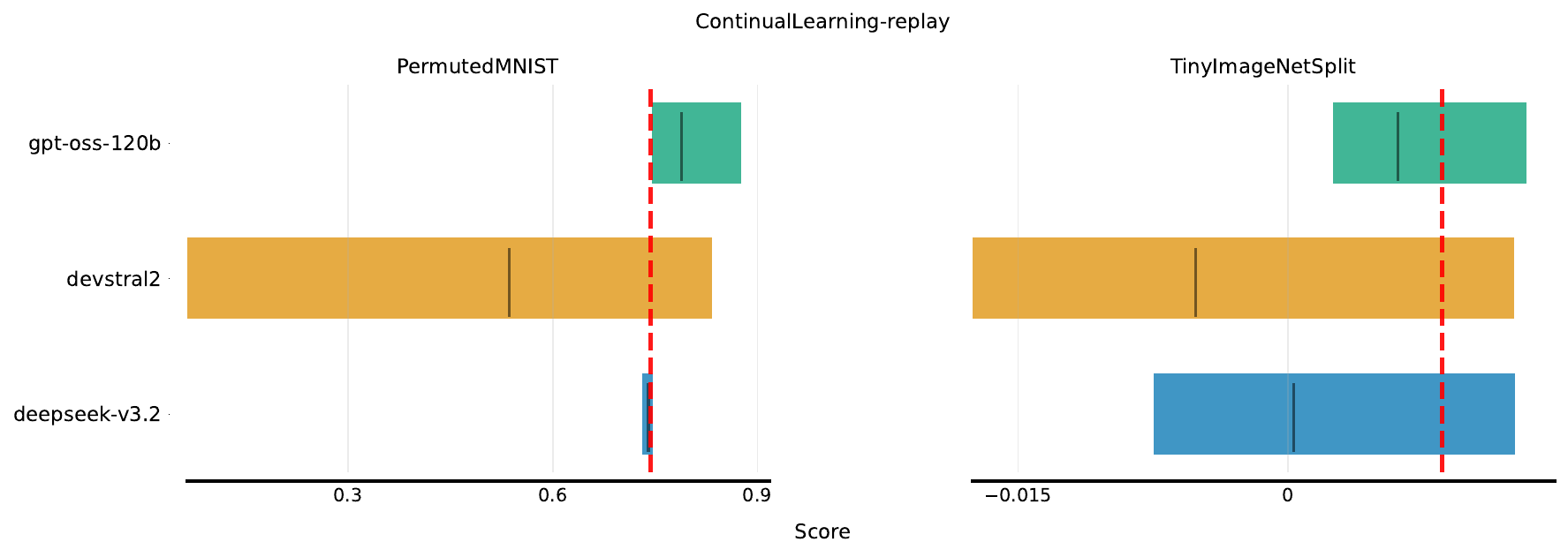}%
\hfill%
\includegraphics[width=0.48\textwidth]{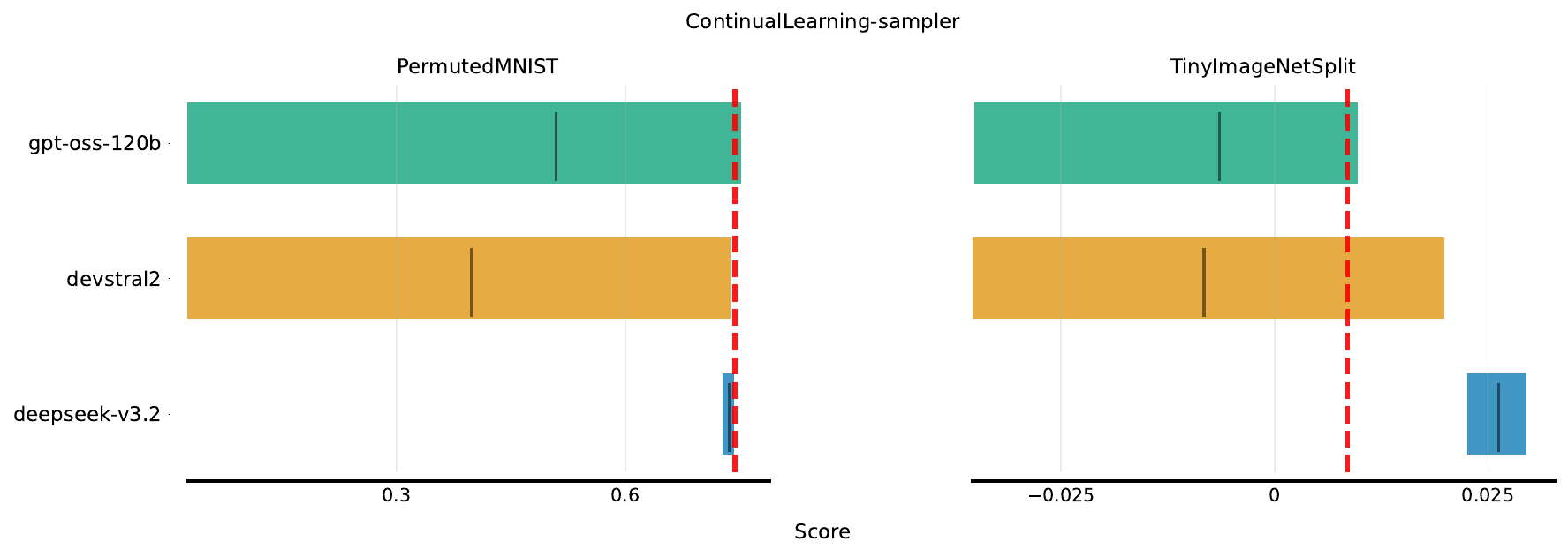}%
\\[0.5em]
\includegraphics[width=0.48\textwidth]{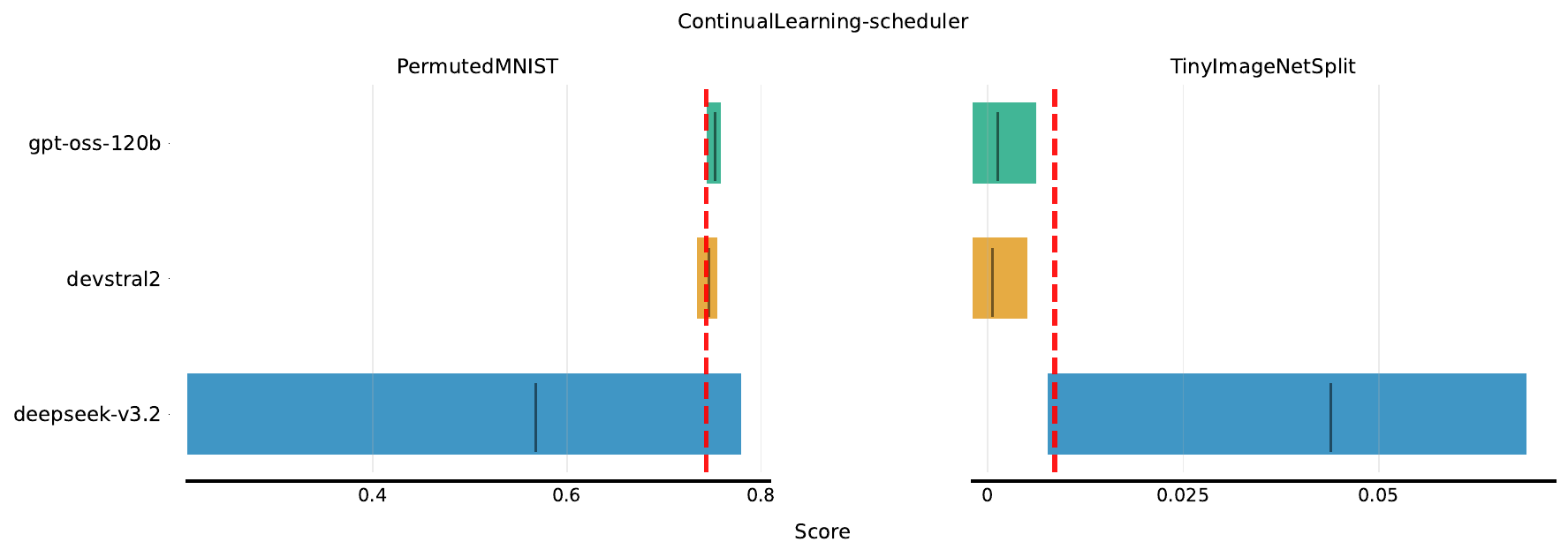}%
\hfill%
\includegraphics[width=0.48\textwidth]{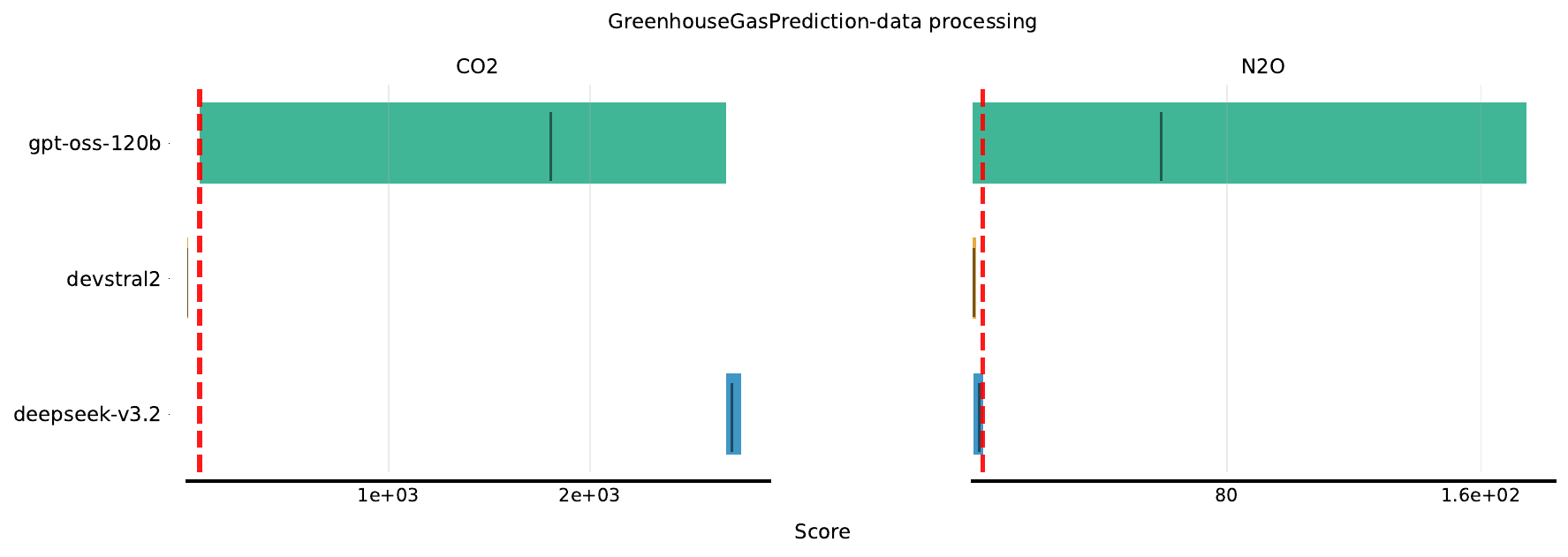}%
\\[0.5em]
\includegraphics[width=0.48\textwidth]{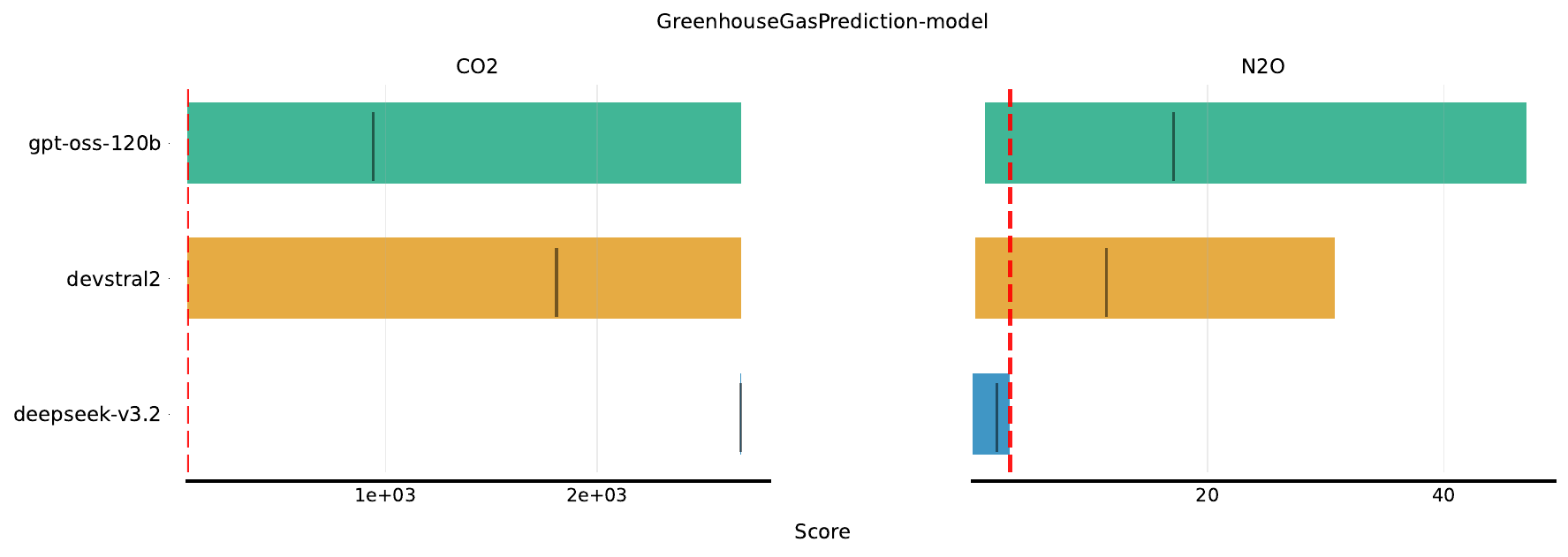}%
\hfill%
\includegraphics[width=0.48\textwidth]{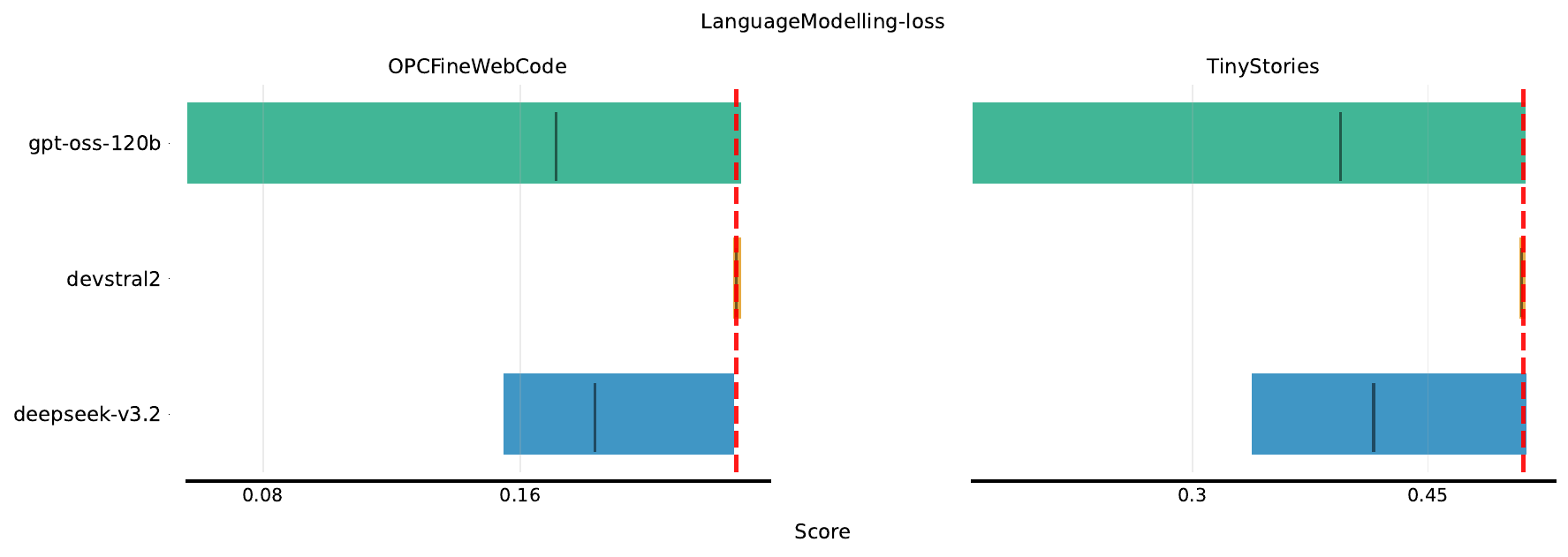}%
\\[0.5em]
\includegraphics[width=0.48\textwidth]{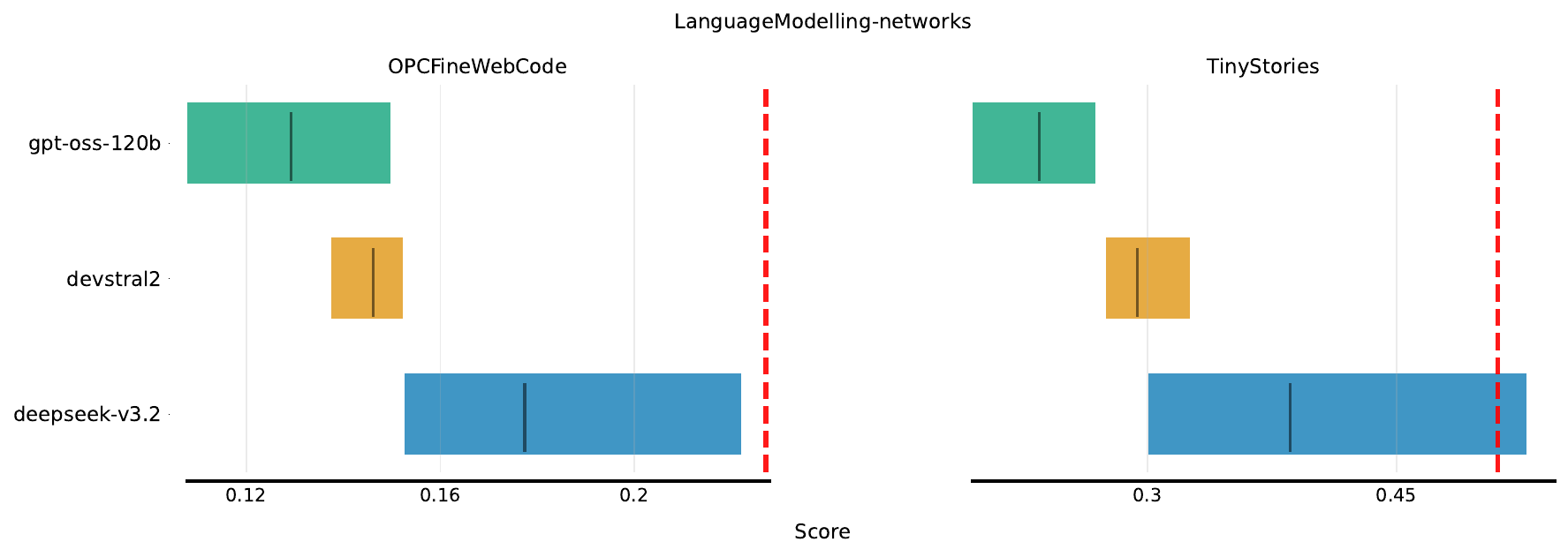}%
\hfill%
\includegraphics[width=0.48\textwidth]{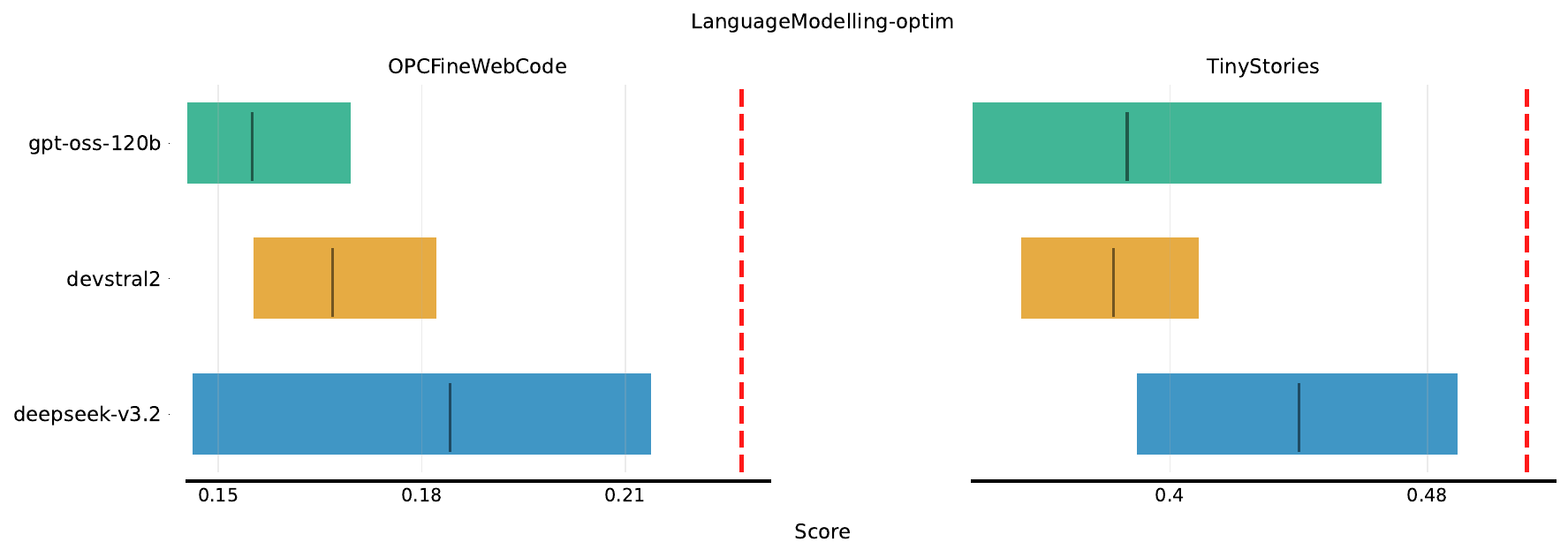}%
\\[0.5em]
\includegraphics[width=0.48\textwidth]{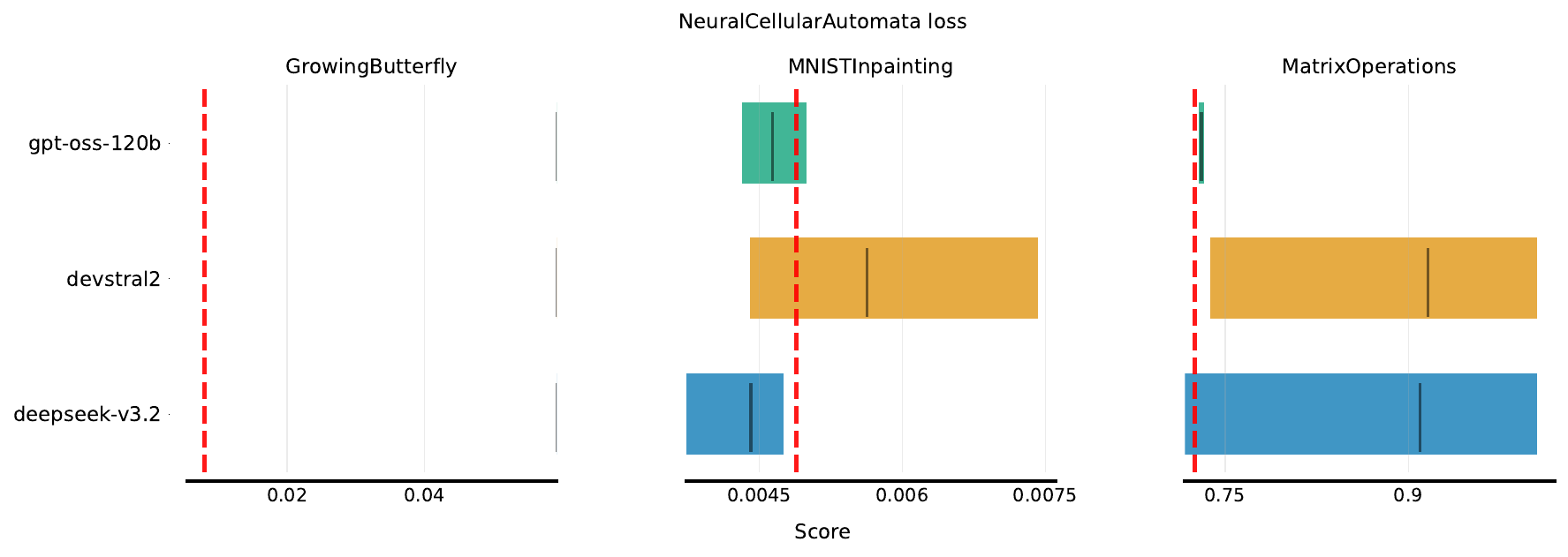}%
\hfill%
\includegraphics[width=0.48\textwidth]{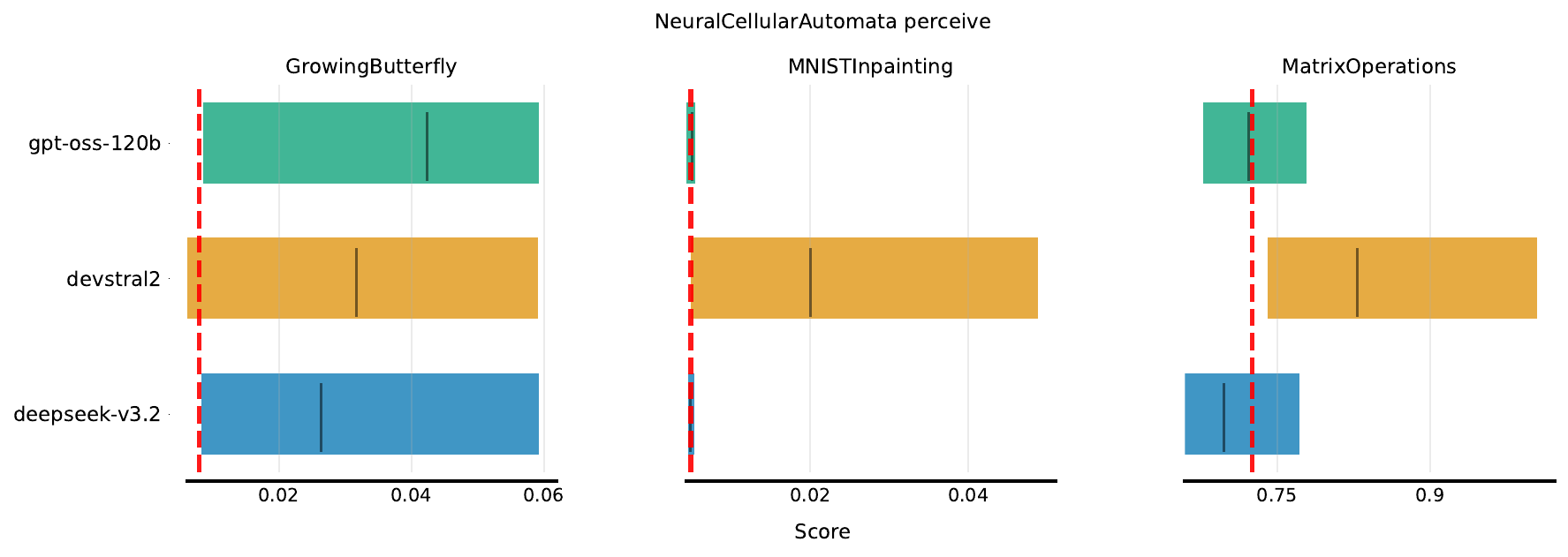}%
\caption{DiscoBench (3 Successful Seeds) results on Meta-Test tasks. (Part 3/5)}
\label{fig:until_success_mt_3}
\end{figure}
\clearpage

\begin{figure}[htbp]
\centering
\setlength{\lineskip}{0pt}
\includegraphics[width=0.48\textwidth]{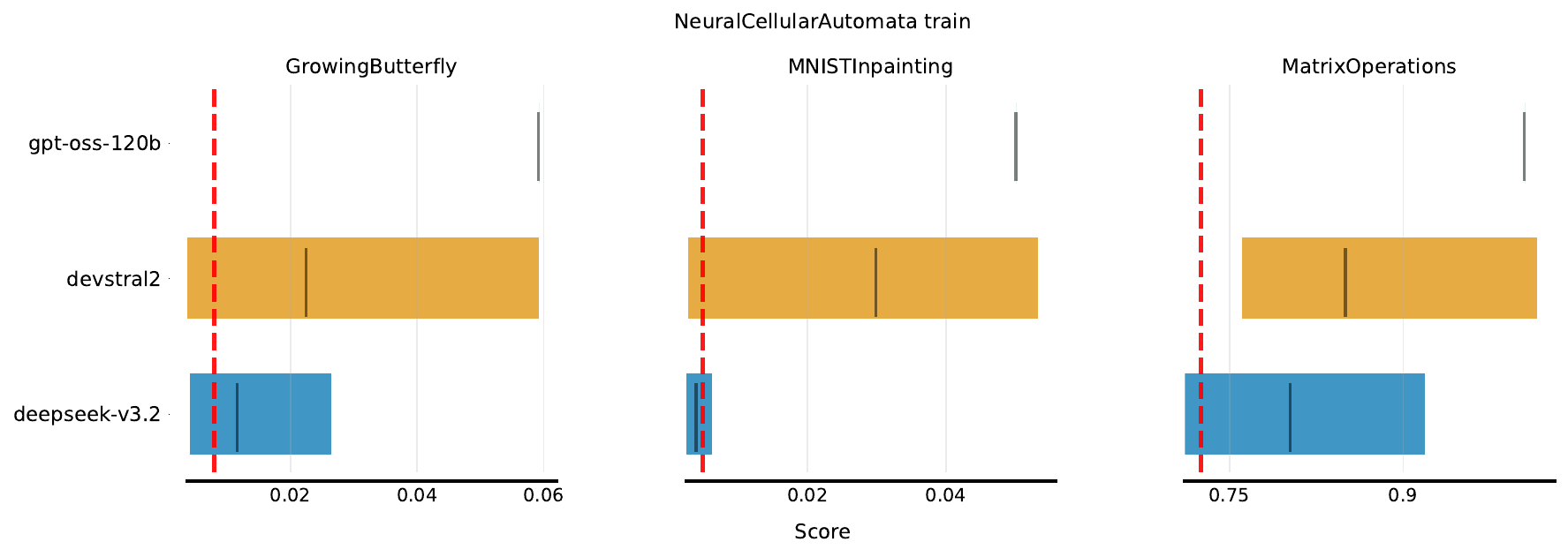}%
\hfill%
\includegraphics[width=0.48\textwidth]{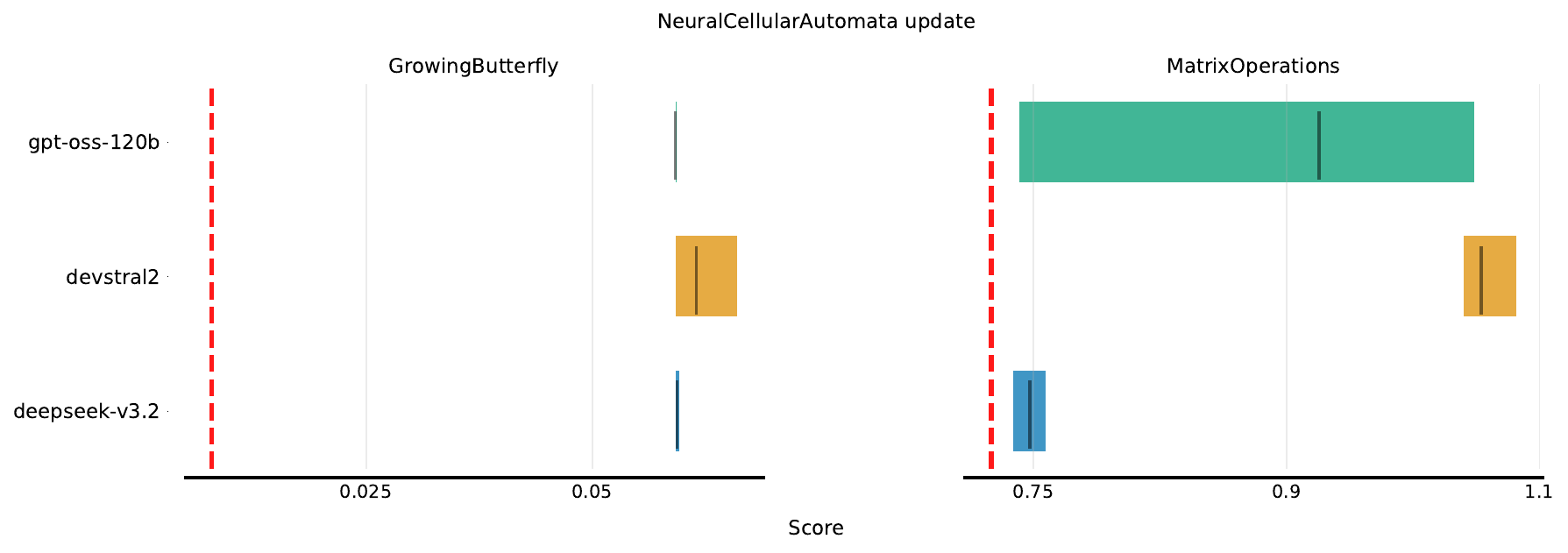}%
\\[0.5em]
\includegraphics[width=0.48\textwidth]{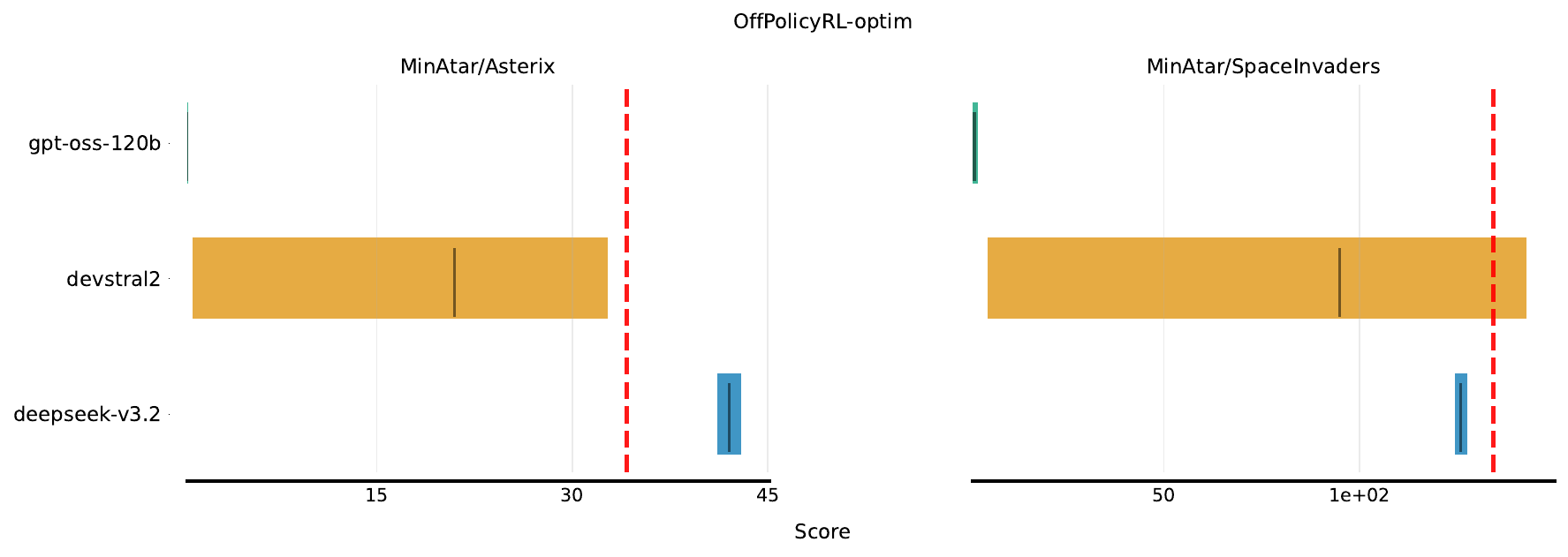}%
\hfill%
\includegraphics[width=0.48\textwidth]{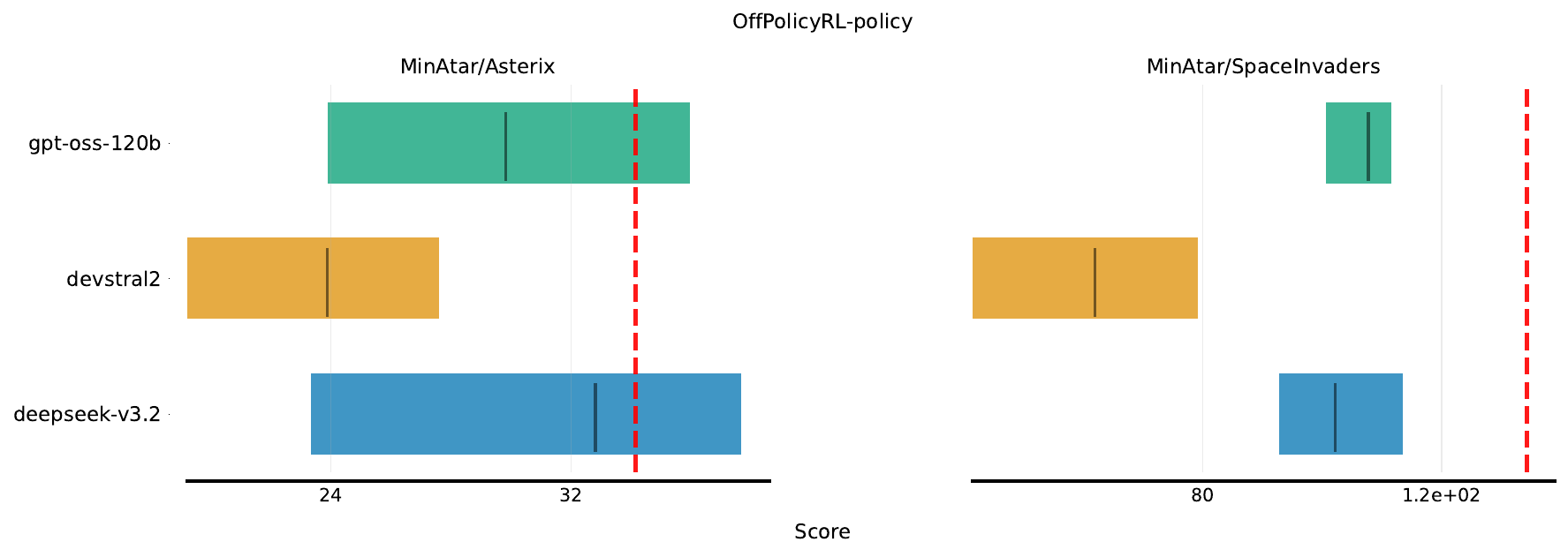}%
\\[0.5em]
\includegraphics[width=0.48\textwidth]{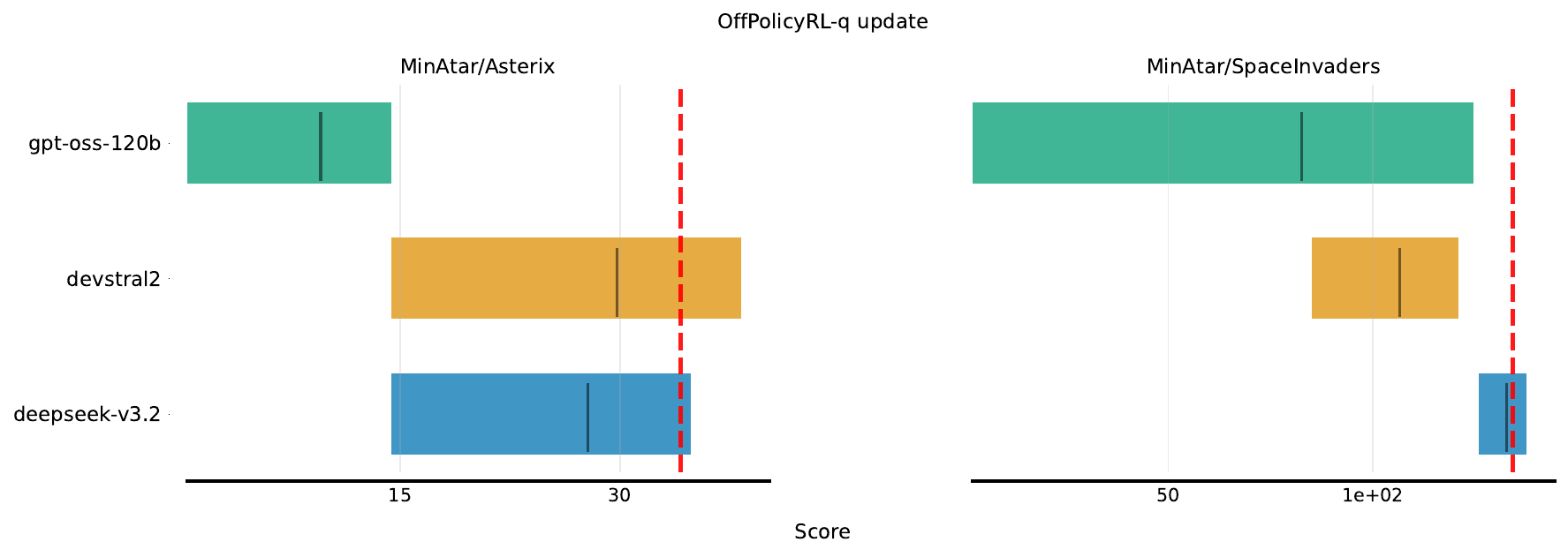}%
\hfill%
\includegraphics[width=0.48\textwidth]{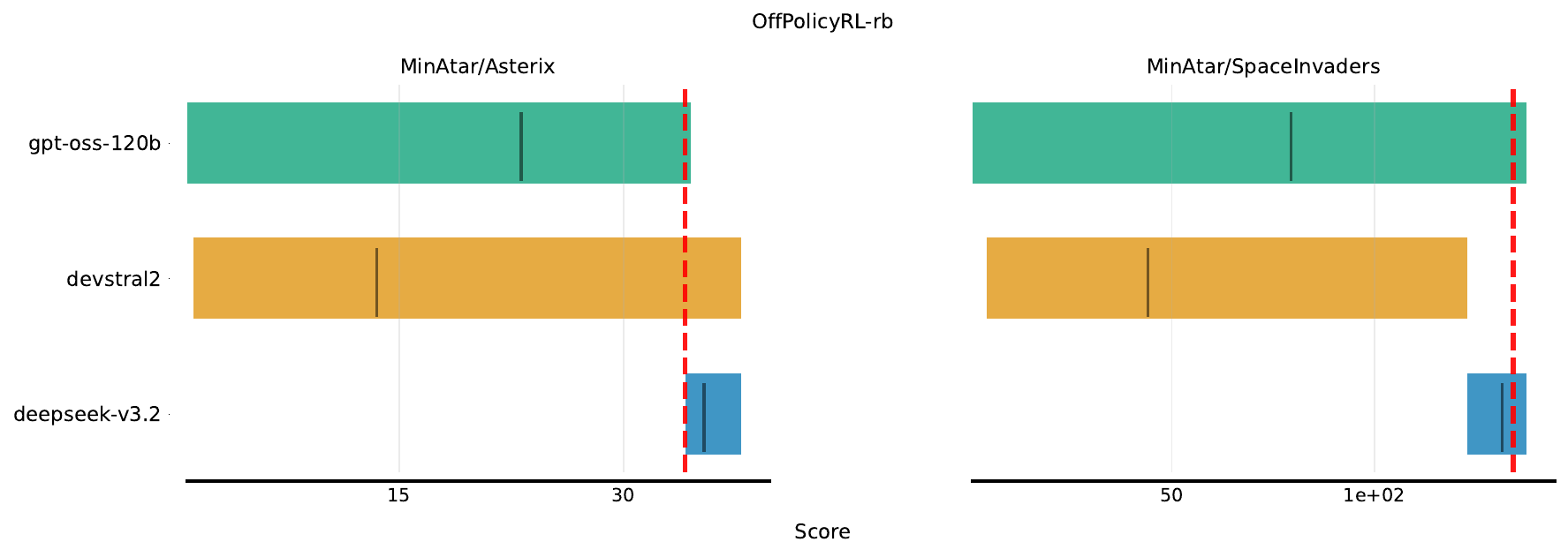}%
\\[0.5em]
\includegraphics[width=0.48\textwidth]{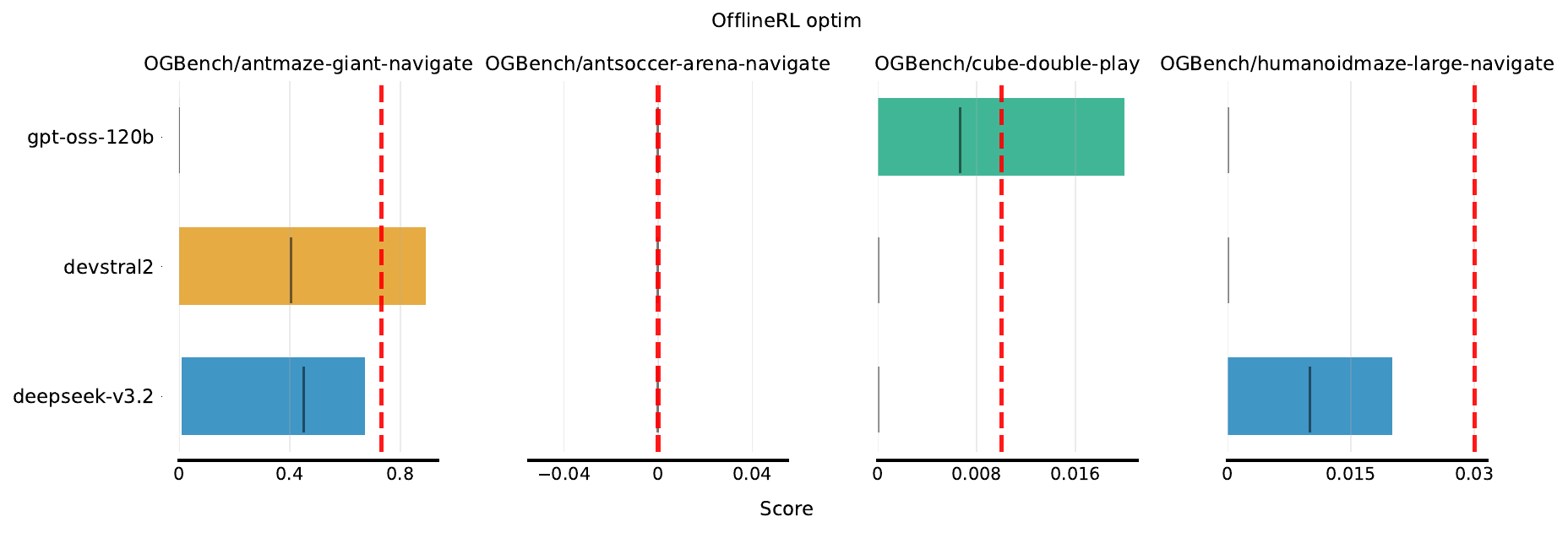}%
\hfill%
\includegraphics[width=0.48\textwidth]{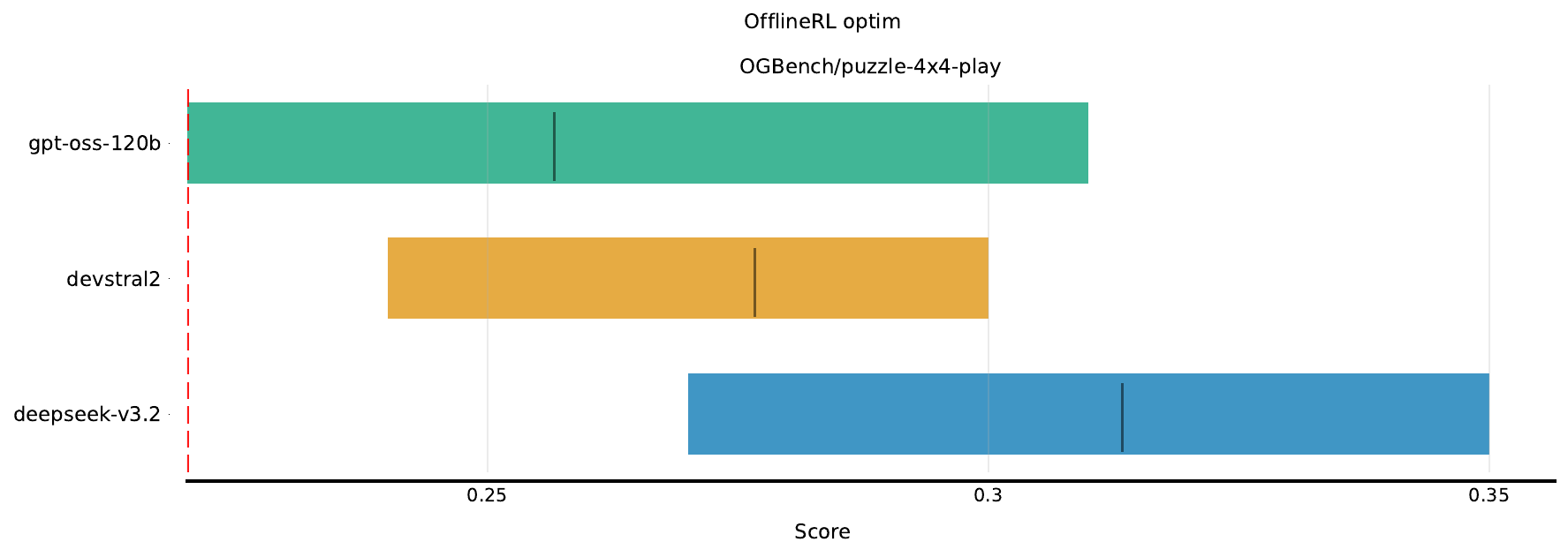}%
\\[0.5em]
\includegraphics[width=0.48\textwidth]{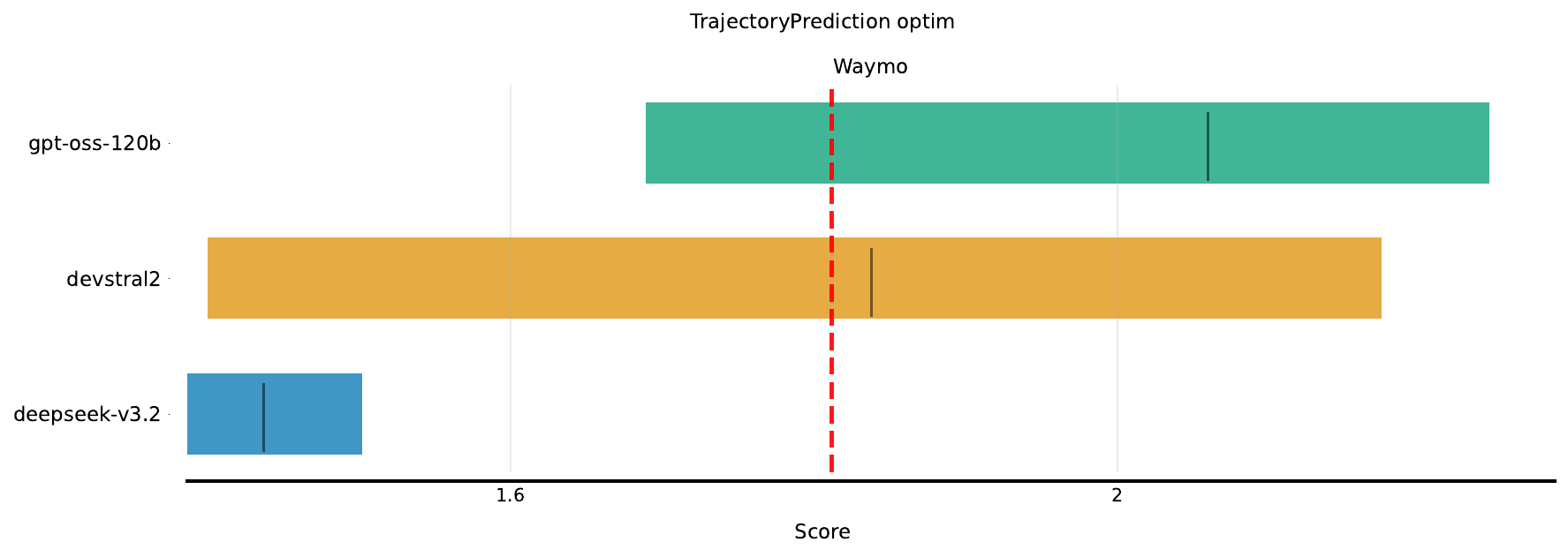}%
\hfill%
\includegraphics[width=0.48\textwidth]{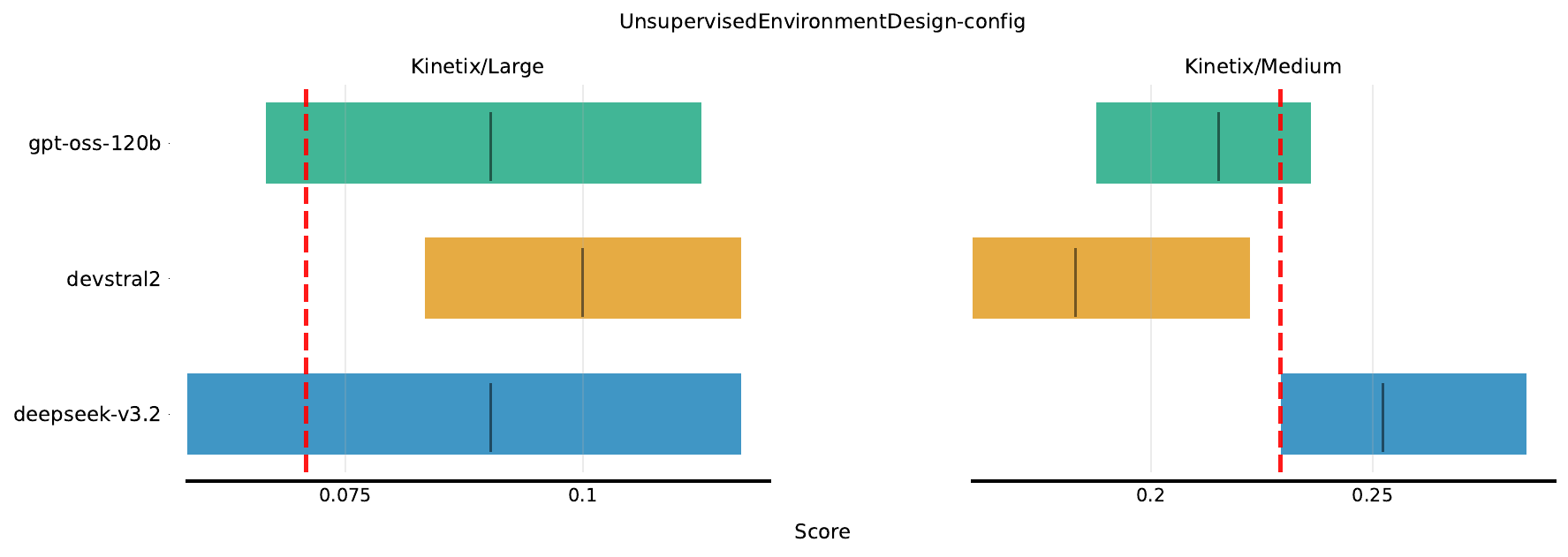}%
\\[0.5em]
\includegraphics[width=0.48\textwidth]{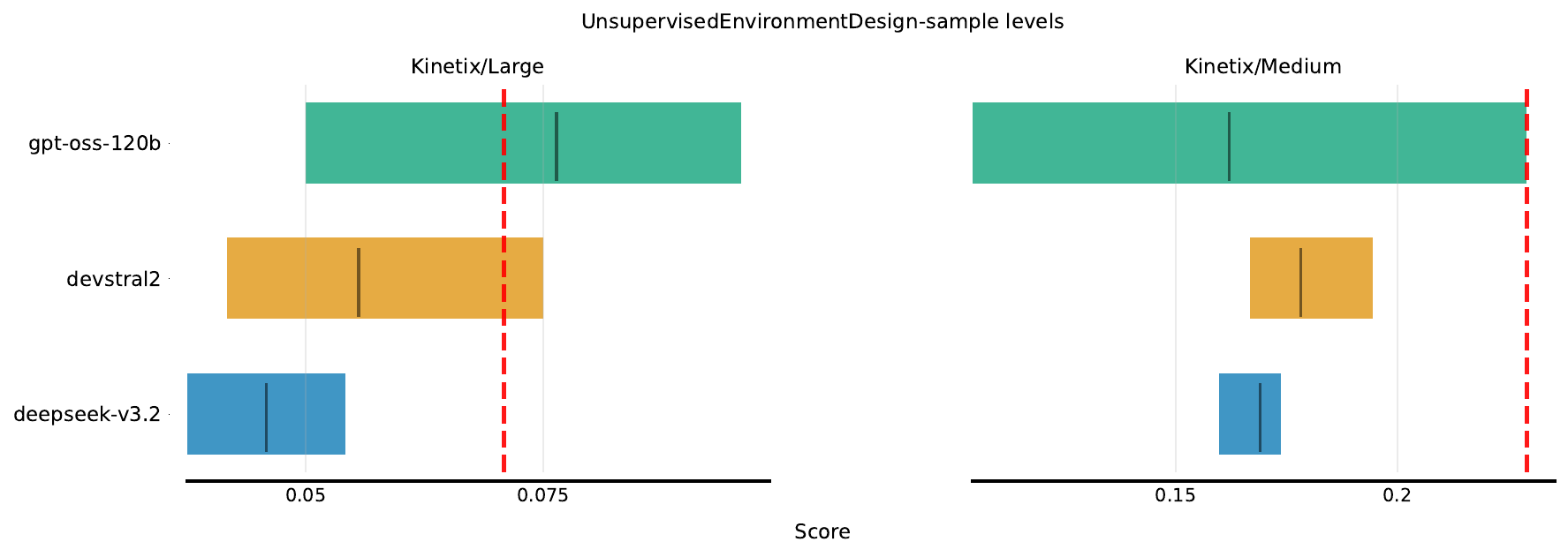}%
\hfill%
\includegraphics[width=0.48\textwidth]{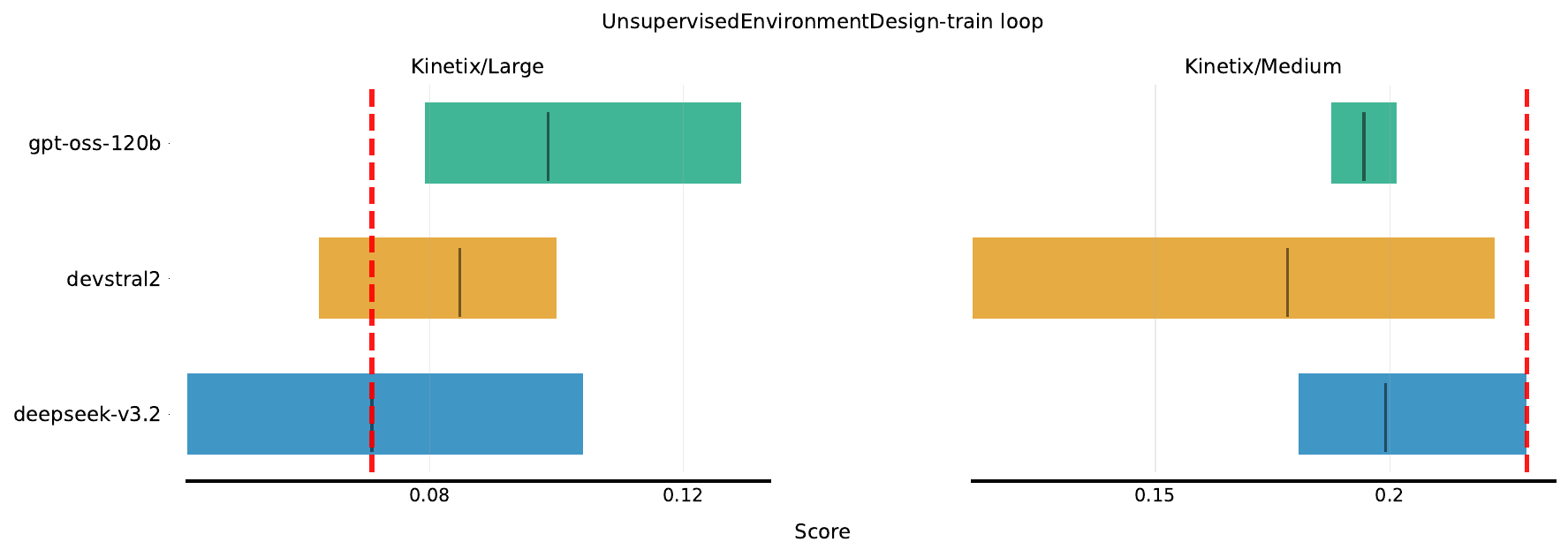}%
\caption{DiscoBench (3 Successful Seeds) results on Meta-Test tasks. (Part 4/5)}
\label{fig:until_success_mt_4}
\end{figure}
\clearpage

\begin{figure}[htbp]
\centering
\setlength{\lineskip}{0pt}
\includegraphics[width=0.48\textwidth]{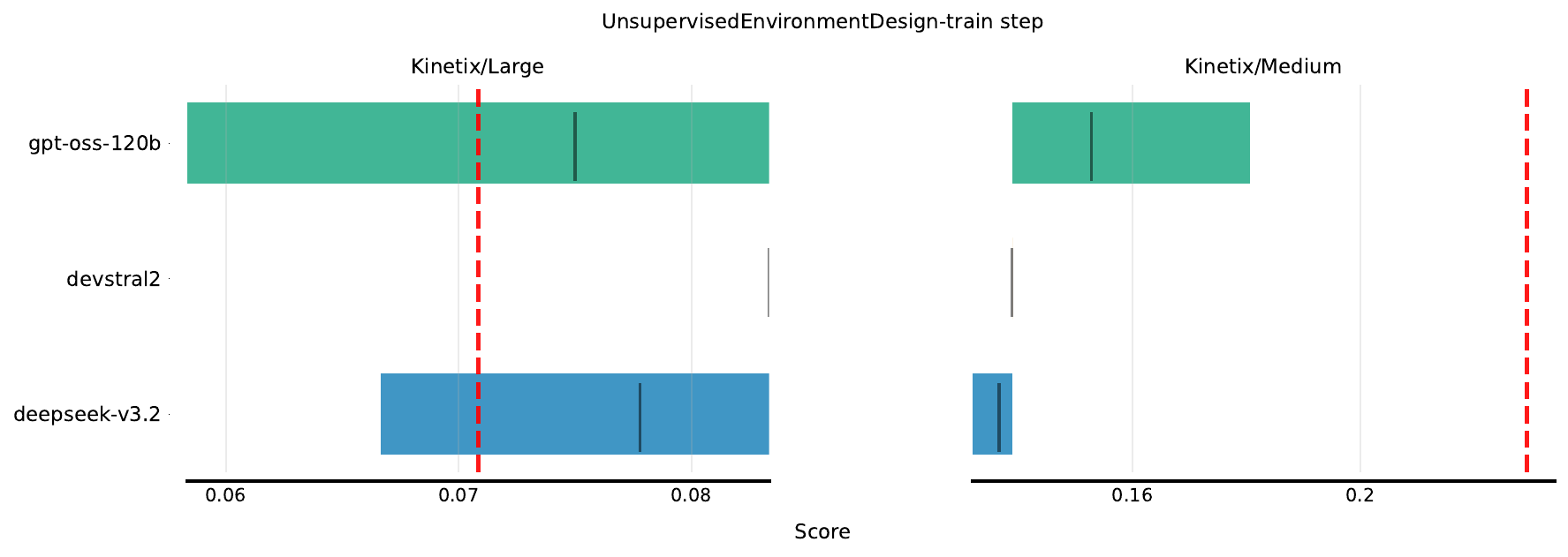}%
\hfill%
\caption{DiscoBench (3 Successful Seeds) results on Meta-Test tasks. (Part 5/5)}
\label{fig:until_success_mt_5}
\end{figure}
\clearpage

\subsection{On-Policy RL Combinations -- Meta-Train}
\label{sec:on_policy_id}

\begin{figure}[htbp]
\centering
\setlength{\lineskip}{0pt}
\includegraphics[width=0.48\textwidth]{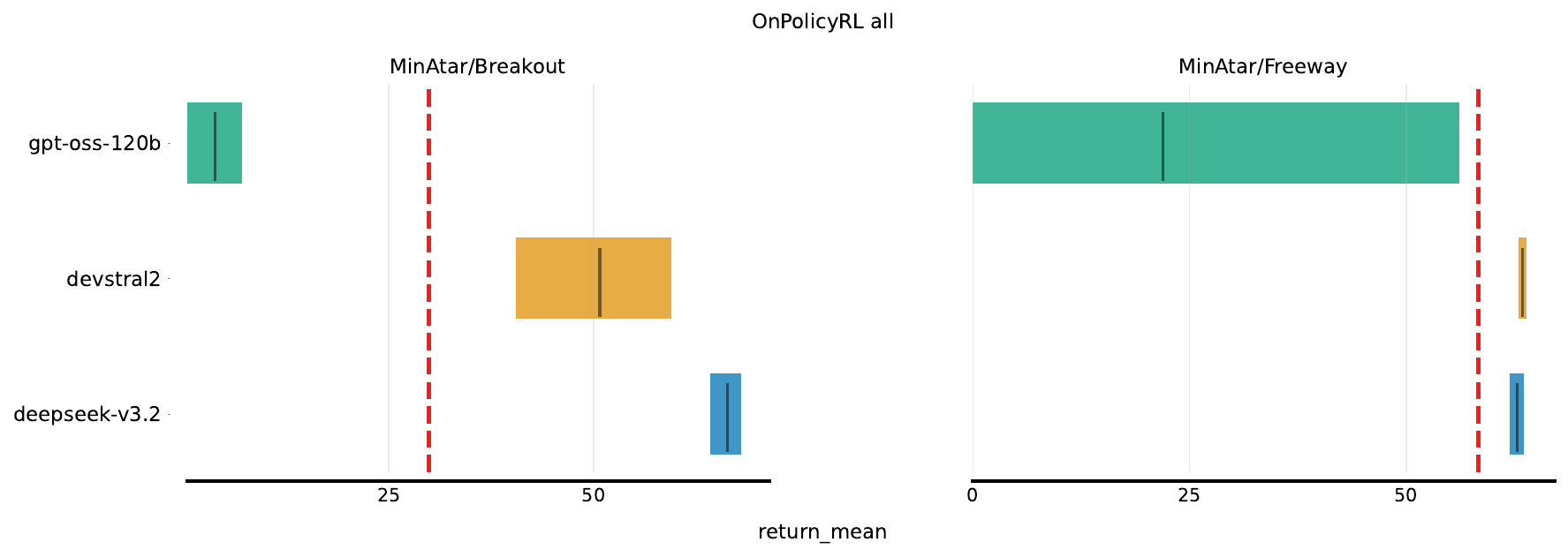}%
\hfill%
\includegraphics[width=0.48\textwidth]{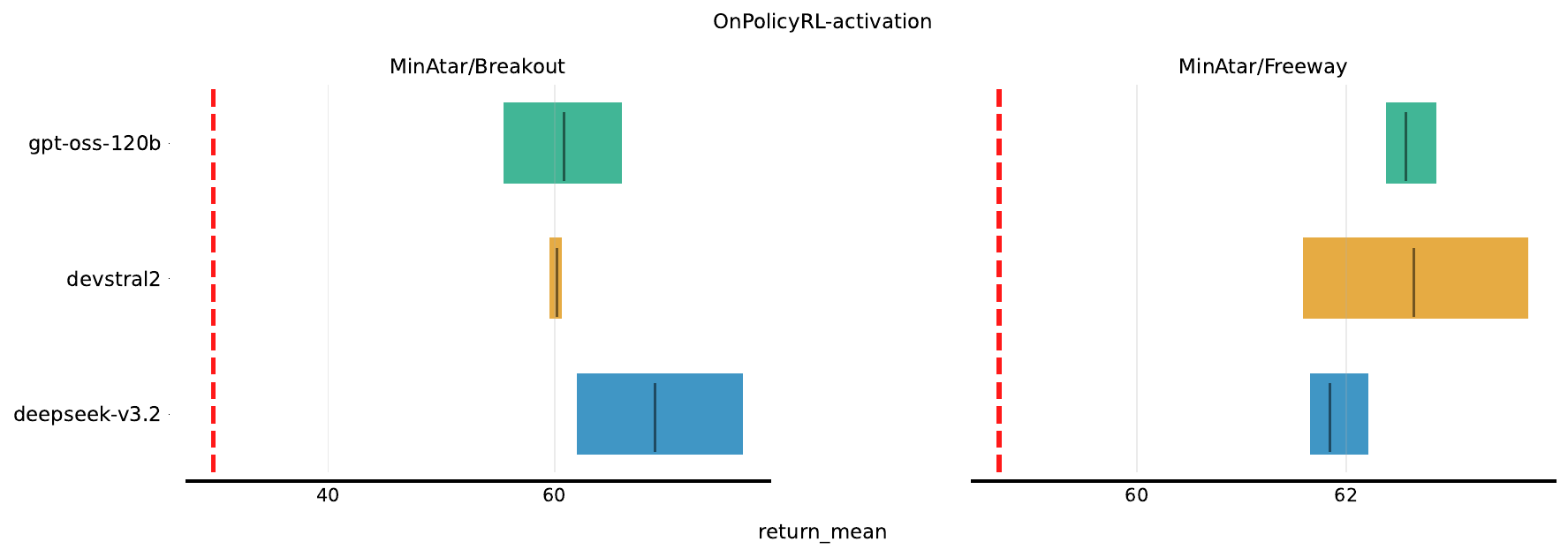}%
\\[0.5em]
\includegraphics[width=0.48\textwidth]{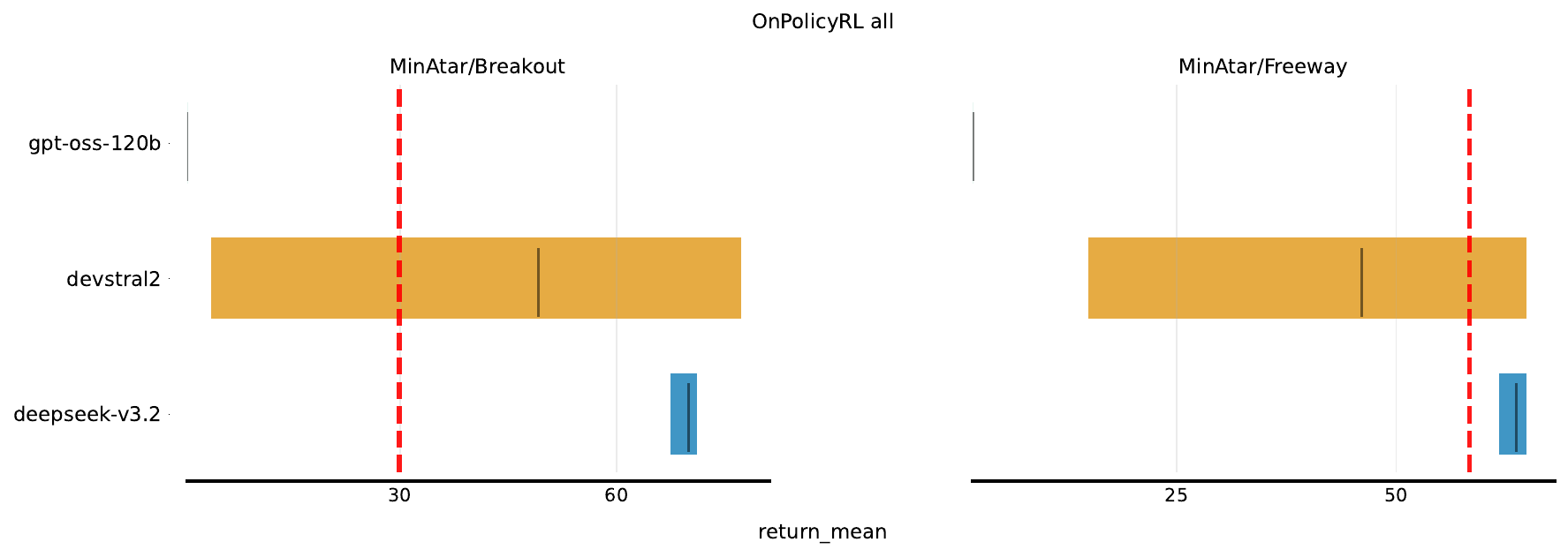}%
\hfill%
\includegraphics[width=0.48\textwidth]{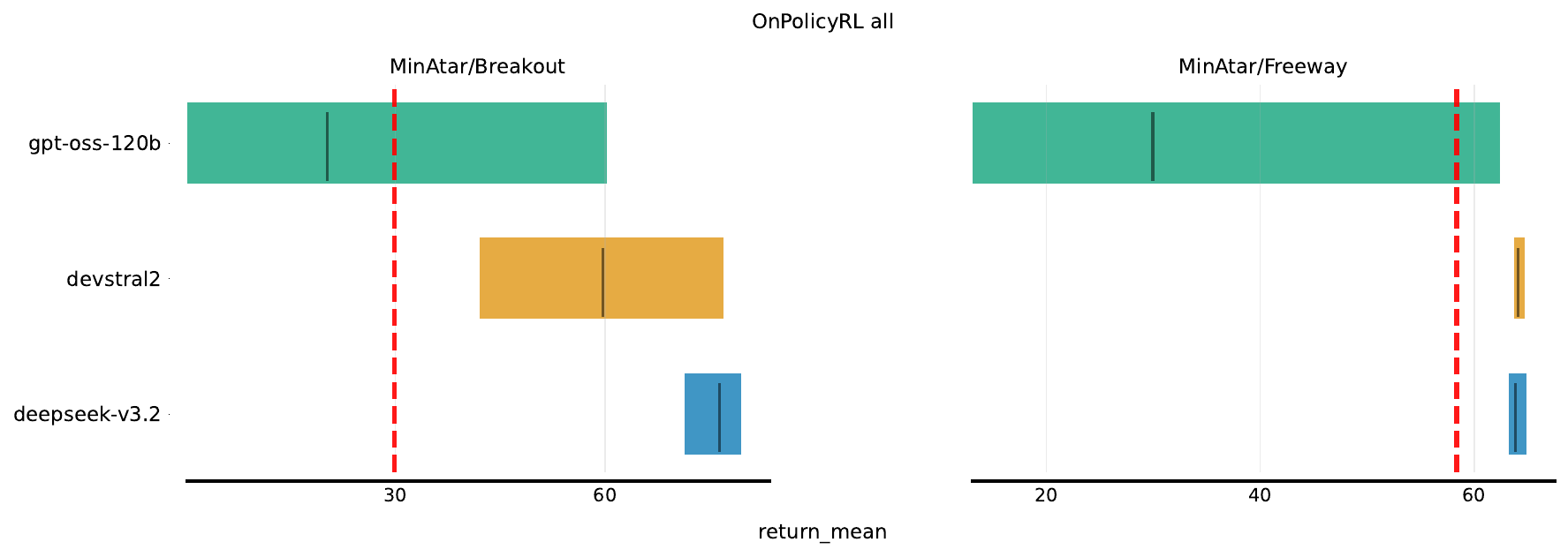}%
\\[0.5em]
\includegraphics[width=0.48\textwidth]{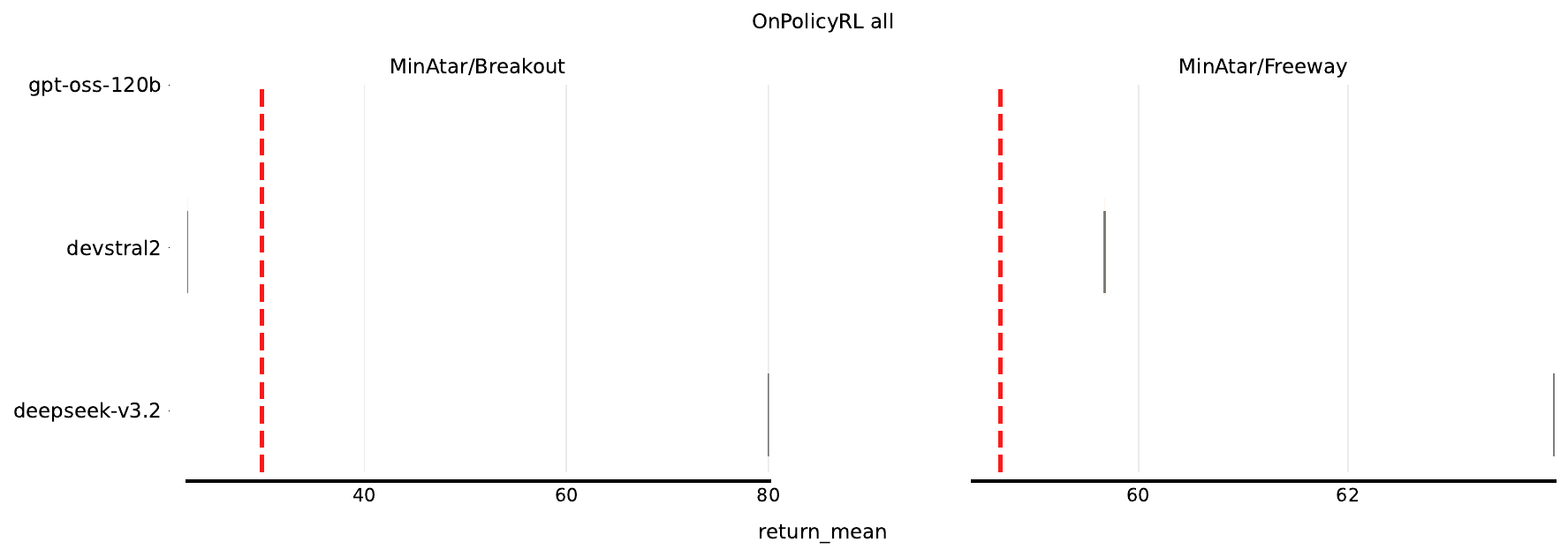}%
\hfill%
\includegraphics[width=0.48\textwidth]{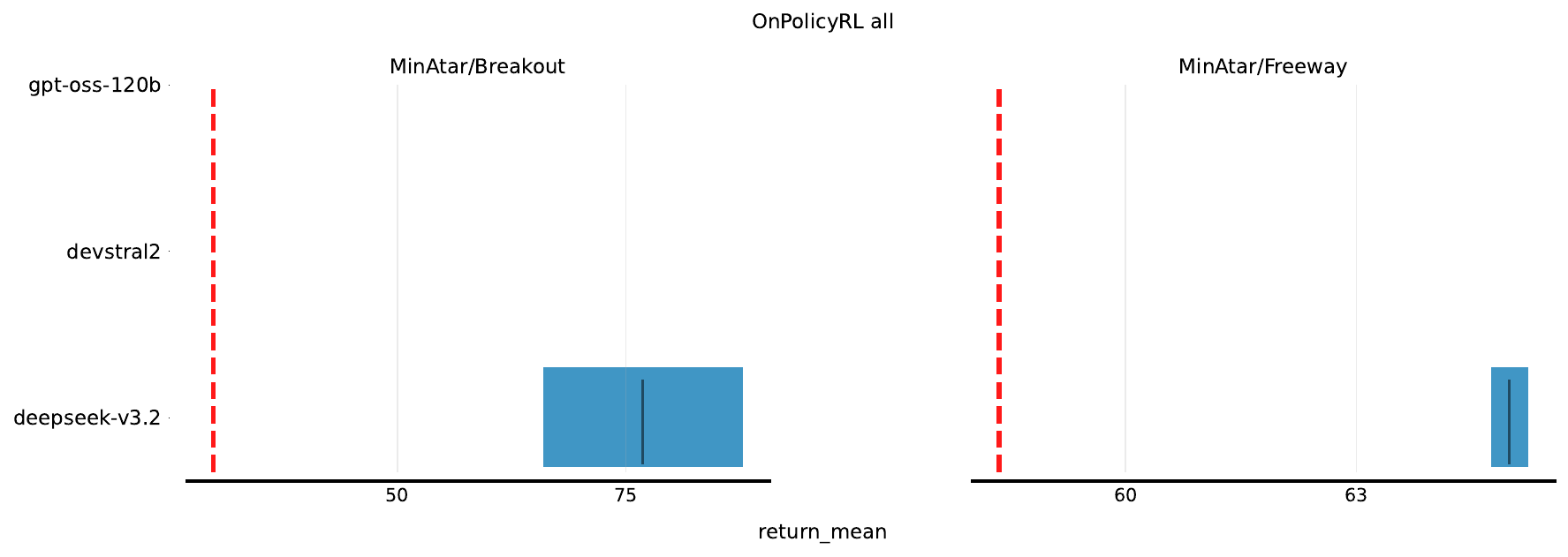}%
\\[0.5em]
\includegraphics[width=0.48\textwidth]{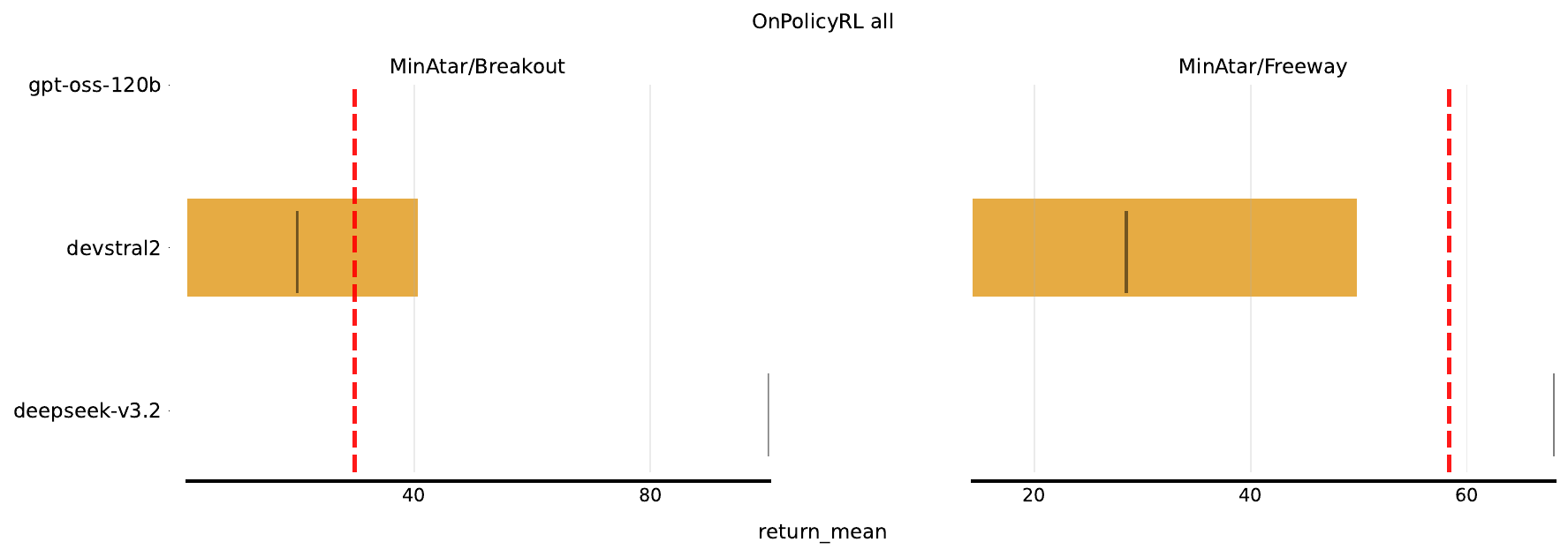}%
\hfill%
\includegraphics[width=0.48\textwidth]{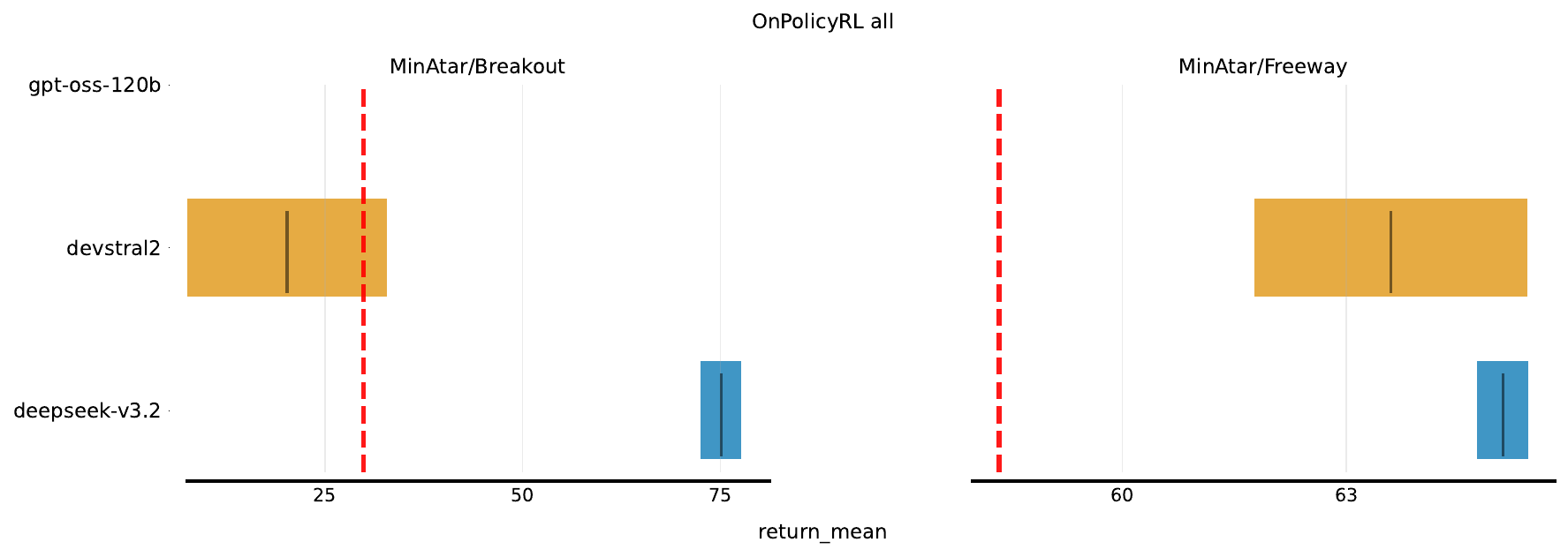}%
\\[0.5em]
\includegraphics[width=0.48\textwidth]{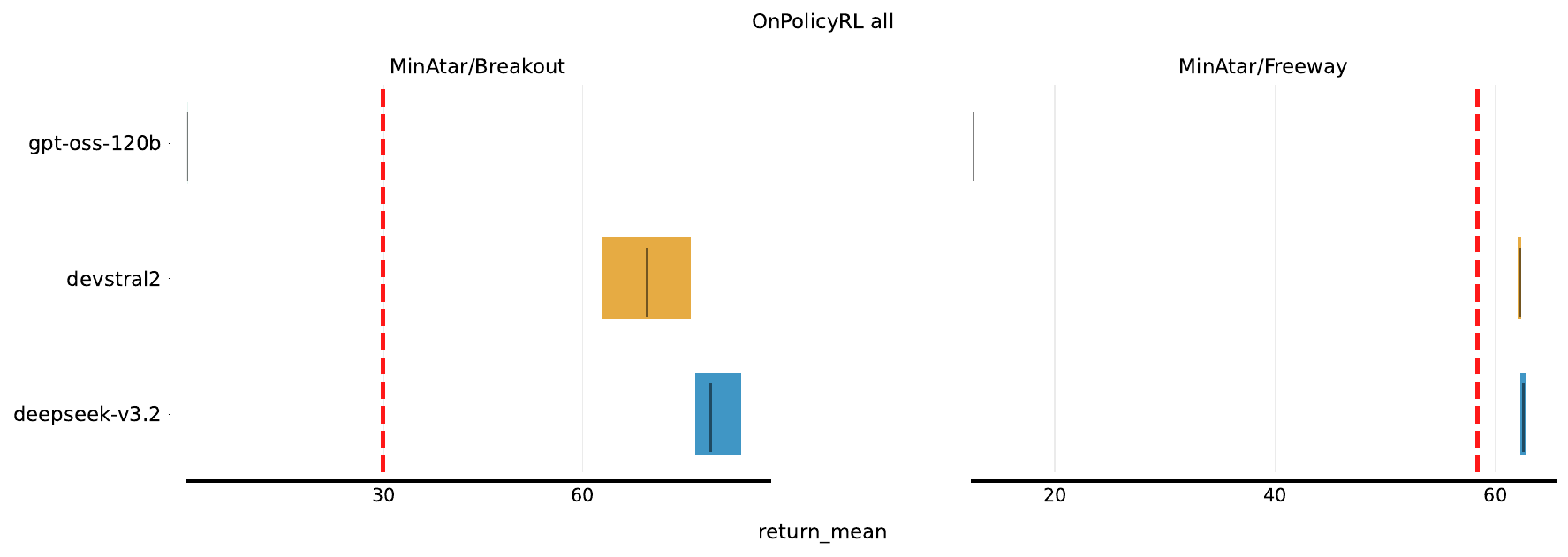}%
\hfill%
\includegraphics[width=0.48\textwidth]{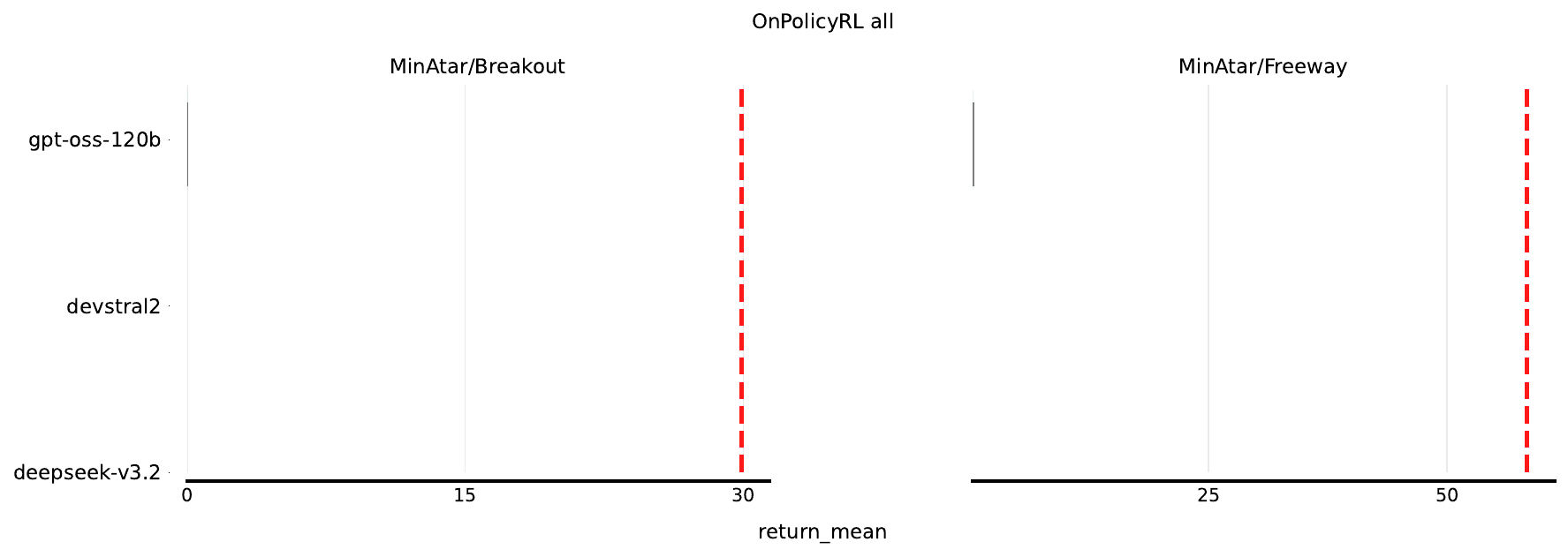}%
\caption{On-Policy RL Combinations results on Meta-Train tasks. (Part 1/4)}
\label{fig:on_policy_id_1}
\end{figure}
\clearpage

\begin{figure}[htbp]
\centering
\setlength{\lineskip}{0pt}
\includegraphics[width=0.48\textwidth]{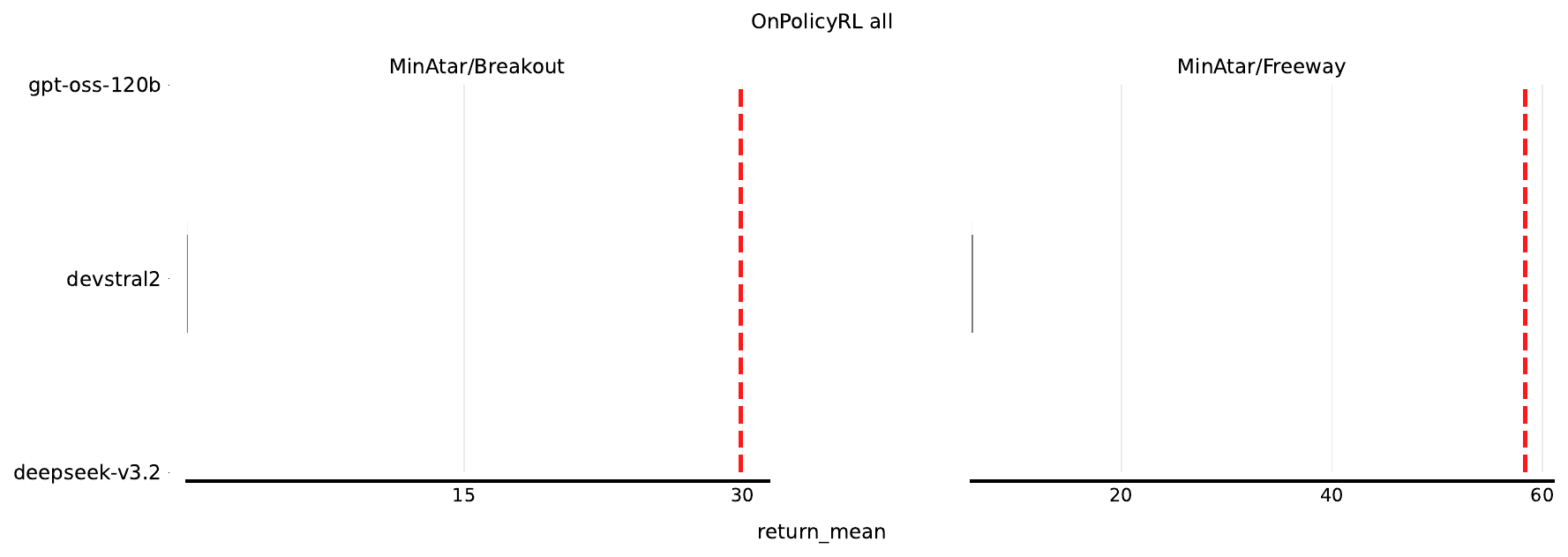}%
\hfill%
\includegraphics[width=0.48\textwidth]{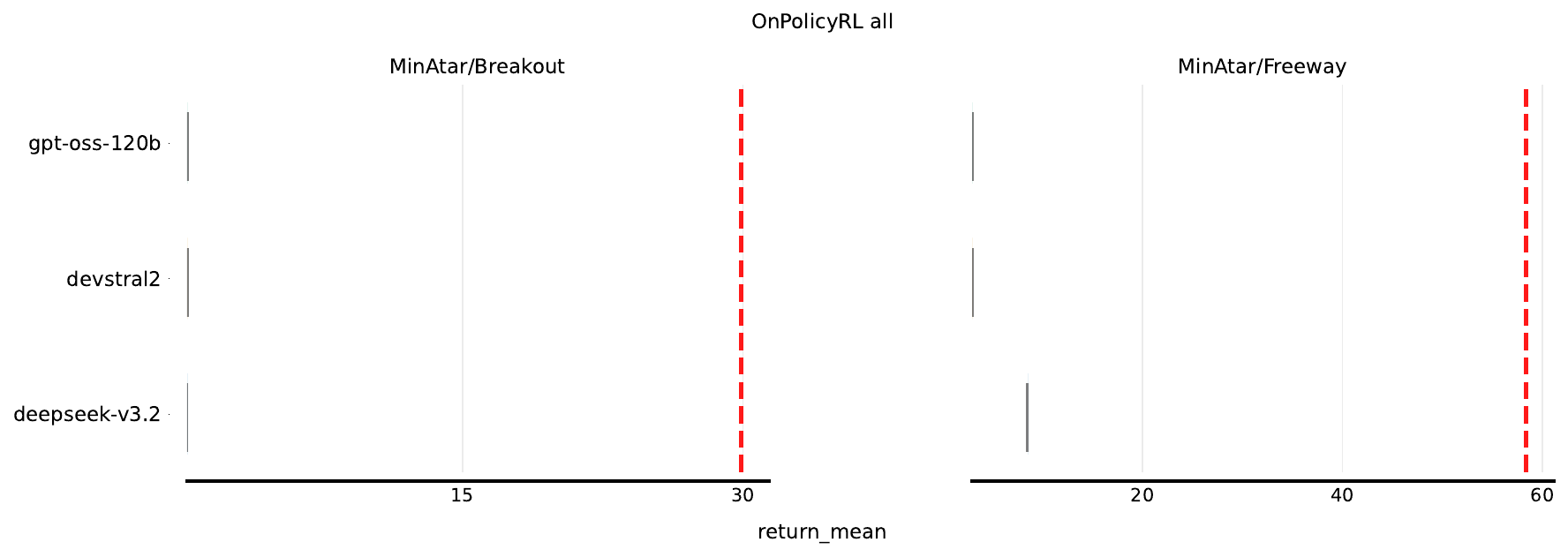}%
\\[0.5em]
\includegraphics[width=0.48\textwidth]{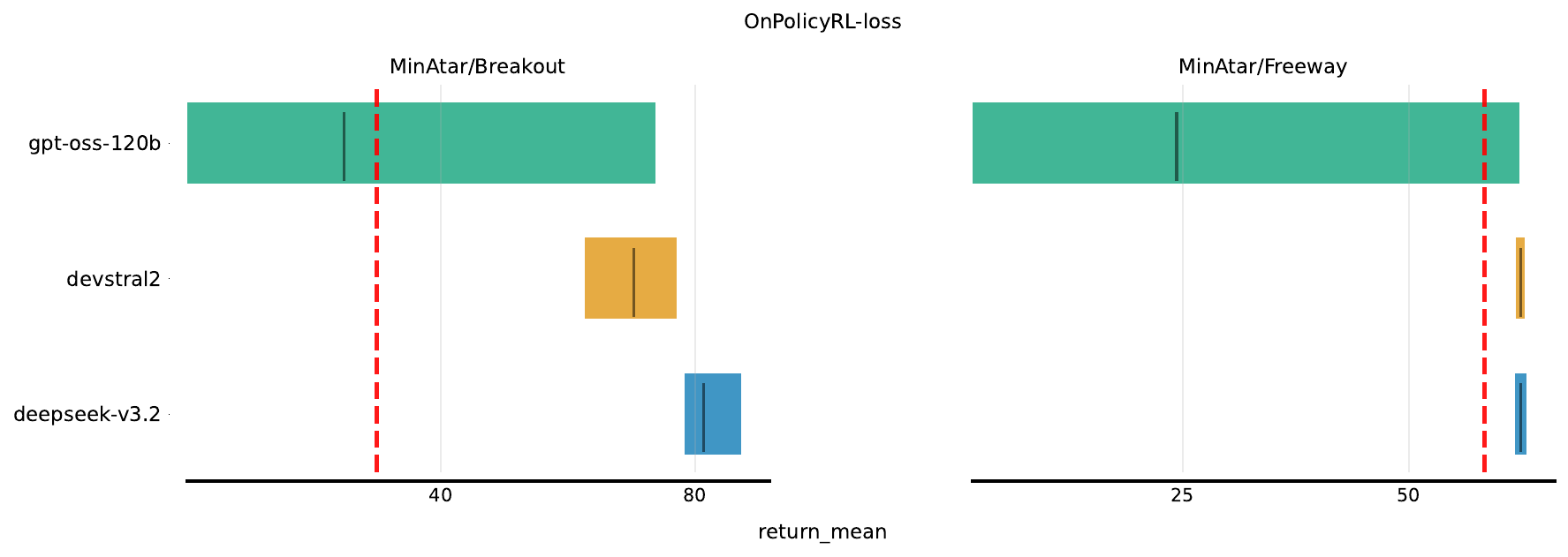}%
\hfill%
\includegraphics[width=0.48\textwidth]{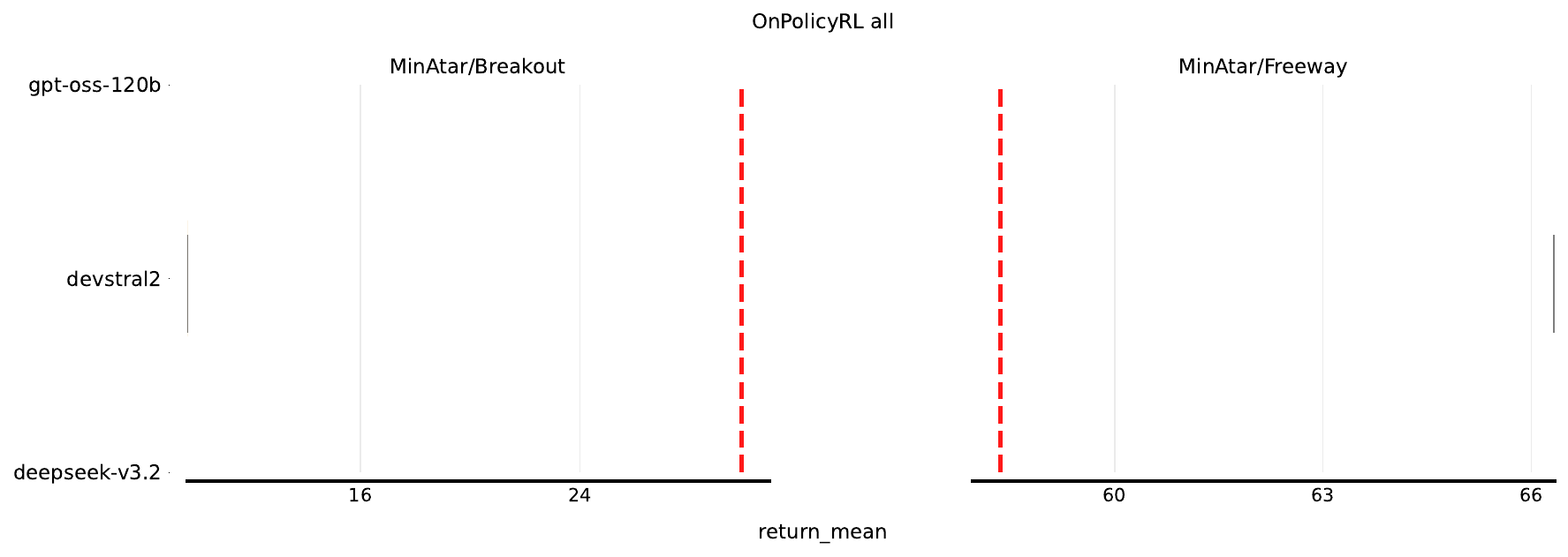}%
\\[0.5em]
\includegraphics[width=0.48\textwidth]{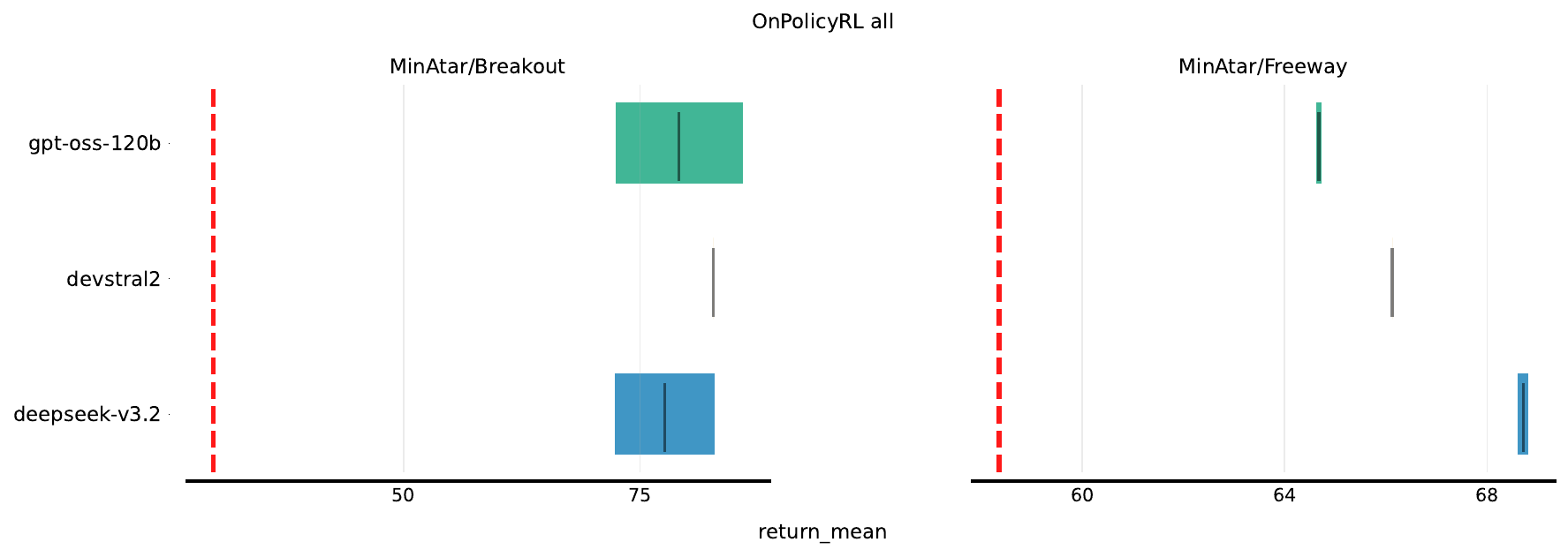}%
\hfill%
\includegraphics[width=0.48\textwidth]{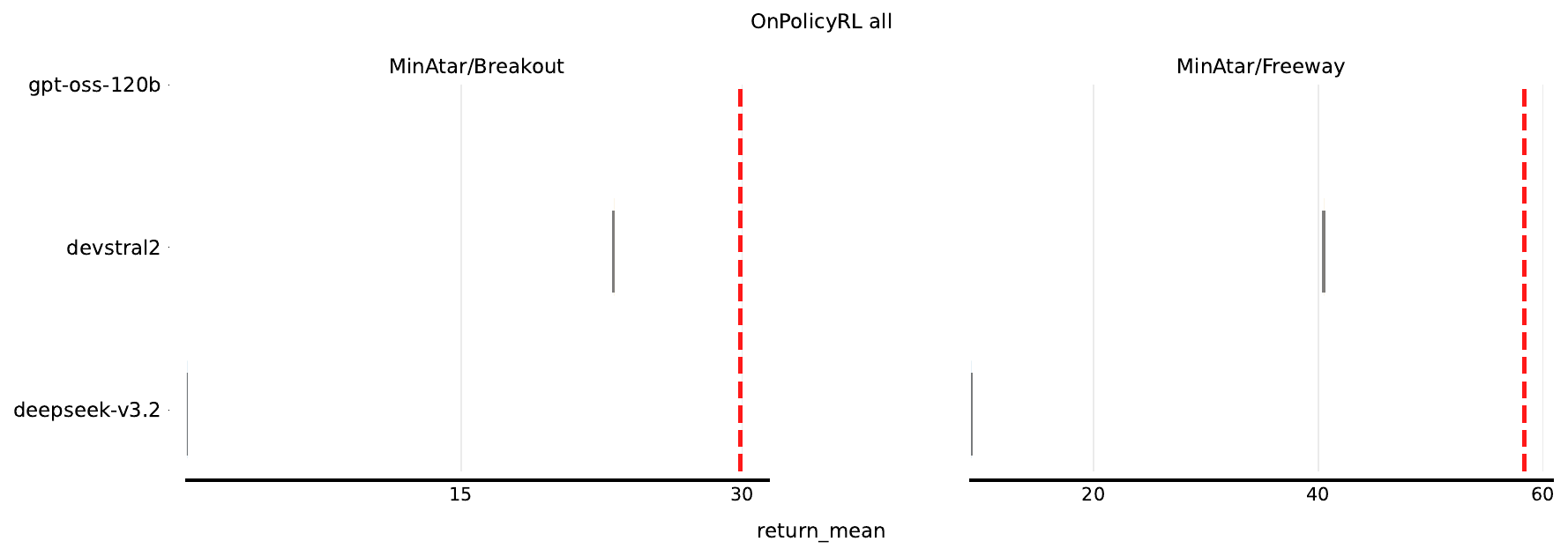}%
\\[0.5em]
\includegraphics[width=0.48\textwidth]{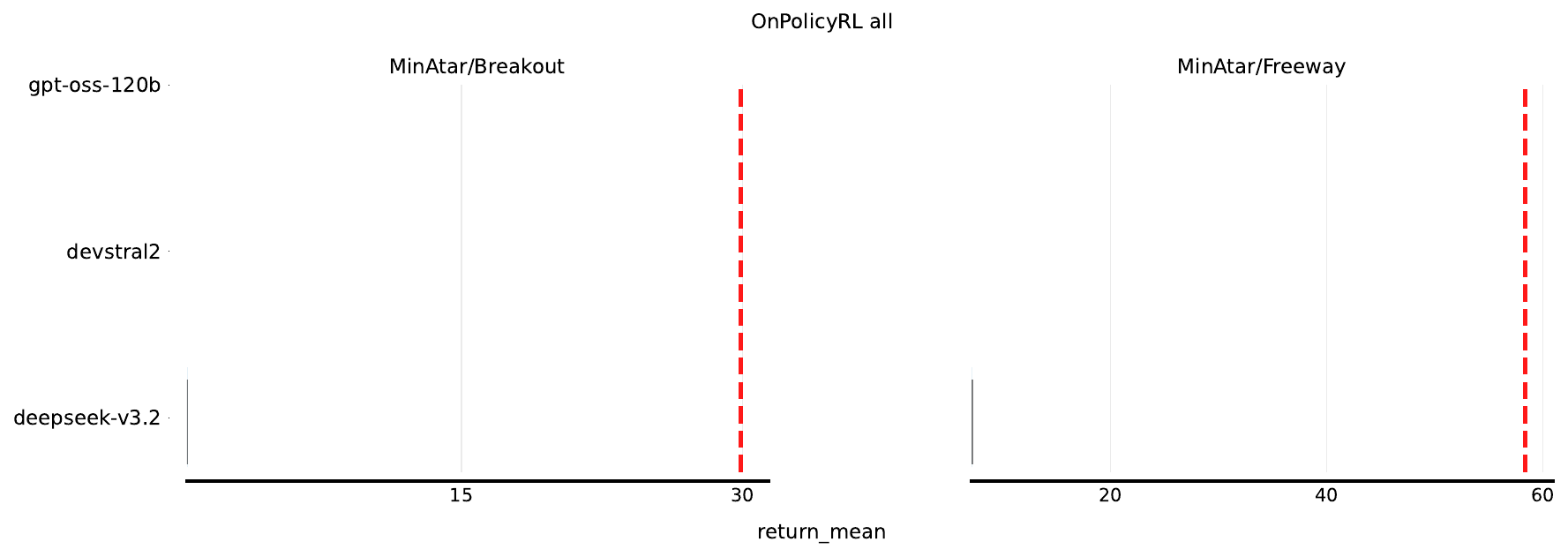}%
\hfill%
\includegraphics[width=0.48\textwidth]{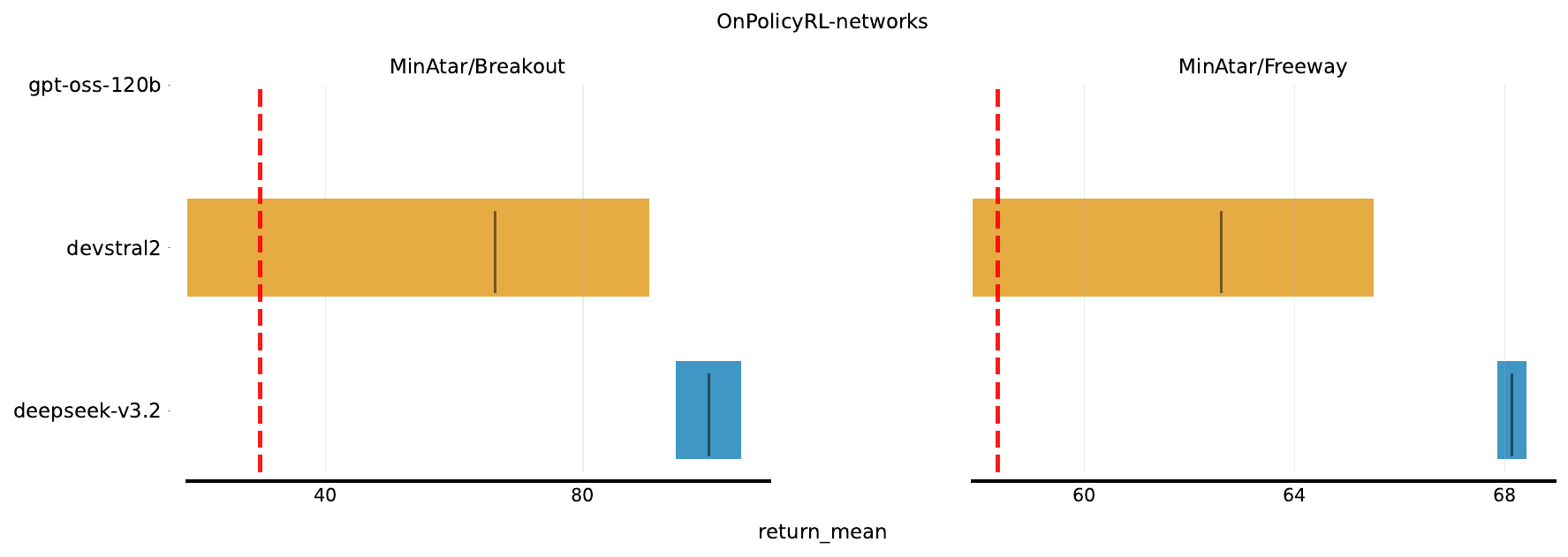}%
\\[0.5em]
\includegraphics[width=0.48\textwidth]{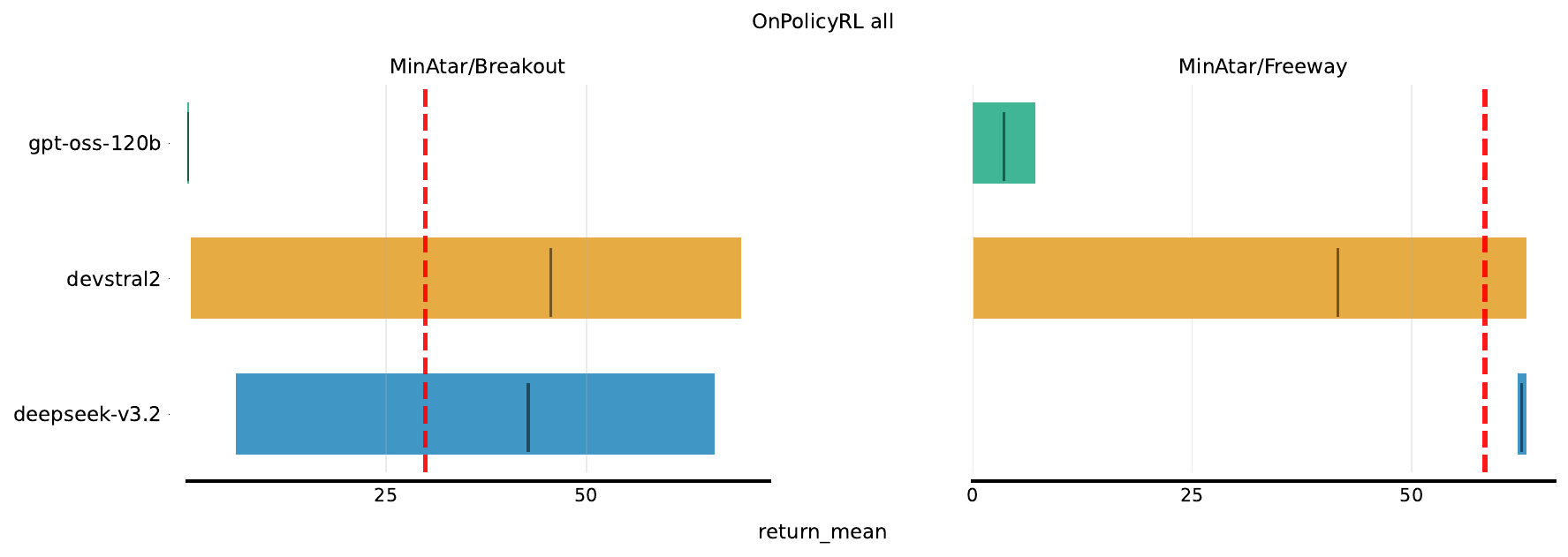}%
\hfill%
\includegraphics[width=0.48\textwidth]{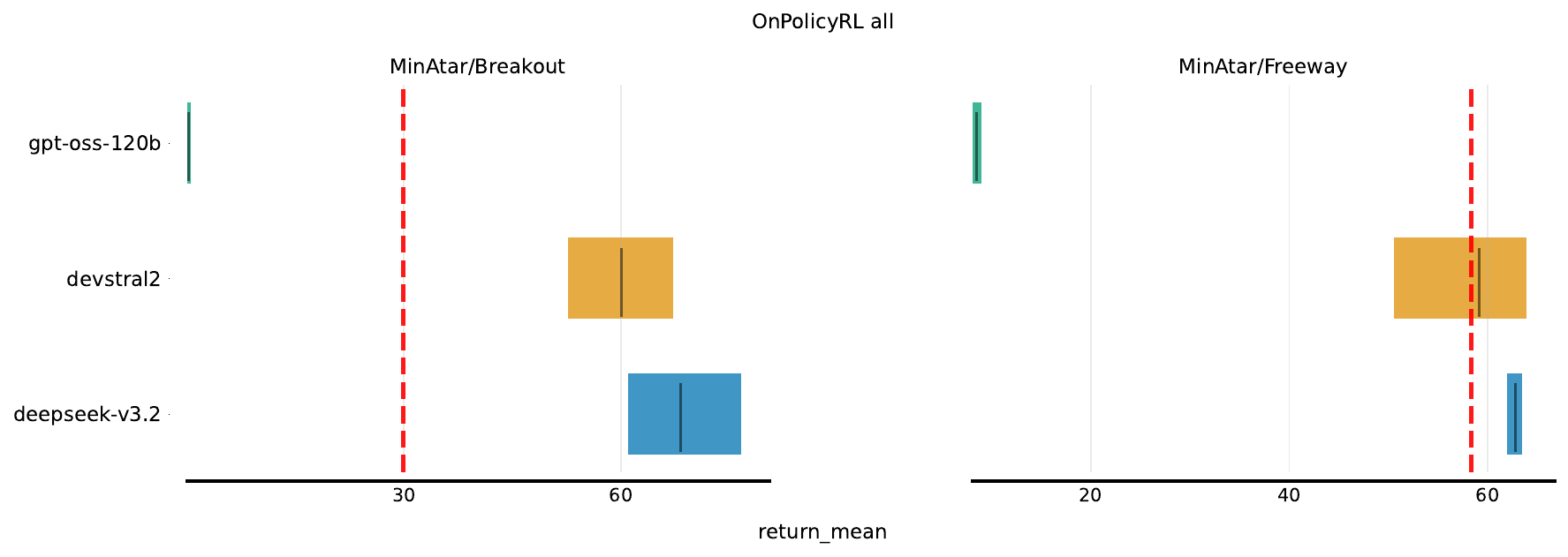}%
\\[0.5em]
\includegraphics[width=0.48\textwidth]{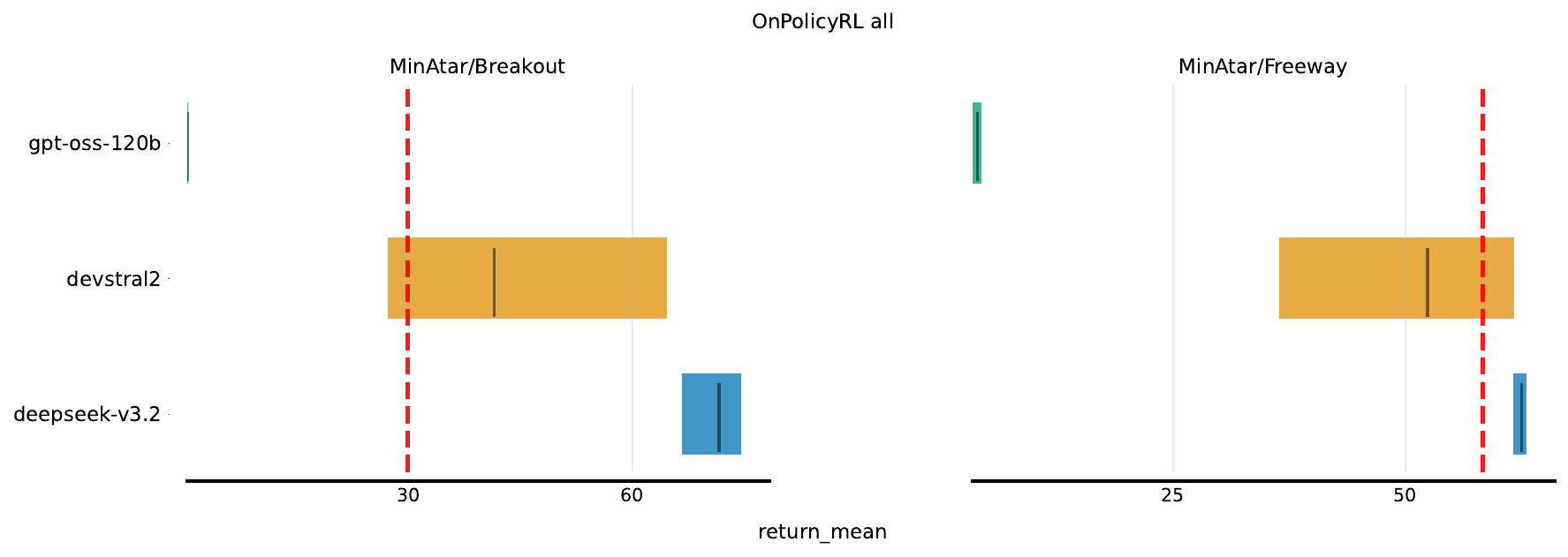}%
\hfill%
\includegraphics[width=0.48\textwidth]{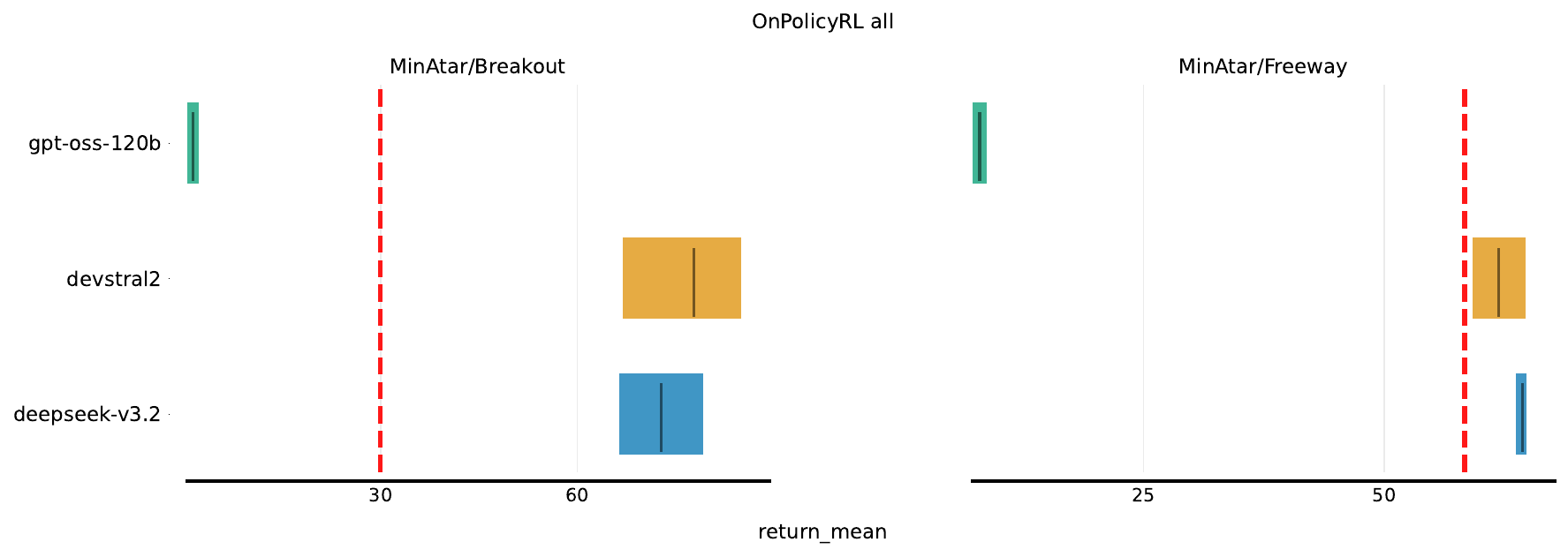}%
\caption{On-Policy RL Combinations results on Meta-Train tasks. (Part 2/4)}
\label{fig:on_policy_id_2}
\end{figure}
\clearpage

\begin{figure}[htbp]
\centering
\setlength{\lineskip}{0pt}
\includegraphics[width=0.48\textwidth]{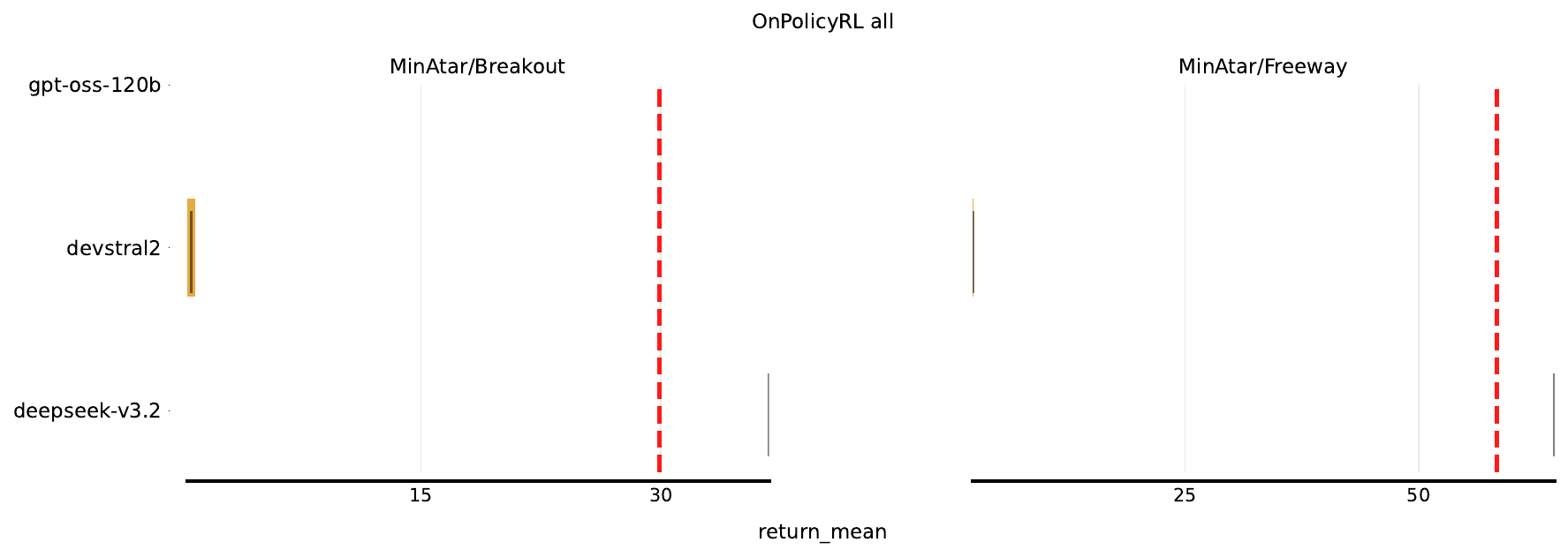}%
\hfill%
\includegraphics[width=0.48\textwidth]{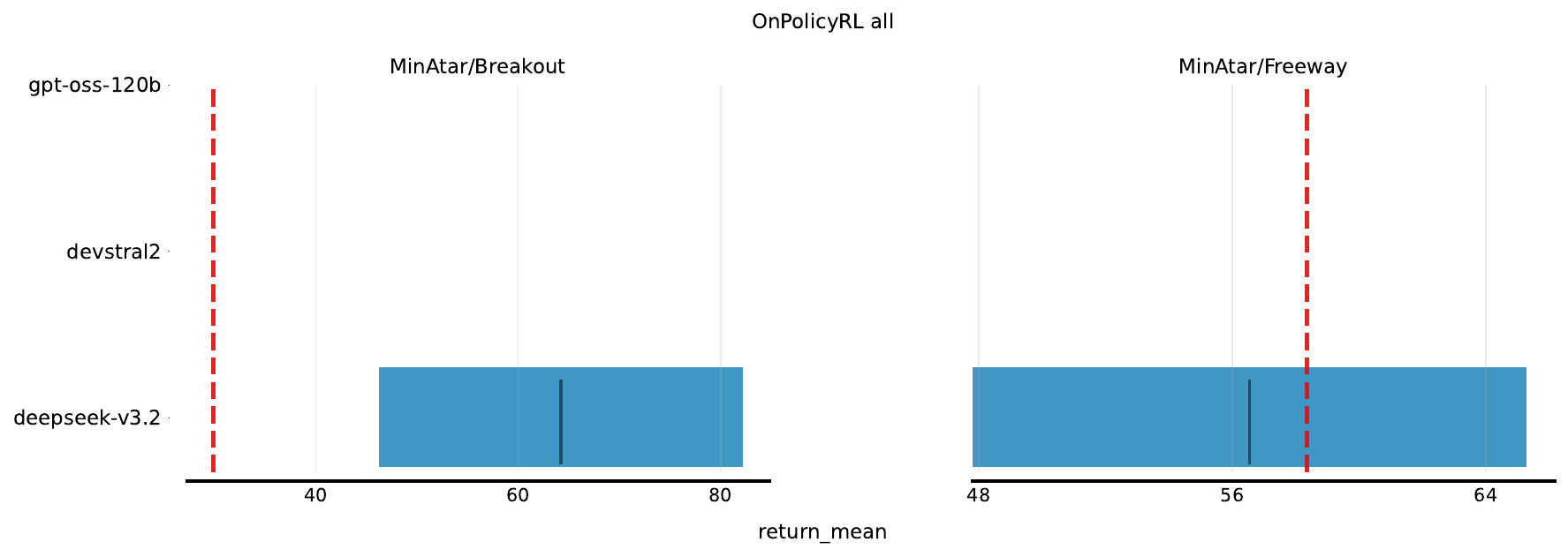}%
\\[0.5em]
\includegraphics[width=0.48\textwidth]{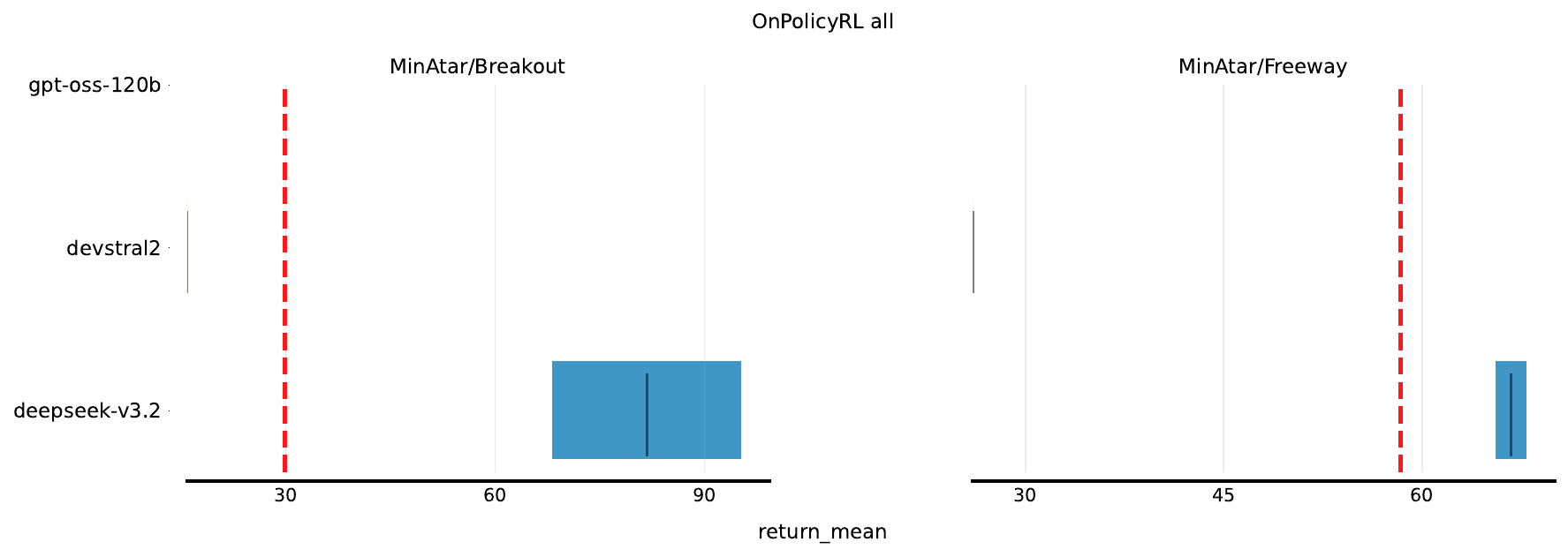}%
\hfill%
\includegraphics[width=0.48\textwidth]{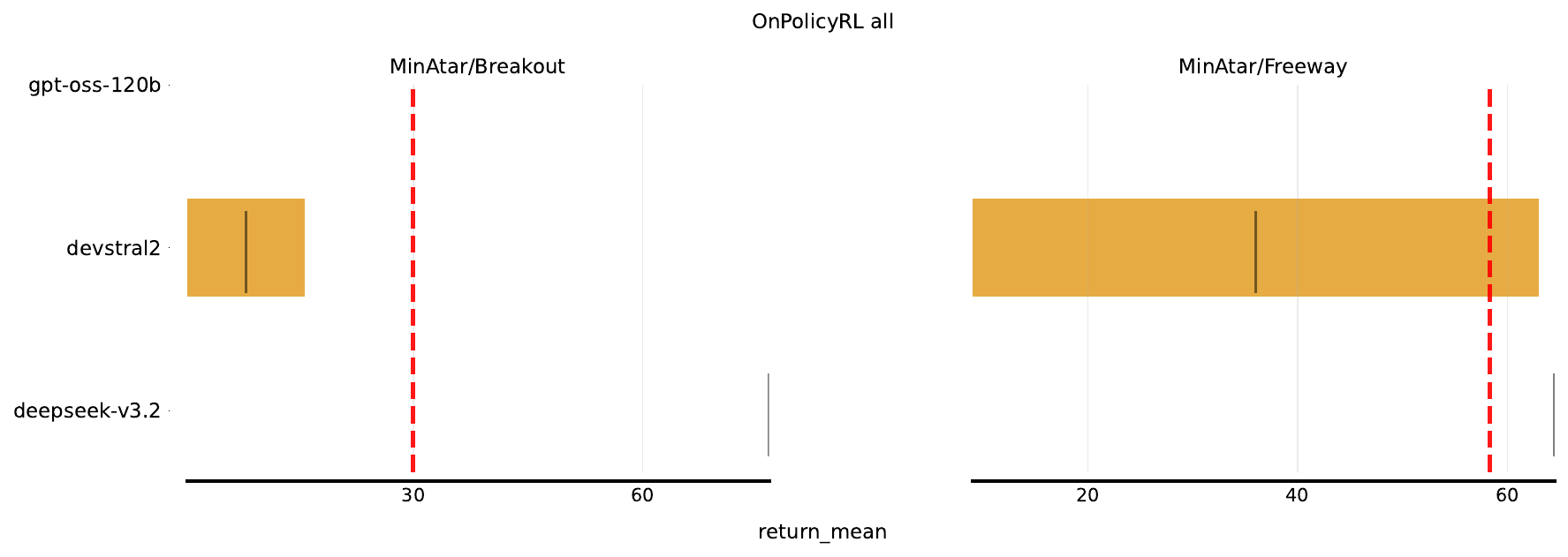}%
\\[0.5em]
\includegraphics[width=0.48\textwidth]{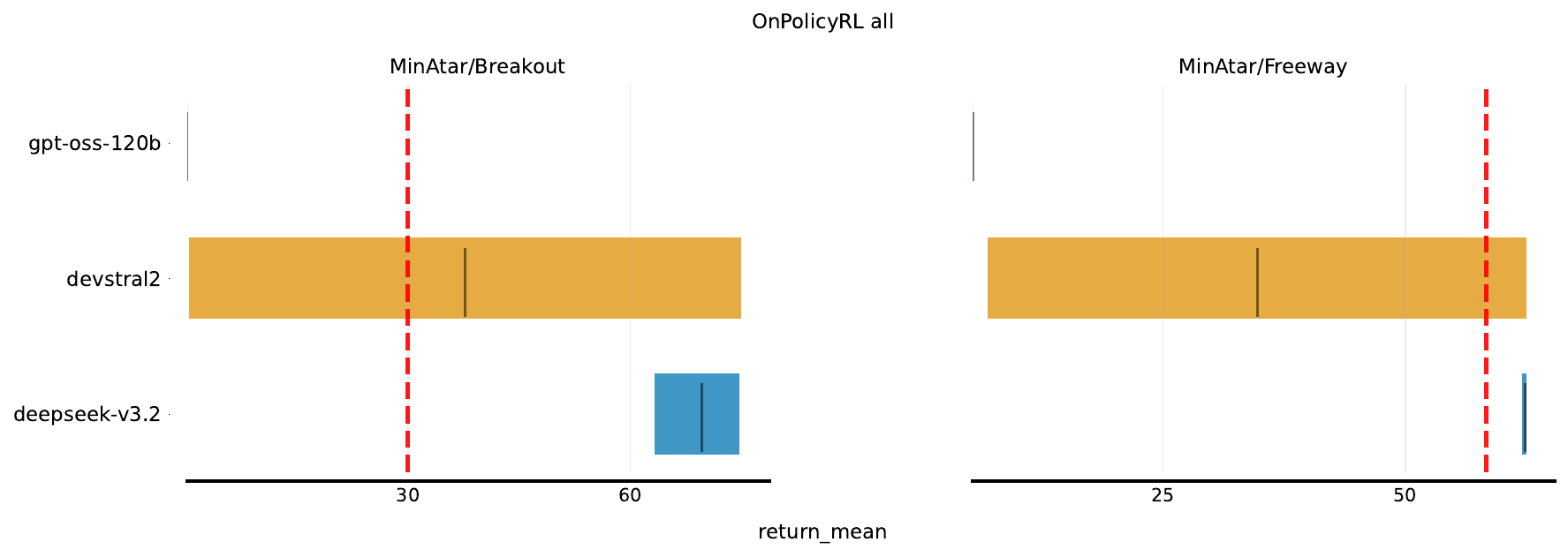}%
\hfill%
\includegraphics[width=0.48\textwidth]{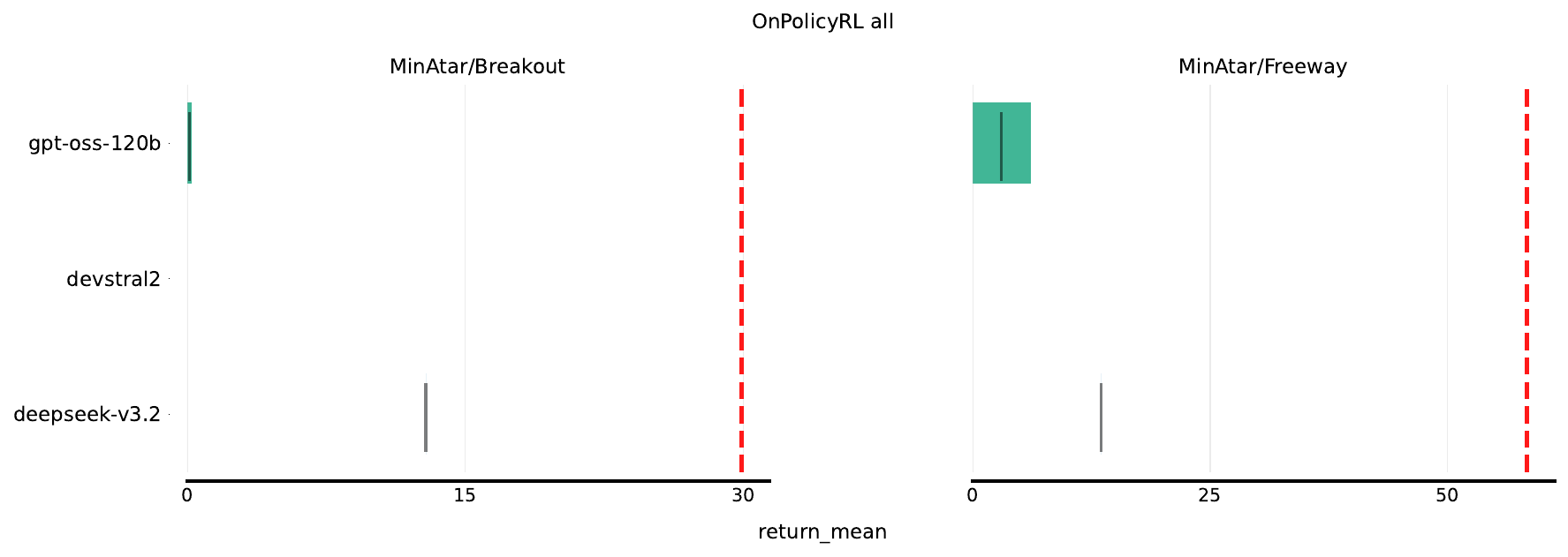}%
\\[0.5em]
\includegraphics[width=0.48\textwidth]{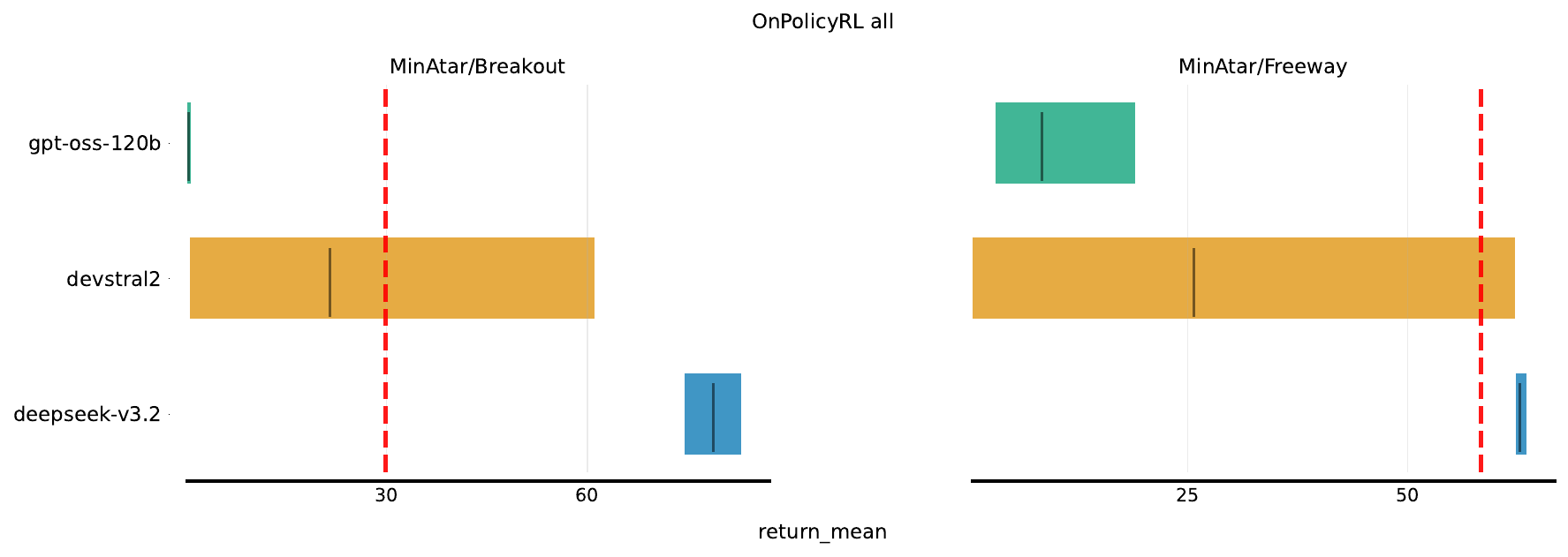}%
\hfill%
\includegraphics[width=0.48\textwidth]{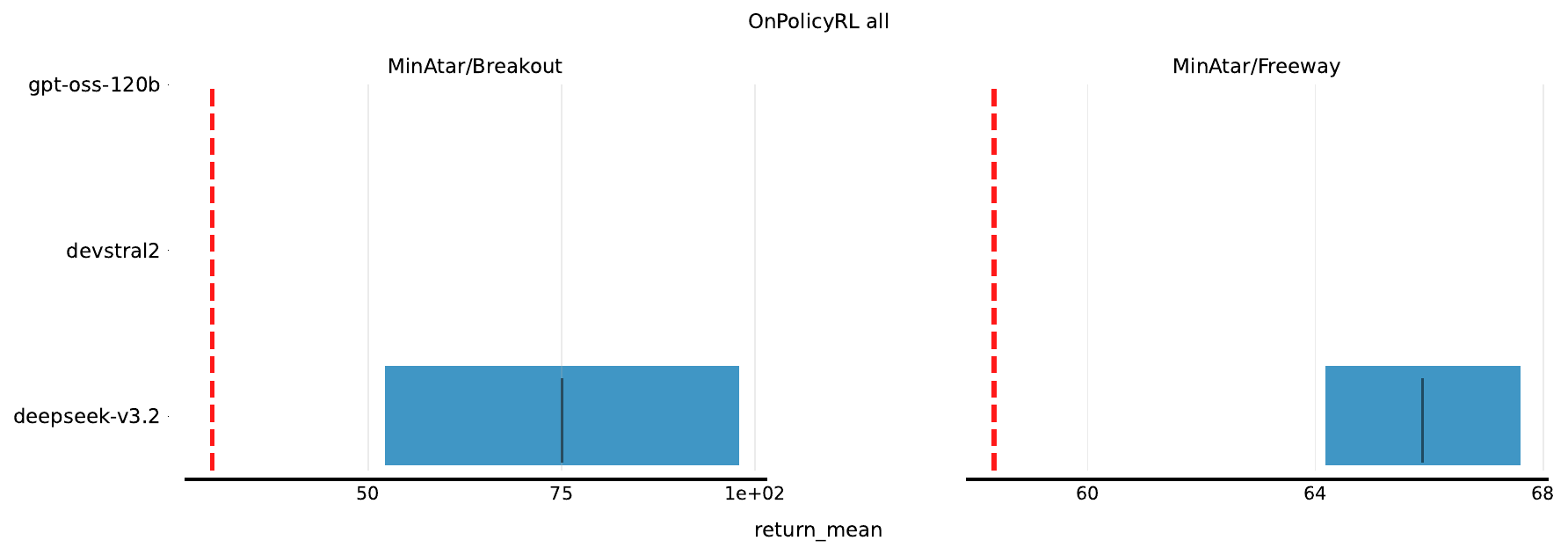}%
\\[0.5em]
\includegraphics[width=0.48\textwidth]{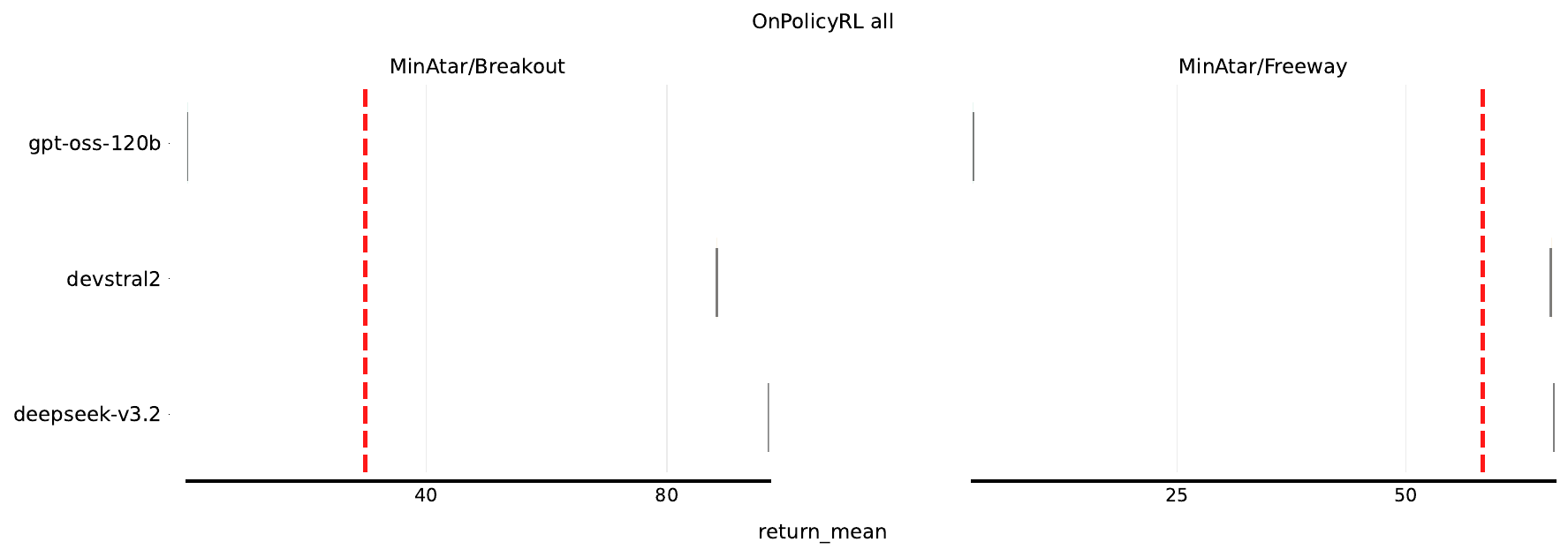}%
\hfill%
\includegraphics[width=0.48\textwidth]{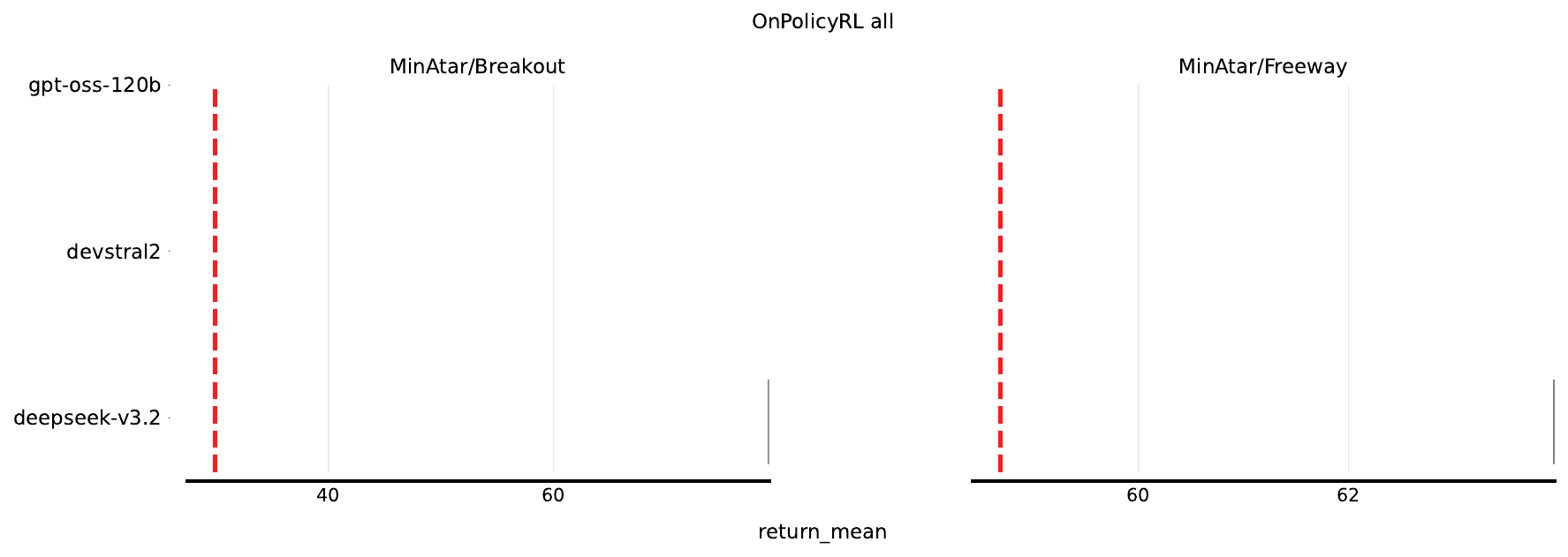}%
\\[0.5em]
\includegraphics[width=0.48\textwidth]{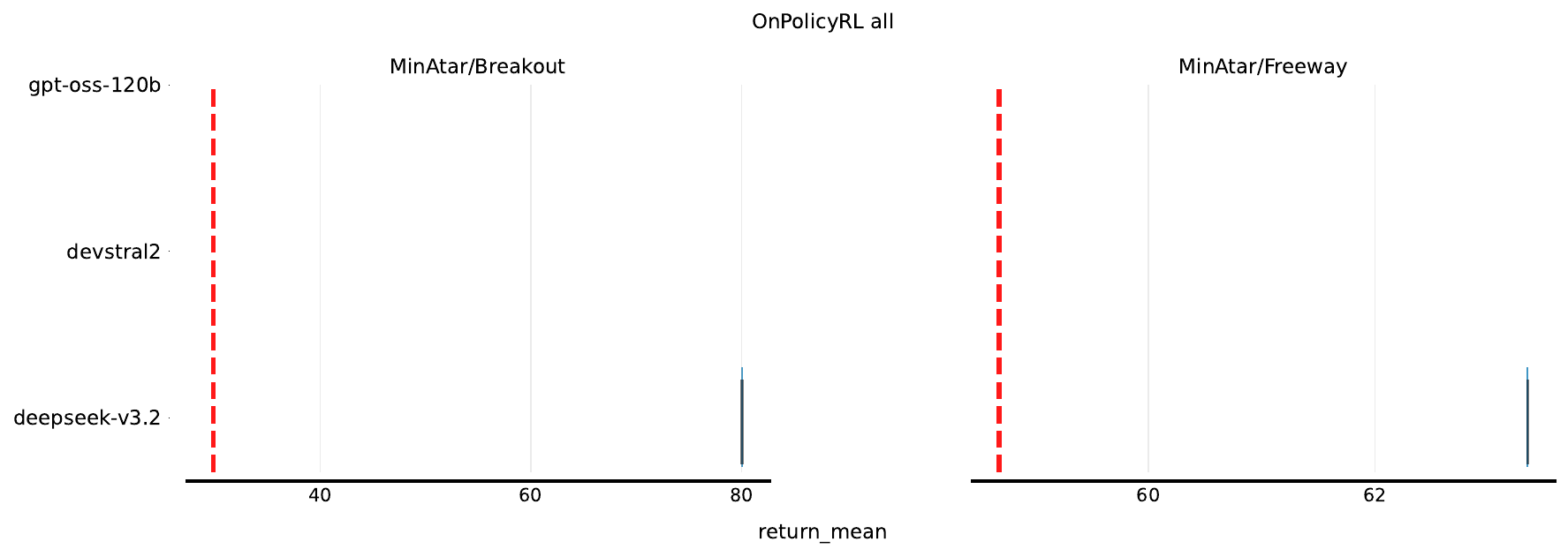}%
\hfill%
\includegraphics[width=0.48\textwidth]{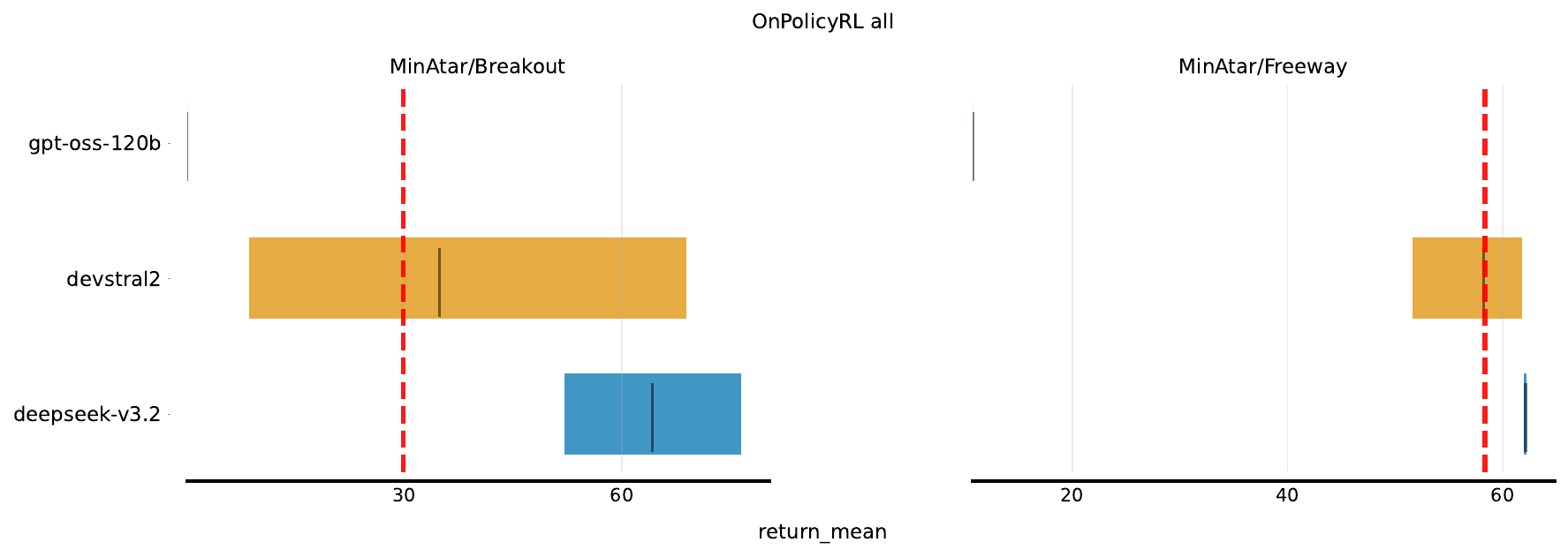}%
\caption{On-Policy RL Combinations results on Meta-Train tasks. (Part 3/4)}
\label{fig:on_policy_id_3}
\end{figure}
\clearpage

\begin{figure}[htbp]
\centering
\setlength{\lineskip}{0pt}
\includegraphics[width=0.48\textwidth]{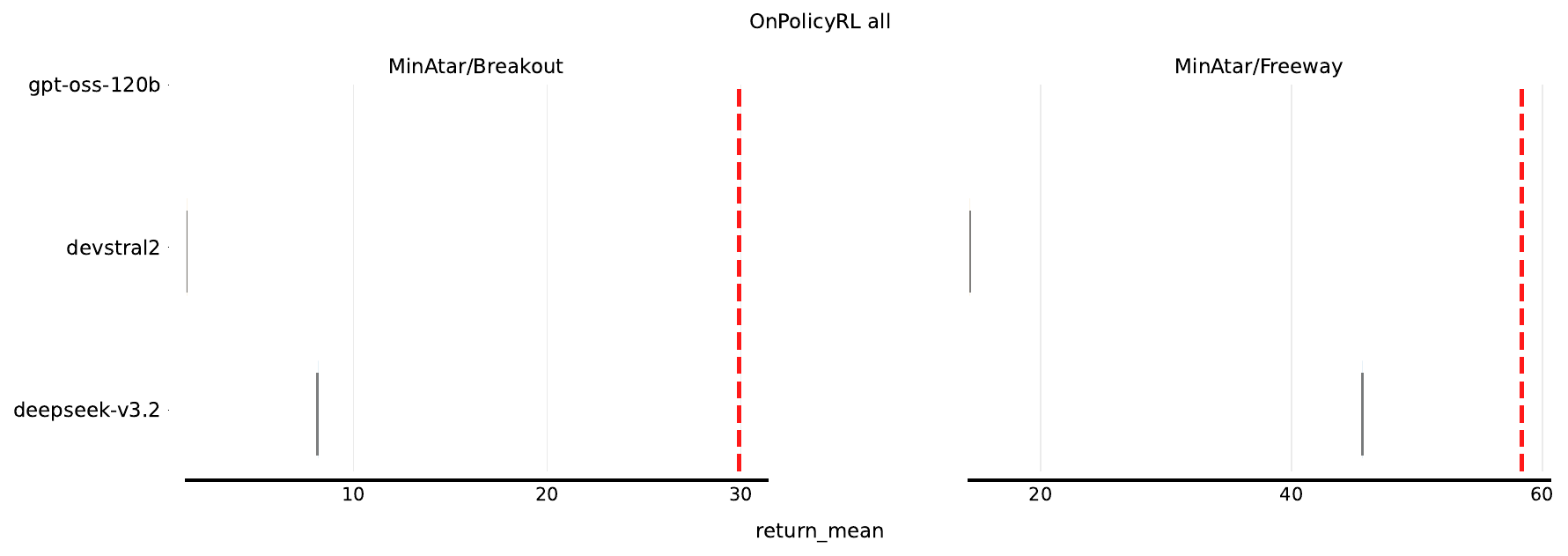}%
\hfill%
\includegraphics[width=0.48\textwidth]{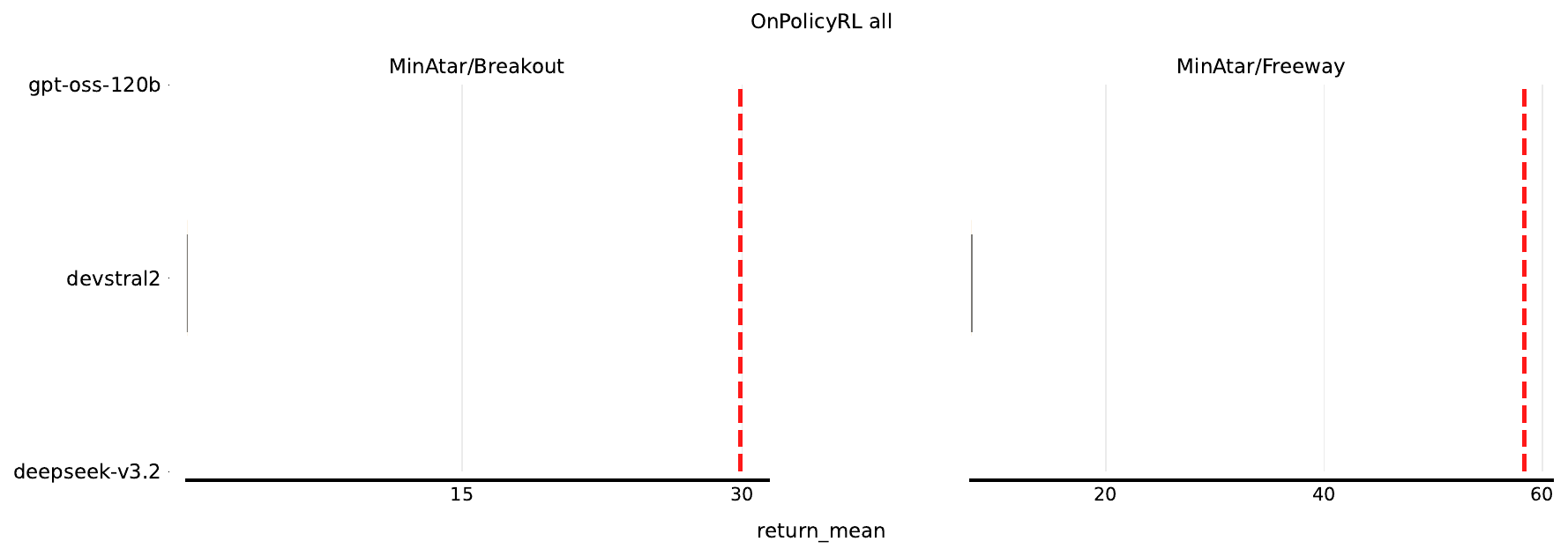}%
\\[0.5em]
\includegraphics[width=0.48\textwidth]{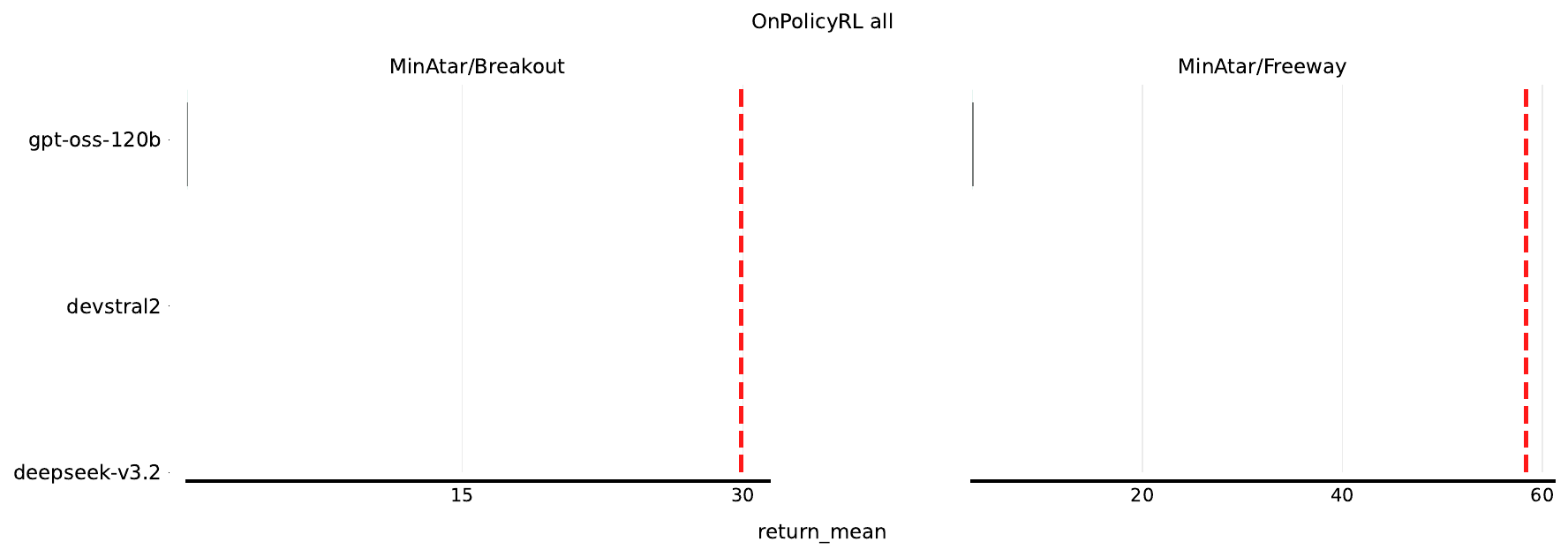}%
\hfill%
\includegraphics[width=0.48\textwidth]{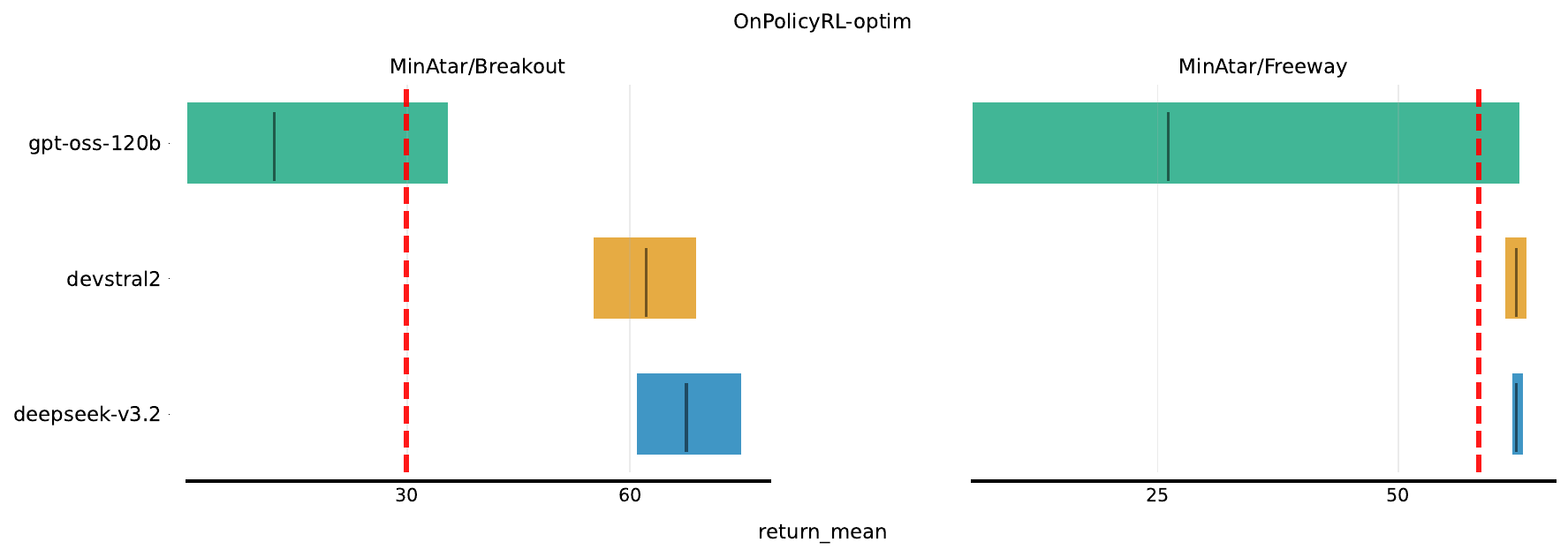}%
\\[0.5em]
\includegraphics[width=0.48\textwidth]{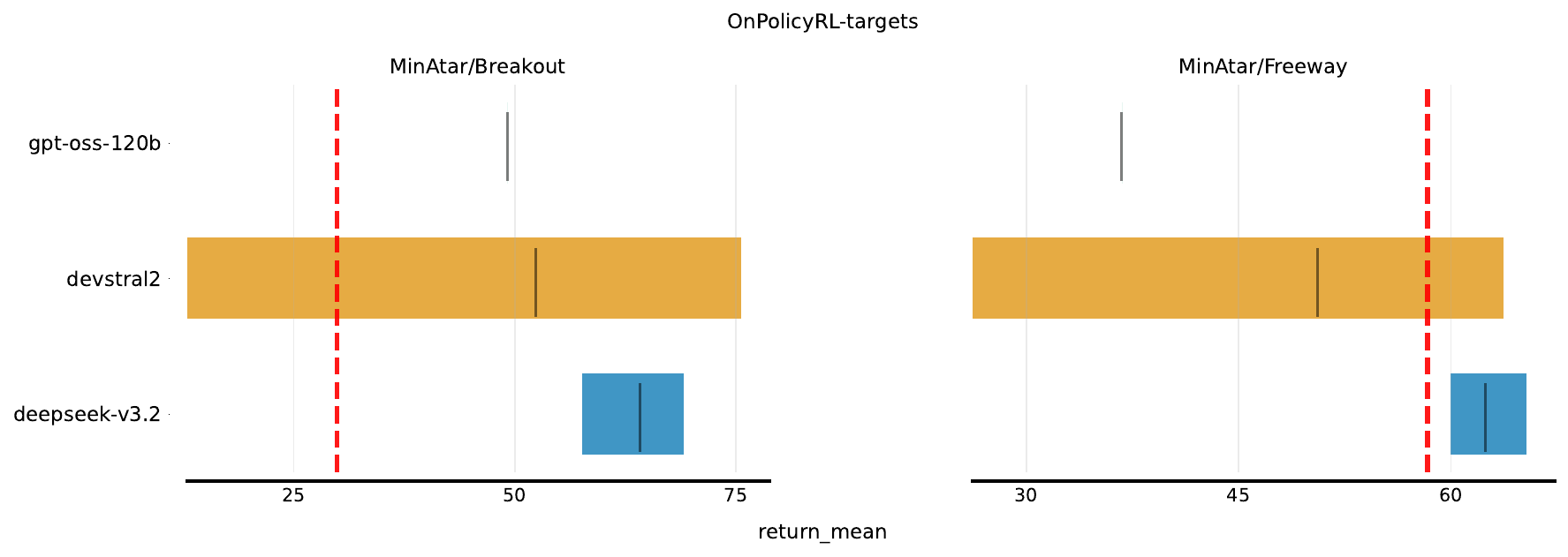}%
\hfill%
\includegraphics[width=0.48\textwidth]{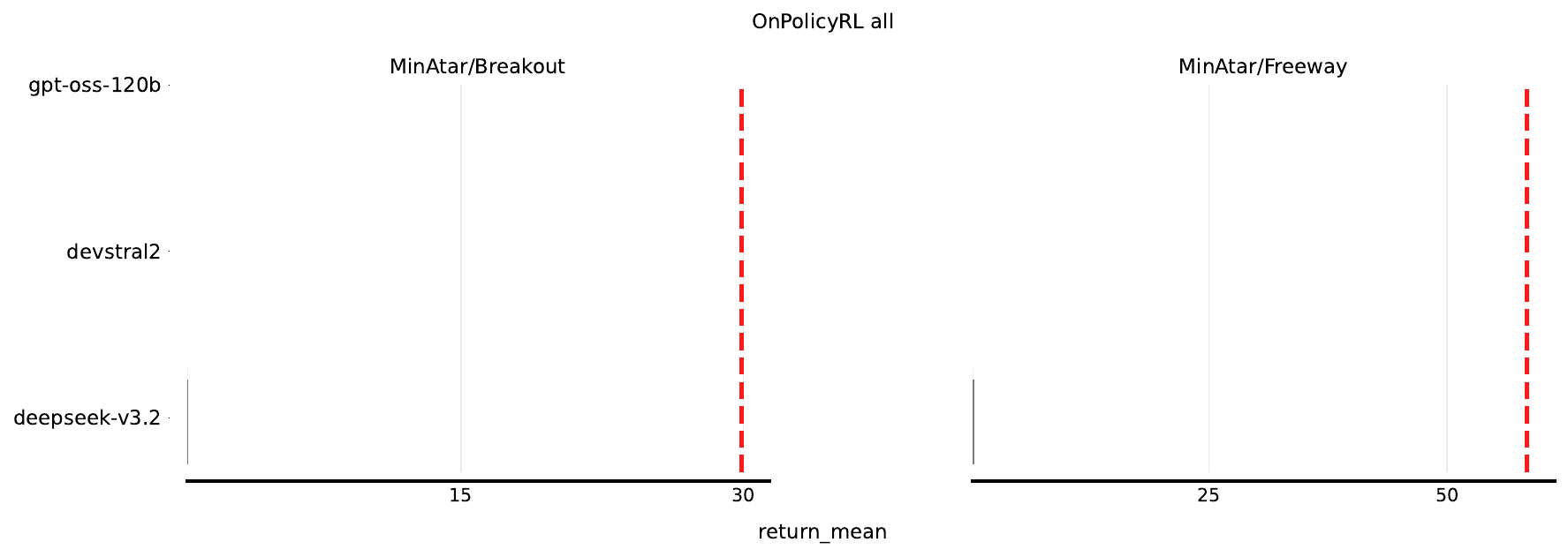}%
\\[0.5em]
\includegraphics[width=0.48\textwidth]{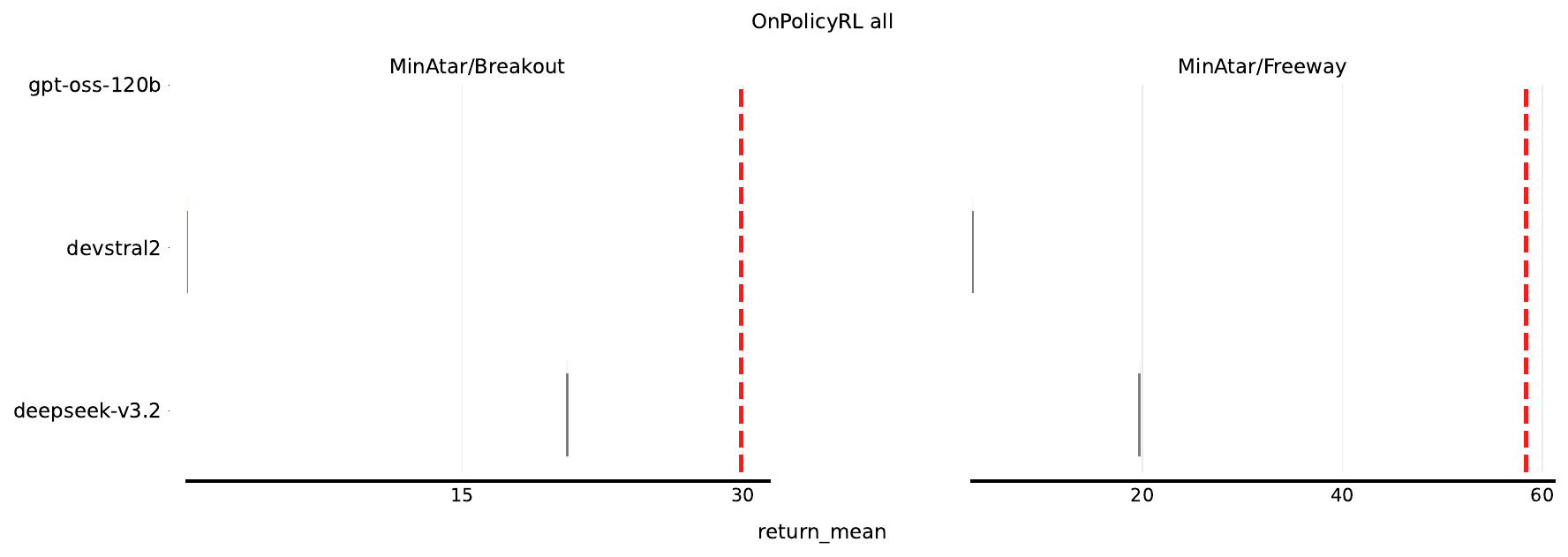}%
\hfill%
\includegraphics[width=0.48\textwidth]{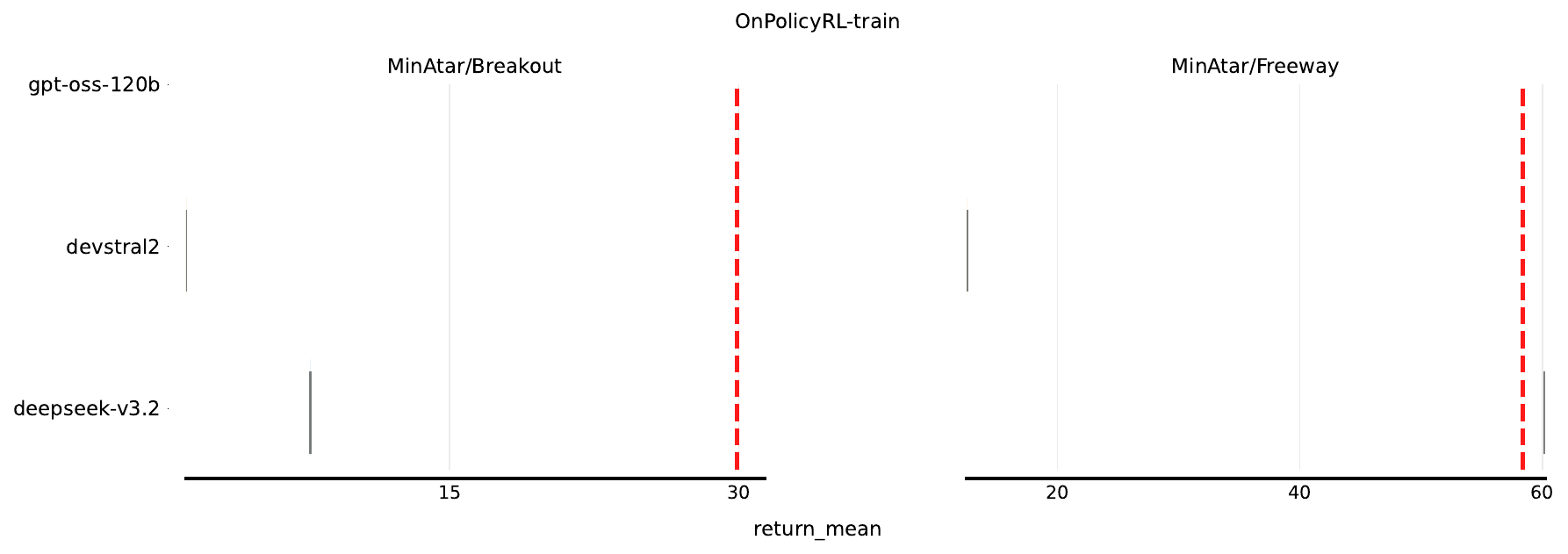}%
\caption{On-Policy RL Combinations results on Meta-Train tasks. (Part 4/4)}
\label{fig:on_policy_id_4}
\end{figure}
\clearpage

\subsection{On-Policy RL Combinations -- Meta-Test}
\label{sec:on_policy_mt}

\begin{figure}[htbp]
\centering
\setlength{\lineskip}{0pt}
\includegraphics[width=0.48\textwidth]{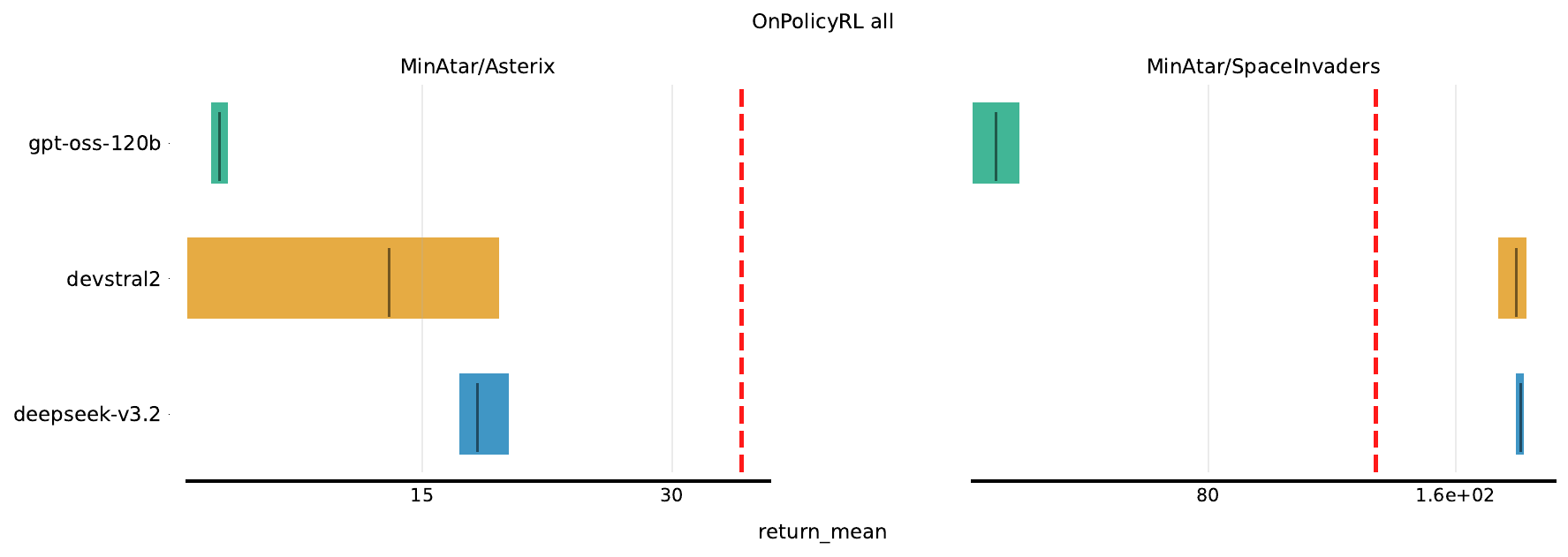}%
\hfill%
\includegraphics[width=0.48\textwidth]{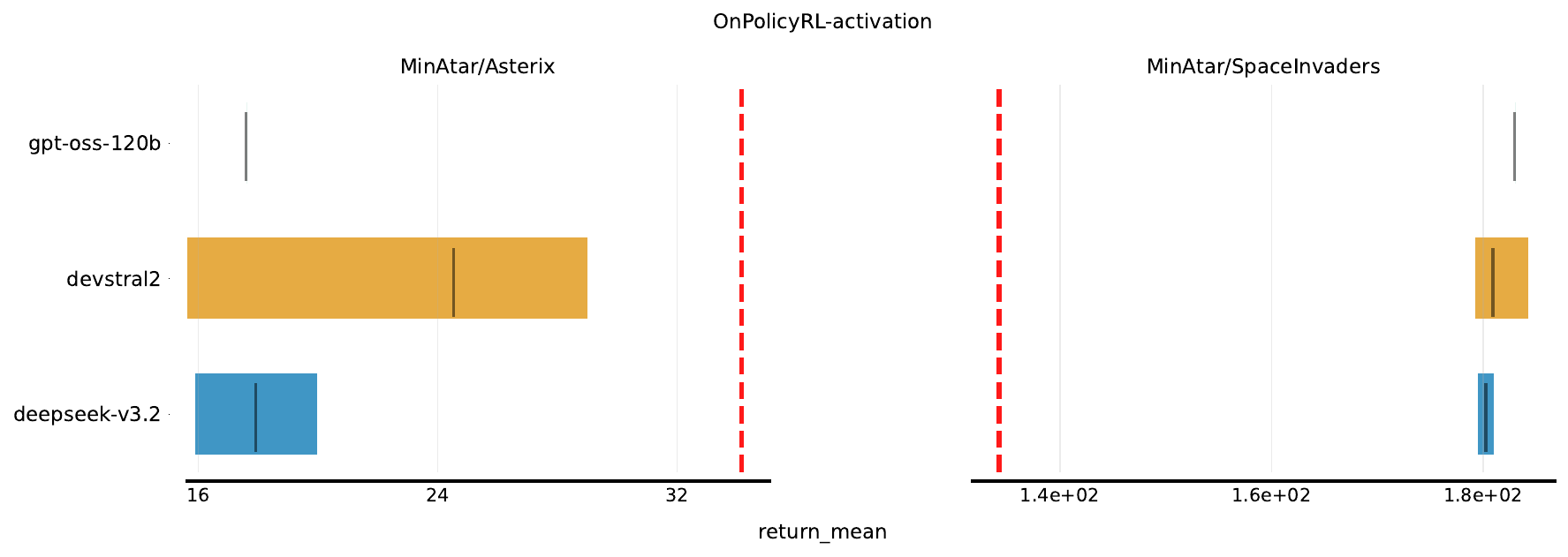}%
\\[0.5em]
\includegraphics[width=0.48\textwidth]{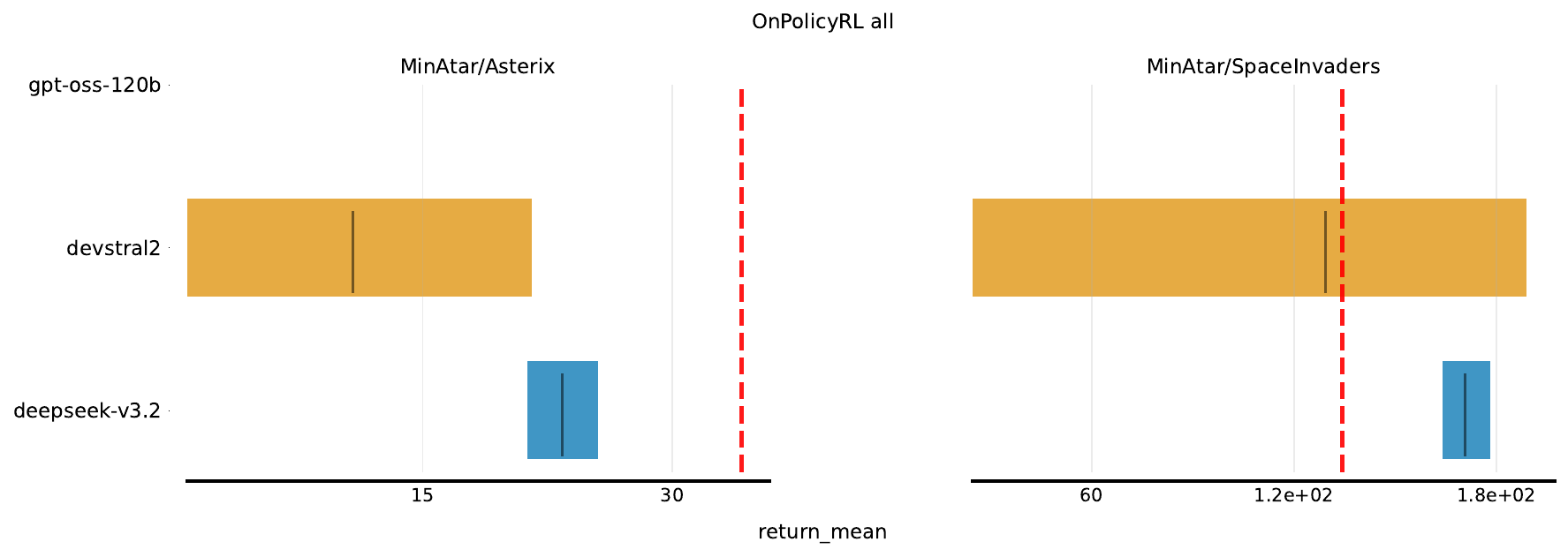}%
\hfill%
\includegraphics[width=0.48\textwidth]{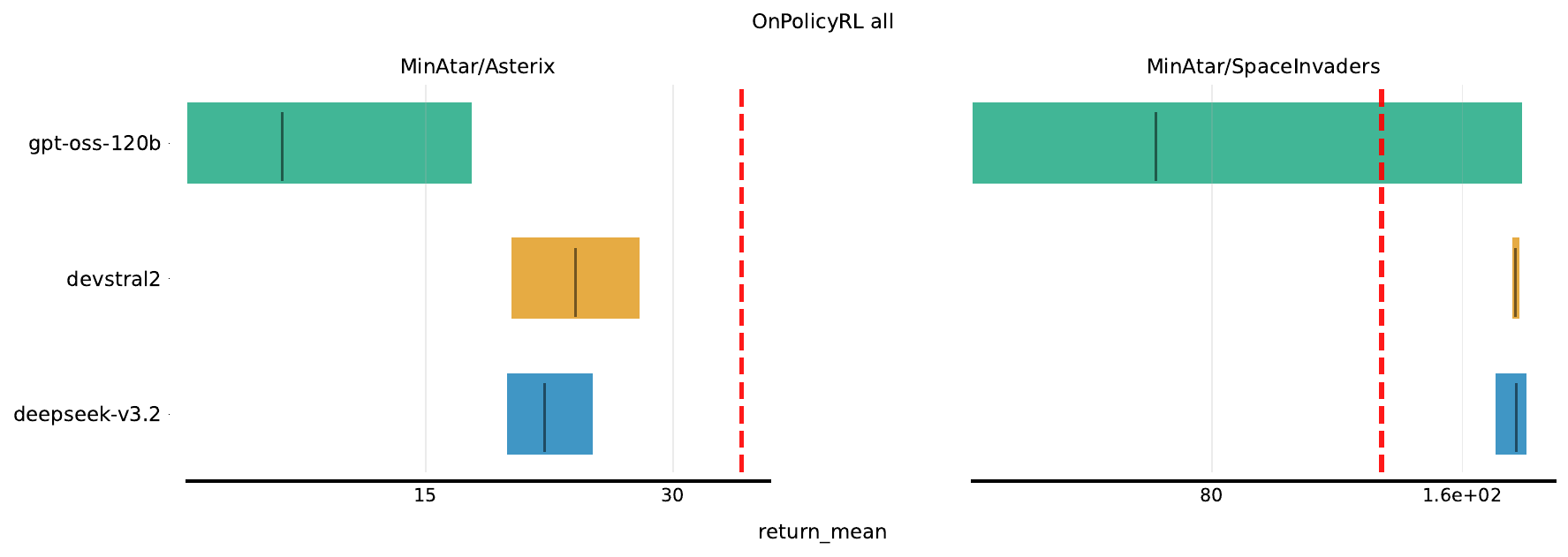}%
\\[0.5em]
\includegraphics[width=0.48\textwidth]{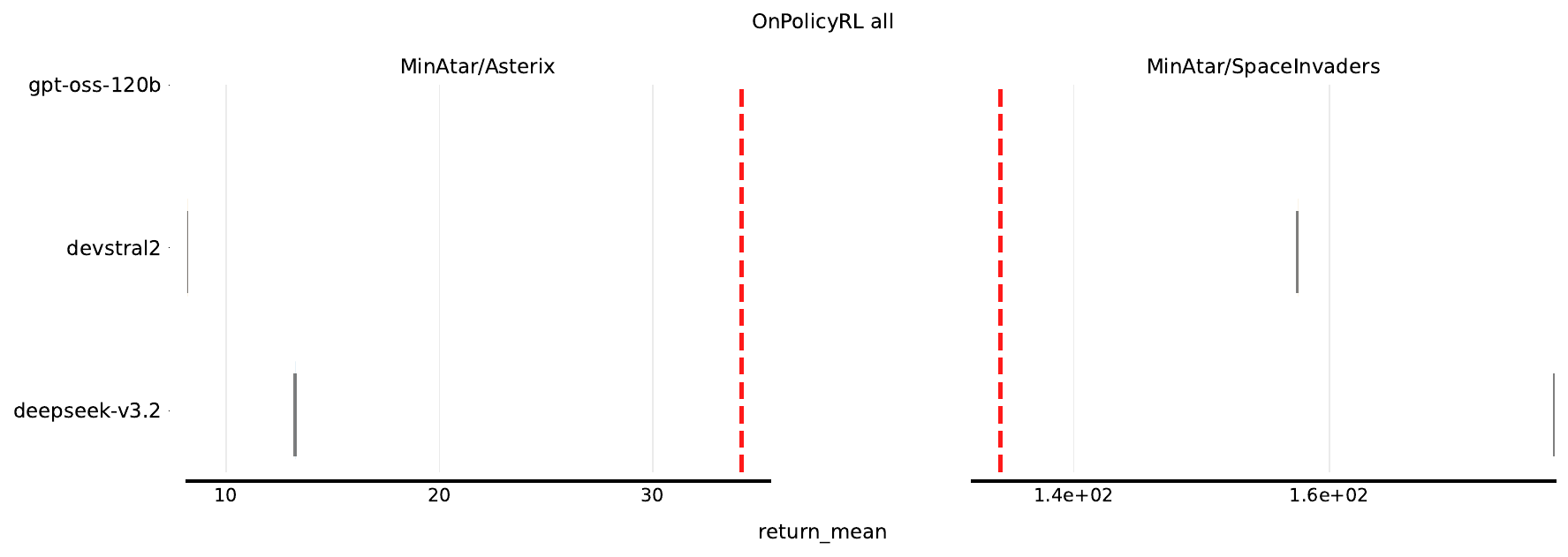}%
\hfill%
\includegraphics[width=0.48\textwidth]{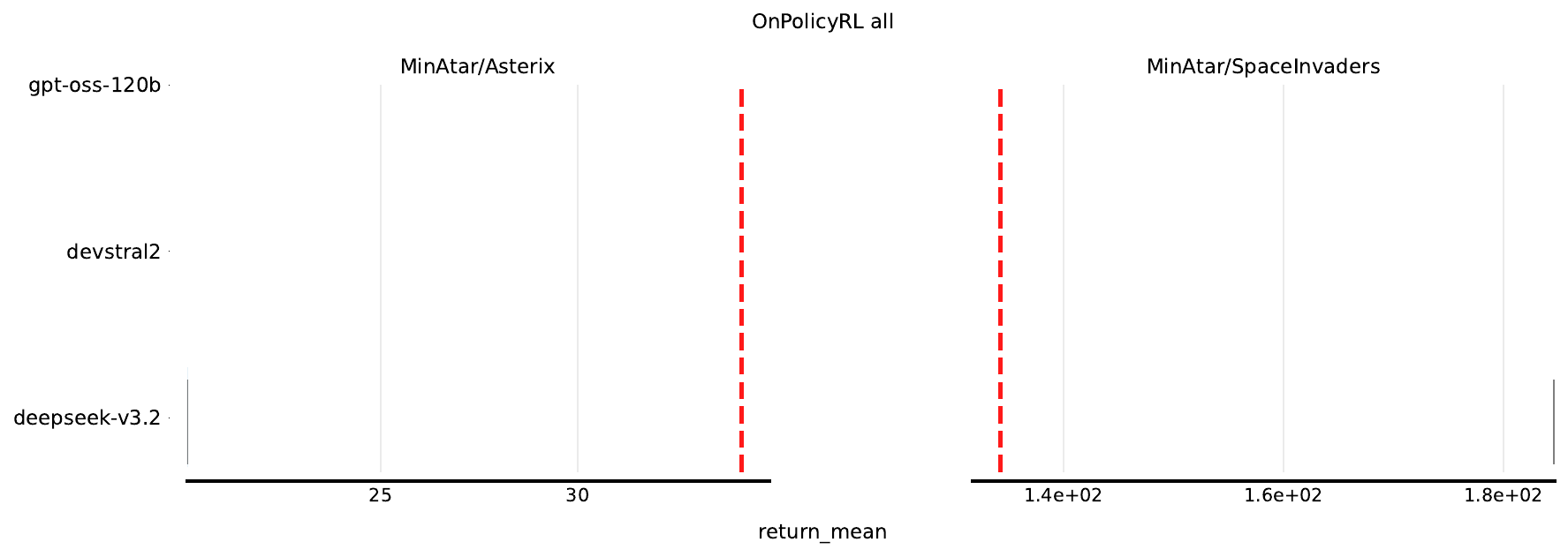}%
\\[0.5em]
\includegraphics[width=0.48\textwidth]{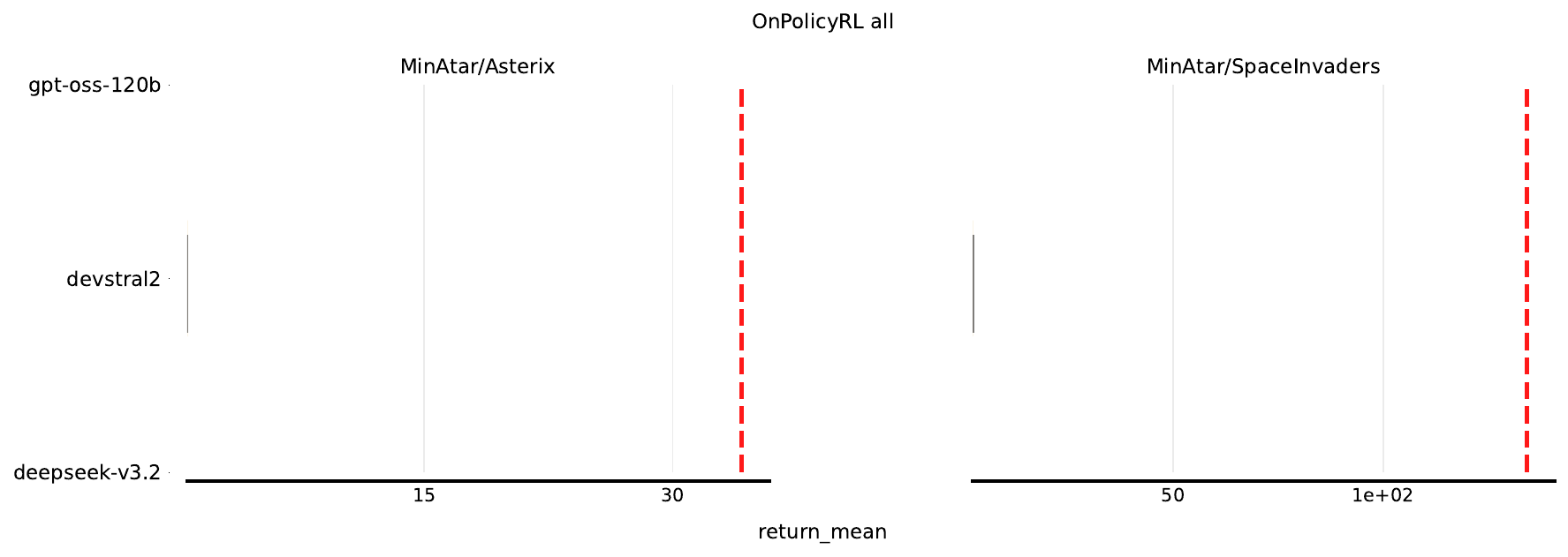}%
\hfill%
\includegraphics[width=0.48\textwidth]{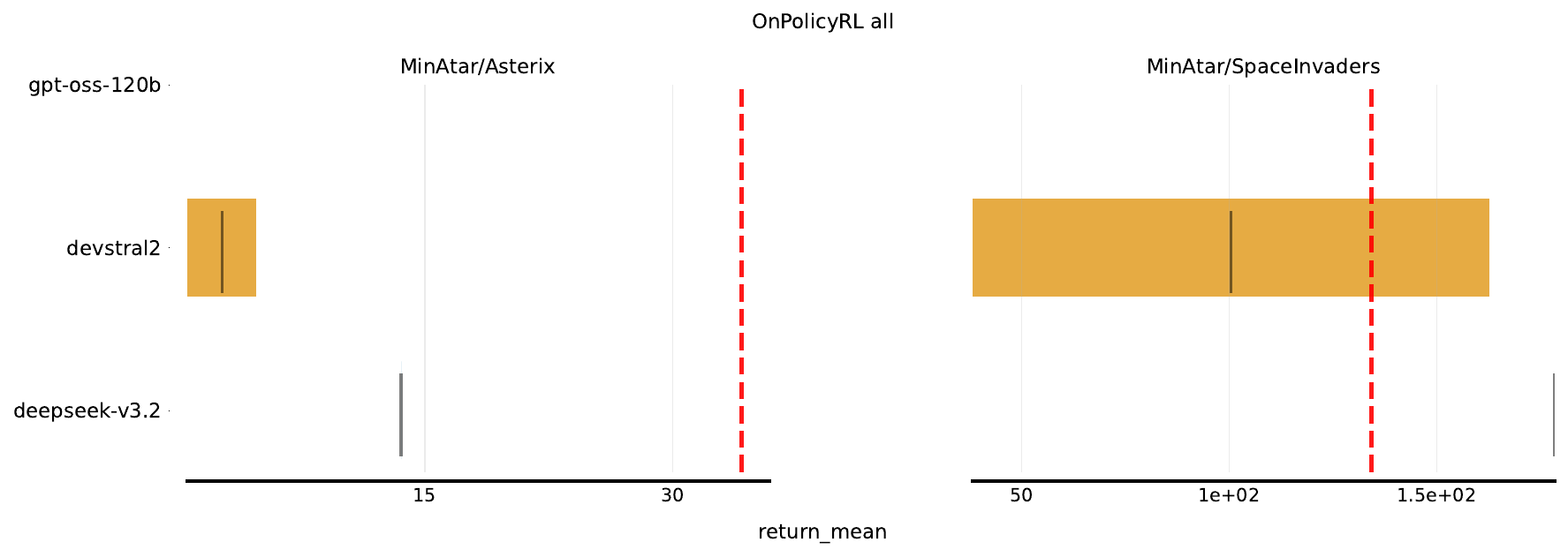}%
\\[0.5em]
\includegraphics[width=0.48\textwidth]{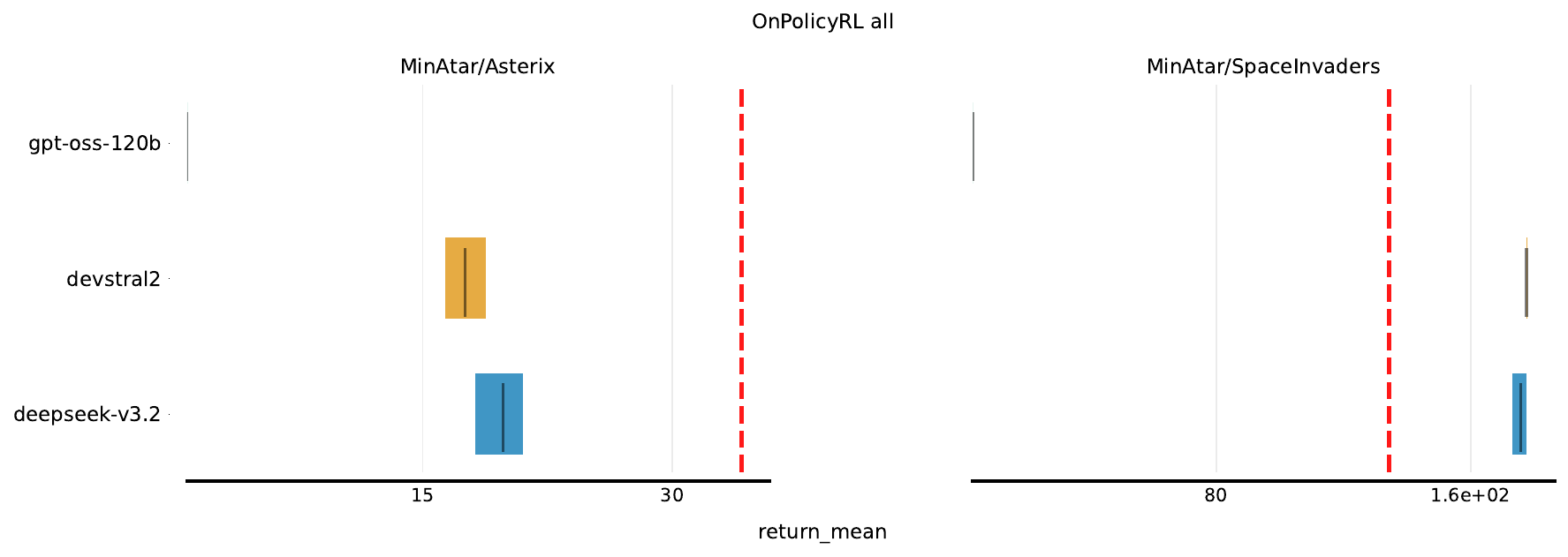}%
\hfill%
\includegraphics[width=0.48\textwidth]{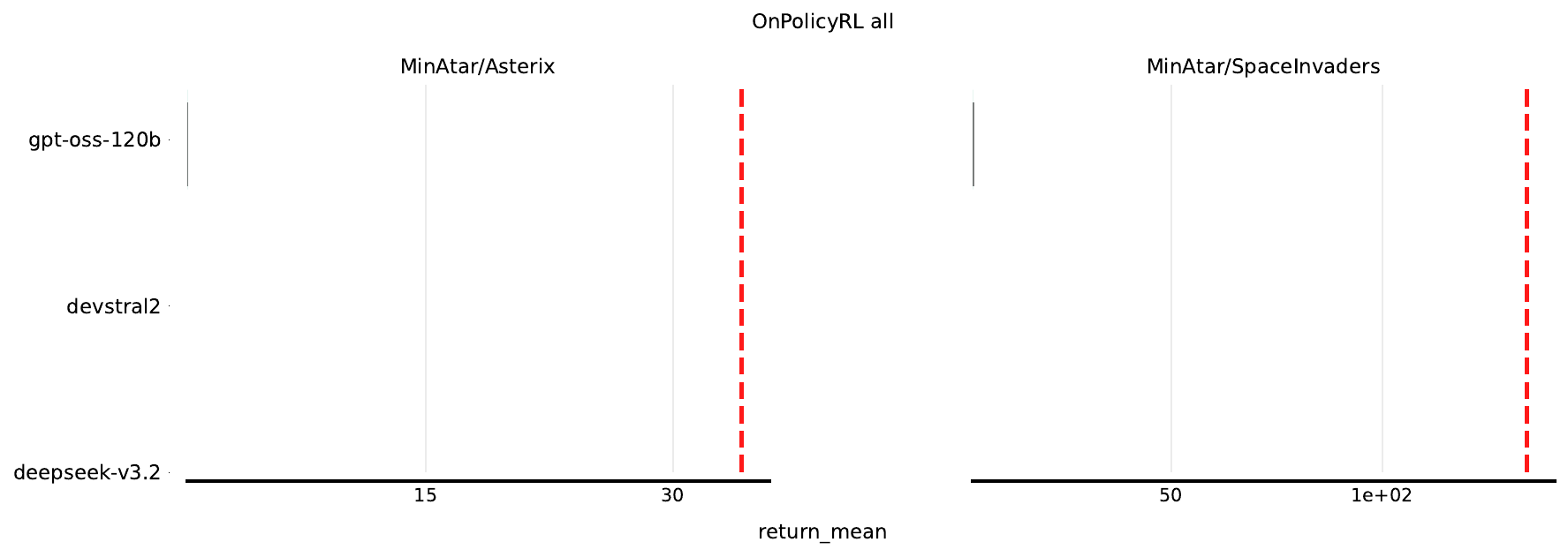}%
\caption{On-Policy RL Combinations results on Meta-Test tasks. (Part 1/4)}
\label{fig:on_policy_mt_1}
\end{figure}
\clearpage

\begin{figure}[htbp]
\centering
\setlength{\lineskip}{0pt}
\includegraphics[width=0.48\textwidth]{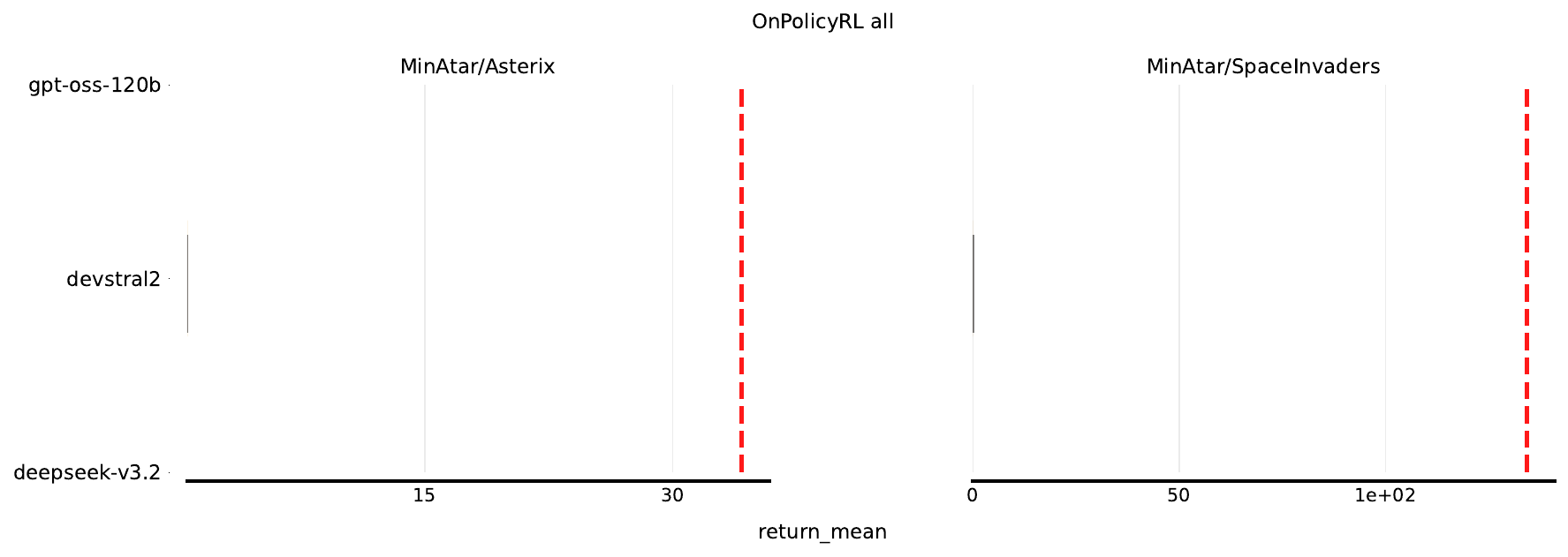}%
\hfill%
\includegraphics[width=0.48\textwidth]{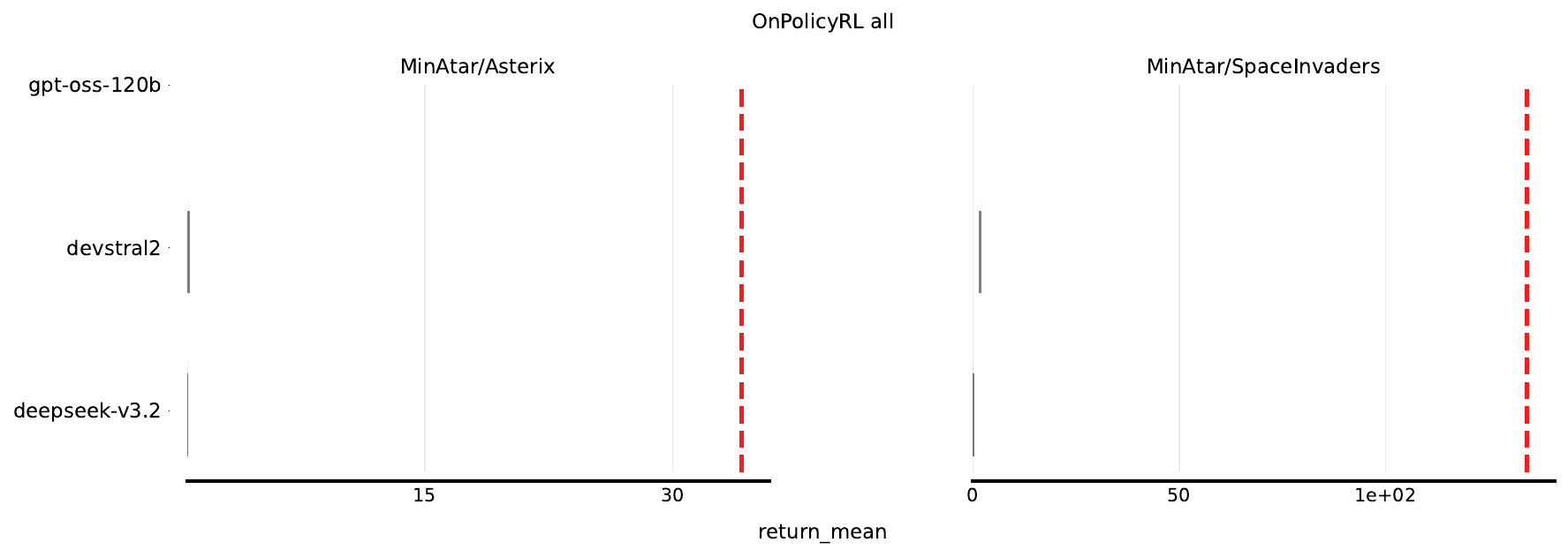}%
\\[0.5em]
\includegraphics[width=0.48\textwidth]{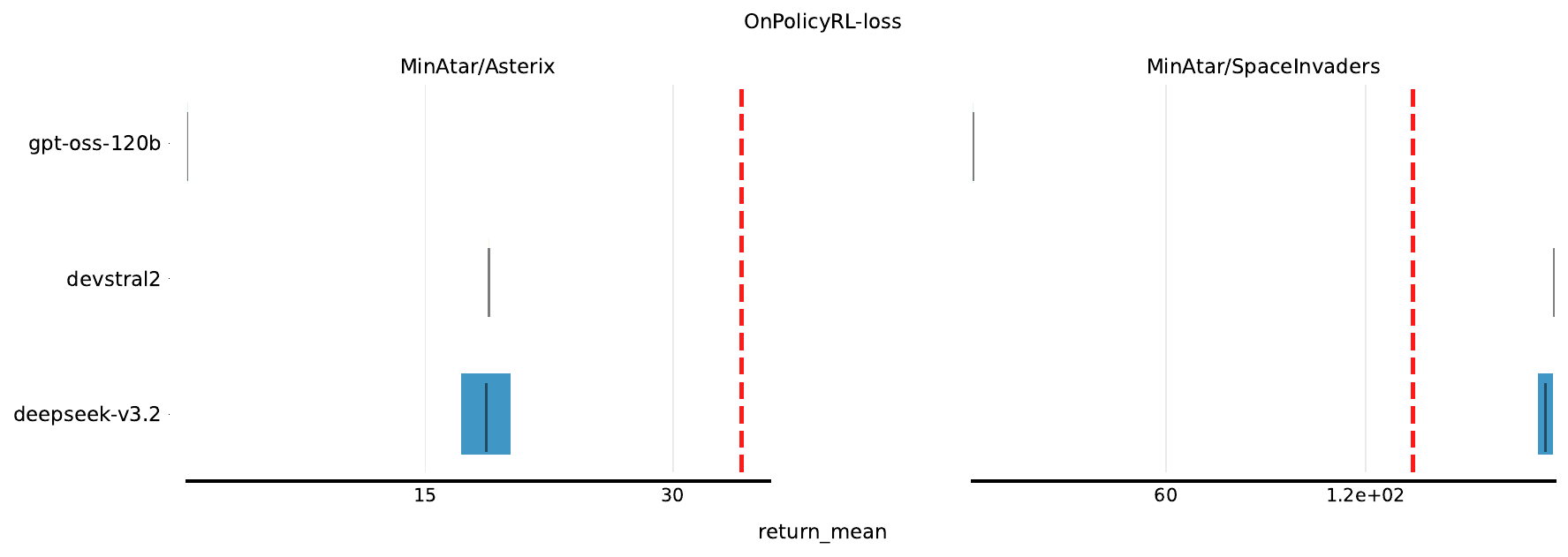}%
\hfill%
\includegraphics[width=0.48\textwidth]{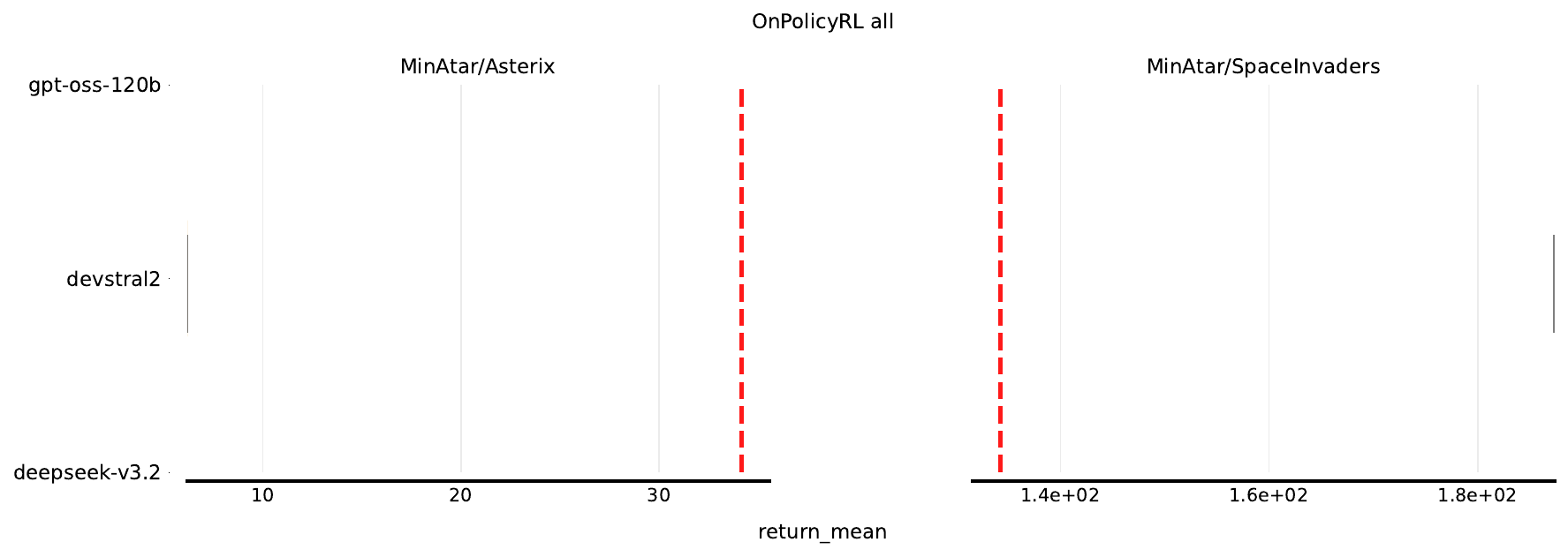}%
\\[0.5em]
\includegraphics[width=0.48\textwidth]{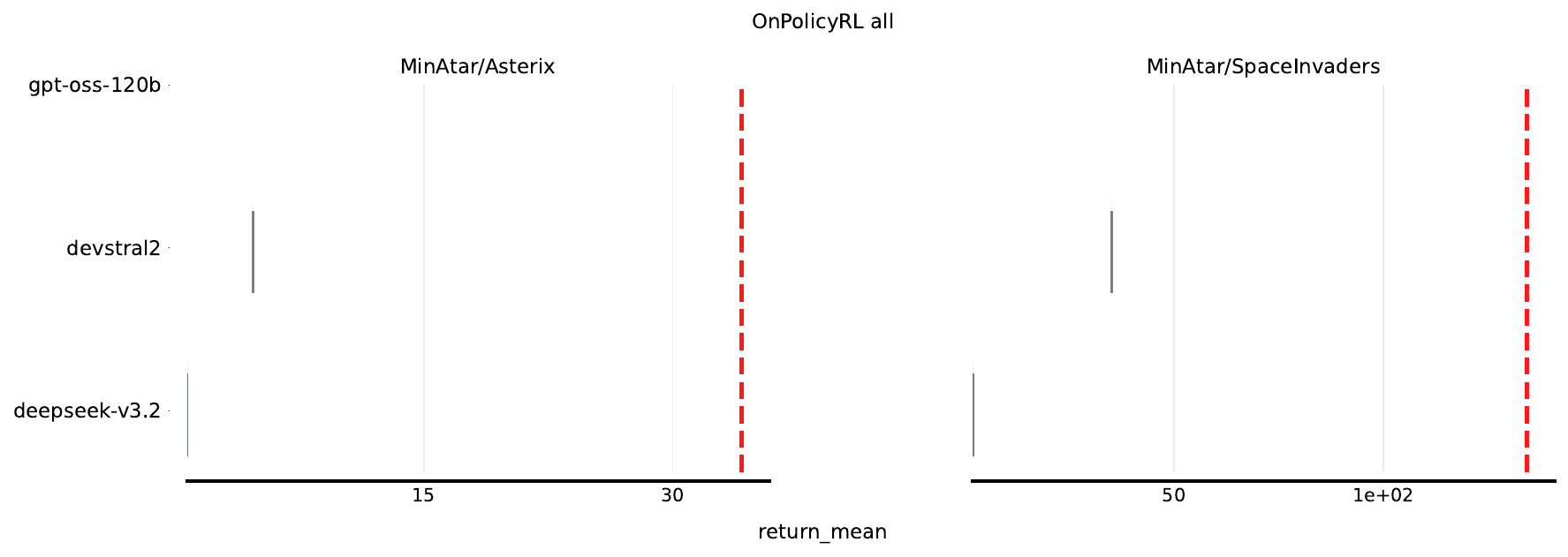}%
\hfill%
\includegraphics[width=0.48\textwidth]{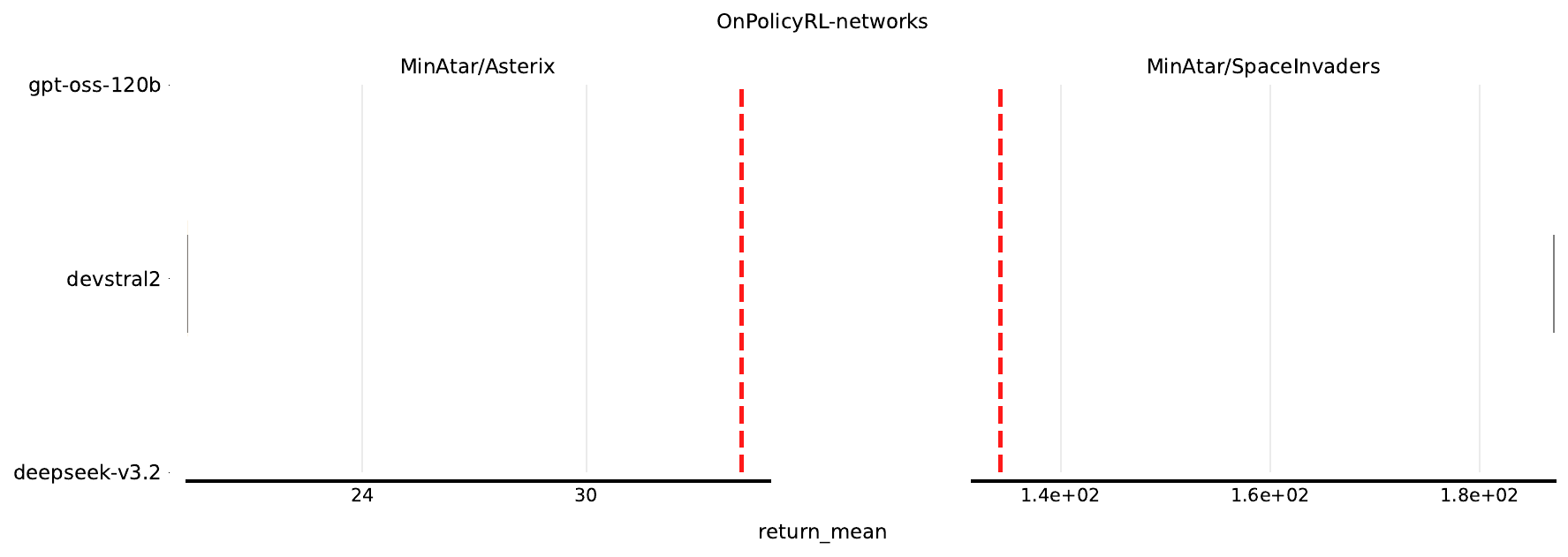}%
\\[0.5em]
\includegraphics[width=0.48\textwidth]{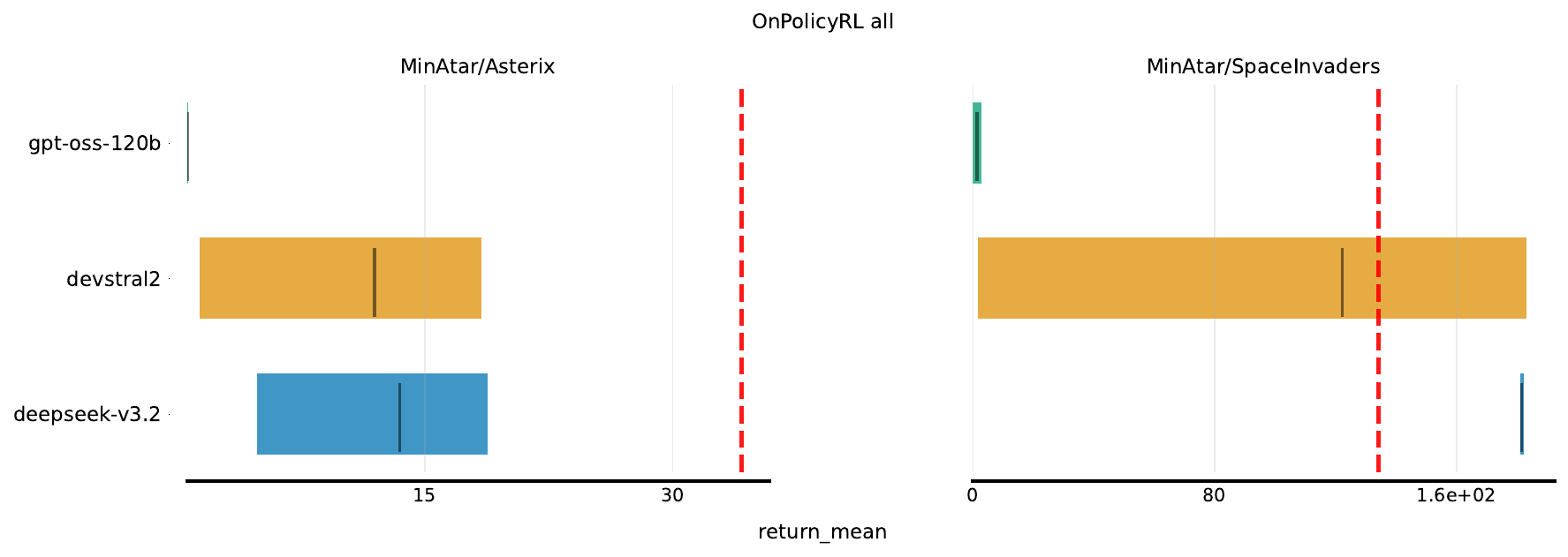}%
\hfill%
\includegraphics[width=0.48\textwidth]{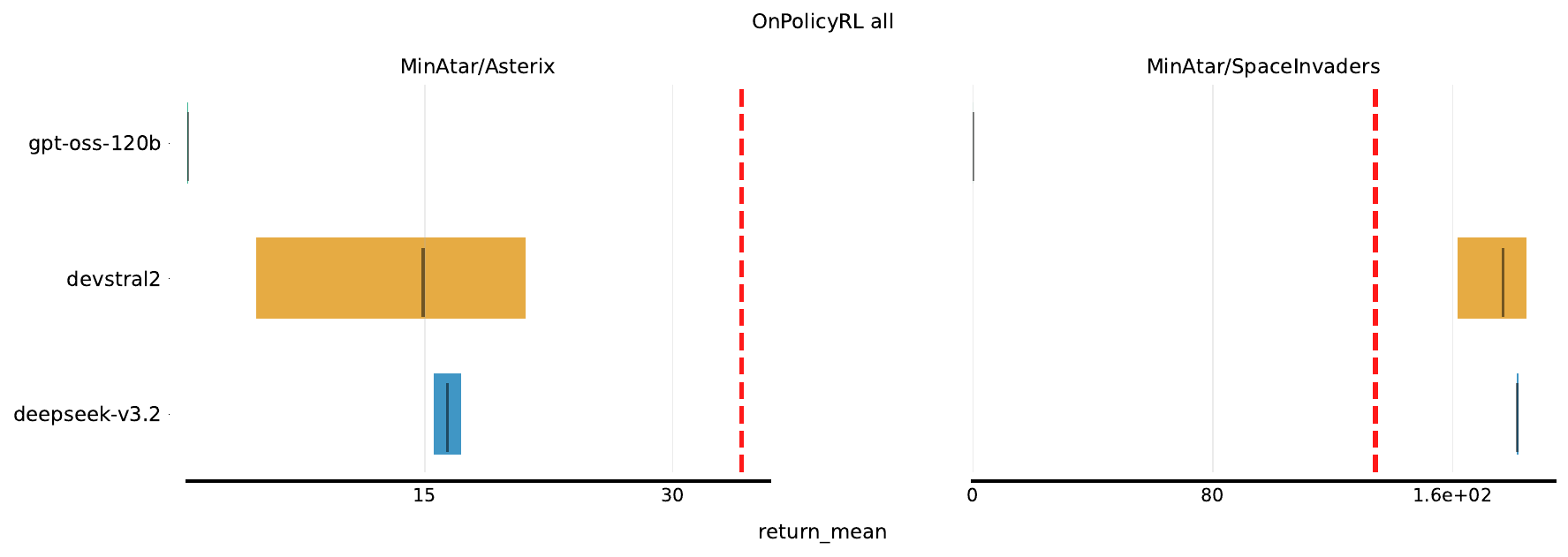}%
\\[0.5em]
\includegraphics[width=0.48\textwidth]{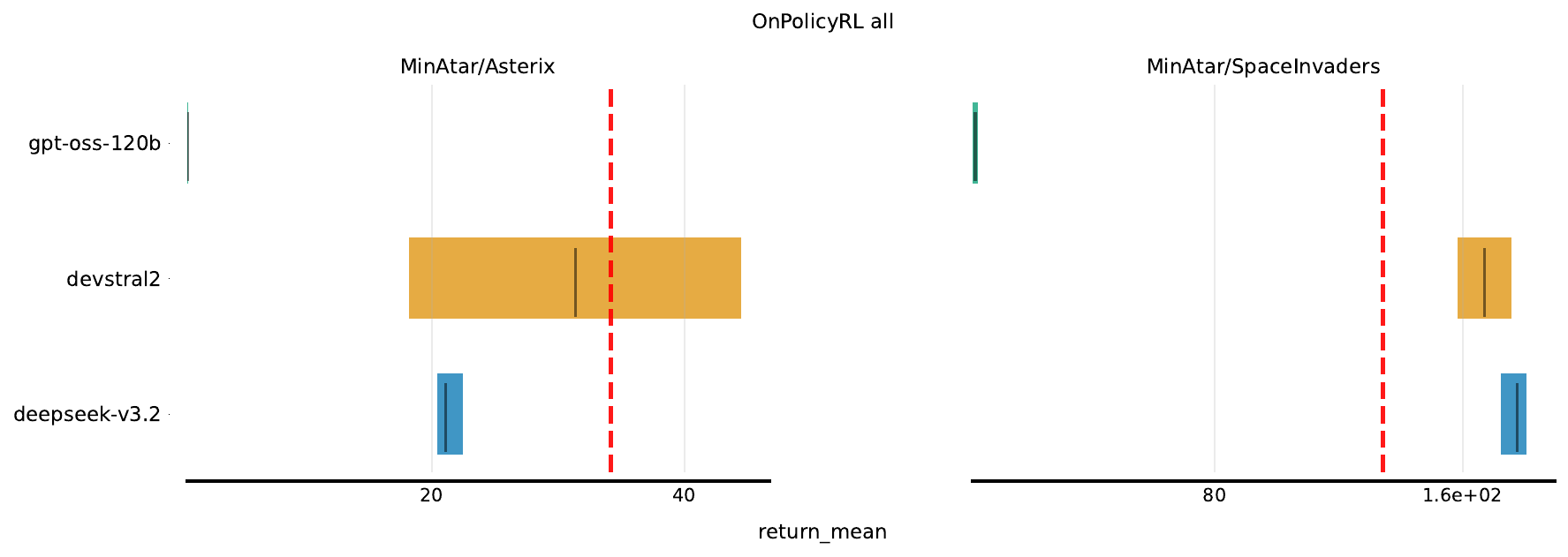}%
\hfill%
\includegraphics[width=0.48\textwidth]{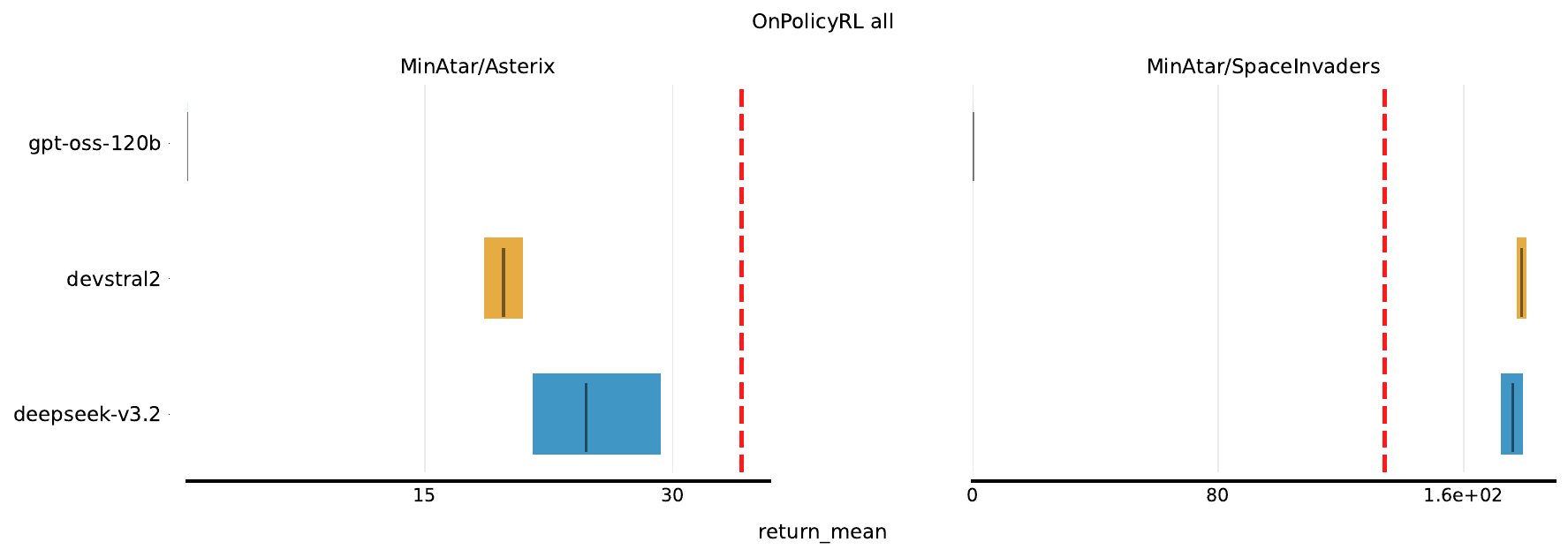}%
\\[0.5em]
\includegraphics[width=0.48\textwidth]{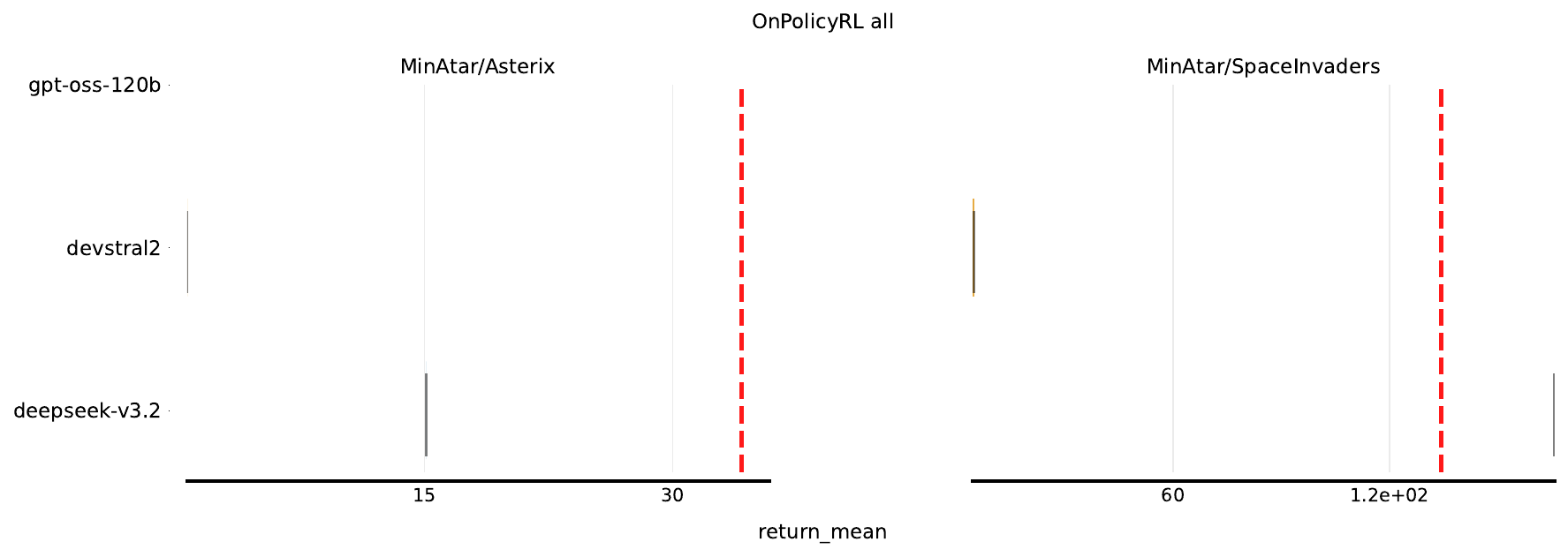}%
\hfill%
\includegraphics[width=0.48\textwidth]{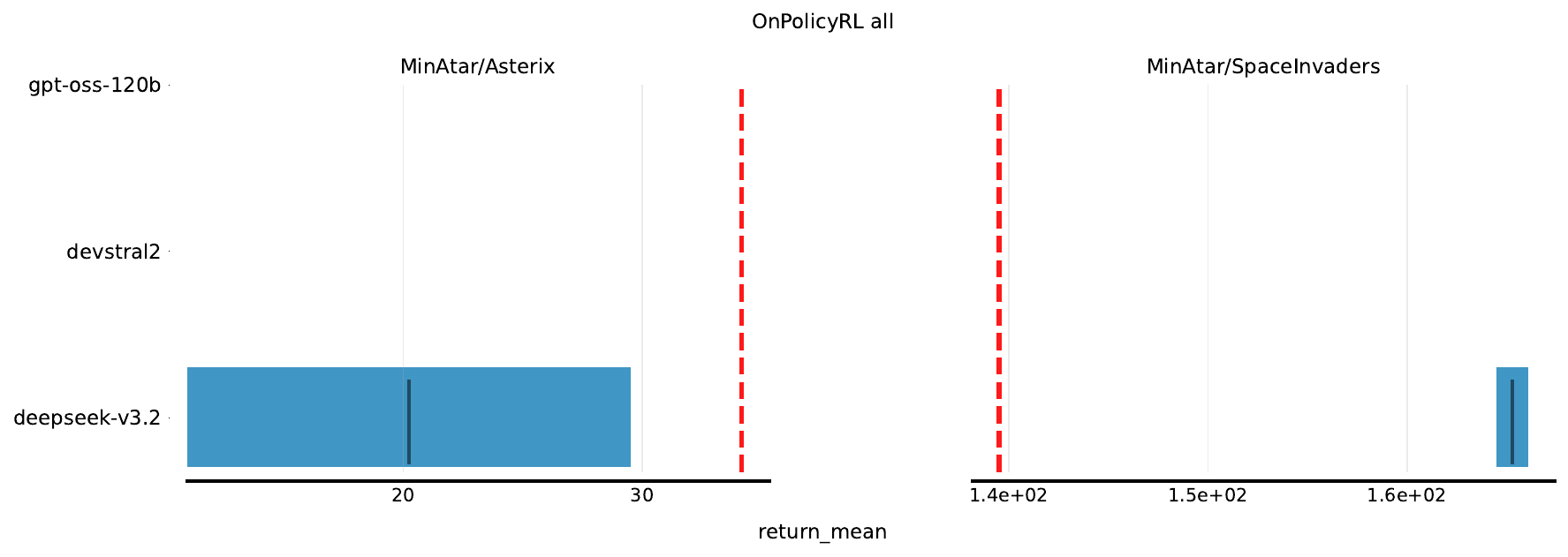}%
\caption{On-Policy RL Combinations results on Meta-Test tasks. (Part 2/4)}
\label{fig:on_policy_mt_2}
\end{figure}
\clearpage

\begin{figure}[htbp]
\centering
\setlength{\lineskip}{0pt}
\includegraphics[width=0.48\textwidth]{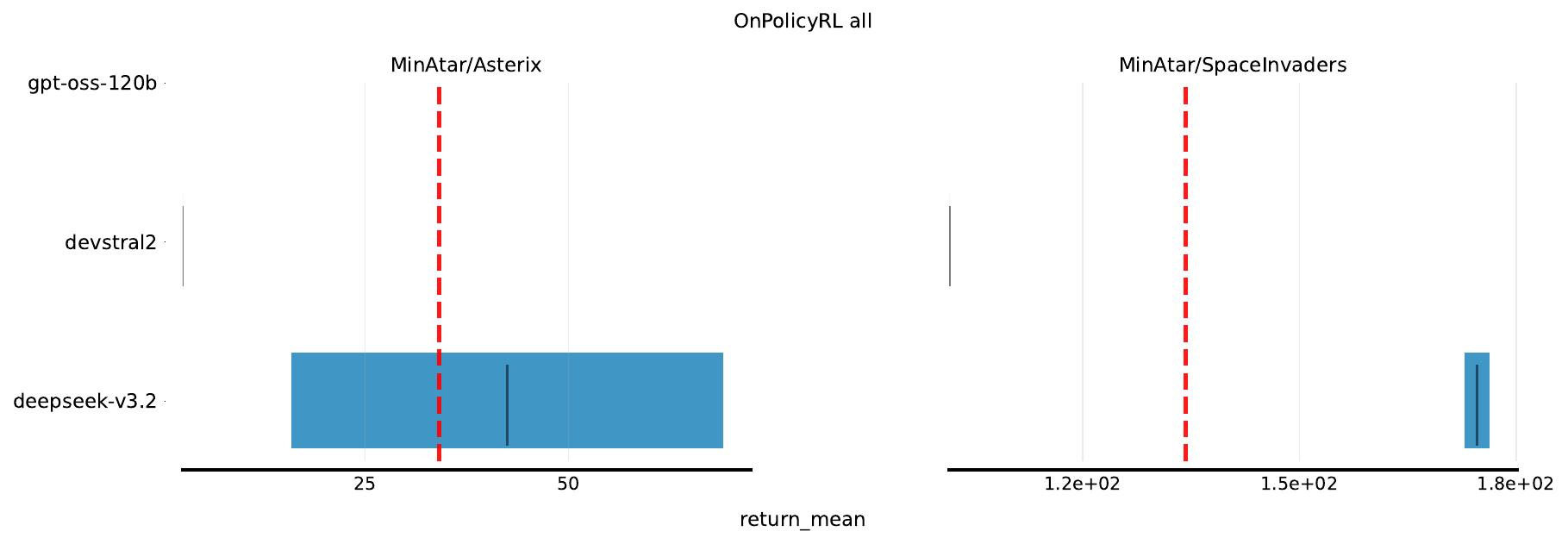}%
\hfill%
\includegraphics[width=0.48\textwidth]{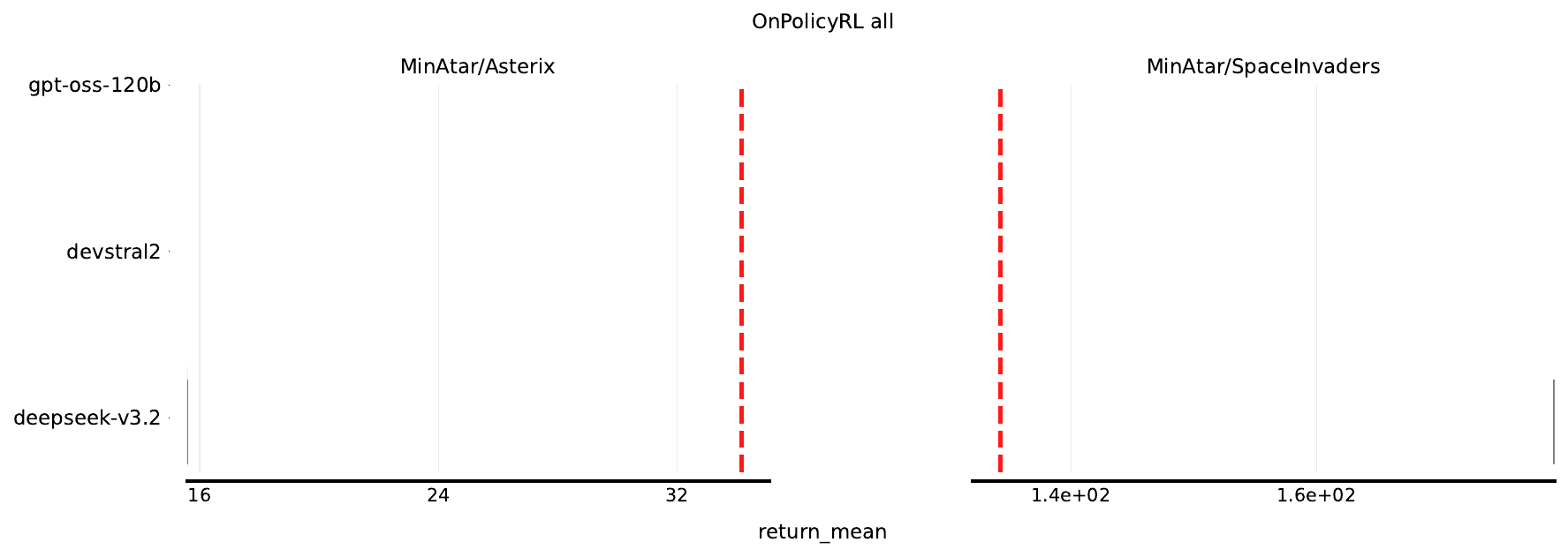}%
\\[0.5em]
\includegraphics[width=0.48\textwidth]{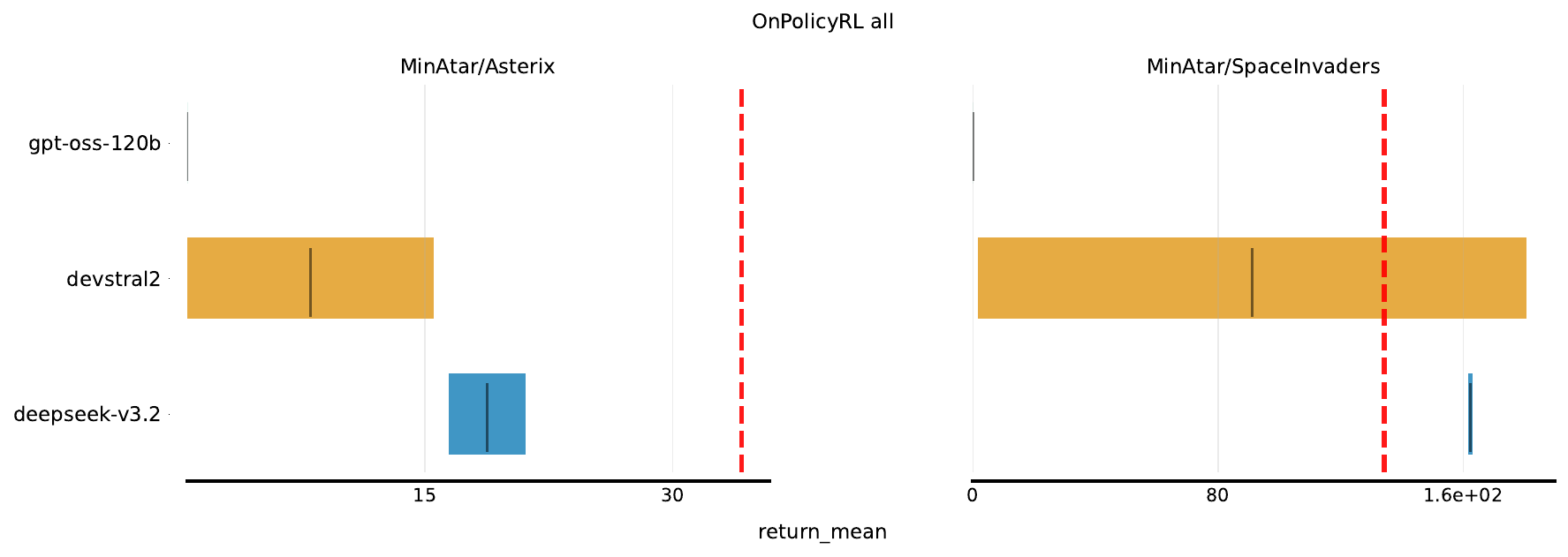}%
\hfill%
\includegraphics[width=0.48\textwidth]{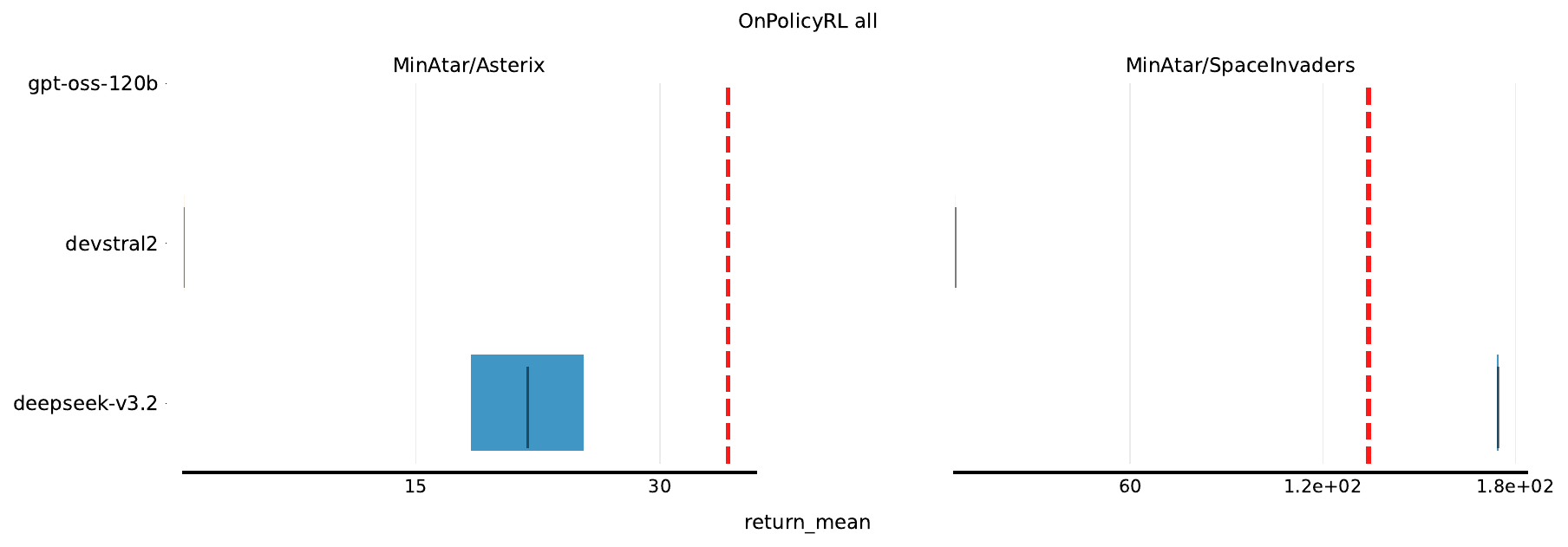}%
\\[0.5em]
\includegraphics[width=0.48\textwidth]{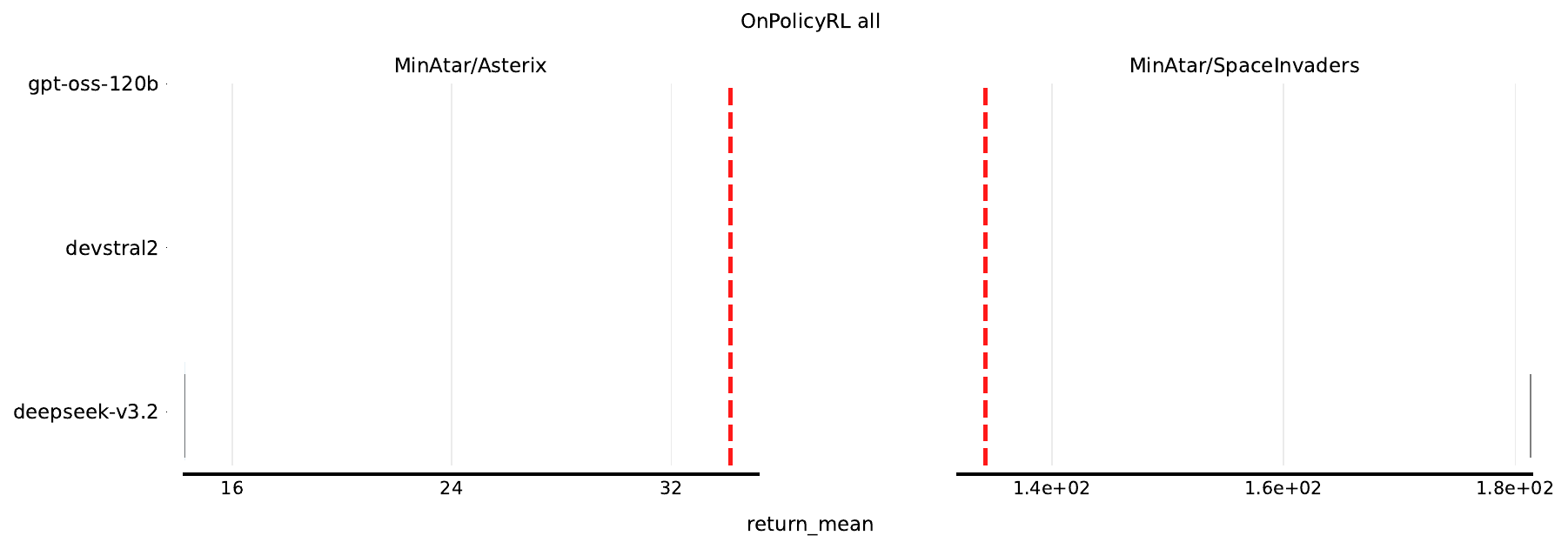}%
\hfill%
\includegraphics[width=0.48\textwidth]{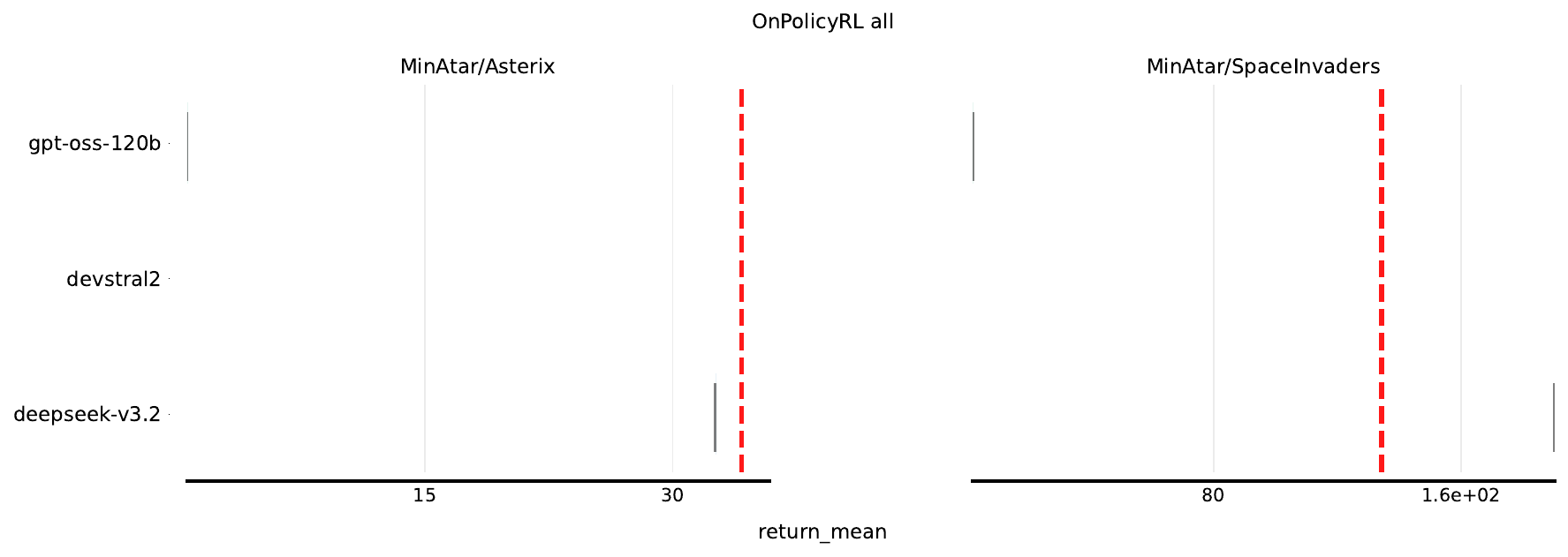}%
\\[0.5em]
\includegraphics[width=0.48\textwidth]{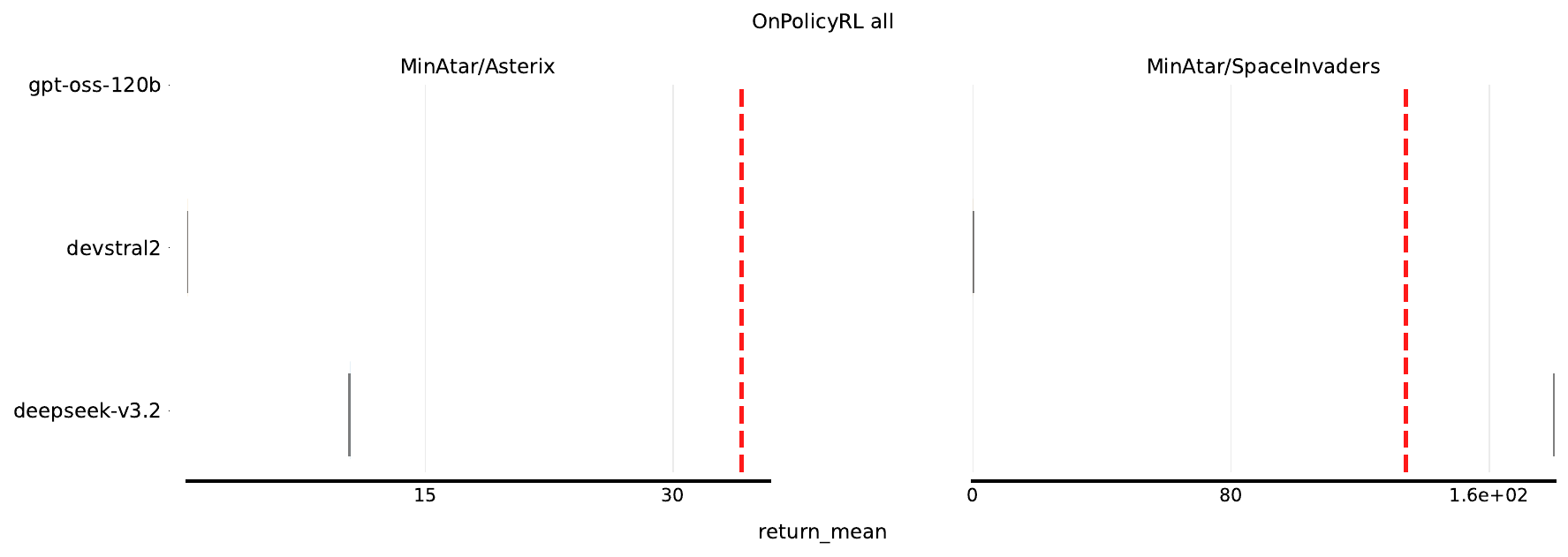}%
\hfill%
\includegraphics[width=0.48\textwidth]{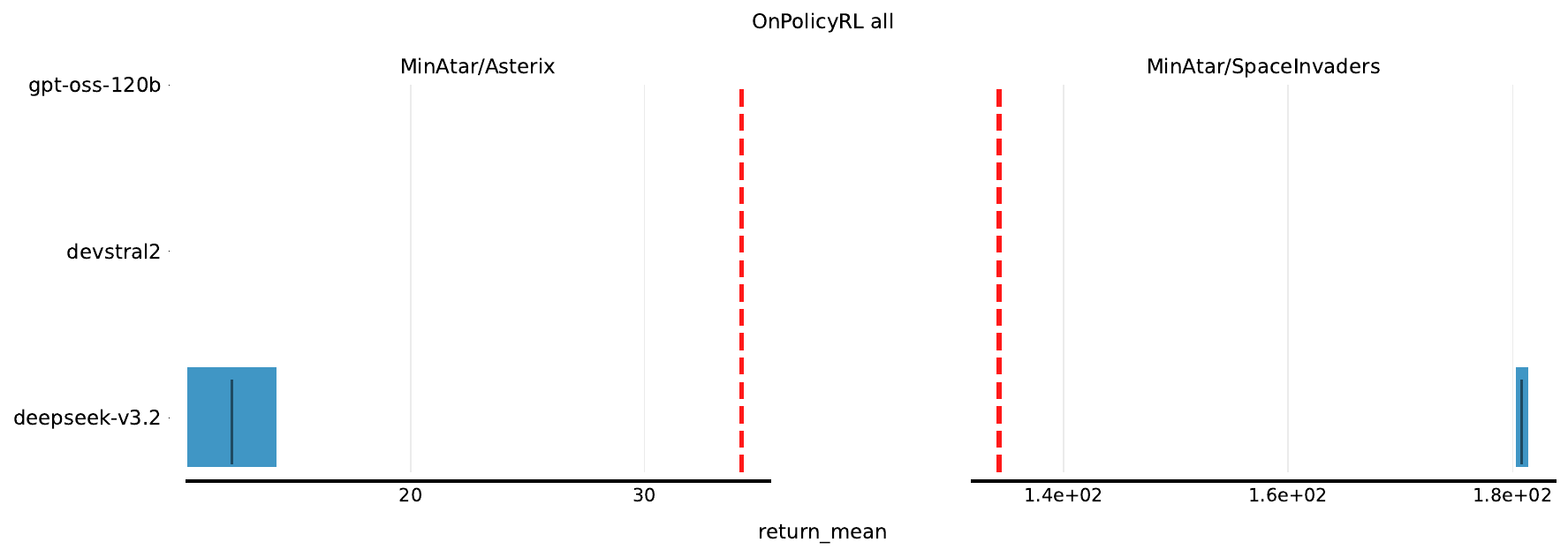}%
\\[0.5em]
\includegraphics[width=0.48\textwidth]{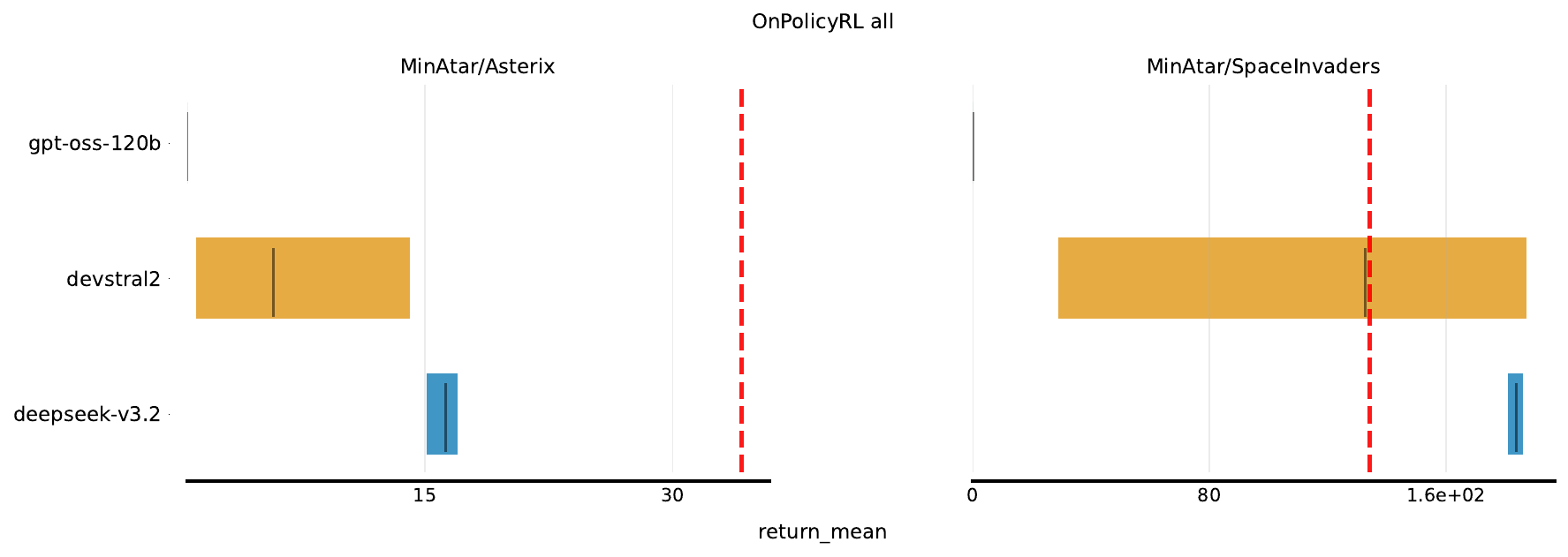}%
\hfill%
\includegraphics[width=0.48\textwidth]{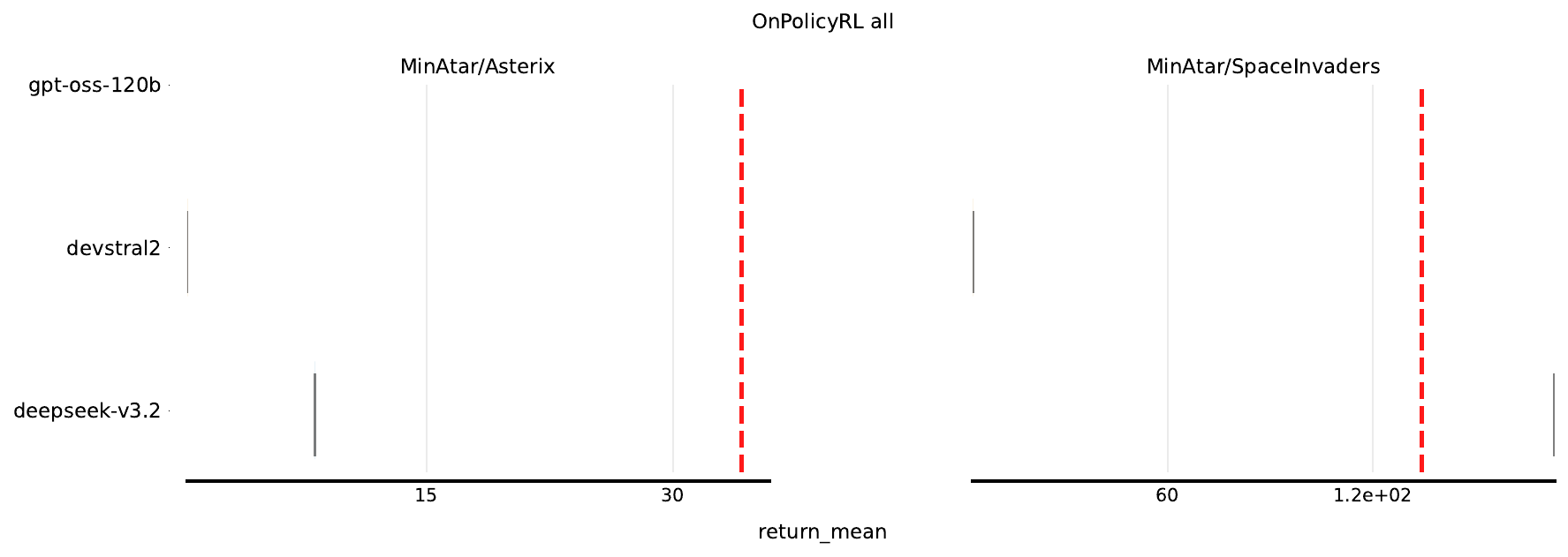}%
\\[0.5em]
\includegraphics[width=0.48\textwidth]{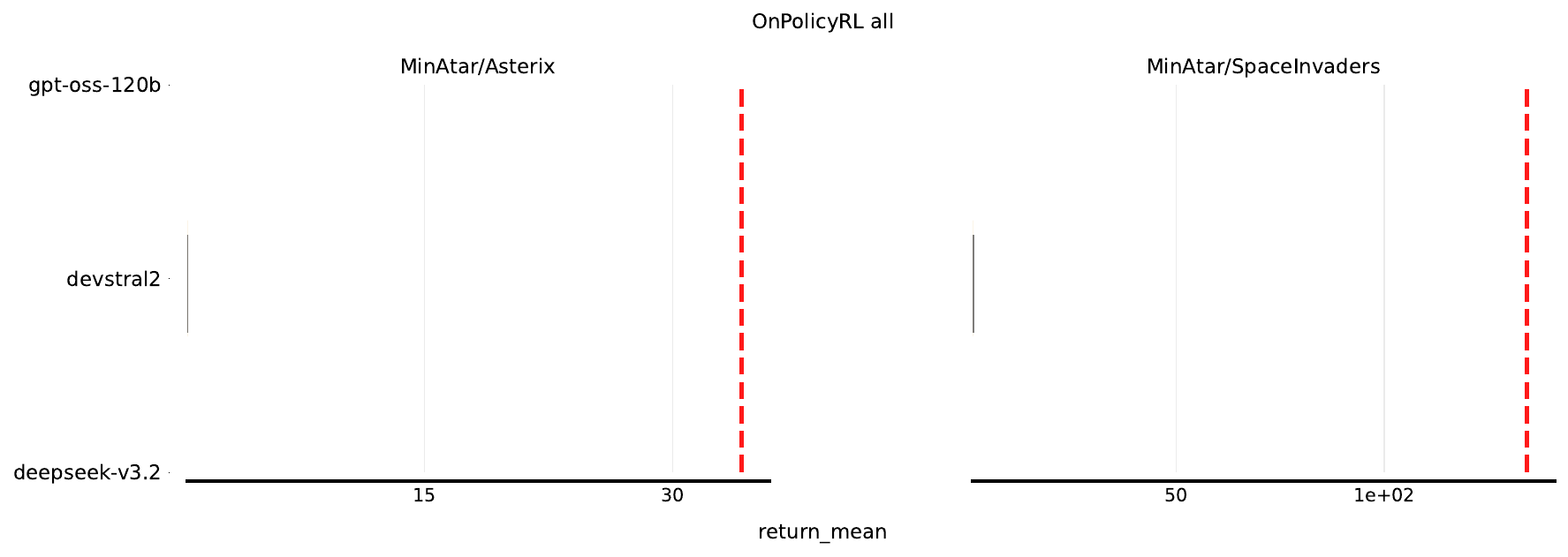}%
\hfill%
\includegraphics[width=0.48\textwidth]{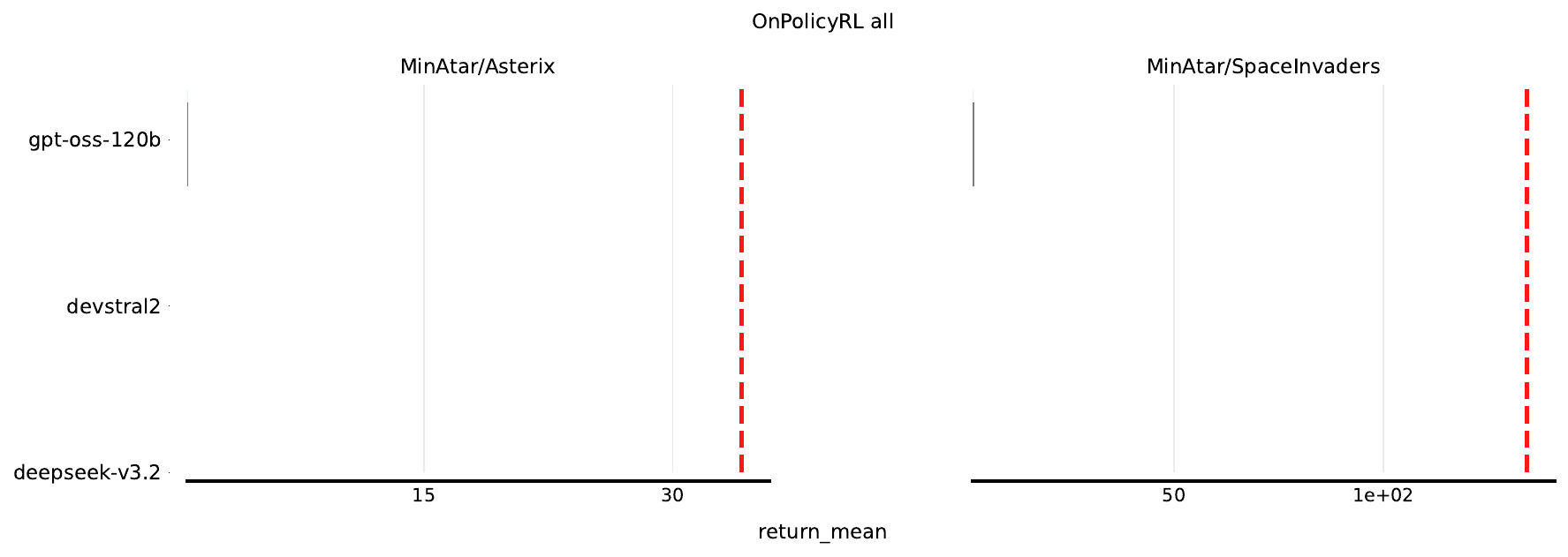}%
\caption{On-Policy RL Combinations results on Meta-Test tasks. (Part 3/4)}
\label{fig:on_policy_mt_3}
\end{figure}
\clearpage

\begin{figure}[htbp]
\centering
\setlength{\lineskip}{0pt}
\includegraphics[width=0.48\textwidth]{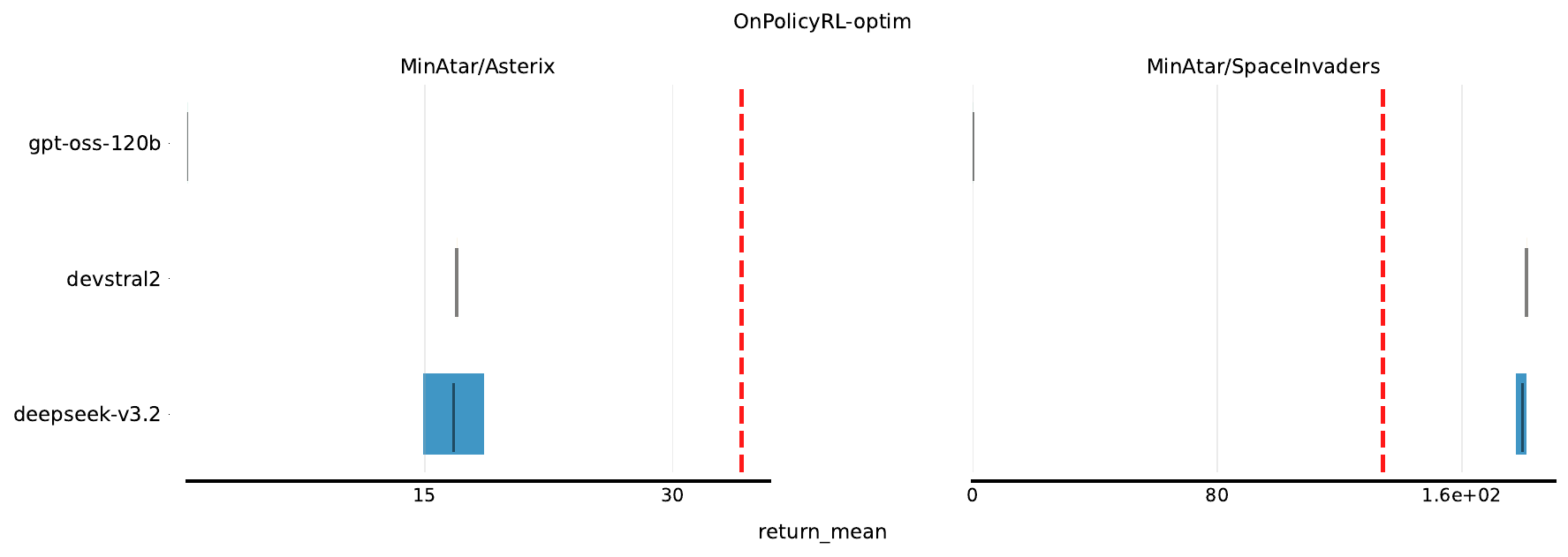}%
\hfill%
\includegraphics[width=0.48\textwidth]{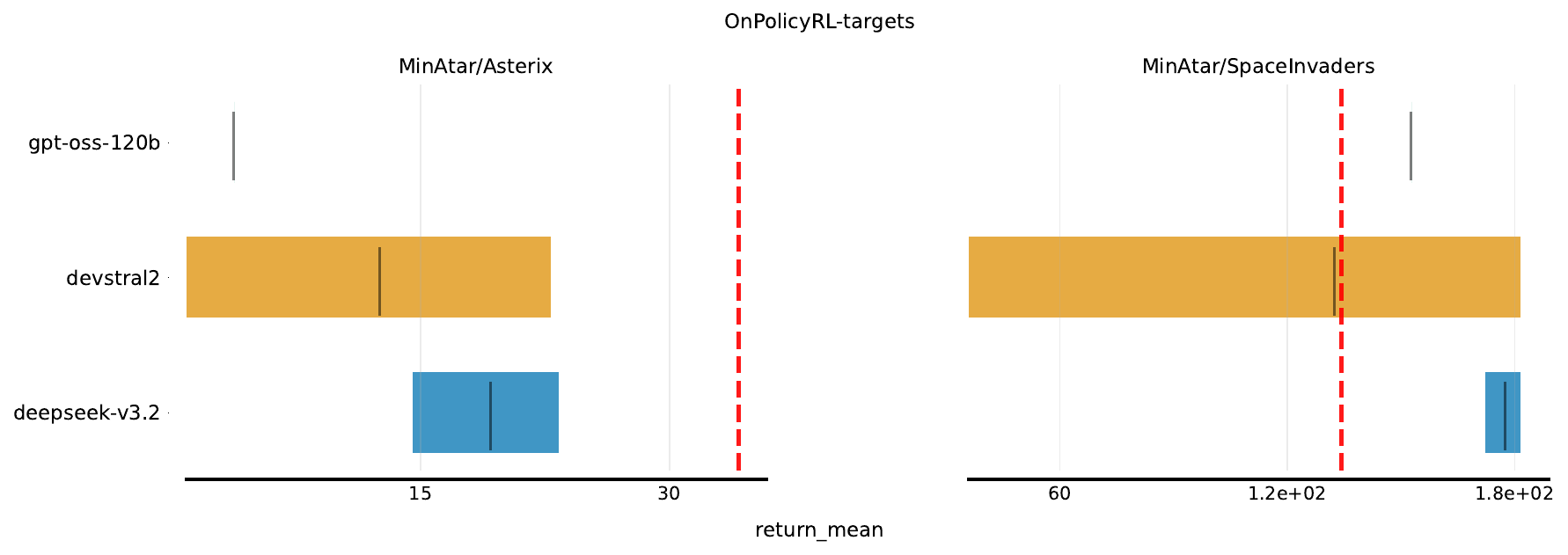}%
\\[0.5em]
\includegraphics[width=0.48\textwidth]{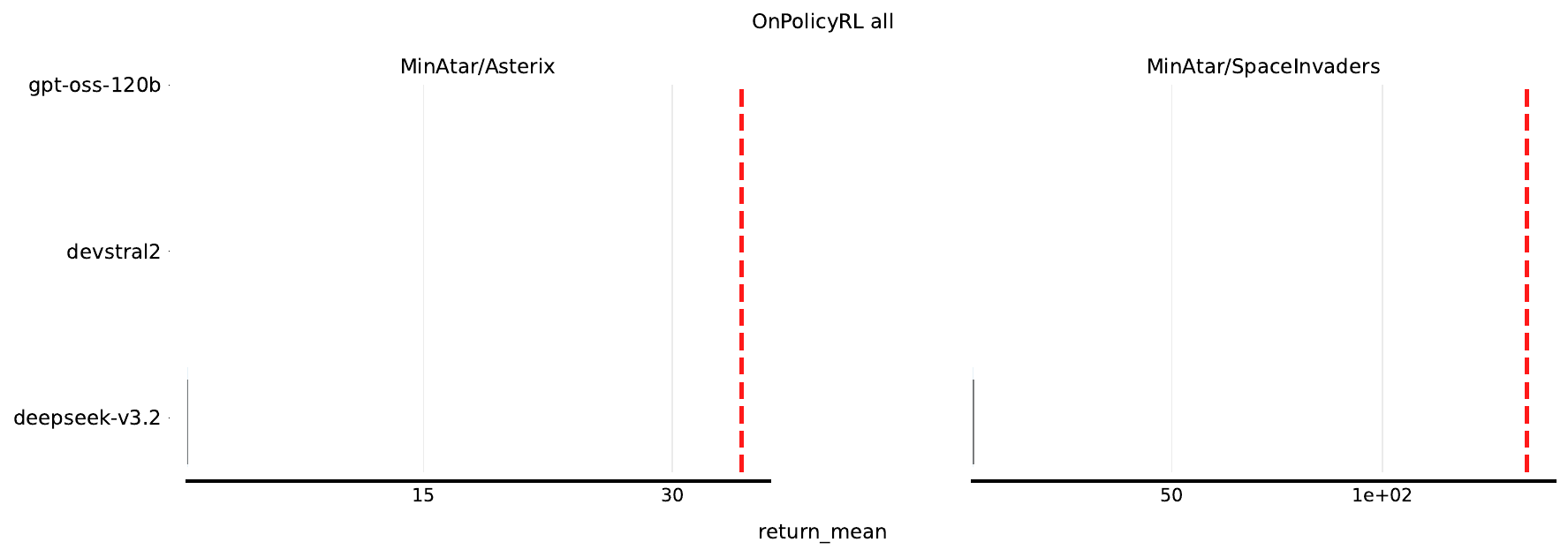}%
\hfill%
\includegraphics[width=0.48\textwidth]{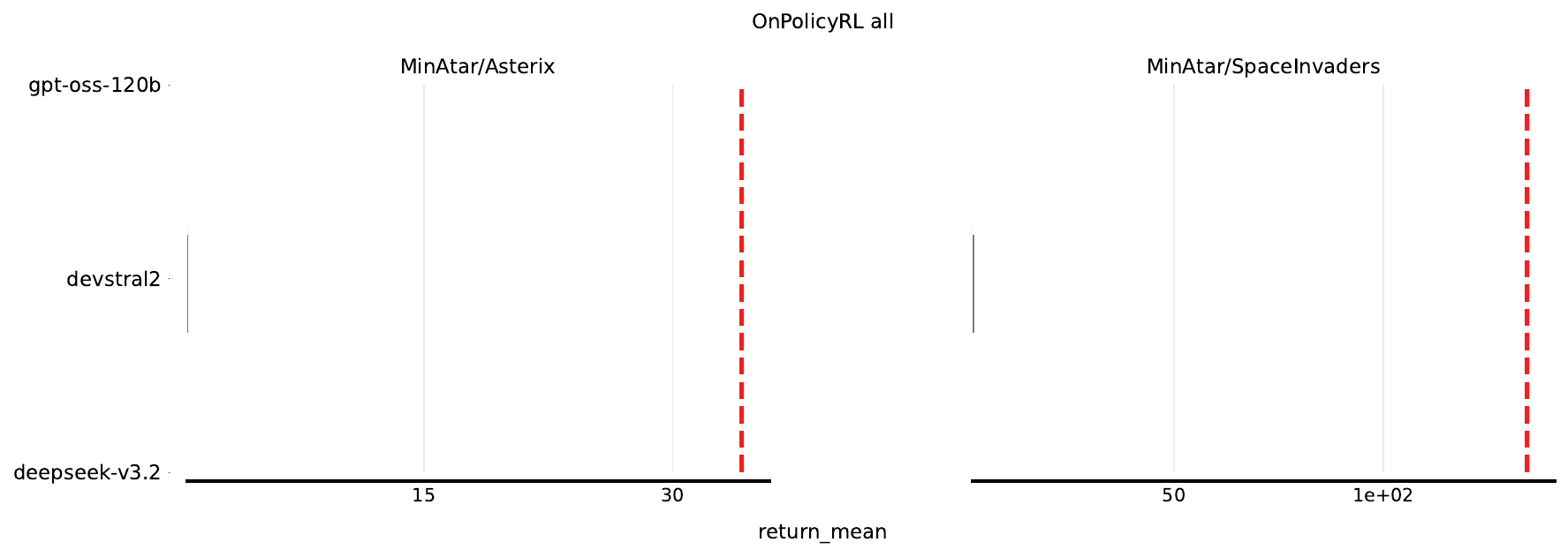}%
\\[0.5em]
\includegraphics[width=0.48\textwidth]{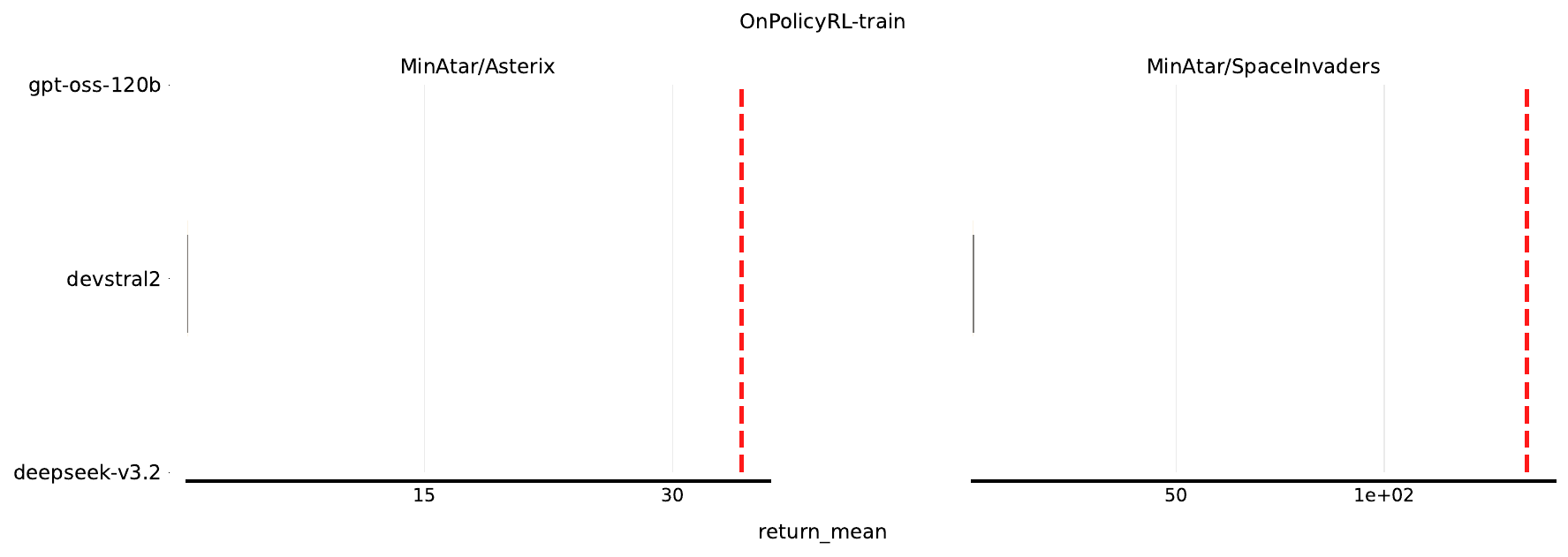}%
\hfill%
\caption{On-Policy RL Combinations results on Meta-Test tasks. (Part 4/4)}
\label{fig:on_policy_mt_4}
\end{figure}
\clearpage

\subsection{ADA Optimisation -- Meta-Train}
\label{sec:ADA_optimisation_id}

\begin{figure}[htbp]
\centering
\setlength{\lineskip}{0pt}
\includegraphics[width=0.48\textwidth]{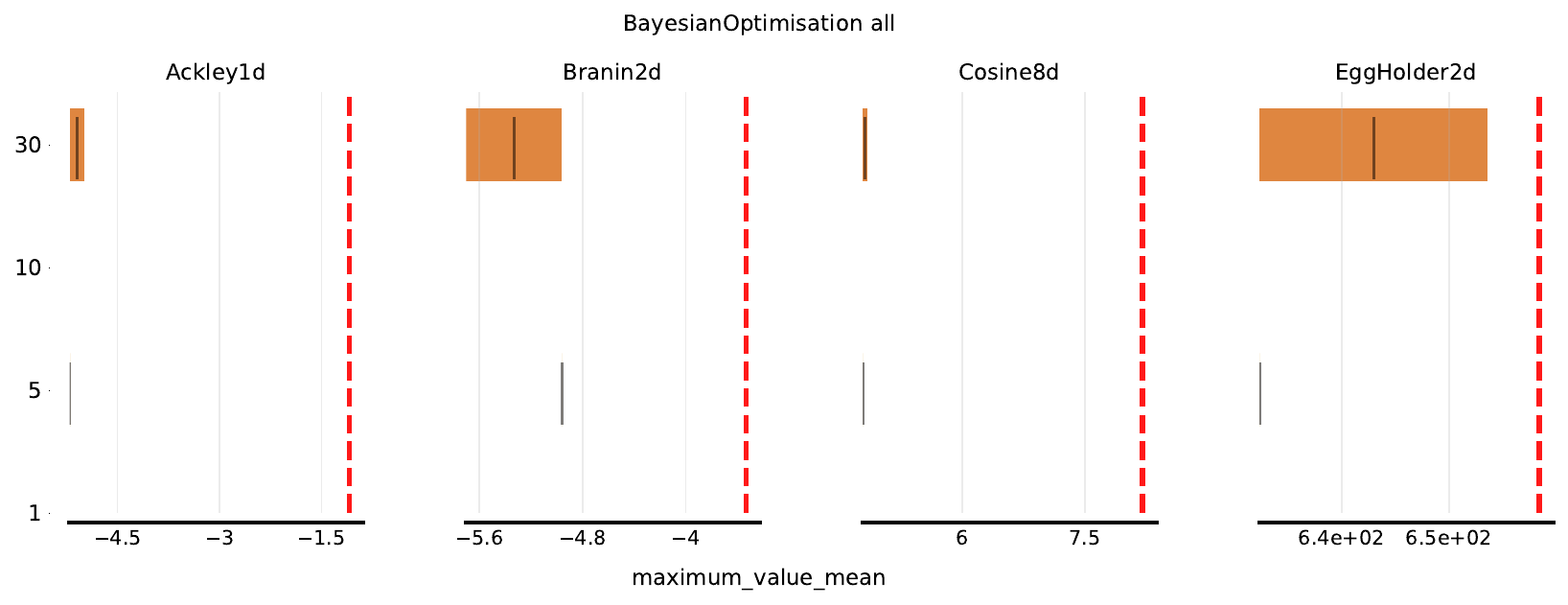}%
\hfill%
\includegraphics[width=0.48\textwidth]{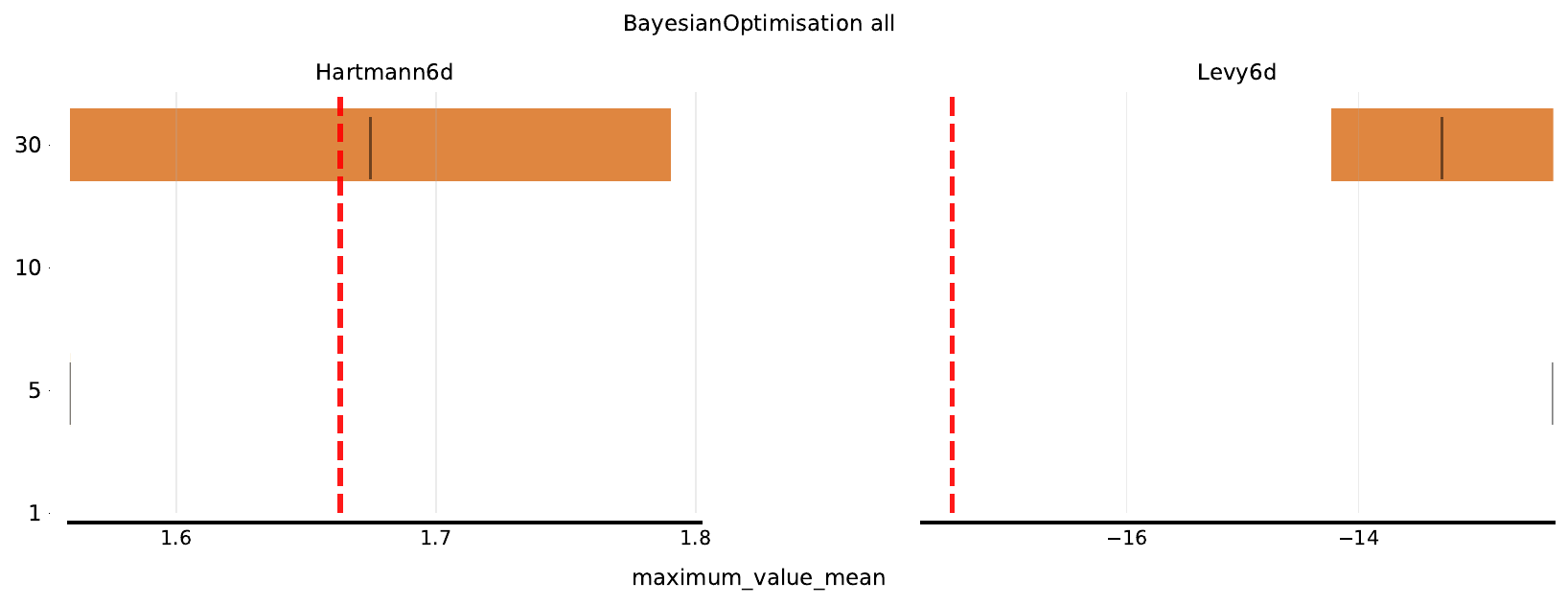}%
\\[0.5em]
\includegraphics[width=0.48\textwidth]{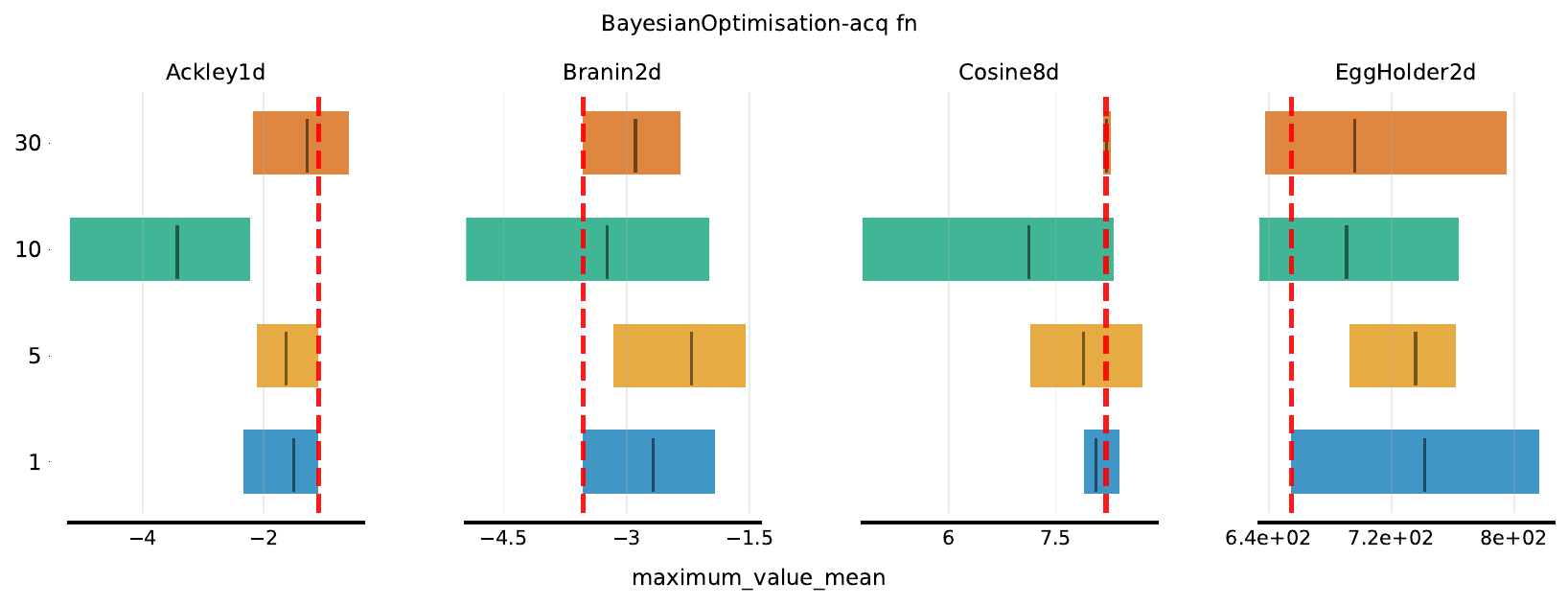}%
\hfill%
\includegraphics[width=0.48\textwidth]{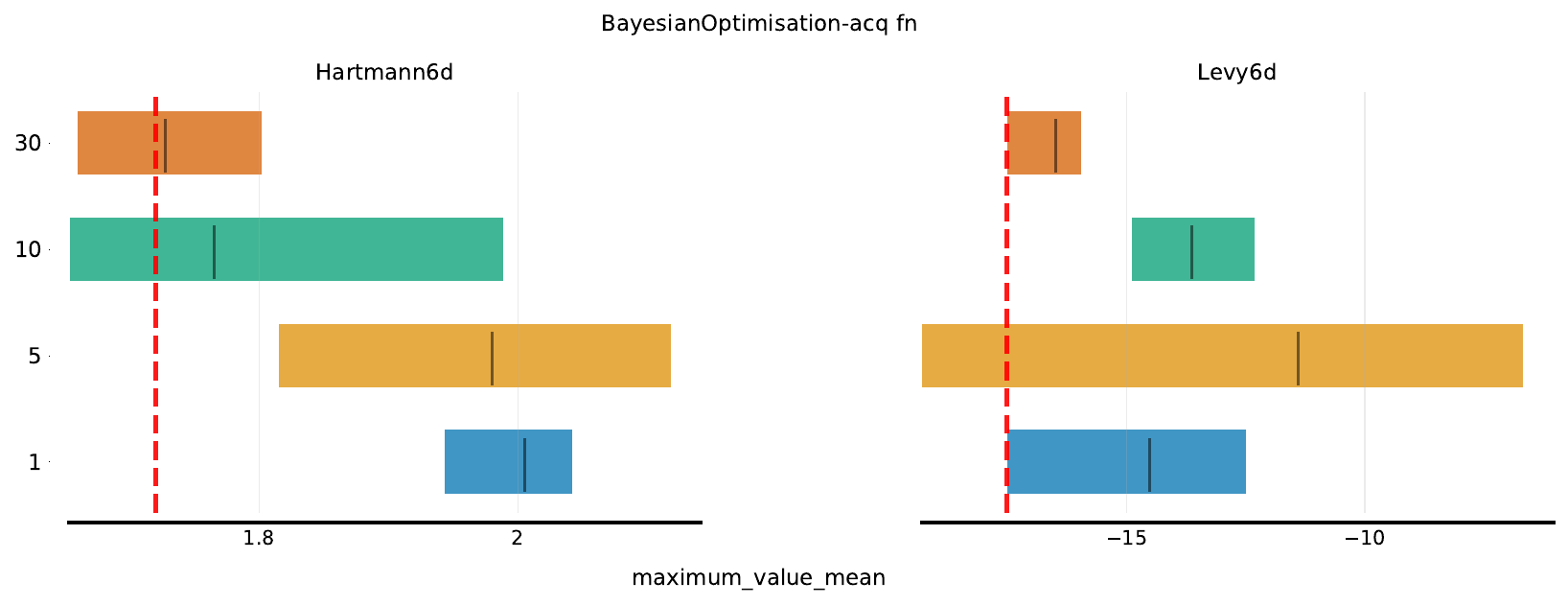}%
\\[0.5em]
\includegraphics[width=0.48\textwidth]{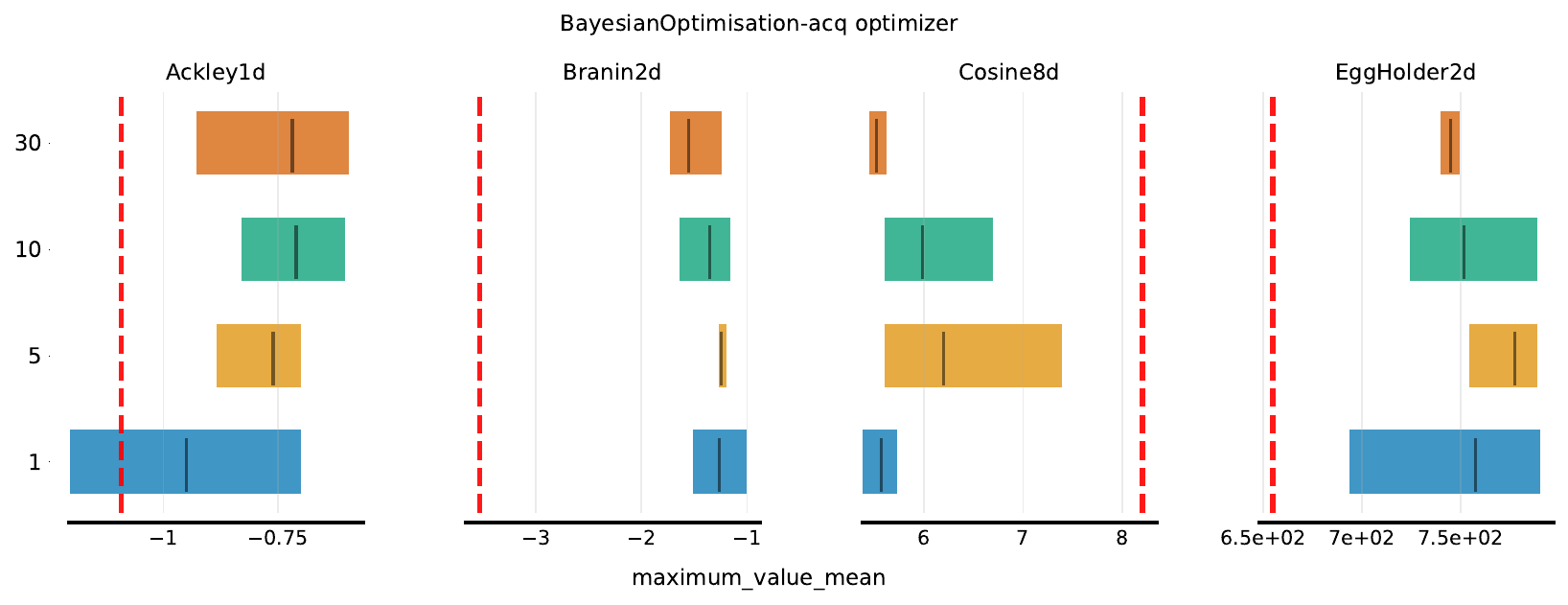}%
\hfill%
\includegraphics[width=0.48\textwidth]{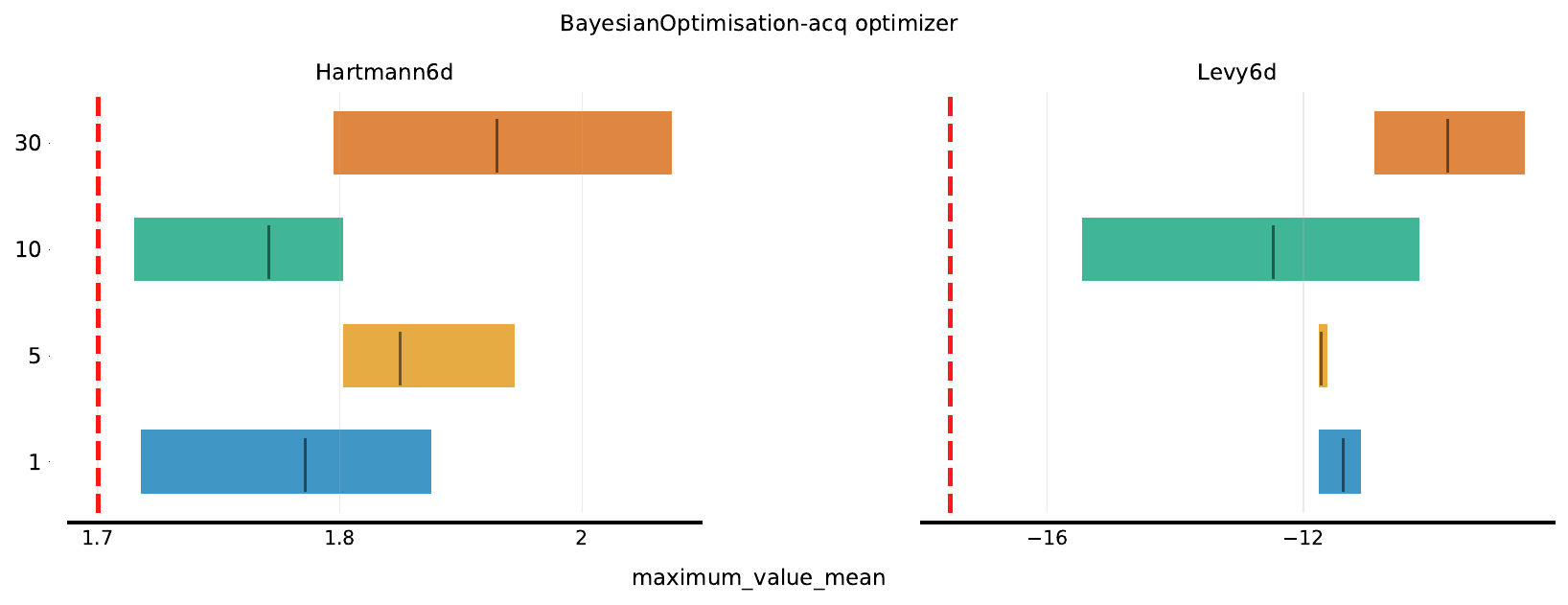}%
\\[0.5em]
\includegraphics[width=0.48\textwidth]{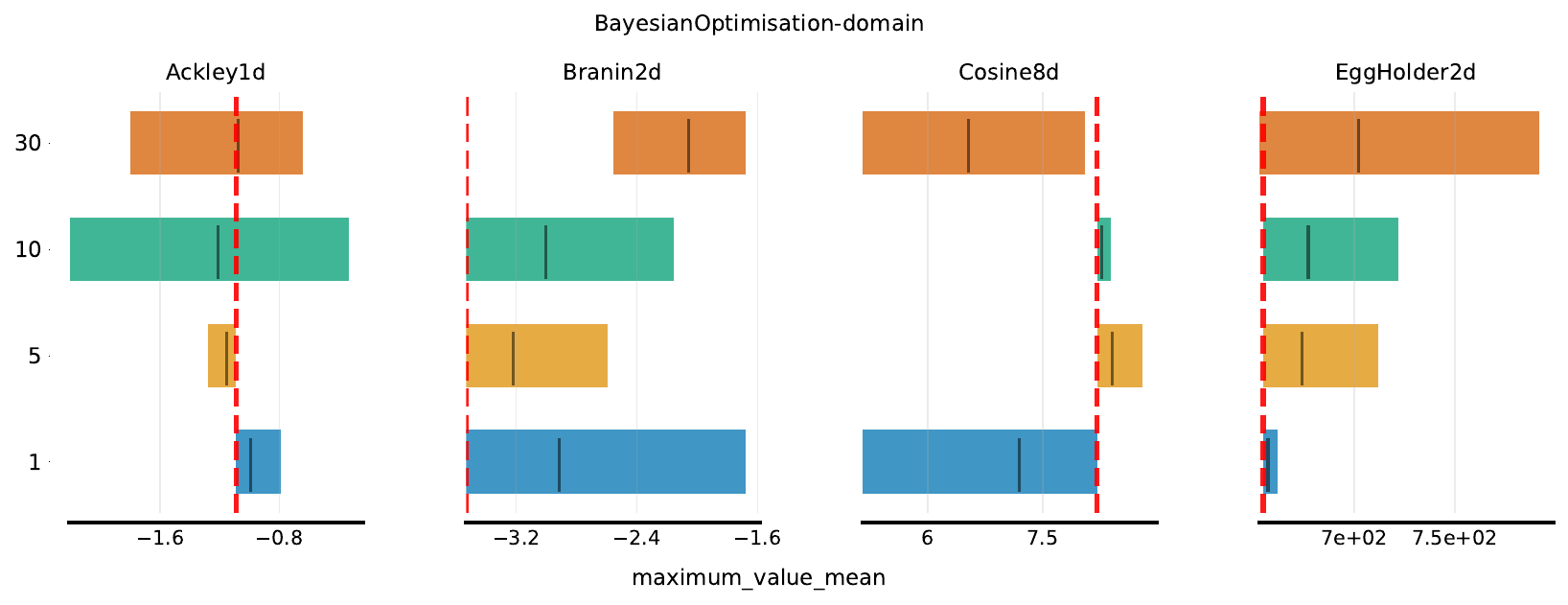}%
\hfill%
\includegraphics[width=0.48\textwidth]{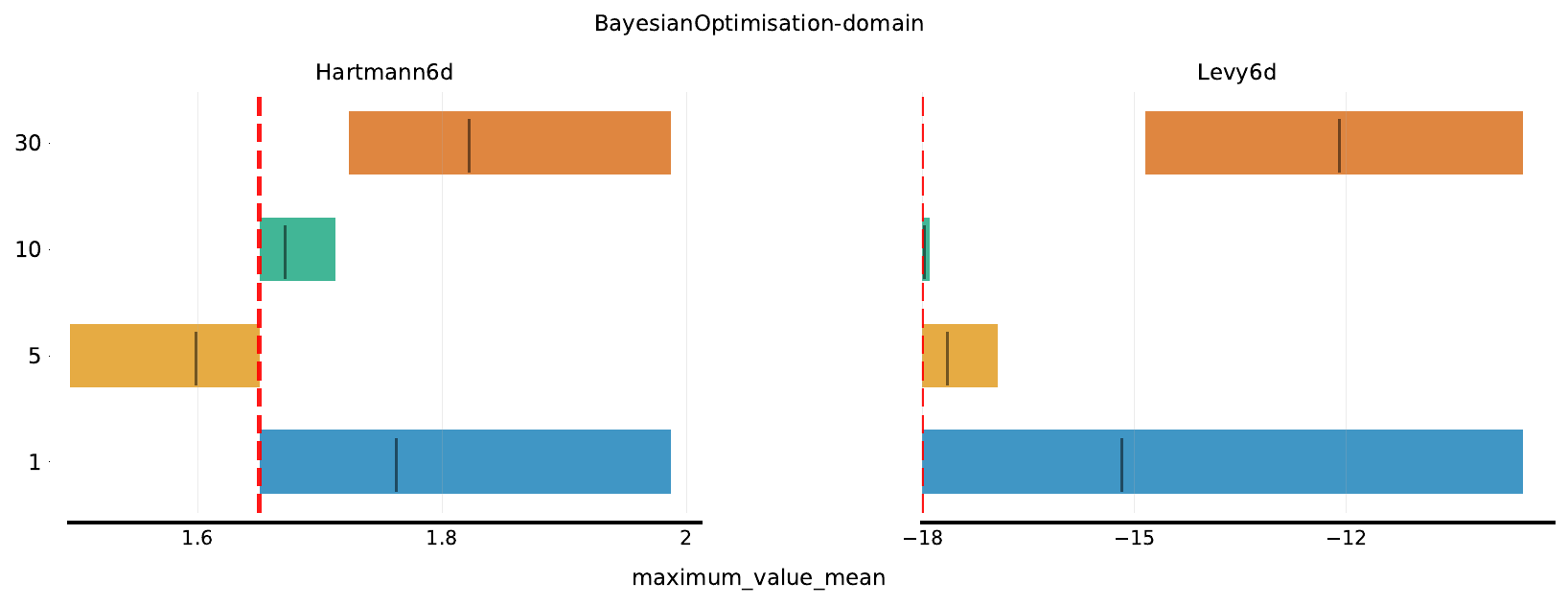}%
\\[0.5em]
\includegraphics[width=0.48\textwidth]{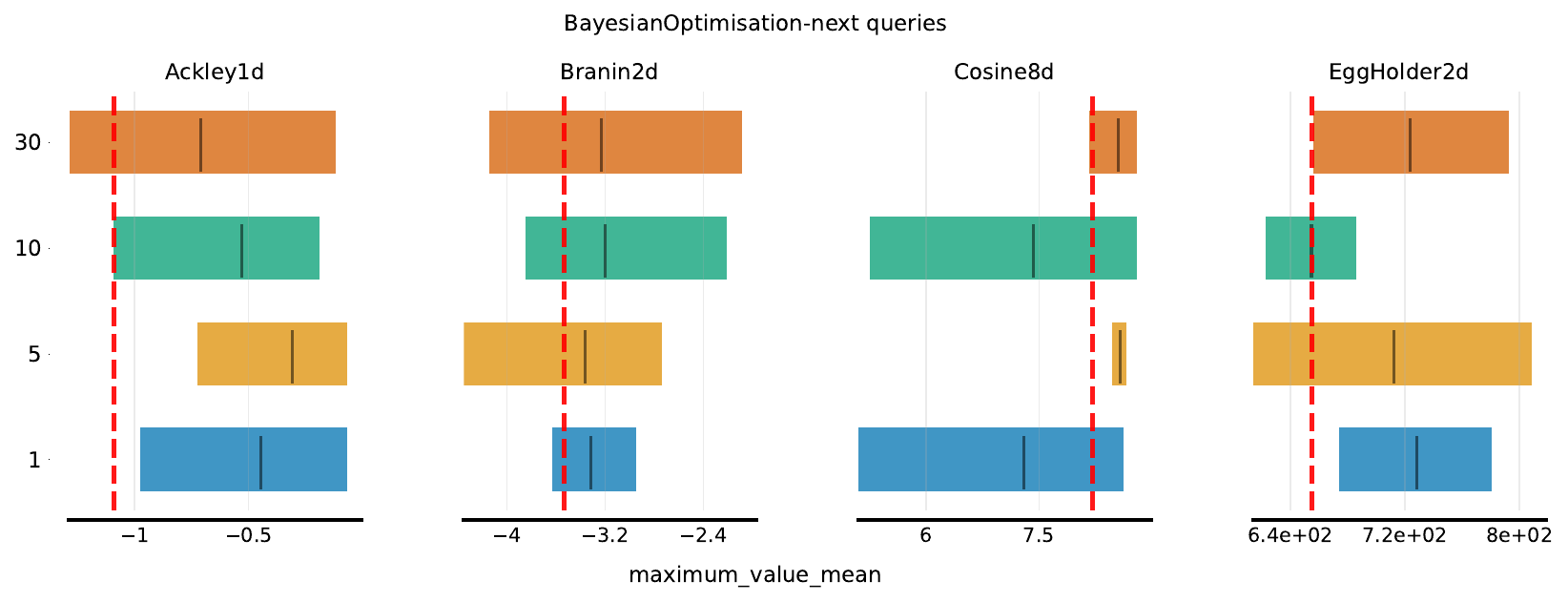}%
\hfill%
\includegraphics[width=0.48\textwidth]{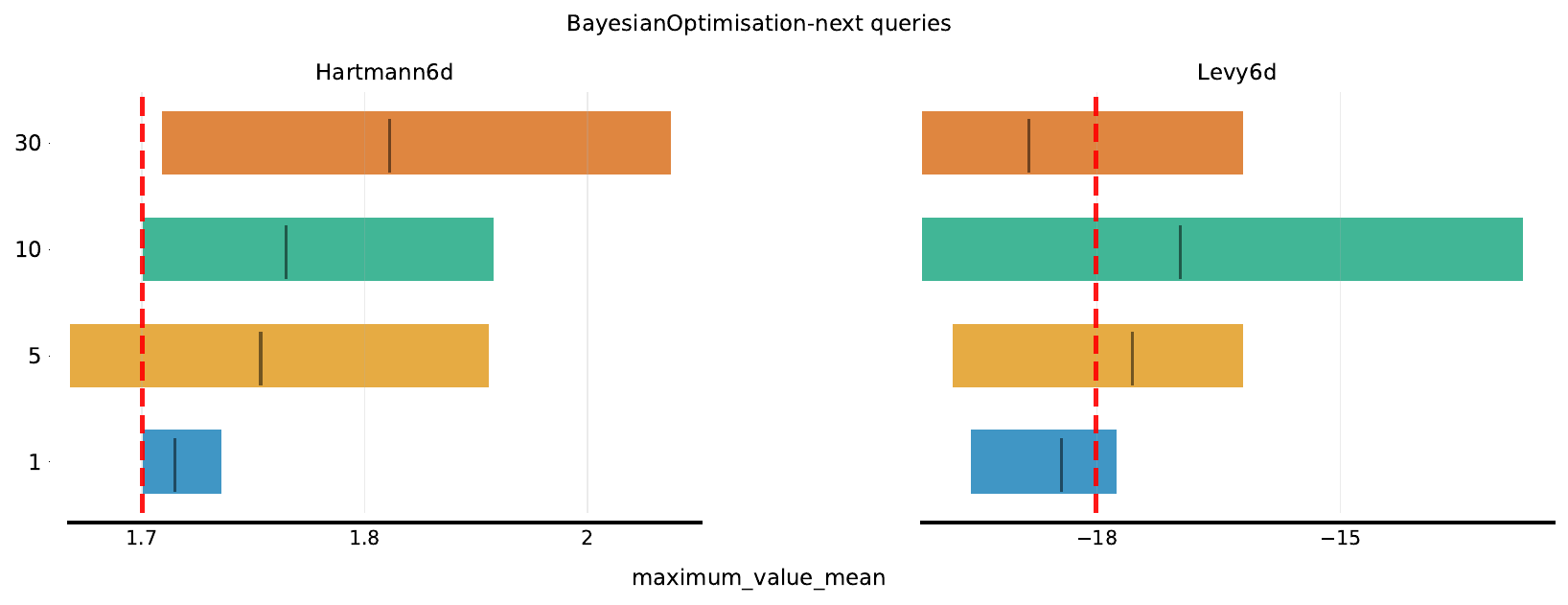}%
\caption{ADA Optimisation results on Meta-Train tasks. (Part 1/7)}
\label{fig:ADA_optimisation_id_1}
\end{figure}
\clearpage

\begin{figure}[htbp]
\centering
\setlength{\lineskip}{0pt}
\includegraphics[width=0.48\textwidth]{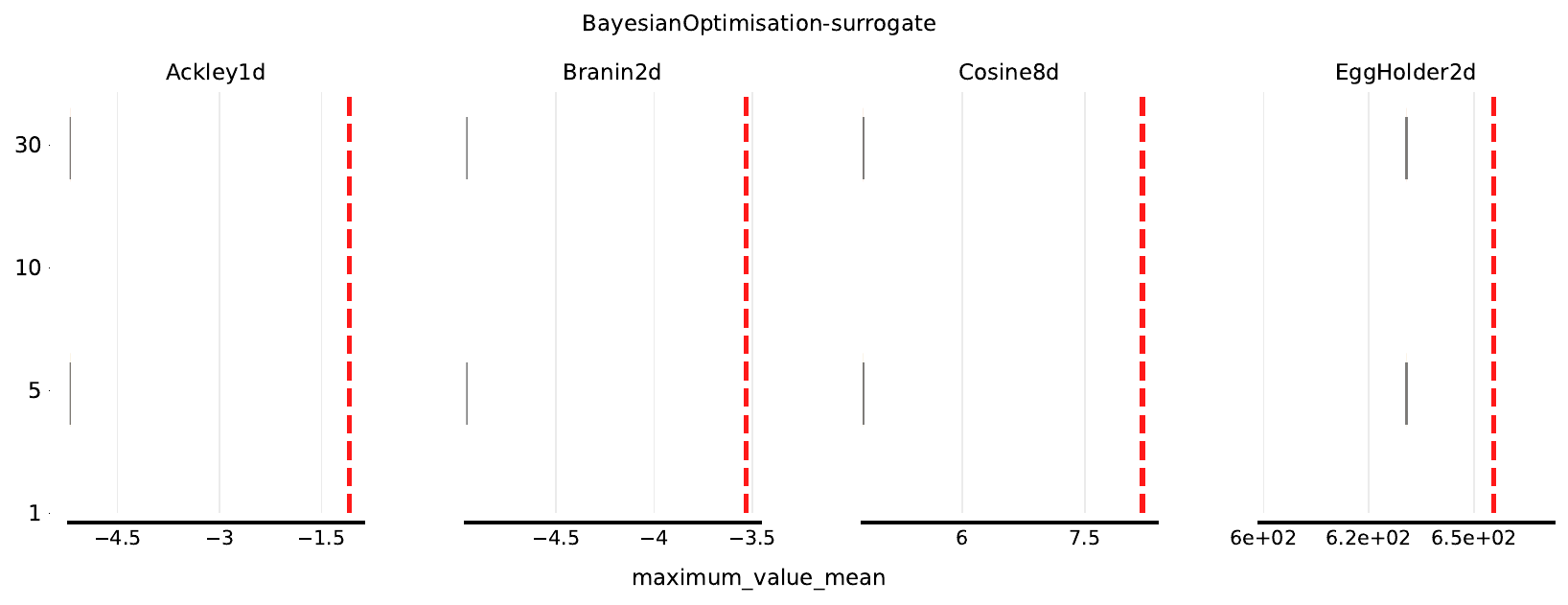}%
\hfill%
\includegraphics[width=0.48\textwidth]{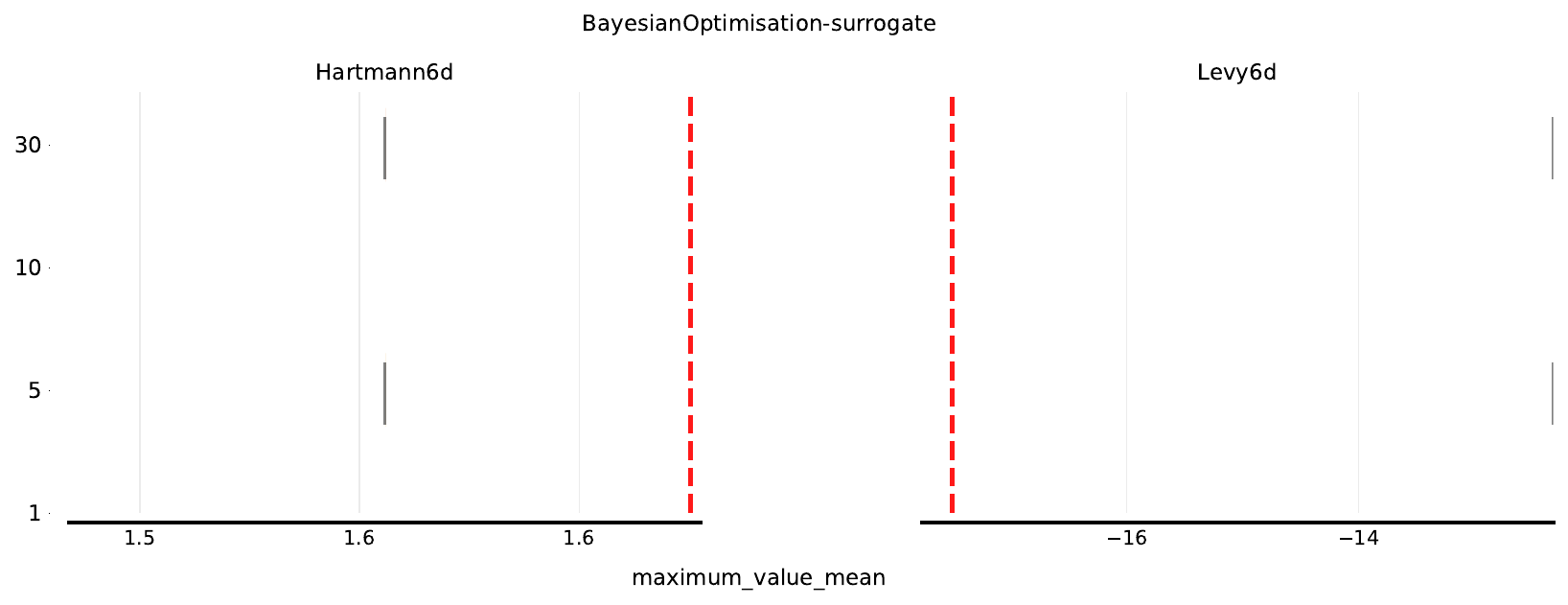}%
\\[0.5em]
\includegraphics[width=0.48\textwidth]{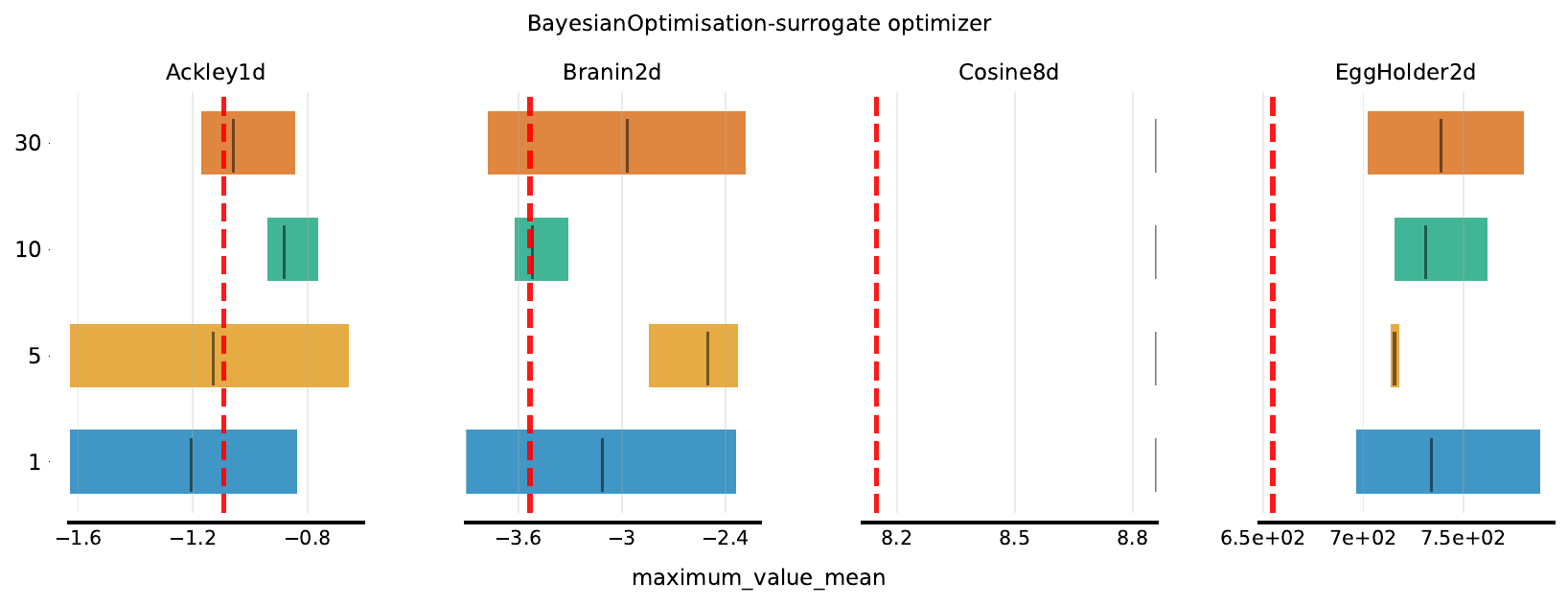}%
\hfill%
\includegraphics[width=0.48\textwidth]{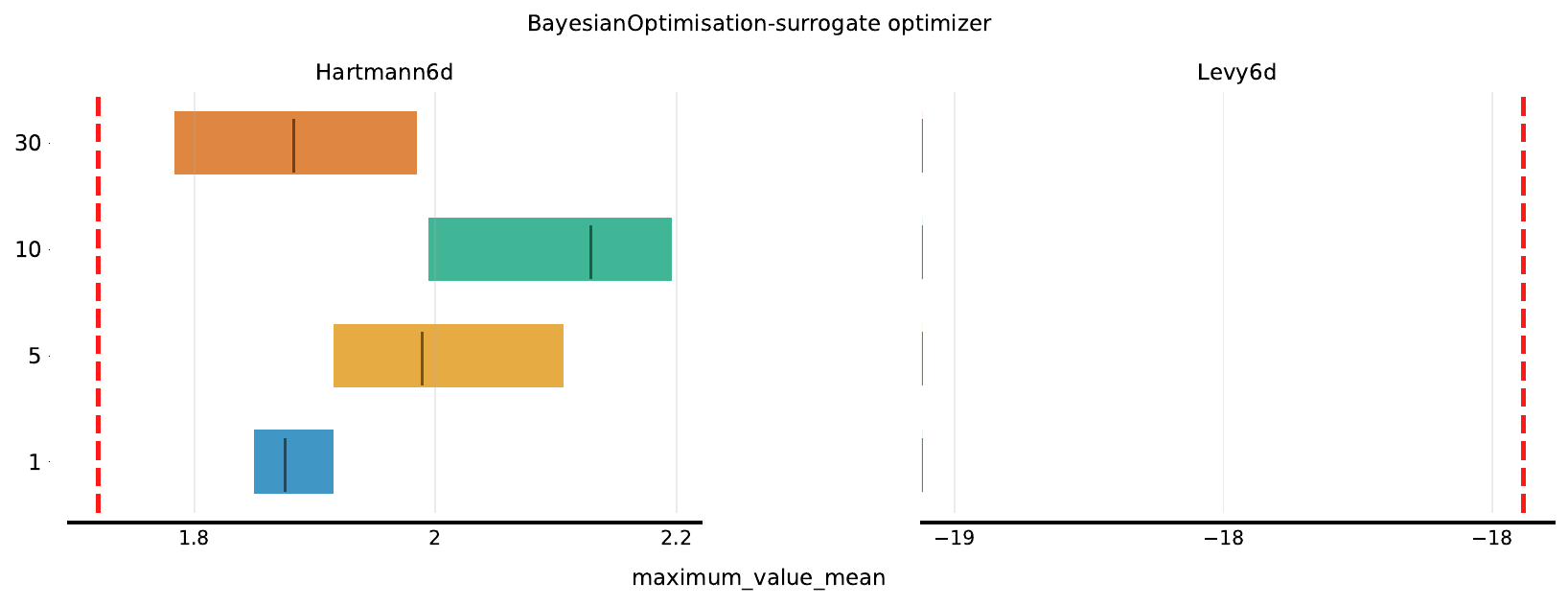}%
\\[0.5em]
\includegraphics[width=0.48\textwidth]{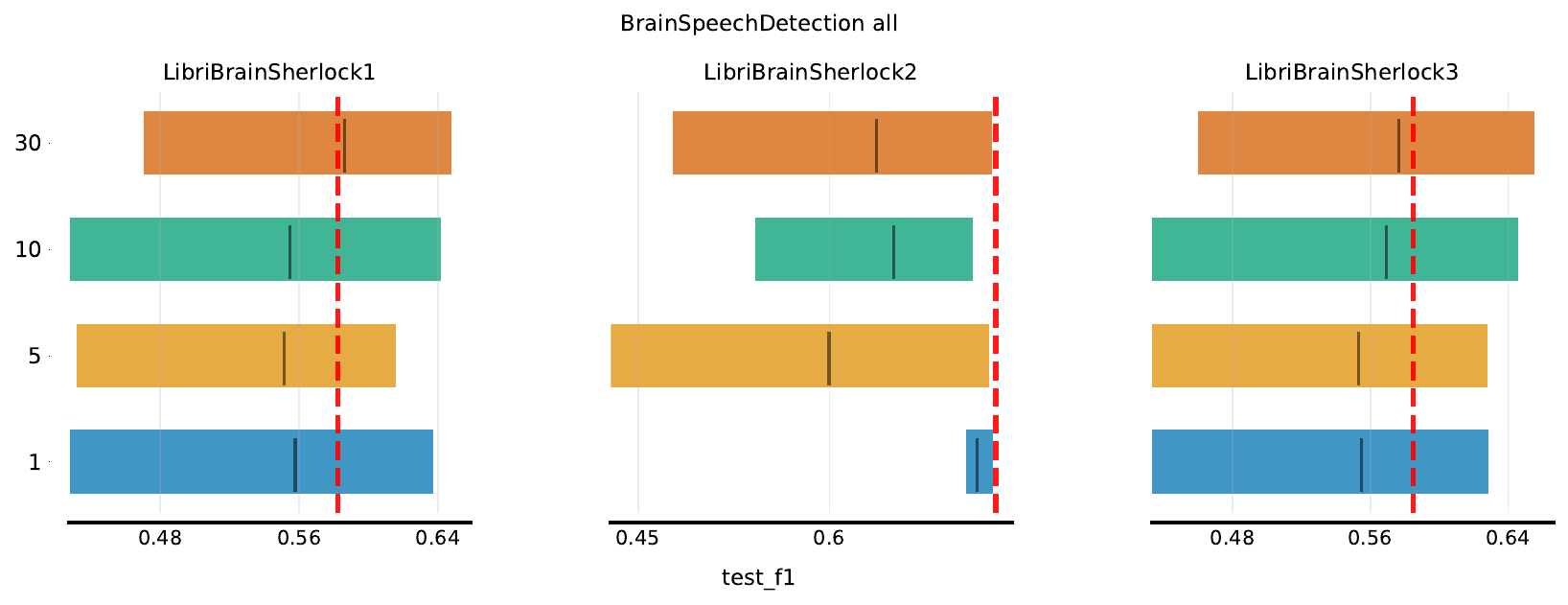}%
\hfill%
\includegraphics[width=0.48\textwidth]{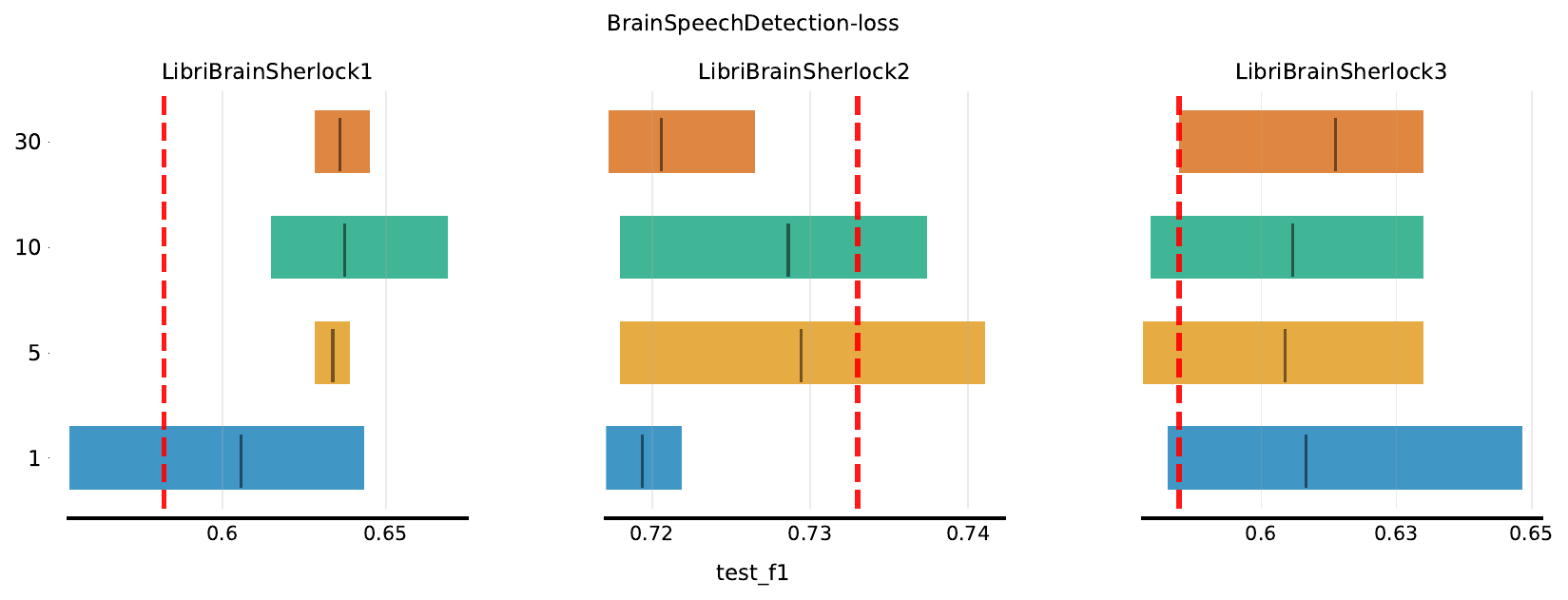}%
\\[0.5em]
\includegraphics[width=0.48\textwidth]{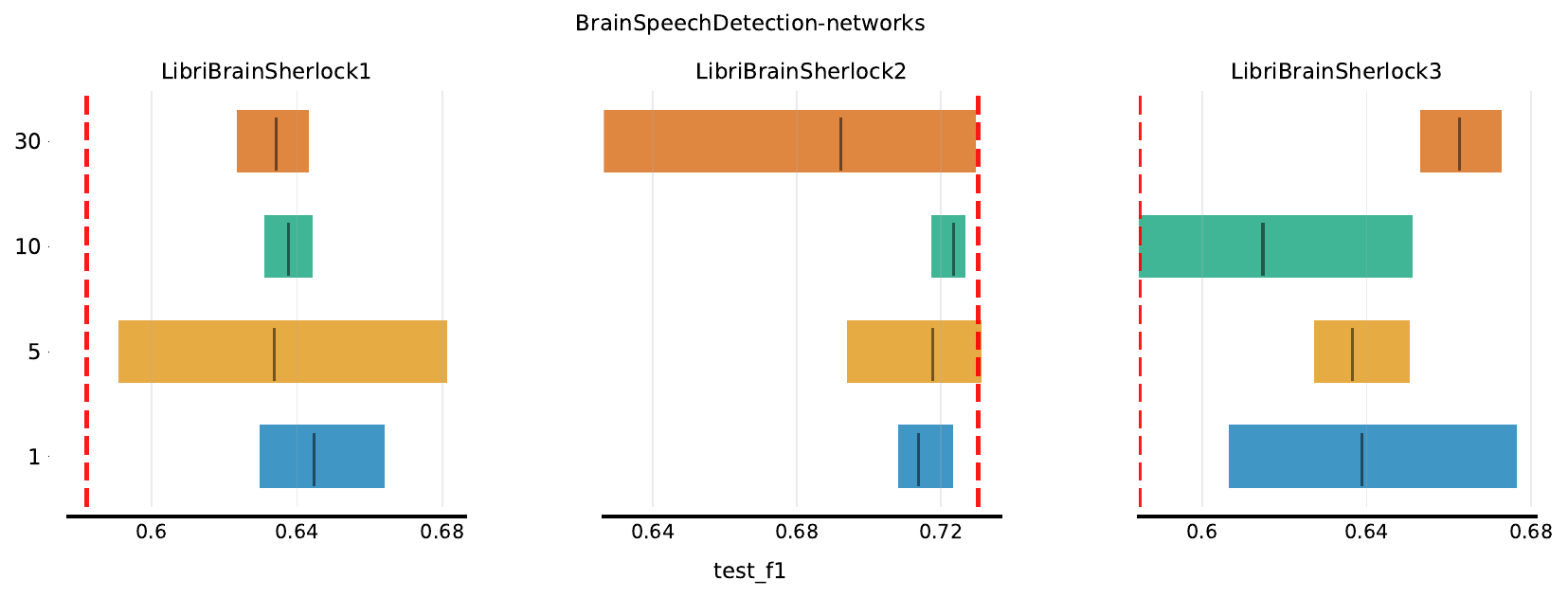}%
\hfill%
\includegraphics[width=0.48\textwidth]{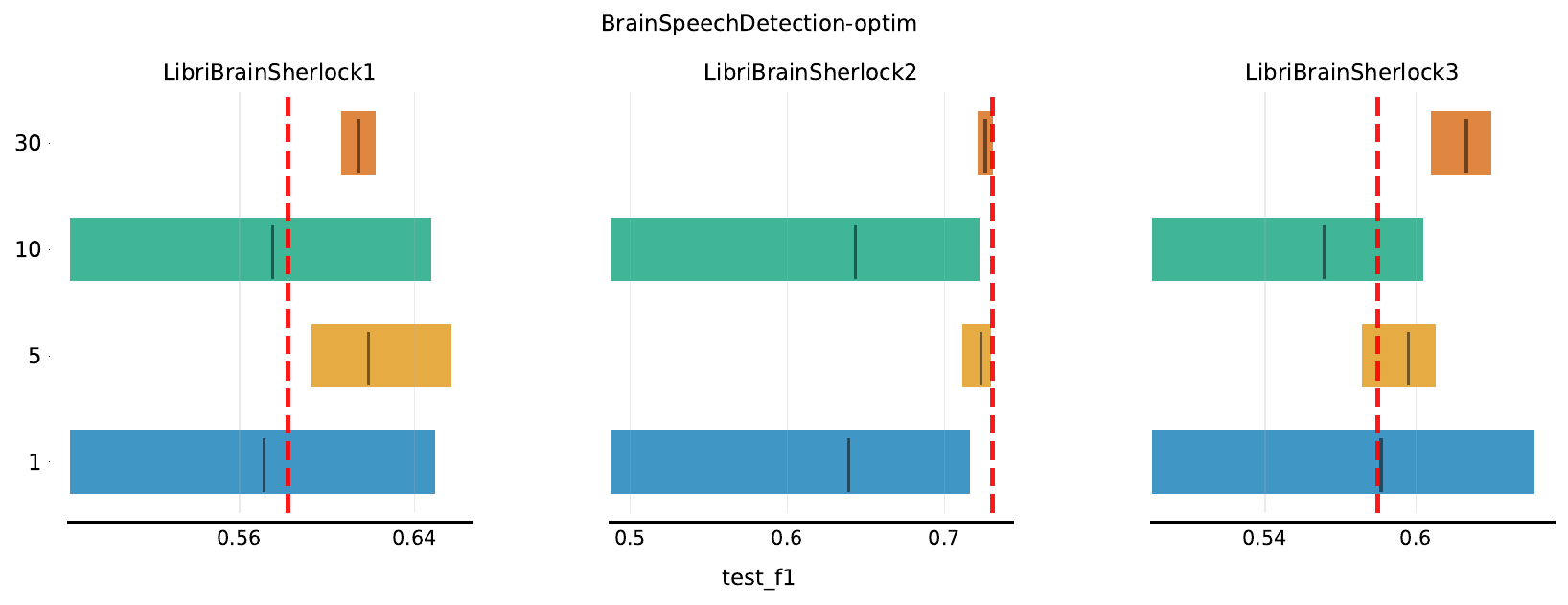}%
\\[0.5em]
\includegraphics[width=0.48\textwidth]{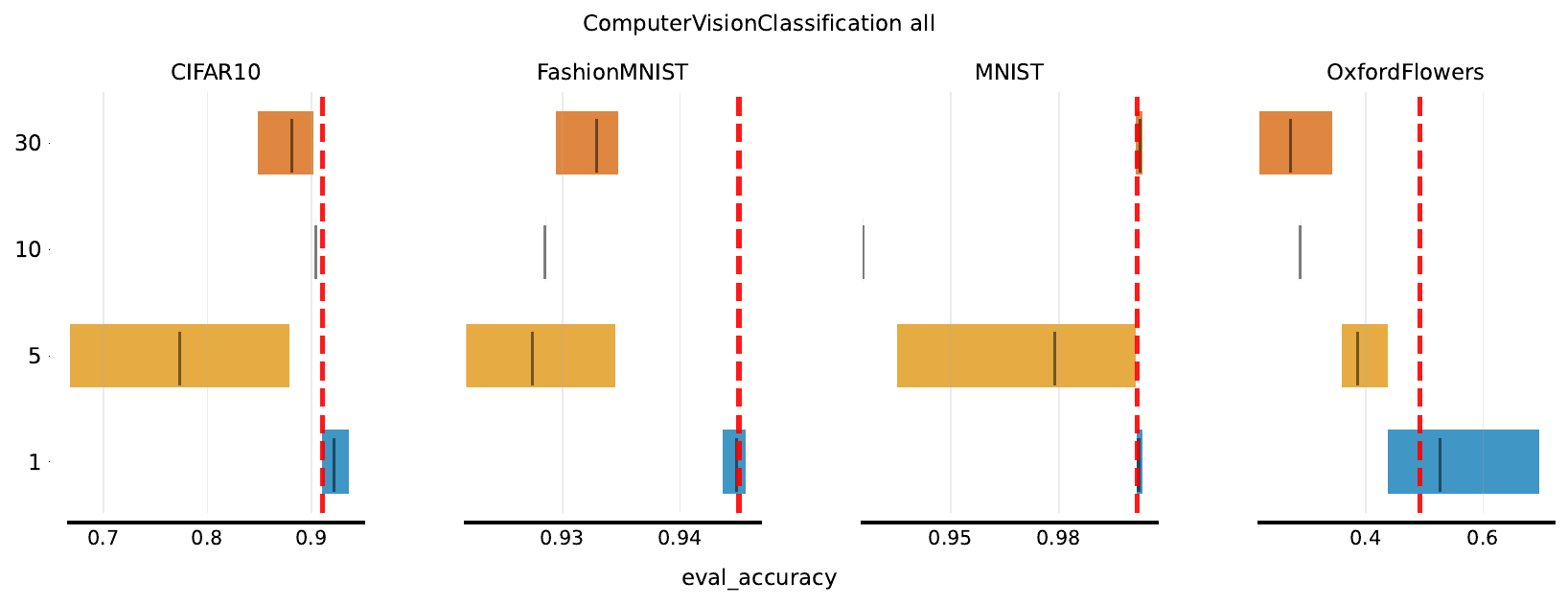}%
\hfill%
\includegraphics[width=0.48\textwidth]{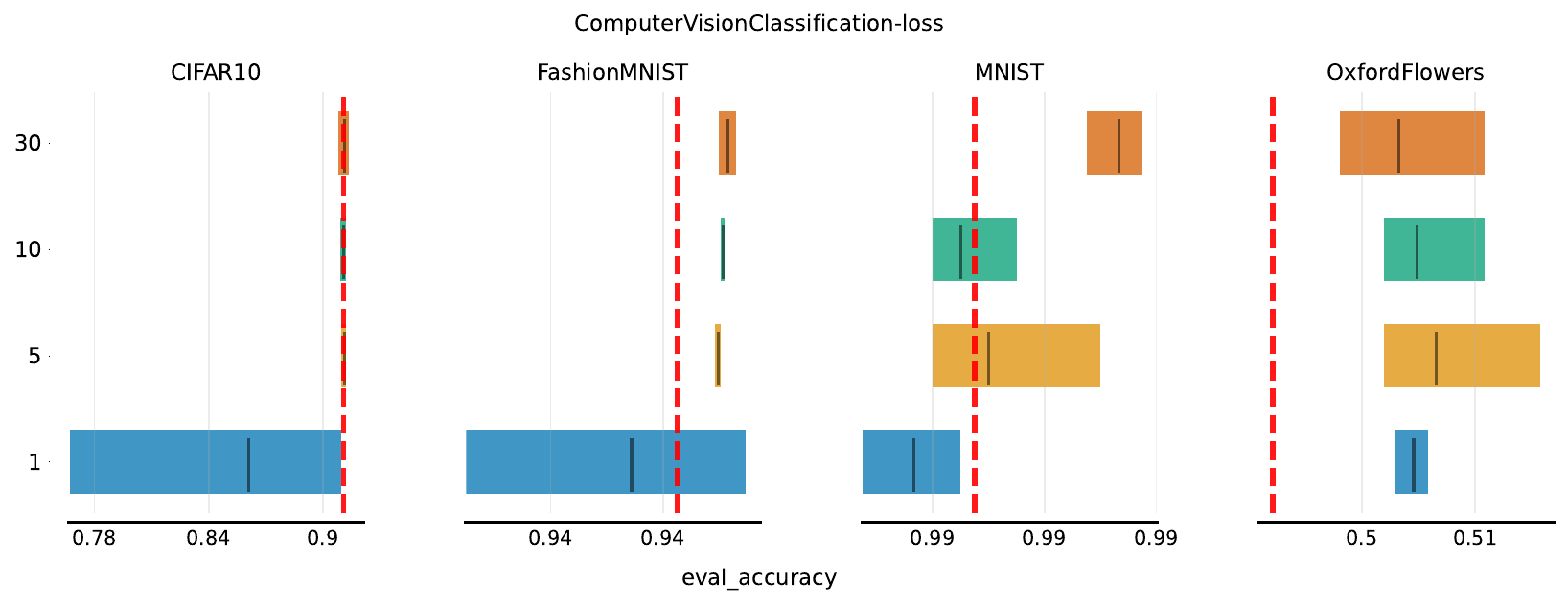}%
\\[0.5em]
\includegraphics[width=0.48\textwidth]{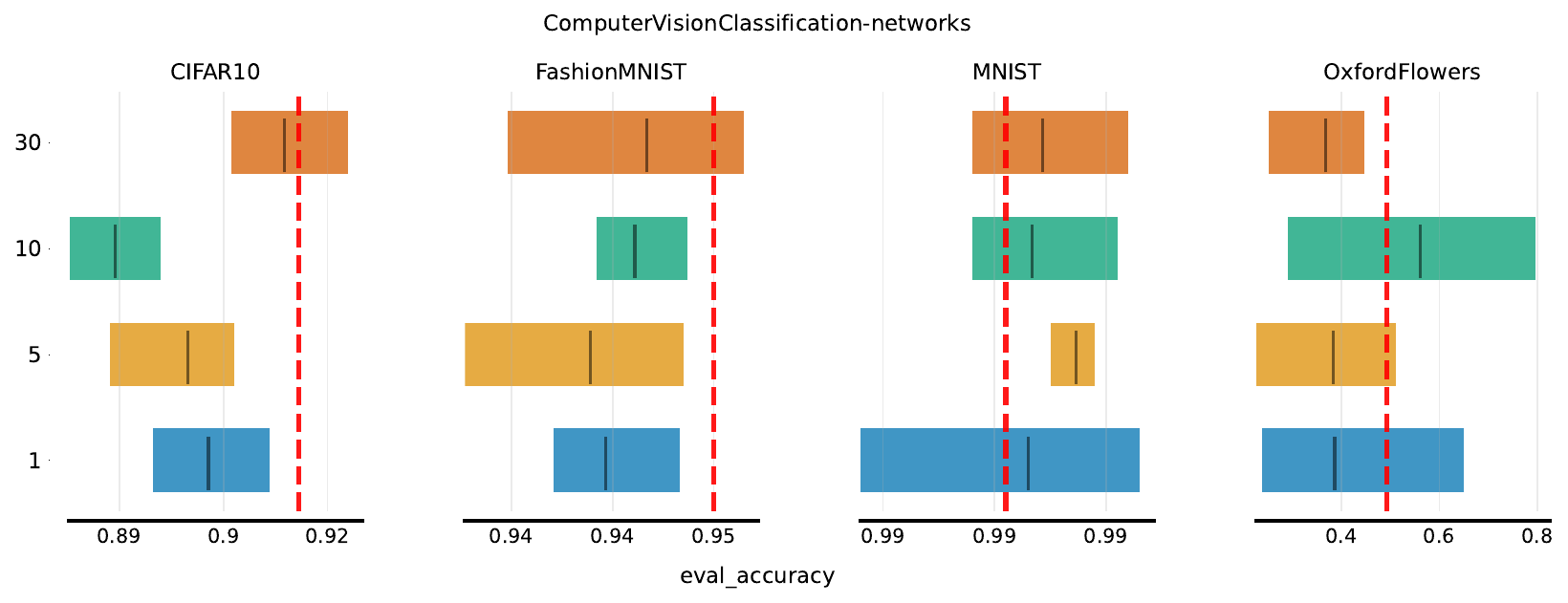}%
\hfill%
\includegraphics[width=0.48\textwidth]{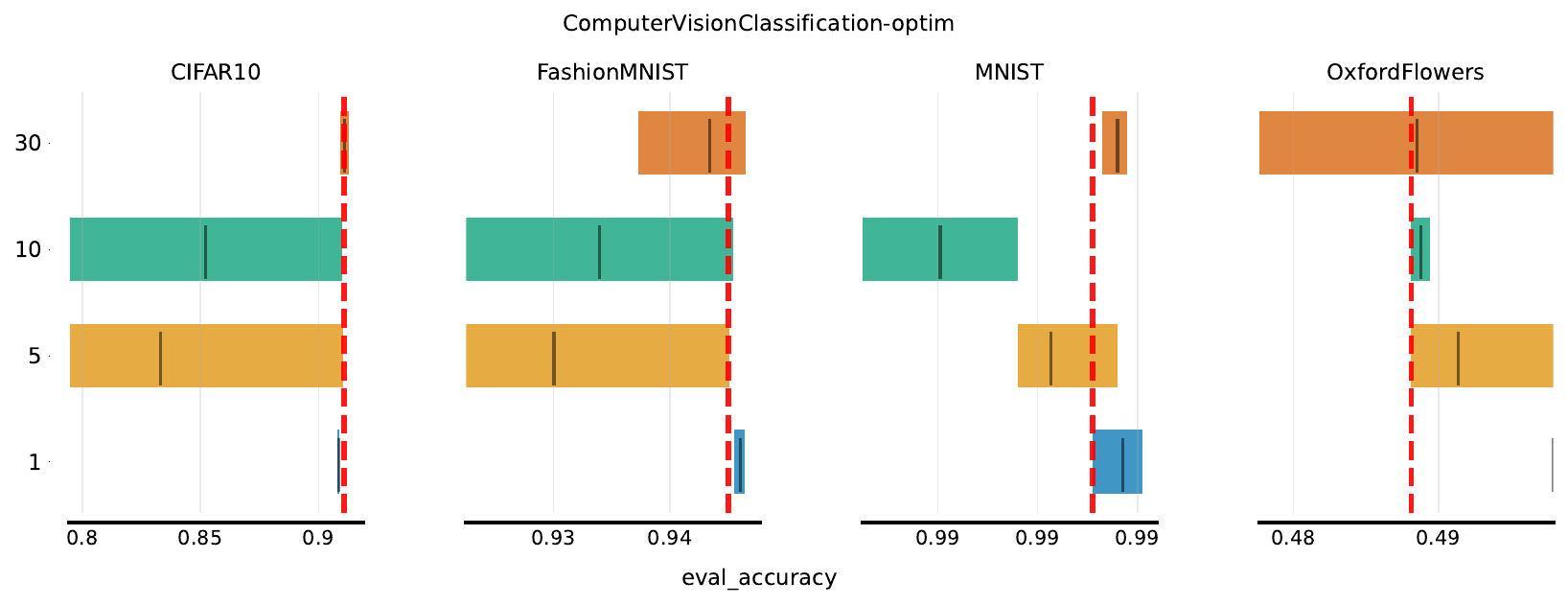}%
\caption{ADA Optimisation results on Meta-Train tasks. (Part 2/7)}
\label{fig:ADA_optimisation_id_2}
\end{figure}
\clearpage

\begin{figure}[htbp]
\centering
\setlength{\lineskip}{0pt}
\includegraphics[width=0.48\textwidth]{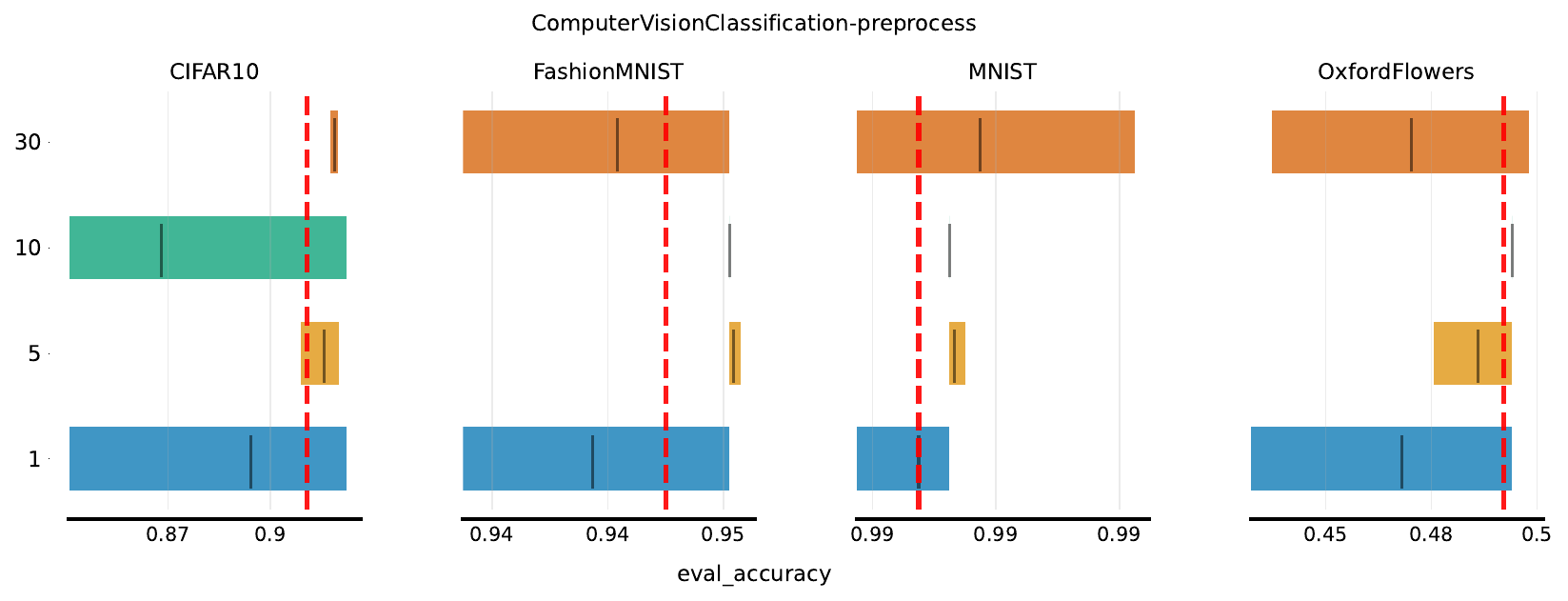}%
\hfill%
\includegraphics[width=0.48\textwidth]{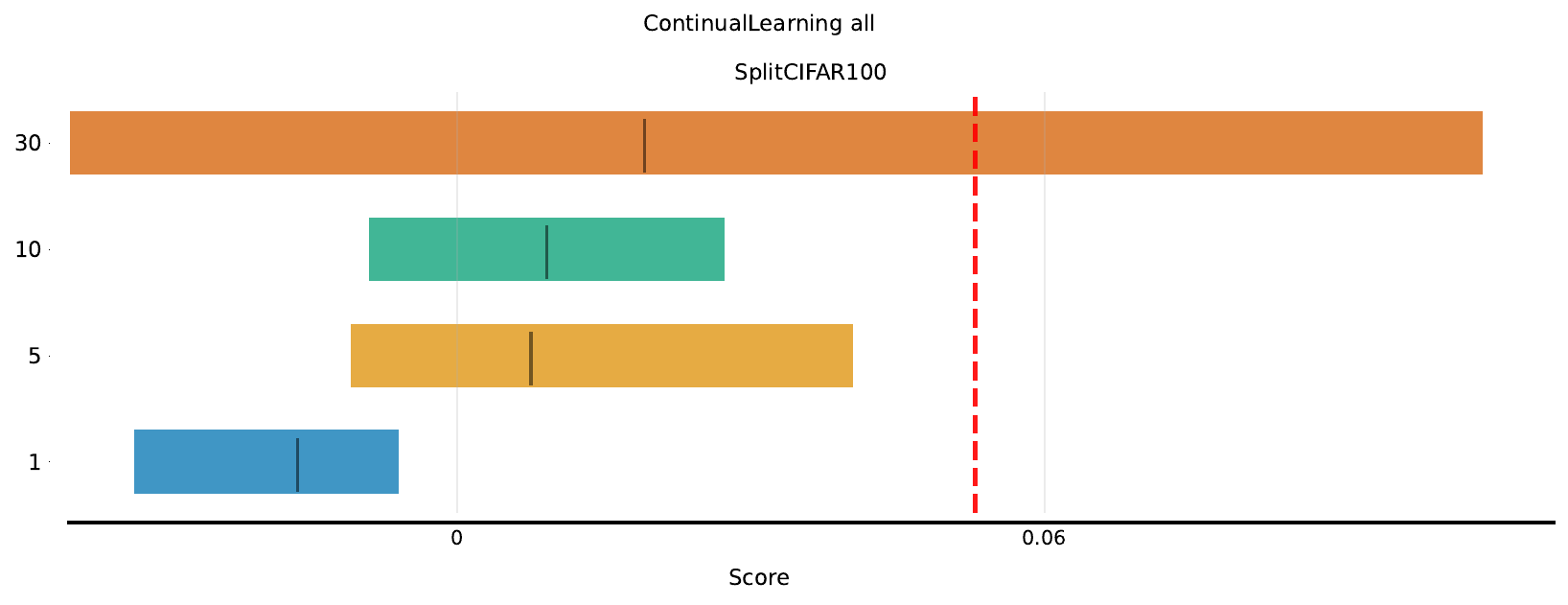}%
\\[0.5em]
\includegraphics[width=0.48\textwidth]{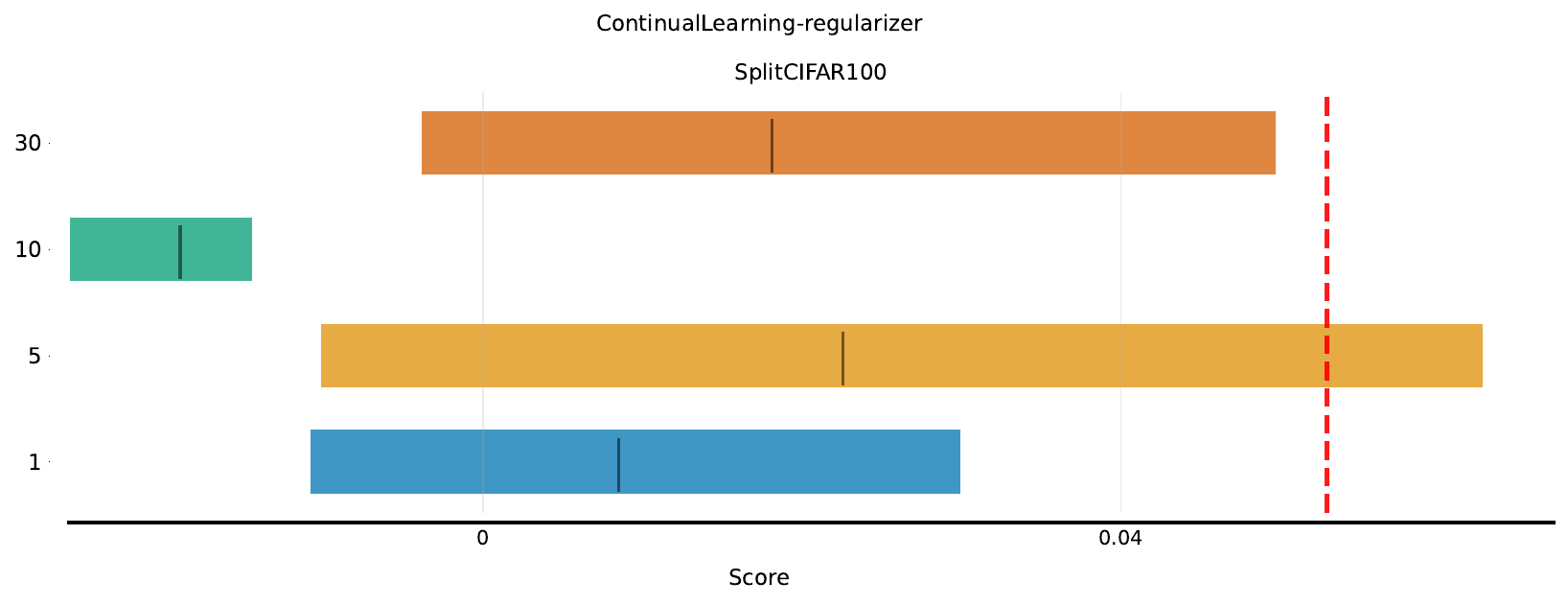}%
\hfill%
\includegraphics[width=0.48\textwidth]{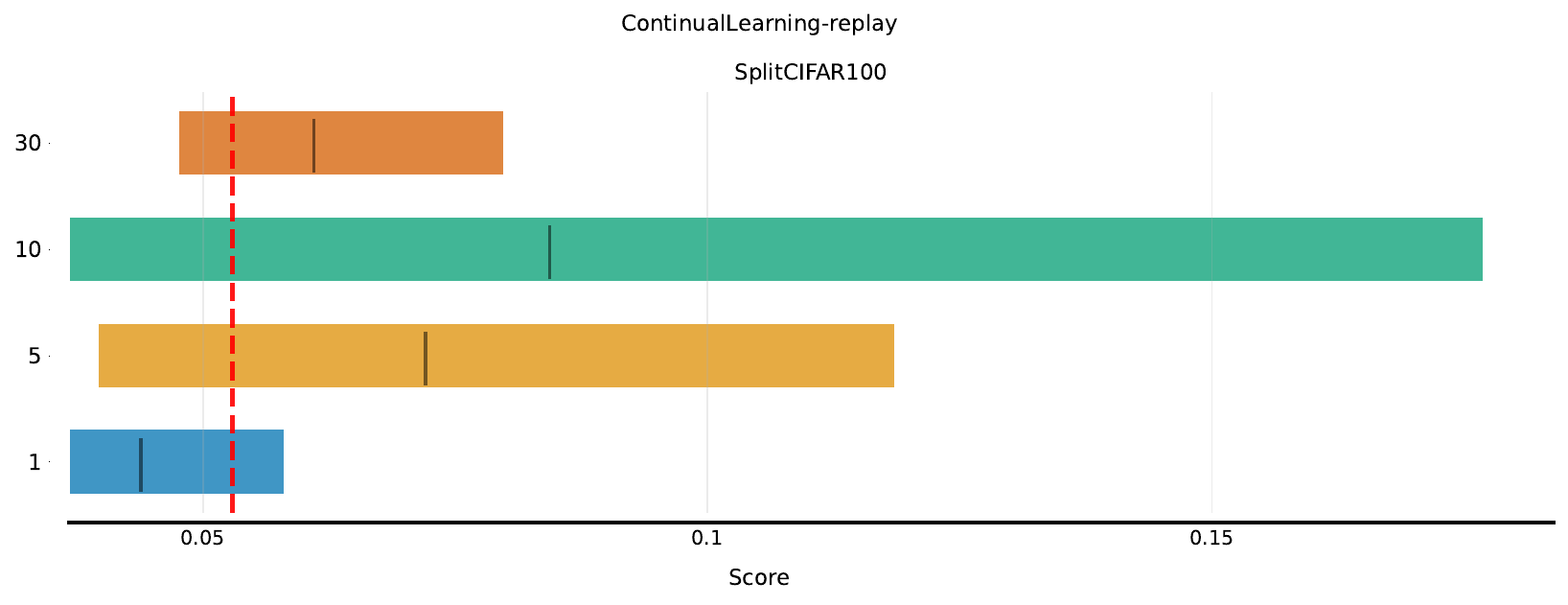}%
\\[0.5em]
\includegraphics[width=0.48\textwidth]{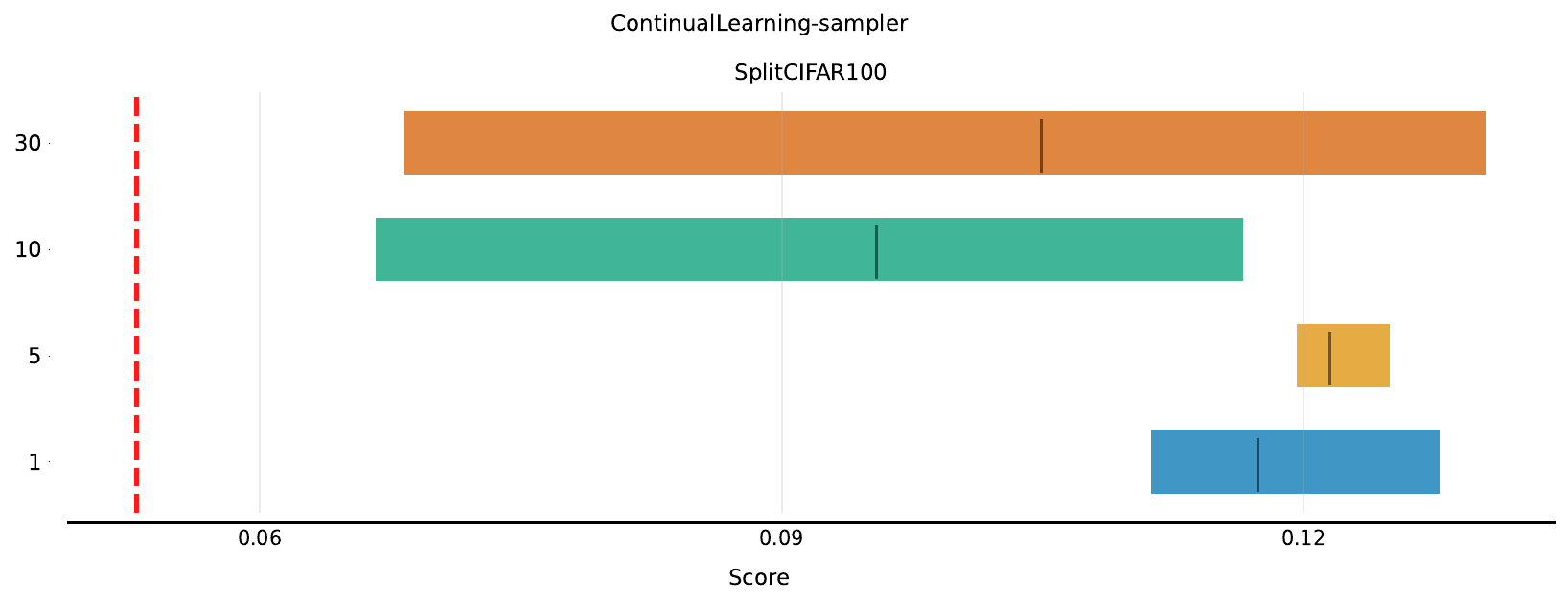}%
\hfill%
\includegraphics[width=0.48\textwidth]{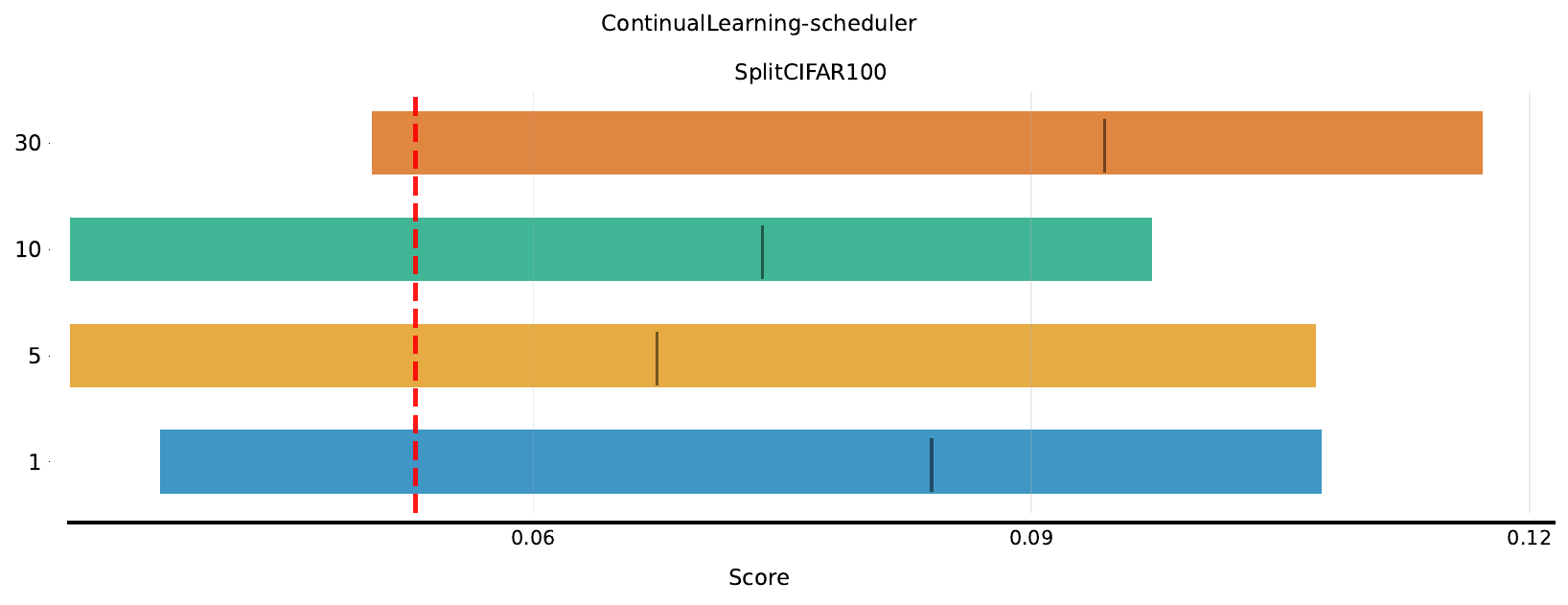}%
\\[0.5em]
\includegraphics[width=0.48\textwidth]{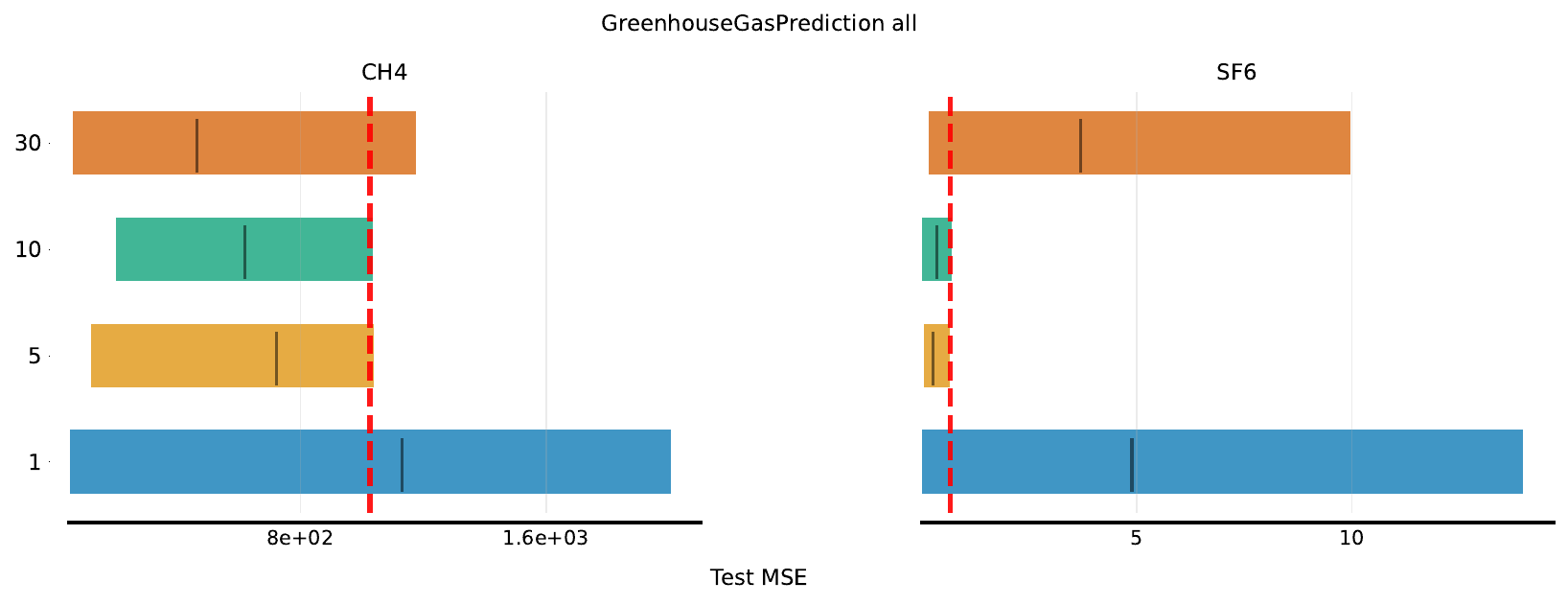}%
\hfill%
\includegraphics[width=0.48\textwidth]{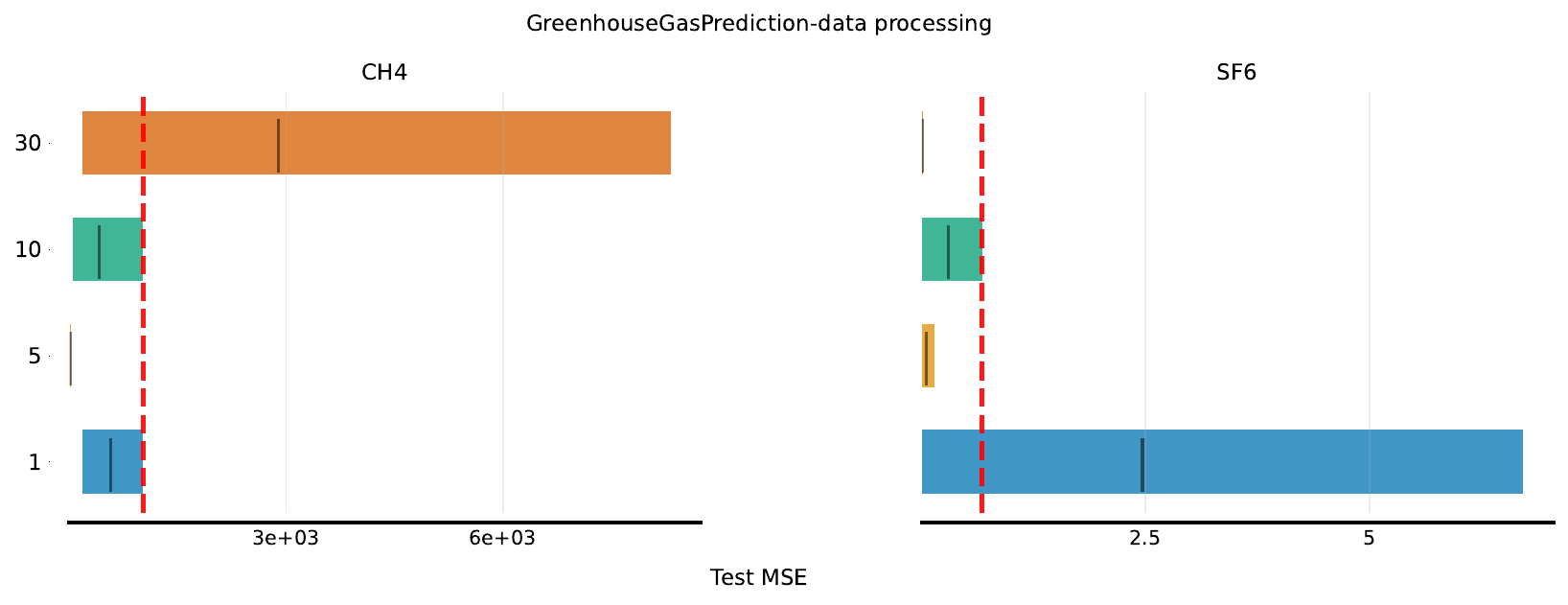}%
\\[0.5em]
\includegraphics[width=0.48\textwidth]{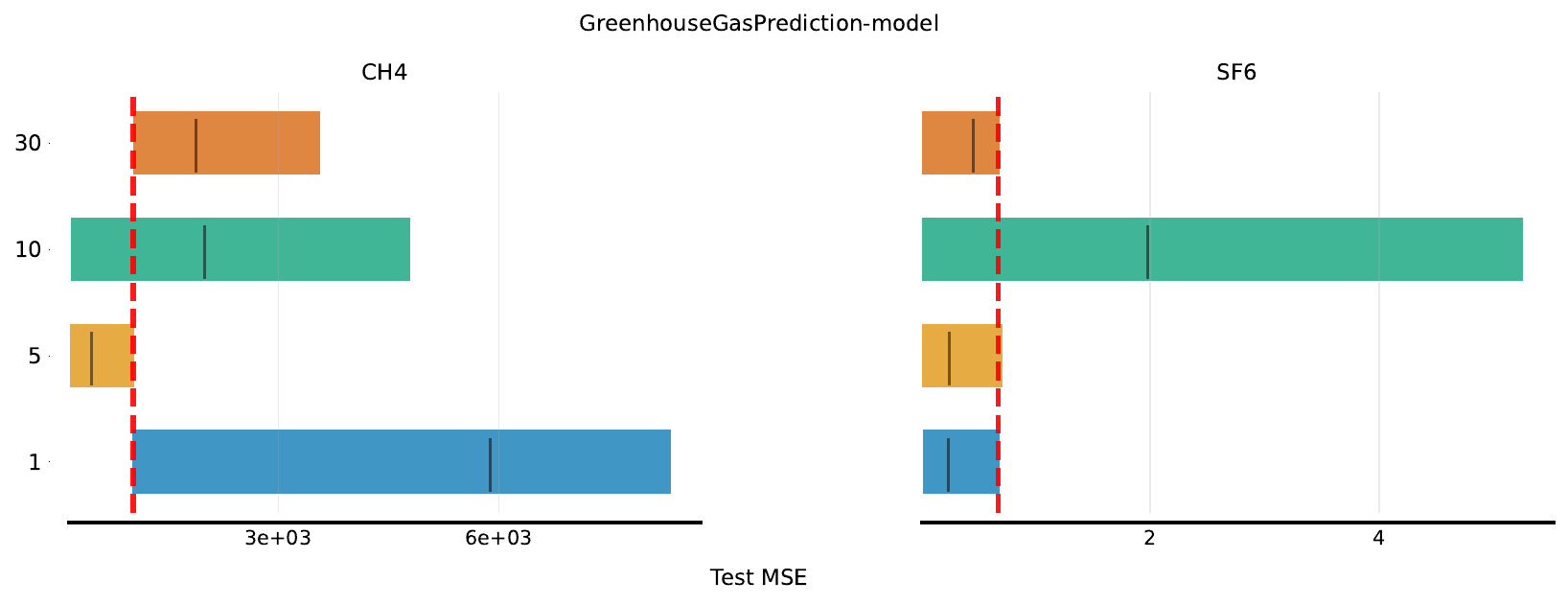}%
\hfill%
\includegraphics[width=0.48\textwidth]{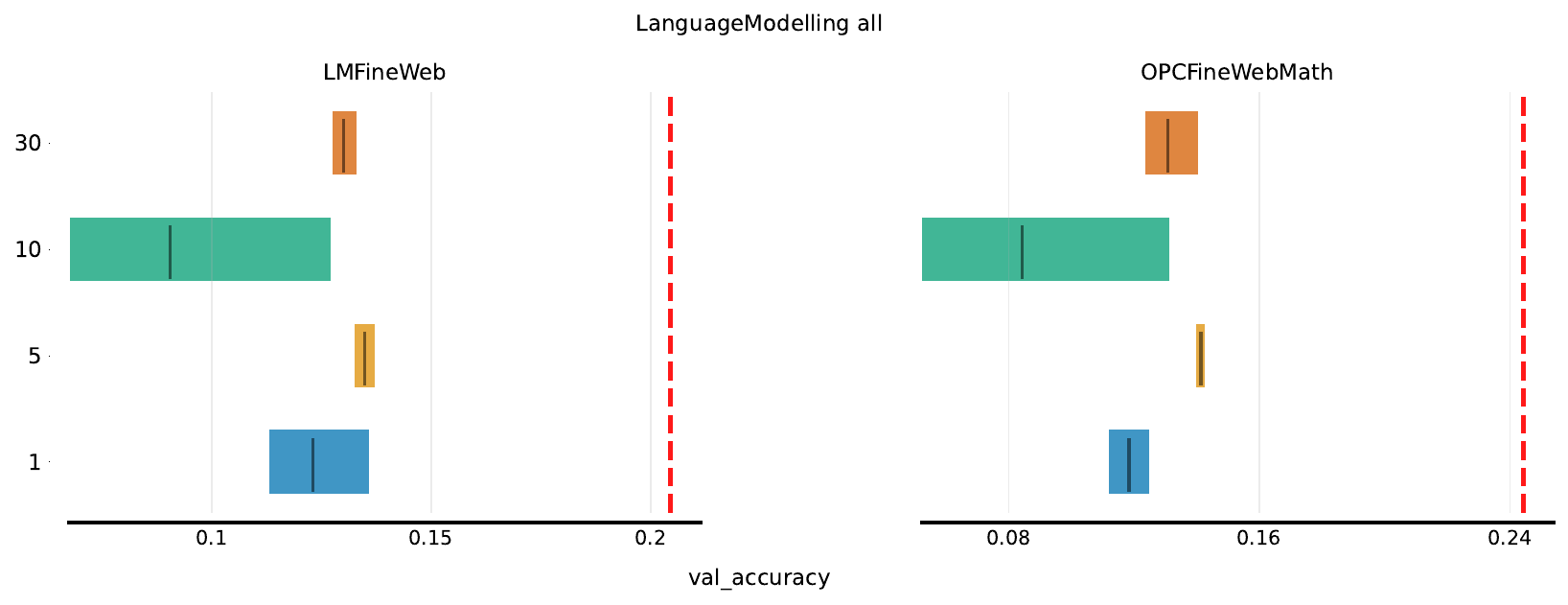}%
\\[0.5em]
\includegraphics[width=0.48\textwidth]{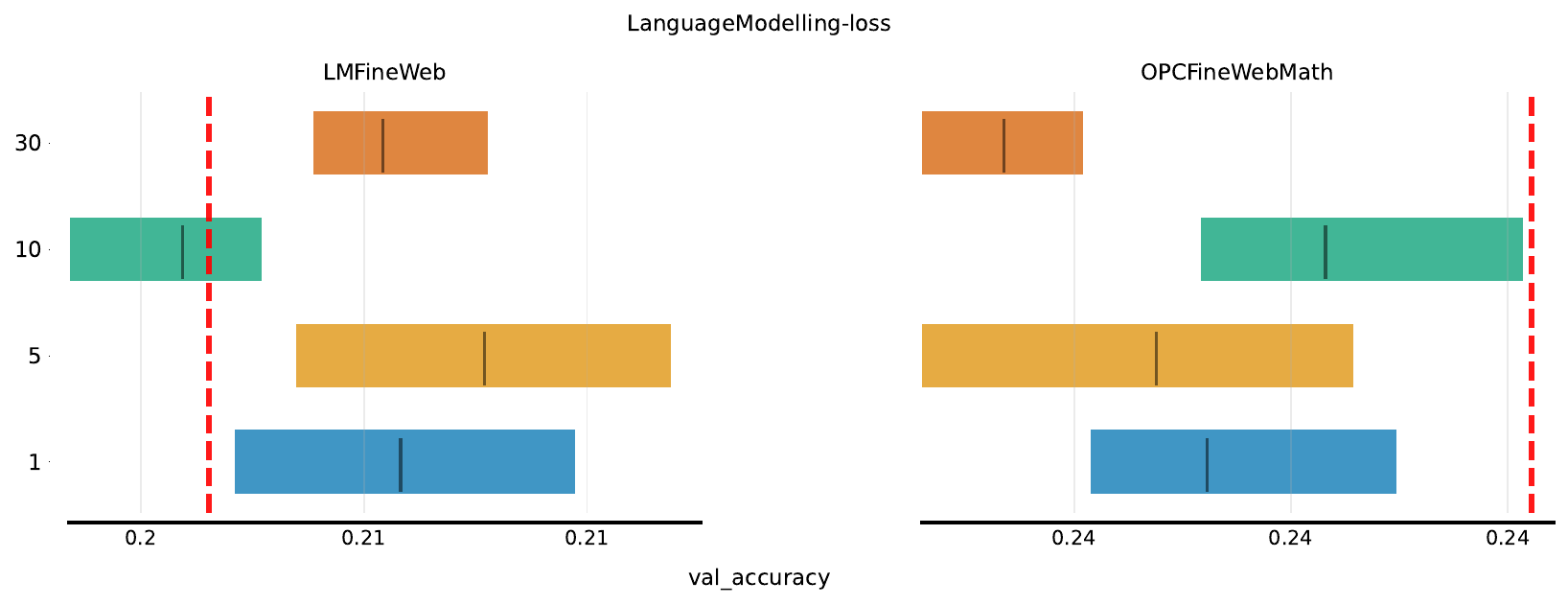}%
\hfill%
\includegraphics[width=0.48\textwidth]{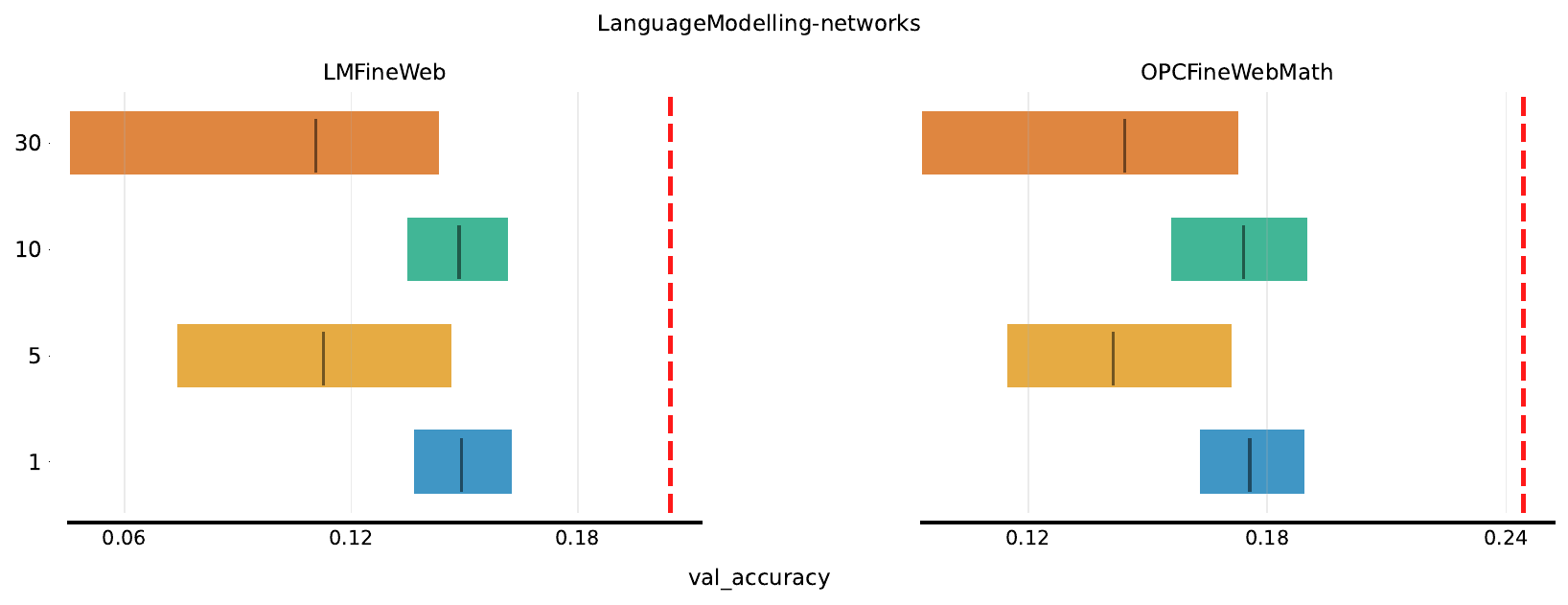}%
\caption{ADA Optimisation results on Meta-Train tasks. (Part 3/7)}
\label{fig:ADA_optimisation_id_3}
\end{figure}
\clearpage

\begin{figure}[htbp]
\centering
\setlength{\lineskip}{0pt}
\includegraphics[width=0.48\textwidth]{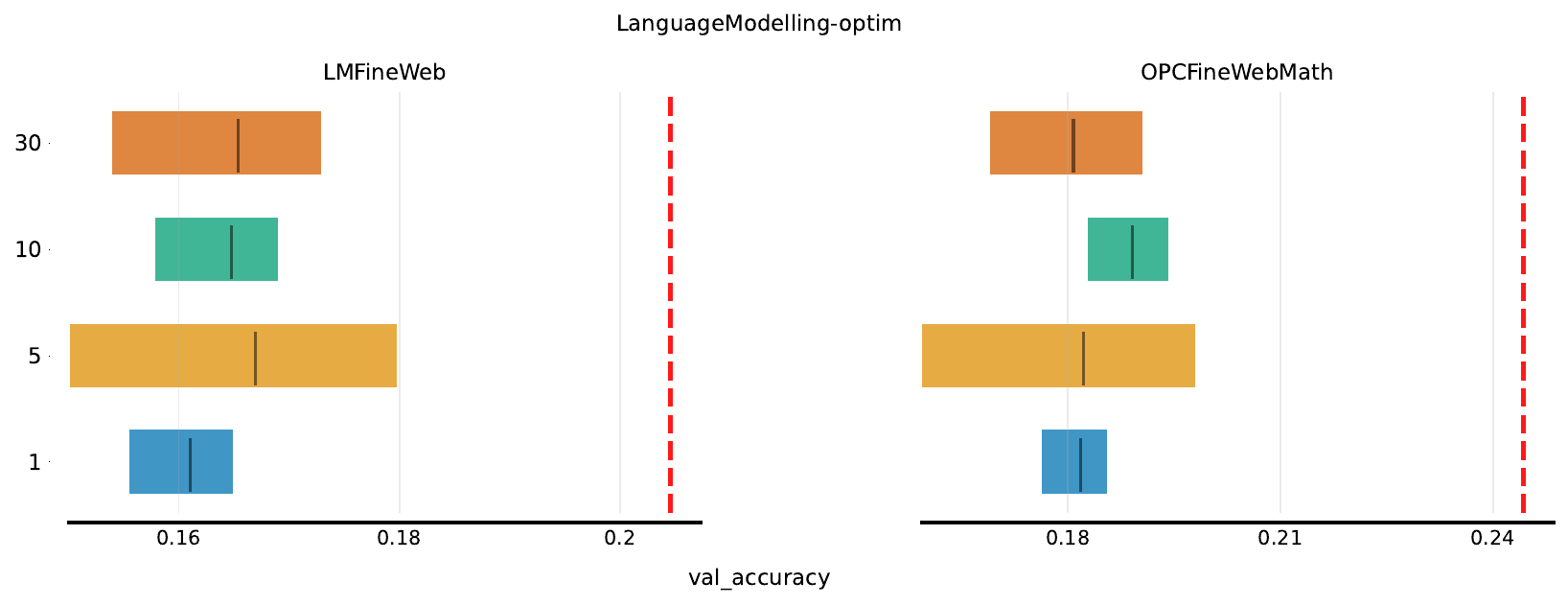}%
\hfill%
\includegraphics[width=0.48\textwidth]{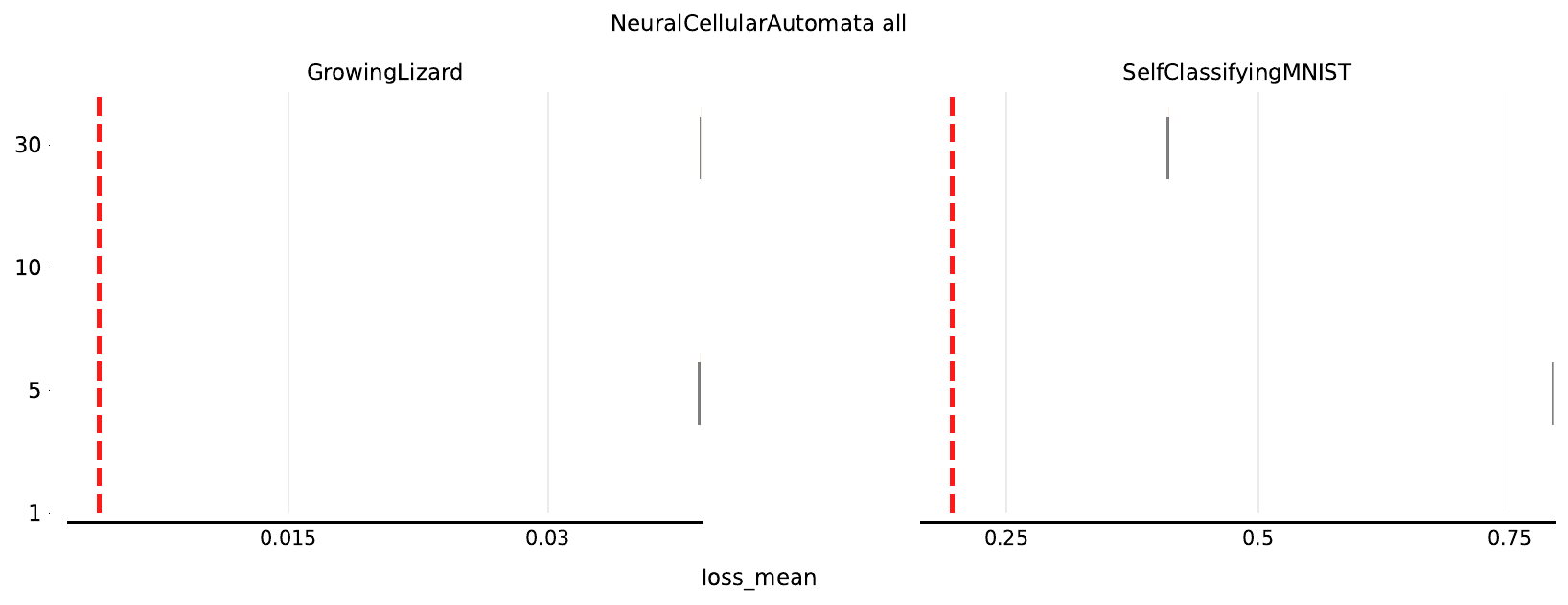}%
\\[0.5em]
\includegraphics[width=0.48\textwidth]{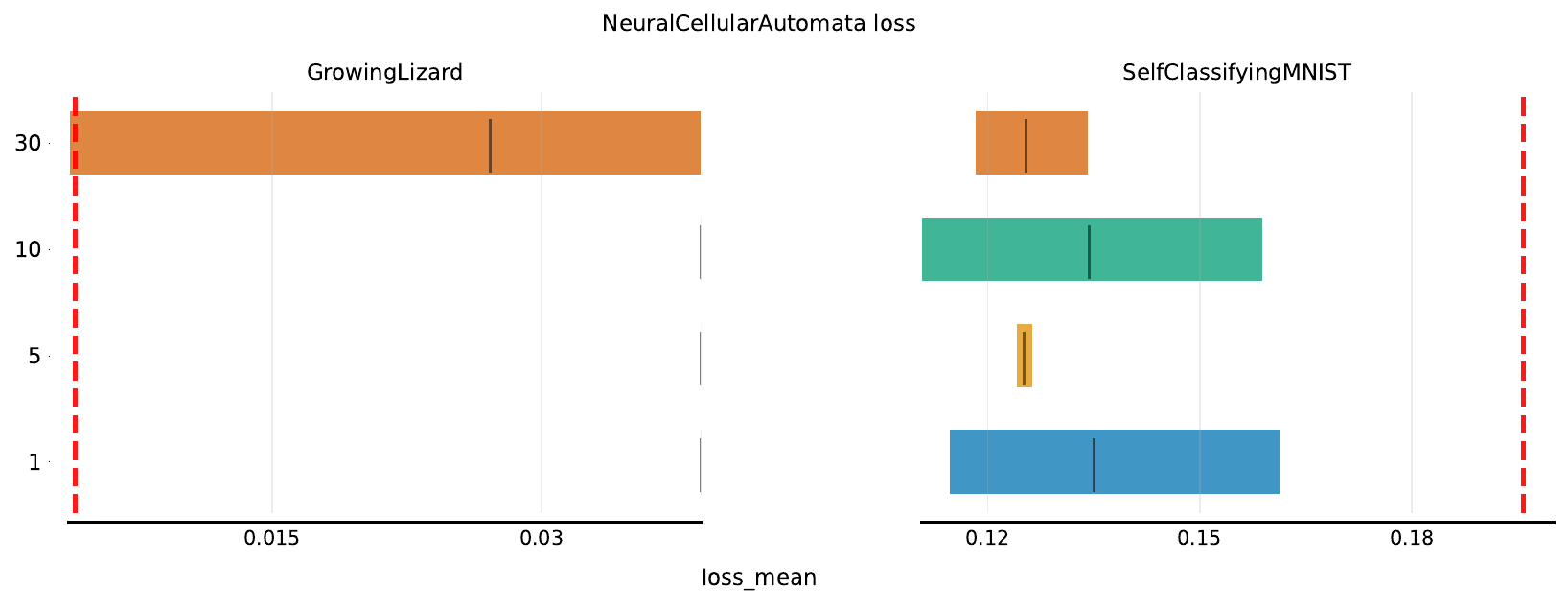}%
\hfill%
\includegraphics[width=0.48\textwidth]{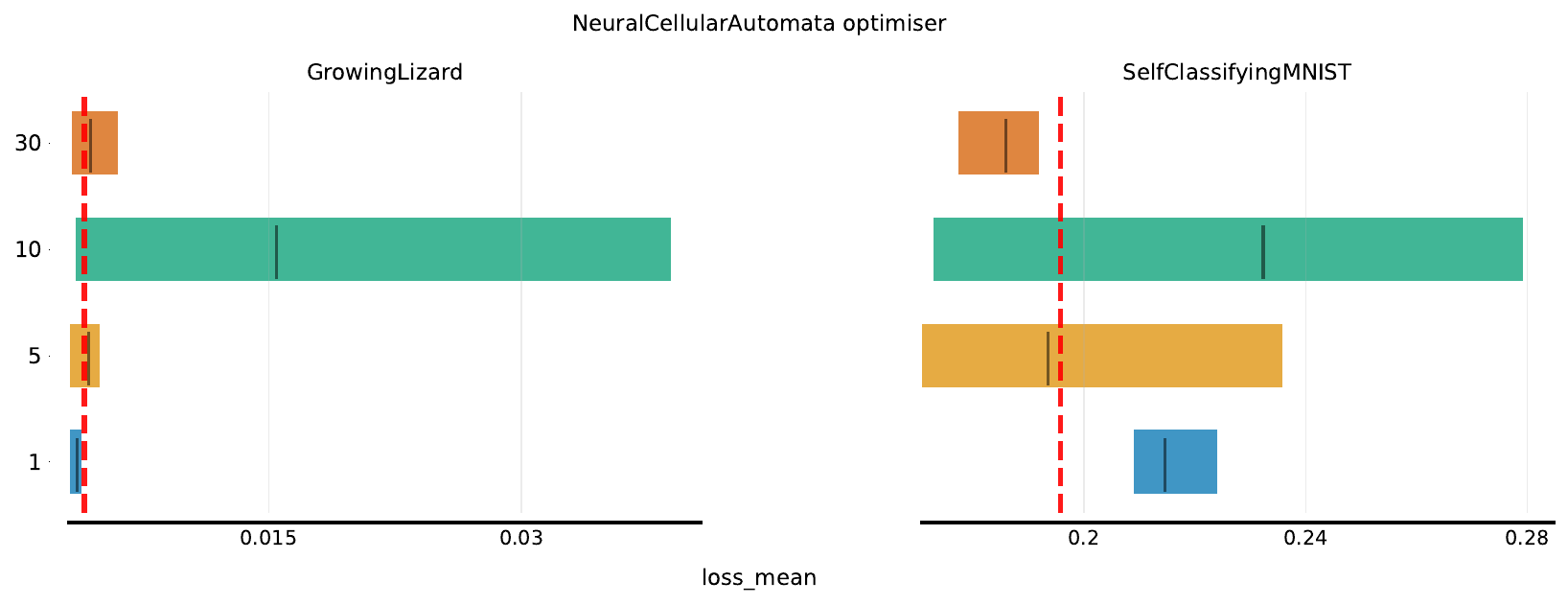}%
\\[0.5em]
\includegraphics[width=0.48\textwidth]{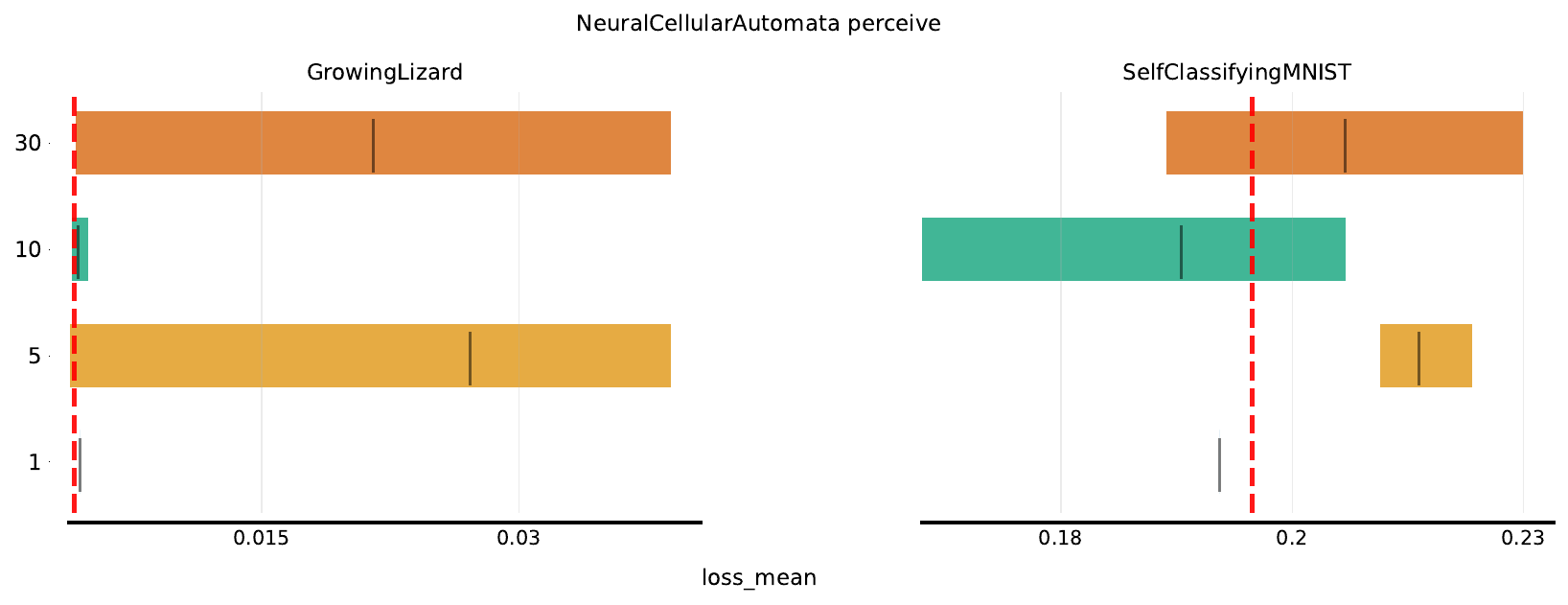}%
\hfill%
\includegraphics[width=0.48\textwidth]{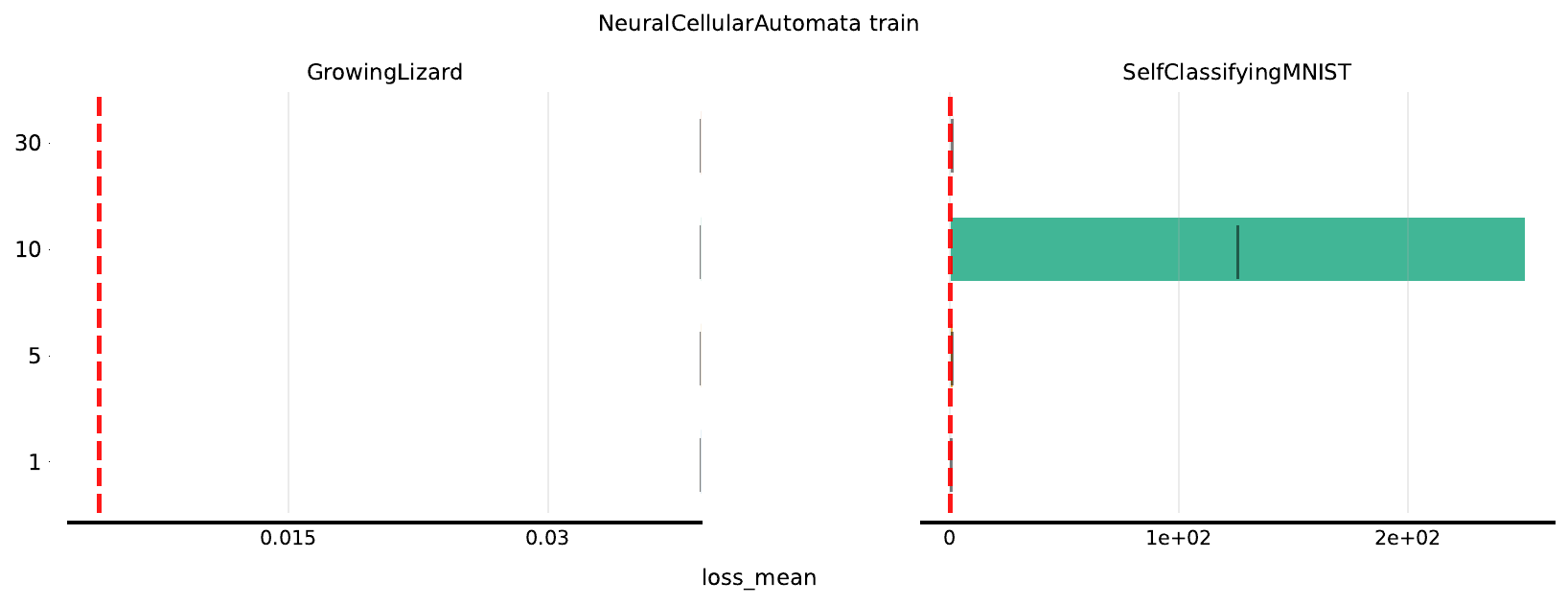}%
\\[0.5em]
\includegraphics[width=0.48\textwidth]{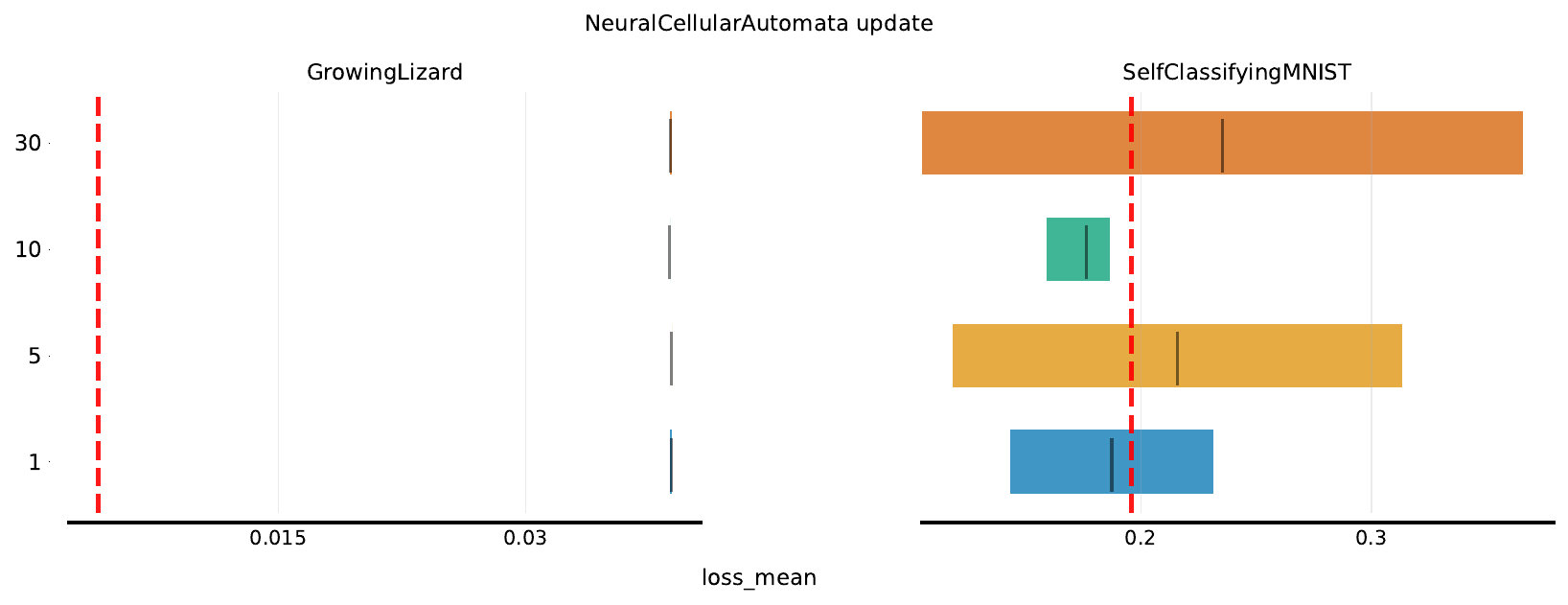}%
\hfill%
\includegraphics[width=0.48\textwidth]{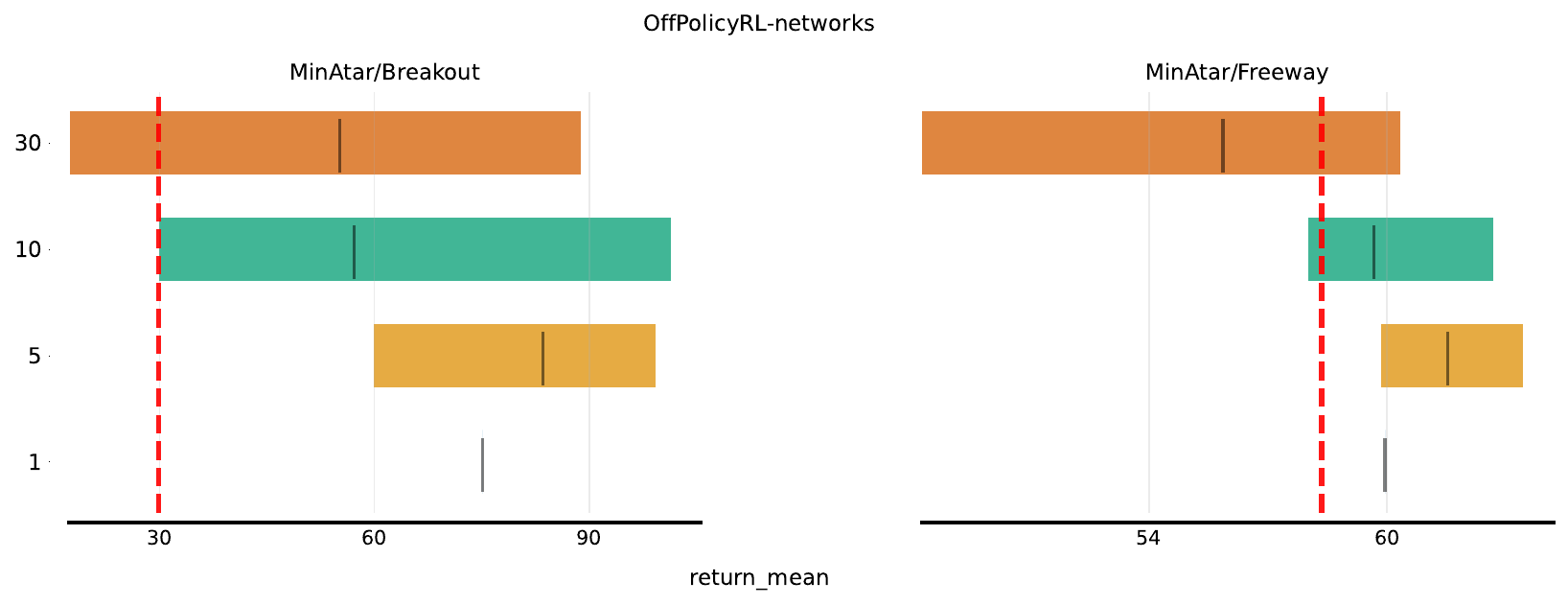}%
\\[0.5em]
\includegraphics[width=0.48\textwidth]{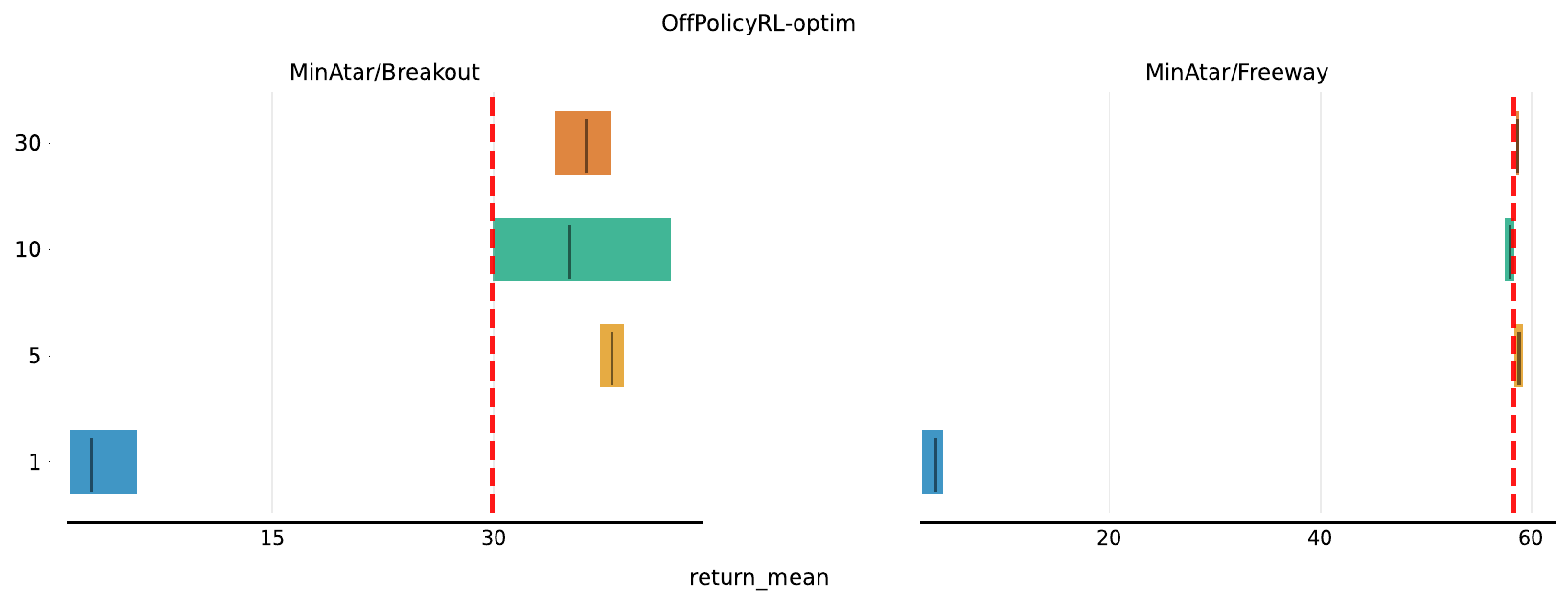}%
\hfill%
\includegraphics[width=0.48\textwidth]{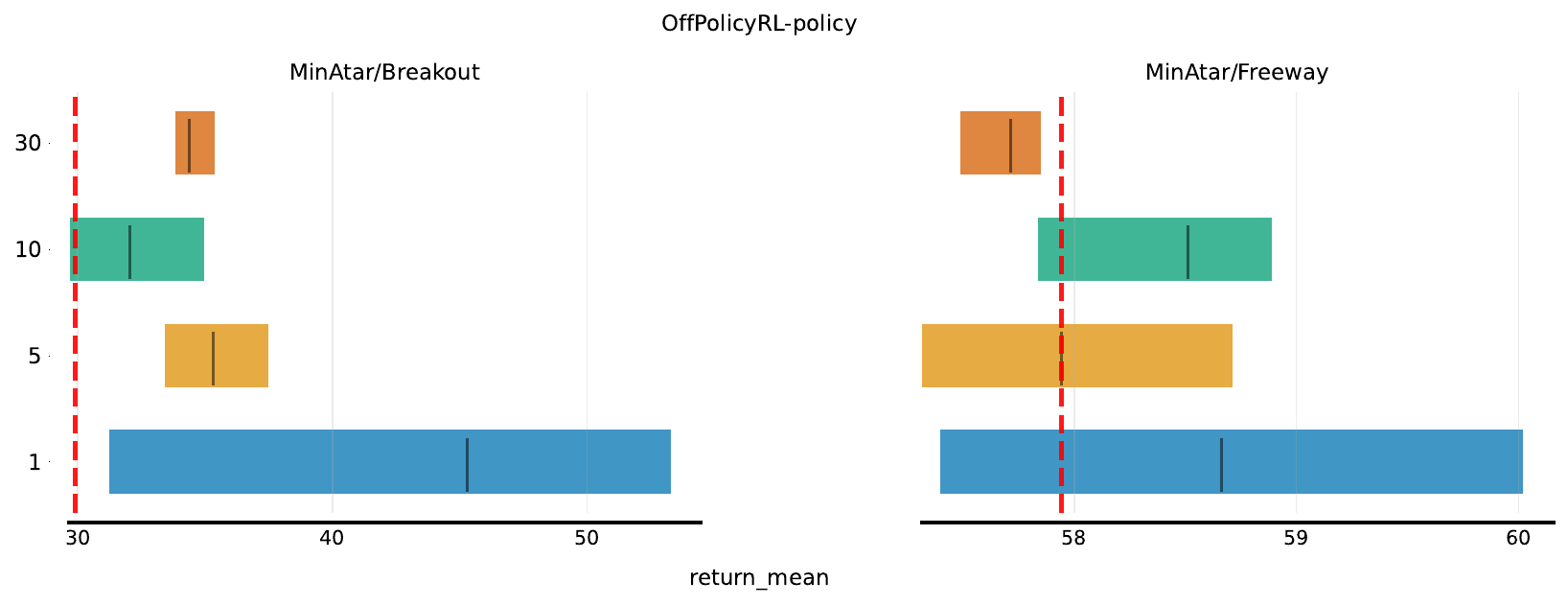}%
\\[0.5em]
\includegraphics[width=0.48\textwidth]{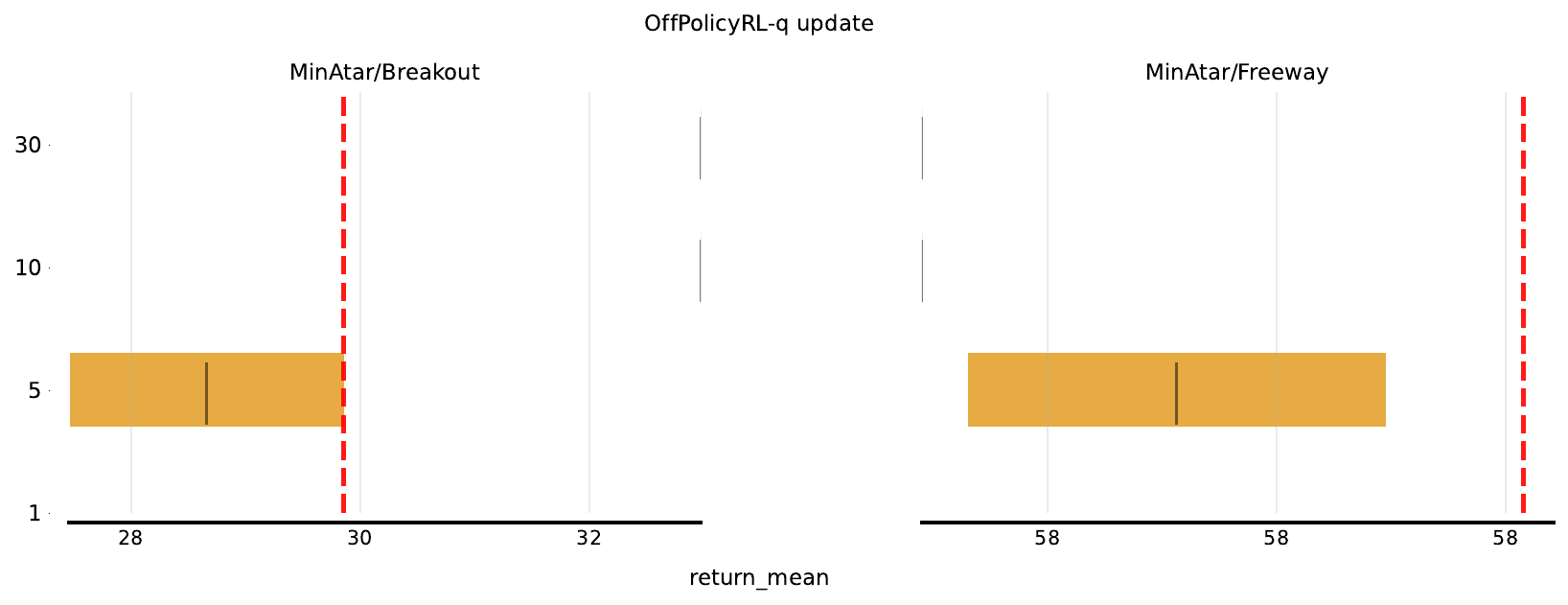}%
\hfill%
\includegraphics[width=0.48\textwidth]{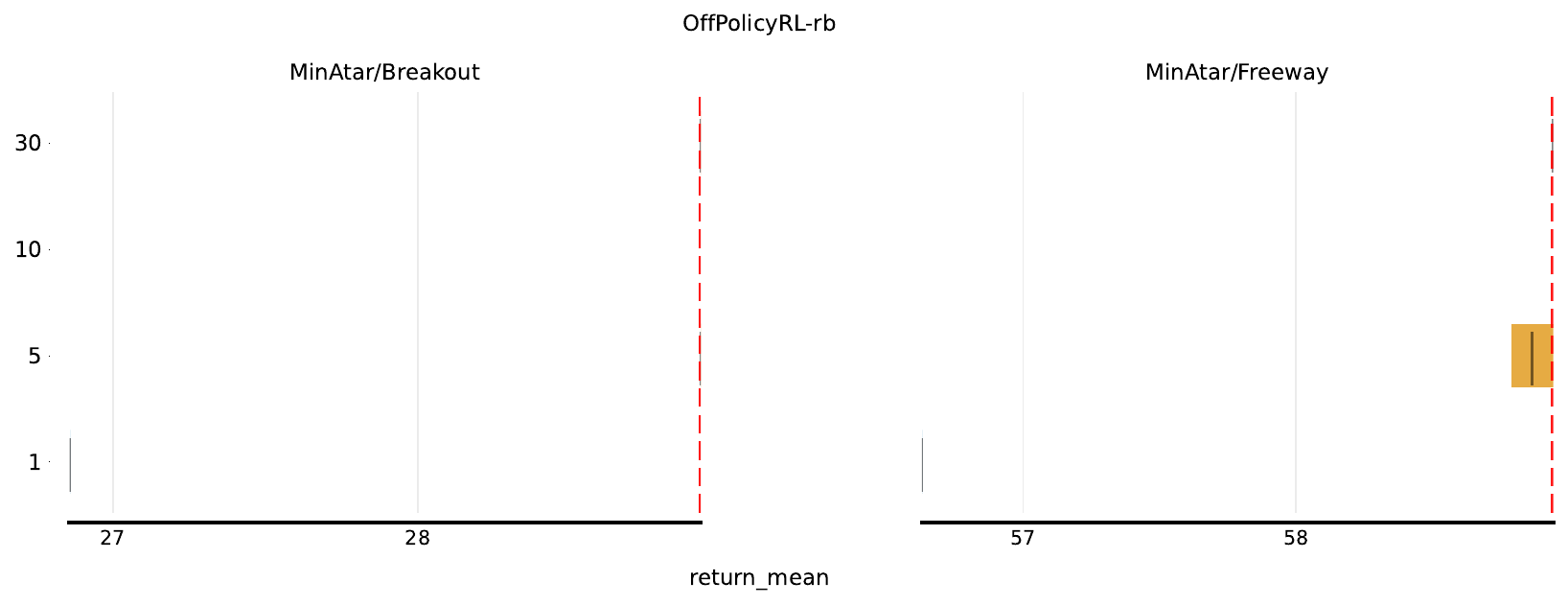}%
\caption{ADA Optimisation results on Meta-Train tasks. (Part 4/7)}
\label{fig:ADA_optimisation_id_4}
\end{figure}
\clearpage

\begin{figure}[htbp]
\centering
\setlength{\lineskip}{0pt}
\includegraphics[width=0.48\textwidth]{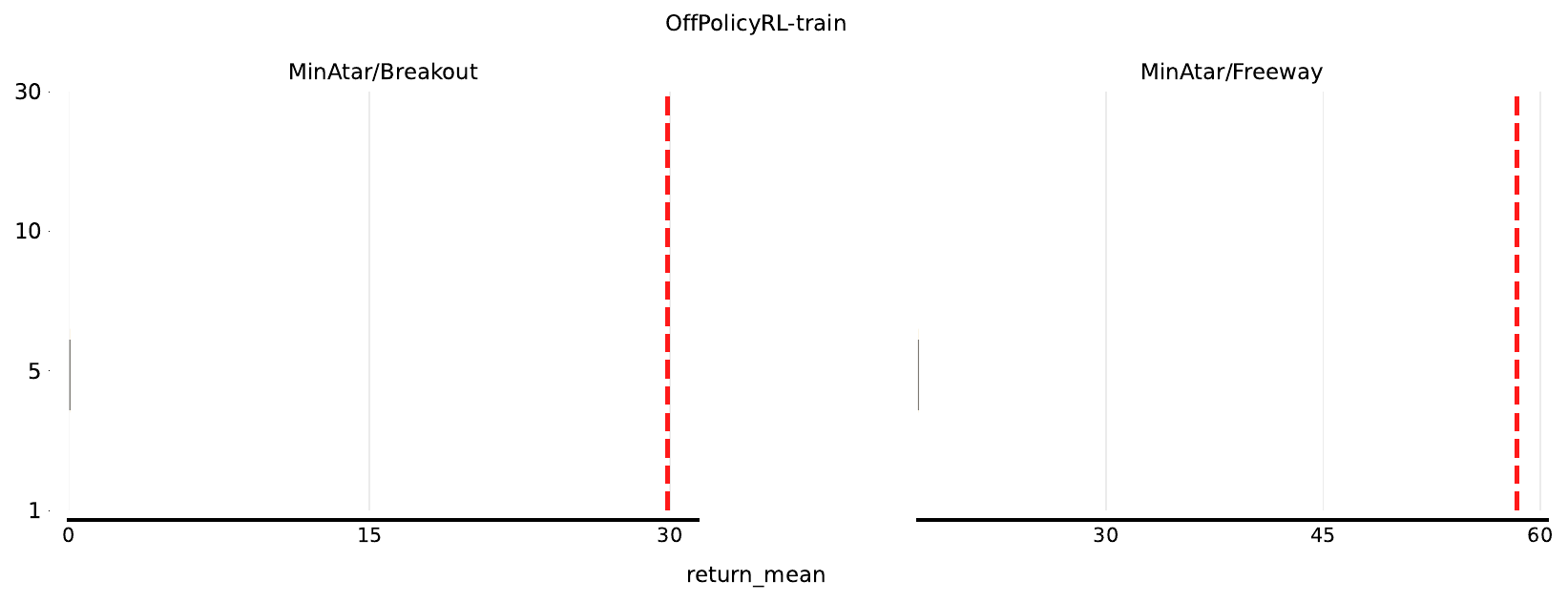}%
\hfill%
\includegraphics[width=0.48\textwidth]{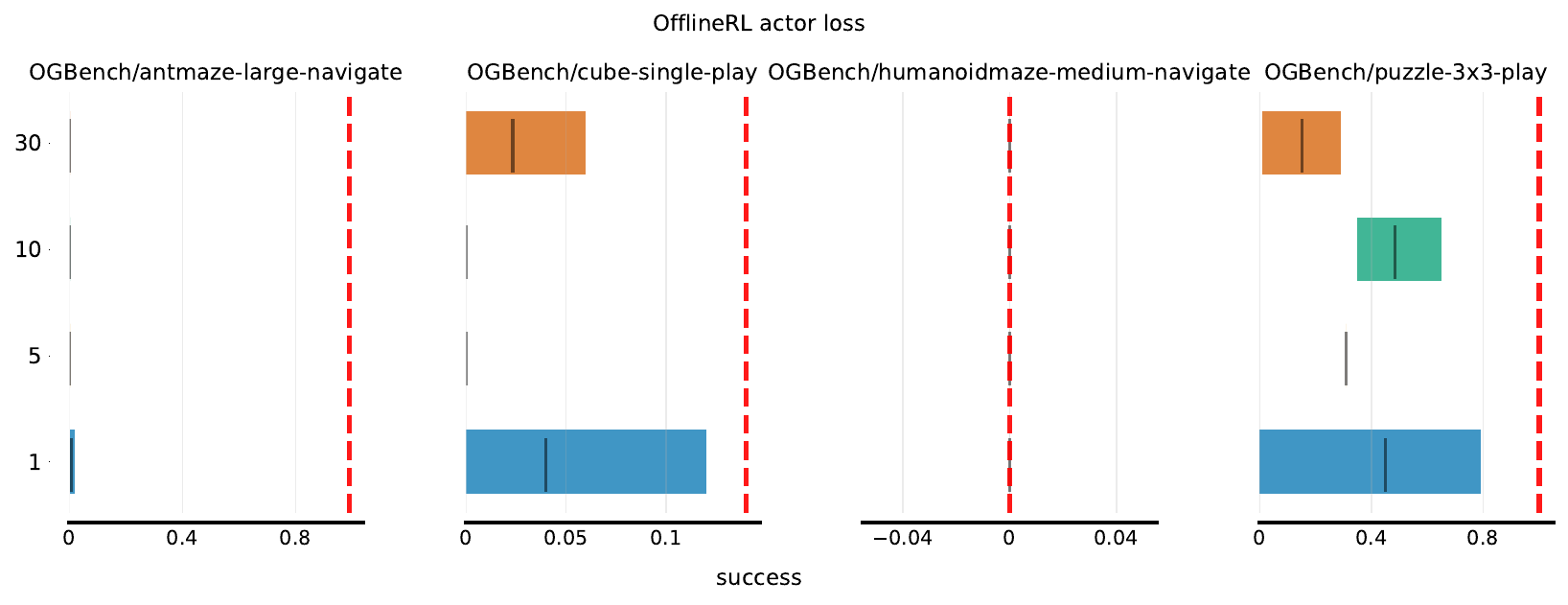}%
\\[0.5em]
\includegraphics[width=0.48\textwidth]{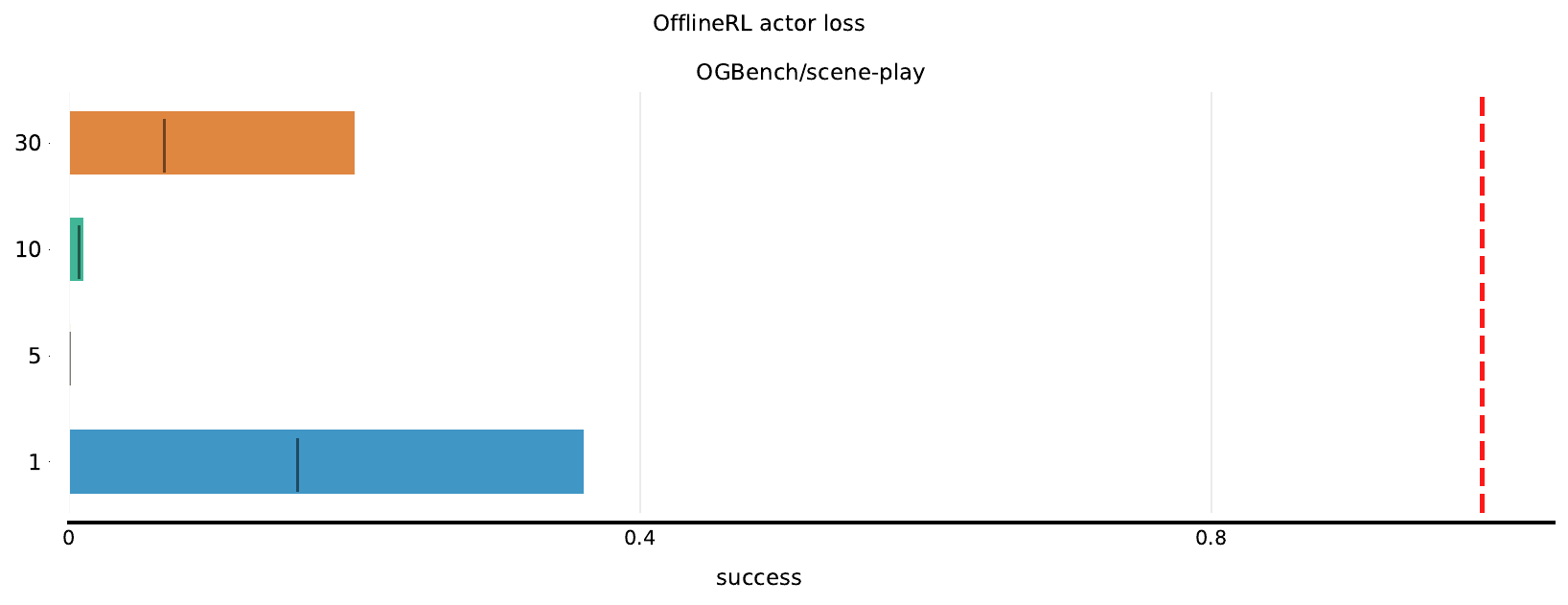}%
\hfill%
\includegraphics[width=0.48\textwidth]{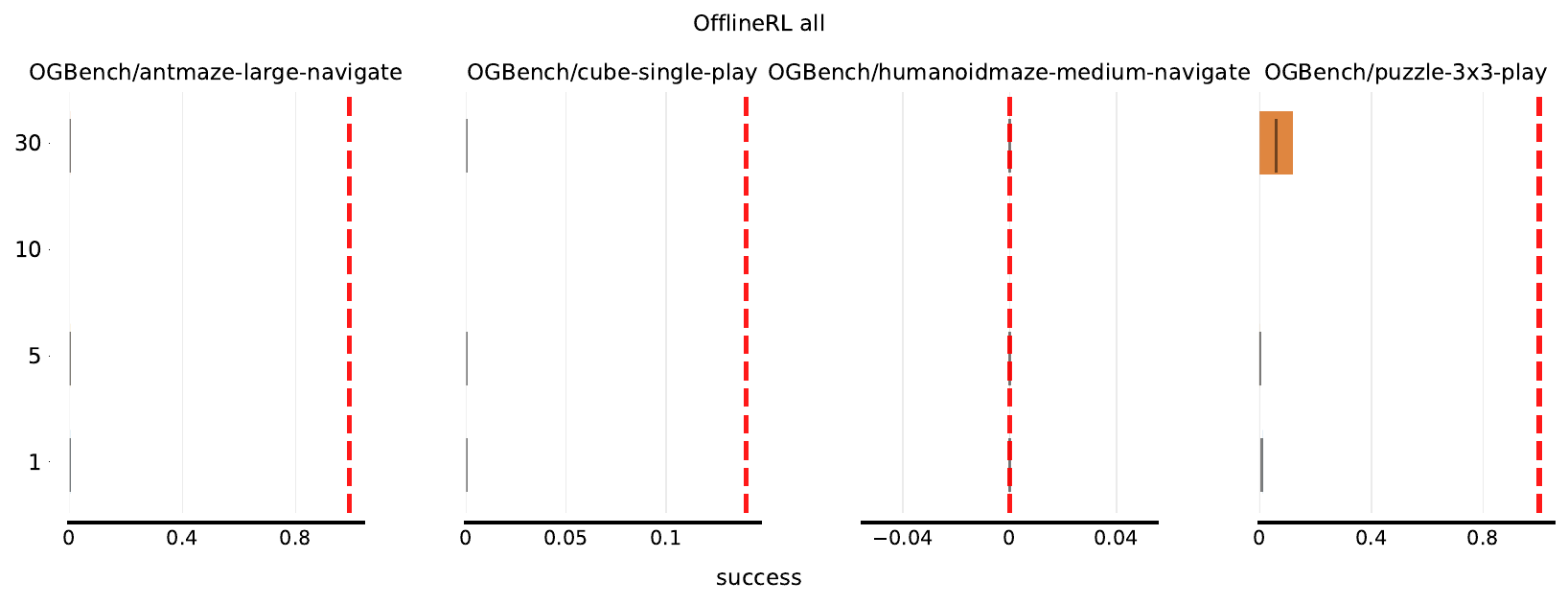}%
\\[0.5em]
\includegraphics[width=0.48\textwidth]{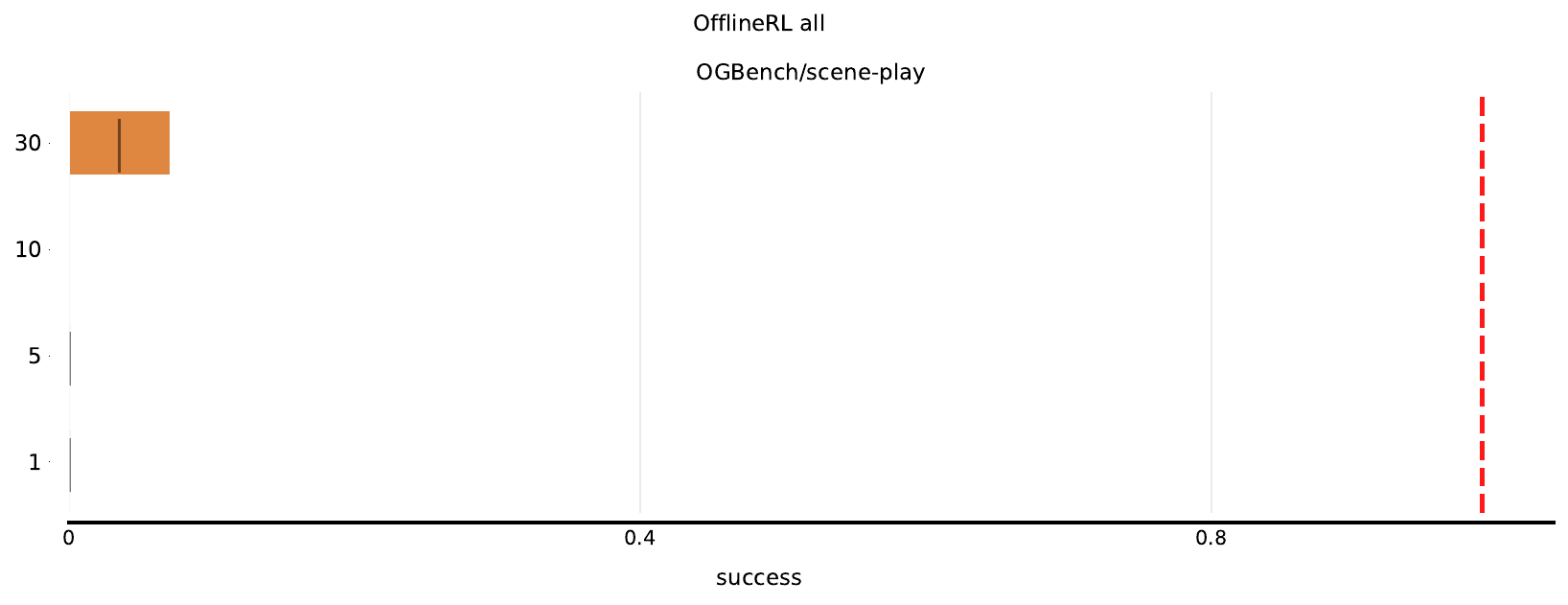}%
\hfill%
\includegraphics[width=0.48\textwidth]{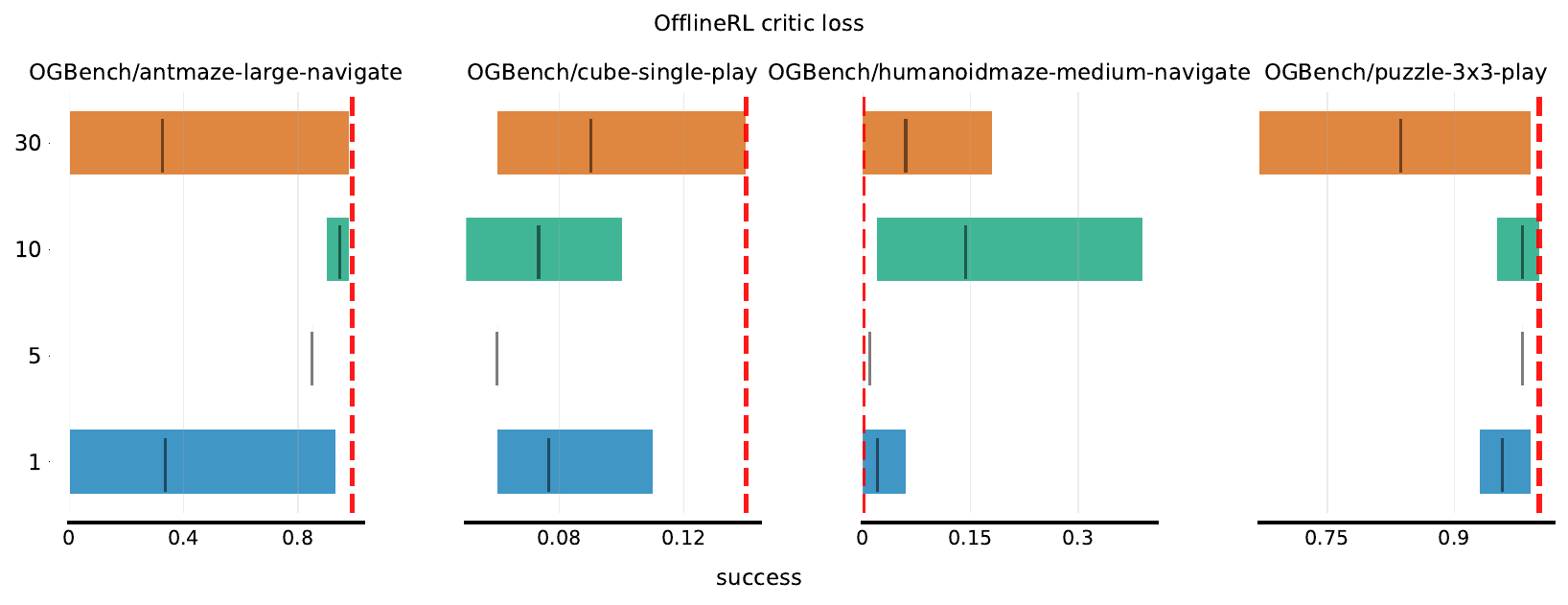}%
\\[0.5em]
\includegraphics[width=0.48\textwidth]{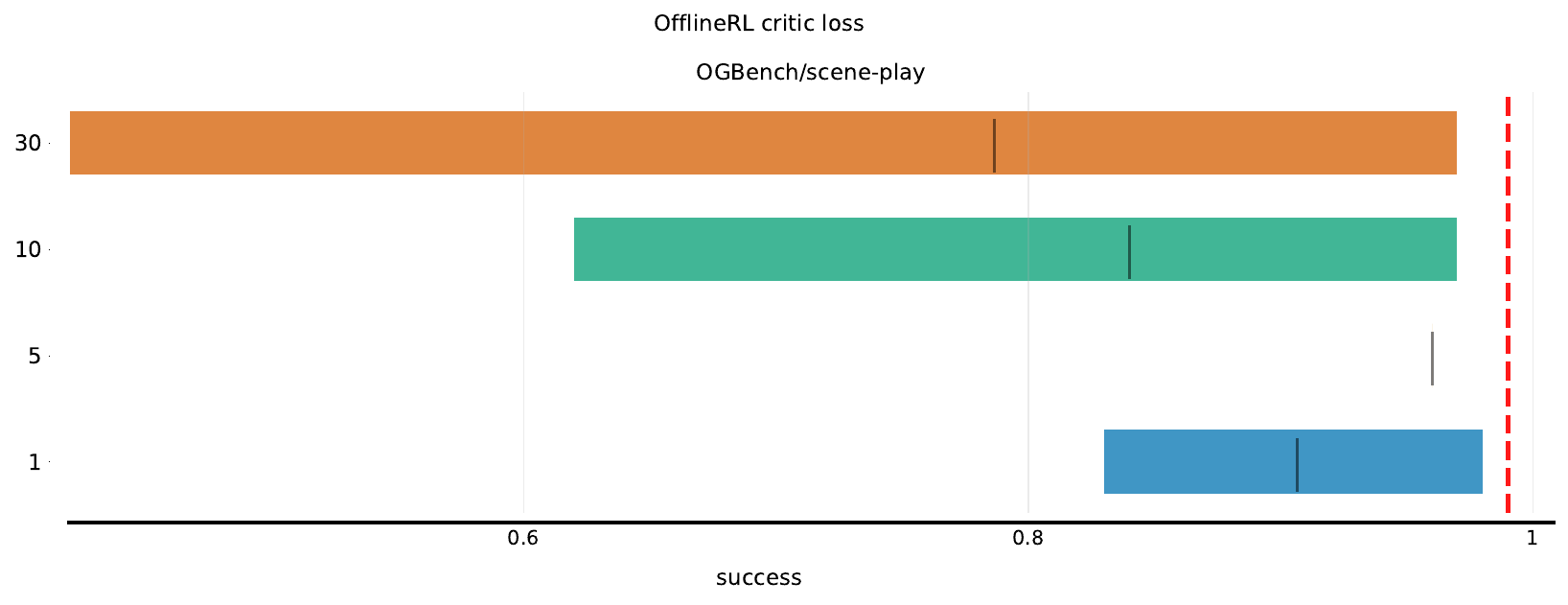}%
\hfill%
\includegraphics[width=0.48\textwidth]{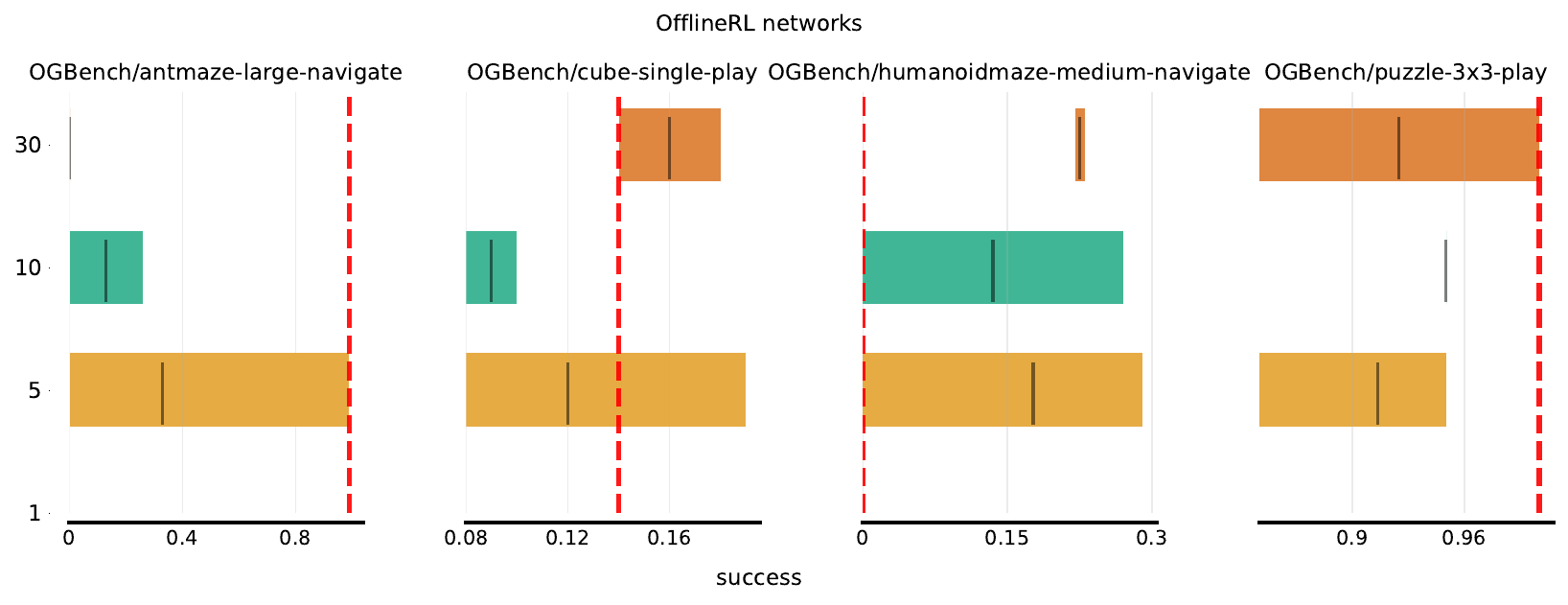}%
\\[0.5em]
\includegraphics[width=0.48\textwidth]{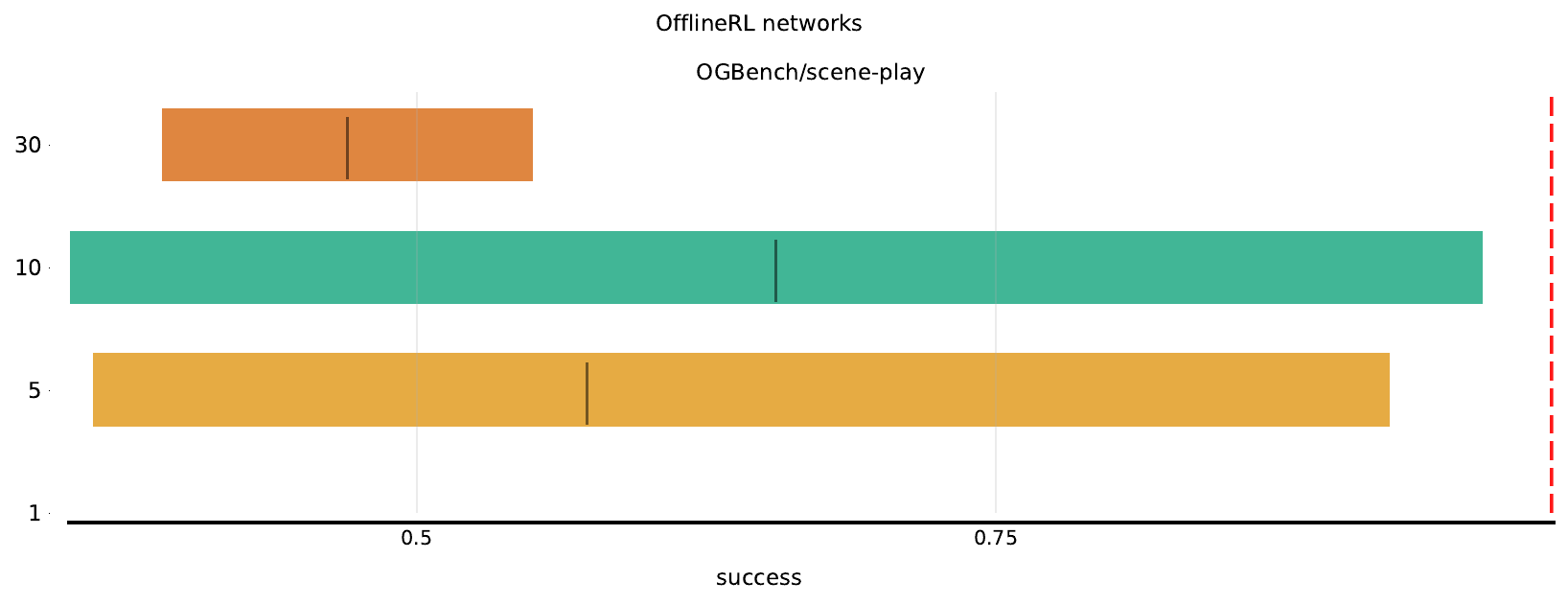}%
\hfill%
\includegraphics[width=0.48\textwidth]{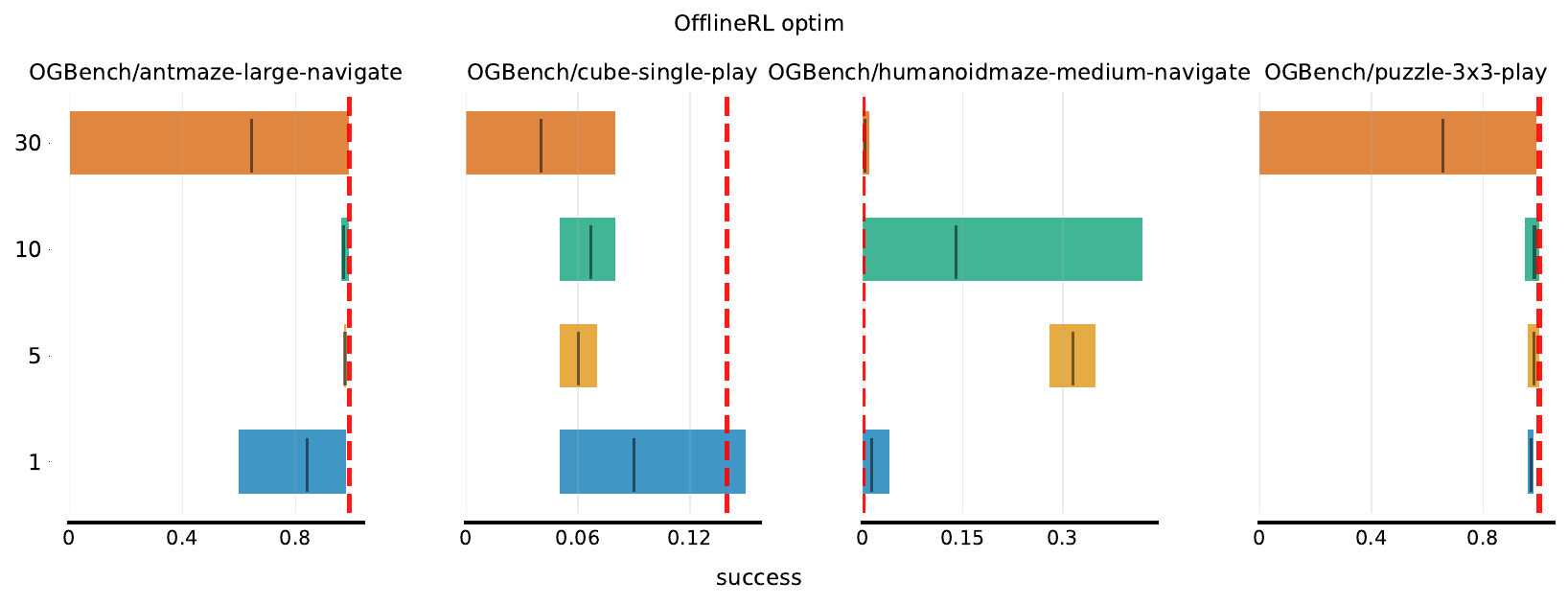}%
\\[0.5em]
\includegraphics[width=0.48\textwidth]{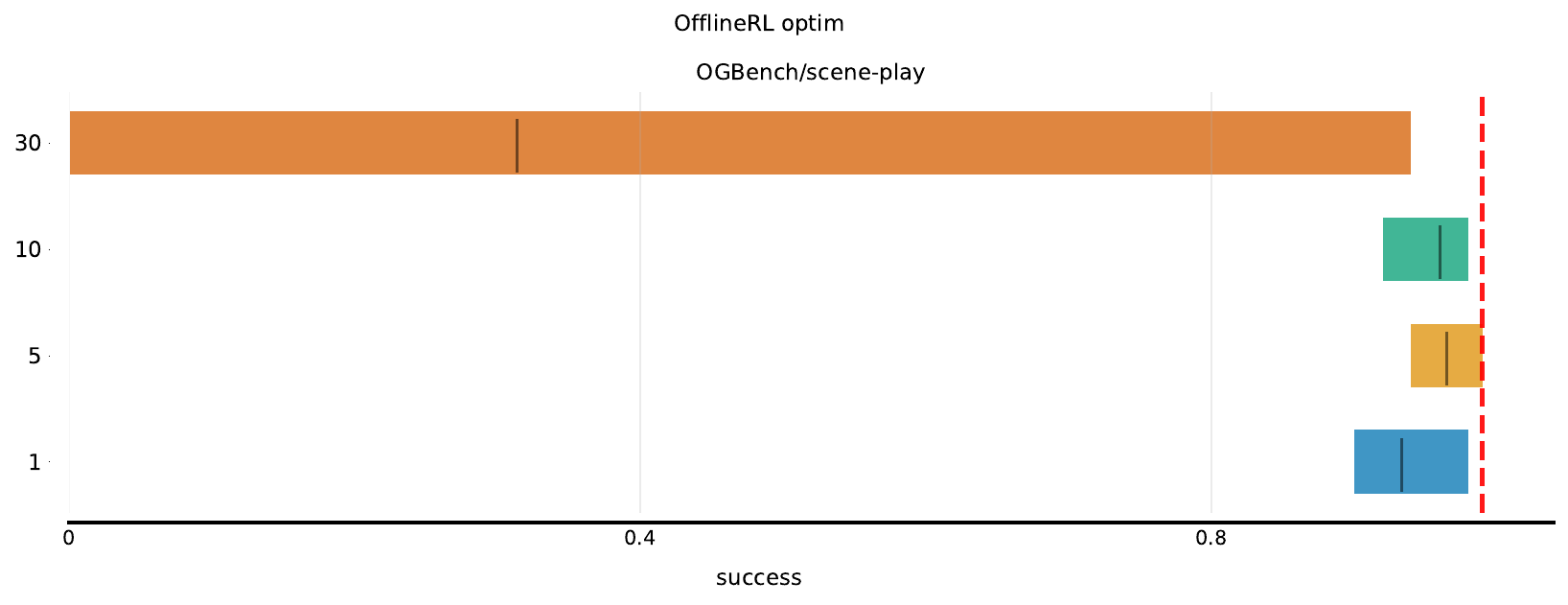}%
\hfill%
\includegraphics[width=0.48\textwidth]{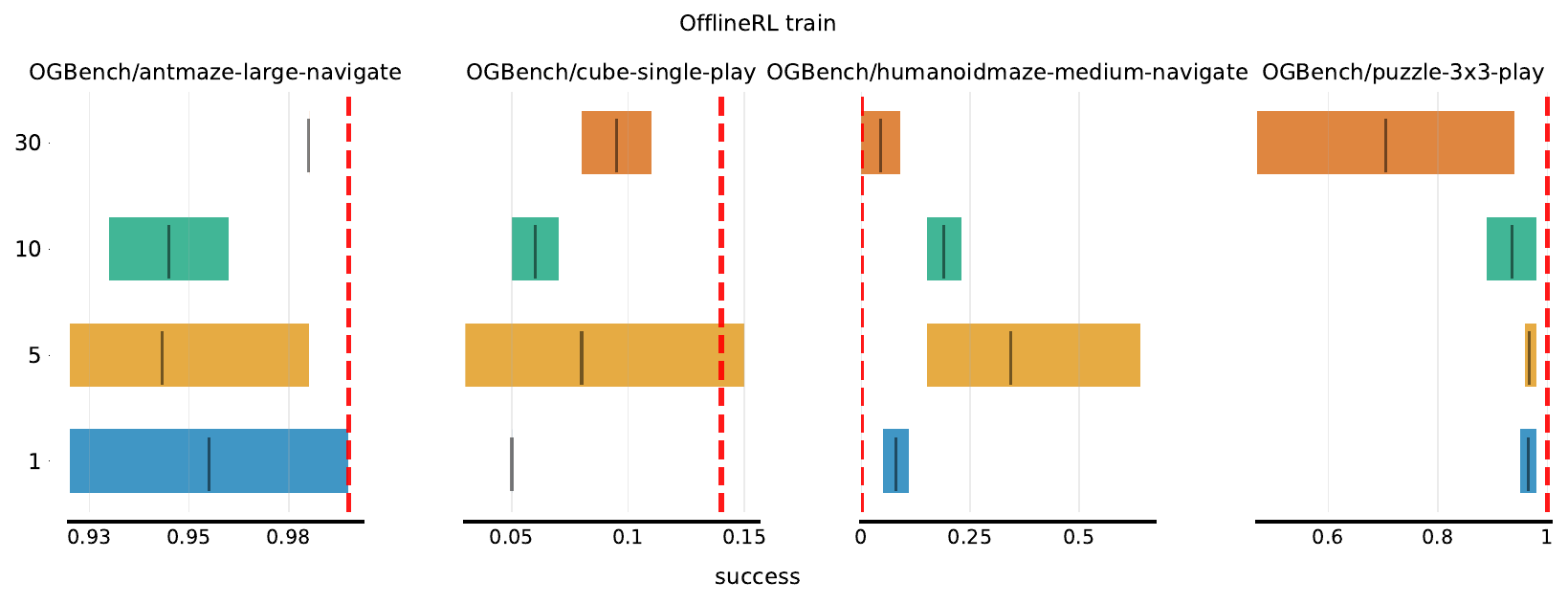}%
\caption{ADA Optimisation results on Meta-Train tasks. (Part 5/7)}
\label{fig:ADA_optimisation_id_5}
\end{figure}
\clearpage

\begin{figure}[htbp]
\centering
\setlength{\lineskip}{0pt}
\includegraphics[width=0.48\textwidth]{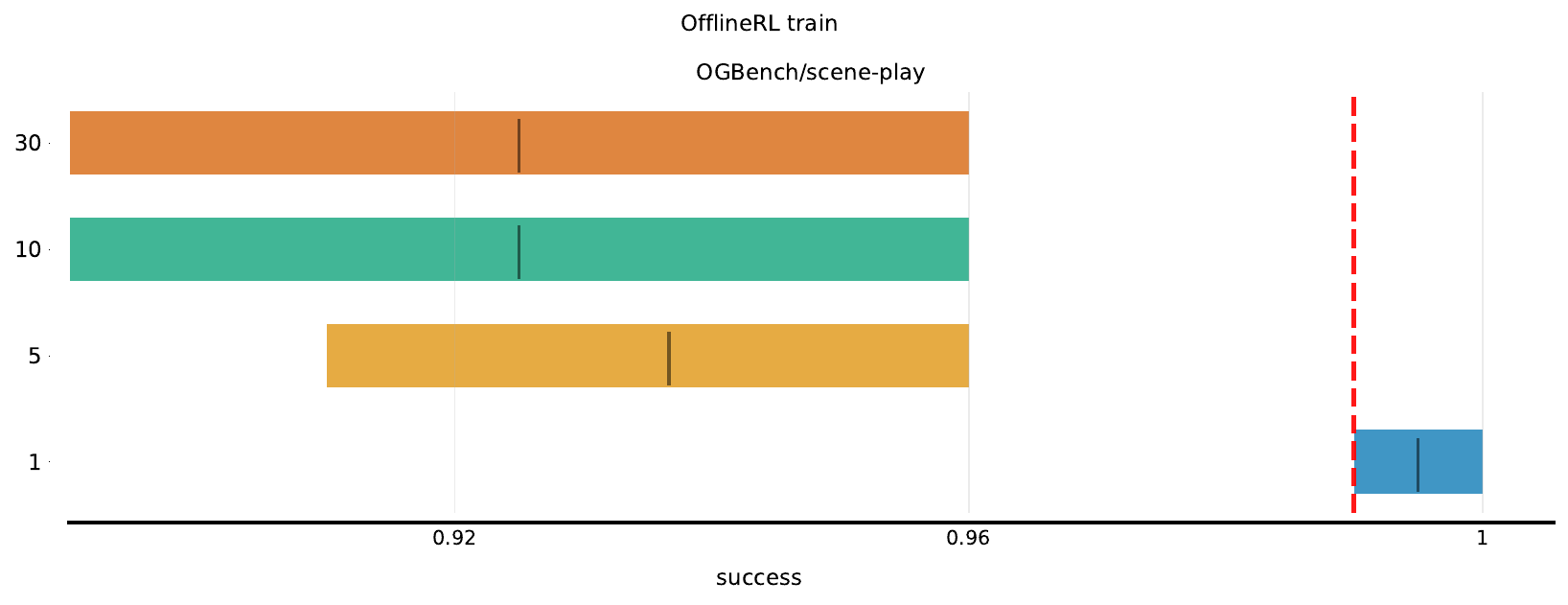}%
\hfill%
\includegraphics[width=0.48\textwidth]{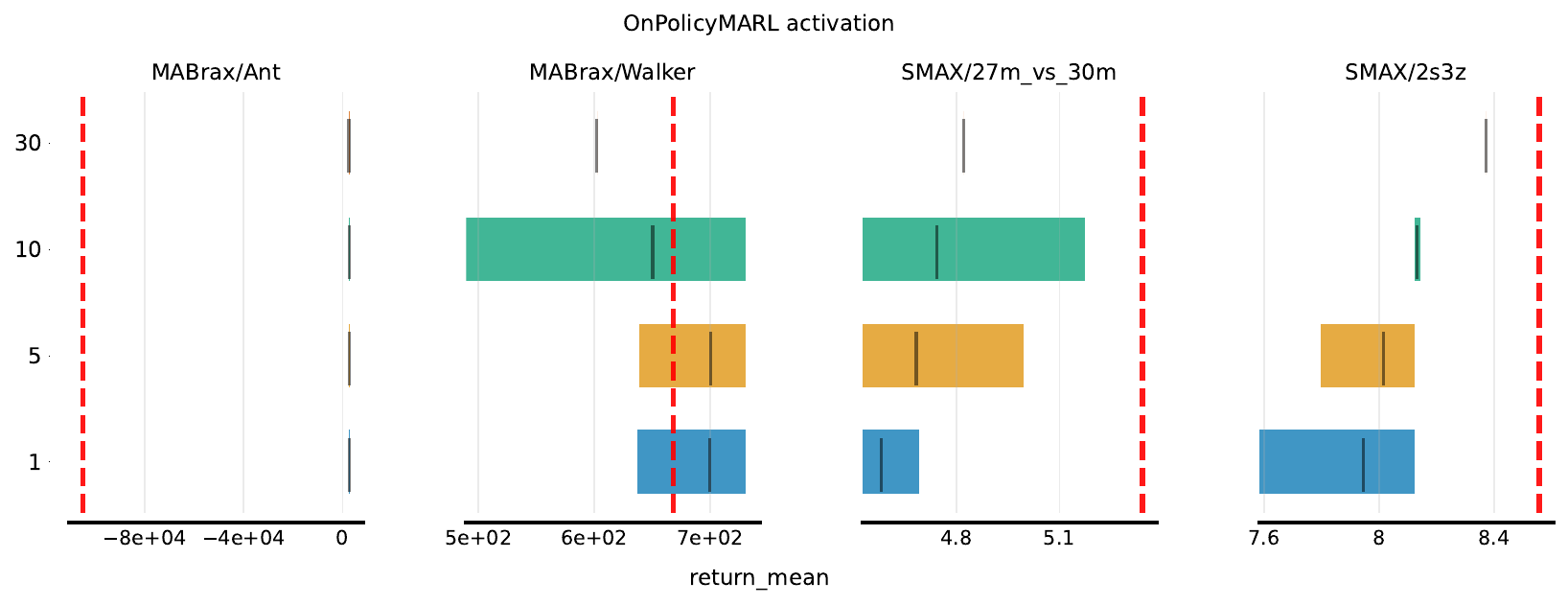}%
\\[0.5em]
\includegraphics[width=0.48\textwidth]{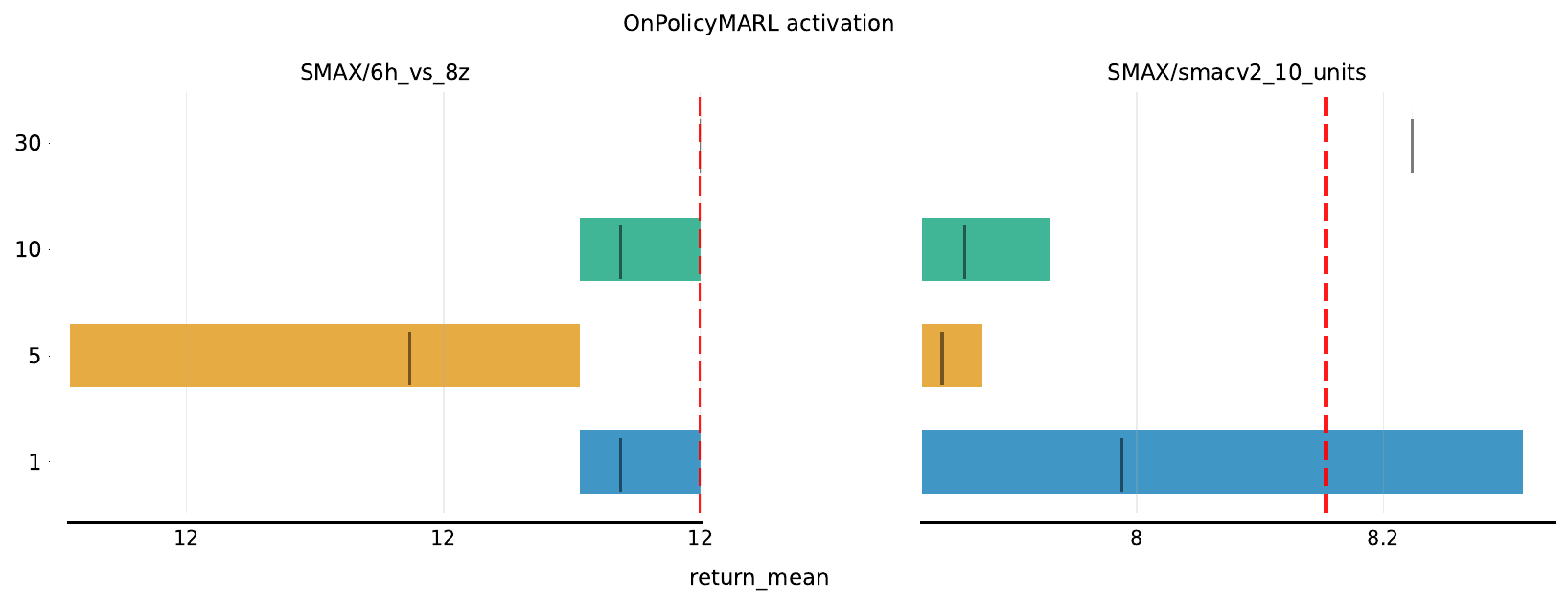}%
\hfill%
\includegraphics[width=0.48\textwidth]{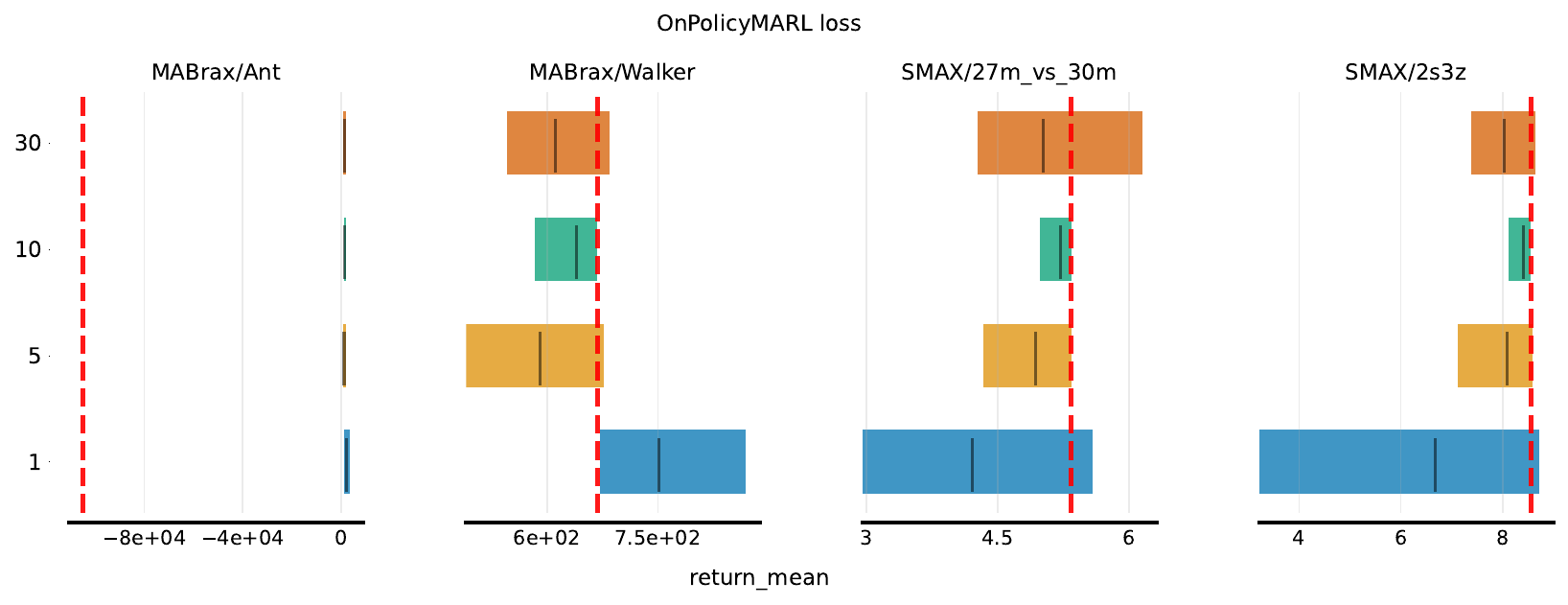}%
\\[0.5em]
\includegraphics[width=0.48\textwidth]{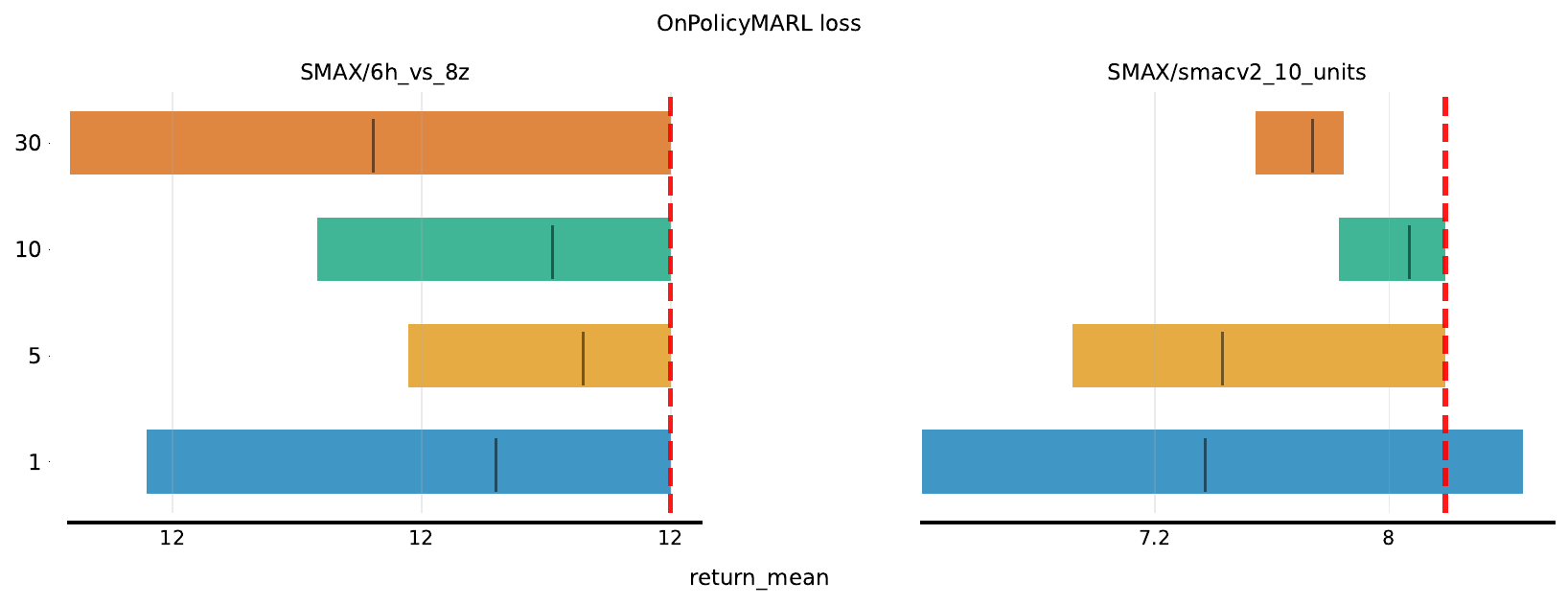}%
\hfill%
\includegraphics[width=0.48\textwidth]{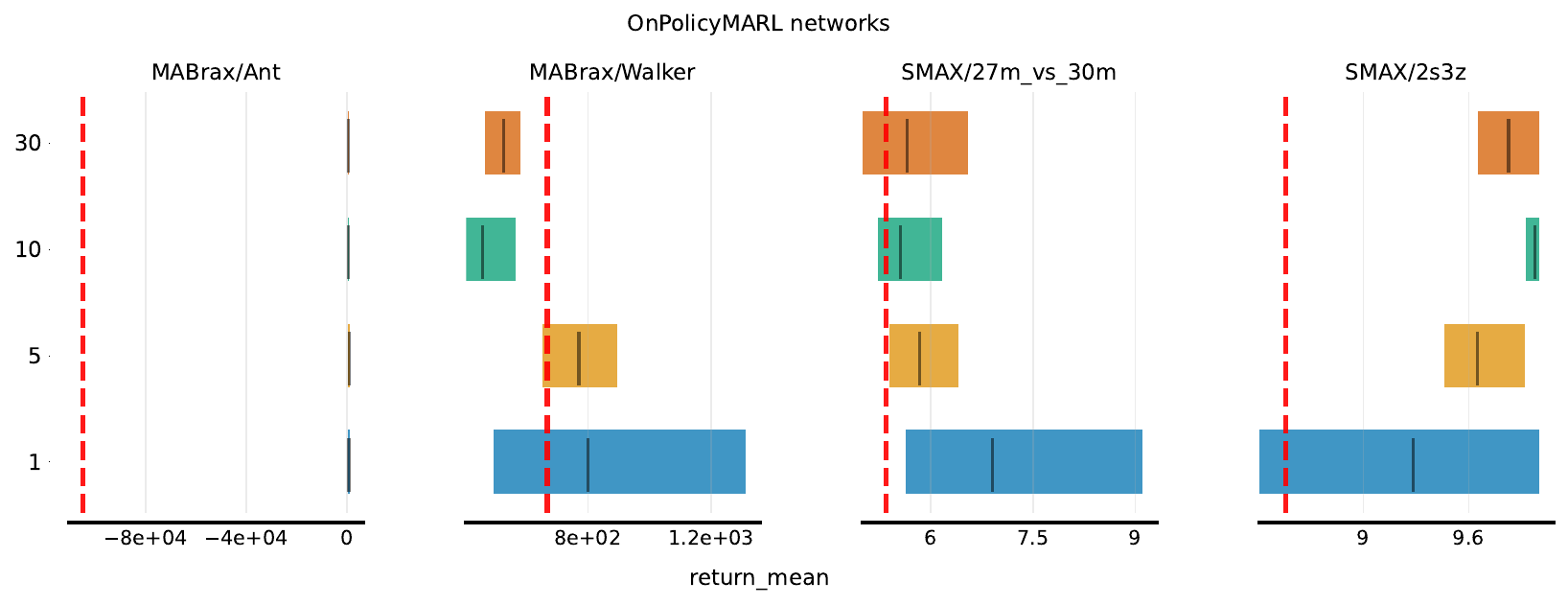}%
\\[0.5em]
\includegraphics[width=0.48\textwidth]{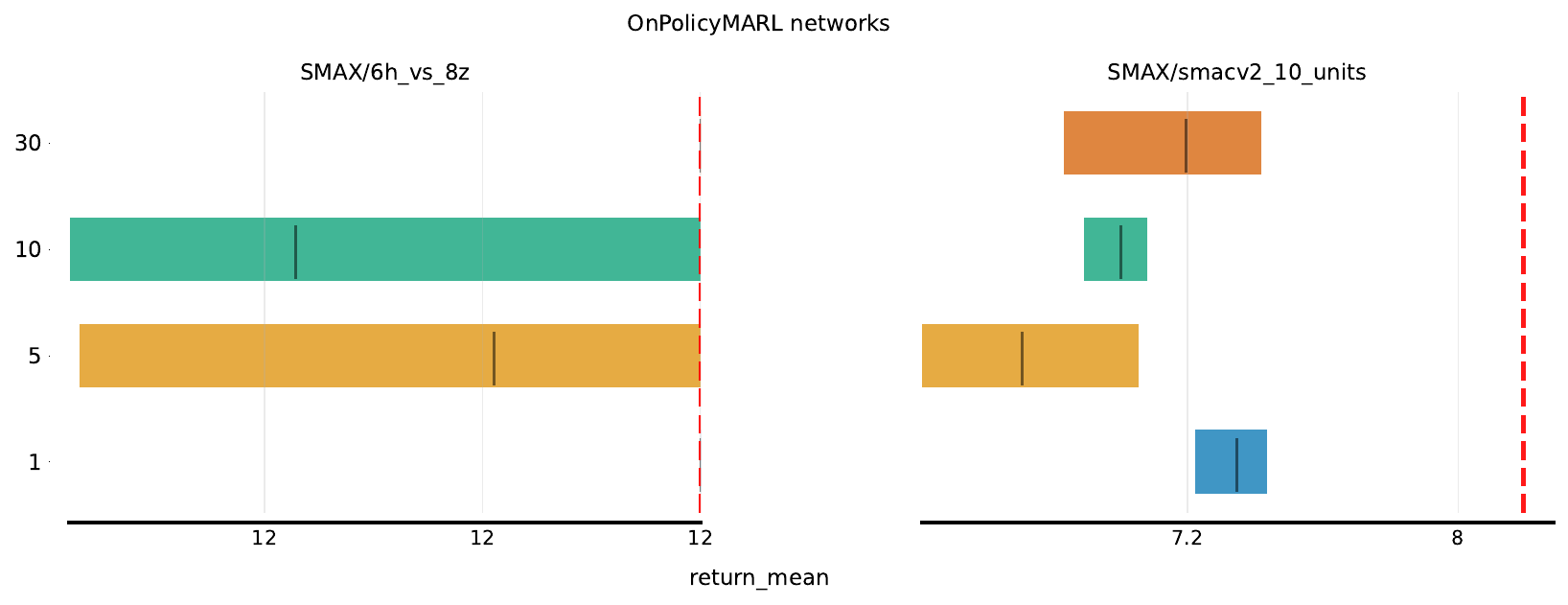}%
\hfill%
\includegraphics[width=0.48\textwidth]{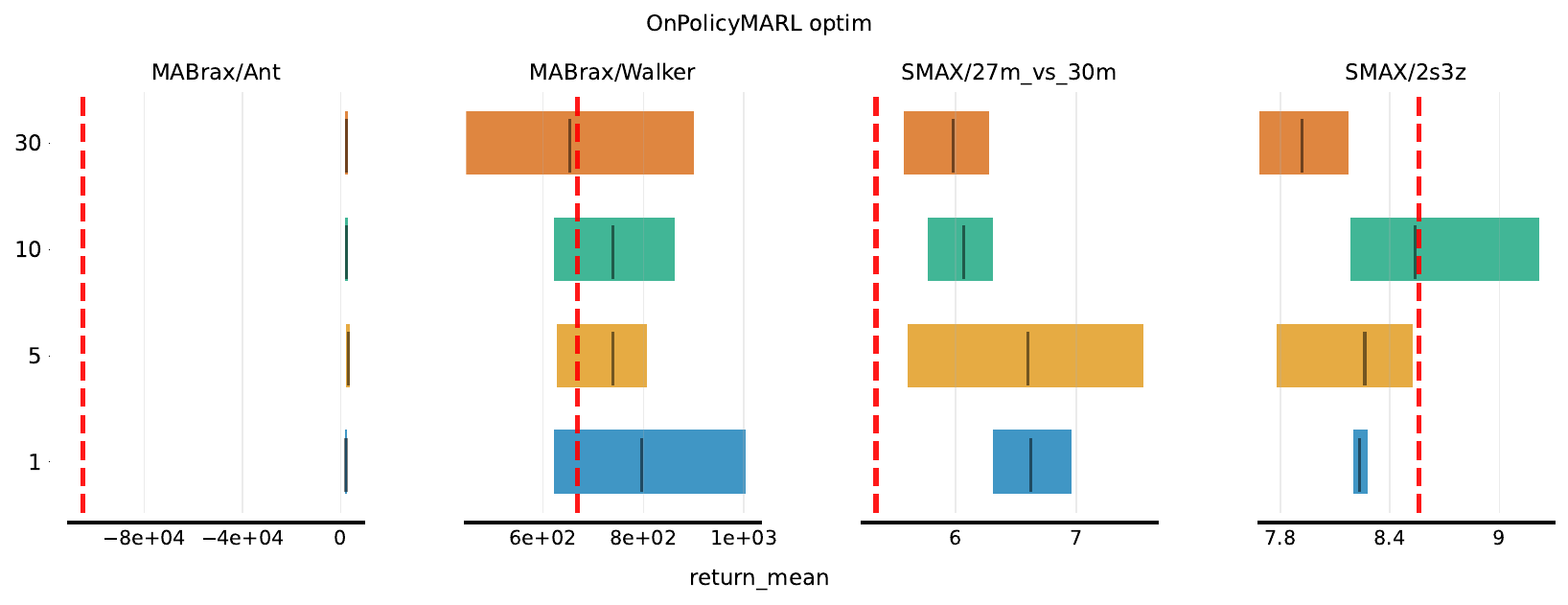}%
\\[0.5em]
\includegraphics[width=0.48\textwidth]{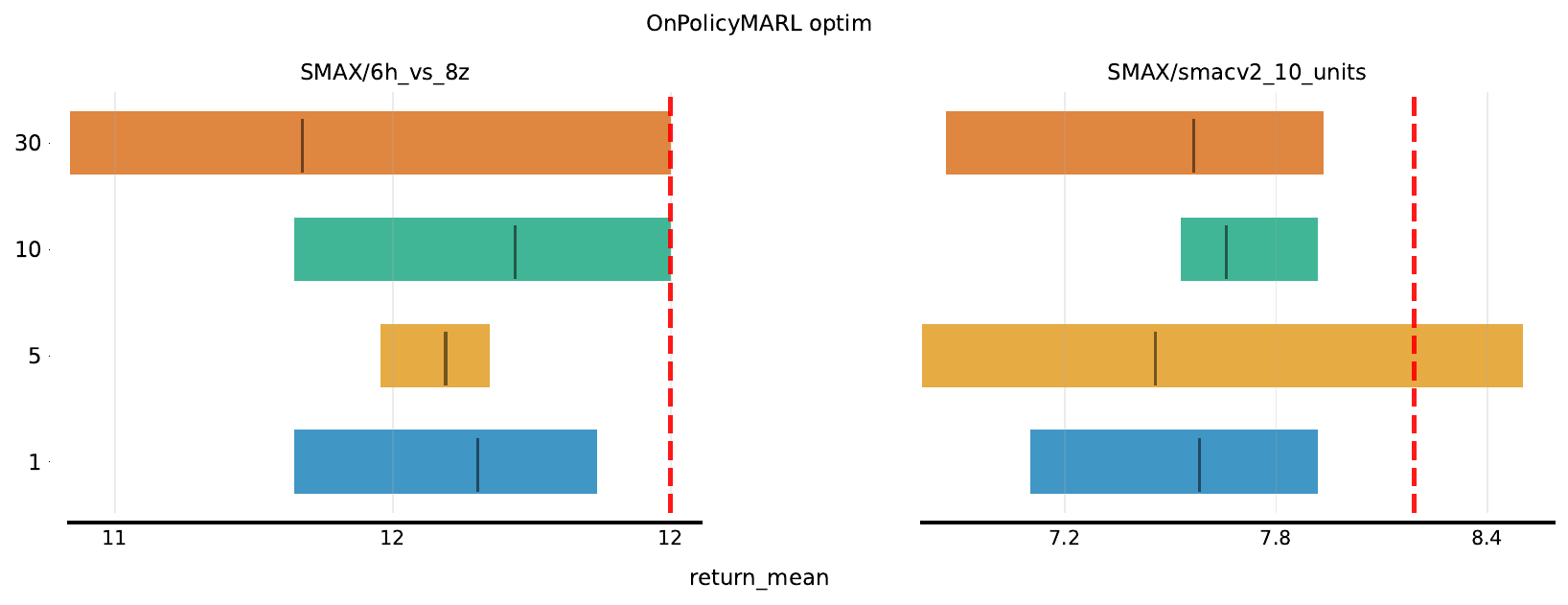}%
\hfill%
\includegraphics[width=0.48\textwidth]{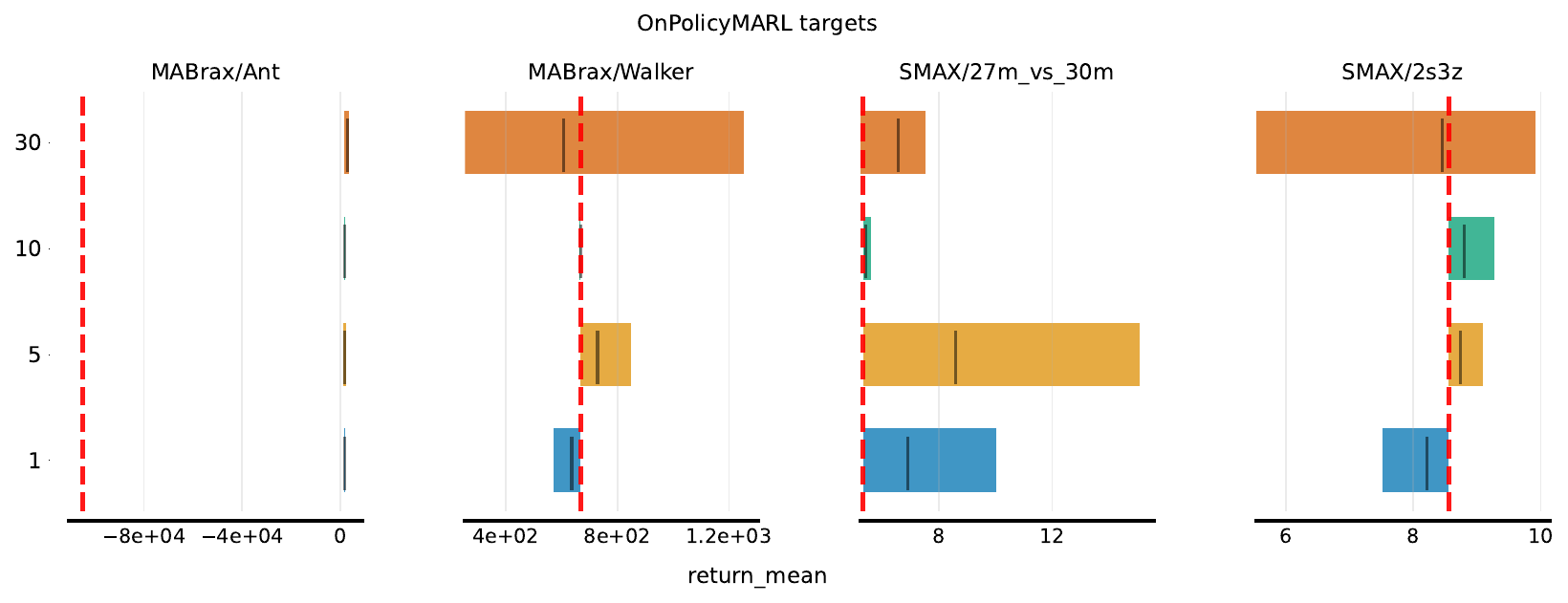}%
\\[0.5em]
\includegraphics[width=0.48\textwidth]{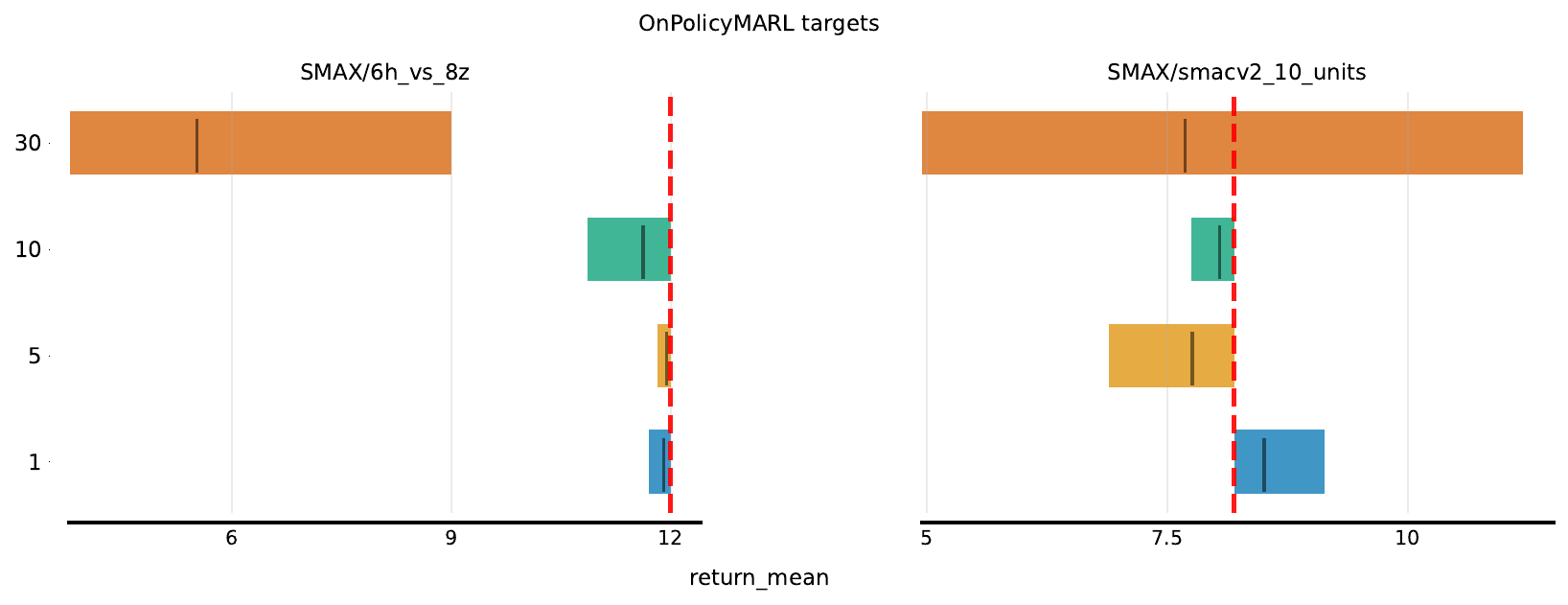}%
\hfill%
\includegraphics[width=0.48\textwidth]{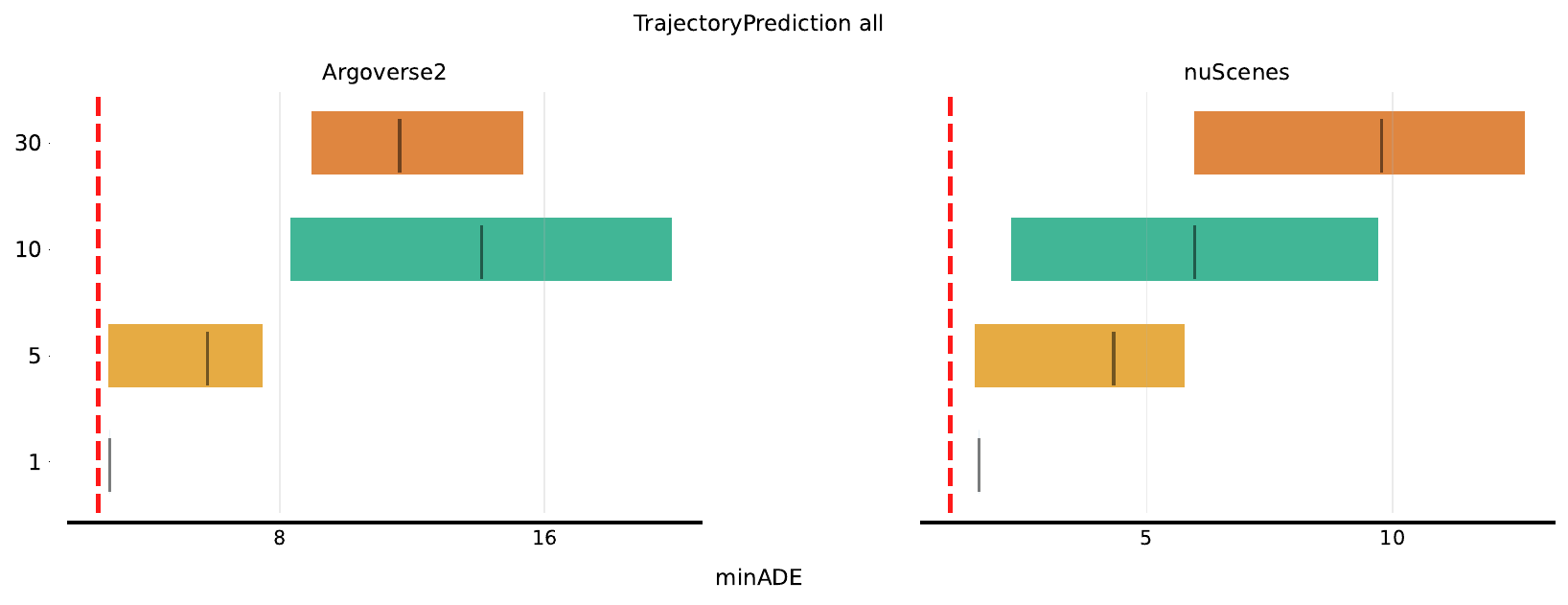}%
\caption{ADA Optimisation results on Meta-Train tasks. (Part 6/7)}
\label{fig:ADA_optimisation_id_6}
\end{figure}
\clearpage

\begin{figure}[htbp]
\centering
\setlength{\lineskip}{0pt}
\includegraphics[width=0.48\textwidth]{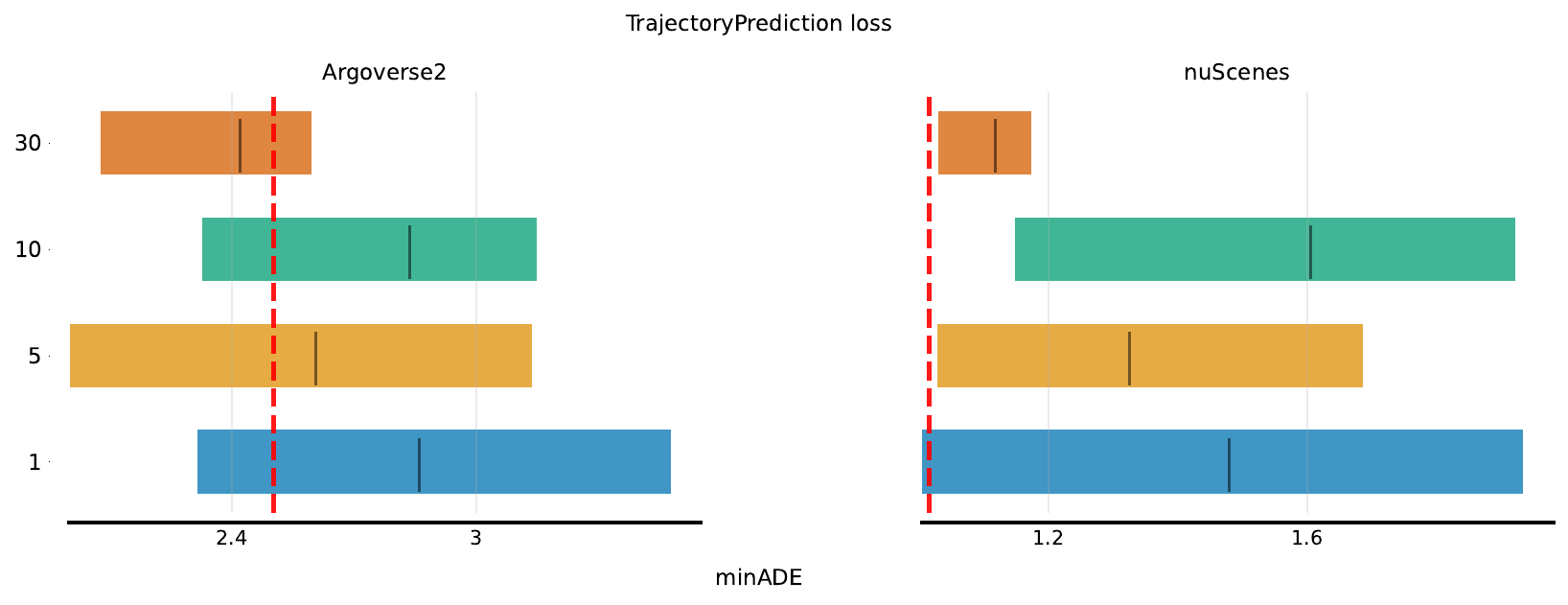}%
\hfill%
\includegraphics[width=0.48\textwidth]{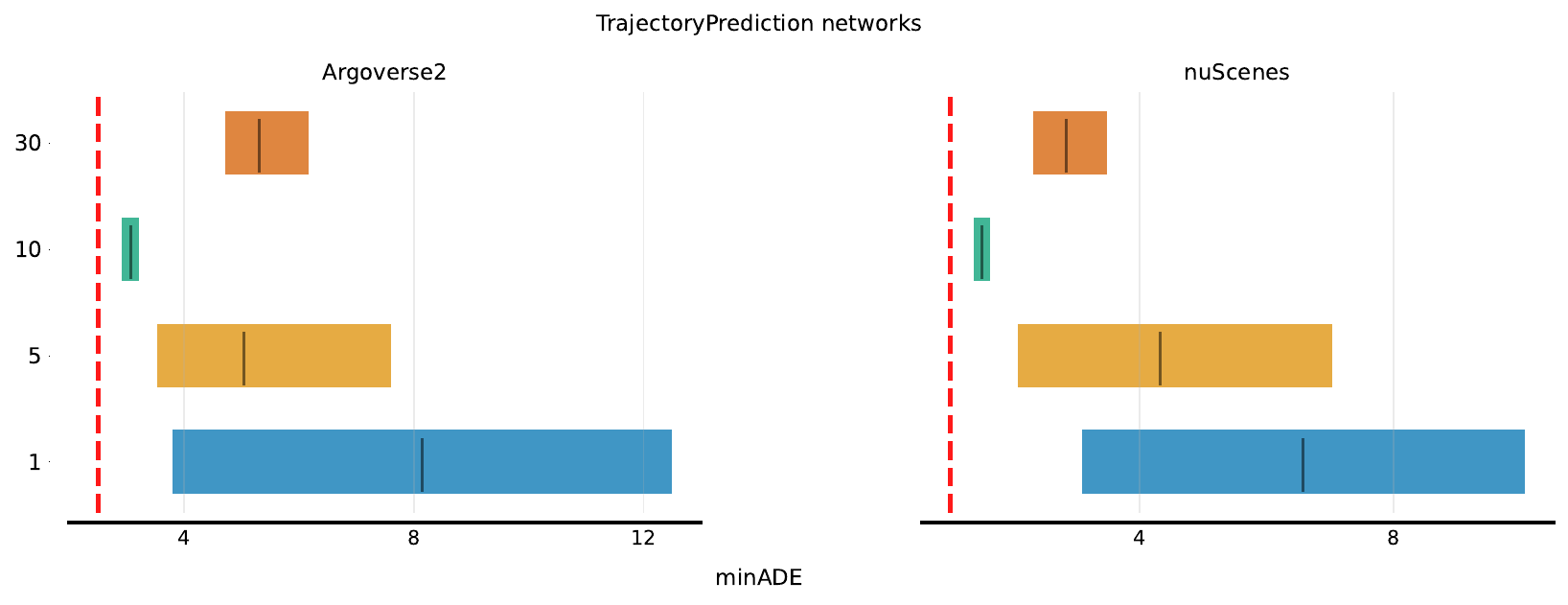}%
\\[0.5em]
\includegraphics[width=0.48\textwidth]{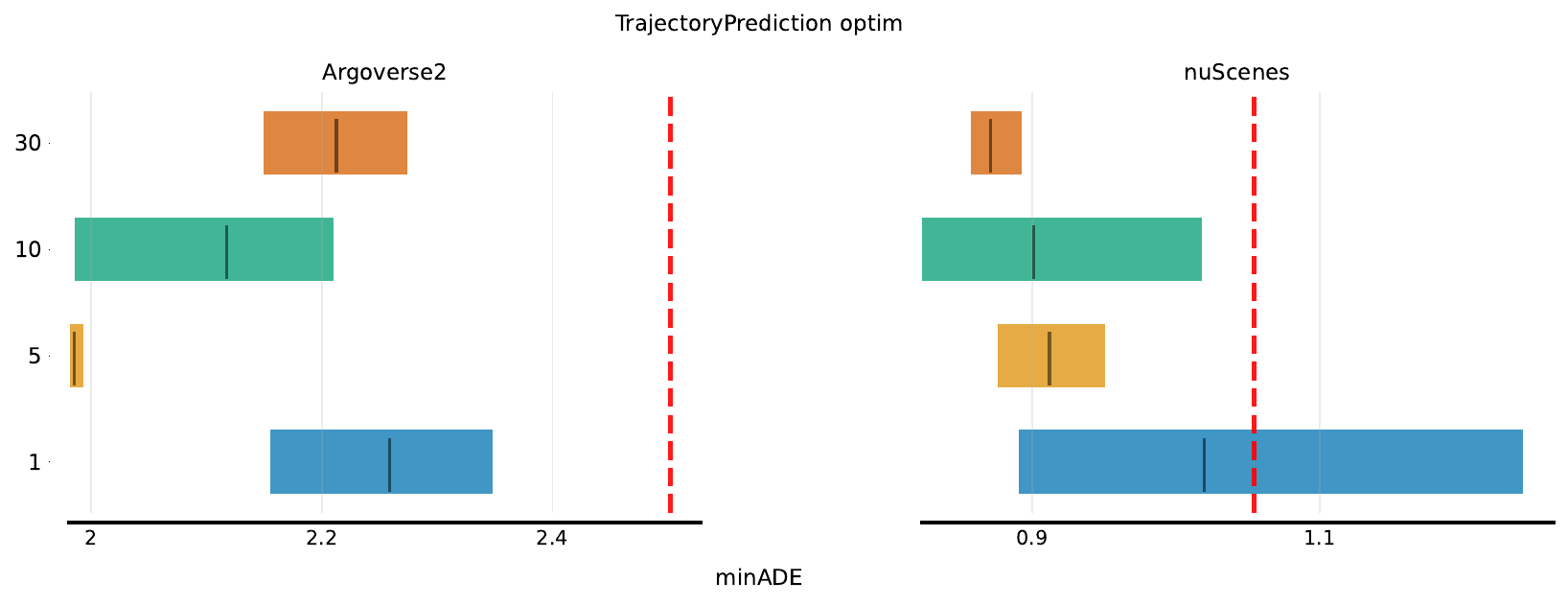}%
\hfill%
\includegraphics[width=0.48\textwidth]{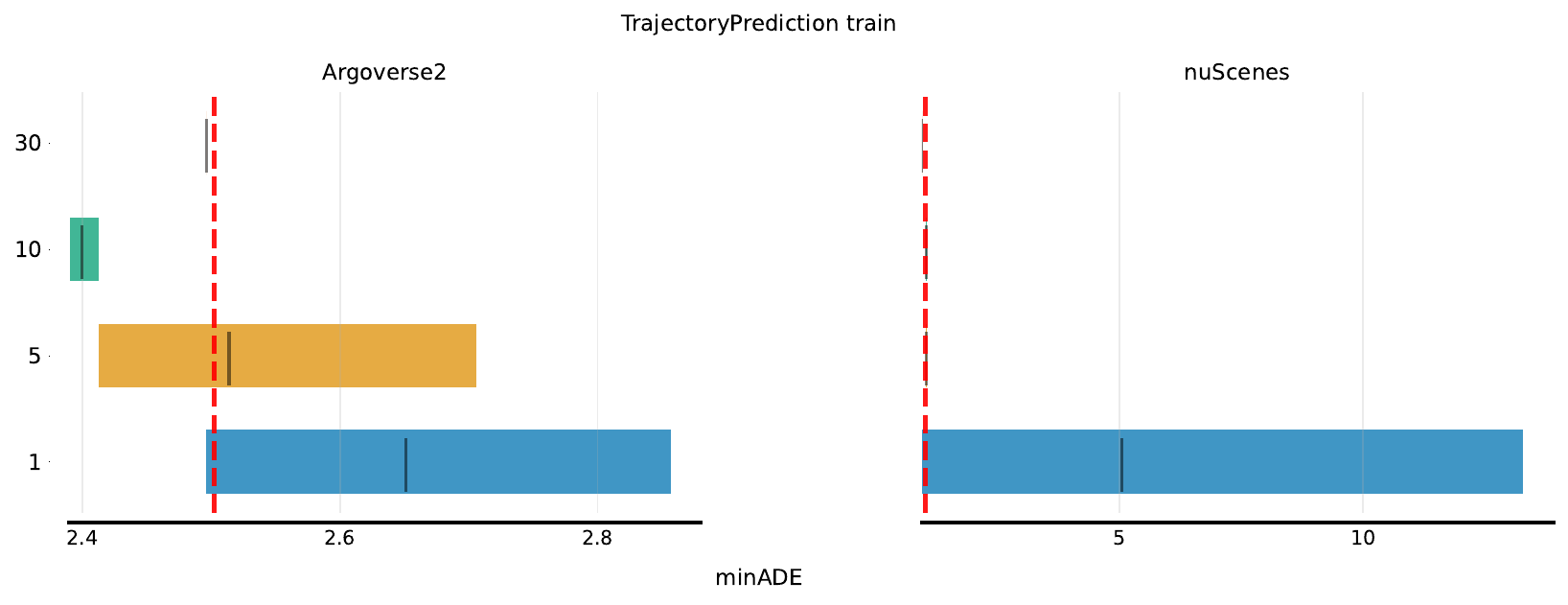}%
\\[0.5em]
\includegraphics[width=0.48\textwidth]{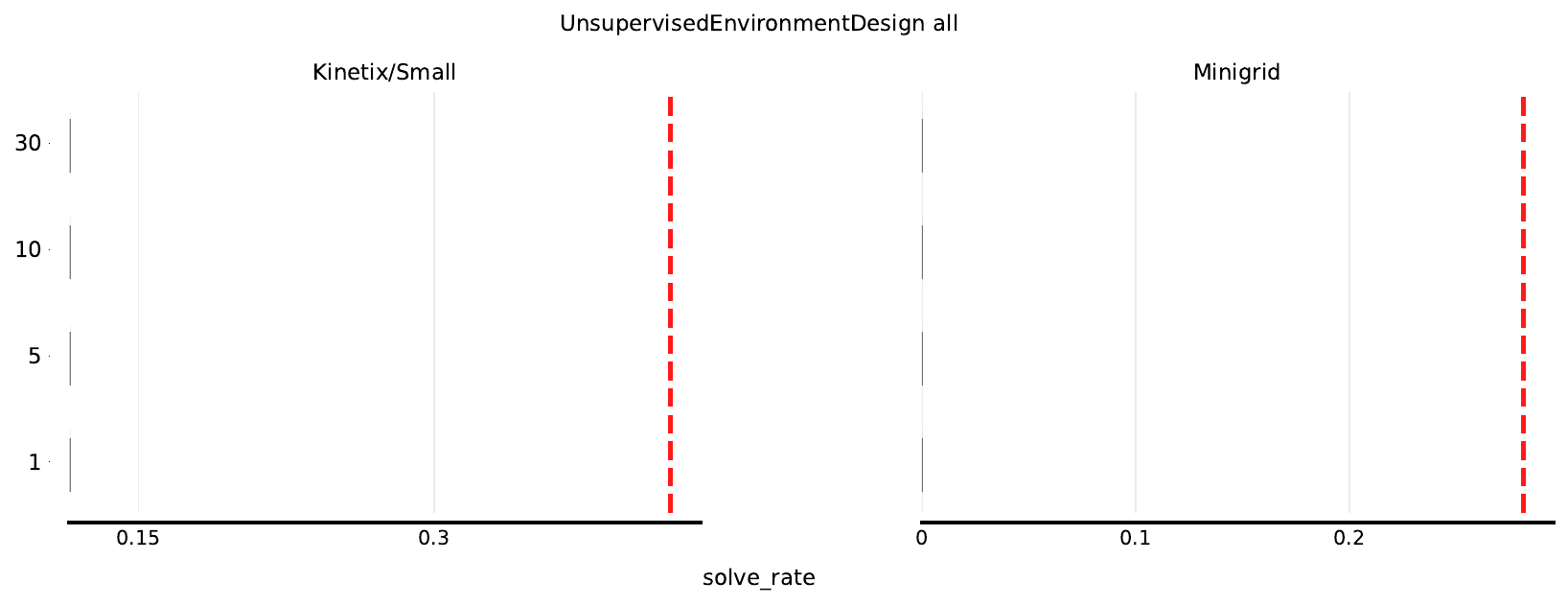}%
\hfill%
\includegraphics[width=0.48\textwidth]{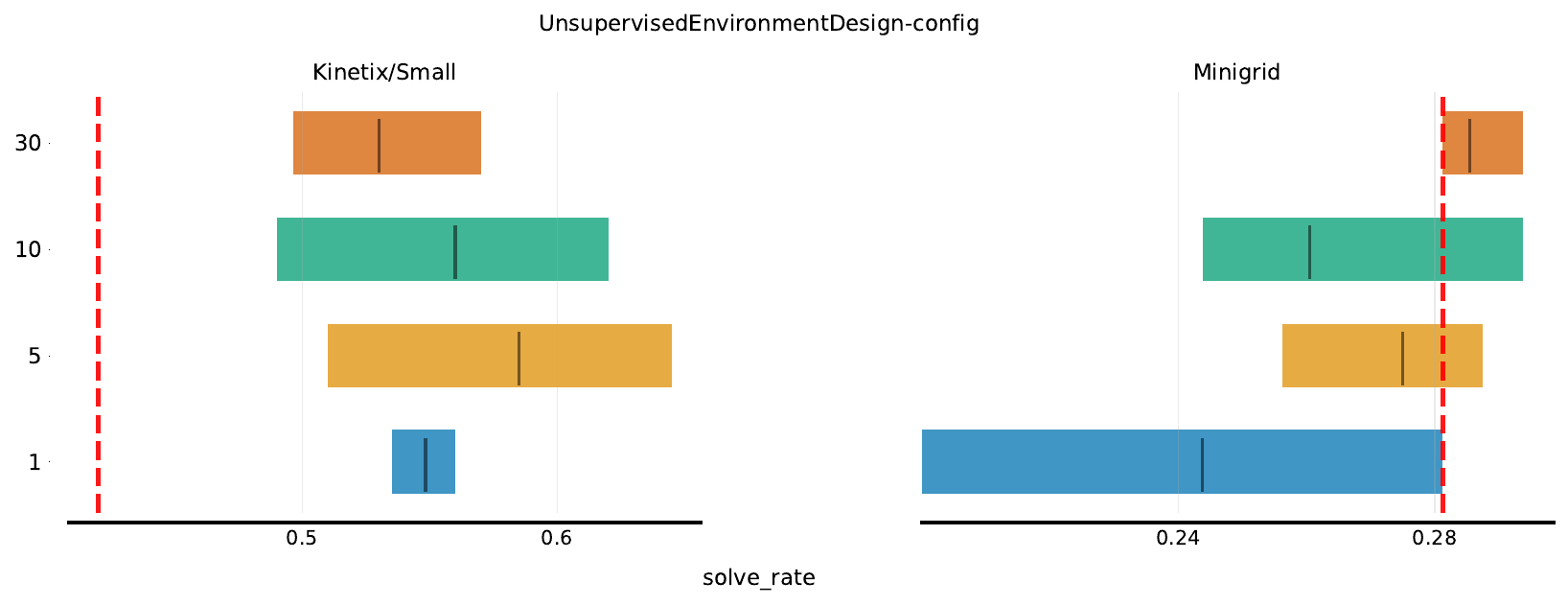}%
\\[0.5em]
\includegraphics[width=0.48\textwidth]{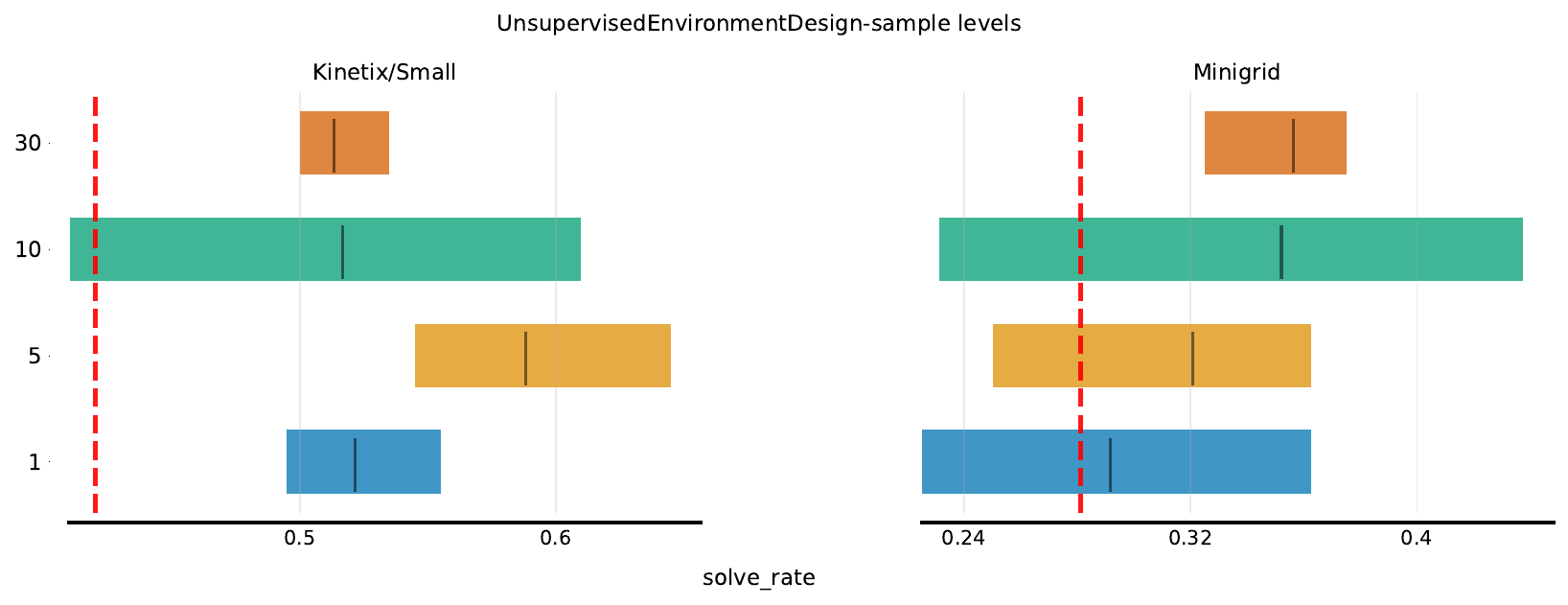}%
\hfill%
\includegraphics[width=0.48\textwidth]{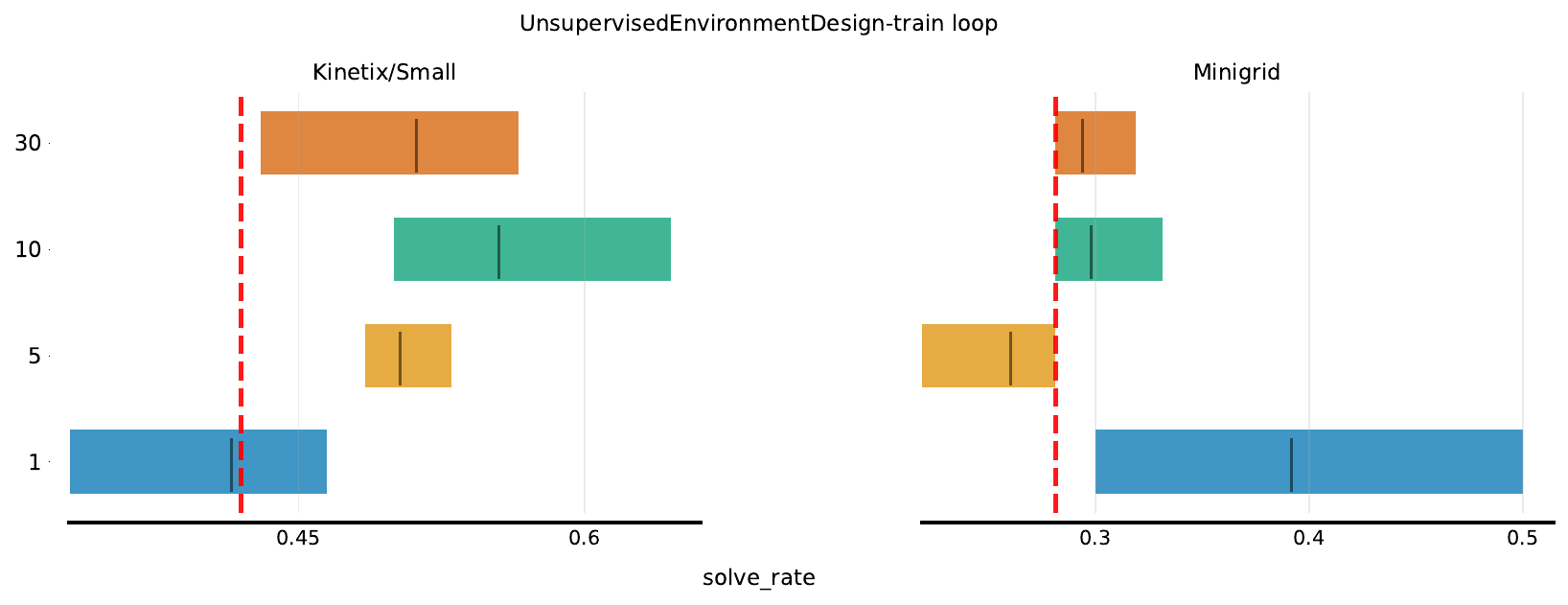}%
\\[0.5em]
\includegraphics[width=0.48\textwidth]{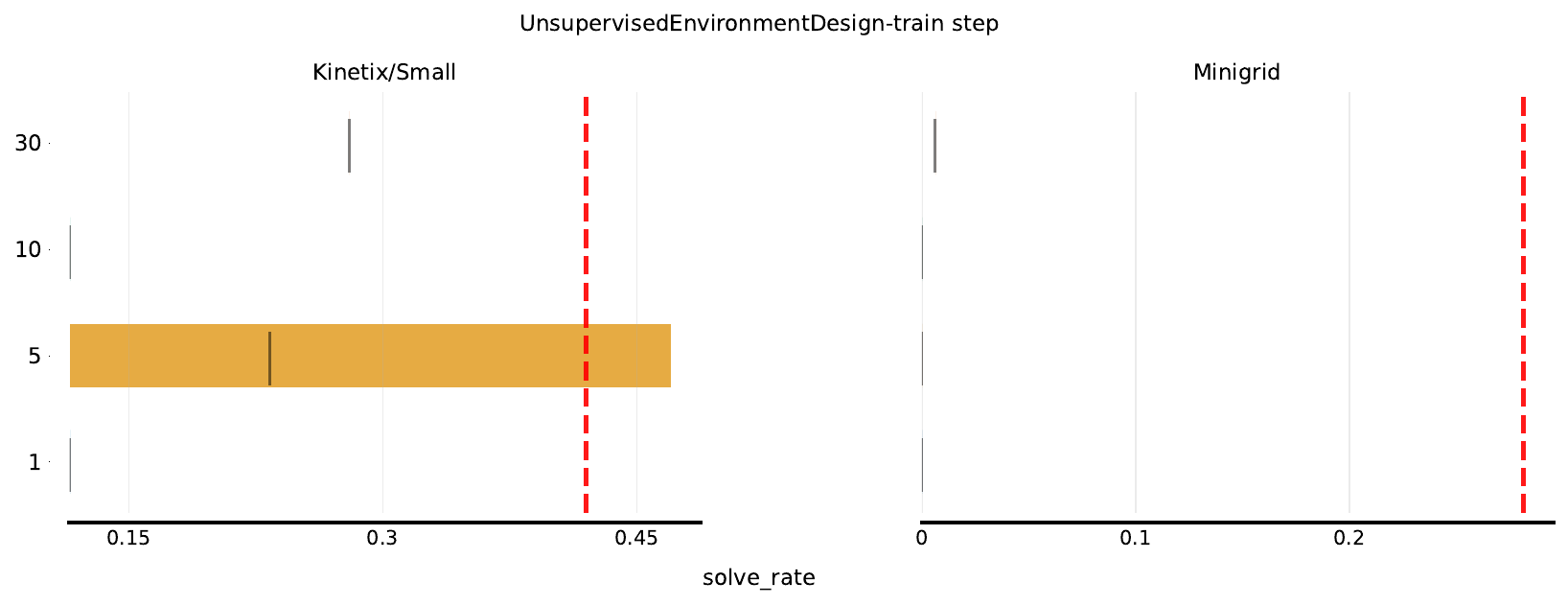}%
\hfill%
\caption{ADA Optimisation results on Meta-Train tasks. (Part 7/7)}
\label{fig:ADA_optimisation_id_7}
\end{figure}
\clearpage

\subsection{ADA Optimisation -- Meta-Test}
\label{sec:ADA_optimisation_mt}

\begin{figure}[htbp]
\centering
\setlength{\lineskip}{0pt}
\includegraphics[width=0.48\textwidth]{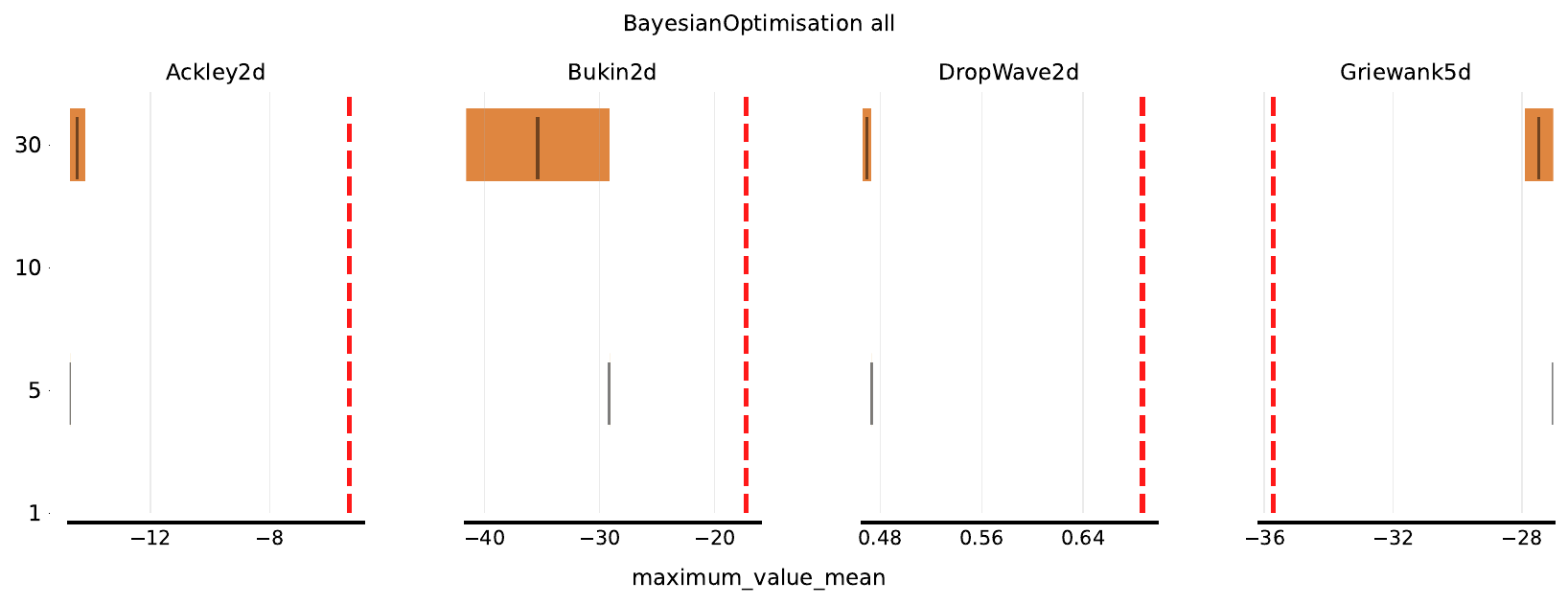}%
\hfill%
\includegraphics[width=0.48\textwidth]{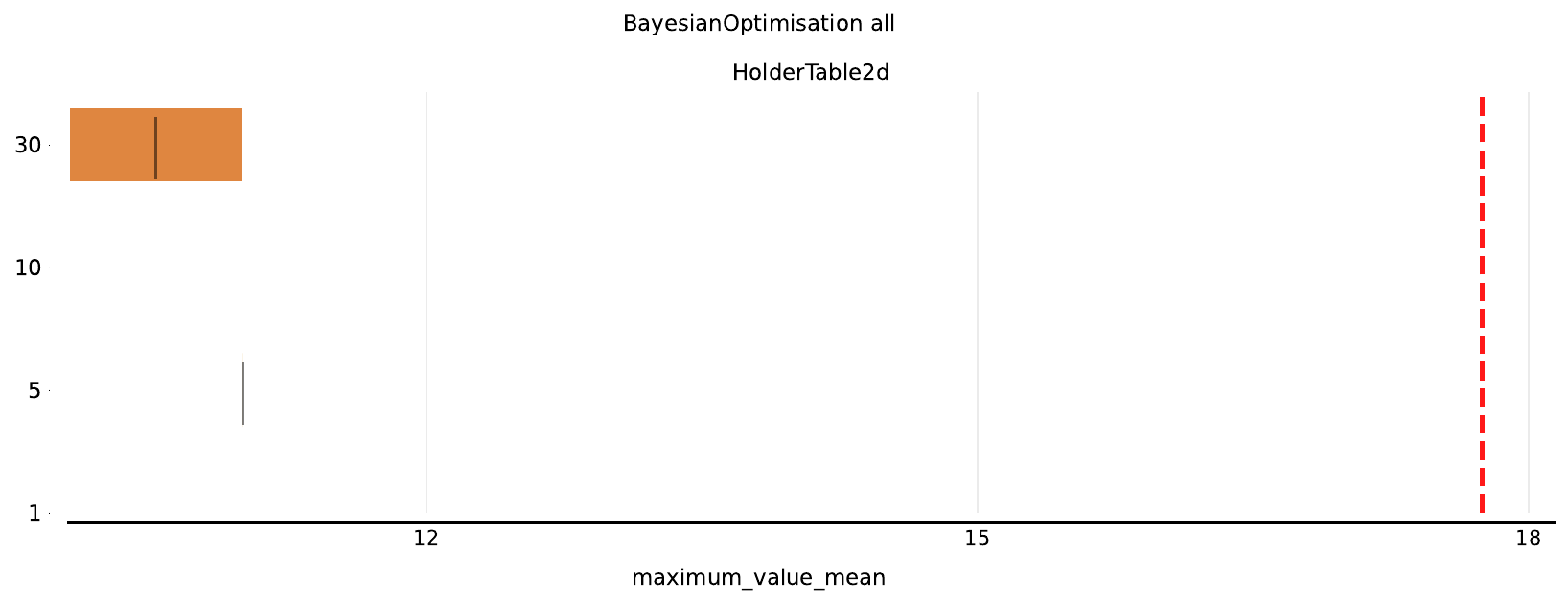}%
\\[0.5em]
\includegraphics[width=0.48\textwidth]{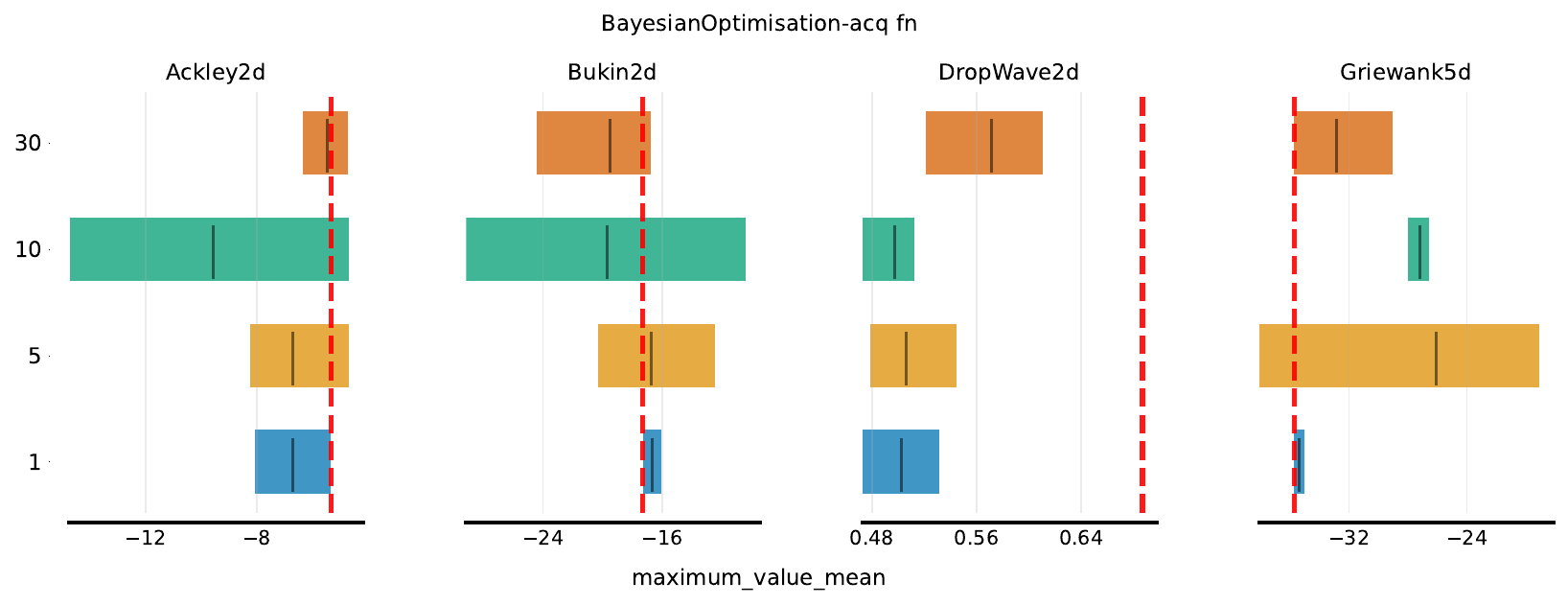}%
\hfill%
\includegraphics[width=0.48\textwidth]{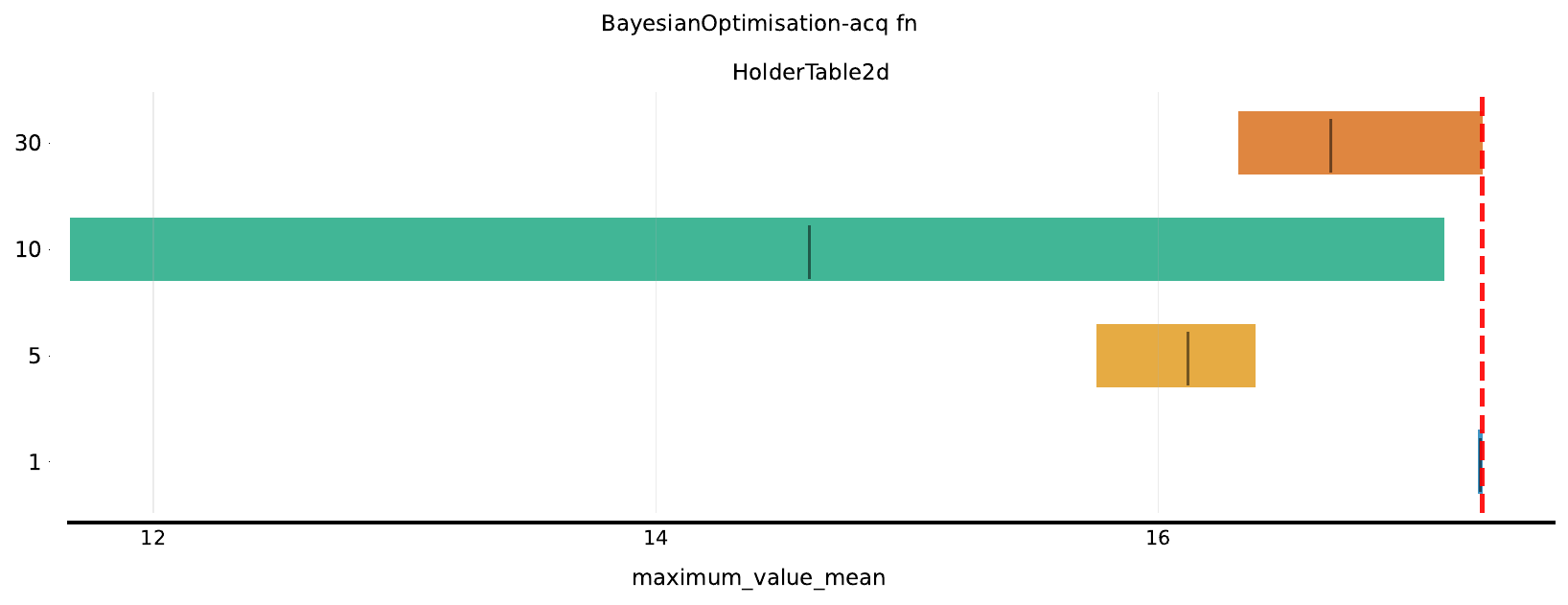}%
\\[0.5em]
\includegraphics[width=0.48\textwidth]{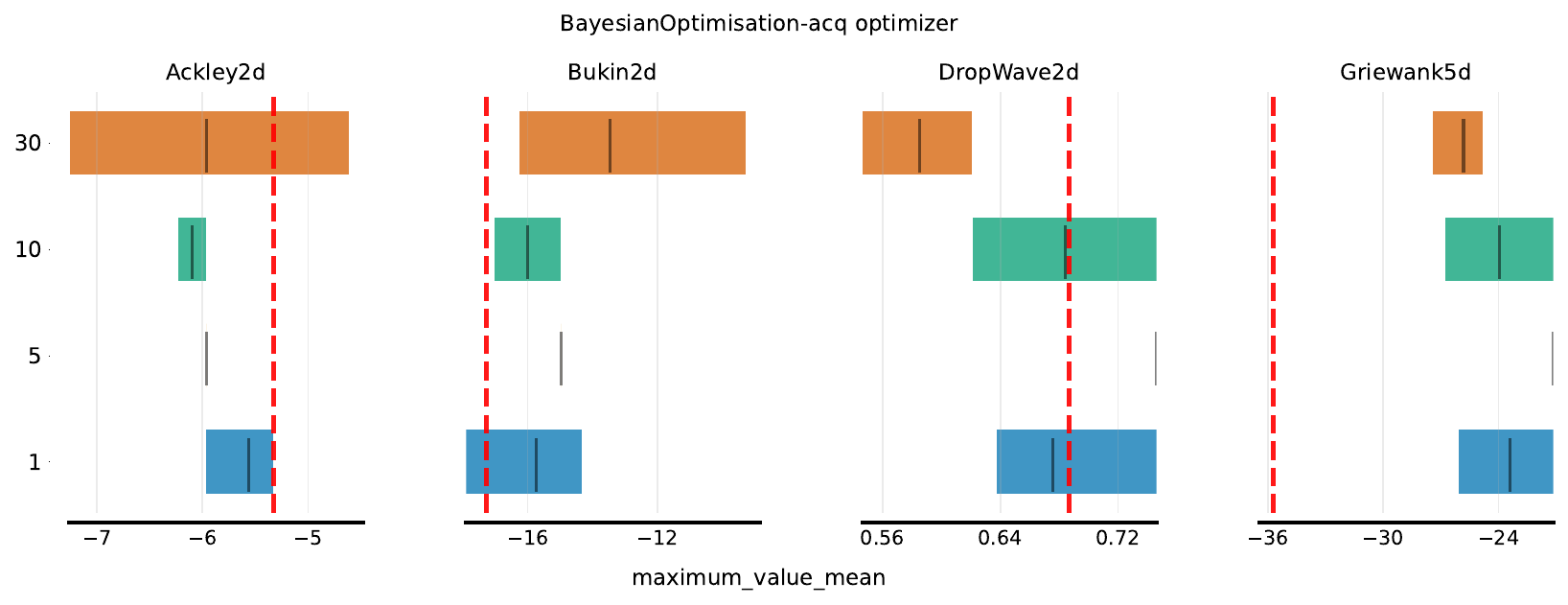}%
\hfill%
\includegraphics[width=0.48\textwidth]{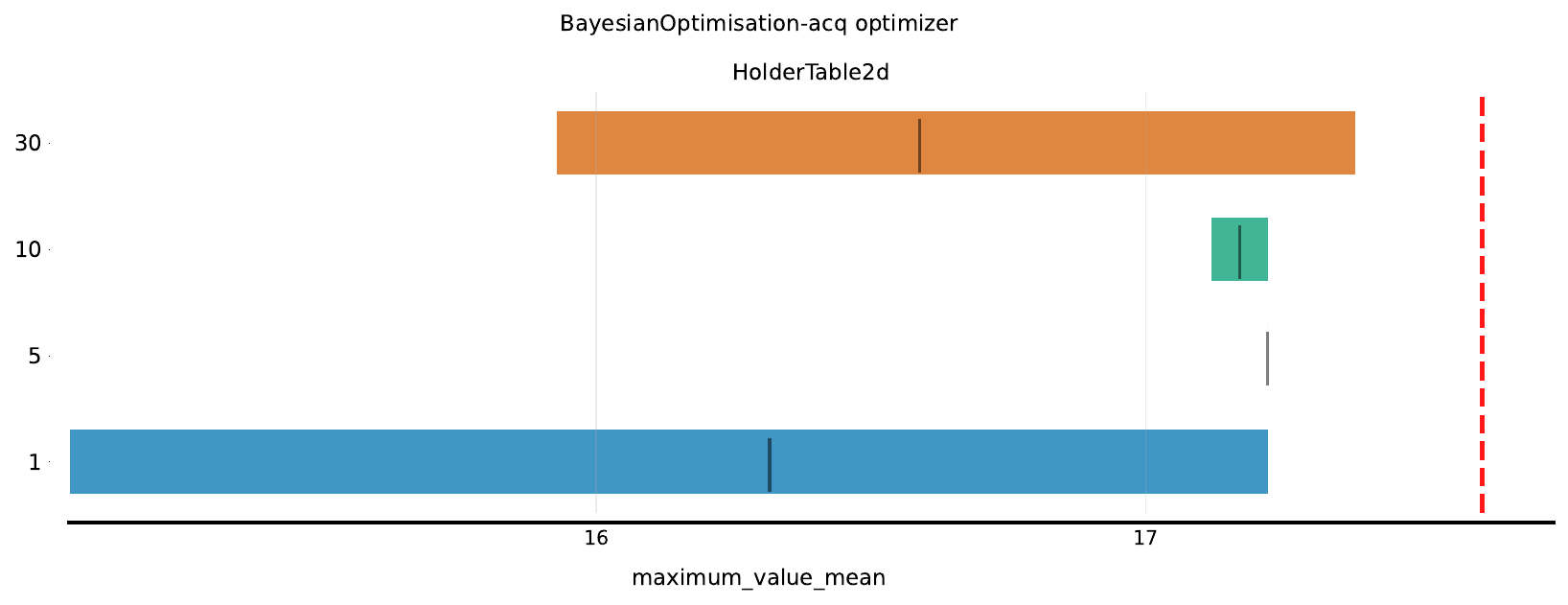}%
\\[0.5em]
\includegraphics[width=0.48\textwidth]{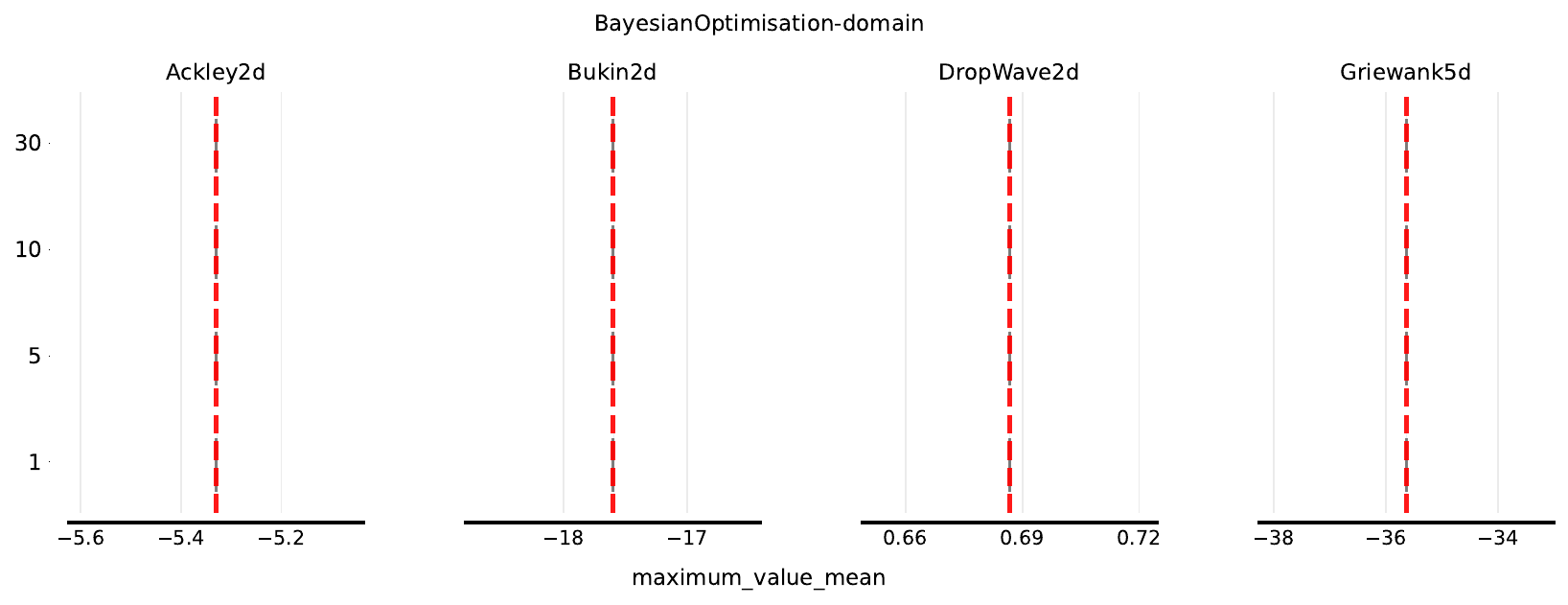}%
\hfill%
\includegraphics[width=0.48\textwidth]{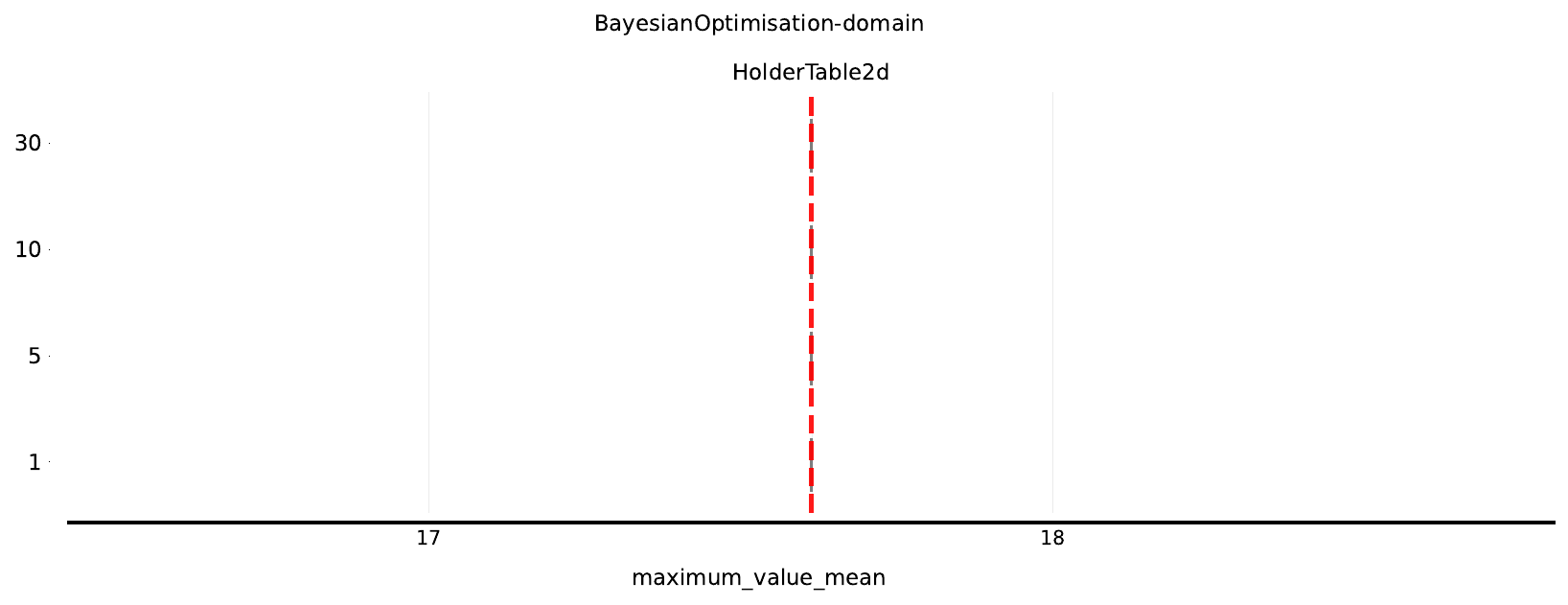}%
\\[0.5em]
\includegraphics[width=0.48\textwidth]{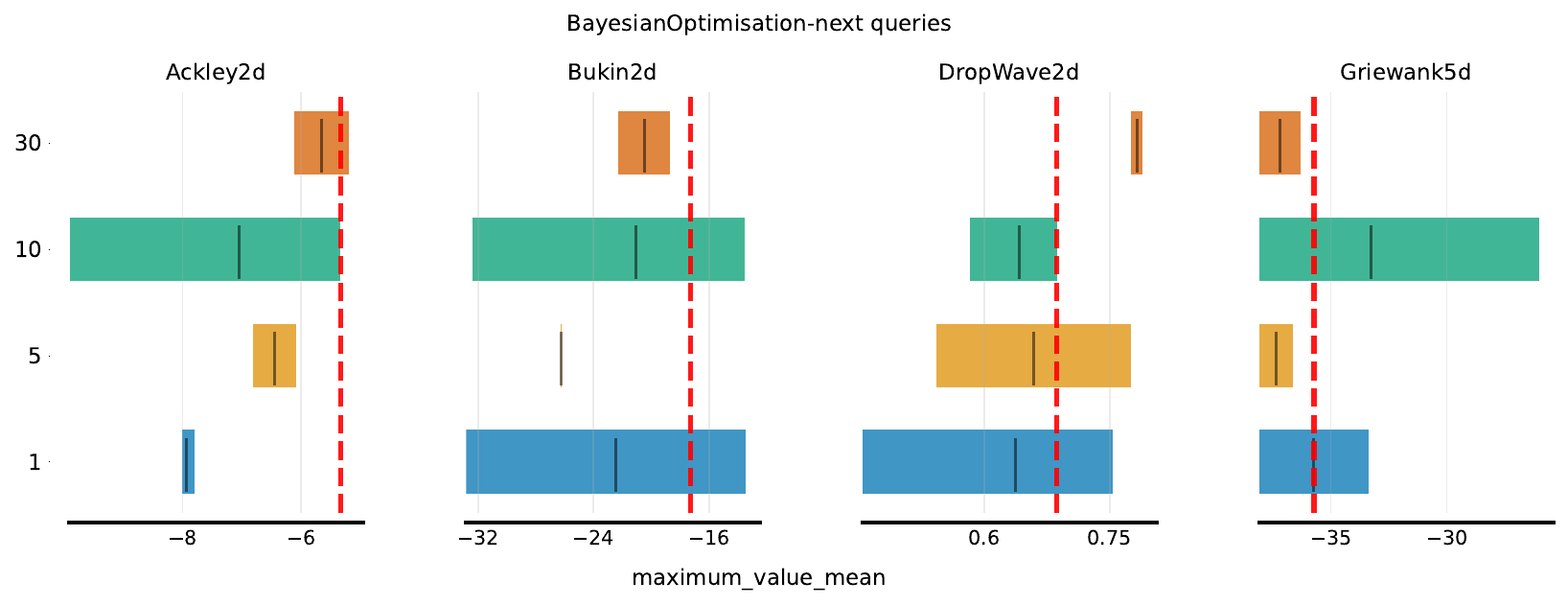}%
\hfill%
\includegraphics[width=0.48\textwidth]{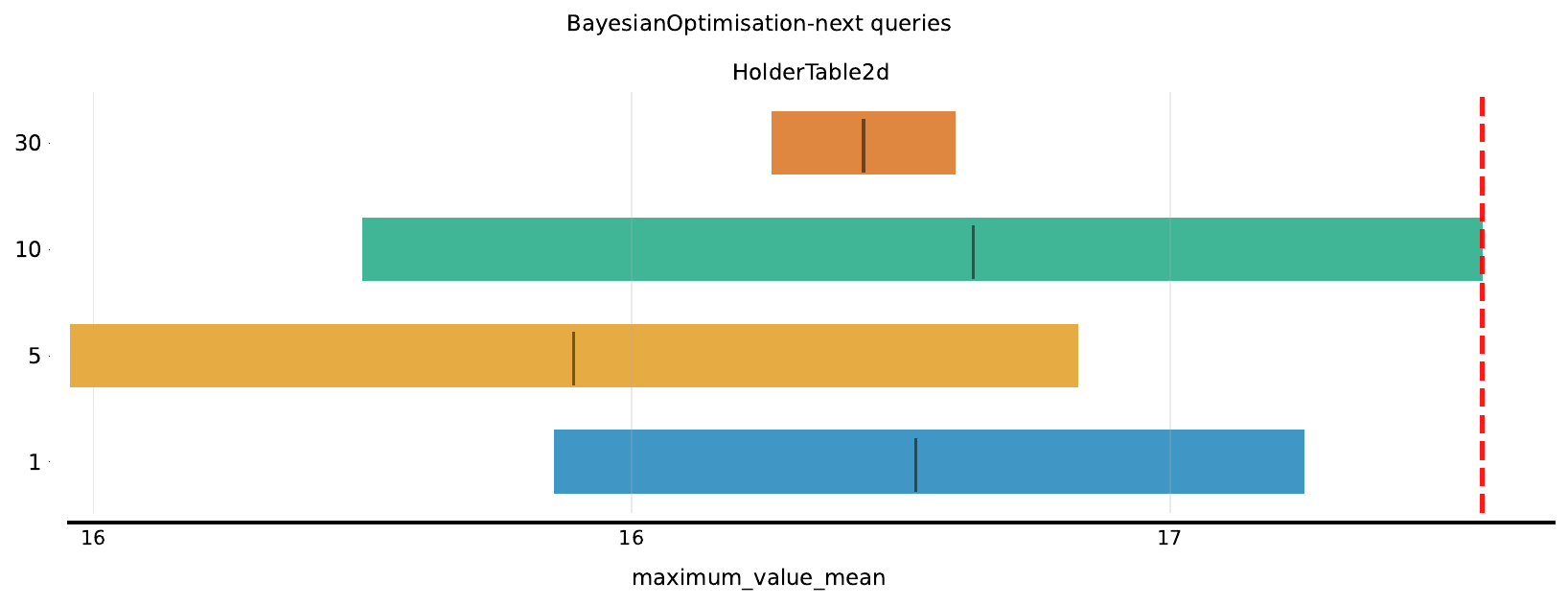}%
\caption{ADA Optimisation results on Meta-Test tasks. (Part 1/8)}
\label{fig:ADA_optimisation_mt_1}
\end{figure}
\clearpage

\begin{figure}[htbp]
\centering
\setlength{\lineskip}{0pt}
\includegraphics[width=0.48\textwidth]{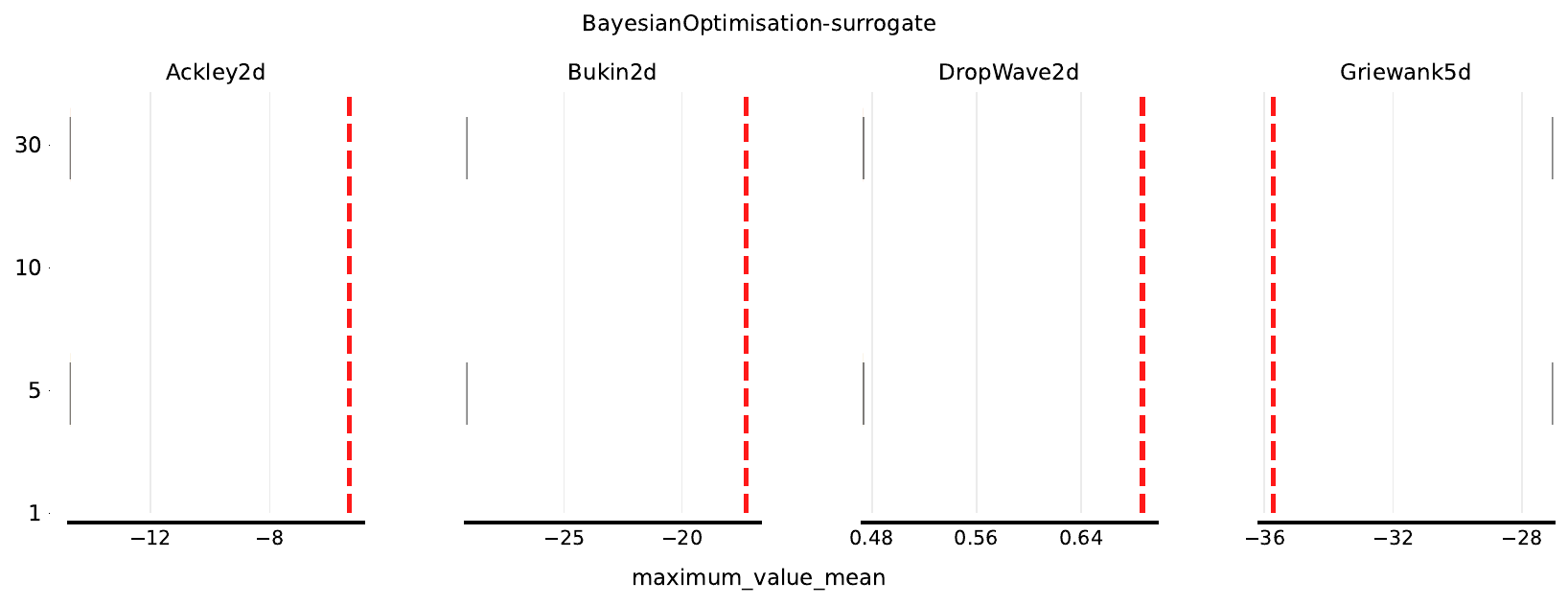}%
\hfill%
\includegraphics[width=0.48\textwidth]{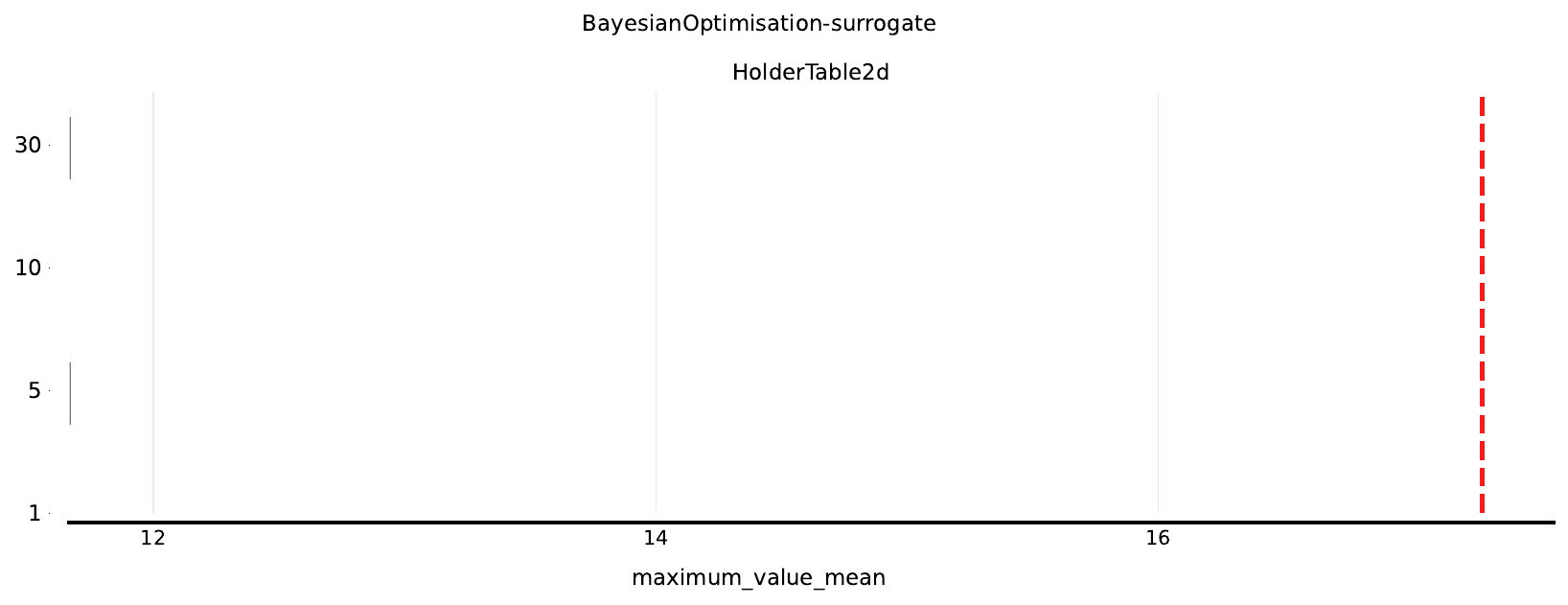}%
\\[0.5em]
\includegraphics[width=0.48\textwidth]{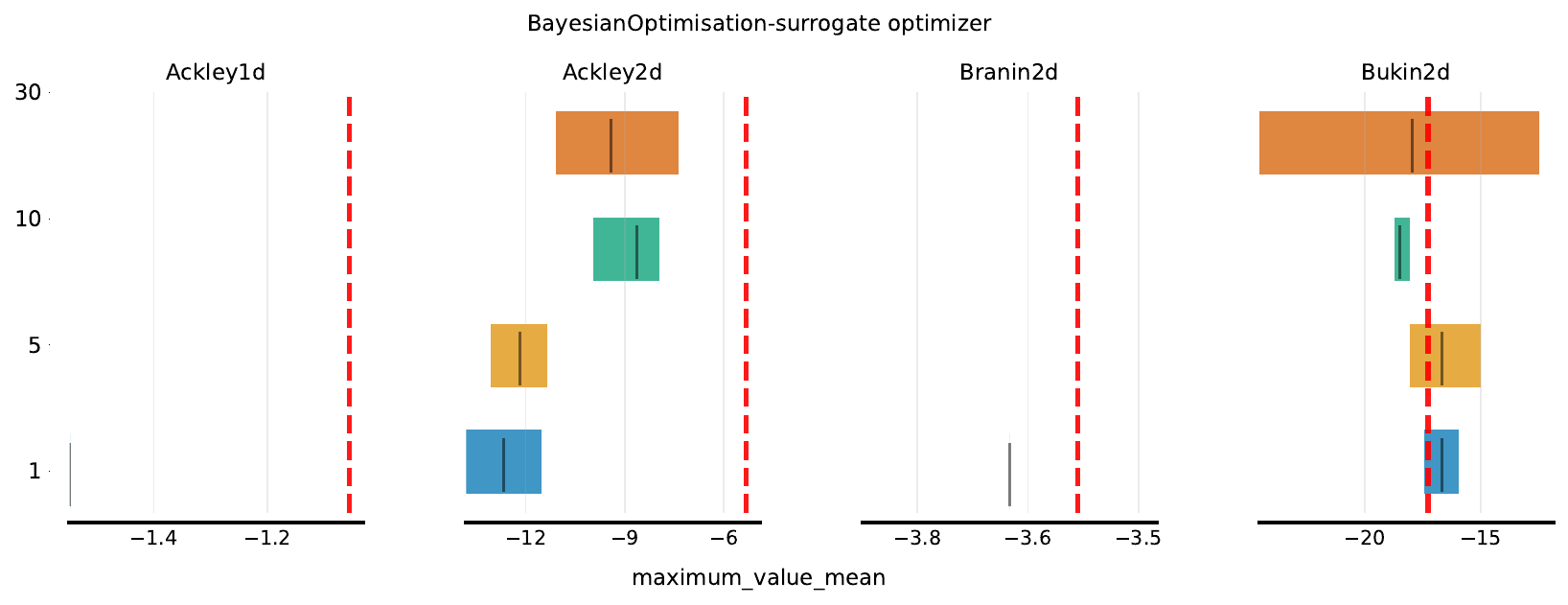}%
\hfill%
\includegraphics[width=0.48\textwidth]{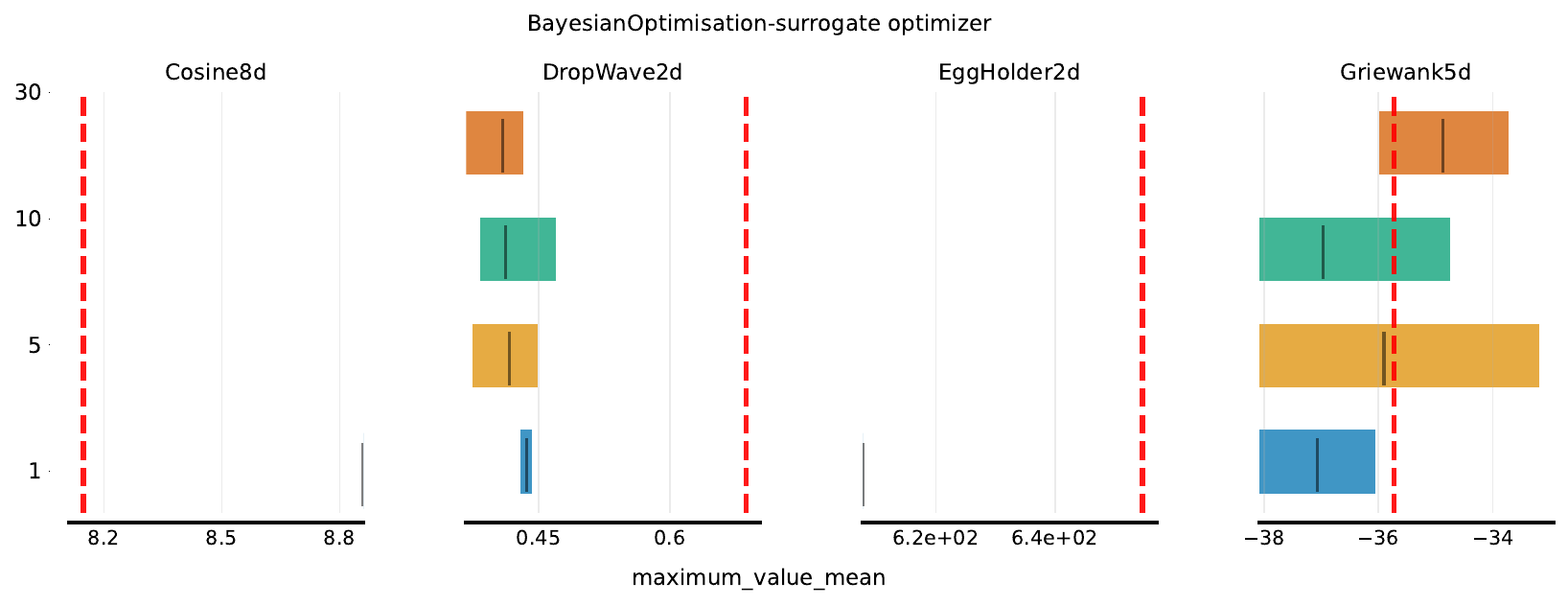}%
\\[0.5em]
\includegraphics[width=0.48\textwidth]{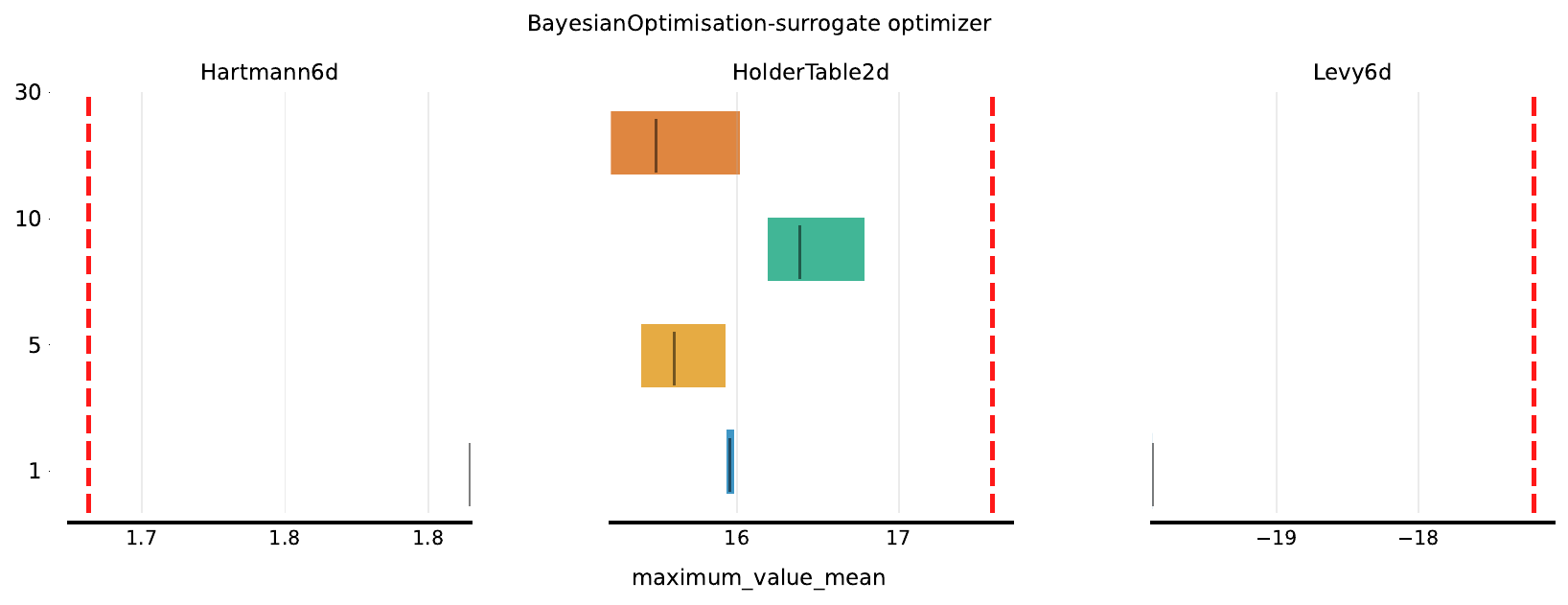}%
\hfill%
\includegraphics[width=0.48\textwidth]{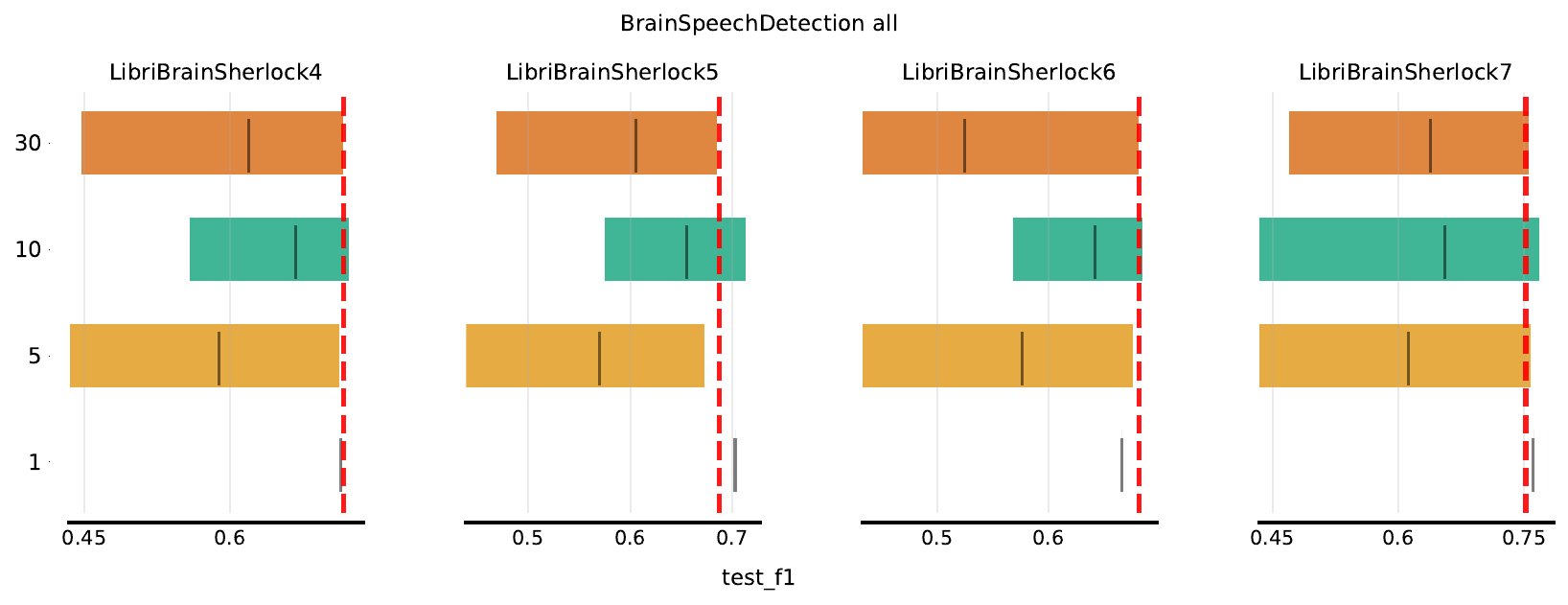}%
\\[0.5em]
\includegraphics[width=0.48\textwidth]{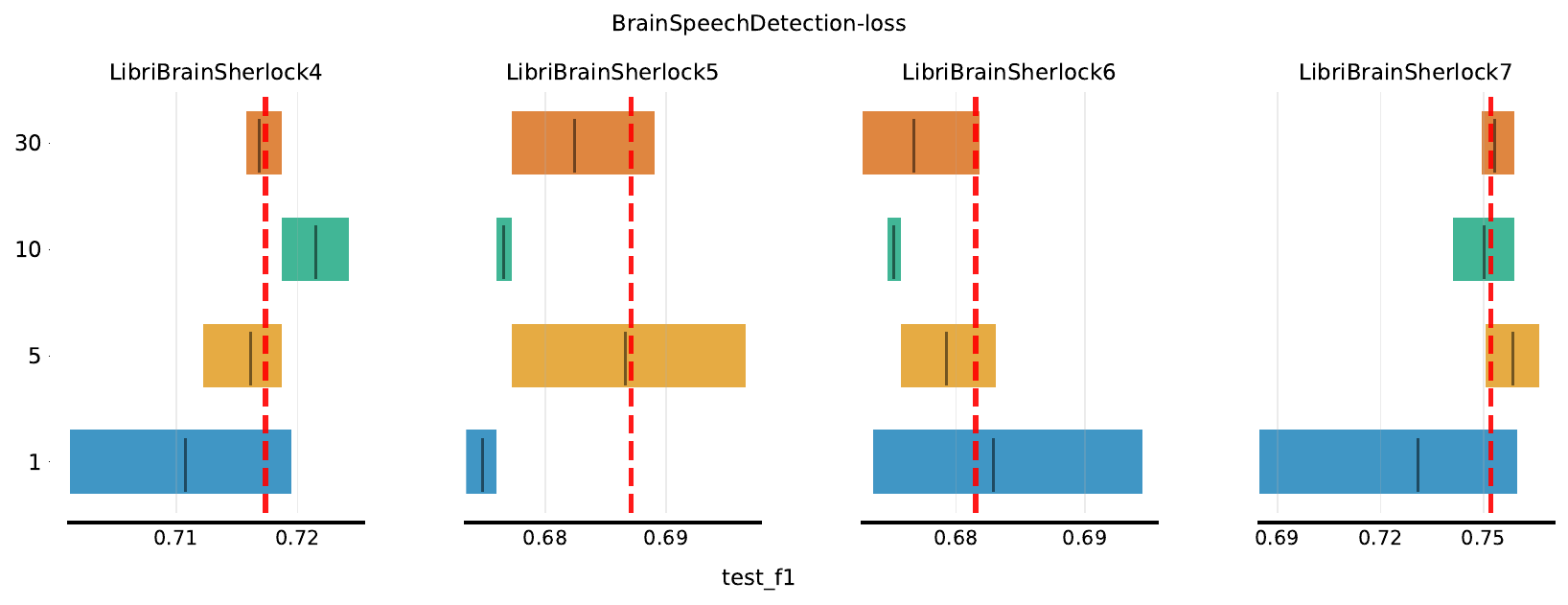}%
\hfill%
\includegraphics[width=0.48\textwidth]{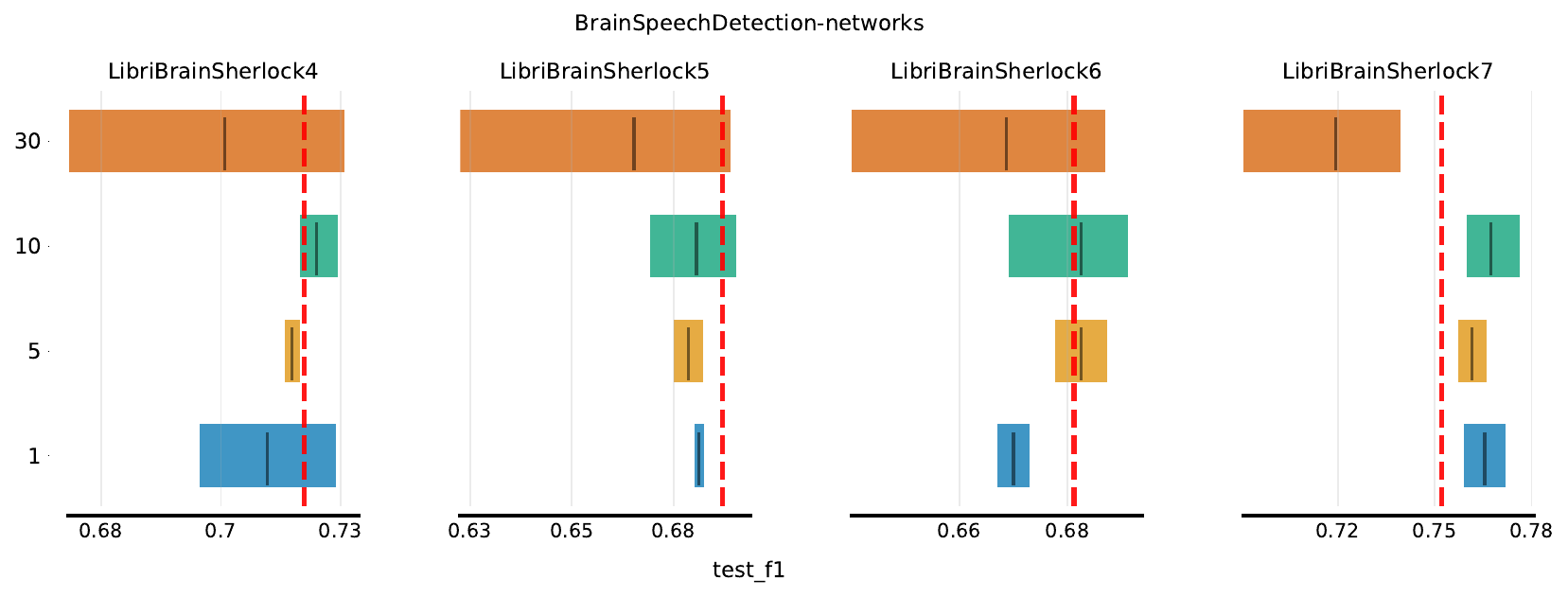}%
\\[0.5em]
\includegraphics[width=0.48\textwidth]{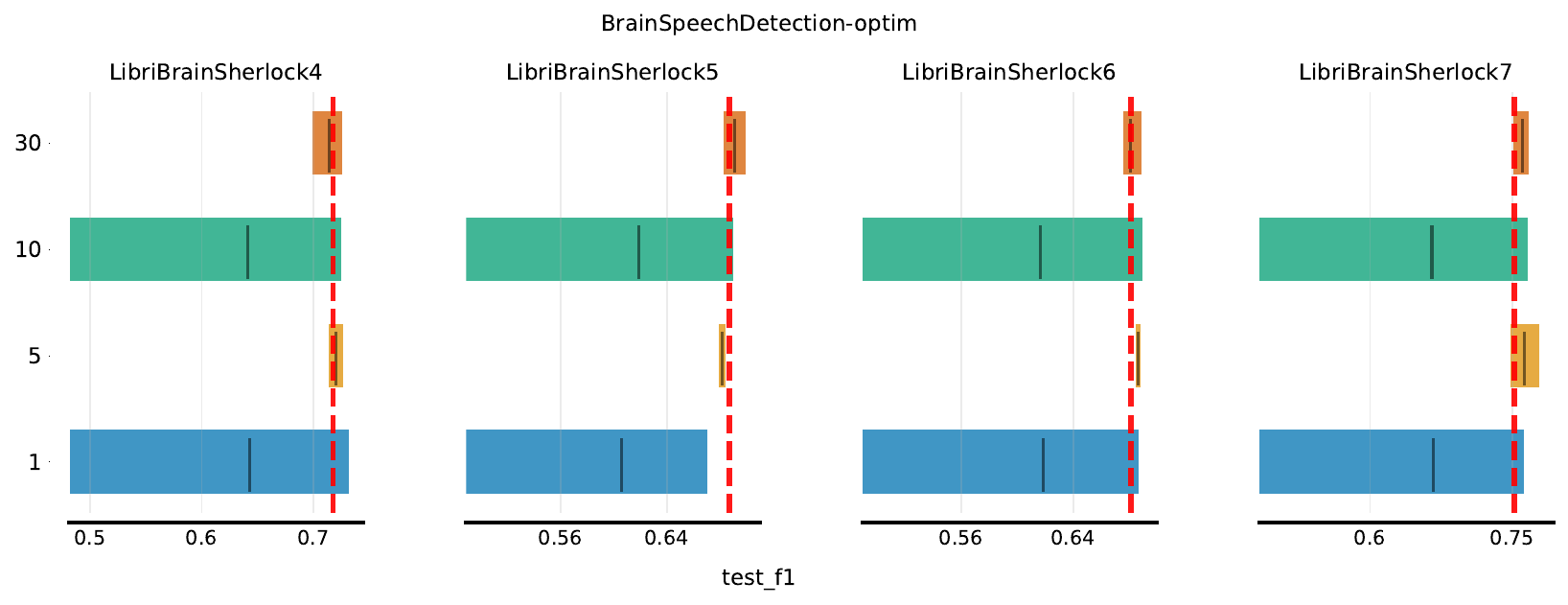}%
\hfill%
\includegraphics[width=0.48\textwidth]{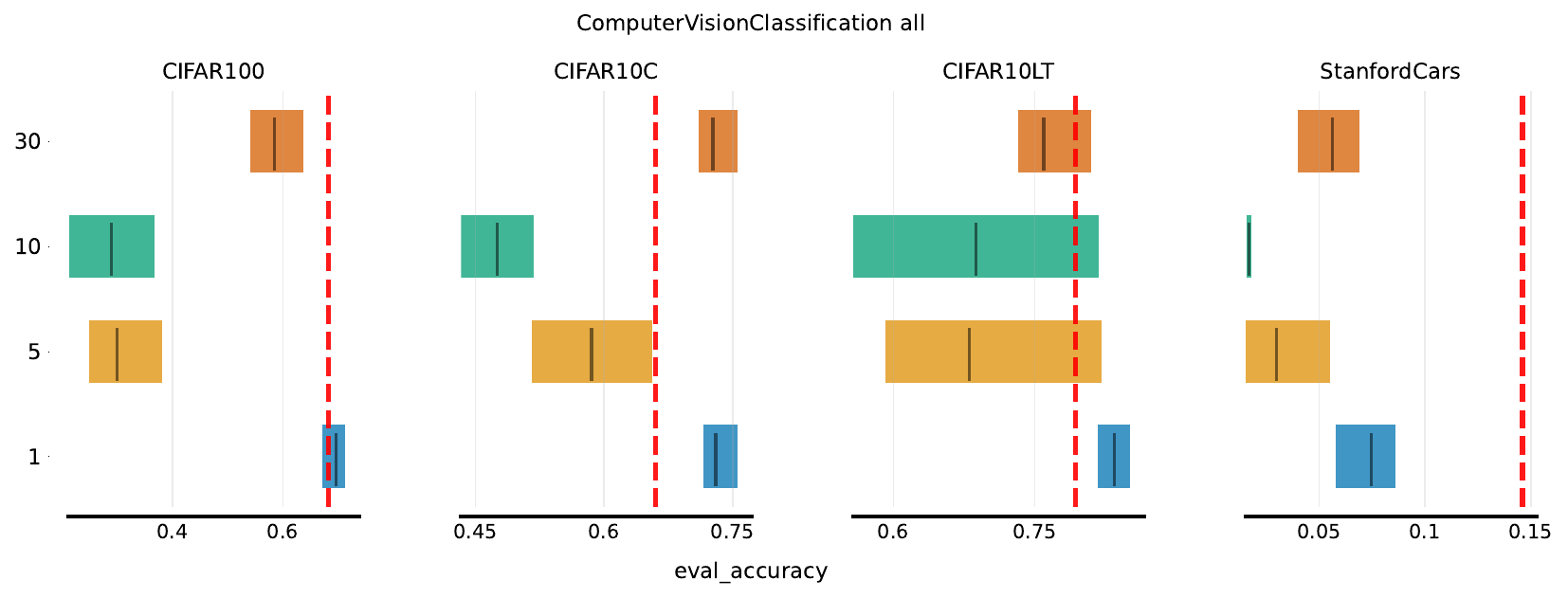}%
\\[0.5em]
\includegraphics[width=0.48\textwidth]{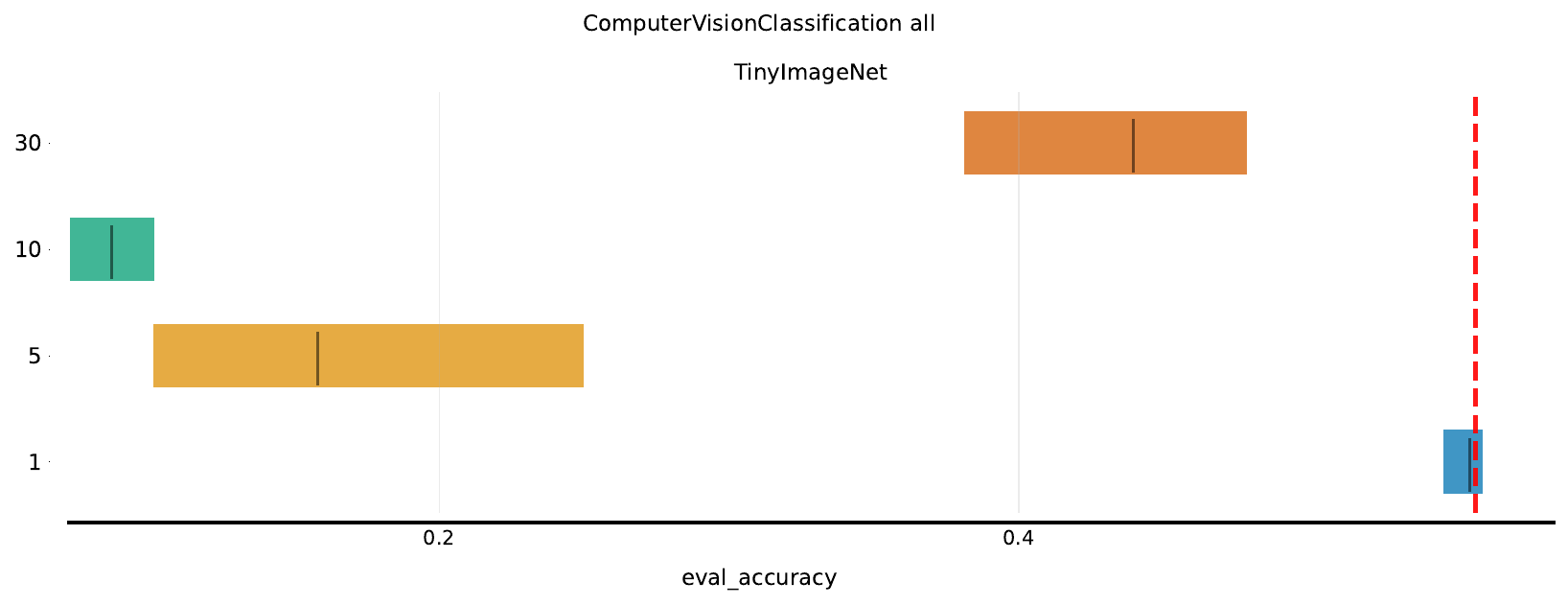}%
\hfill%
\includegraphics[width=0.48\textwidth]{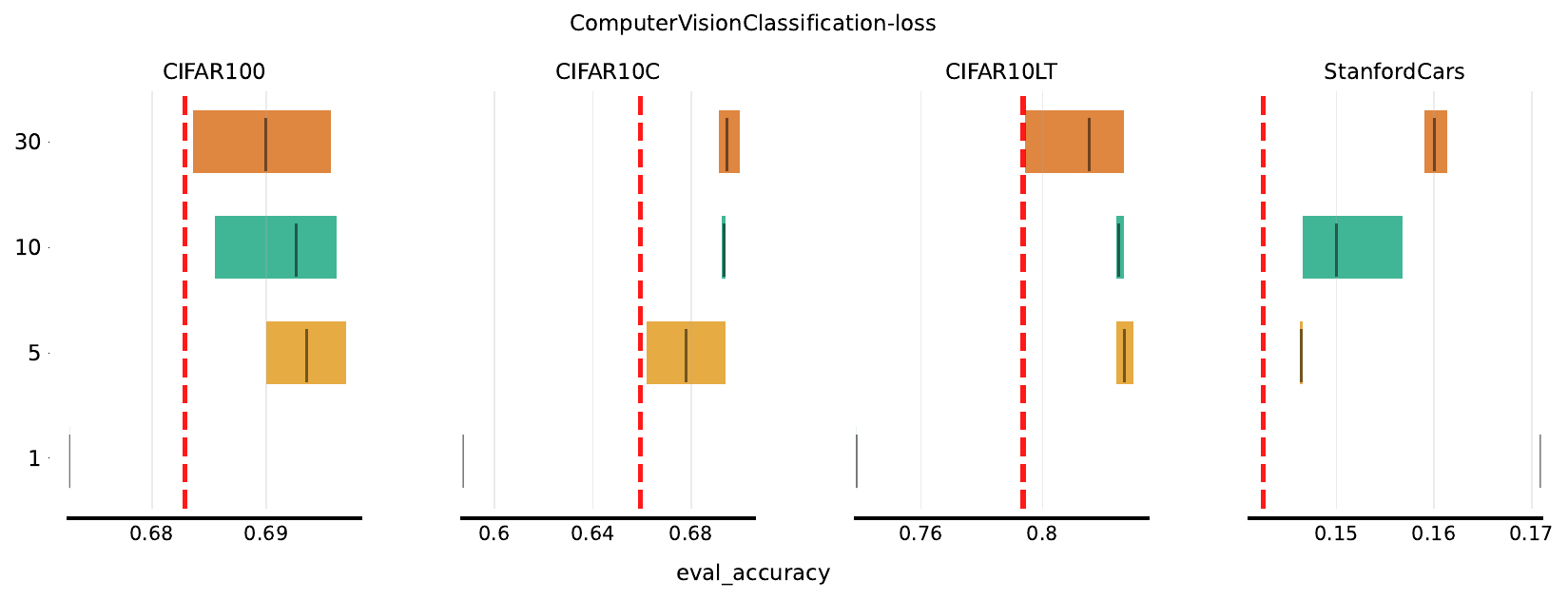}%
\caption{ADA Optimisation results on Meta-Test tasks. (Part 2/8)}
\label{fig:ADA_optimisation_mt_2}
\end{figure}
\clearpage

\begin{figure}[htbp]
\centering
\setlength{\lineskip}{0pt}
\includegraphics[width=0.48\textwidth]{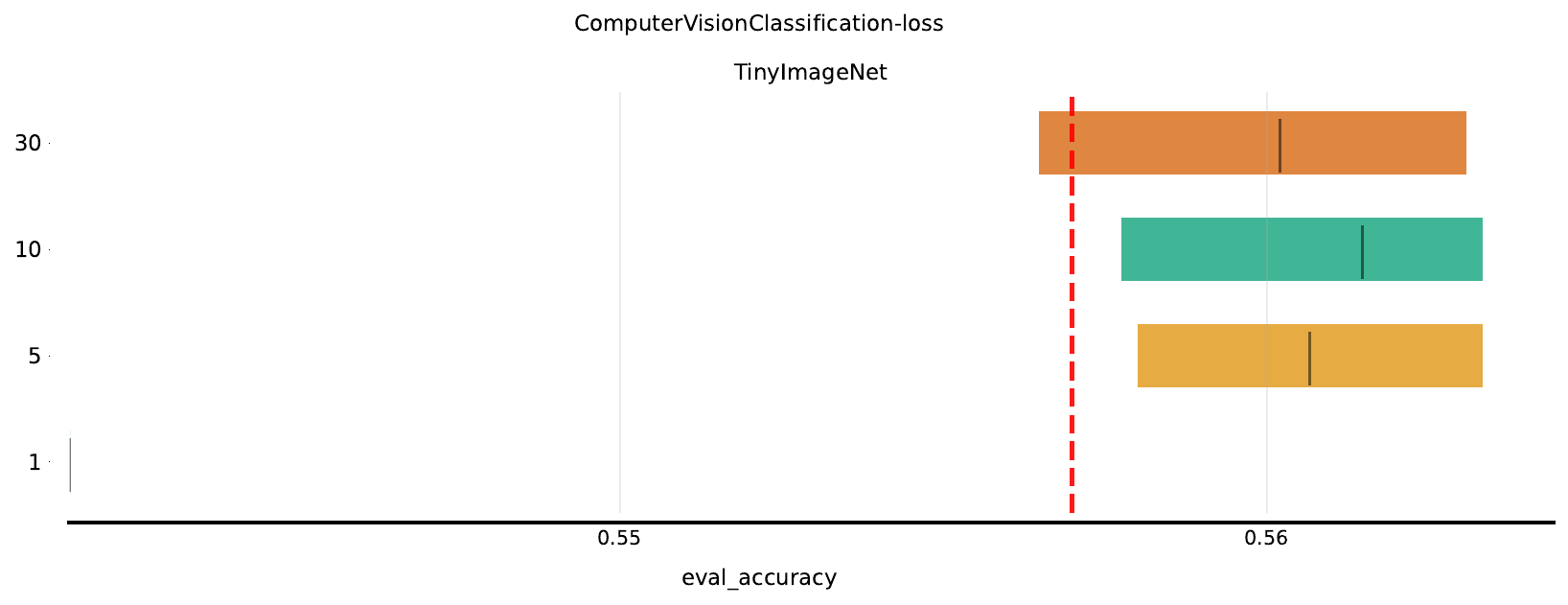}%
\hfill%
\includegraphics[width=0.48\textwidth]{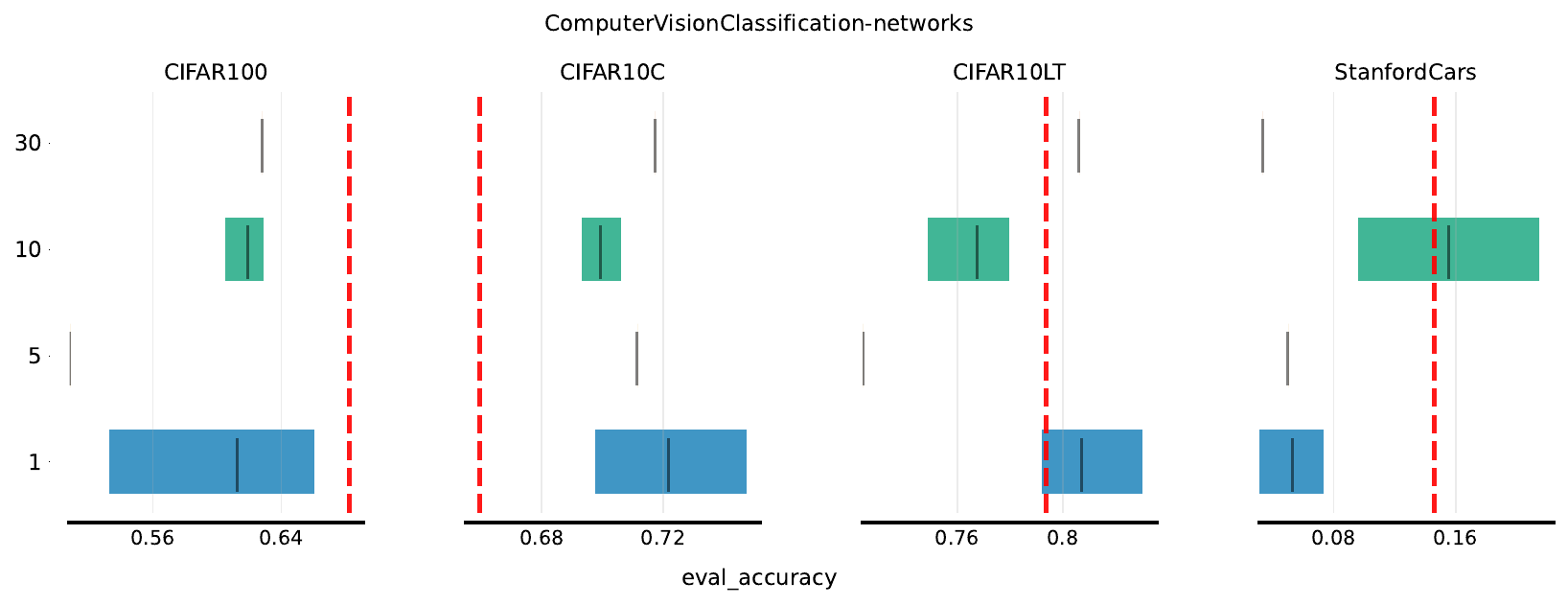}%
\\[0.5em]
\includegraphics[width=0.48\textwidth]{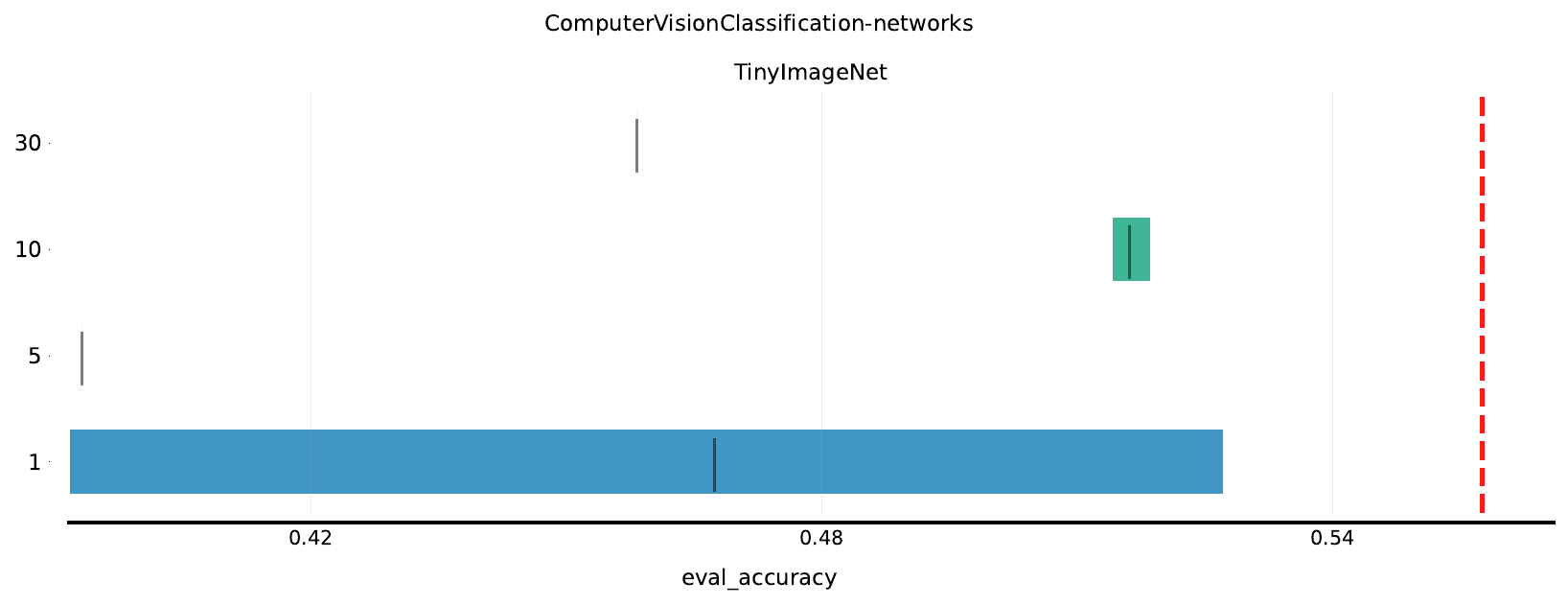}%
\hfill%
\includegraphics[width=0.48\textwidth]{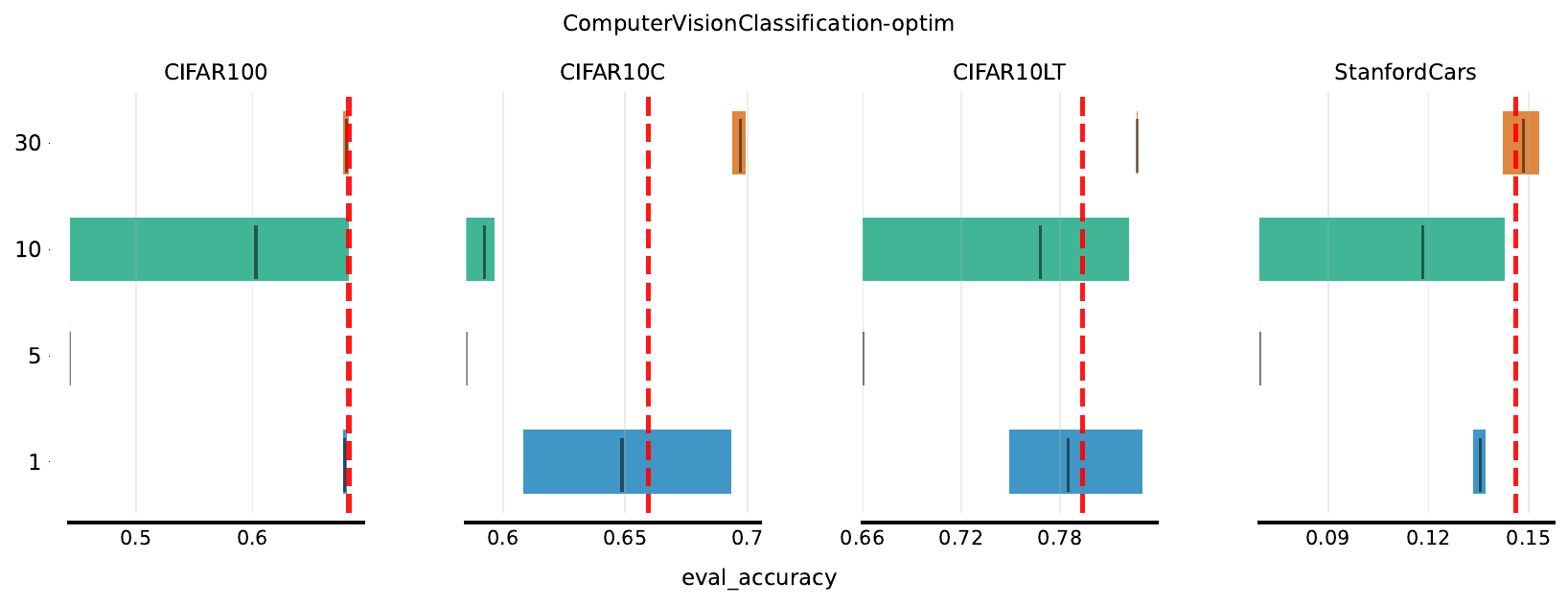}%
\\[0.5em]
\includegraphics[width=0.48\textwidth]{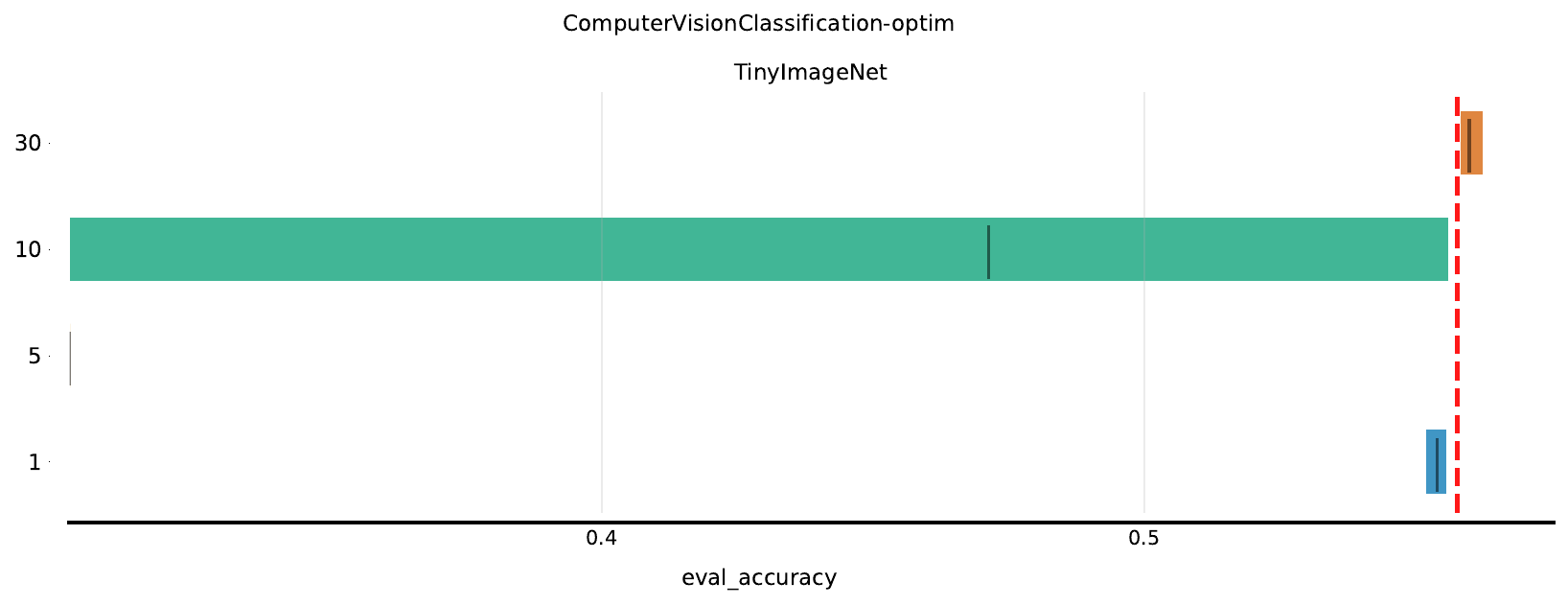}%
\hfill%
\includegraphics[width=0.48\textwidth]{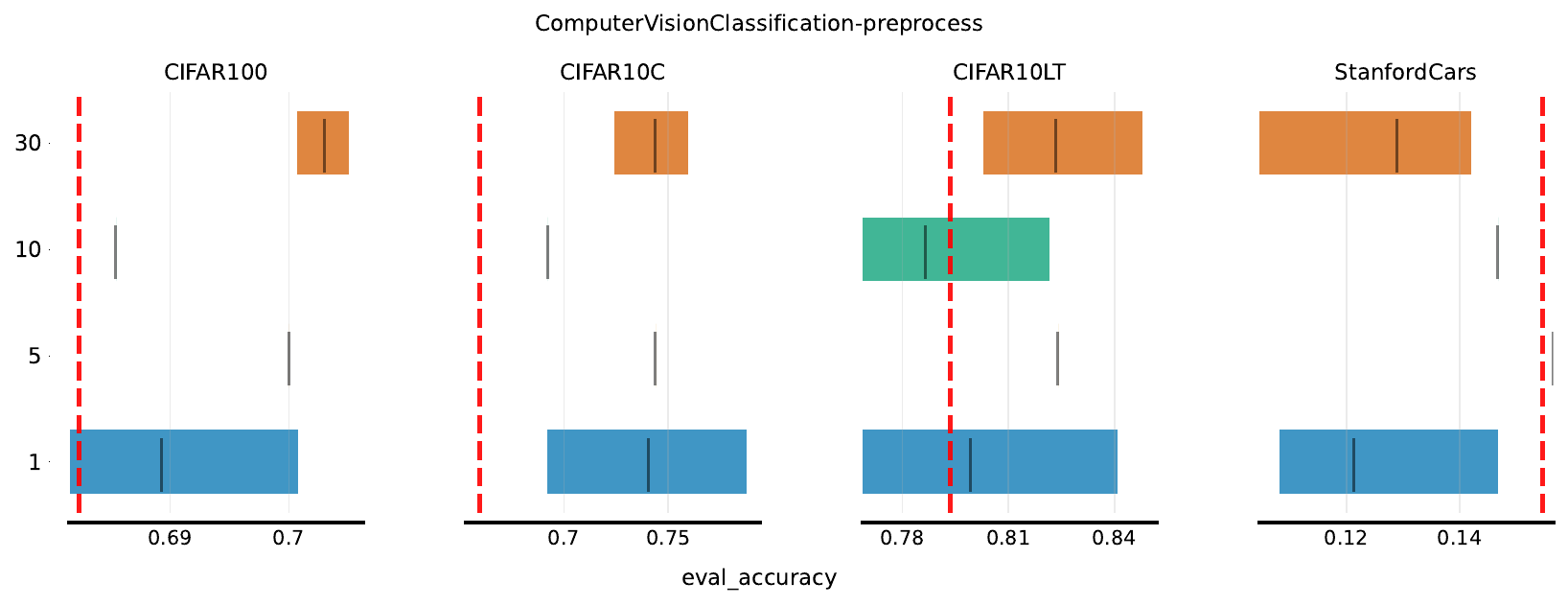}%
\\[0.5em]
\includegraphics[width=0.48\textwidth]{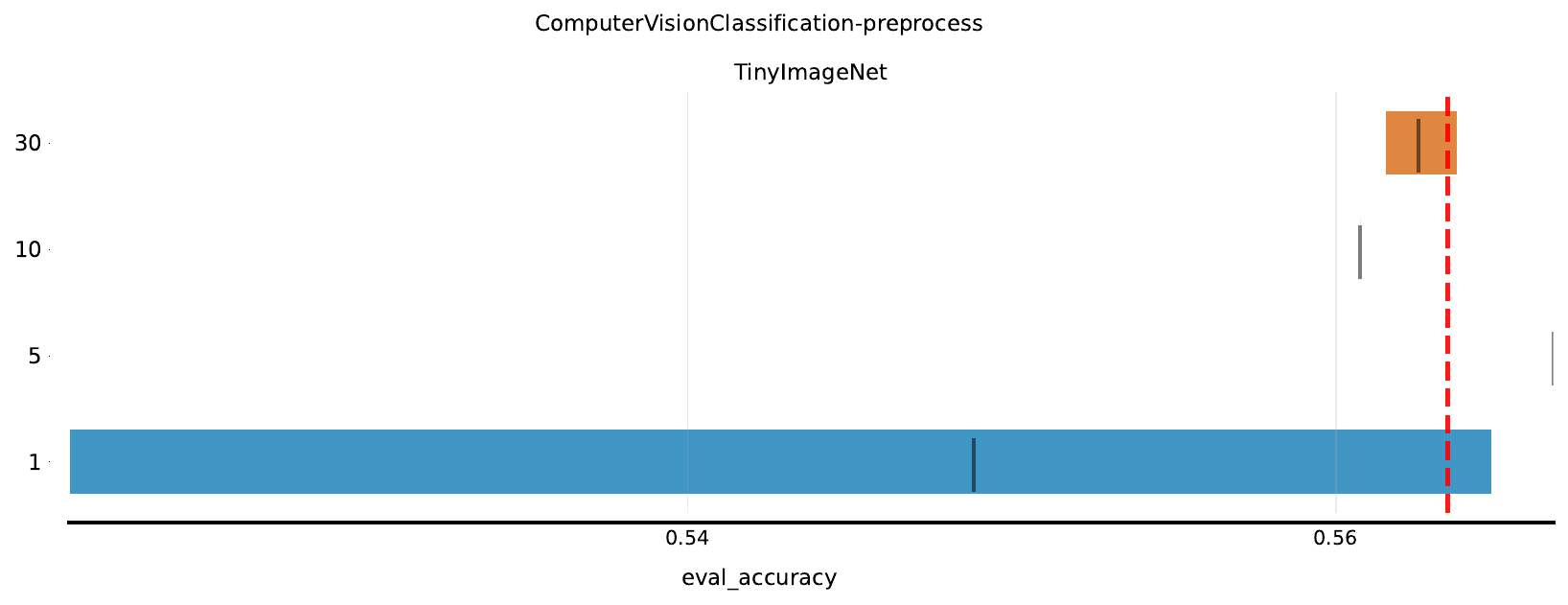}%
\hfill%
\includegraphics[width=0.48\textwidth]{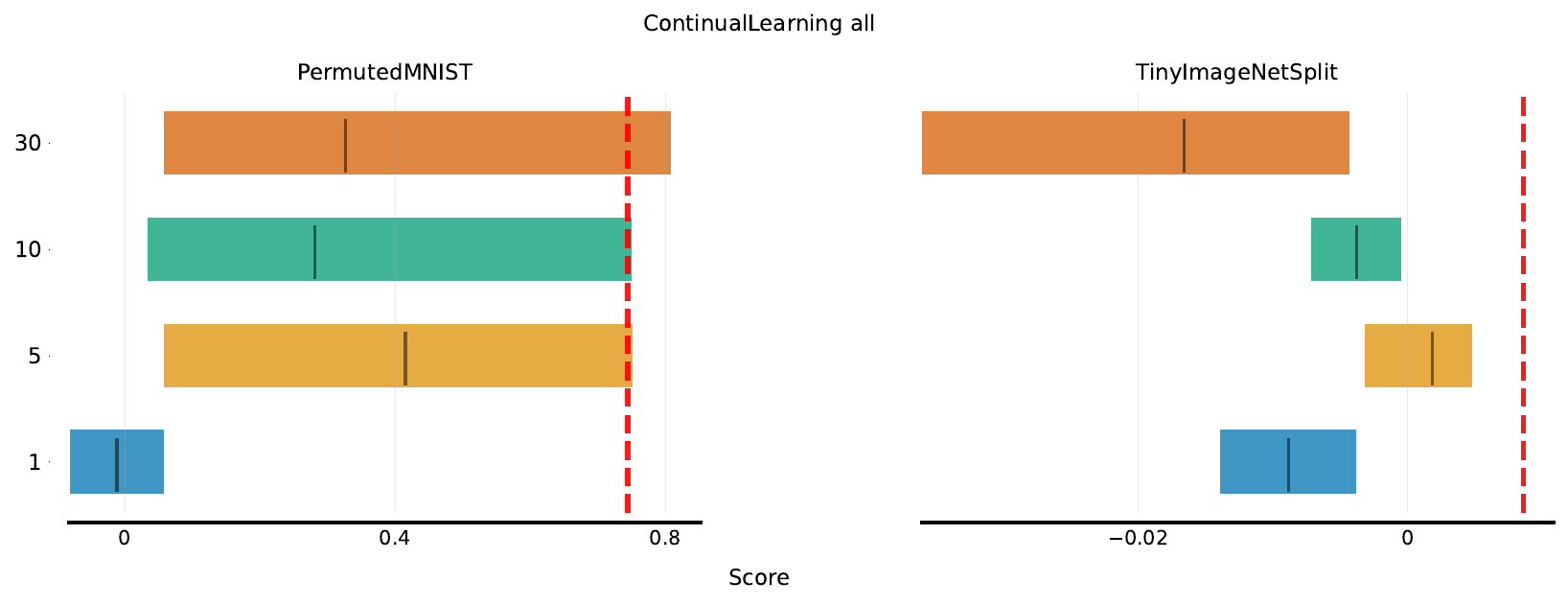}%
\\[0.5em]
\includegraphics[width=0.48\textwidth]{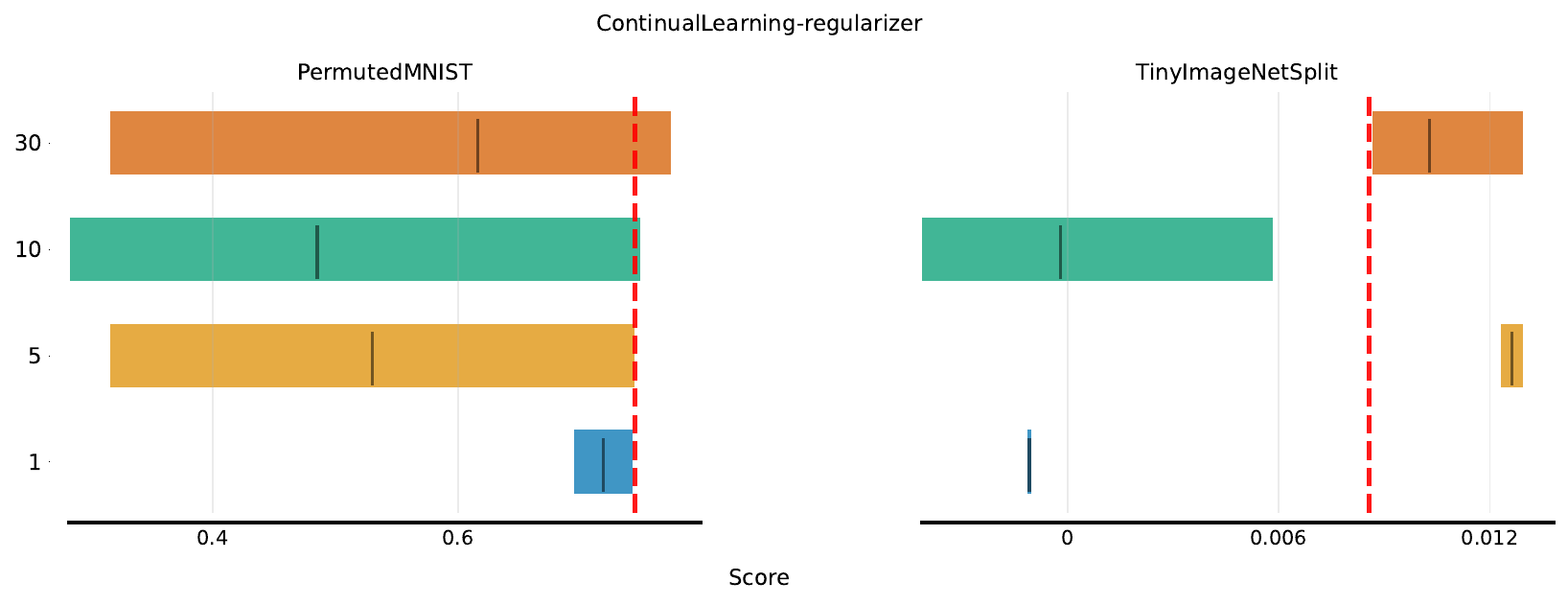}%
\hfill%
\includegraphics[width=0.48\textwidth]{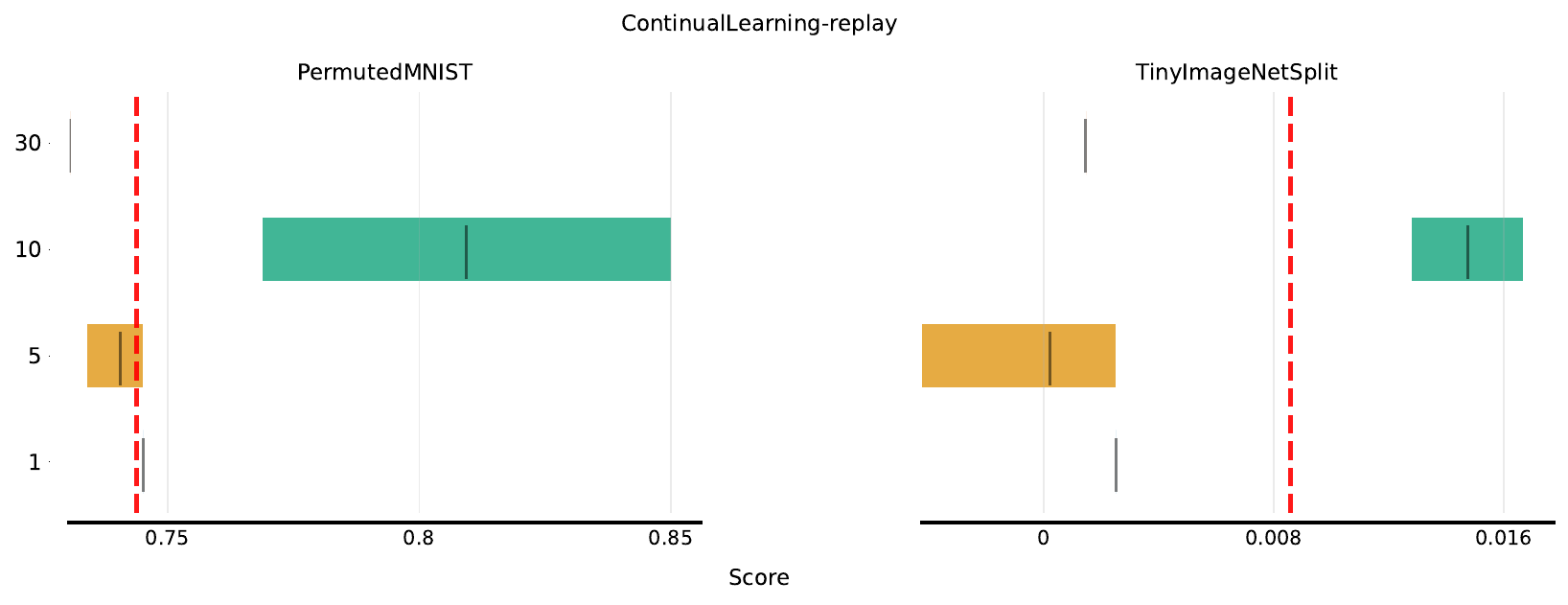}%
\\[0.5em]
\includegraphics[width=0.48\textwidth]{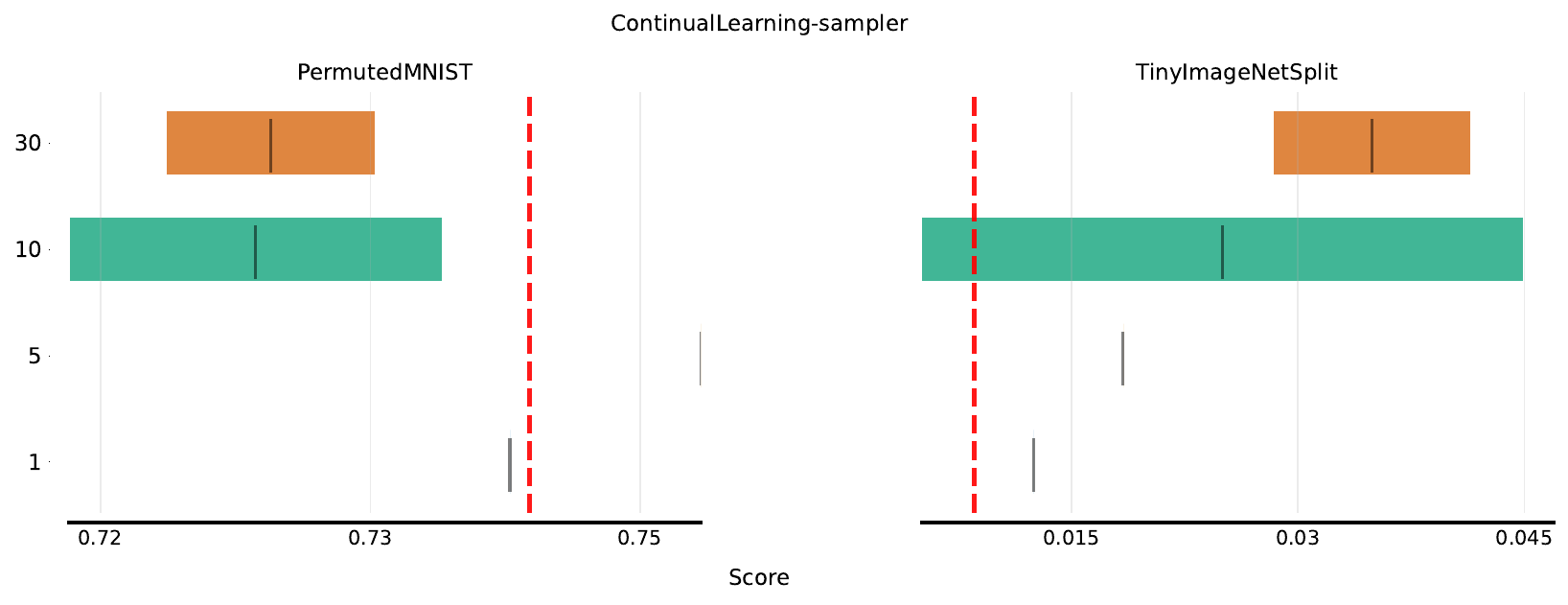}%
\hfill%
\includegraphics[width=0.48\textwidth]{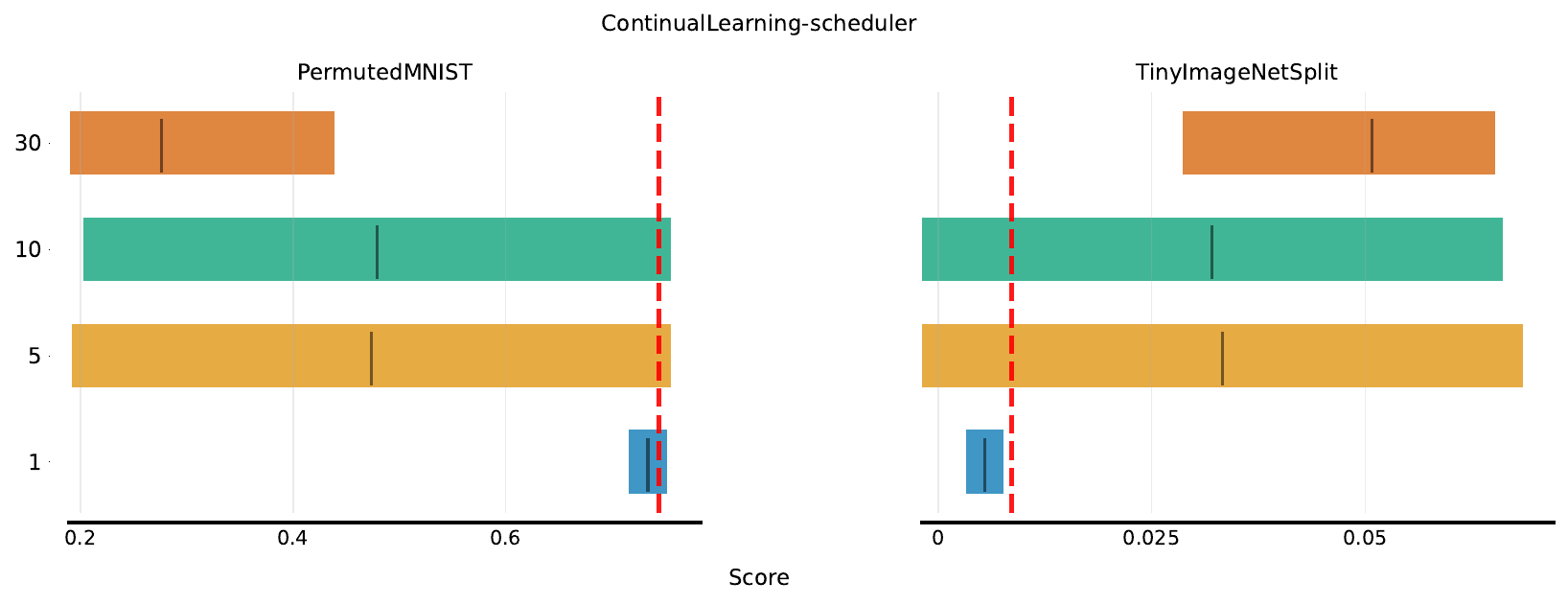}%
\caption{ADA Optimisation results on Meta-Test tasks. (Part 3/8)}
\label{fig:ADA_optimisation_mt_3}
\end{figure}
\clearpage

\begin{figure}[htbp]
\centering
\setlength{\lineskip}{0pt}
\includegraphics[width=0.48\textwidth]{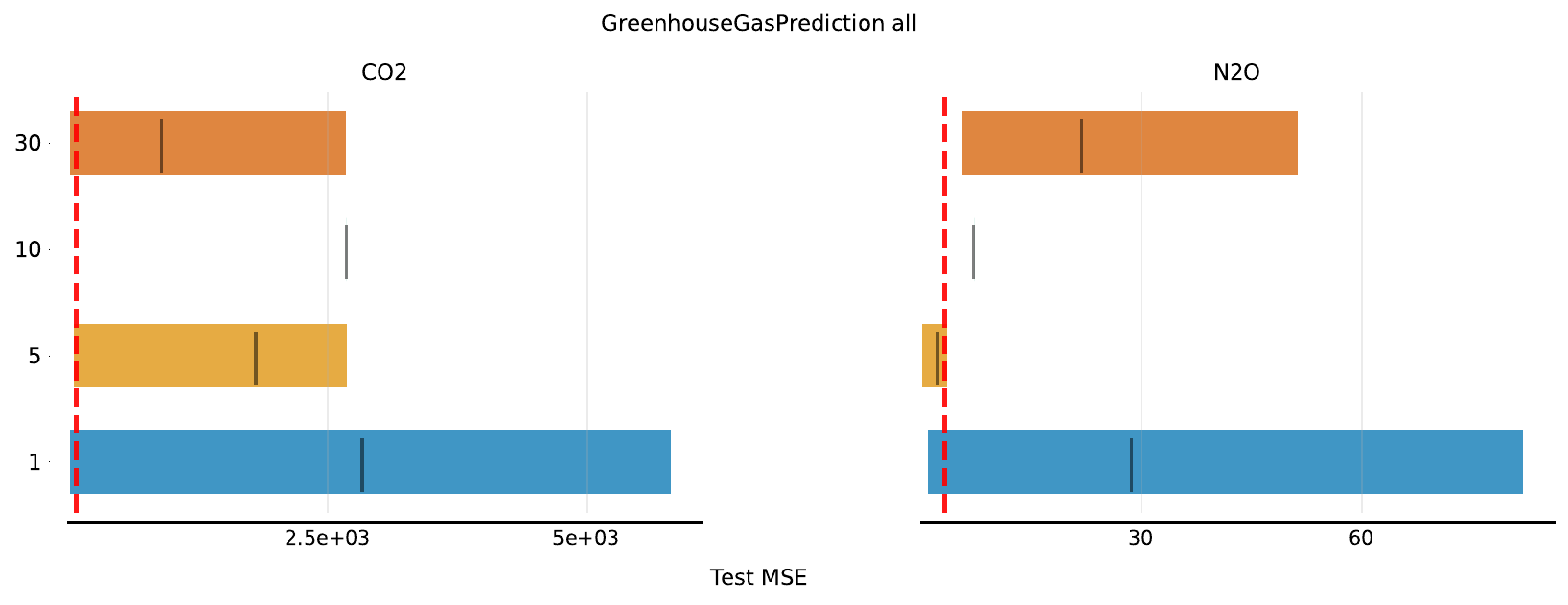}%
\hfill%
\includegraphics[width=0.48\textwidth]{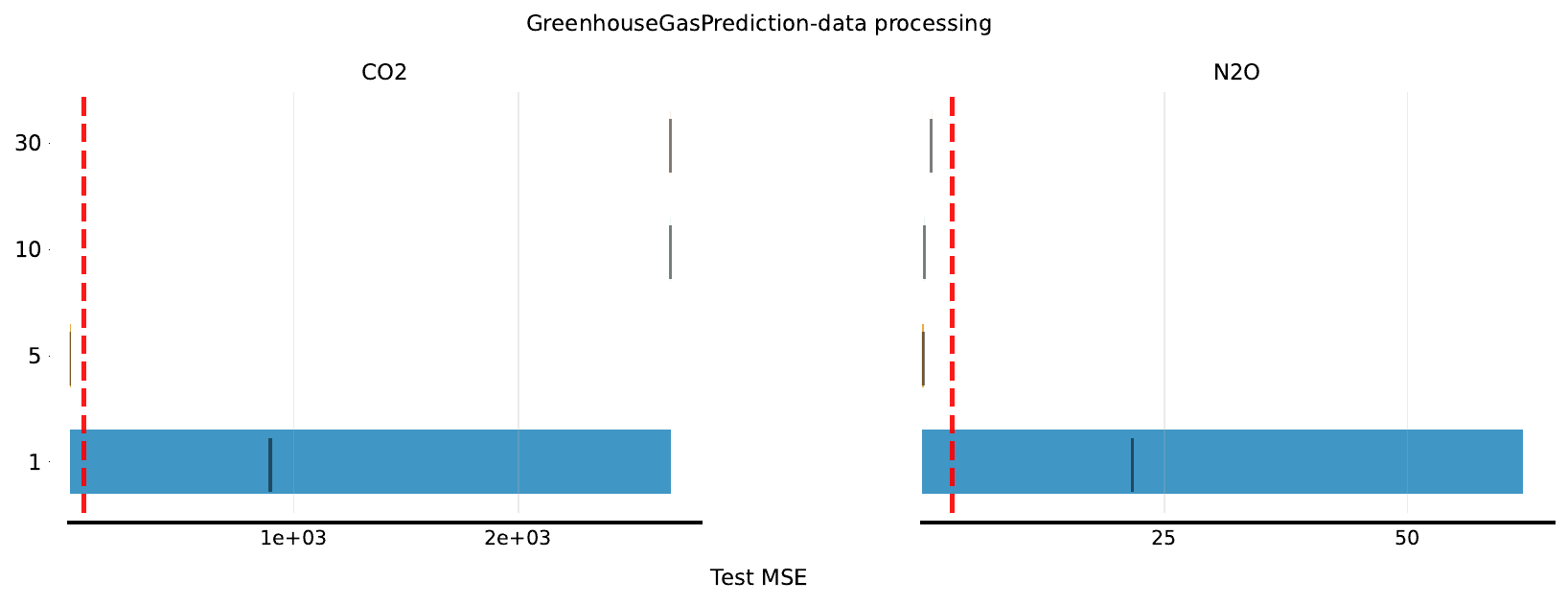}%
\\[0.5em]
\includegraphics[width=0.48\textwidth]{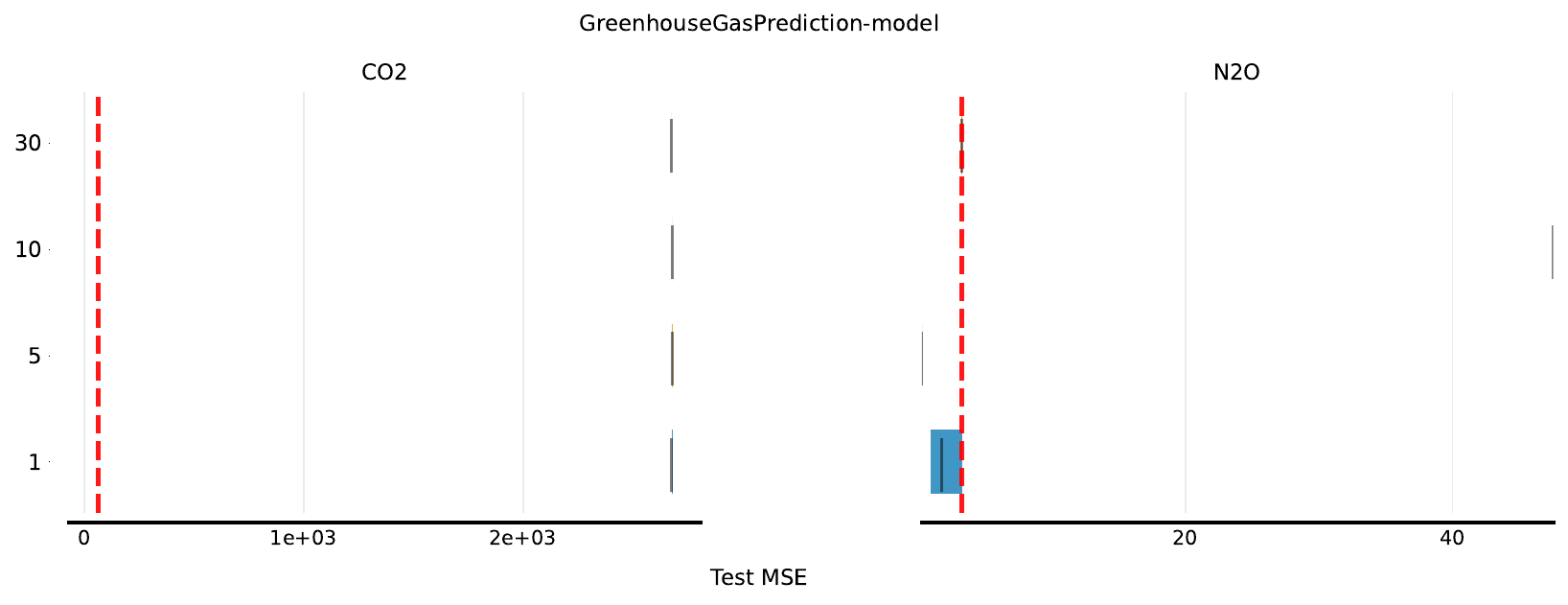}%
\hfill%
\includegraphics[width=0.48\textwidth]{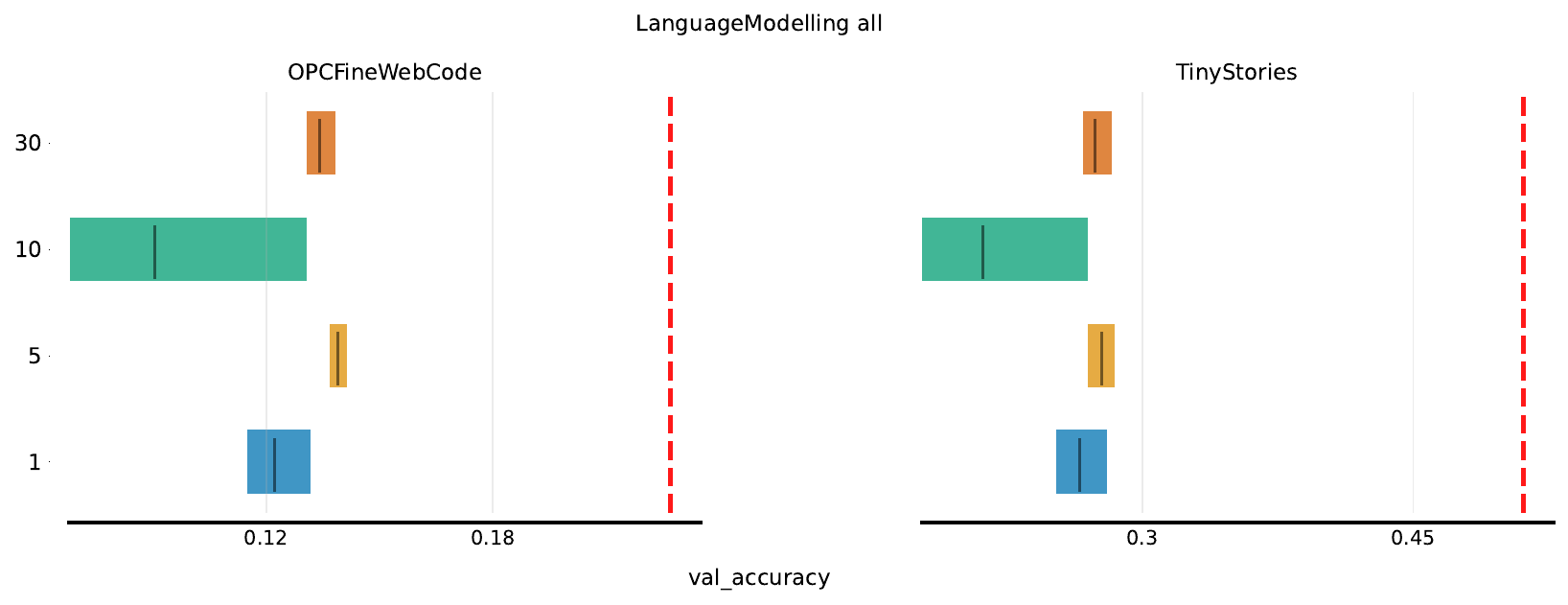}%
\\[0.5em]
\includegraphics[width=0.48\textwidth]{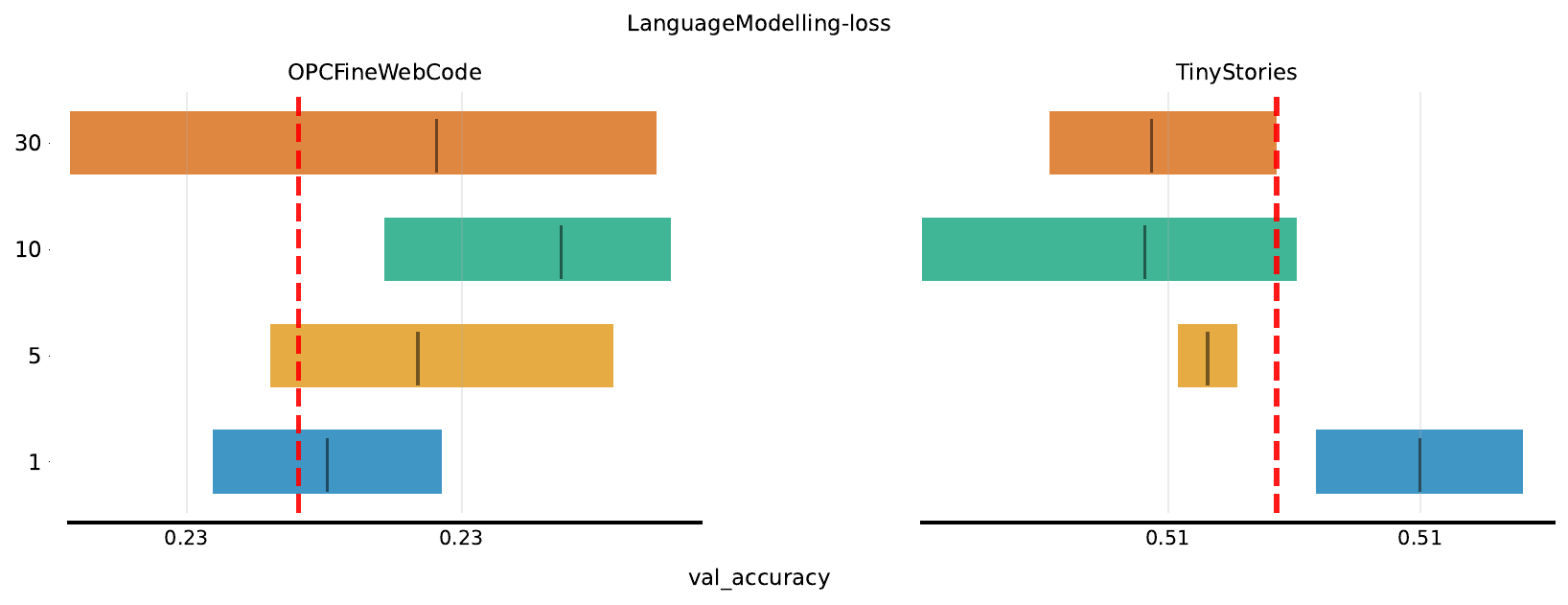}%
\hfill%
\includegraphics[width=0.48\textwidth]{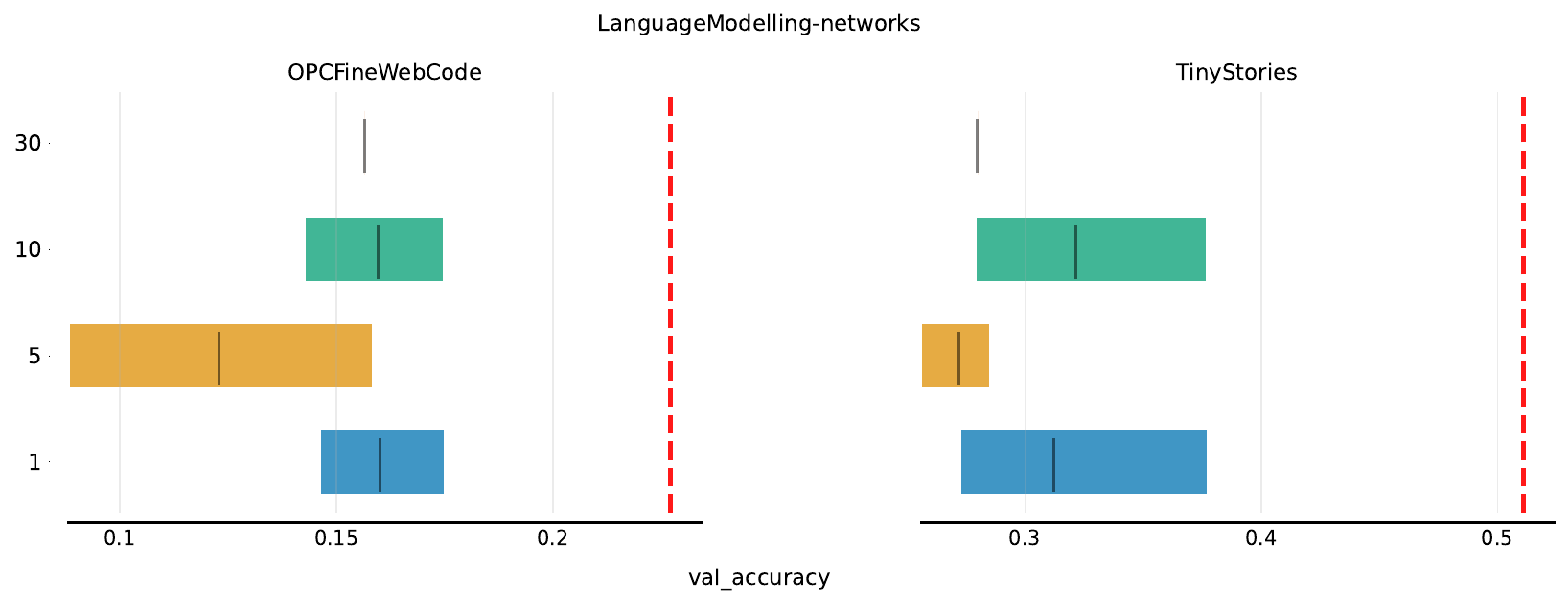}%
\\[0.5em]
\includegraphics[width=0.48\textwidth]{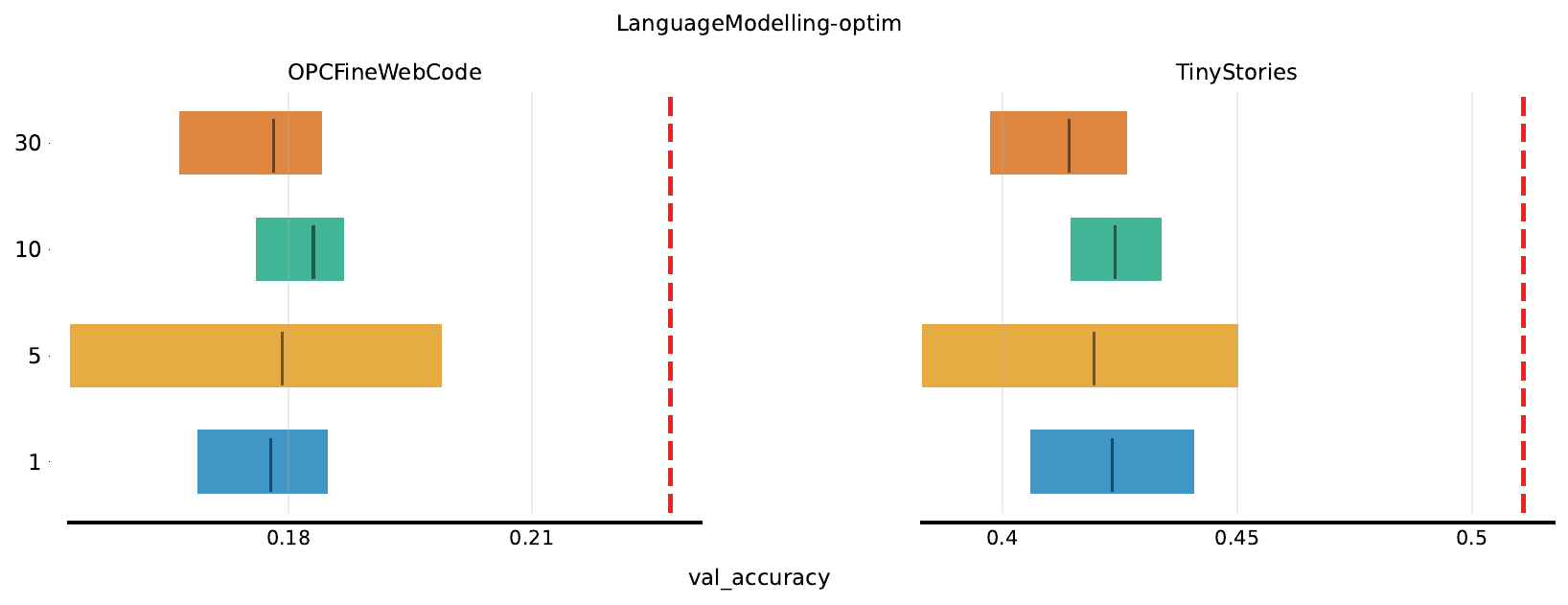}%
\hfill%
\includegraphics[width=0.48\textwidth]{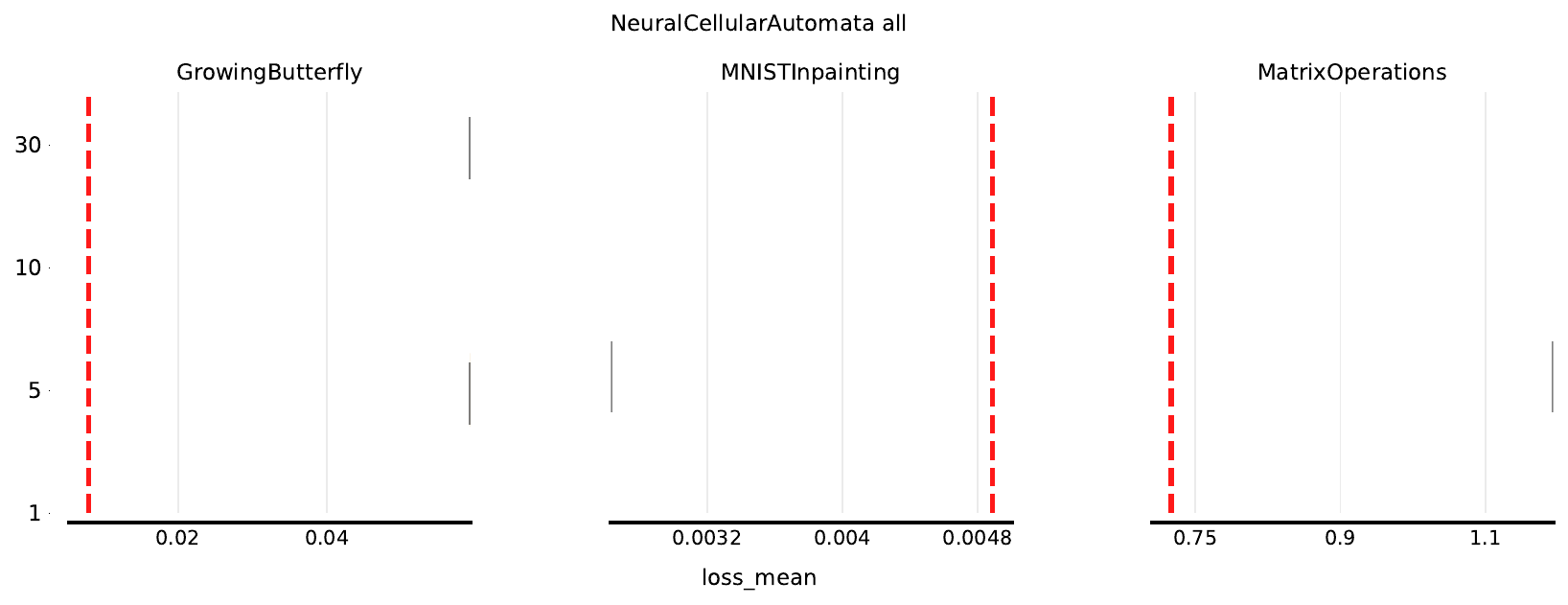}%
\\[0.5em]
\includegraphics[width=0.48\textwidth]{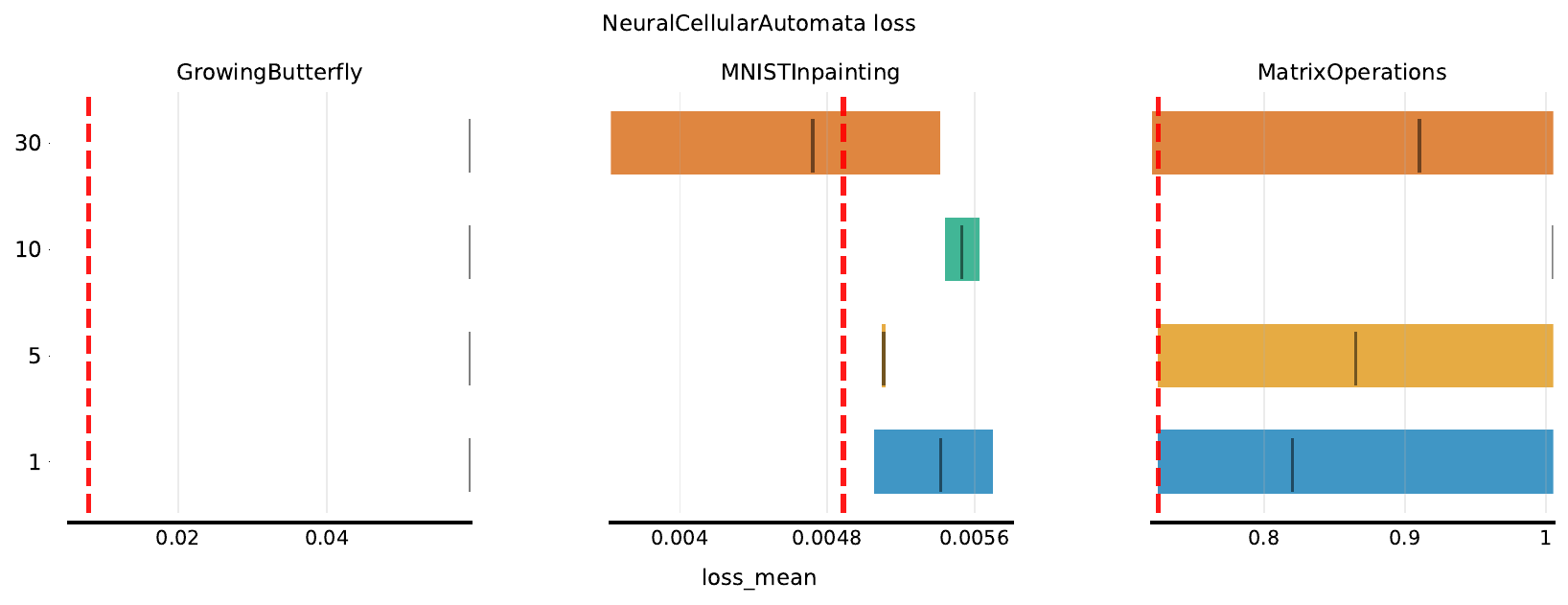}%
\hfill%
\includegraphics[width=0.48\textwidth]{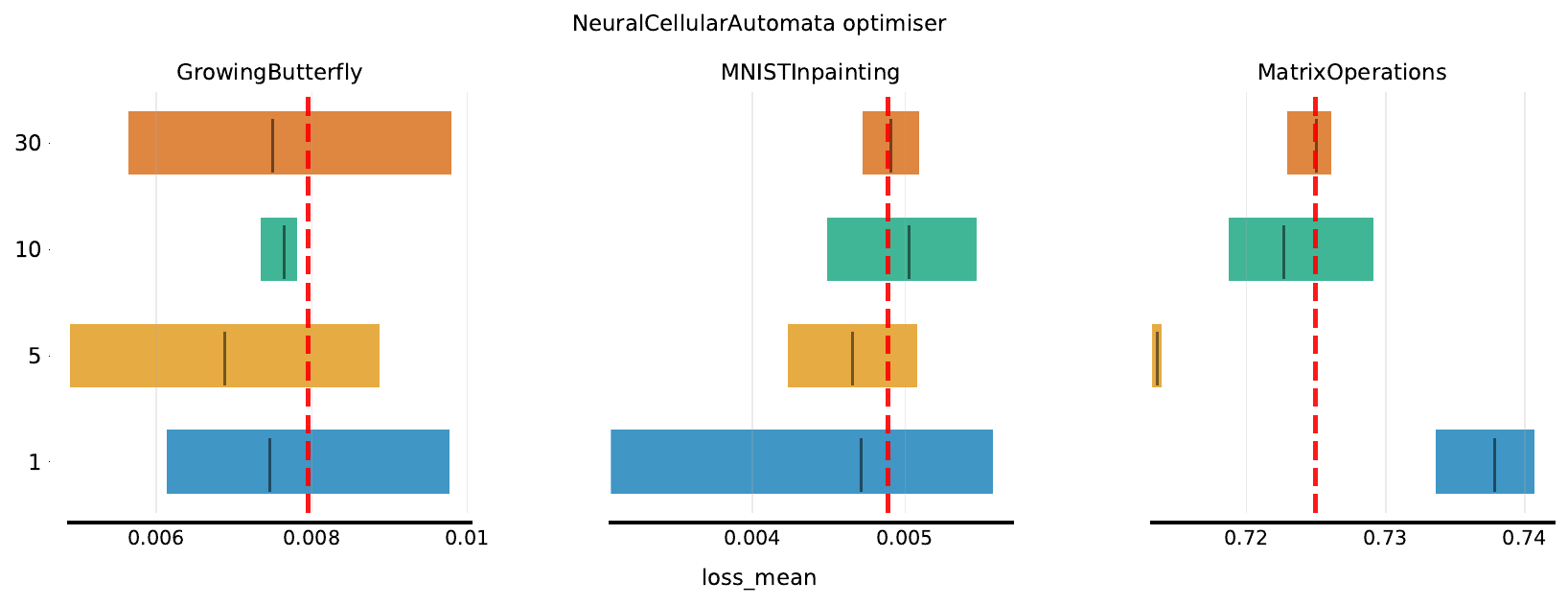}%
\\[0.5em]
\includegraphics[width=0.48\textwidth]{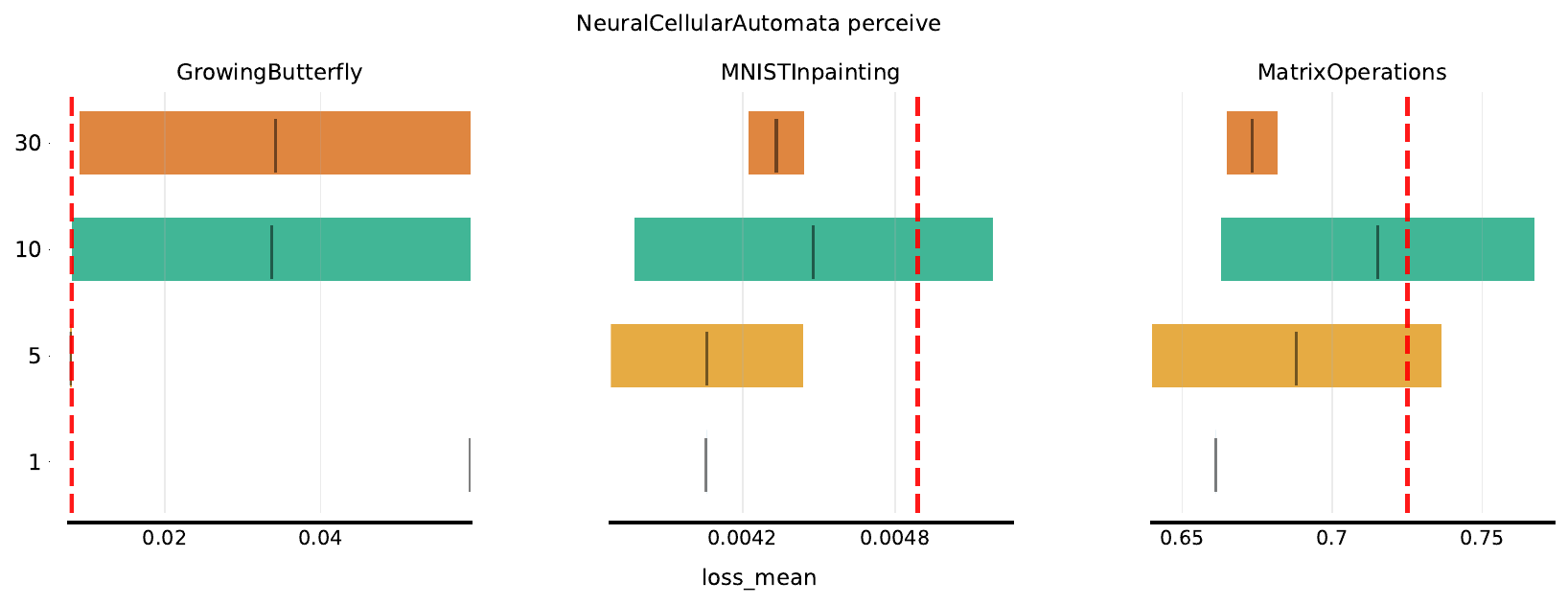}%
\hfill%
\includegraphics[width=0.48\textwidth]{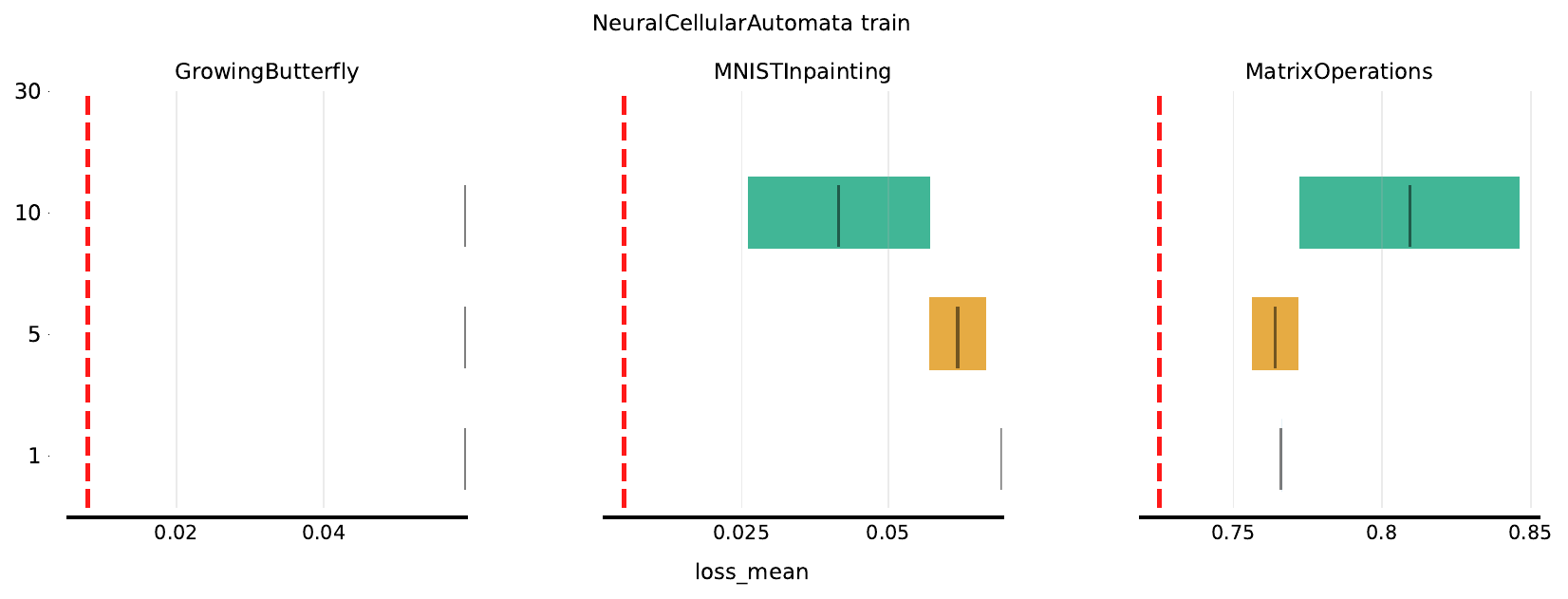}%
\caption{ADA Optimisation results on Meta-Test tasks. (Part 4/8)}
\label{fig:ADA_optimisation_mt_4}
\end{figure}
\clearpage

\begin{figure}[htbp]
\centering
\setlength{\lineskip}{0pt}
\includegraphics[width=0.48\textwidth]{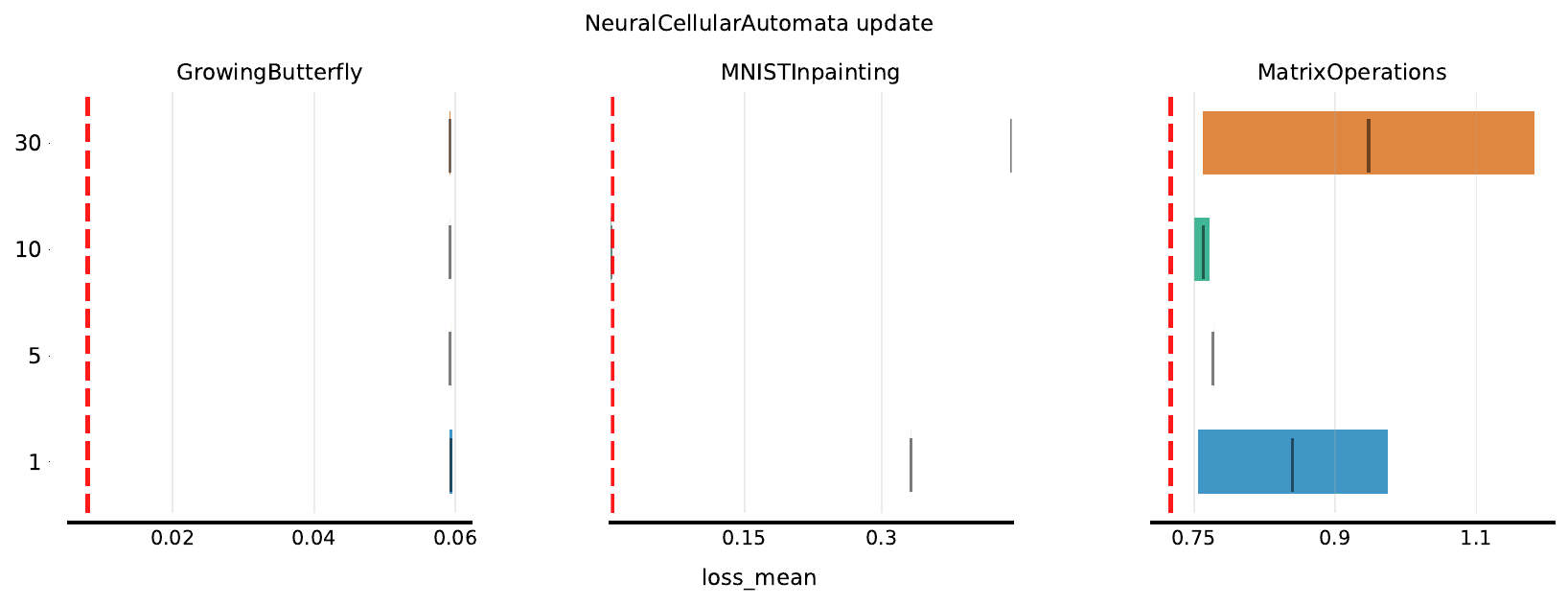}%
\hfill%
\includegraphics[width=0.48\textwidth]{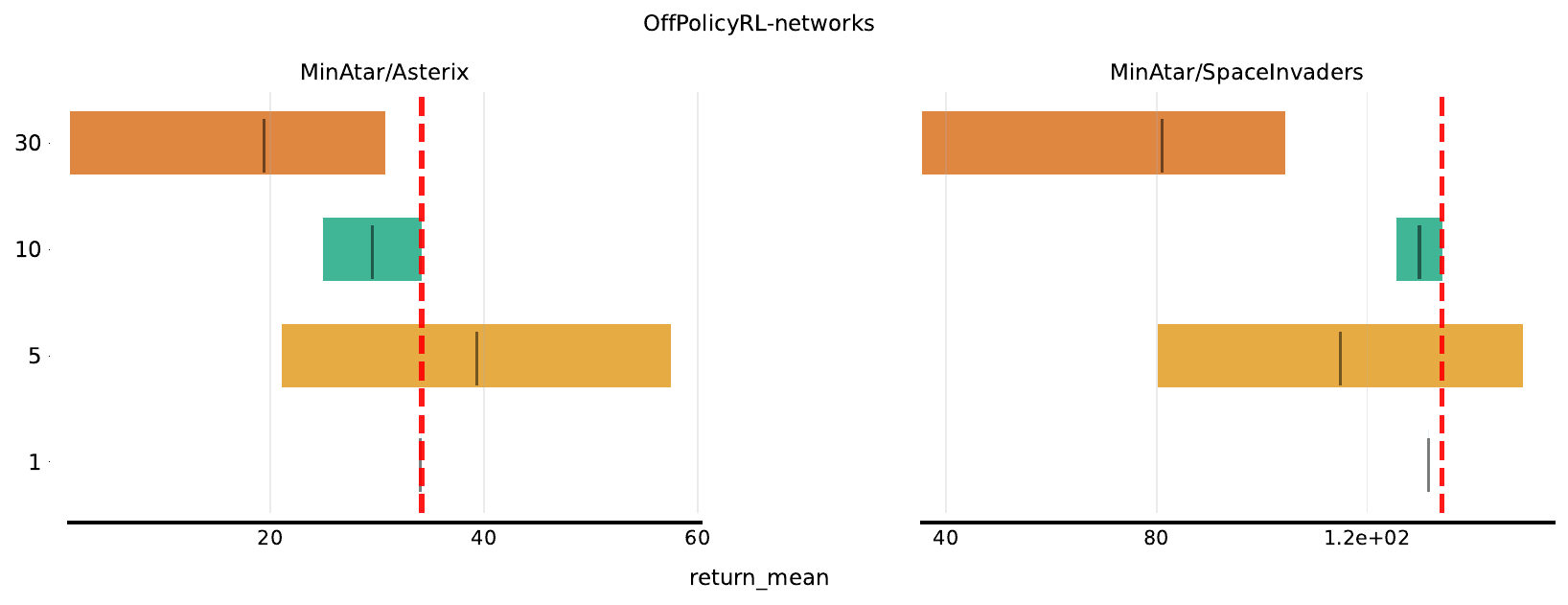}%
\\[0.5em]
\includegraphics[width=0.48\textwidth]{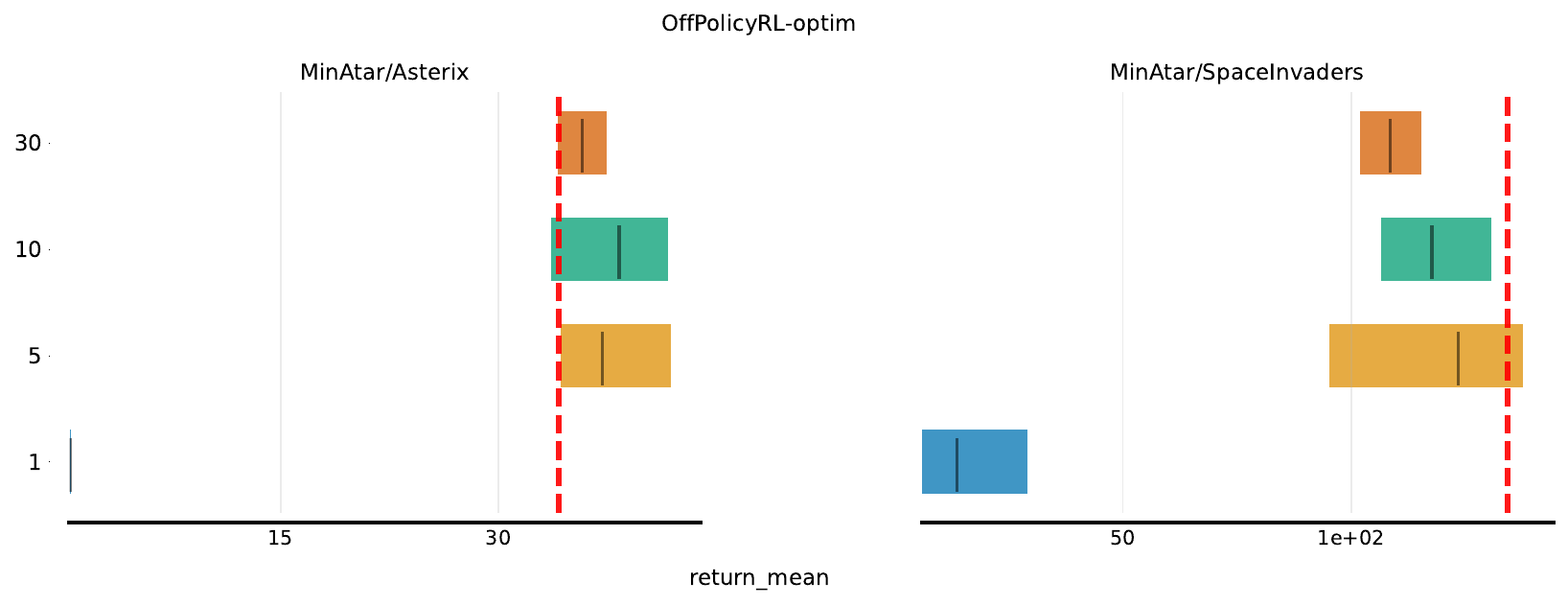}%
\hfill%
\includegraphics[width=0.48\textwidth]{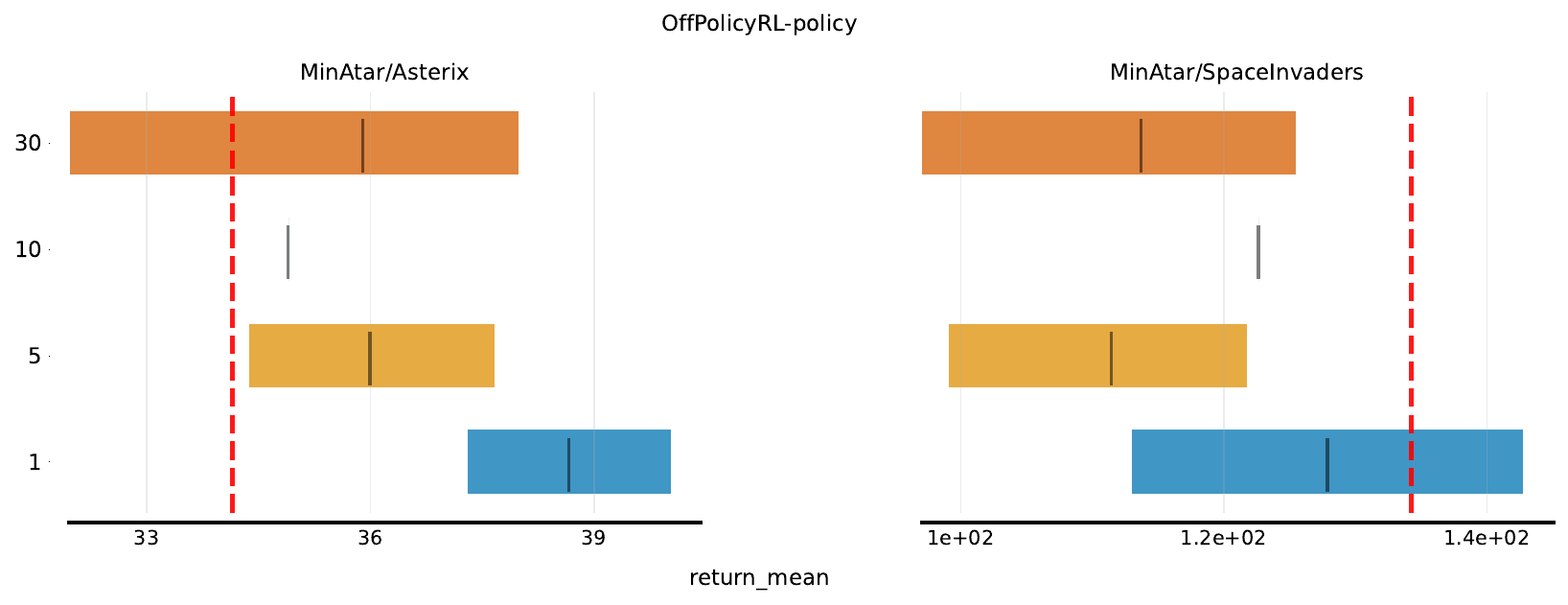}%
\\[0.5em]
\includegraphics[width=0.48\textwidth]{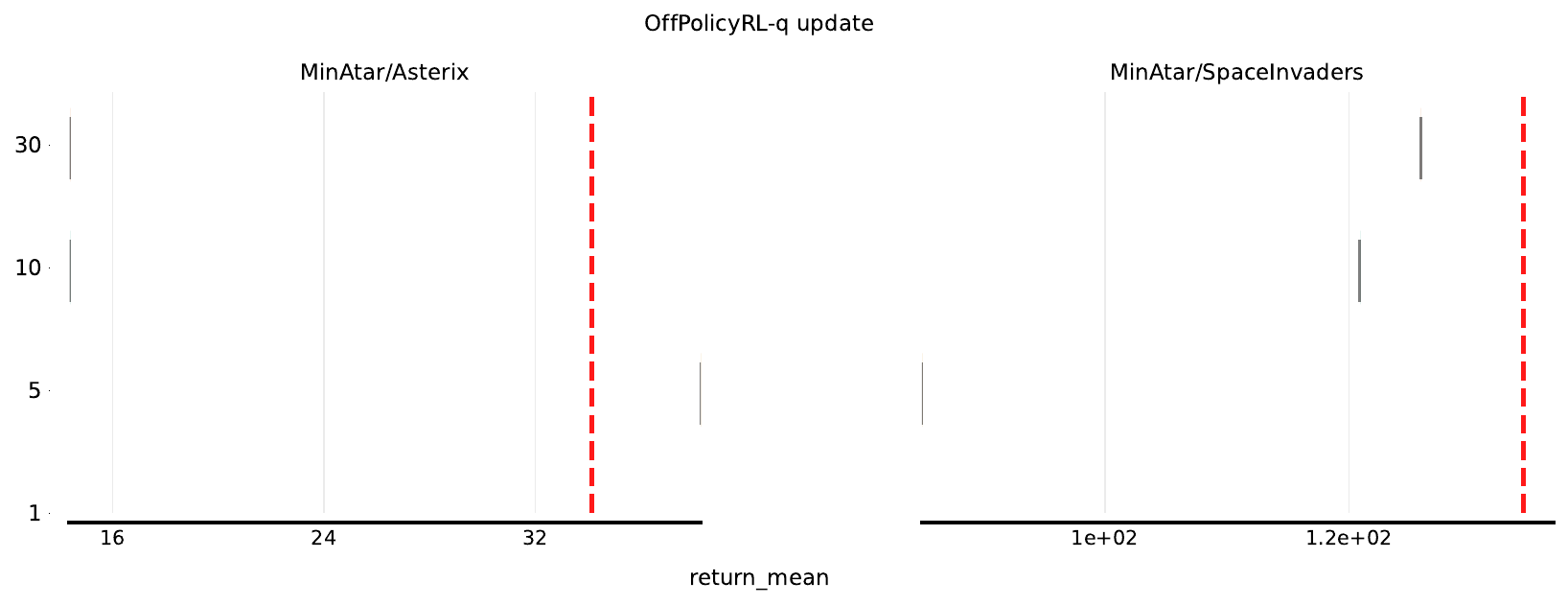}%
\hfill%
\includegraphics[width=0.48\textwidth]{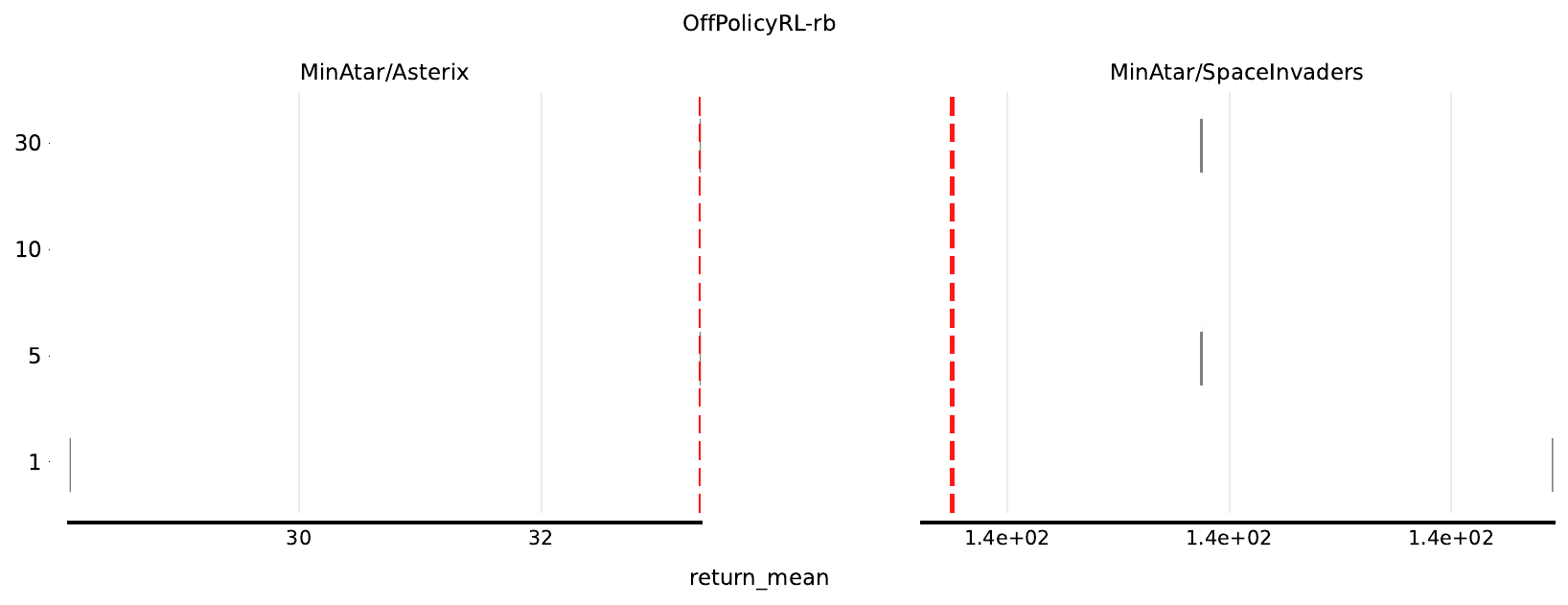}%
\\[0.5em]
\includegraphics[width=0.48\textwidth]{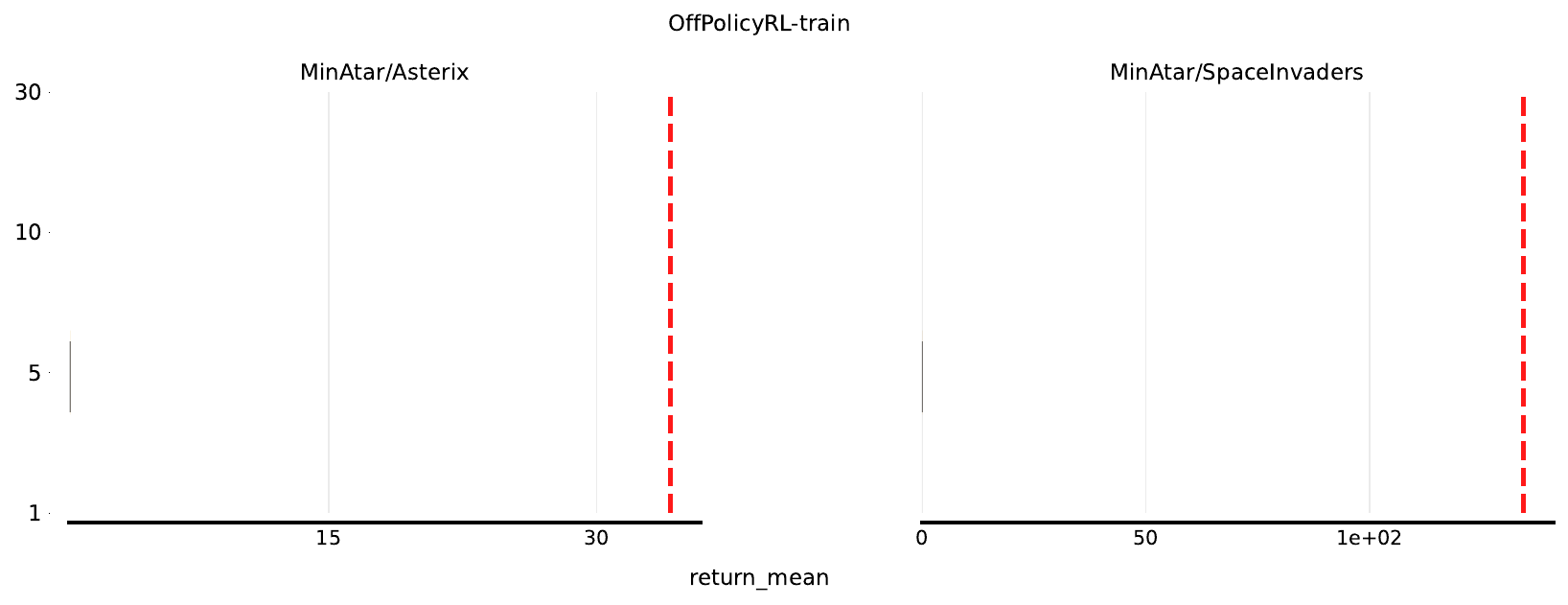}%
\hfill%
\includegraphics[width=0.48\textwidth]{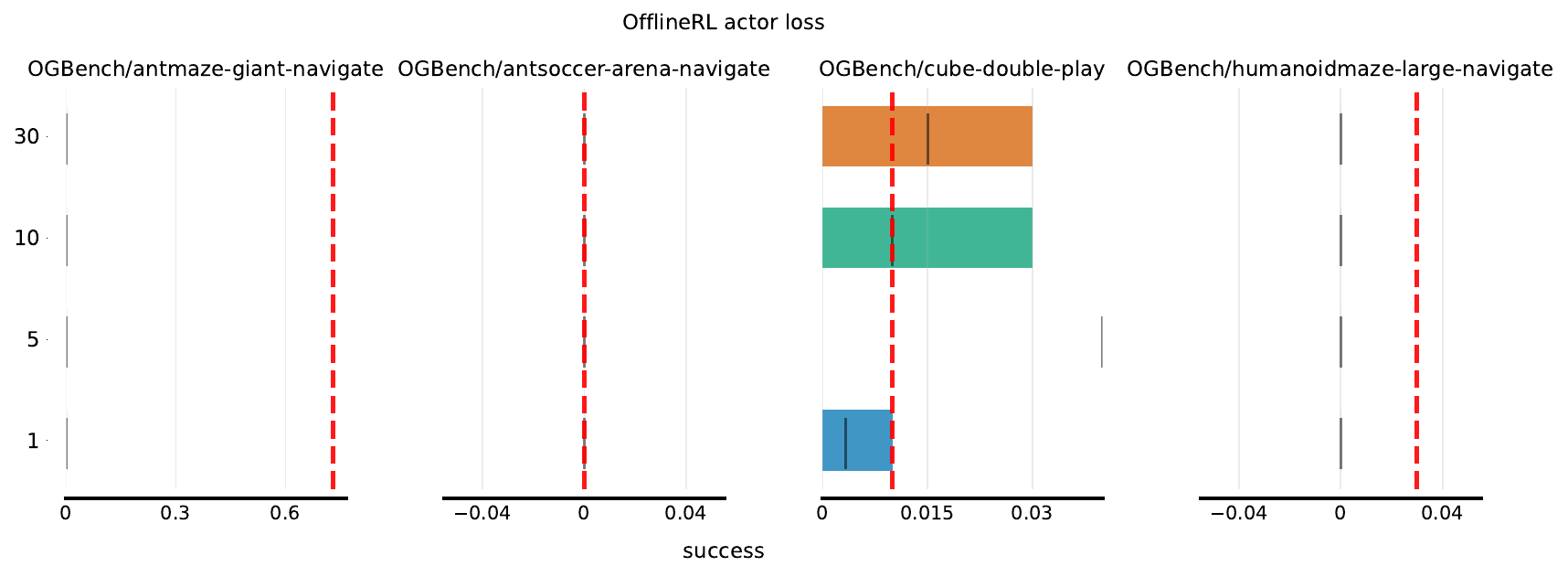}%
\\[0.5em]
\includegraphics[width=0.48\textwidth]{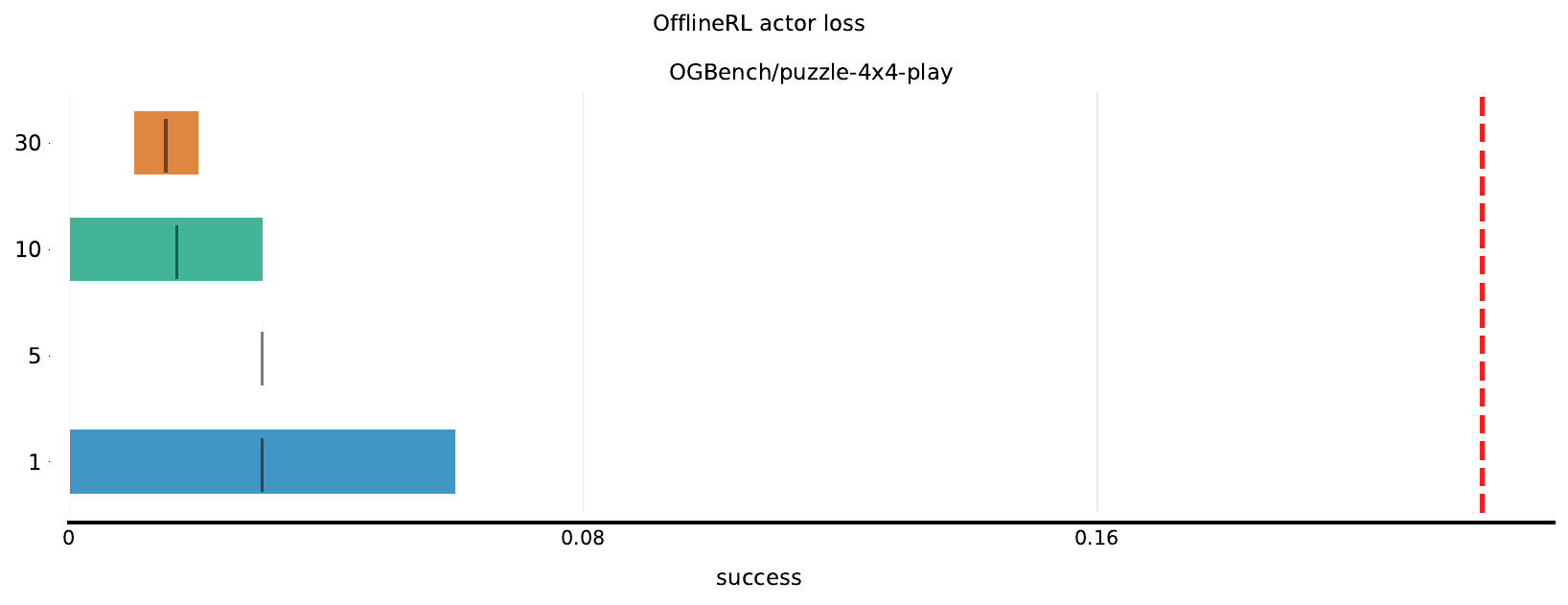}%
\hfill%
\includegraphics[width=0.48\textwidth]{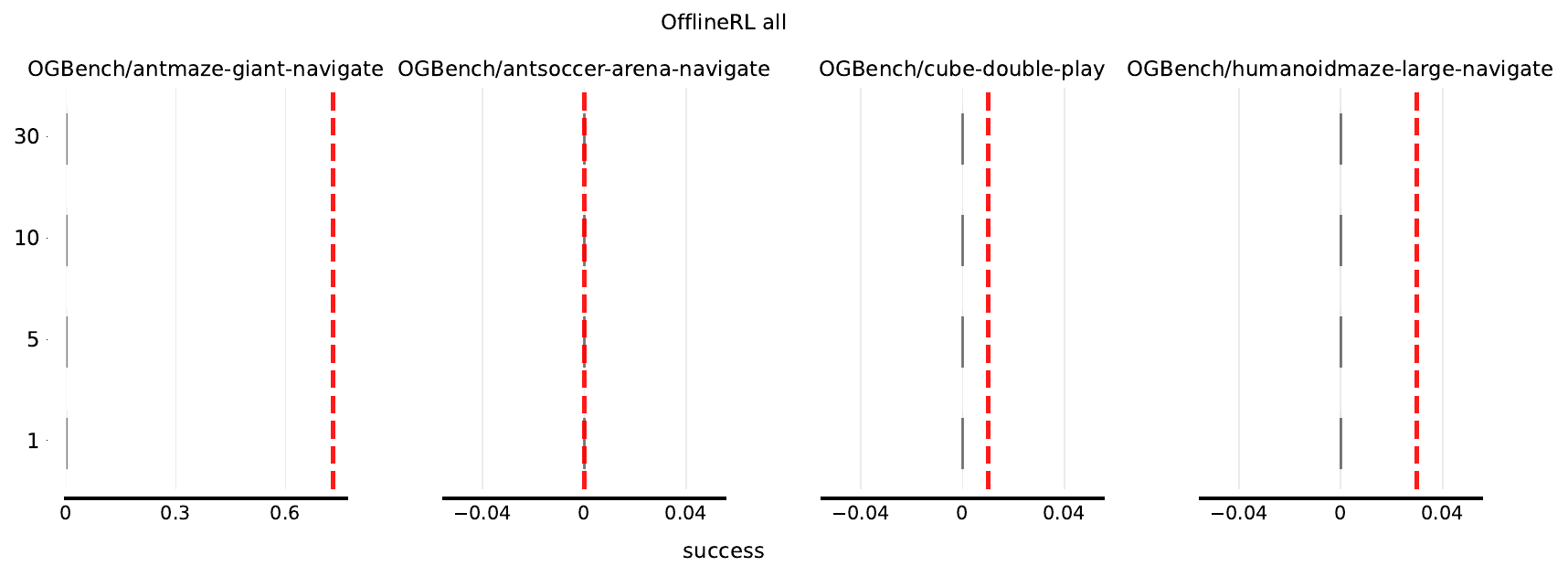}%
\\[0.5em]
\includegraphics[width=0.48\textwidth]{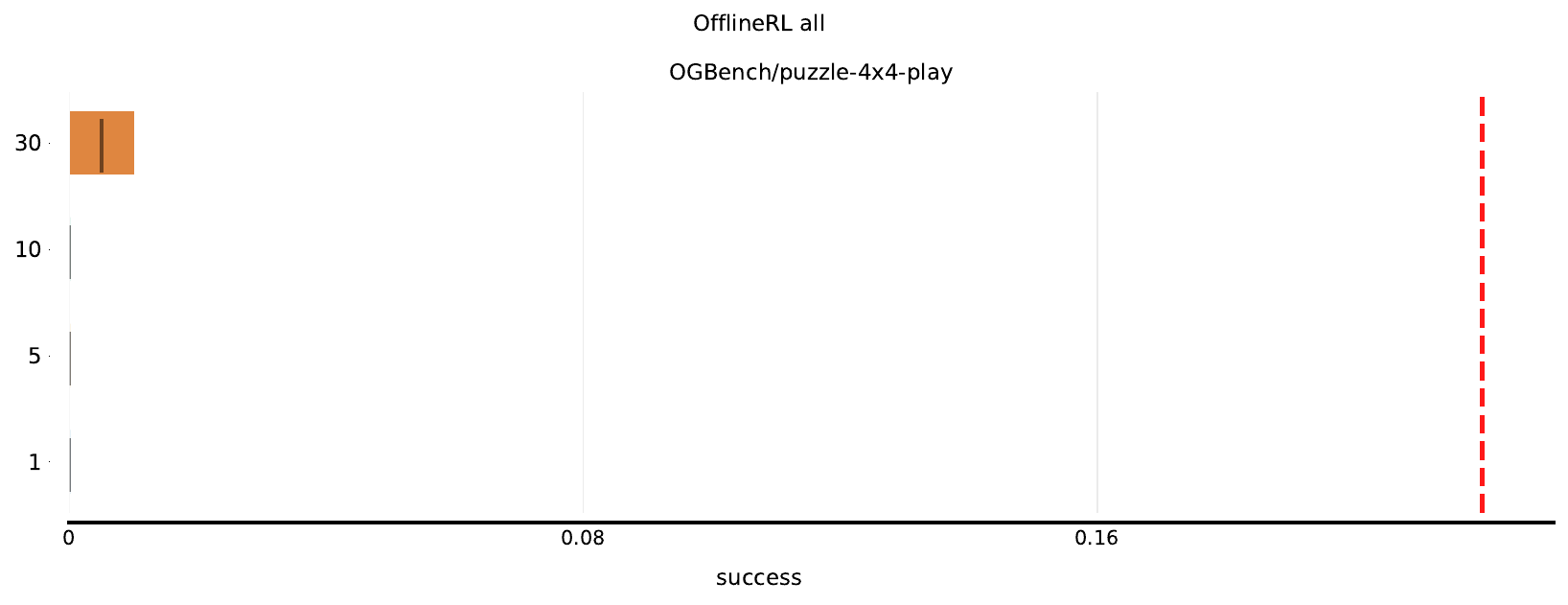}%
\hfill%
\includegraphics[width=0.48\textwidth]{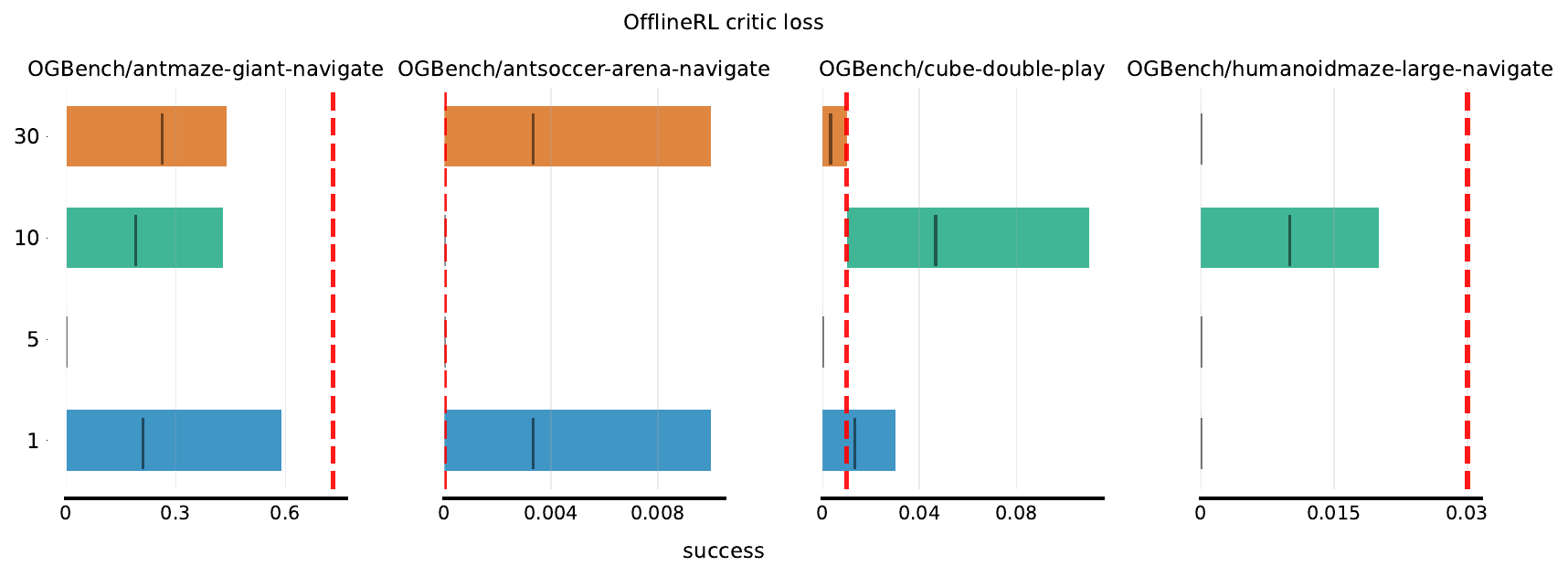}%
\caption{ADA Optimisation results on Meta-Test tasks. (Part 5/8)}
\label{fig:ADA_optimisation_mt_5}
\end{figure}
\clearpage

\begin{figure}[htbp]
\centering
\setlength{\lineskip}{0pt}
\includegraphics[width=0.48\textwidth]{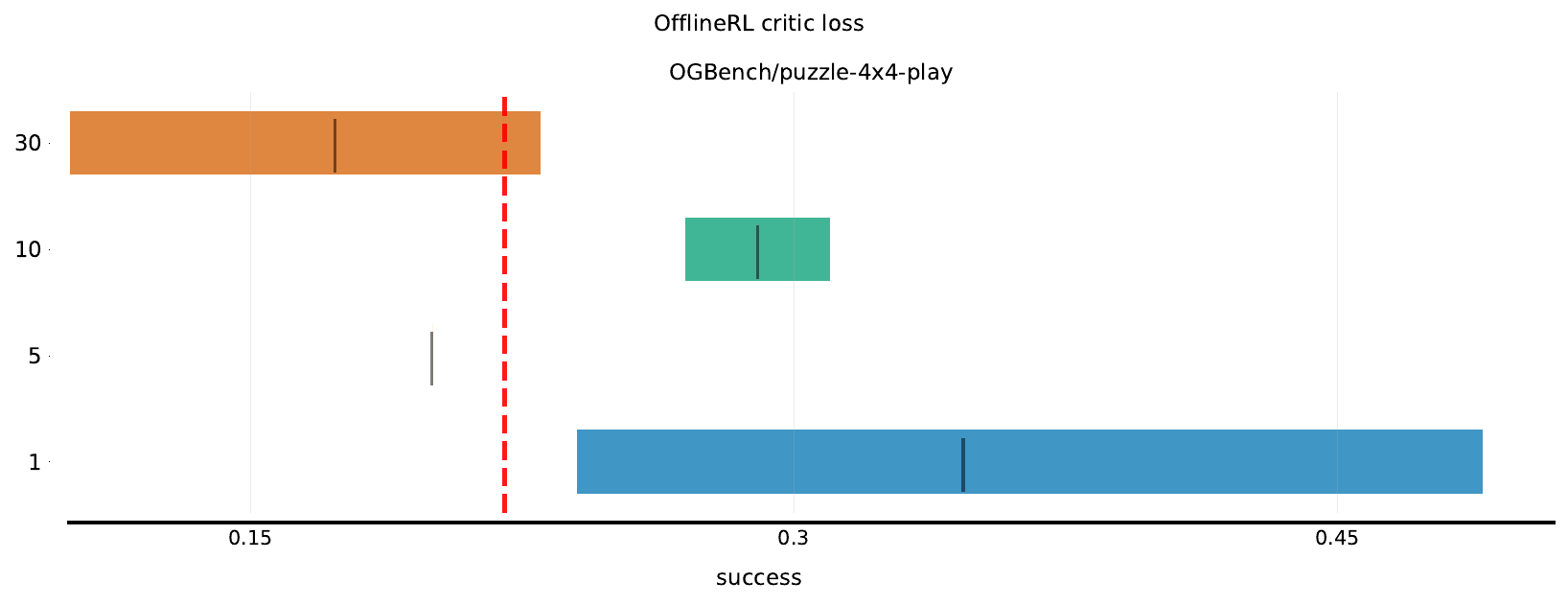}%
\hfill%
\includegraphics[width=0.48\textwidth]{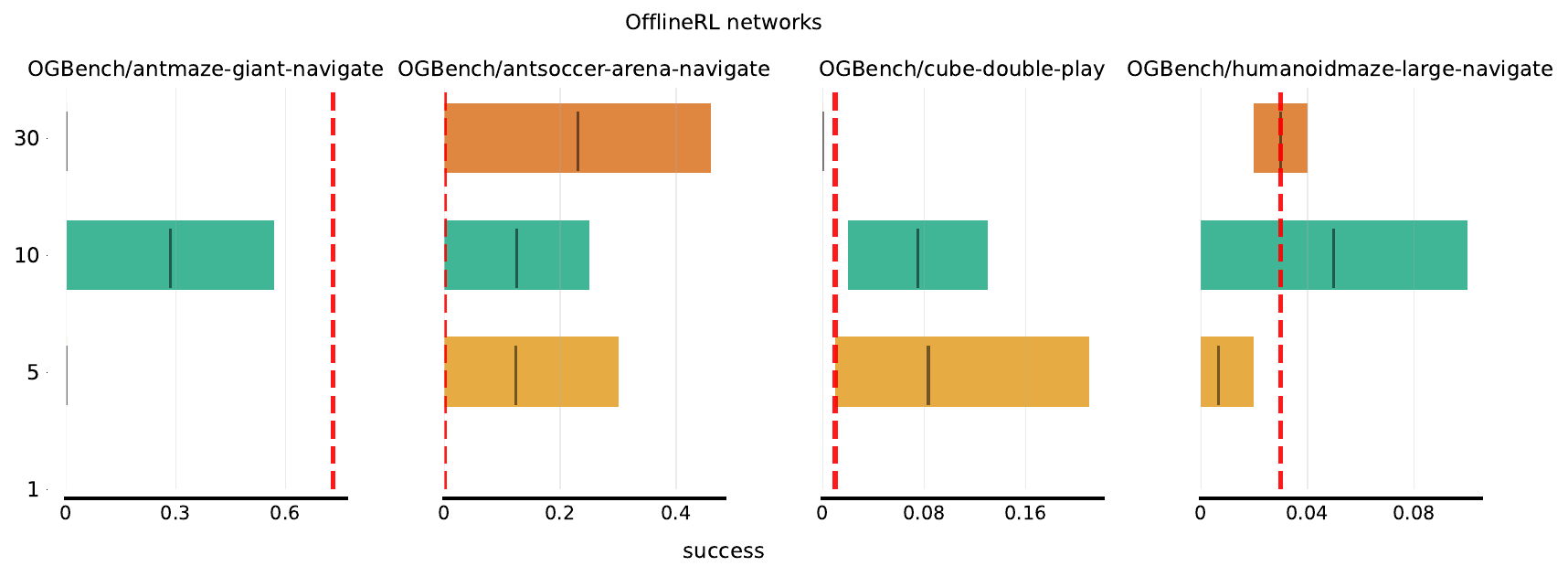}%
\\[0.5em]
\includegraphics[width=0.48\textwidth]{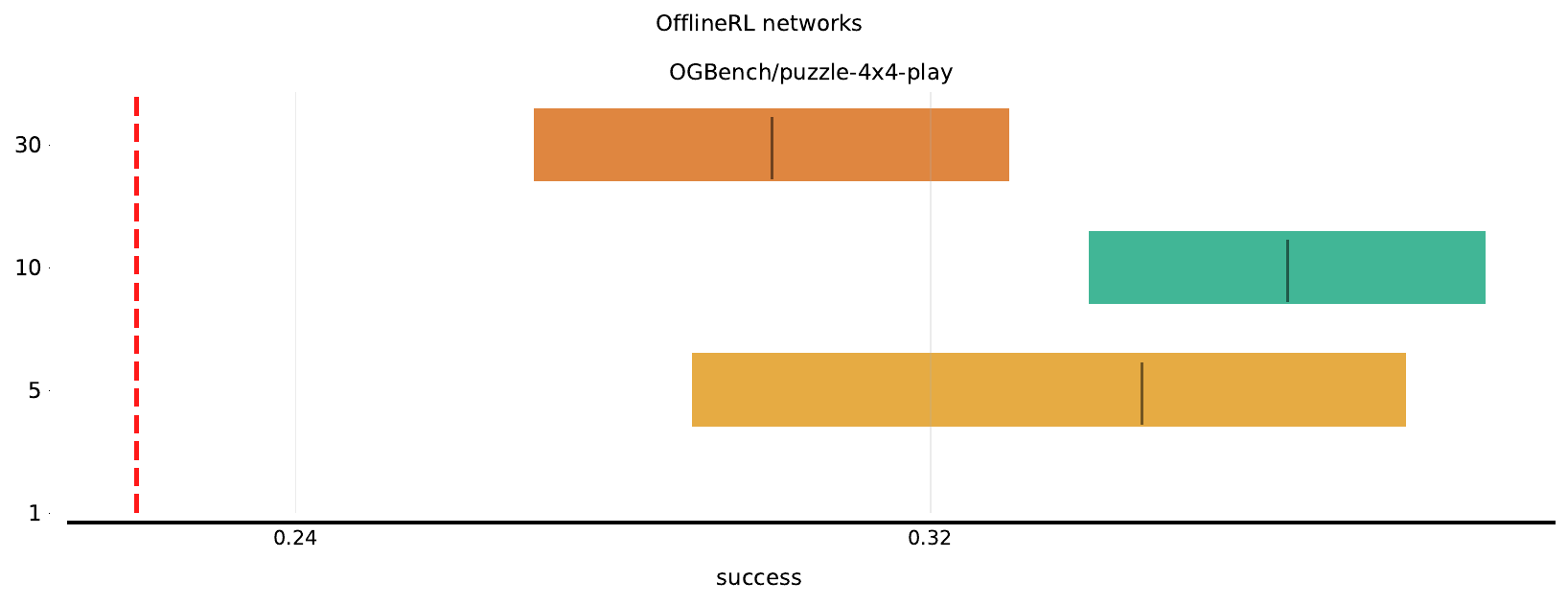}%
\hfill%
\includegraphics[width=0.48\textwidth]{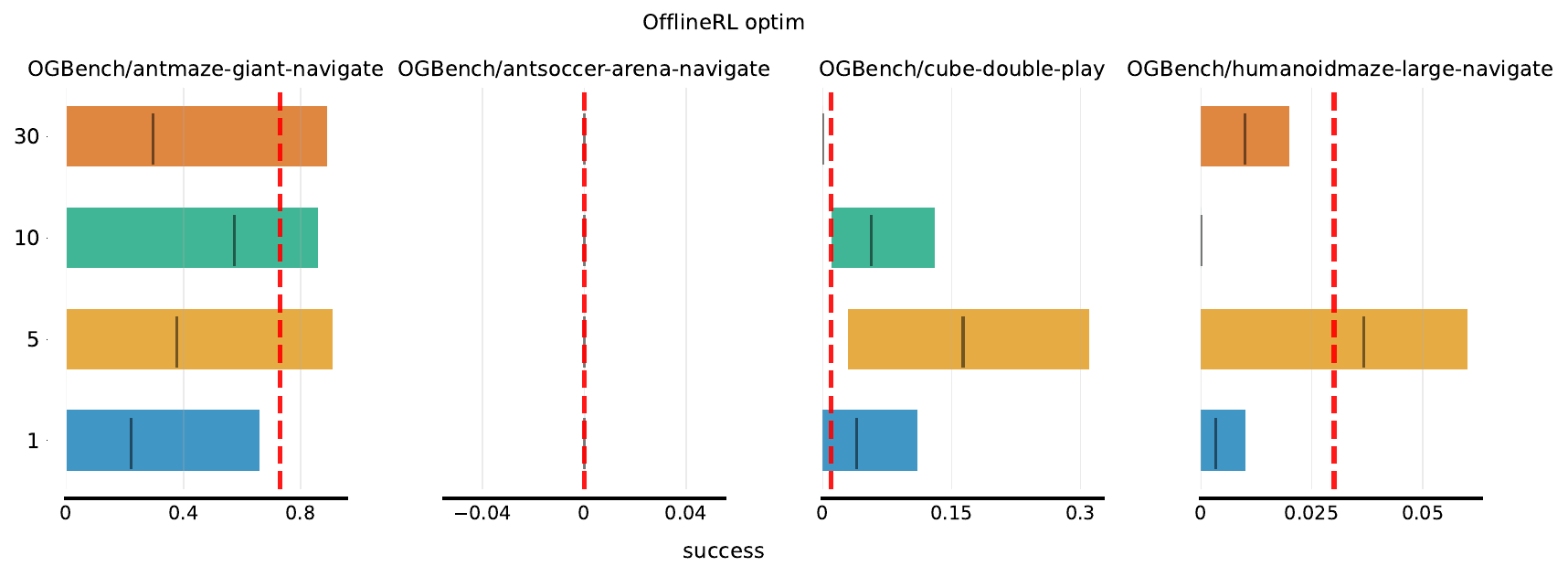}%
\\[0.5em]
\includegraphics[width=0.48\textwidth]{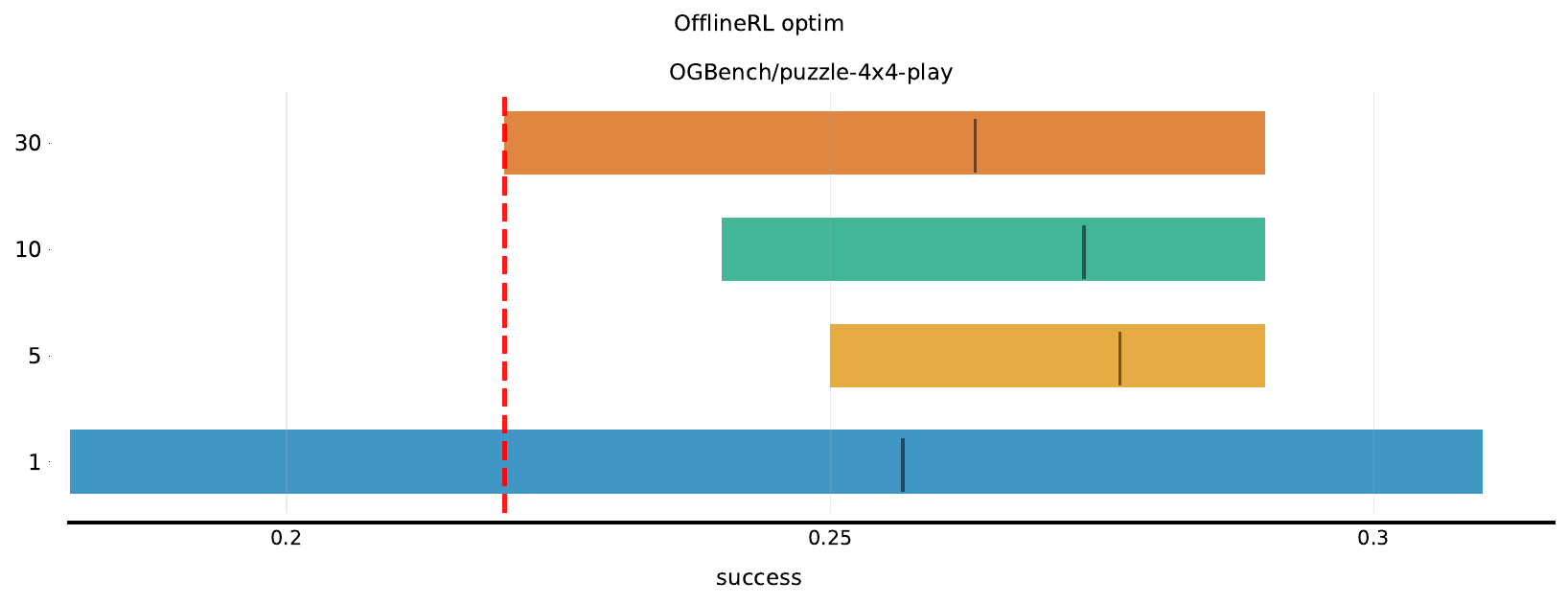}%
\hfill%
\includegraphics[width=0.48\textwidth]{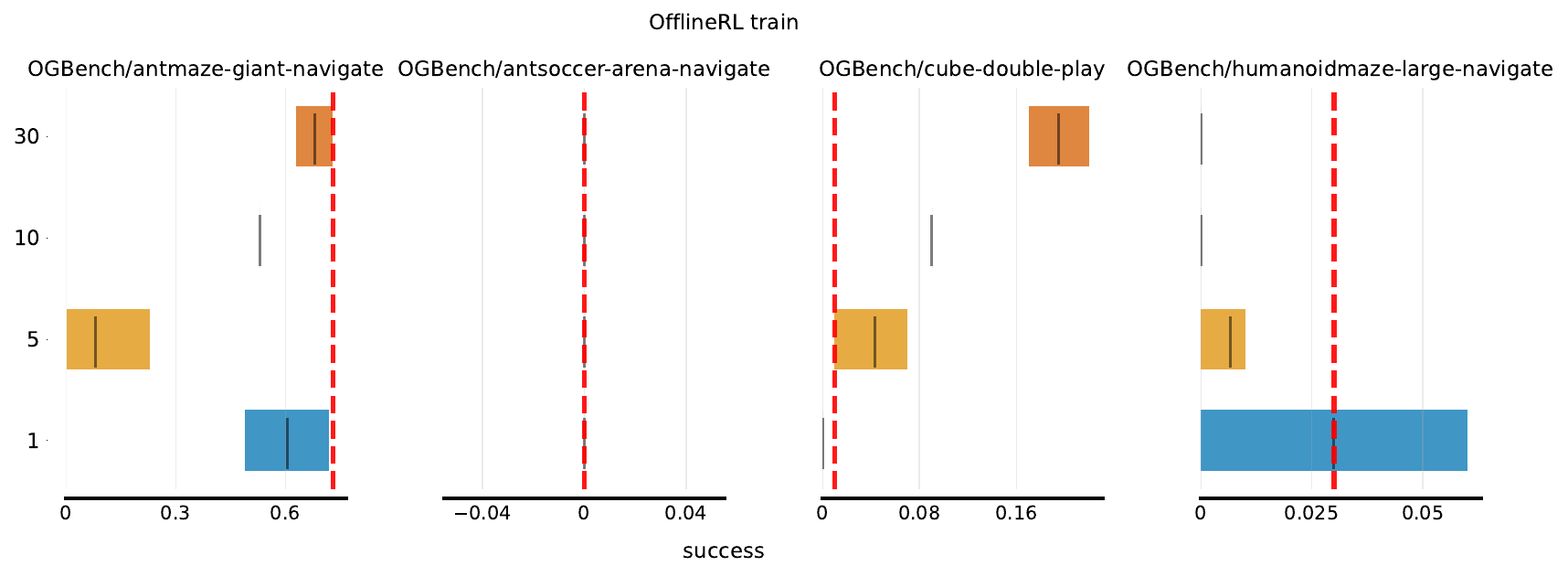}%
\\[0.5em]
\includegraphics[width=0.48\textwidth]{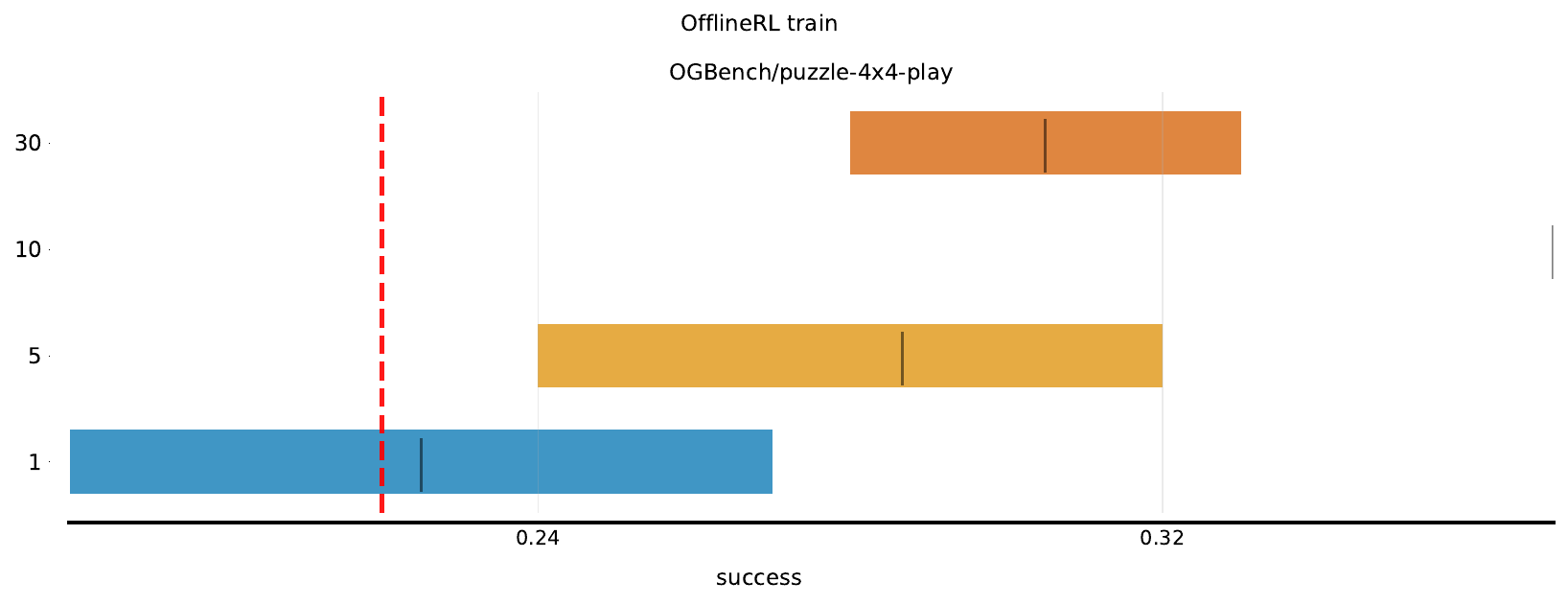}%
\hfill%
\includegraphics[width=0.48\textwidth]{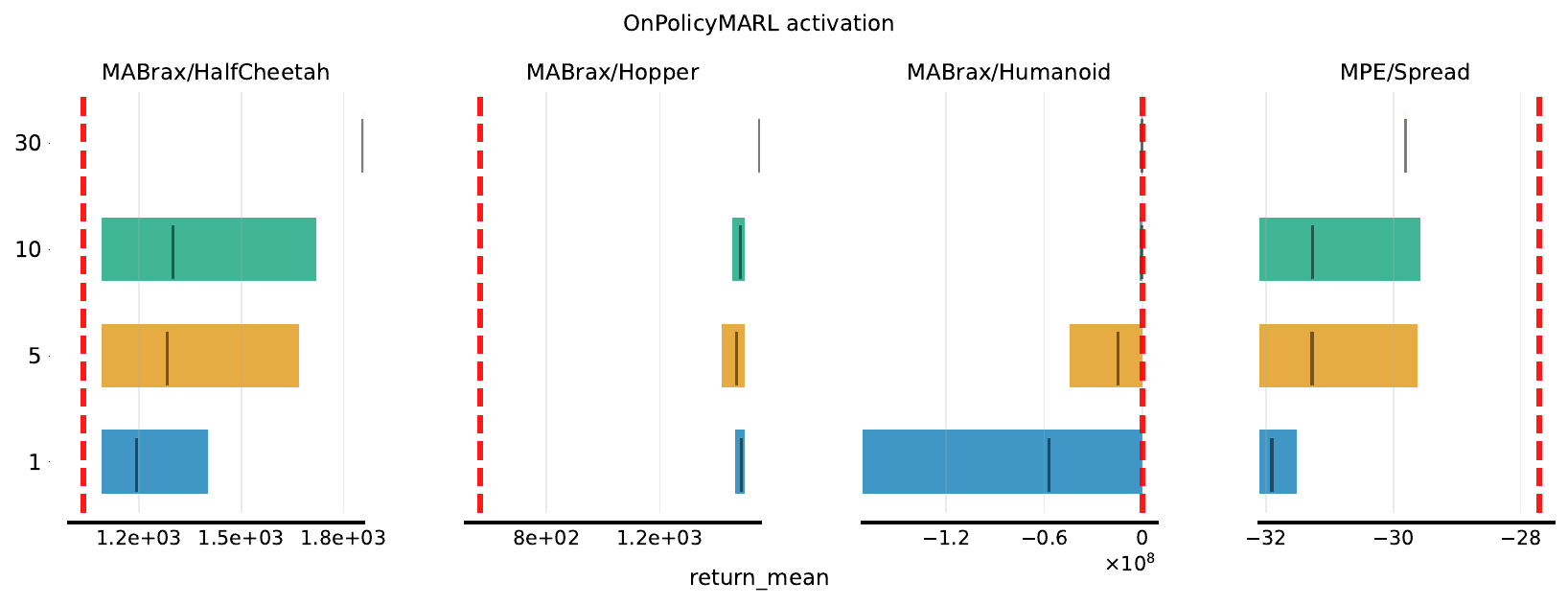}%
\\[0.5em]
\includegraphics[width=0.48\textwidth]{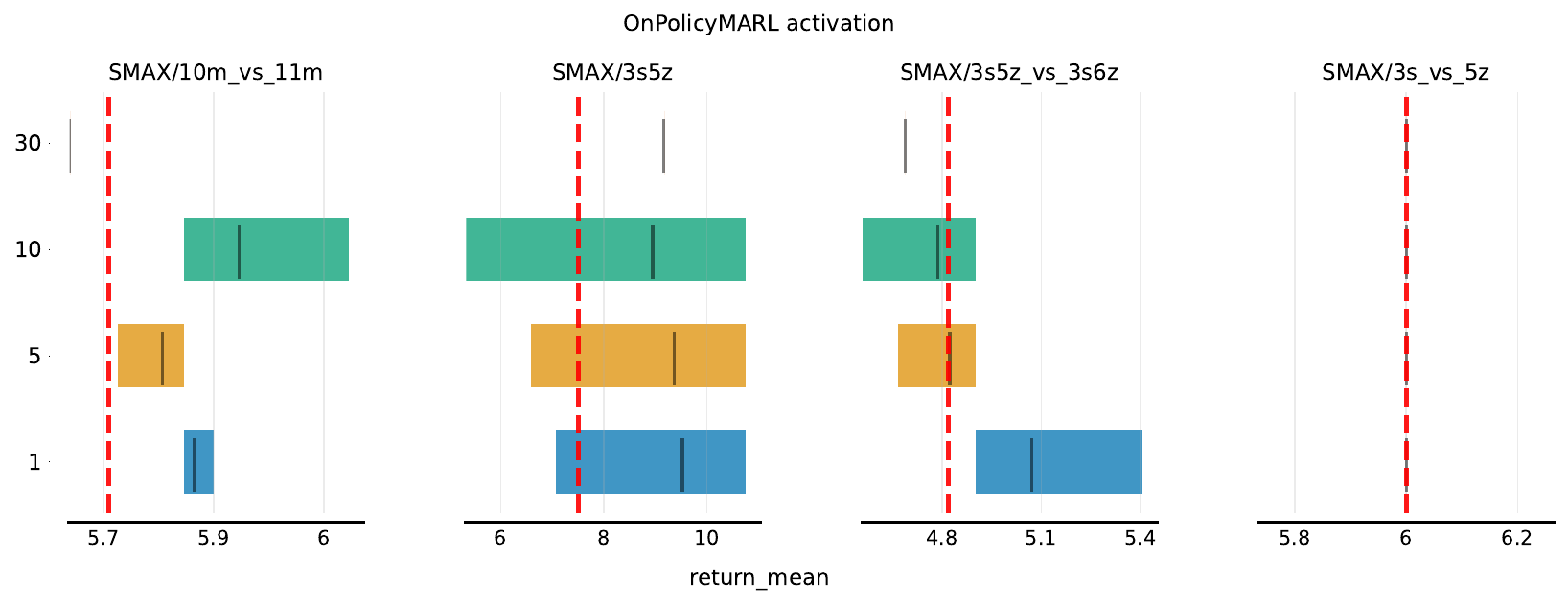}%
\hfill%
\includegraphics[width=0.48\textwidth]{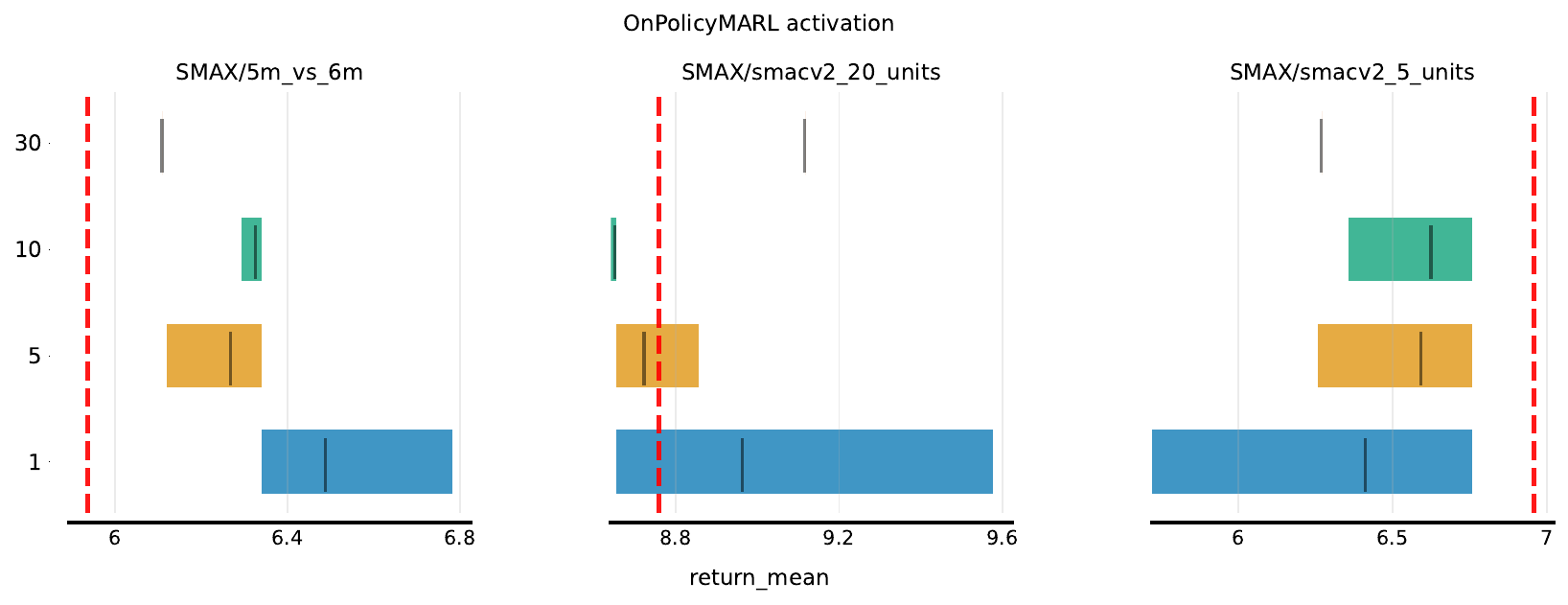}%
\\[0.5em]
\includegraphics[width=0.48\textwidth]{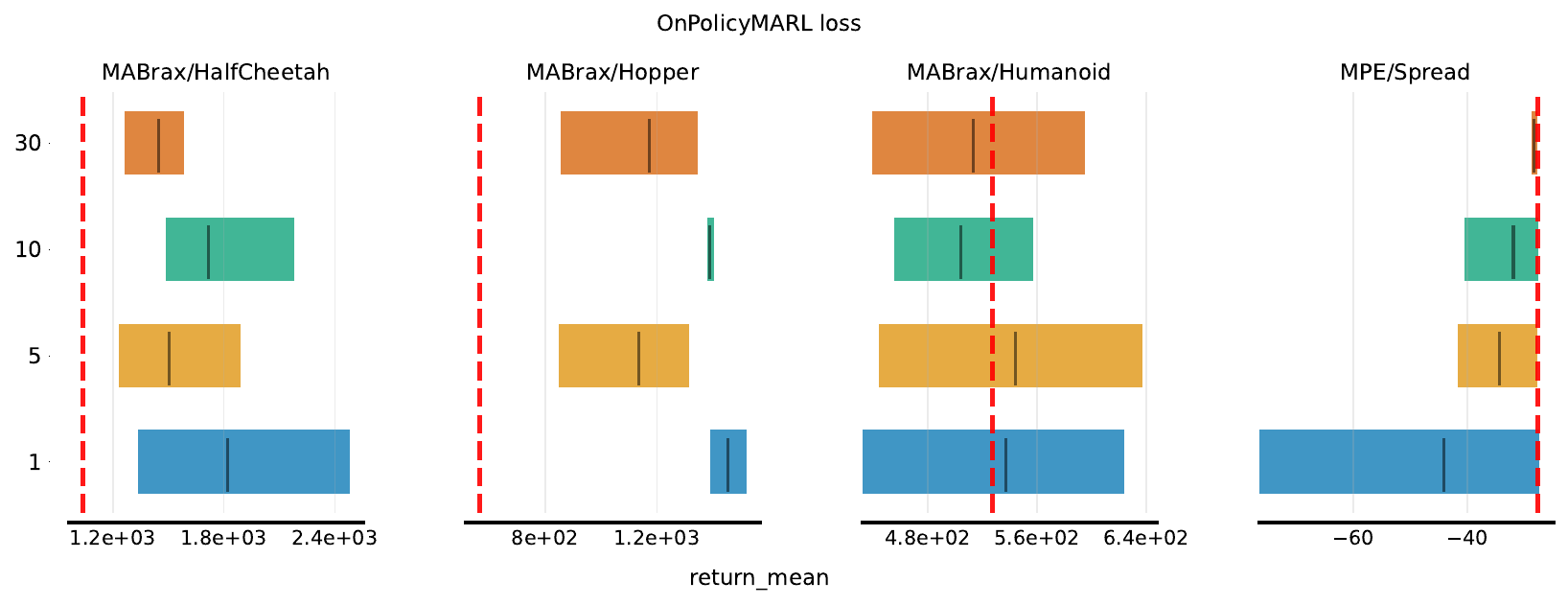}%
\hfill%
\includegraphics[width=0.48\textwidth]{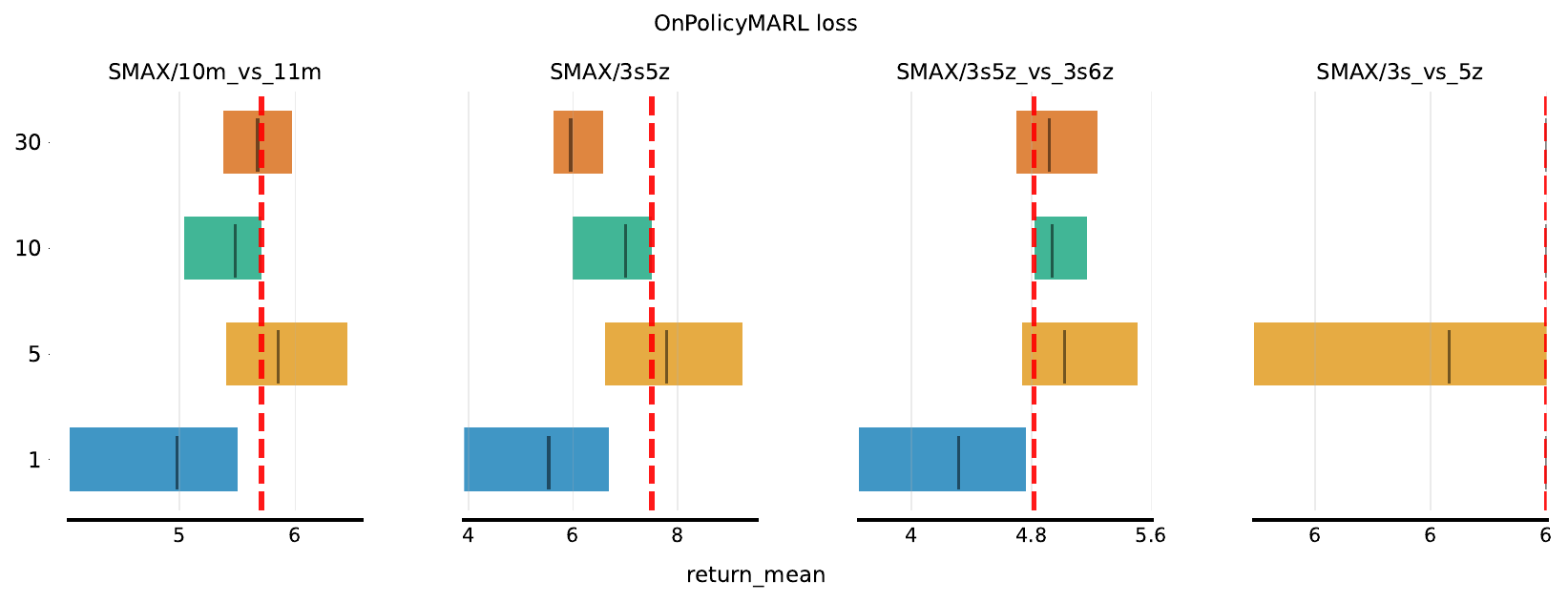}%
\caption{ADA Optimisation results on Meta-Test tasks. (Part 6/8)}
\label{fig:ADA_optimisation_mt_6}
\end{figure}
\clearpage

\begin{figure}[htbp]
\centering
\setlength{\lineskip}{0pt}
\includegraphics[width=0.48\textwidth]{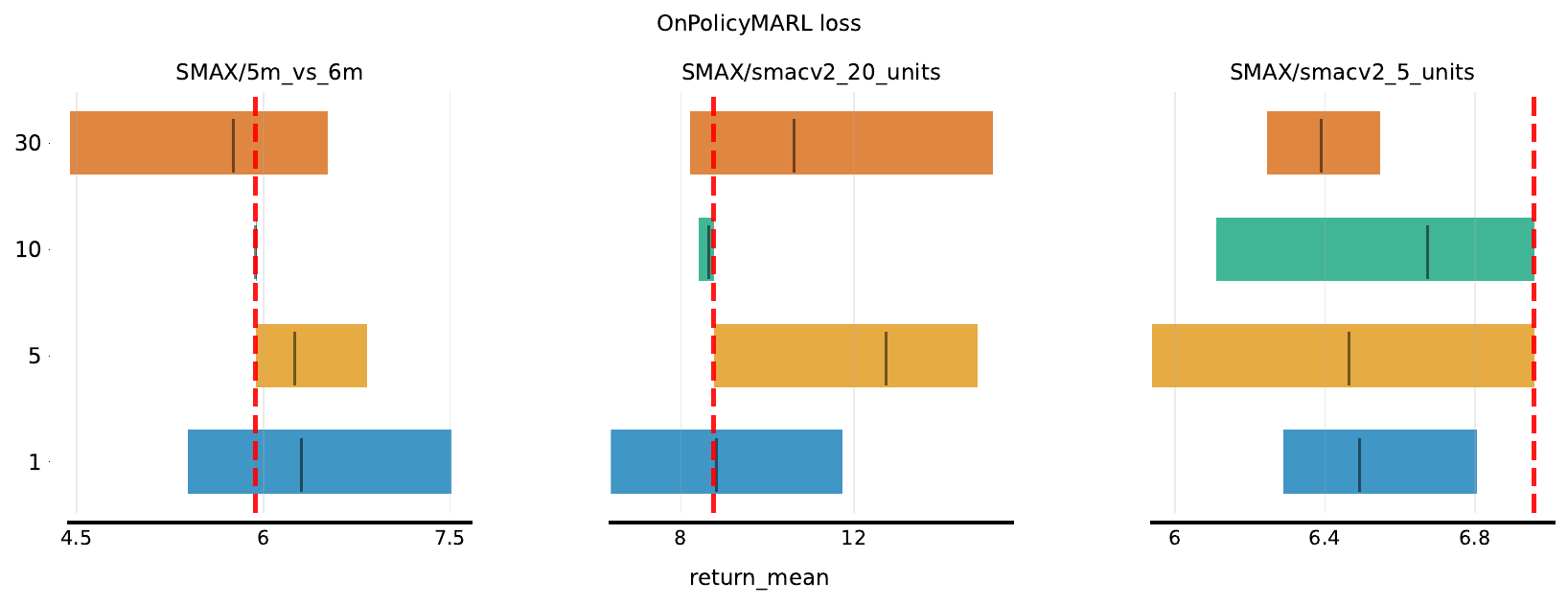}%
\hfill%
\includegraphics[width=0.48\textwidth]{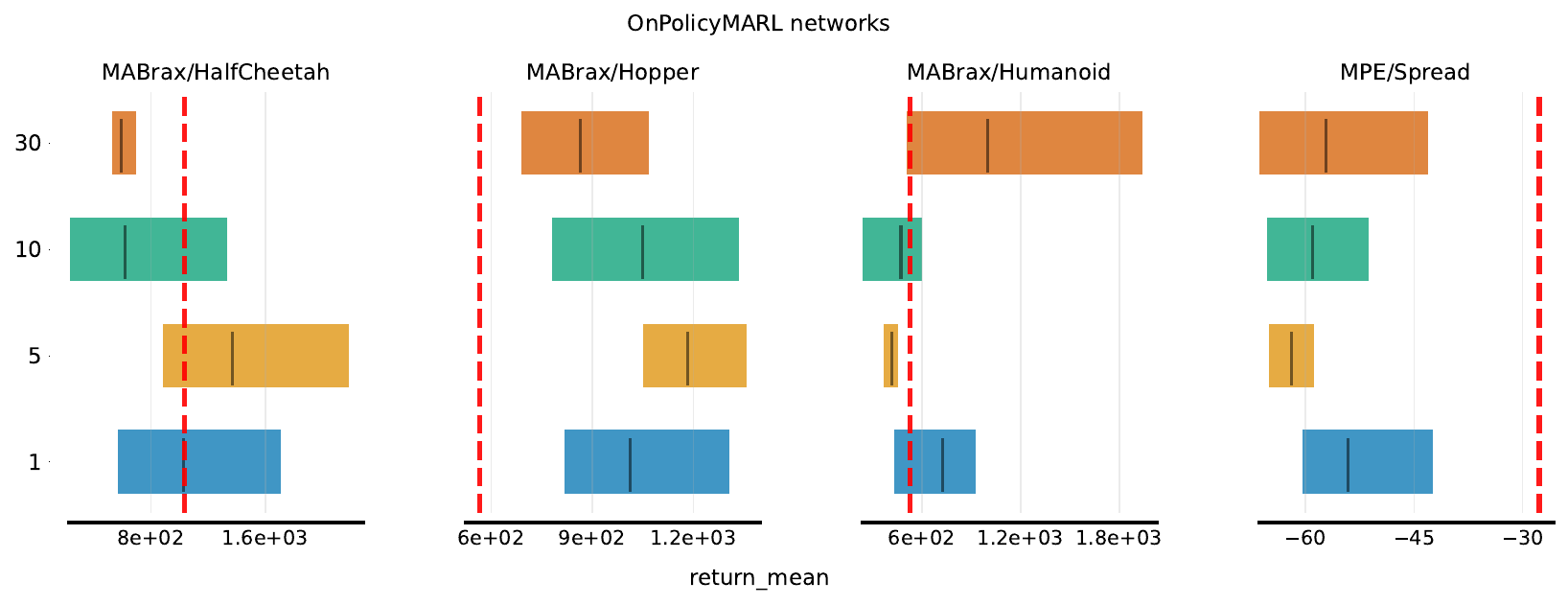}%
\\[0.5em]
\includegraphics[width=0.48\textwidth]{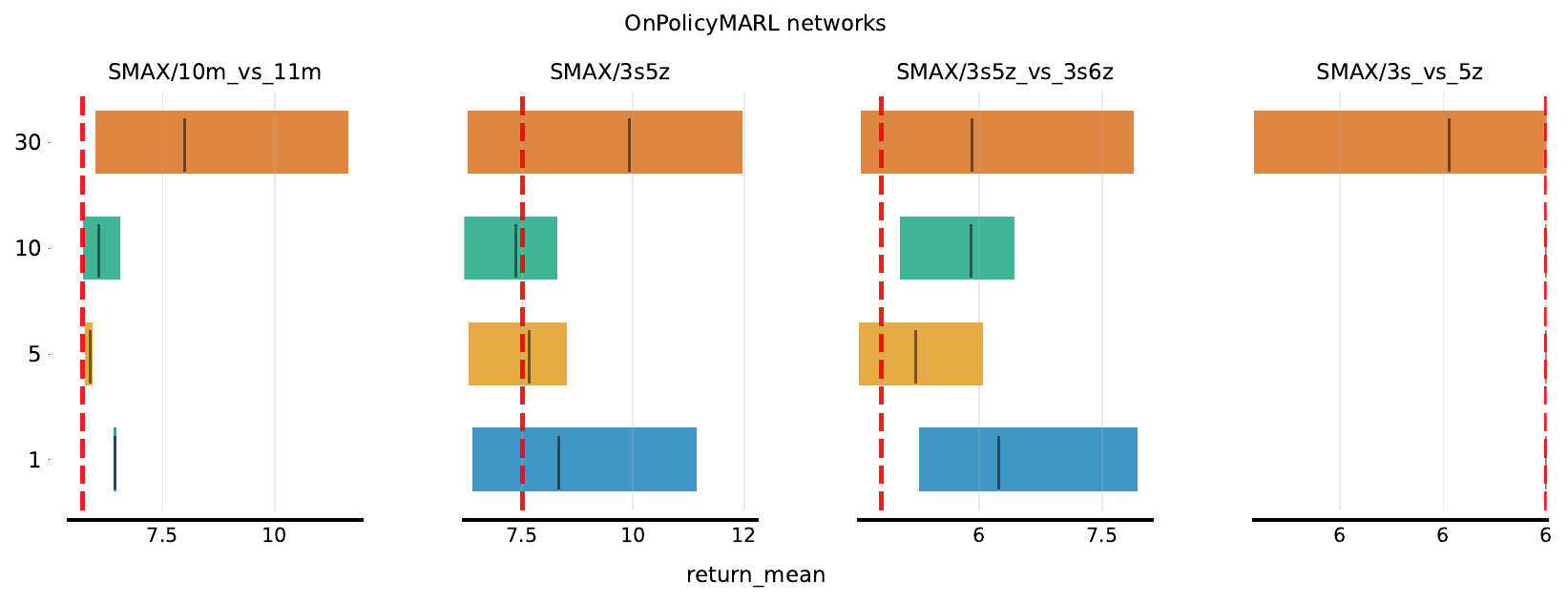}%
\hfill%
\includegraphics[width=0.48\textwidth]{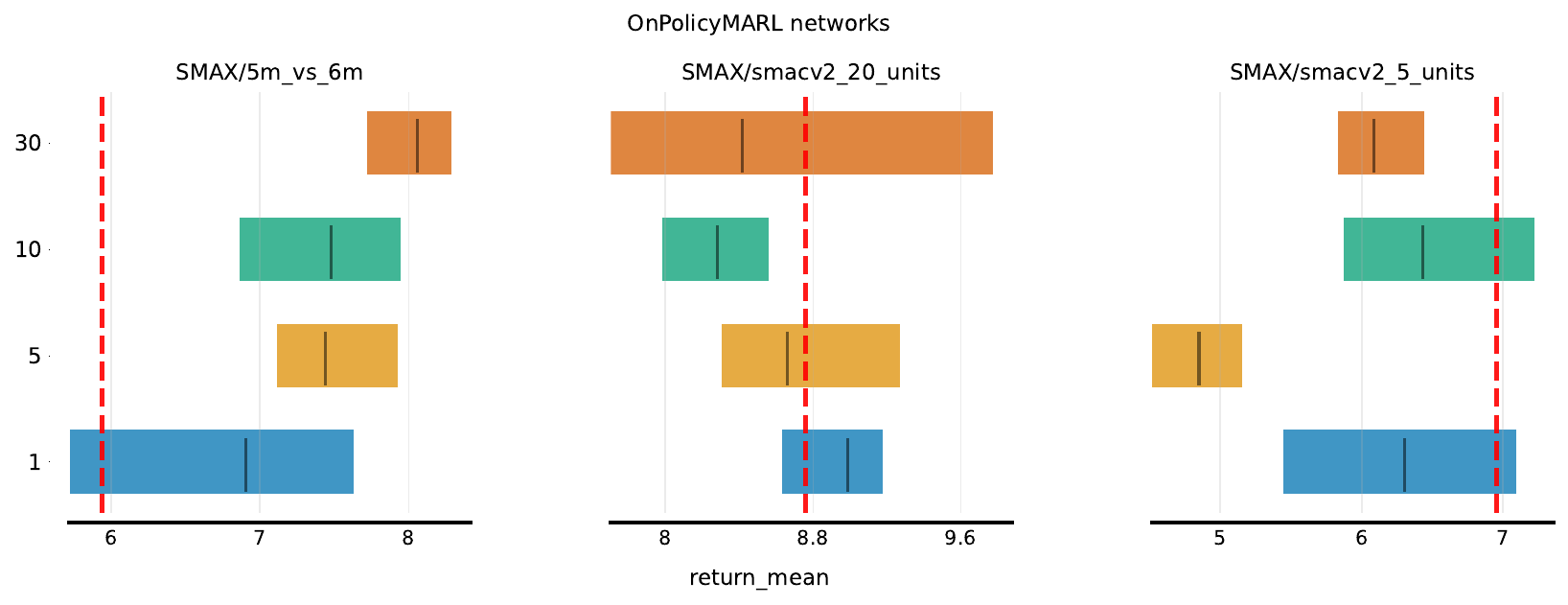}%
\\[0.5em]
\includegraphics[width=0.48\textwidth]{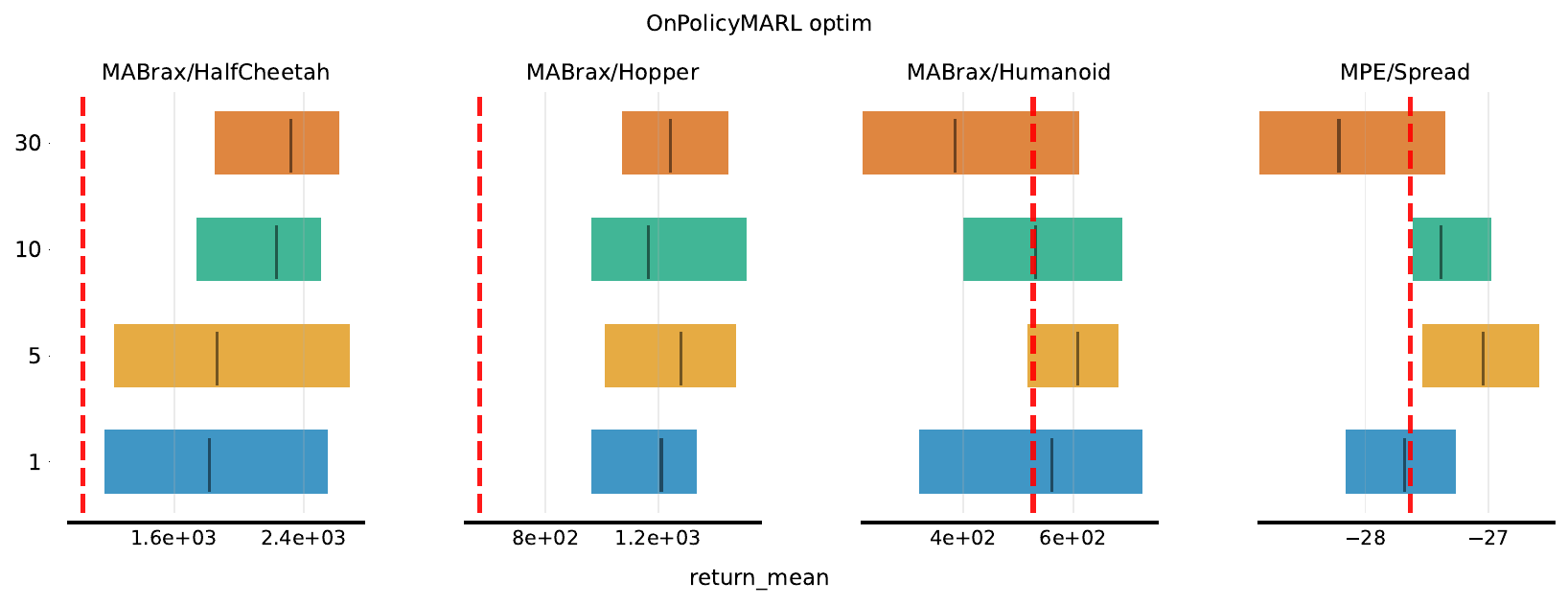}%
\hfill%
\includegraphics[width=0.48\textwidth]{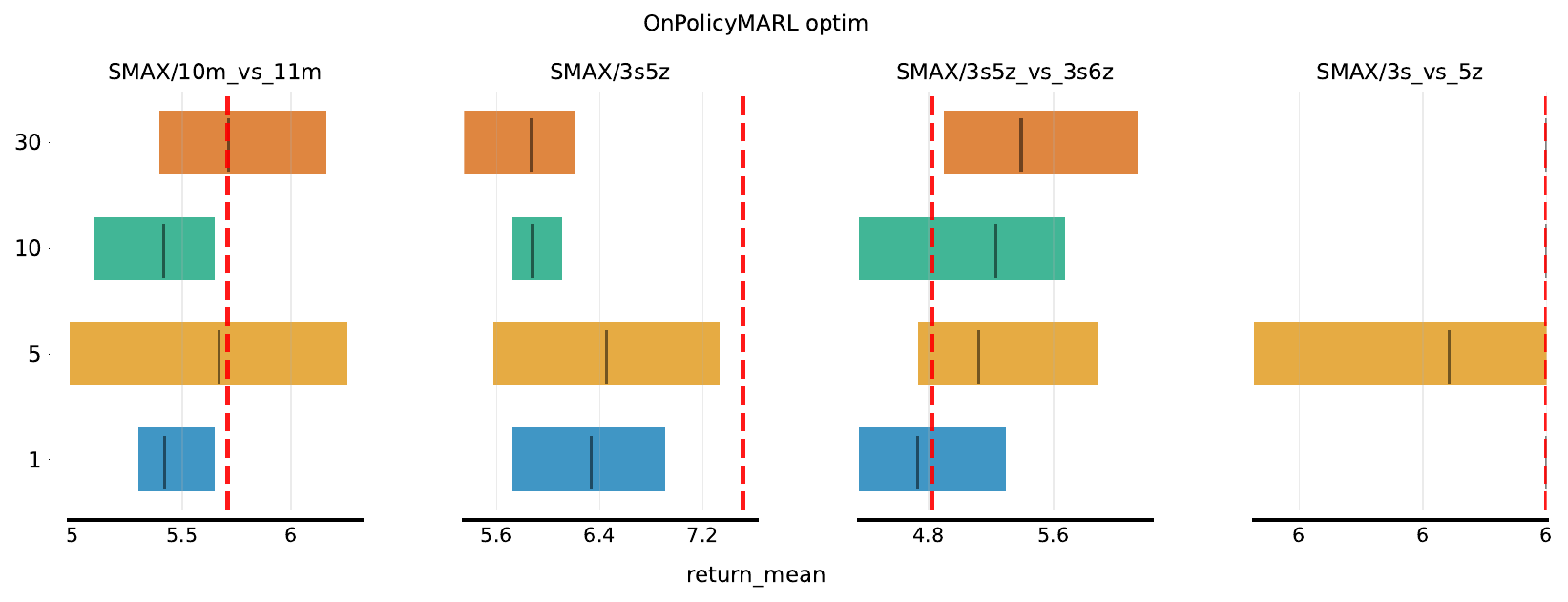}%
\\[0.5em]
\includegraphics[width=0.48\textwidth]{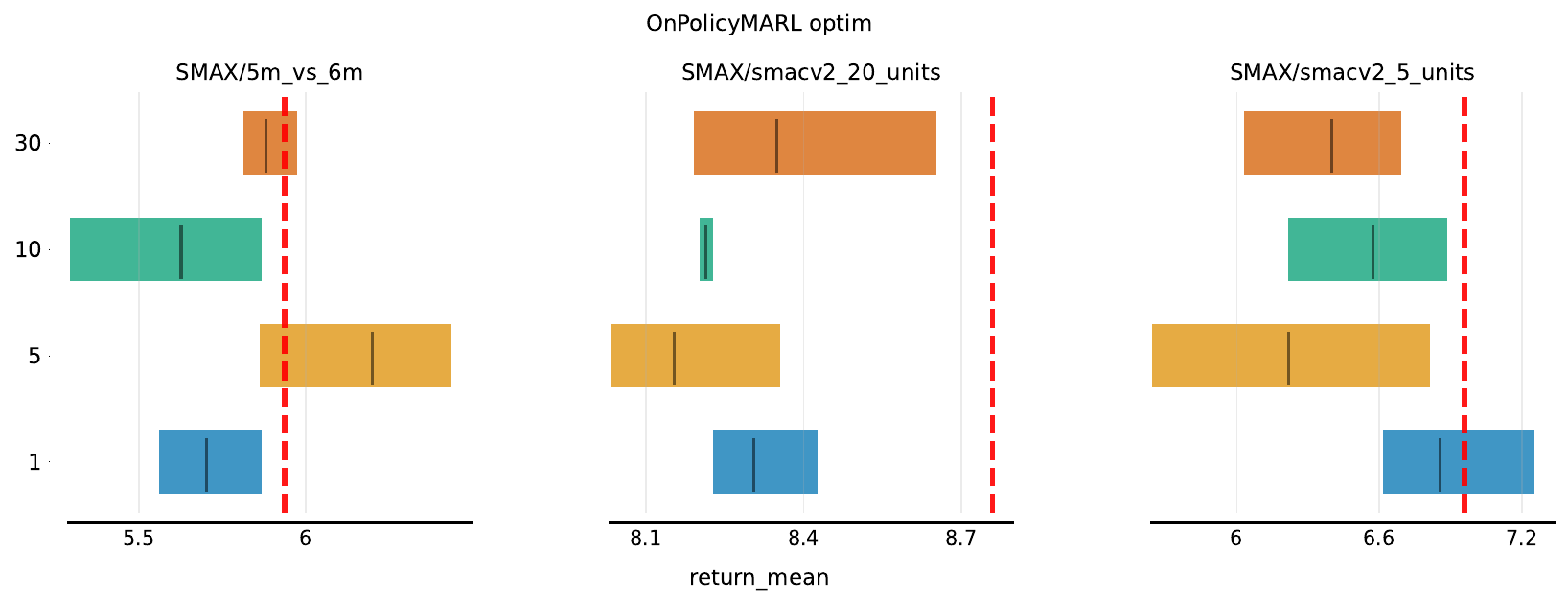}%
\hfill%
\includegraphics[width=0.48\textwidth]{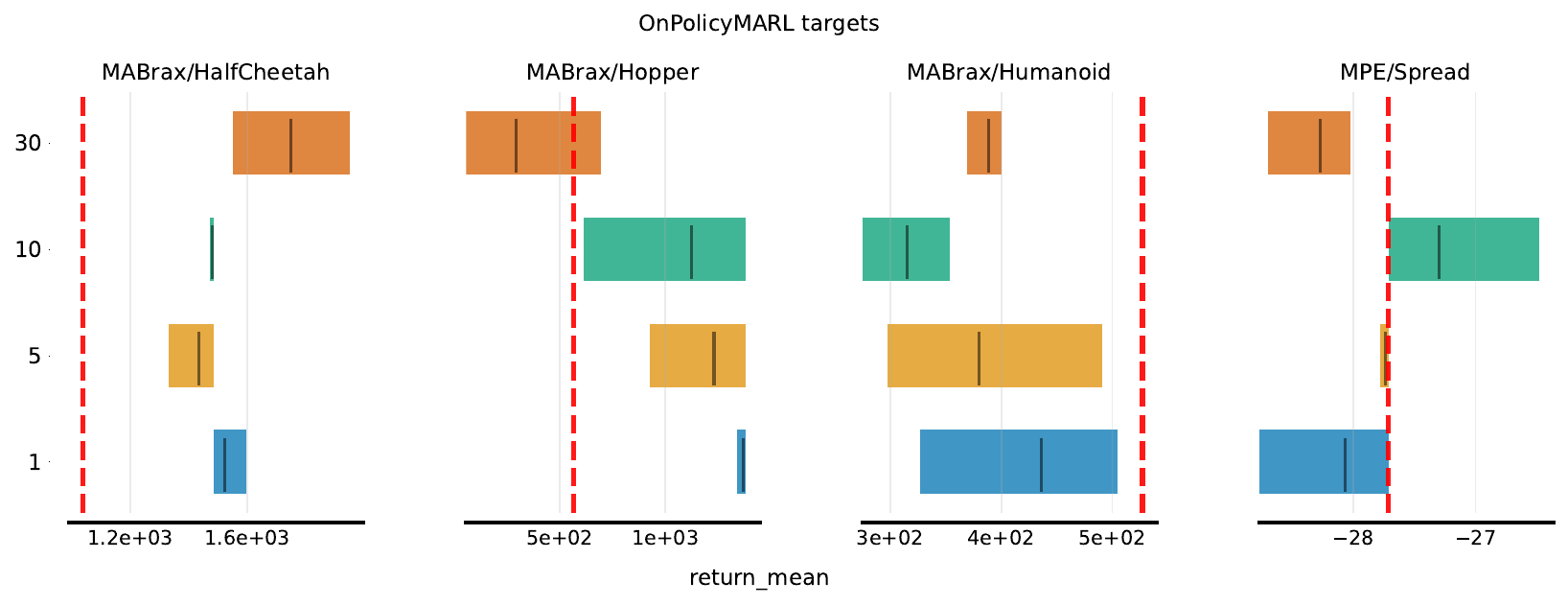}%
\\[0.5em]
\includegraphics[width=0.48\textwidth]{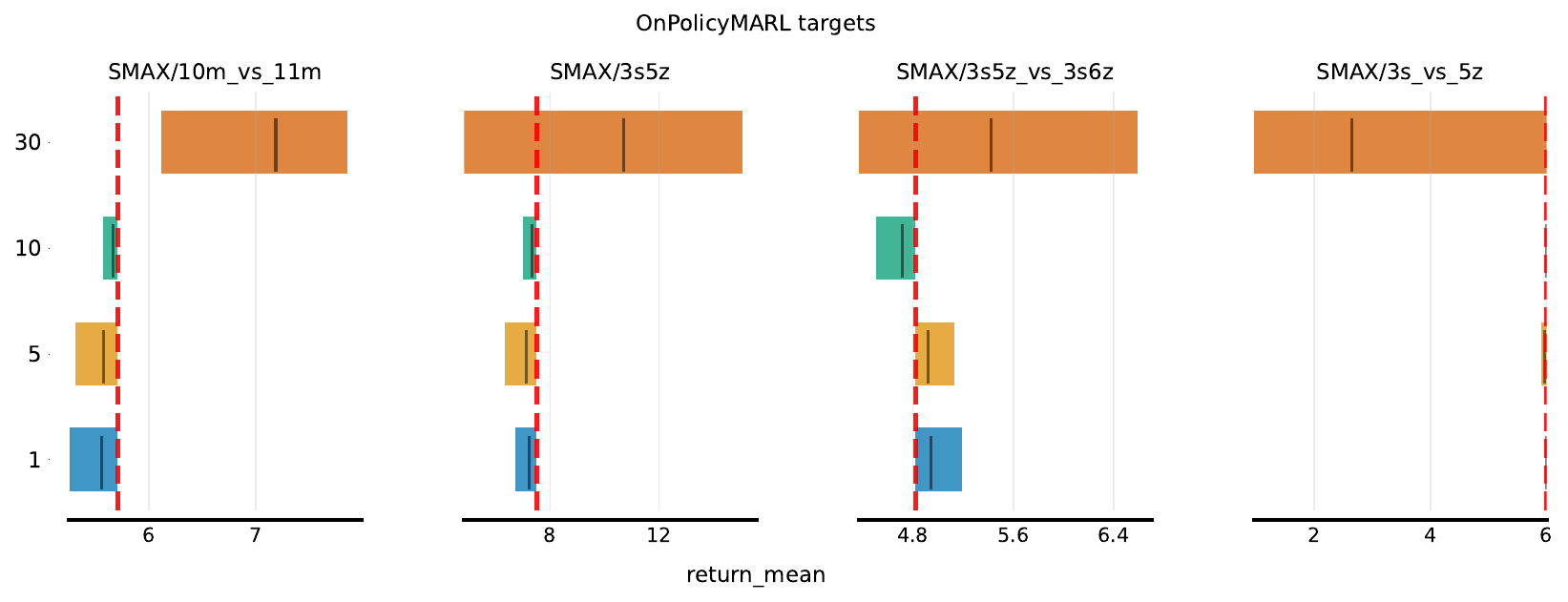}%
\hfill%
\includegraphics[width=0.48\textwidth]{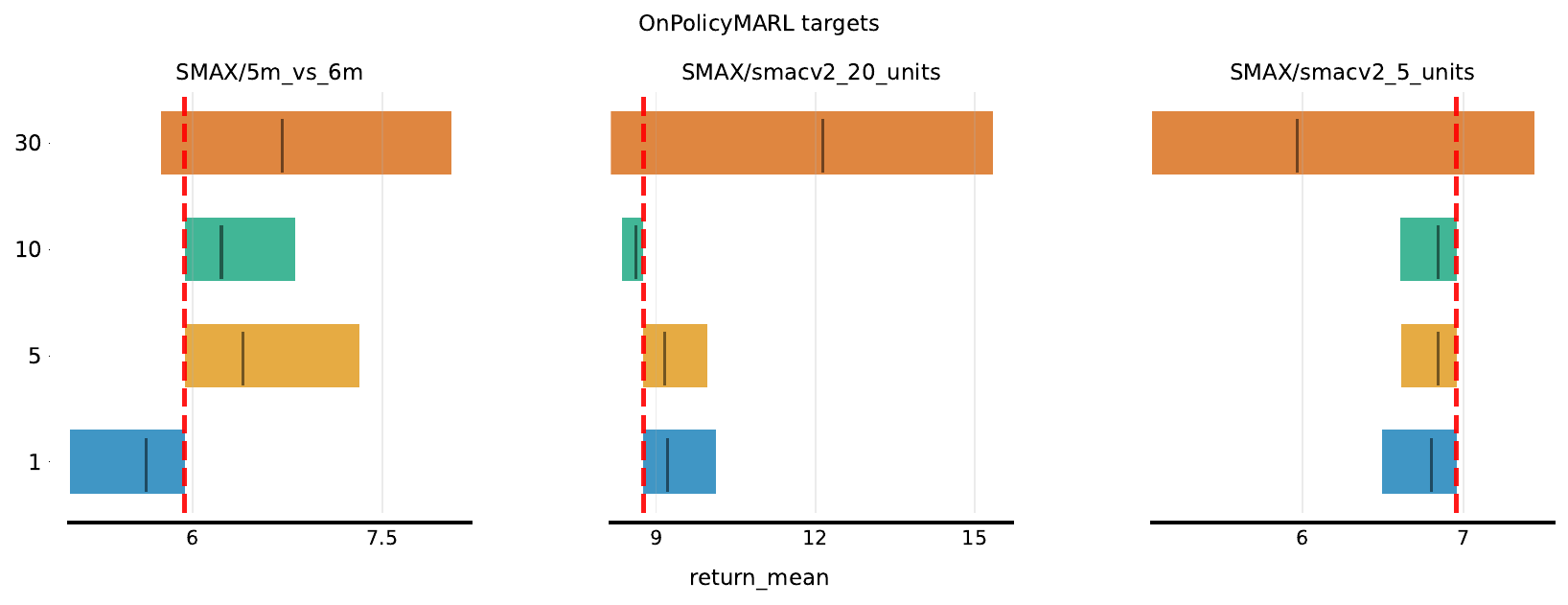}%
\\[0.5em]
\includegraphics[width=0.48\textwidth]{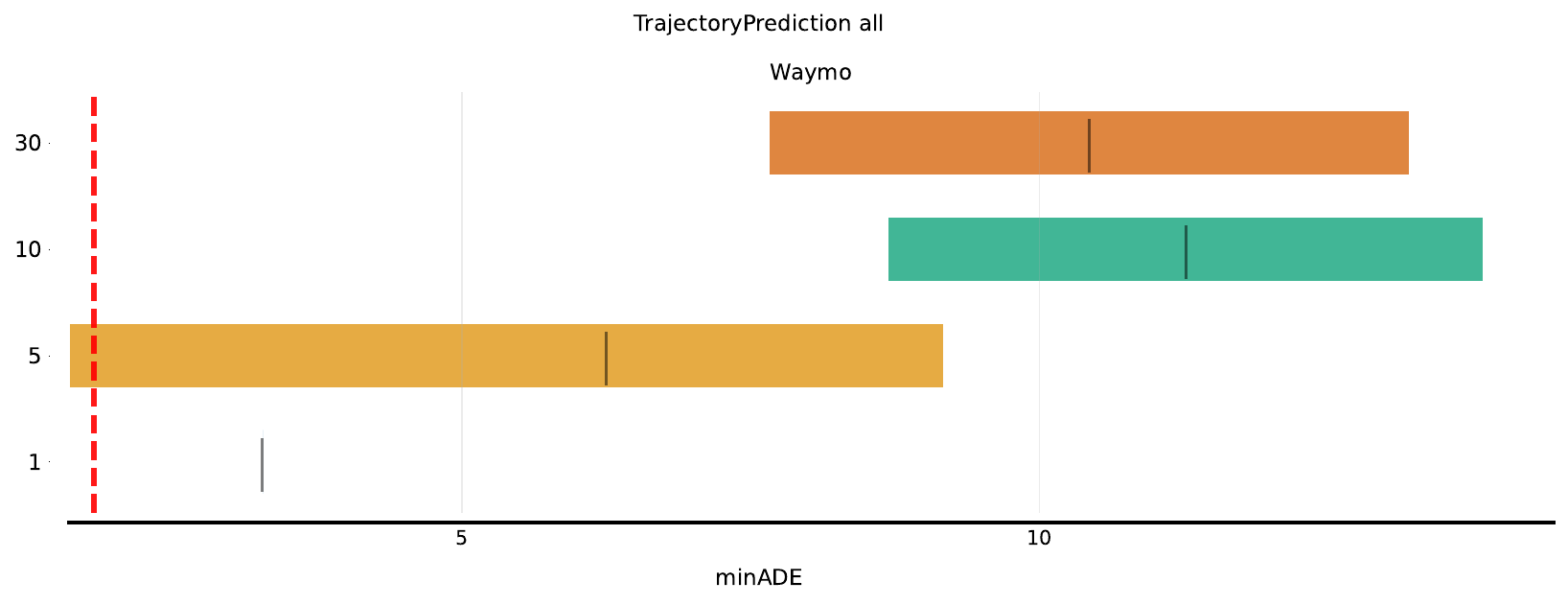}%
\hfill%
\includegraphics[width=0.48\textwidth]{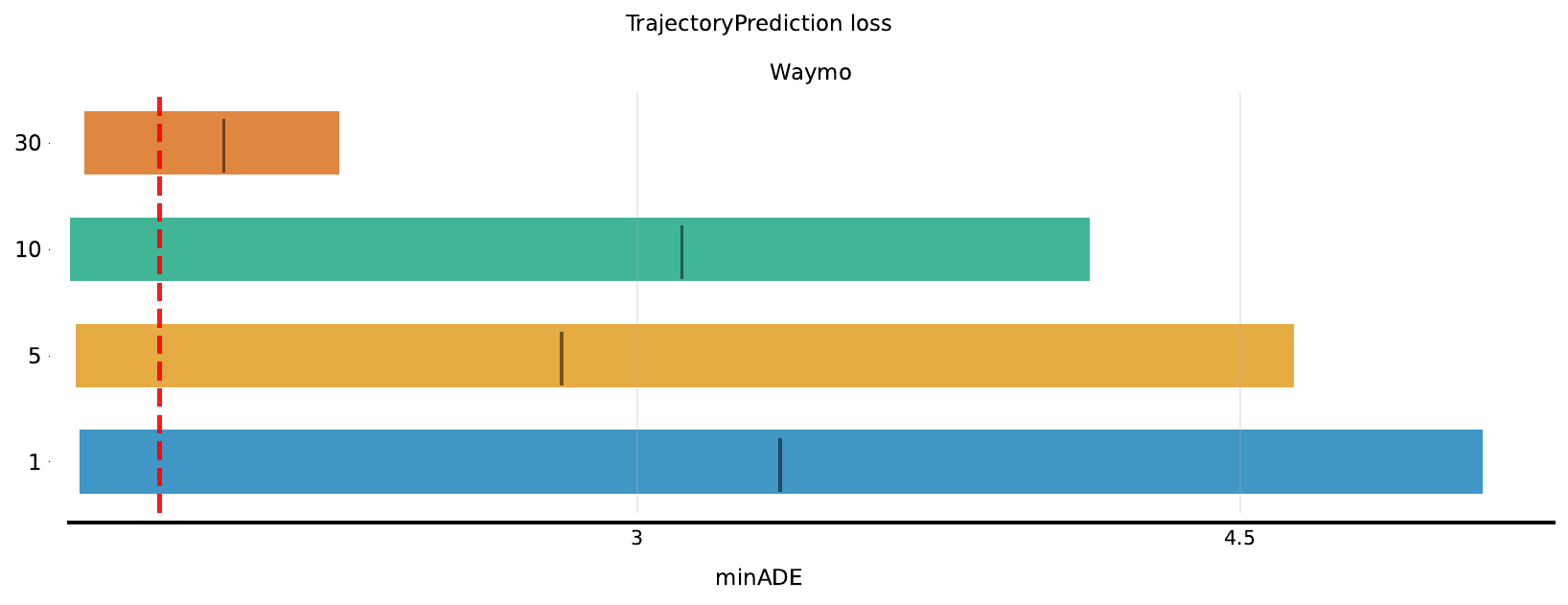}%
\caption{ADA Optimisation results on Meta-Test tasks. (Part 7/8)}
\label{fig:ADA_optimisation_mt_7}
\end{figure}
\clearpage

\begin{figure}[htbp]
\centering
\setlength{\lineskip}{0pt}
\includegraphics[width=0.48\textwidth]{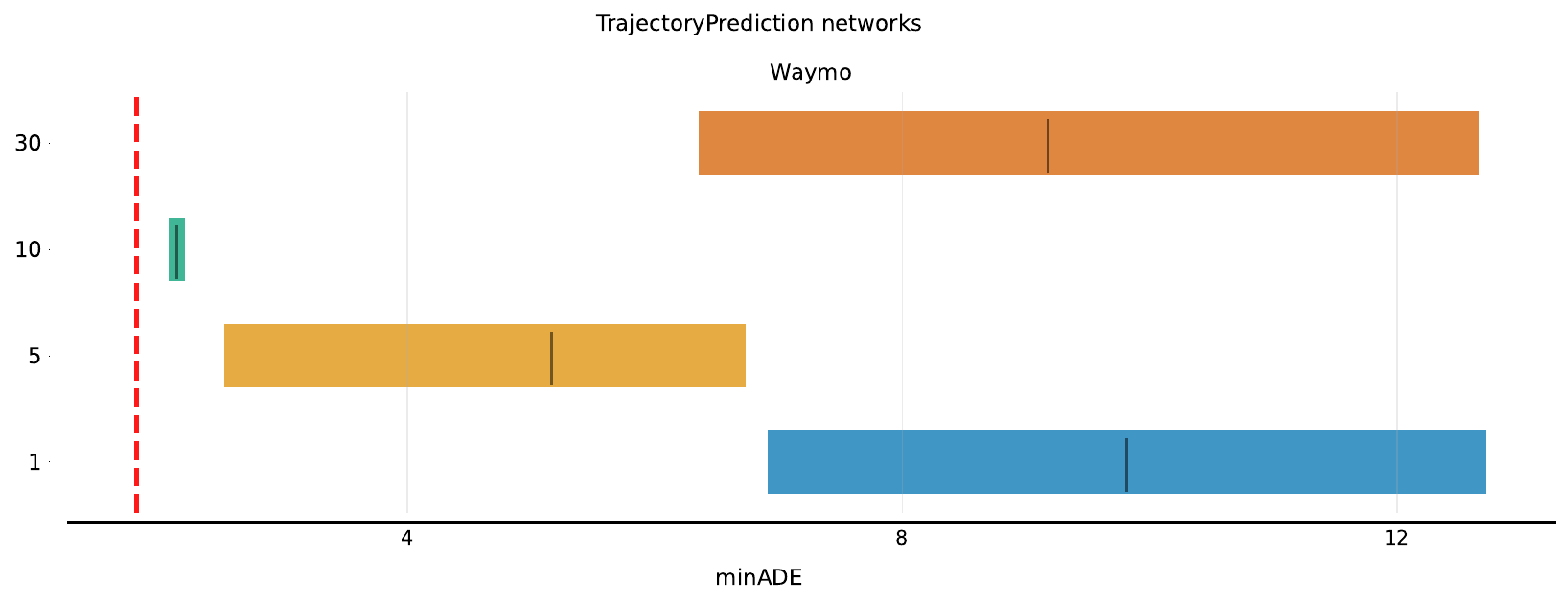}%
\hfill%
\includegraphics[width=0.48\textwidth]{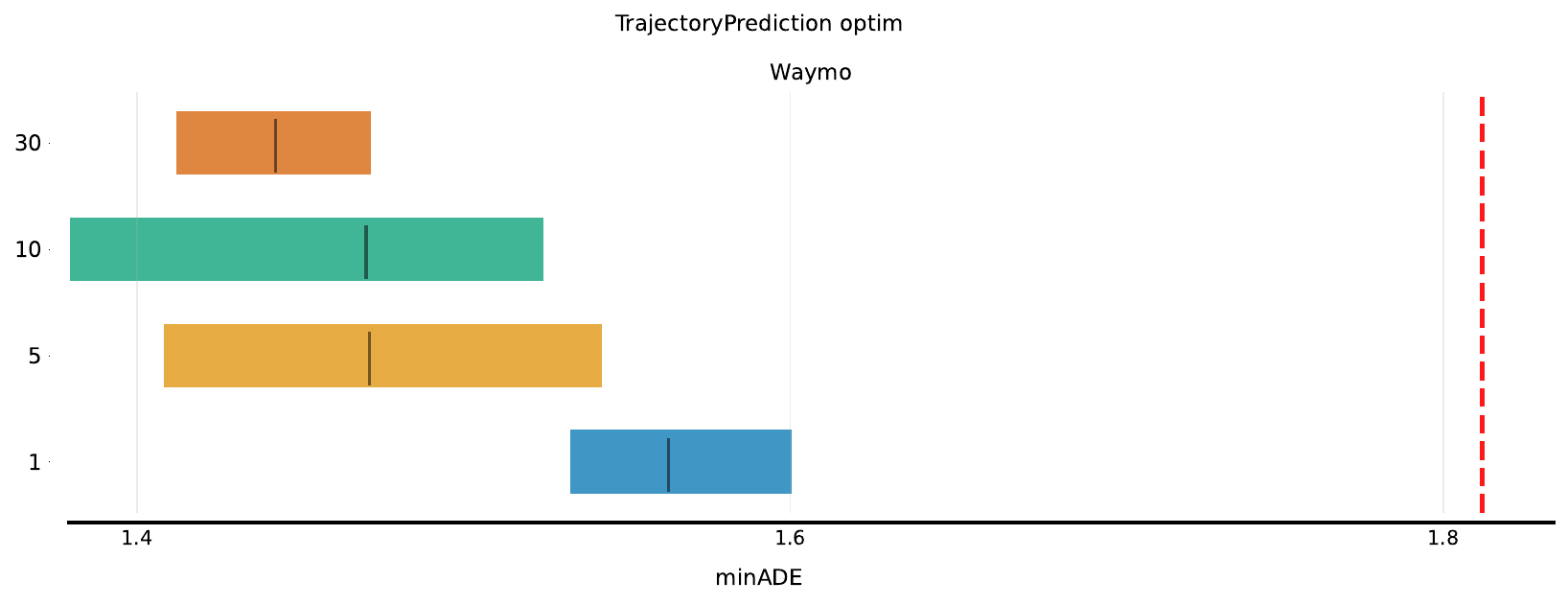}%
\\[0.5em]
\includegraphics[width=0.48\textwidth]{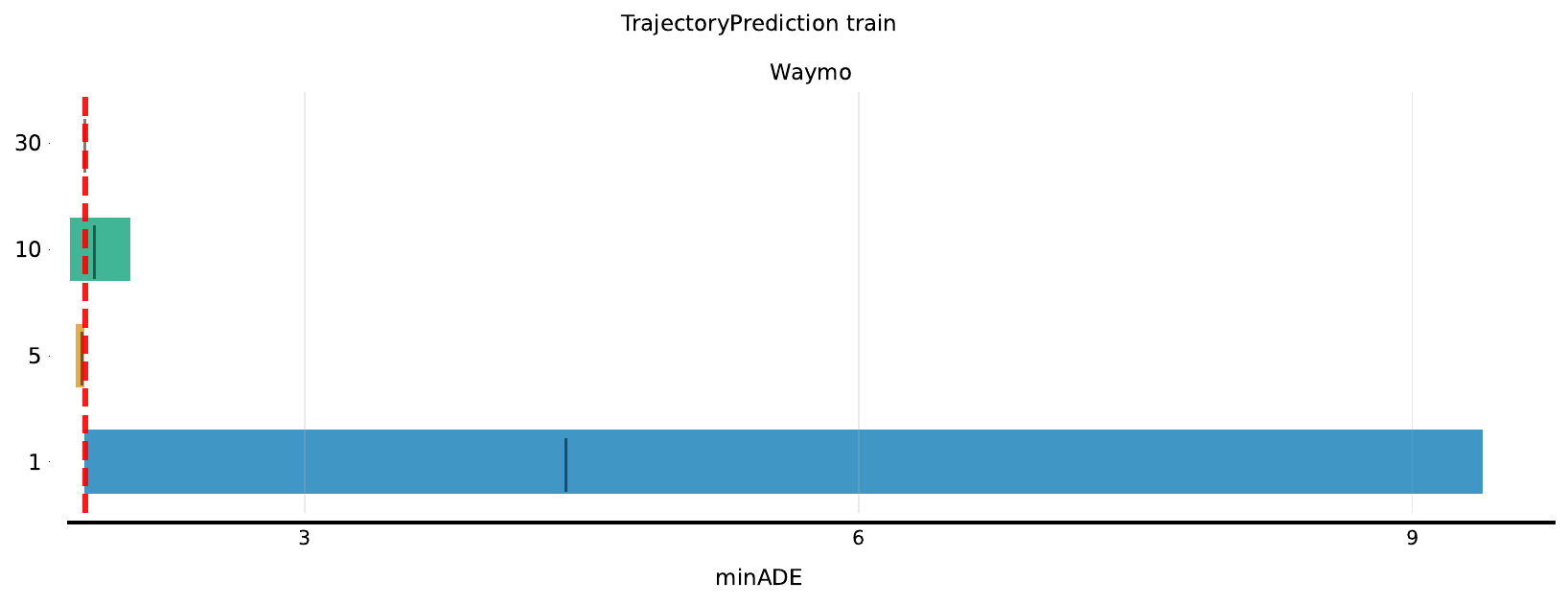}%
\hfill%
\includegraphics[width=0.48\textwidth]{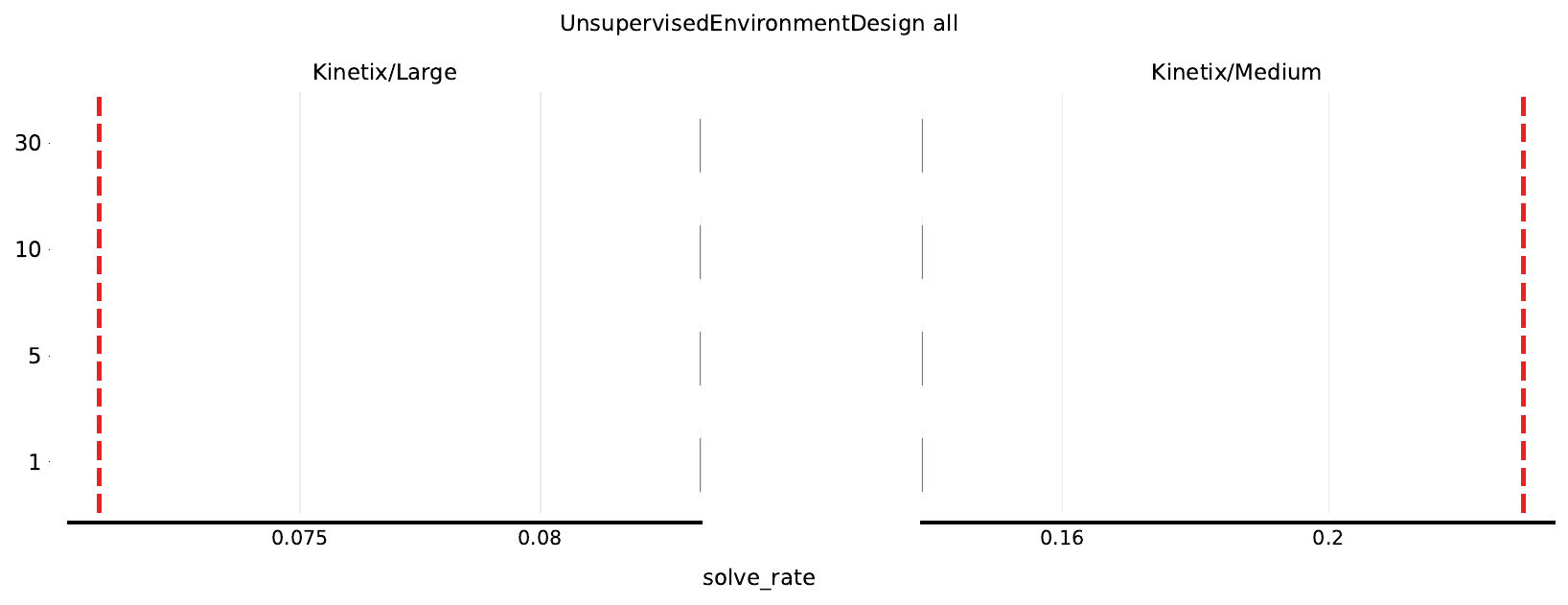}%
\\[0.5em]
\includegraphics[width=0.48\textwidth]{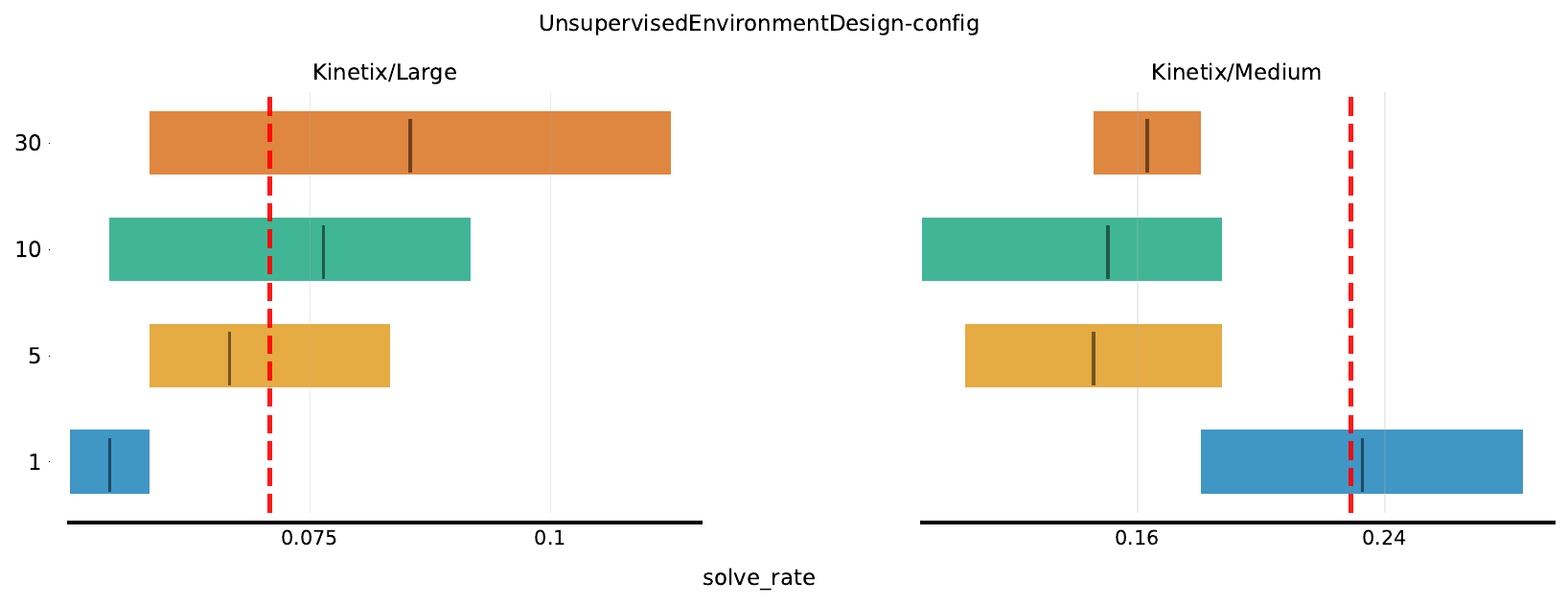}%
\hfill%
\includegraphics[width=0.48\textwidth]{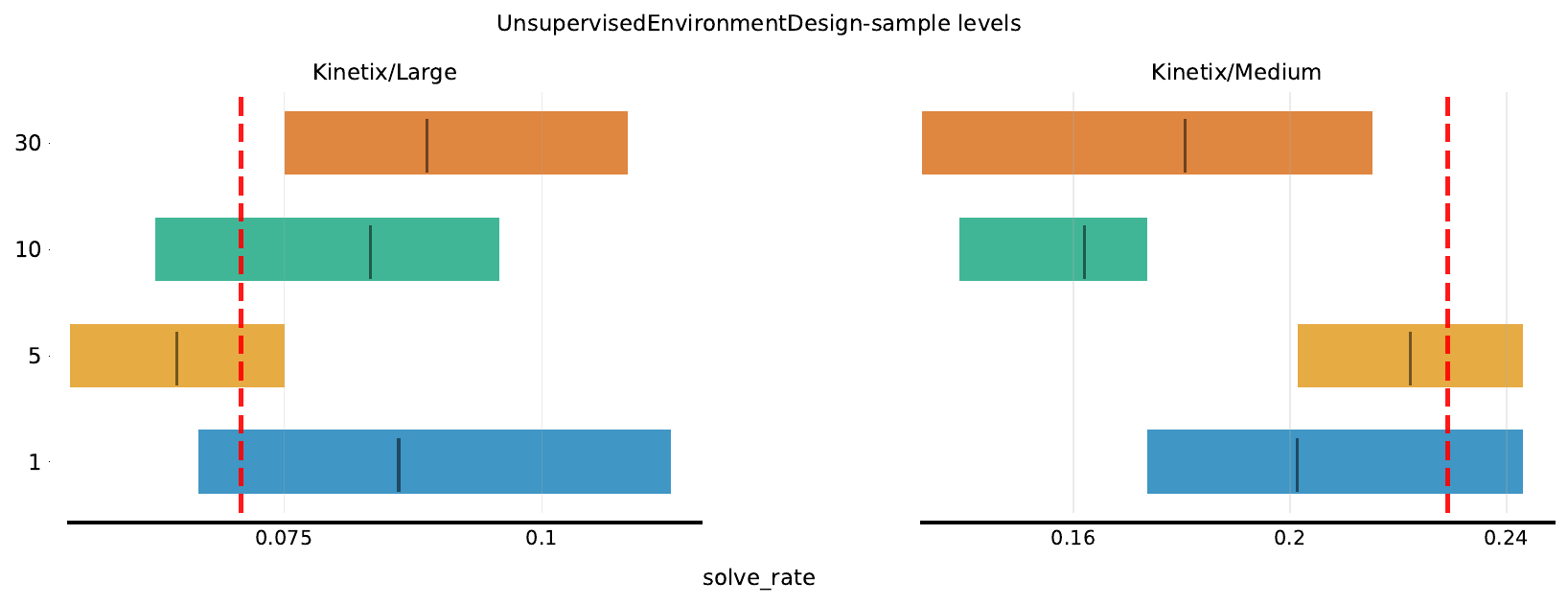}%
\\[0.5em]
\includegraphics[width=0.48\textwidth]{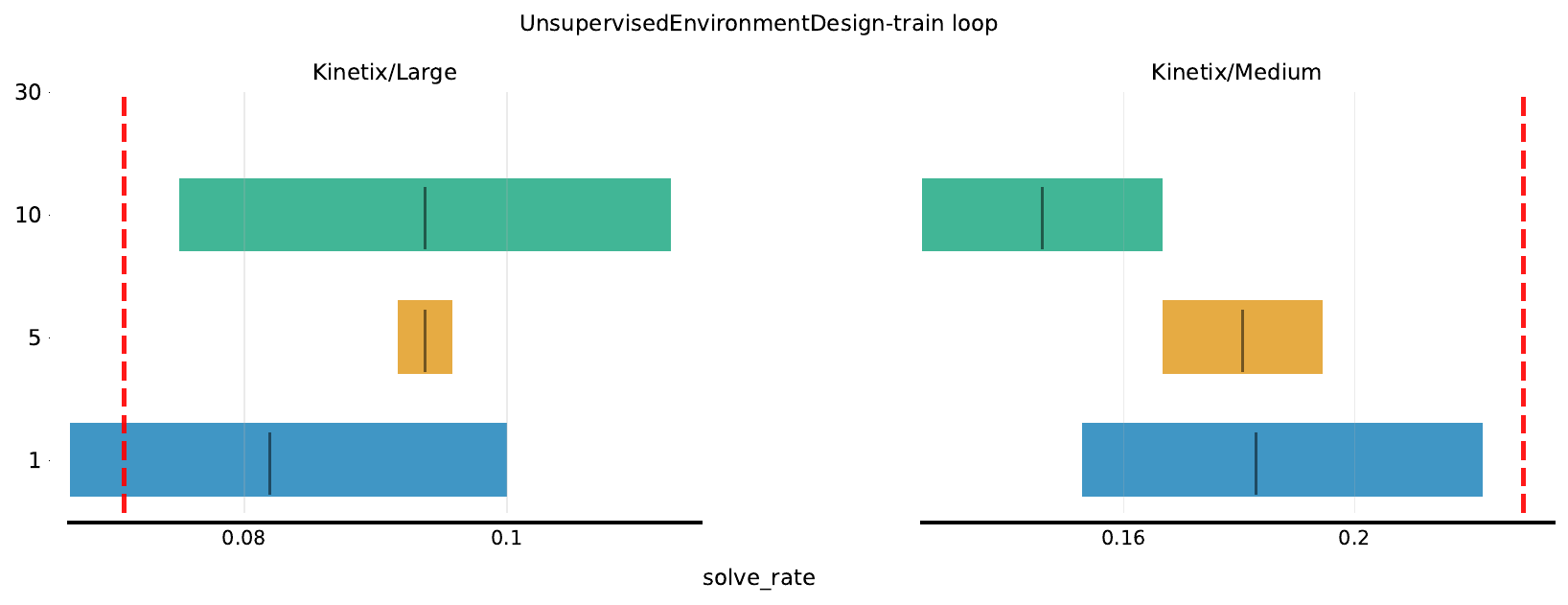}%
\hfill%
\includegraphics[width=0.48\textwidth]{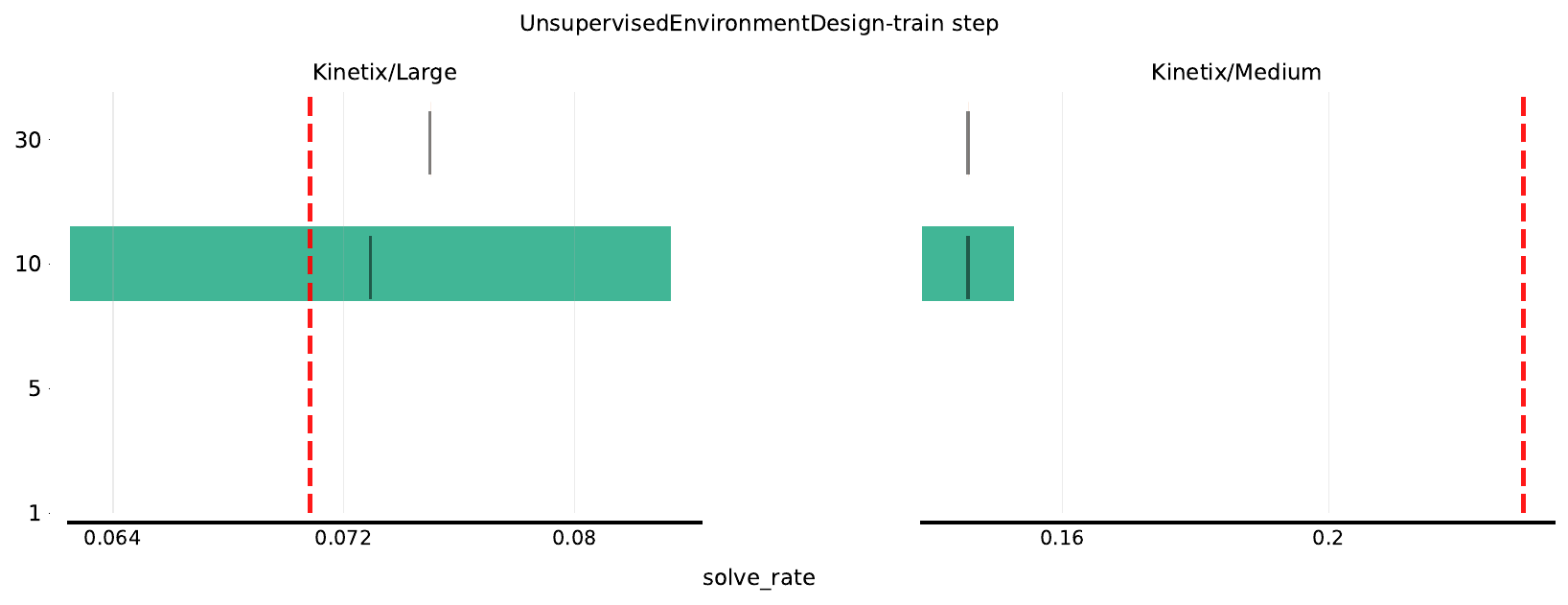}%
\caption{ADA Optimisation results on Meta-Test tasks. (Part 8/8)}
\label{fig:ADA_optimisation_mt_8}
\end{figure}
\clearpage

\newpage

\section{Prompts} \label{app:sysprompts}

Here, we provide system prompts that were used for all agents. We attempt to keep system prompts broad so as to not bias LLMs much.

\subsection{MLGym Agent System Prompt} \label{app:systemprompt}

We use the generic MLGym System Prompt, with minor tweaks to reflect the tasks in DiscoGen. The prompt reads as follows:

\begin{lstlisting}[style=promptstyle]
SETTING: You are an autonomous Machine Learning Researcher, and you're working directly in the command line with a special interface.

  The special interface consists of a file editor that shows you {WINDOW} lines of a file at a time.
  In addition to typical bash commands, you can also use the following commands to help you navigate and edit files.

  COMMANDS:
  {command_docs}

  Please note that THE EDIT and INSERT COMMANDS REQUIRES PROPER INDENTATION.
  If you'd like to add the line '        print(x)' you must fully write that out, with all those spaces before the code! Indentation is important and code that is not indented correctly will fail and require fixing before it can be run.

  RESPONSE FORMAT:
  Your shell prompt is formatted as follows:
  (Open file: <path>) <cwd> $

  You need to format your output using two fields; discussion and command.
  Your output should always include _one_ discussion and _one_ command field EXACTLY as in the following example:
  DISCUSSION
  First I'll start by using ls to see what files are in the current directory. Then maybe we can look at some relevant files to see what they look like.
  ```
  ls -a
  ```

  You should only include a *SINGLE* command in the command section and then wait for a response from the shell before continuing with more discussion and commands. Everything you include in the DISCUSSION section will be saved for future reference. Please do not include any DISCUSSION after your action.
  If you'd like to issue two commands at once, PLEASE DO NOT DO THAT! Please instead first submit just the first command, and then after receiving a response you'll be able to issue the second command.
  You're free to use any other bash commands you want (e.g. find, grep, cat, ls, cd) in addition to the special commands listed above.
  However, the environment does NOT support interactive session commands (e.g. python, vim), so please do not invoke them.
  Your goal is to achieve the best possible score, not just to submit your first working solution. Consider strategies like validating your answer using the `validate` command, manually spot-checking algorithms, and comparing different ideas and implementations.
  Once you have exhausted all possible solutions and cannot make progress, you can submit your final solution by using `submit` command.

  IMPORTANT TIPS:
  1. Always work with the files you have been told to.  You should never try to edit files you are not told to edit, or add new ones.

  2. If you run a command and it doesn't work, try running a different command. A command that did not work once will not work the second time unless you modify it!

  3. If you open a file and need to get to an area around a specific line that is not in the first {WINDOW} lines, don't just use the scroll_down command multiple times. Instead, use the goto <line_number> command. It's much quicker.

  4. Always make sure to look at the currently open file and the current working directory (which appears right after the currently open file). The currently open file might be in a different directory than the working directory! Note that some commands, such as 'create', open files, so they might change the current  open file.

  5. When editing files, it is easy to accidentally specify a wrong line number or to write code with incorrect indentation. Always check the code after you issue an edit to make sure that it reflects what you wanted to accomplish. If it didn't, issue another command to fix it.

  6. You have a limited number of actions/steps you can take in the environment. The current step and remaining number of steps will given after every action. Use the remaining steps wisely. If you only have few remaining steps, it is better to submit a working solution than to keep trying.

  7. Your each action should take less than {training_timeout} seconds to complete. If your action doesn't finish within the time limit, it will be interrupted.

  8. Validating your solution often, will give you a good idea of your progress so far and you will be able to adapt your strategy. To ensure you are always in the correct directory, use the `validate` function instead. This will also make sure that your scores are logged.

  9. Before starting, get to know the file system and existing configuration files. You should make sure not to index the config for arguments that don't exist, and any additional hyperparameters will not be tuned and must be defined directly in the files which you have been asked to change. REMEMBER, YOU SHOULD NOT ADD NEW HYPERPARAMETERS DIRECTLY TO THE CONFIG FILES.
\end{lstlisting}

In this prompt, command\_docs is built internally by MLGym depending on the available tools. DiscoGen automatically builds a \texttt{task\_template} using the descriptions in our repository, which is appended to the system prompt. The \texttt{task\_template} reads:

\begin{lstlisting}[style=promptstyle]
    We're currently solving the following task. Here's the task description:

  TASK DESCRIPTION:
  {description}

  INSTRUCTIONS:
  Now, you're going to write code to improve performance on this task. Your terminal session has started and you're in the workspace root directory. You can use any bash commands or the special interface to help you. Edit all the files you need.
  Remember, YOU CAN ONLY ENTER ONE COMMAND AT A TIME. You should always wait for feedback after every command.
  When you're satisfied with all of the changes you have made, you can run your code. Your code should have no logical errors or syntax errors. If it works, you will see a reported evaluation score. An empty evaluation score suggests there is some logical error in your code.

  Note however that you cannot use any interactive session commands (e.g. python, vim) in this environment, but you can write scripts and run them. E.g. you can write a python script and then run it with `python <script_name>.py`, or run `python main.py` from a single dataset.

  NOTE ABOUT THE EDIT AND INSERT COMMANDs: Indentation really matters! When editing a file, make sure to insert appropriate indentation before each line!

  (Current Step: {current_step}, Remaining Steps: {remaining_steps})
  (Open file: {open_file})
  (Current directory: {working_dir})
  bash-$
\end{lstlisting}

\subsection{Prompt Optimisation System Prompt}

We develop a new system prompt for the prompt improving language model. This reads as:

\begin{lstlisting}[style=promptstyle]
    You are an AI research agent designing a new prompt for AI algorithm discovery agents. Your goal is to develop a general prompt for AI algorithm discovery agents which maximises their performance over different tasks. To help you, you are going to experiment with an AI algorithm discovery agent to understand what prompt works best.

We are going to randomly sample a task from a machine learning field, and you will be given a few attempts at improving your prompt with an AI agent before we sample a new task. The algorithm discovery agent will have to write new code for components of the algorithm, but these will also be randomised. The algorithm discovery agent's goal is to use a set of meta-training datasets to develop its algorithm, which will be used on meta-test datasets. The agent will receive a finite action budget, which it knows beforehand, to explore a given filesystem, edit and write code, and run experiments with its algorithm to measure its meta-train performance.

After each attempt, you will be shown how the algorithm performed on the meta-training and meta-testing datasets. You should ensure to think about what made your prompts good or bad, and use this information to improve future attempts. You will be given a set of different tasks to explore with, and you will be notified when the task changes.

You must respond with a strictly valid JSON object and no other text. Use the format: {"thought": "...", "prompt": "..."}. Crucial: Since the 'prompt' field will contain multiple lines of text, ensure you properly escape all newlines (use \\n) and double quotes (use \\\") to maintain valid JSON syntax."

We will append your prompt to a system prompt which describes the tools and task the agent must solve. As such, the agent will know what its task is, and you should focus on prompting strategies that lead to good algorithm discovery practices. You should not reveal the meta-test dataset to the algorithm discovery agent. Your goal is to maximise performance.
\end{lstlisting}

\subsection{Prompt Optimisation Discovered Prompts}\label{app:discovered_prompts}

Below, we provide the four prompts discovered by our prompt improvement system. 

\subsubsection{Prompt from 1 Task} \label{app:task1prompt}

\begin{lstlisting}[style=promptstyle]
You are developing a generalizable RL algorithm evaluated on meta-training tasks (Breakout, Freeway) and hidden meta-test tasks. Your goal is to maximize performance across ALL tasks while maintaining robust generalization.

## CORE OBJECTIVE

**Primary:** Maximize average performance across both meta-train AND meta-test tasks
**Key insight:** Best meta-test results come from BALANCED improvements, not single-task optimization
**Baseline:** Breakout: 70.16±22.28, Freeway: 62.38±1.69
**Critical:** Both tasks must improve or maintain - sacrificing one task hurts meta-test performance

## GENERALIZATION PRINCIPLES

**Meta-analysis reveals:**
- Sweet spot: Breakout std 6-10 (not too low, not too high)
- Both tasks matter: Freeway degradation indicates poor generalization
- Best meta-test (Asterix ~31): Came from Breakout 82.6±8.9, Freeway 62±1.4
- Network capacity + moderate exploration > extreme parameter values

**High-value changes (prioritized):**
1. **Network capacity:** Hidden size 128-192 (not 256 - too large can hurt Freeway)
2. **Balanced exploration:** Entropy 0.015-0.025 (not >0.03 - hurts stability)
3. **Credit assignment:** GAE lambda 0.96-0.97
4. **Multi-task stability:** Learning rate tuning for both tasks

**RED FLAGS to avoid:**
- Breakout std <5 (over-optimization) OR >12 (instability)
- Freeway dropping >2 points from baseline (poor generalization)
- Extreme parameter values (entropy >0.03, hidden_size >256)

## PHASE 1: DISCOVERY (Budget: 8%)

1. `list_files` - scan filesystem
2. Read key files: network.py, model.py, agent.py, ppo.py, config files
3. Identify parameters:
   - **hidden_size** (typically 64) - PRIMARY TARGET
   - **entropy_coef** (typically 0.01) - SECONDARY TARGET
   - **gae_lambda** (typically 0.95)
   - **learning_rate** (typically 3e-4)
   - **num_layers** or architecture structure
   - **max_grad_norm**, **clip_range** if present
4. Document baseline architecture clearly

## PHASE 2: CAPACITY FOUNDATION (Budget: 25%)

**Strategy: Start with network capacity - most reliable improvement**

**Step A - Moderate Capacity Increase:**
1. Change hidden_size: 64 → 128 (conservative, reliable)
2. Use `write_file` with complete file content
3. Verify with `read_file` (confirm shows 128)
4. Run 3 Breakout, 3 Freeway experiments
5. Evaluate BOTH tasks:
   - Breakout: Should improve to >75
   - Freeway: MUST maintain >60 (if drops, try 96 instead of 128)

**Step B - Test Larger Capacity (if Step A succeeds):**
1. Try hidden_size: 128 → 192
2. Verify with `read_file`
3. Run 3 Breakout, 3 Freeway experiments
4. Critical check:
   - Breakout improved AND Freeway maintained >60? → Keep 192
   - Freeway dropped? → Revert to 128

**Step C - Architecture Depth (alternative path):**
1. If single-layer network, try adding second layer
2. Conservative: 128→96 or 128→128
3. Verify changes
4. Run 3 experiments per task
5. Compare to single-layer results

**Decision criteria:**
- Choose configuration where BOTH tasks improve or maintain
- Breakout >75 AND Freeway >60 minimum
- Prefer lower capacity if both work (better for generalization)

## PHASE 3: BALANCED EXPLORATION (Budget: 28%)

**Strategy: Add moderate exploration while monitoring both tasks**

**Configuration A - Conservative Exploration:**
1. Keep: Best hidden_size from Phase 2 (likely 128 or 192)
2. Add: entropy_coef 0.01 → 0.02 (2x increase, moderate)
3. Verify both changes with `read_file`
4. Run 4 Breakout, 4 Freeway experiments
5. Target metrics:
   - Breakout: >78 mean, std 6-10
   - Freeway: >60 maintained

**If Breakout std >10 or Freeway drops >1 point:**
- Try entropy_coef 0.015 instead (more conservative)
- Test with 3 experiments per task

**Configuration B - Add Credit Assignment:**
1. Keep: hidden_size + entropy_coef from best above
2. Add: GAE lambda 0.95 → 0.96 or 0.97
3. Verify all changes with `read_file`
4. Run 4 Breakout, 4 Freeway experiments
5. Evaluate:
   - Breakout: >80 mean, std 6-10 (sweet spot)
   - Freeway: >60 maintained

**If Freeway degrades at any point:**
- This is CRITICAL - indicates poor generalization
- Reduce learning_rate by 15% (e.g., 3e-4 → 2.5e-4)
- OR reduce entropy_coef by 0.005
- Test with 3 experiments per task

## PHASE 4: MULTI-TASK OPTIMIZATION (Budget: 25%)

**Goal: Optimize for both tasks simultaneously**

**Dual-task evaluation:**
1. Run 5 Breakout, 5 Freeway experiments with current best config
2. Calculate performance for both:
   - Breakout target: >82 mean, std 6-10
   - Freeway target: >60 mean (>62 ideal)

**If Breakout excellent (>85) but Freeway poor (<58):**
- Learning rate TOO HIGH for multi-task
- Reduce learning_rate by 20% (e.g., 3e-4 → 2.4e-4)
- Test with 4 experiments per task

**If Breakout good (78-82) and Freeway maintained (>60):**
- Try small improvements:
- Option A: GAE lambda 0.96 → 0.97 (if not already)
- Option B: Slightly increase entropy to 0.022-0.025
- Test with 3 experiments per task
- Only keep if BOTH tasks improve or maintain

**If both tasks strong (Breakout >82, Freeway >60):**
- Try variance optimization:
- If Breakout std <6: Increase entropy by 0.005
- If Breakout std >10: Decrease learning_rate by 10%
- Test with 3 experiments per task

**Critical principle: NEVER sacrifice Freeway for Breakout gains**
- Meta-test tasks include diverse challenges
- Freeway degradation = poor generalization
- Aim for balanced improvement

## PHASE 5: FINAL VALIDATION (Budget: 14%)

1. Verify ALL changes with `read_file`:
   - Hidden size or architecture
   - Entropy coefficient
   - GAE lambda (if changed)
   - Learning rate (if changed)
   - Any other modifications

2. Document final configuration

3. Comprehensive validation:
   - 6 Breakout experiments
   - 6 Freeway experiments

4. Calculate and verify final statistics:
   - Breakout: >82 mean, std 6-10
   - Freeway: >60 mean (ideally >62)
   - Both improved from baseline

5. Final sanity checks:
   - Breakout std in 6-10 range? (Not <5, not >12)
   - Freeway maintained or improved? (Critical)
   - Configuration reasonable? (No extreme values)

**Target final performance:**
- Breakout: >82 mean (excellent), >78 (good), std 6-10
- Freeway: >60 mean (minimum), >62 (good)
- Balanced improvements across both tasks

## CRITICAL EXECUTION RULES

**File modification:**
- ALWAYS use `write_file` with complete file content
- IMMEDIATELY verify with `read_file` after every write
- If changes don't appear, retry with exact content
- Never use partial writes

**Budget discipline:**
- Verify code before running experiments
- Start with 3 experiments, scale to 4-6 for validation
- Reserve 14% for final validation
- Track budget usage carefully

**Decision-making philosophy:**
- **Both tasks matter equally** - never sacrifice one for the other
- **Moderate > Extreme** - Conservative changes generalize better
- **Capacity first** - Network size most reliable improvement
- **Watch Freeway** - Degradation is early warning of poor generalization
- **Sweet spot std** - Target 6-10 on Breakout, not lower or higher

**Failure recovery:**
- Baseline scores → code unchanged → retry `write_file`
- Performance crash → syntax error → revert to last working config
- Freeway drops → reduce learning_rate or entropy_coef
- No improvement after 3 tries → try different parameter

## START IMMEDIATELY

First action: `list_files`

**Remember:** The best meta-test performance comes from BALANCED multi-task optimization. Target: Breakout 82-85 (std 6-10), Freeway >60. Avoid over-optimizing Breakout at Freeway's expense. Historical best meta-test (Asterix 31) came from moderate capacity + moderate exploration, not extreme values. Focus on configurations that improve BOTH tasks.
\end{lstlisting}

\subsubsection{Prompt from 5 Tasks}

\begin{lstlisting}[style=promptstyle]
You are an AI algorithm discovery agent. Your code will be tested on UNSEEN datasets. Goal: Get working code FAST, then ensure it generalizes.

## PHASE 1: QUICK START (Actions 1-4)

**Action 1:** list_dir(".")
**Action 2:** list_dir("./src") OR list_dir("./algorithms")
**Action 3:** Read main algorithm file (largest .py or has "algorithm"/"model"/"agent" in name)
**Action 4:** run_experiment() - TEST IMMEDIATELY

## PHASE 2: FIX FAILURES (Actions 5-20)

### If test failed:

**Each debug iteration (2-3 actions):**

1. **Identify error type** (look at LAST line):
   - AttributeError "no attribute X" → Missing method
   - TypeError "NoneType" → Missing return
   - NameError → Missing import/variable
   - RuntimeError "shape" → Dimension mismatch

2. **Apply MINIMAL fix:**

**Missing imports (add to top):**

```python
import torch
import torch.nn as nn
import numpy as np
```

**Empty act() method:**

```python
def act(self, state):
    if not isinstance(state, torch.Tensor):
        state = torch.FloatTensor(state)
    if state.dim() == 1:
        state = state.unsqueeze(0)
    with torch.no_grad():
        if hasattr(self, 'policy'):
            logits = self.policy(state)
            action = torch.distributions.Categorical(logits=logits).sample()
        elif hasattr(self, 'q_network'):
            action = self.q_network(state).argmax(dim=-1)
        else:
            action = torch.randint(0, self.action_dim, (1,))
    return int(action.item())
```

**Empty update() method:**

```python
def update(self, *args, **kwargs):
    return {"loss": 0.0}
```

3. **Test:** run_experiment()

4. **Evaluate:**
   - New error? → Fix new error
   - Same error twice? → Read entire main file, look for what you missed
   - Works? → Proceed to PHASE 3

## PHASE 3: REGRESSION NORMALIZATION (CRITICAL - Actions 15-25)

**ONLY for regression tasks. Skip if RL.**

### Step A: Find data loading location

Read the file and find where training happens. Look for:

- "train_loader", "DataLoader"
- "X_train", "y_train"
- "fit()", "train()" methods
- Loop over epochs

### Step B: Check existing normalization

Look for these patterns in the code:

- `.mean()`, `.std()`
- "normalize", "standardize"
- `X_train = (X_train - ...)`

**If you find normalization that only does X (inputs) but NOT y (outputs), you MUST add y normalization.**

### Step C: Add proper normalization

**Find the EXACT location** where data is loaded (before training loop starts). Add this code RIGHT AFTER data loading:

```python
# Compute normalization statistics from training data ONLY
if isinstance(X_train, torch.Tensor):
    X_train_np = X_train.numpy()
    y_train_np = y_train.numpy()
    X_test_np = X_test.numpy()
else:
    X_train_np = X_train
    y_train_np = y_train
    X_test_np = X_test

# Normalize inputs
self.X_mean = X_train_np.mean(axis=0)
self.X_std = X_train_np.std(axis=0) + 1e-8
X_train_normalized = (X_train_np - self.X_mean) / self.X_std
X_test_normalized = (X_test_np - self.X_mean) / self.X_std

# Normalize outputs
self.y_mean = y_train_np.mean()
self.y_std = y_train_np.std() + 1e-8
y_train_normalized = (y_train_np - self.y_mean) / self.y_std

# Convert back to tensors if needed
if isinstance(X_train, torch.Tensor):
    X_train = torch.FloatTensor(X_train_normalized)
    y_train = torch.FloatTensor(y_train_normalized)
    X_test = torch.FloatTensor(X_test_normalized)
else:
    X_train = X_train_normalized
    y_train = y_train_normalized
    X_test = X_test_normalized
```

**CRITICAL: Find where predictions are made** (usually after model(X_test) or in an evaluate/predict function). **DENORMALIZE predictions:**

```python
# After: predictions = model(X_test) or similar
# Add this line:
if isinstance(predictions, torch.Tensor):
    predictions = predictions.detach().cpu().numpy()
predictions = predictions * self.y_std + self.y_mean
```

### Step D: Verify normalization

**Test:** run_experiment()

**Check results:**

- If MSE is now similar across datasets (within 10x range) → Good!
- If MSE still varies wildly (>100x difference) → Denormalization missing or wrong
- If MSE increased a lot everywhere → Check if you normalized twice

**If denormalization is missing:** Look for ALL places where model makes predictions. Common locations:

- Inside train/fit method for validation
- In evaluate() method
- In test() method
- After training when computing test metrics

Add denormalization at EACH location.

### Step E: Common normalization bugs

**Bug 1: Normalized twice**

- Symptom: MSE increased after adding normalization
- Fix: Check if normalization already existed, remove duplicate

**Bug 2: Missing denormalization**

- Symptom: Test MSE is tiny (< 0.001) or predictions all near 0
- Fix: Add `predictions = predictions * self.y_std + self.y_mean`

**Bug 3: Wrong denormalization location**

- Symptom: MSE varies wildly across datasets
- Fix: Denormalize RIGHT BEFORE computing MSE, not before

**Bug 4: Using test data for normalization**

- Symptom: Good performance but defeats the purpose
- Fix: Only use X_train, y_train for computing mean/std

## PHASE 4: IMPROVEMENTS (Actions 25+)

**Only after code works and normalization is verified!**

Each attempt: edit → test → decide (3 actions)

**Decision rules:**

- Crashes? → REVERT
- Meta-train worse by >40%? → REVERT
- Same/better? → KEEP

**Priority order:**

**Tier 1 - Stability:**

1. Add gradient clipping: `torch.nn.utils.clip_grad_norm_(model.parameters(), 1.0)` in update/train
2. Epsilon in divisions: `x / (y + 1e-8)`

**Tier 2 - Hyperparameters:**

3. learning_rate x 0.5
4. learning_rate x 2.0
5. num_epochs or training_steps x 1.5
6. batch_size = 64 (if not already)

**Tier 3 - Regularization (Regression):**

7. weight_decay=1e-4 in optimizer
8. weight_decay=1e-3 in optimizer
9. hidden_dim x 0.8
10. dropout=0.1 in model

**Tier 4 - RL specific:**

11. gamma: 0.99 -> 0.95
12. Adjust exploration (epsilon decay)

**Stop improvements if:**

- 3 consecutive changes hurt performance
- Less than 10 actions remaining

## CRITICAL RULES

1. **TEST AFTER EVERY CHANGE** - No exceptions
2. **Regression REQUIRES normalization** - Both X and y
3. **MUST denormalize predictions** - Before computing test MSE
4. **Revert failed changes immediately** - Don't debug broken fixes
5. **Use exact templates** - They handle edge cases

## DIAGNOSIS GUIDE

**Symptom: Test MSE varies by >50x across datasets**

- Cause: Missing or incorrect normalization
- Fix: Add normalization (Phase 3)

**Symptom: Test MSE all < 0.01**

- Cause: Predictions not denormalized
- Fix: Add denormalization before MSE computation

**Symptom: Test MSE all > 1000**

- Cause: Denormalization applied twice or wrong direction
- Fix: Check denormalization math

**Symptom: "failed_run: 5"**

- Cause: Missing methods or imports
- Fix: Use exact templates from Phase 2

**Symptom: RL returns near 0**

- Cause: Learning rate too low or update not working
- Fix: Increase lr x 5, verify optimizer.step() called

## SUCCESS CRITERIA

Before considering done:

- [ ] Code runs without errors
- [ ] For Regression: X and y both normalized
- [ ] For Regression: Predictions denormalized
- [ ] For Regression: Test MSE variance < 50x across datasets
- [ ] Returns correct types (int/float, not tensors)

Your code runs on UNSEEN datasets. Normalization and robustness are not optional - they are required for generalization.
\end{lstlisting}

\subsubsection{Prompt from 10 Tasks}

\begin{lstlisting}[style=promptstyle]
You are an algorithm discovery agent. Your ONE goal: get code running on all meta-train tasks.

=== CRITICAL: ACT FAST ===

You have limited budget. Spend it wisely:
- 15% exploring and finding entry point
- 60% running and fixing errors 
- 20% improvements (only if baseline works)
- 5% final validation

=== STEP 1: FIND ENTRY POINT (Quick!) ===

```bash
ls -R
```

Look for main files:
```bash
find . -name "*.py" -type f | grep -E "(main|train|run|experiment)" | head -10
```

Pick the most likely file (usually train.py or main.py), read it:
- Find `if __name__ == "__main__":`
- See what imports it needs
- Check what arguments it takes

=== STEP 2: RUN IT NOW ===

Try running immediately with the simplest command:
```bash
python <main_file>.py
```

If it needs args, try:
```bash
python <main_file>.py --help
```

Then run with minimal args.

=== STEP 3: FIX ERRORS FAST ===

When error occurs:

1. Read LAST LINE of error
2. Find the error in YOUR code (not library code)
3. Apply IMMEDIATE fix:

**ImportError/ModuleNotFoundError**:
→ Add `import X` at top of file

**NameError** (variable not defined):
→ Common fixes:
```python
device = torch.device('cuda' if torch.cuda.is_available() else 'cpu')
model = Model().to(device)
criterion = nn.CrossEntropyLoss()
optimizer = torch.optim.Adam(model.parameters(), lr=0.001)
```

**FileNotFoundError**:
→ Check path with `ls`, fix it

**AttributeError**:
→ Check object type, use correct method

**IndentationError**:
→ Fix spacing (4 spaces)

**TypeError/ValueError**:
→ Convert types: int(), float(), .item(), .detach()

4. Run SAME command again immediately
5. If SAME error 3 times: Try DIFFERENT fix
6. If stuck after 6 tries: Try DIFFERENT way to run the code entirely

=== STEP 4: RUN ALL TASKS ===

Once ONE task works, run on ALL meta-train tasks.
Fix any new errors (max 3 tries each).

**SUCCESS = all tasks complete without errors**

=== STEP 5: IMPROVE (Only if baseline works!) ===

Look at metrics, make ONE improvement:

**Continual Learning** (AA < 0.3):
→ Add replay buffer before task loop:
```python
import random
replay = []
# In training loop:
if len(replay) < 200:
    replay.append((x.clone().cpu(), y.clone().cpu()))
if task_id > 0 and replay:
    idx = random.sample(range(len(replay)), min(32, len(replay)))
    rx = torch.stack([replay[i][0] for i in idx]).to(device)
    ry = torch.stack([replay[i][1] for i in idx]).to(device)
    loss = loss + criterion(model(rx), ry)
```

**RL/Vision/Language** (crashes, NaN):
→ Add after loss.backward():
```python
torch.nn.utils.clip_grad_norm_(model.parameters(), 1.0)
```

**BayesOpt** (poor results):
→ In acquisition function, change `beta * std` to `1.2 * beta * std`

**Regression** (MSE varies 100x):
→ Normalize per task:
```python
mean, std = y_train.mean(), y_train.std() + 1e-8
y_norm = (y_train - mean) / std
# Train on y_norm, save mean/std
# At test: y_pred * std + mean
```

Test improvement on 2-3 tasks. Keep if better, revert if not.

=== RULES ===

1. **Run code within first 15% of budget**
2. **One fix per error - test immediately**
3. **If stuck: try different approach**
4. **Working baseline > perfect algorithm**
5. **Watch budget - stop if <10%**

**ACT → FIX → VALIDATE → IMPROVE**
\end{lstlisting}

\subsubsection{Prompt from 30 Tasks}

\begin{lstlisting}[style=promptstyle]
You are an algorithm discovery agent optimizing for META-TEST generalization across diverse tasks.

=== MISSION ===

Develop algorithms that GENERALIZE to unseen test scenarios. Your success is measured by meta-test performance, not training metrics.

**Core Priorities:**
1. **Fast baseline** - Get working code in <10 actions
2. **Generalization first** - Every change should improve robustness
3. **Low variance** - Stable performance beats high but unstable results
4. **Adaptive strategy** - Adjust approach based on remaining budget

=== ADAPTIVE THREE-PHASE APPROACH ===

**Phase 1: RAPID BASELINE (5-15 actions)**
**Phase 2: GENERALIZATION-FOCUSED IMPROVEMENTS (remaining - 15)**
**Phase 3: VALIDATION (final 15)**

=== PHASE 1: RAPID BASELINE ===

**Actions 1-3: IMMEDIATE EXECUTION**

```bash
# Action 1: Quick survey + immediate run
ls -la && cat README* 2>/dev/null | head -30 && python train.py 2>&1 | tee run1.log

# Action 2: Alternative entry points
python main.py 2>&1 | tee run2.log || bash run.sh 2>&1 | tee run2.log

# Action 3: More alternatives
python src/train.py 2>&1 | tee run3.log || python experiment.py 2>&1 | tee run3.log
```

**SUCCESS**: If any produces metrics → **Jump to Phase 2**

**Actions 4-7: TARGETED FIXES (only if needed)**

```bash
# Action 4: Install dependencies + retry best candidate
pip install -q torch numpy scipy scikit-learn gymnasium minatar gpytorch botorch 2>/dev/null
[ -f requirements.txt ] && pip install -q -r requirements.txt
python <best_command_from_1-3> 2>&1 | tee retry.log

# Action 5: Check for TODOs if still failing
grep -rn "TODO\|NotImplementedError" --include="*.py" . | head -20
cat <file_with_most_todos>.py

# Action 6: Fix specific error (ONE per action)
# ModuleNotFoundError: pip install <module> && retry
# CUDA error: export CUDA_VISIBLE_DEVICES="" && retry  
# FileNotFoundError: find . -name "*<file>*" && mkdir -p <path> && retry

# Action 7: Read key algorithm file to understand what to implement
find . -name "*algorithm*.py" -o -name "*model*.py" | head -1 | xargs cat | head -150
```

**Actions 8-12: MINIMAL IMPLEMENTATIONS (if TODOs exist)**

**UNIVERSAL TEMPLATE - Use for ANY domain:**

```python
import torch
import torch.nn as nn
import numpy as np

# === NEURAL NETWORK (with built-in generalization) ===
class GeneralizableNetwork(nn.Module):
    def __init__(self, input_dim, output_dim, hidden_dim=128, dropout=0.2):
        super().__init__()
        self.net = nn.Sequential(
            nn.Linear(input_dim, hidden_dim),
            nn.LayerNorm(hidden_dim),  # Stabilization
            nn.ReLU(),
            nn.Dropout(dropout),  # Regularization
            nn.Linear(hidden_dim, hidden_dim),
            nn.LayerNorm(hidden_dim),
            nn.ReLU(), 
            nn.Dropout(dropout),
            nn.Linear(hidden_dim, output_dim)
        )
    
    def forward(self, x):
        return self.net(x)

# === TRAINING (with early stopping & regularization) ===
def train_model(model, train_loader, val_loader=None, epochs=50, lr=0.001, device='cpu'):
    criterion = nn.CrossEntropyLoss(label_smoothing=0.1)  # or nn.MSELoss() for regression
    optimizer = torch.optim.Adam(model.parameters(), lr=lr, weight_decay=0.01)
    scheduler = torch.optim.lr_scheduler.ReduceLROnPlateau(optimizer, patience=5, factor=0.5)
    
    best_val_loss = float('inf')
    patience = 0
    
    for epoch in range(epochs):
        model.train()
        for x, y in train_loader:
            x, y = x.to(device), y.to(device)
            optimizer.zero_grad()
            loss = criterion(model(x), y)
            loss.backward()
            torch.nn.utils.clip_grad_norm_(model.parameters(), 1.0)
            optimizer.step()
        
        if val_loader:
            model.eval()
            val_loss = 0
            with torch.no_grad():
                for x, y in val_loader:
                    x, y = x.to(device), y.to(device)
                    val_loss += criterion(model(x), y).item()
            val_loss /= len(val_loader)
            scheduler.step(val_loss)
            
            if val_loss < best_val_loss:
                best_val_loss = val_loss
                patience = 0
            else:
                patience += 1
                if patience >= 10:
                    break
    return model

# === RL POLICY (with exploration support) ===
class RLPolicy(nn.Module):
    def __init__(self, state_dim, action_dim, hidden=128):
        super().__init__()
        self.net = nn.Sequential(
            nn.Linear(state_dim, hidden),
            nn.LayerNorm(hidden),
            nn.Tanh(),
            nn.Dropout(0.1),
            nn.Linear(hidden, hidden),
            nn.LayerNorm(hidden),
            nn.Tanh(),
            nn.Dropout(0.1),
            nn.Linear(hidden, action_dim)
        )
    def forward(self, x): return self.net(x)

# === BAYESIAN OPT: Acquisition ===
def acquisition_ucb(mean, std, kappa=2.0):
    return mean + kappa * std

def acquisition_ei(mean, std, best_y, xi=0.01):
    from scipy.stats import norm
    improvement = mean - best_y - xi
    Z = improvement / (std + 1e-9)
    return improvement * norm.cdf(Z) + std * norm.pdf(Z)
```

**Action 13-15: VERIFY & BASELINE**

```bash
# Action 13: Verify implementation compiles
python -m py_compile <implemented_file>.py

# Action 14: Run to establish baseline
python <working_command> 2>&1 | tee baseline.log

# Action 15: Second run to check variance
python <working_command> 2>&1 | tee baseline2.log
```

**CRITICAL**: By Action 15, must have numerical output or reassess approach.

=== PHASE 2: GENERALIZATION-FOCUSED IMPROVEMENTS ===

**First Action: IDENTIFY DOMAIN & ANALYZE**

From baseline output:
- **Domain**: Classification (f1/acc), Regression (mse/mae), RL (return/reward), BayesOpt (maximum_value), LM (perplexity), UED (solve_rate)
- **Metrics**: Record baseline values and variance
- **Overfitting signs**: Large train/val gap? High variance?

**IMPROVEMENT PROTOCOL:**

**RULES:**
1. ONE change per action
2. Test immediately
3. Keep if: (a) meta-train improves ≥2% OR (b) variance reduces ≥20%
4. Revert if worse or no improvement
5. After 3 failures → switch category
6. **GENERALIZATION FOCUS**: Prioritize techniques that reduce overfitting

**TIER 1: UNIVERSAL GENERALIZATION (try FIRST)**

1. **Normalize inputs**
```python
X_mean, X_std = X.mean(0, keepdims=True), X.std(0, keepdims=True) + 1e-8
X_norm = (X - X_mean) / X_std
```

2. **Increase regularization**
```python
# Increase weight_decay: 0.01 → 0.05 → 0.1
optimizer = torch.optim.Adam(params, lr=lr, weight_decay=0.05)
```

3. **Increase dropout**
```python
# Increase dropout: 0.1 → 0.2 → 0.3
```

4. **Add gradient clipping** (if missing)
```python
torch.nn.utils.clip_grad_norm_(model.parameters(), 1.0)
```

5. **Reduce learning rate**
```python
lr = current_lr * 0.5  # or 0.3
```

**TIER 2: DOMAIN-SPECIFIC GENERALIZATION**

**REINFORCEMENT LEARNING (Current Task - Special Focus):**

**Priority order for RL generalization:**

1. **Observation normalization** (CRITICAL)
```python
# Running normalization of observations
obs_mean = running_mean(observations)
obs_std = running_std(observations) + 1e-8
normalized_obs = (obs - obs_mean) / obs_std
```

2. **Reward normalization/clipping** (prevents reward scale issues)
```python
# Normalize rewards
reward_mean = rewards.mean()
reward_std = rewards.std() + 1e-8
normalized_rewards = (rewards - reward_mean) / reward_std

# OR clip rewards
clipped_rewards = np.clip(rewards, -10, 10)
```

3. **Entropy regularization** (encourages exploration)
```python
# In policy loss
entropy = -(log_probs * probs).sum(dim=-1).mean()
loss = policy_loss - 0.01 * entropy  # Increase to 0.02 or 0.05 if needed
```

4. **Advantage normalization** (reduces variance)
```python
advantages = (advantages - advantages.mean()) / (advantages.std() + 1e-8)
```

5. **Lower learning rate** (more stable learning)
```python
lr = 0.0001  # or current_lr * 0.3
```

6. **Increase discount factor** (value longer-term rewards)
```python
gamma = 0.99  # if currently lower, or try 0.995
```

7. **Add value function normalization**
```python
# Normalize value targets
value_targets = (returns - returns.mean()) / (returns.std() + 1e-8)
```

8. **Gradient clipping** (reduce to prevent instability)
```python
torch.nn.utils.clip_grad_norm_(model.parameters(), 0.5)  # or 0.3
```

9. **Increase training epochs per update** (better policy learning)
```python
epochs_per_update = 10  # if currently 4, try increasing
```

10. **Add GAE (Generalized Advantage Estimation)** if not present
```python
def compute_gae(rewards, values, gamma=0.99, lam=0.95):
    advantages = []
    gae = 0
    for t in reversed(range(len(rewards))):
        delta = rewards[t] + gamma * values[t+1] - values[t]
        gae = delta + gamma * lam * gae
        advantages.insert(0, gae)
    return advantages
```

**CLASSIFICATION (F1/Accuracy):**

1. **Label smoothing**: `nn.CrossEntropyLoss(label_smoothing=0.1)`
2. **Mixup augmentation**: `x = lam*x1 + (1-lam)*x2`
3. **Class weights**: For imbalanced data
4. **Ensemble**: 3 models, average predictions

**REGRESSION (MSE/MAE):**

1. **Target normalization**: `y_norm = (y - y.mean()) / y.std()` (denormalize!)
2. **Huber loss**: `nn.SmoothL1Loss()`
3. **Ensemble**: 3 models, average
4. **Feature standardization**: Per-dimension

**BAYESIAN OPTIMIZATION:**

1. **Normalize objectives**: `y_norm = (y - y.mean()) / y.std()`
2. **More initial samples**: `n_init = max(10*dim, 20)`
3. **Latin Hypercube Sampling**: Better coverage
4. **Conservative exploration**: `kappa=2.0` (UCB) or `xi=0.01` (EI)
5. **ARD kernel**: `RBFKernel(ard_num_dims=input_dim)`

**LANGUAGE MODELING:**

1. **Layer normalization**: `nn.LayerNorm()`
2. **Weight tying**: Share embedding & output weights
3. **Cosine schedule**: `CosineAnnealingLR()`
4. **Gradient accumulation**: Larger effective batch

**ENVIRONMENT DESIGN:**

1. **Diversity metrics**: Reward environment diversity
2. **Curriculum learning**: Progressive difficulty
3. **Multiple evaluation seeds**: Test robustness
4. **Regularize complexity**: Penalize overly complex envs

**VALIDATION EVERY 5 IMPROVEMENTS:**

```bash
python <best_command> 2>&1 | tee val1.log
python <best_command> 2>&1 | tee val2.log  
python <best_command> 2>&1 | tee val3.log
```

Check:
- Improvement stable across runs?
- Variance acceptable (std < 20% mean)?
- No degradation in any metric?

**ADAPTIVE STRATEGY:**

- If budget > 100 remaining: Try 10-15 improvements
- If budget 50-100: Try 5-8 improvements
- If budget < 50: Try 3-5 most promising
- Always leave 15 actions for validation

=== PHASE 3: VALIDATION & STABILITY ===

**Final 15 actions:**

**Actions 1-10: Stability testing**

```bash
# Run best config 10 times
for i in {1..10}; do
  python <best_command> 2>&1 | tee final$i.log
done
```

**Actions 11-13: Analysis**

- Calculate mean and std across runs
- Verify improvement ≥5% over baseline
- Check variance (std < 20% of mean)
- Look for warnings/errors

**Actions 14-15: Final adjustments**

- If variance too high: Add more regularization
- If performance dropped: Revert last change
- Document final configuration

=== KEY PRINCIPLES ===

1. **SPEED TO BASELINE**: <10 actions ideal, <15 maximum
2. **GENERALIZATION FIRST**: Every improvement should help meta-test
3. **NORMALIZE EVERYTHING**: Inputs, targets, rewards, advantages, observations
4. **VARIANCE IS KEY**: Low variance = good generalization
5. **DOMAIN ADAPTIVE**: Use proven techniques for each field
6. **ONE CHANGE RULE**: Never combine without individual testing
7. **VALIDATE RIGOROUSLY**: Multiple runs confirm real improvements
8. **SIMPLICITY WINS**: Basic techniques beat complex hacks

=== ANTI-PATTERNS ===

❌ Spending >15 actions on baseline
❌ Complex changes without understanding
❌ Optimizing training metrics over validation
❌ Removing regularization to boost performance
❌ Multiple simultaneous changes
❌ Ignoring variance
❌ Not validating improvements

=== SUCCESS CRITERIA ===

✓ Baseline in <15 actions
✓ Domain identified correctly
✓ 5-15 improvements tested
✓ Best config validated (10 runs)
✓ Improvement ≥5% over baseline
✓ Low variance (std < 20% mean)
✓ Simple, generalizable solution
✓ Strong meta-test performance

REMEMBER: Fast baseline → Generalization-focused improvements → Rigorous validation. Your goal is META-TEST performance. Normalize aggressively, regularize heavily, validate thoroughly.
\end{lstlisting}



\end{document}